%% file: main.tex
\documentclass[a4paper,11pt,fleqn]{book}
\usepackage{subcaption}
\expandafter\def\csname ver@subfig.sty\endcsname{}
\usepackage{url}
\usepackage{enumitem}
\usepackage{multirow}
\usepackage{amsthm}
\usepackage{bbm}
\usepackage{epigraph}
\usepackage{xspace}
\usepackage{multirow}
\usepackage{booktabs}
\usepackage{graphicx}
\usepackage{acronym}
\newacro{HAIM}{Holistic AI in Medicine}
\newacro{MultiModN}{Multimodal Modular Networks}
\newacro{MML}{Multimodal Learning}
\newcommand{\md}{\texttt{MultiModN}\xspace}
\newcommand{\pf}{\texttt{P-Fusion}\xspace}
\usepackage{booktabs}
\usepackage{xr}
\usepackage[dvipsnames]{xcolor}         % colors
\usepackage{graphicx}
\usepackage{float}
\usepackage{caption}
\usepackage{subcaption}
\usepackage{amsmath}
\usepackage{wrapfig}
\usepackage{longtable}
\usepackage{enumitem}
\setlist[enumerate]{nosep, leftmargin=*, label=\textbf{[\arabic*]}, itemsep=2pt}
\usepackage{mdframed}
\newmdenv[
  backgroundcolor=gray!20,
  linecolor=gray!20,
  linewidth=0pt,
  innertopmargin=6pt,
  innerbottommargin=6pt,
  innerleftmargin=6pt,
  innerrightmargin=6pt
]{graybox}
\usepackage{multicol}

\usepackage[breakable, theorems, skins]{tcolorbox}
\tcbset{enhanced}

\DeclareRobustCommand{\mybox}[2][gray!20]{%
\begin{tcolorbox}[   %% Adjust the following parameters at will.
        breakable,
        left=0pt,
        right=0pt,
        top=0pt,
        bottom=0pt,
        colback=#1,
        colframe=#1,
        width=\dimexpr\textwidth\relax, 
        enlarge left by=0mm,
        boxsep=1pt,
        arc=0pt,outer arc=0pt,
        ]
        #2
\end{tcolorbox}
}

\usepackage{makecell}
\usepackage{ragged2e}

\definecolor{darkorange}{HTML}{724209}
\definecolor{lightorange}{HTML}{f7ce9d}
\definecolor{verylightorange}{HTML}{fbe6ce}
\definecolor{darkblue}{RGB}{31,78,120}
\definecolor{lightblue}{RGB}{221,235,247}
\definecolor{puce}{HTML}{513b41}
\definecolor{tifblue}{HTML}{c8f4f9}
\definecolor{lightbluegrey}{RGB}{230,240,255}
\definecolor{darkgrey}{RGB}{50,50,50}
\definecolor{darkgreen}{RGB}{27,64,74}
\definecolor{lightgreen}{RGB}{233,245,211}
\definecolor{brandblue}{HTML}{AFD2E9}
\definecolor{brandyellow}{HTML}{F9DB6D}
\definecolor{brandred}{HTML}{6e4047}

\usepackage[utf8]{inputenc}
\usepackage[T1]{fontenc}
\usepackage{amssymb}
\usepackage{dsfont}
\usepackage{hyphenat}
\usepackage{xspace}
\usepackage{url}
\usepackage{hyperref}

\usepackage{tikz}
\usetikzlibrary{positioning, calc, shapes.geometric, shapes, shapes.multipart, arrows.meta, arrows, decorations.markings, external, trees}
\tikzset{
node distance=0.5cm, % specifies the minimum distance between two nodes. Change if necessary.
}
\tikzstyle{Arrow} = [
	thick, 
	decoration={
		markings,
		mark=at position 1 with {
			\arrow[thick]{latex}
			}
		}, 
	shorten >= 0.5pt, preaction = {decorate},
	]

\input{head/settings_epfl_template.tex}
\input{head/settings_custom.tex} 
\begin{document}
\frontmatter
\input{head/titlepage.tex}

\input{head/dedication}

\setcounter{page}{0}
\input{head/acknowledgements}
\input{head/abstracts}

\tableofcontents
\cleardoublepage
\phantomsection
\addcontentsline{toc}{chapter}{List of figures} % adds an entry to the table of contents
\listoffigures
\cleardoublepage
\phantomsection
\addcontentsline{toc}{chapter}{List of tables} % adds an entry to the table of contents
\listoftables
% your list of symbols here, if needed.

% space before each new paragraph according to the template guidelines.
%(needs to be after titlepage and frontmatter to keep the table of contents lists short)
\setlength{\parskip}{1em}
\newcommand{\il}{\texttt{iLLuMinaTE}\xspace}

%%%%%%%%%%%%%%%%%%%%%%%%%%%%%%%%%%%%%%%%%%%%%%
%%%%% MAIN: The chapters of the thesis
%%%%%%%%%%%%%%%%%%%%%%%%%%%%%%%%%%%%%%%%%%%%%%
\mainmatter
\chapter{Introduction}
\input{main/01_intro}

\chapter{Background}
\input{main/02_background}

\part{Investigations of Explainable AI in Personalized Education}
\chapter{Generalized Student Modeling through Meta Transfer Learning}
\input{main/03_student_modeling}

\chapter{Evaluating the Explainers} 
\input{main/04_evaluating_explainers}

\chapter{Trusting the Explainers} 
\input{main/05_trusting_explainers}

\part{Advancing Human-Centric Explainable AI}
\chapter{Sequential Multimodal Fusion for Interpretability} 
\input{main/06_multimodn}

\chapter{Optimizing the Interpretability vs. Accuracy Tradeoff} 
\input{main/07_interpretcc}

\chapter{Towards Explainer Consistency}
\input{main/09_robustness_of_explainers}

\chapter{From Explanations to Action} 
\input{main/08_illuminate}

\chapter{Discussion and Conclusion}
\input{main/10_discussion}

%%%%%%%%%%%%%%%%%%%%%%%%%%%%%%%%%%%%%%%%%%%%%%
%%%%% TAIL: Bibliography, Appendix, CV
%%%%%%%%%%%%%%%%%%%%%%%%%%%%%%%%%%%%%%%%%%%%%%
\input{tail/appendix}

\backmatter
\input{tail/biblio}
% Add your glossary here
% Add your index here
% Photographic credits (list of pictures&images that have been used with names of the person holding the copyright for them)
\input{tail/cv}

\end{document}

%% file: head/settings_epfl_template.tex
\usepackage[T1]{fontenc}
\usepackage[utf8]{inputenc}
\usepackage[french,german,english]{babel}

\usepackage{fourier} % Utopia font-typesetting including mathematical formula compatible with newer TeX-Distributions (>2010)
\usepackage{setspace} % increase interline spacing slightly

\usepackage{graphicx,xcolor}
\graphicspath{{images/}}

\usepackage{subfig}
\usepackage{booktabs}
\usepackage{lipsum}
\usepackage{microtype}
\usepackage{url}
\usepackage[final]{pdfpages}

\usepackage{fancyhdr}

\fancypagestyle{plain}{
	\fancyhf{}

	\fancyfoot[R]{\thepage}
	}
\fancypagestyle{addpagenumbersforpdfimports}{
	\fancyhead{}
	
	\fancyfoot{}
    \fancyfoot[R]{\thepage}
}

\usepackage{listings}
\usepackage{hyperref}
\hypersetup{pdfborder={0 0 0},
	colorlinks=true,
	linkcolor=black,
	citecolor=black,
	urlcolor=black}
\makeatletter
\def\cleardoublepage{\clearpage\if@twoside \ifodd\c@page\else
    \hbox{}
    \thispagestyle{empty}
    \newpage
    \if@twocolumn\hbox{}\newpage\fi\fi\fi}
\makeatother \clearpage{\pagestyle{plain}\cleardoublepage}

\usepackage{color}
\usepackage{tikz}
\usepackage[explicit]{titlesec}
\newcommand*\chapterlabel{}
\titleformat{\chapter}[display]  % type (section,chapter,etc...) to vary,  shape (eg display-type)
	{\normalfont\bfseries\Huge} % format of the chapter
	{\gdef\chapterlabel{\thechapter\ }}     % the label 
 	{0pt} % separation between label and chapter-title
 	  {\begin{tikzpicture}[remember picture,overlay]
    \node[yshift=-8cm] at (current page.north west)
      {\begin{tikzpicture}[remember picture, overlay]
        \draw[fill=black] (0,0) rectangle(35.5mm,15mm);
        \node[anchor=north east,yshift=-7.2cm,xshift=34mm,minimum height=30mm,inner sep=0mm] at (current page.north west)
        {\parbox[top][30mm][t]{15mm}{\raggedleft $\phantom{\textrm{l}}$\color{white}\chapterlabel}};  %the black l is just to get better base-line alingement
        \node[anchor=north west,yshift=-7.2cm,xshift=37mm,text width=\textwidth,minimum height=30mm,inner sep=0mm] at (current page.north west)
              {\parbox[top][30mm][t]{\textwidth}{\color{black}#1}};
       \end{tikzpicture}
      };
   \end{tikzpicture}
   \gdef\chapterlabel{}
  } % code before the title body

\titlespacing*{\chapter}{0pt}{50pt}{30pt}
\titlespacing*{\section}{0pt}{13.2pt}{*0}  % 13.2pt is line spacing for a text with 11pt font size
\titlespacing*{\subsection}{0pt}{13.2pt}{*0}
\titlespacing*{\subsubsection}{0pt}{13.2pt}{*0}

\newcounter{myparts}
\newcommand*\partlabel{}
\titleformat{\part}[display]  % type (section,chapter,etc...) to vary,  shape (eg display-type)
	{\normalfont\bfseries\Huge} % format of the part
	{\gdef\partlabel{\thepart\ }}     % the label 
 	{0pt} % separation between label and part-title
 	  {\setlength{\unitlength}{20mm}
	  \addtocounter{myparts}{1}
	  \begin{tikzpicture}[remember picture,overlay]
    \node[anchor=north west,xshift=-65mm,yshift=-6.9cm-\value{myparts}*20mm] at (current page.north east) % for unknown reasons: 3mm missing -> 65 instead of 62
      {\begin{tikzpicture}[remember picture, overlay]
        \draw[fill=black] (0,0) rectangle(62mm,20mm);   % -\value{myparts}\unitlength
        \node[anchor=north west,yshift=-6.1cm-\value{myparts}*20mm,xshift=-60.5mm,minimum height=30mm,inner sep=0mm] at (current page.north east)
        {\parbox[top][30mm][t]{55mm}{\raggedright \color{white}Part \partlabel $\phantom{\textrm{l}}$}};  %the phantom l is just to get better base-line alingement
        \node[anchor=north east,yshift=-6.1cm-\value{myparts}*20mm,xshift=-63.5mm,text width=\textwidth,minimum height=30mm,inner sep=0mm] at (current page.north east)
              {\parbox[top][30mm][t]{\textwidth}{\raggedleft \color{black}#1}};
       \end{tikzpicture}
      };
   \end{tikzpicture}
   \gdef\partlabel{}
  } % code before the title body

\usepackage{amsmath}
\makeatletter
\def\resetMathstrut@{%
  \setbox\z@\hbox{%
    \mathchardef\@tempa\mathcode`\(\relax
      \def\@tempb##1"##2##3{\the\textfont"##3\char"}%
      \expandafter\@tempb\meaning\@tempa \relax
  }%
  \ht\Mathstrutbox@1.2\ht\z@ \dp\Mathstrutbox@1.2\dp\z@
}
\makeatother

\usepackage{subfiles}

%% file: head/settings_custom.tex
\usepackage{csquotes} % needed for biblatex
\usepackage[backend=biber,style=alphabetic]{biblatex} % uses APA style

%% file: head/titlepage.tex
\begin{titlepage}
\begin{center}
%\large
\sffamily

\null\vspace{2cm}
{\huge A Human-Centric Approach to Explainable AI \\for Personalized Education} \\[24pt] 

\vfill

\begin{tabular} {cc}
\parbox{0.3\textwidth}{\includegraphics[width=4cm]{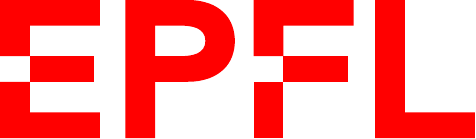}}
&
\parbox{0.7\textwidth}{%
Faculté informatique et communications\\
Laboratoire de science des données\\
Programme doctoral en informatique et communications\\
%
%	ÉCOLE POLYTECHNIQUE FÉDÉRALE DE LAUSANNE\\
	École Polytechnique Fédérale de Lausanne\\[6pt]
	pour l’obtention du grade de Docteur ès Sciences\\
	par\\ [4pt]
	\null \hspace{3em} Vinitra Swamy\\[9pt]
\small
acceptée sur proposition du jury:\\[4pt]
    Prof Patrick Thiran, président du jury\\
    Prof Tanja Käser, directeur de thèse\\
    Prof Martin Jaggi, directeur de thèse\\
    Prof Karl Aberer, rapporteur\\
    Prof Cristina Conati, rapporteur\\
    Prof Ryan Baker, rapporteur\\[12pt]
Lausanne, EPFL, 2025}
\end{tabular}
\end{center}
\vspace{2cm}
\end{titlepage}

%% file: head/dedication.tex
% \cleardoublepage
\thispagestyle{empty}

\vspace*{3cm}

\vspace{4cm}

\begin{center}
    To my loving family -- Swami, Bhuvana, Varsh and Jish.
\end{center}

%% file: head/acknowledgements.tex
\chapter*{Acknowledgements}
~\newline~\newline~
\markboth{Acknowledgements}{Acknowledgements}
\addcontentsline{toc}{chapter}{Acknowledgements}
% put your text here

\bigskip
If you had told college Vini that she would move halfway across the world and start a PhD, she would have thought you were crazy. If you had told her about even 1\% of the adventures, the memories, the mentors, the friends as close as family, the research excitement, and the life she would build here in Switzerland, she would have signed up again in a heartbeat. \\ 

It takes a village to write a thesis. \\

First, to \textbf{Tanja} -- I am extremely lucky to have an advisor that cares so deeply. The years passed so quickly: sparks of research ideas scribbled on the white board in your office, laughing over lunch in the kitchen, chatting with MLBD students, and magically finding which words to cut during late-night zoom calls on deadlines. It's rare to find someone who is simultaneously so brilliant, hard-working, and down-to-earth. From Tanja, I learned how to pay attention to the details, how to conduct strong and rigorous research, how to find the right level of clarity in technical talks, and how to take into account the user perspective at every stage of the process. We often joke about having a TanjaGPT sitting in our heads as we write; I think it's one of my most valuable takeaways from the PhD. It is an honor to be the first graduate of the ML4ED lab, of many, many more to come. \\

\textbf{Martin}, thank you for taking a chance on me. It was wonderful to have you as my advisor -- you were always an unwavering support, willing to look at any idea I was excited about and finding the perfect way to frame it in the surrounding research. I am a huge fan of your positivity and kindness in the way you treat your students and always aim to emulate both. Thank you for always inviting me to join projects, helping me find collaborators, and supporting me to dream big. MLO has been a very welcoming lab, and I have many fond memories of times on the INJ patio.\\

To \textbf{Patrick}, \textbf{Karl}, \textbf{Cristina}, and \textbf{Ryan}, you were the best jury I could ask for. I greatly enjoyed our discussions, and found your feedback extremely valuable. I'm especially honored by the detail to which you read the text and the enthusiasm you had for the topic. It means a lot to have experts I look up to engage so deeply with the work. \\

To my \textbf{ML4ED} family (because they truly are more of a family than a lab): Paola, Jade Mai, and Lucas were my first friends here in Switzerland. \textbf{Paola}, to have both immense technical skills and people skills is no small feat, I continually look up to how you make your friends and collaborators feel so heard. I'm really looking forward to founding a company and team together, and I could not imagine a better person to spend all of my time with. \textbf{Jade Mai}, your thoughtfulness, your drive, your passion, and your empathy are so inspiring. Jade Mai is the kind of person to plan a surprise birthday party because "you always organize so many social events and you might not have had someone plan a surprise for \textit{you} before." \textbf{Lucas}, you have a remarkable ability to bring people together, and the days we all spent at your dining table eating delicious food and discussing everything and anything will remain my fondest lab memories. There is nothing quite as comforting as a tea break eating differential diet cookies in the lab kitchen with you three. \\

\textbf{Bahar}, my first intern turned lab-mate, you have a very bright future ahead of you. Thank you for getting pho with me on rainy days, always being willing to chat about research, and making the best tea. For me, the lab means hearing the latest tech updates from \textbf{Parsa}, dressing as gossip girls with \textbf{Kate}, seeing \textbf{Betül}'s beautifully organized desk in comparison with the controlled chaos on mine, waving hi in the morning to \textbf{Peter} (and stealing a chocolate from the little bowl on the shelf), and discussing the latest cluster updates with \textbf{Hugues}. \textbf{Marta}, \textbf{Fares}, \textbf{Dominik} and \textbf{Nandu}: you four are amazing additions to the lab, and I'm so happy to have a glimpse of the culture you're spreading to the new generation of ML4ED. Fares, thank you for continuing our XAI research to heights far beyond where I could have envisioned. \textbf{Marie}, thank you for making us look forward to going to the lab every day; your sweet gestures make each of us feel happy, from little advent calendars to team photos, fresh oranges to a lab Christmas tree. Right before my private defense, Marie left the cutest pink post-it note near my laptop that means a lot to me. To my favorite postdocs, \textbf{Mirko}, \textbf{Jibril}, \textbf{Tanya}, \textbf{Chris}, \textbf{Syrielle}, \textbf{Thiemo}, and \textbf{Kostas}, thank you for your support. Through working with each of you, I learned so much about growth in my research and growth as a person. Parts of this thesis can be attributed directly to your guidance: Mirko was my very first lab collaborator, Jibril was always willing to debug with me, and Tanya always figured out the perfect way to evaluate our user studies. I look up to you a lot.\\

I've had the privilege to advise a number of extremely talented Master's and Bachelor's students: \textbf{Bahar} (EPFL), \textbf{Natasa} (Imperial), \textbf{Mohammed} (Stanford), \textbf{Julian} (BNS), \textbf{Gabriele} (CERN), \textbf{Su} (ETH), \textbf{Skye} (Oracle), \textbf{Isadora} (UFMG), \textbf{Juliette} (HEP), \textbf{Davide} (EPFL), \textbf{Lan} (EPFL). You'll see many of their names on chapters of this thesis and the work that comes afterwards. I know they have great things in their future, and I'd love to thank them for going above and beyond as researchers and friends; we had so much fun conducting experiments and writing papers that it was always bittersweet when the semester ended.\\

To \textbf{Annie Hartley} and the LiGHT lab, you really deserve the accolade of "making the world a better place" - keep shining, I'm cheering team LiGHT on. \textbf{Keivan}, \textbf{Tatjana}, \textbf{Anastasia}, \textbf{Matteo}, \textbf{Alex}, \textbf{Maksym}, \textbf{Atli}, \textbf{Dongyang}, \textbf{Bettina}, \textbf{Gizem}, \textbf{Oguz}, \textbf{Aditya}, any many more -- thank you for making such a happy and positive environment at (T)MLO. To \textbf{Oana-Maria Camburu} and the UCL AI Center as well as \textbf{Advait Sarkar} and the MSR Cambridge HCI + AI Team, thank you for mentoring me and welcoming me to London. I had an incredible time, and feel even more motivated that user-centric explainable AI is the way forward. \\

Those of you who knew me at EPFL knew that my heart was always with EPIC (the EPFL PhDs of IC). We must have planned close to a hundred events over the years for our IC PhDs, and I enjoyed every single one. Together, we wrote the EPIC guide, revitalized the PhD Slack, launched the EPIC Games, and ran the buddy program, RAMP, coffee chats, board game nights, summer parties, winter breakfasts, lunch speakers, scavenger hunts, lightning talks, and so many more. It meant a lot to have a community on campus, and that's what EPIC was for me. To my wonderful EPIC committee members of past and present -- \textbf{Can}, \textbf{Uday}, \textbf{Federico}, \textbf{Jade Mai}, \textbf{Viktor}, \textbf{Shashwat}, \textbf{Neelu}, \textbf{Hamish}, \textbf{Yugesh}, \textbf{Paola}, \textbf{Laurane}, \textbf{Yist}, \textbf{Cemre}, \textbf{Boya}, \textbf{Sevda}, \textbf{Coco}, \textbf{Shailesh}, \textbf{Charly}, \textbf{Samah}, \textbf{Mikhail}, \textbf{Aditya}, \textbf{Rishubh}, \textbf{Madhur}, \textbf{Elsa}, \textbf{Evann}, and the new gen of EPIC (\textbf{Ayan}, \textbf{Sami}, \textbf{Martin}, \textbf{Parsa}) -- thank you for going all-in on this crazy adventure with me. I think we built something really important and impactful together. To the IC department team (\textbf{Marta}, \textbf{Cecilia}, \textbf{Annalisa}, \textbf{Eileen}, \textbf{Antonella}, \textbf{Deirdre}, \textbf{Tanya}, and \textbf{Tania}) thank you for your constant love and warmth, as well as supporting all of our initiatives. You are honestly incredible people, and give direct credence to the phrase "Swiss efficiency" when it comes to anything organization related. Marta, I hope fast and furious treat you well and that you always have a cup of tea for the next too enthusiastic student that sits in your office with many ideas. To my seniors \textbf{Akhil}, \textbf{Solal}, \textbf{Manoel}, \textbf{Sandra}, and \textbf{Ehsan}, you left very big shoes to fill and I tried my best. \\

To my book club (\textbf{Ankita}, \textbf{Akshar}, \textbf{Anna}, \textbf{Charlie}, \textbf{Can}), thank you for the snacks, the laughs, and the thoughtful (and hilarious) discussions. I am always looking forward to our next book, but I think we all know the books are an excuse for us to spend time together. To my trivia team (the Spanish Inquizition / the Holy Quizzades) -- \textbf{Jess}, \textbf{Solal}, \textbf{Anna}, \textbf{Can}, \textbf{Charlie}, \textbf{Chiara}, \textbf{Nick}, \textbf{Uday}, \textbf{Ethan}, \textbf{Ankita}, \textbf{Shardul}, \textbf{Tatiana}, \textbf{Maya} and many more, thank you for giving me exciting plans on Sunday evenings. There's no one else I would rather decipher pictogram puns or popular songs played backwards with. To \textbf{Akash} and \textbf{Saqib}, thanks for going to Golden India with me; I always think we make the perfect order. To my friends from back home that I was able to see over the last few years (\textbf{Shanaaz}, \textbf{Ashlee}, \textbf{Sathvik}, \textbf{Vasilis}, \textbf{Evy}, \textbf{Emma}, \textbf{Diana}, \textbf{Rupali}, \textbf{Mrina}, \textbf{Ana-Maria} and many more) thanks for believing in me, visiting me, and keeping me company across the globe. To my dear flatmates \textbf{Amelia}, \textbf{Lea}, and \textbf{Molly}, it was so nice to spend a few years together doing yoga by the lake and watching the sunset on the balcony. I'd also like to gratefully acknowledge my favorite cafes in Lausanne (Cafe St. Pierre in Bessiéres, Tibits at Lausanne Gare, and Osio in Malley) for being lovely places where much of this thesis was written over many years. \\

To \textbf{Angelika}, my sister in all but name. Being your friend made Lausanne feel like home to me, and our trips to Greece were experiences I will treasure for life. The view in your car, driving into Athens at night, singing out loud to the radio and watching the lights sparkle over the city in the distance made anything feel possible. That, and hot sauce on Souvlaki. To \textbf{Anna}, I'm so glad DB thursdays brought us together. I find myself often admiring how you carry yourself with such wit, strength, and self-knowledge. You both are amazing travel buddies, my absolute favorite brunch company, and extremely talented researchers. \\

To \textbf{Federico}, you're the type of person that would go to the moon and back for his friends, and I consider myself very lucky to be counted among them. Thank you for creating the party playlist, going way too deep into too many games (especially universal paperclips), planning logistics with me, and always bringing the vibes. I also have to thank dear Giulietta (Federico's Alpha Romeo) for the many road trips she's taken us on. To \textbf{Uday} and \textbf{Adi}, thank you for curing my homesickness with biryani, card games, and TV shows. I'm sorry for the alarm clock pranks (but they were very funny!). You have always been a great support to me, and I love laughing about the most random incidents with you both. To \textbf{Can}, thank you for helping me re-discover my love of reading, for sharing my love of teaching, for traveling with me, and for showing me all the little moments of happiness in daily life.\\

Most importantly, to my \textbf{family} -- Appa, Amma, Varsh and Jish, this PhD would not have happened without your love and support. From Appa, I learned how to have real enthusiasm for a topic, and how to express those thoughts in a way that can be relatable and exciting to an audience. If I have any talent in oration or teaching, it comes directly from my Dad. From Amma, I learned how to deeply care for the people around me, to see what needs to be done and do it, and how a hug and a cup of chai can go a long way. From Varsh, I learned the true meaning of hard work (and the benefits of colorful pens). Every time I'm at the lab and it's late at night, I take pride in knowing she must be working even harder than I am at the hospital and is cheering me on from across the ocean. And from Jish, I learned how to enjoy my years here -- how listening to nice music can make anything I'm doing fly by, and how to let go of work to take in the beauty of my surroundings. This PhD is lovingly dedicated to the four of you.\\

\noindent\textit{Lausanne, March 18, 2025}
\hfill Vinitra Swamy.

%% file: head/abstracts.tex
%\begingroup
%\let\cleardoublepage\clearpage

% English abstract
% \cleardoublepage
\chapter*{Abstract}
~\newline~
%\markboth{Abstract}{Abstract}
\addcontentsline{toc}{chapter}{Abstract (English/Français)}

% % German abstract
% \begin{otherlanguage}{french}
% \cleardoublepage
% \chapter*{Zusammenfassung}
% ~\newline~\newline~
% %\markboth{Zusammenfassung}{Zusammenfassung}
% % put your text here
% \lipsum[1-2]
% \vskip0.5cm
% Stichwörter: 
% %put your text here
% \end{otherlanguage}

\begin{raggedleft}
Science is what we understand well enough to explain to a computer. \\
Art is everything else we do. \\
— Donald Knuth \\
\end{raggedleft}
\vspace{2em}
Deep neural networks form the backbone of artificial intelligence (AI) research, with potential to transform the human experience in areas ranging from autonomous driving to personal assistants, healthcare to education. However, their integration into the daily routines of real-world classrooms remains limited. It is not yet common for a teacher to assign students individualized homework targeting their specific weaknesses, provide students with instant feedback, or simulate student responses to a new exam question. While these models excel in predictive performance, this lack of adoption can be attributed to a significant weakness: the lack of explainability of model decisions, leading to a lack of trust from students, parents, and teachers. \\

This thesis aims to bring human needs to the forefront of eXplainable AI (XAI) research, grounded in the concrete use case of personalized learning and teaching. We frame the contributions along two verticals: technical advances in XAI and their aligned human studies.\\

We begin with a generalizable approach to student modeling evaluated at scale across 26 online courses with over 100,000 students and millions of student interactions. To enable personalized learning interventions, we evaluate five state-of-the-art explainability methods, finding systematic disagreement between explainers when they are evaluated for the same students and the same models. We then turn to expert educators for ground truth evaluation; they find strong actionability value in explanations, but disagree regarding which explanations are trustworthy.\\

This thesis therefore presents a shift away from popular approximation-based explainability methods towards model architectures that are inherently interpretable. We propose four complementary technical contributions to enhance interpretability: \\
\begin{itemize}[leftmargin=*]
    \item \textbf{MultiModN}, an interpretable, modular, multimodal model which offers accurate predictions and explanations even with missing data, at the fraction of the number of parameters.
    \item \textbf{InterpretCC}, an interpretable mixture-of-experts model that uses adaptive sparsity to produce concise explanations without sacrificing performance.
    \item \textbf{An exploration of adversarial training}, to improve the consistency and stability of post-hoc explainers in educational settings.
    \item \textbf{iLLuMinaTE}, an LLM-XAI pipeline that generates user-friendly, actionable, zero-shot explanations as a communication layer for XAI.
\end{itemize}

This thesis also places a strong emphasis on measuring the human perception, actionability, and real-world usefulness of new technical contributions. We conducted four human-AI user studies:
\begin{itemize}[leftmargin=*]
    \item 26 university professors participate in semi-structured interviews to \textbf{validate post-hoc explanations for course design} across LIME, SHAP, and confounder explanations.
    \item 56 teachers evaluate hybrid (visual and text) explanations in the first known \textbf{perception study of intrinsically interpretable models}.
    \item 20 expert learning scientists participate in semi-structured interviews to \textbf{measure perceptions of inconsistency in explanations}.
    \item 114 university students evaluate the \textbf{actionability of mid-semester feedback explanations}.
\end{itemize}

By combining empirical evaluations of existing explainers with novel architectural designs and human studies, our work lays a foundation for human-centric AI systems that balance state-of-the-art performance with built-in transparency and trust.

\vskip0.5cm
Key words: Explainable AI, Human-Centric AI, Educational Data Mining.
% French abstract
\begin{otherlanguage}{french}
% \cleardoublepage
\chapter*{Résumé}
~\newline~\newline~

Les réseaux neuronaux profonds forment l'épine dorsale de la recherche en intelligence artificielle (IA) appliquée, avec le potentiel de transformer l’expérience humaine dans des domaines allant de la conduite autonome aux assistants personnels en passant par la santé et l’éducation. Cependant, leur intégration dans les routines quotidiennes des salles de classe reste limitée. Il n’est pas encore courant pour un enseignant d’assigner aux étudiants des devoirs individualisés ciblant leurs lacunes spécifiques ou de fournir des retours instantanés et détaillés. Bien que les modèles d’IA excellent en performance prédictive, ce manque d’adoption peut être attribué à une faiblesse importante : le manque d’explicabilité des décisions des modèles, conduisant à un manque de confiance de la part des étudiants, des parents et des enseignants. \\

Cette thèse vise à mettre les besoins humains au centre de la recherche sur l’IA eXplainable  (IAX), en s’appuyant sur le cas concret de l’apprentissage et de l’enseignement personnalisés. Nous encadrons les contributions selon deux axes : les avancées techniques en IAX et les études humaines qui s'y rapportent. \\

Nous commençons par une approche généralisable de la modélisation des étudiants, évaluée à grande échelle sur 26 cours en ligne avec plus de 100 000 étudiants et des millions d’interactions. Pour permettre des interventions d’apprentissage personnalisées, nous évaluons cinq méthodes d’explicabilité à l'état de l'art. Nos recherches attirent l’attention sur le problème du désaccord systématique entre les explicateurs lorsqu’ils sont évalués sur les mêmes étudiants et les mêmes modèles. Nous nous tournons ensuite vers des éducateurs experts pour une évaluation fondée; ils trouvent une forte valeur d’actionnabilité dans les explications des étudiants, mais ne s’accordent pas entre eux sur la sélection des explications les plus dignes de confiance. \\

Cette thèse présente donc un changement de paradigme en s’éloignant des méthodes d’explicabilité populaires basées sur des approximations vers des architectures de modèles intrinsèquement interprétables. Nous proposons quatre contributions techniques complémentaires pour améliorer l’interprétabilité : \\

\begin{itemize}[leftmargin=*]
    \item \textbf{MultiModN}, un modèle modulaire, multimodal et interprétable qui offre des prédictions et des explications précises même en cas de données manquantes, avec une fraction du nombre de paramètres.
    \item \textbf{InterpretCC}, un modèle interprétable de type mixture-of-experts qui utilise une parcimonie (\textit{sparsity}) adaptative pour produire des explications concises sans compromettre les performances.
    \item \textbf{Une exploration de l’apprentissage adversarial}, pour améliorer la cohérence et la stabilité des explicateurs post-hoc dans des contextes éducatifs.
    \item \textbf{iLLuMinaTE}, un pipeline LLM-XAI qui génère des explications conviviales, actionnables, et en zero-shot comme une couche de communication pour la XAI.
\end{itemize}

Cette thèse met également un fort accent sur la mesure de la perception humaine, de l’actionnabilité et de l’utilité réelle des nouvelles contributions techniques. Nous avons donc mené quatre études utilisateurs IA-humains :

\begin{itemize}[leftmargin=*]
    \item 26 entretiens semi-structurés avec des éducateurs universitaires pour \textbf{valider des explications pour la conception de cours} à travers LIME, SHAP, et une explication de confondement.
    \item 56 enseignants ont évalué des \textbf{explications hybrides (visuelles et textuelles)} issues de méthodes intrinsèquement interprétables dans la première étude connue impliquant des modèles interprétables par conception.
    \item 20 entretiens semi-structurés avec des experts en sciences de l’apprentissage pour \textbf{décrire à quel point l’incohérence dans les explications est un problème}.
    \item 114 étudiants universitaires ont évalué l’actionnabilité des explications de rétroaction de mi-semestre avec \textbf{iLLuMinaTE}.
\end{itemize}

En combinant des évaluations empiriques d'explicateurs existants avec de nouvelles conceptions architecturales et des études humaines, notre travail pose les bases de systèmes d’IA humains qui équilibrent des performances de pointe avec une transparence et une confiance intégrées.

\vskip0.5cm

Mots clés : Intelligence Artificielle Explicable, Intelligence Artificielle Centrée sur l’Humain, Fouille de Données Éducatives.
\end{otherlanguage}

%\endgroup			
%\vfill

%% file: main/01_intro.tex
\section{A "Golden Age of AI", Tempered by Trust Concerns}

Artificial Intelligence (AI) has become one of the transformative innovations of our modern era, with substantial advances driven by the availability of large-scale compute \cite{gruetzemacher2022transformative}. However, AI adoption in human-centric settings is not yet widespread, as it presents unique challenges: it is not enough for an AI system to excel in predictive accuracy alone — the system must also be \textit{trusted} by its stakeholders (e.g. educators, students, patients, doctors) \cite{webb2021machine, conati2018ai}. In this context, we define ``human-centric" as any application in which a human directly uses AI predictions to make informed decisions \cite{national2021human}.

In education, where learners are often under-aged and equitable learning outcomes are paramount, trust plays a pivotal role \cite{leichtmann2023effects}. Decisions made by AI-driven systems can influence learners’ academic trajectories and long-term futures, making transparency and interpretability essential \cite{conati2018ai}. Teachers need to understand the recommendations provided by an AI model before they can use them to adapt instruction \cite{nazaretsky2022teachers}. Students and parents must feel confident that the system’s predictions are fair, accurate, and actionable \cite{ai-trust}. Across these domains, a pressing question arises: \textit{How can we design AI systems whose decision-making processes are transparent, understandable, and actionable by the students and teachers who rely on them?}

\section{Explainable AI and Its Limitations}

Explainable AI (XAI) has emerged as a prominent area of research aimed at addressing these challenges \cite{molnar2020interpretable}. By providing explanations for model decisions, XAI methods seek to bridge the gap between highly accurate ``black-box" models and the transparency required for user trust and adoption \cite{shneiderman2020human}. To enable interpretability, practitioners tend to use either traditional ML models that are intrinsically interpretable (e.g. decision trees, SVM) \cite{jovanovic2016building,vultureanu2021improving} or apply a single post-hoc explainer \cite{adadi2018peeking,dovsilovic2018explainable}. Unfortunately, recent research shows that post-hoc explanations might be unfaithful to the true model \cite{rudin2019stop}, inconsistent \cite{slack2020fooling,agarwal2022openxai}, or method-dependent \cite{Swamyexplainers2022,krishna2022disagreement,brughmans2023disagreement}. Furthermore, evaluating the quality of the provided explanations is a challenge, since there is often no ground truth \cite{swamy2023trusting,dai2022fairness}. Substantial limitations of these methods (discussed further in Chapters 4 and 5 of this thesis) can be summarized as follows:

\begin{enumerate}[leftmargin=*]
\item \textbf{Potential unfaithfulness}: The nature of post-hoc explanations as approximations of black-box model internals can lead to innaccuracies in depicting the true decision-making process of the model, even if the explanations appear plausible \cite{rudin2019stop, agarwal2022openxai, yadav2022xaudit}.
\item \textbf{High computational cost}: Popular explanation techniques are computationally intensive, rendering them impractical for real-time applications in interactive educational systems \cite{xu2017real}.
\item \textbf{Inconsistency across methods and instances}: Post-hoc explainers often produce contradictory explanations for the same model prediction, leading to confusion and mistrust among users \cite{slack2020fooling, Swamyexplainers2022, swamy2023trusting, krishna2022disagreement, brughmans2023disagreement}.
\item \textbf{Lack of actionability}: Feature-based explanations often fail to provide stakeholders with concrete guidance on how to respond to the system’s outputs \cite{joshi2019towards}.
\item \textbf{Limited comprehensibility}: Explanations that rely on complex numerical or visual representations may be difficult for non-technical users, such as teachers and students, to interpret \cite{swamy2023trusting, swamy2024explanationsactionzeroshottheorydriven, dai2022fairness}.
\end{enumerate}

Despite these limitations, post-hoc explainability methods like LIME \cite{lime} and SHAP \cite{shap} remain widely deployed in real-world applications \cite{hasib-lime, creditrisk, shap-ex}, often with minimal scrutiny of their faithfulness or usability. This uncritical adoption poses significant risks, particularly in high-stakes settings where erroneous explanations can erode trust and have serious real-world consequences.

\section{Toward Human-Centric Explainable AI Solutions}

To address the shortcomings of existing XAI methods, there is a growing need for interpretability solutions that are not only accurate and computationally efficient but also \textbf{human-friendly}. We position several of the contributions of this thesis in the emerging research area of intrinsically interpretable neural networks \cite{agarwal2021neural,molnar2020interpretable}. In light of the specific challenges in human-centric domains \cite{national2021human}, we define five design targets that explanations should fulfill, motivated originally in  \cite{swamy2023future} and aligned with \cite{pinto2024towards, rudin2018stop, conati2018ai}:

\begin{enumerate}[leftmargin=*]
    \item \textbf{Accurate explanations with certainty}: Explanations should accurately represent the true internal reasoning of the model, or be accompanied by a score of confidence \cite{marx2023but,leichtmann2023effects}. This trait is also referred to as having high explanation \textit{fidelity} or \textit{faithfulness} \cite{yeh2019fidelity,dasgupta2022framework,jain2020learning}.
    \item\textbf{Real-Time}: Explanations should be provided in real-time or with minimal delay to support timely decision-making for interactive settings (in the scale of seconds, not tens of minutes) \cite{xu2017real}.
    \item \textbf{Consistent}: Explanations should remain stable for similar instances or contexts, ensuring that users can form reliable expectations of model outputs. In a time series of interactive predictions, the explanations should not drastically differ \cite{li2021algorithmic}.
    \item \textbf{Actionable}: Explanations should provide actionable insights, empowering model users to take appropriate actions or make informed interventions based on the explanation \cite{joshi2019towards}.
    \item \textbf{Human understandable}: The format and language of explanations should be accessible to non-technical users like educators, students, and patients  \cite{hudon2021explainable,haque2023explainable, zytek2024llms}.
\end{enumerate}

This thesis advocates for a \textbf{human-centric} approach to explainable AI that addresses these design targets, with a specific focus on personalized education. In the following chapters, we show how explainability can be made \textbf{accurate} (guaranteed alignment with the model decision making process), \textbf{real-time} (providing explanations instantaneously alongside predictions), \textbf{consistent} (the same model given the same input will produce the same explanation), \textbf{actionable} (students and teachers can derive useful actions from explanations), and \textbf{human understandable} (tailored for the understanding of students, teachers, and learning scientists). In the educational domain, we showcase possible use cases of AI systems that support instructors in identifying at-risk students, personalizing learning material recommendations, and tailoring feedback to individual learners. We additionally demonstrate that the methodological insights presented in this thesis show promise beyond education (for example in medical settings, weather forecasting, or sentiment classification  \cite{swamy2024multimodn,swamy2024interpretcc}) and are relevant to any field where interpretability, transparency, and trustworthiness are essential.

\section{Thesis Outline and Contributions}

\textbf{Part I: Introduction and Background.} The introductory chapters provide an overview of the current landscape of AI in education. Key concepts in interpretable and explainable machine learning are introduced, with a focus on their implications for user trust and adoption.

\textbf{Part II: Investigations of Explainable AI in Personalized Education.} In this section, we begin by exploring generalizable models of student behavior for settings where teachers are not present: online courses. We introduce a generalized model architecture that can accurately predict student performance early across diverse MOOC courses. However, these larger, generalized models have limited interpretability on an individual student scale. To enable interpretations of student learning behavior, we investigate the merits of 5 different post-hoc explainers across diverse educational course models, finding systematic disagreement. We then turn to expert educators for ground truth evaluation, designing a study that looks at explanations across pairs of courses that differ in one learning dimension. Educators find the explanations helpful and actionable, but disagree with each other about which explanations are most trustworthy. The main contributions of each chapter are described below.

Chapter 3 introduces a generalizable approach for early student performance prediction, evaluated across 26 online courses with over 100,000 students. The work aims to find whether training student behavior models accounting for diverse context and course structures could effectively address the "warm start" problem for ongoing or first-time courses that do not yet have end-of-course student performance signals. The key contributions include three novel strategies for transfer learning in education: pre-training on a large set of courses, integrating meta-information about course characteristics statically or at each time-step, and a comparison to fine-tuning on prior course iterations. We provide the trained state-of-the-art models for student performance prediction for public use. This study shows that interaction data combined with course-level information yields predictive performance comparable to models trained on historical course data, addressing a critical gap in "warm-start" predictive modeling for new and ongoing courses.

\mybox{
\textbf{Generalized Student Modeling through Meta Transfer Learning (Chapter 3)}\\\
\textit{Originally published at Learning@Scale 2022 ~\cite{swamy2022meta}.}
}

Chapter 4 presents a comprehensive evaluation of five  explainability methods (LIME, PermutationSHAP, KernelSHAP, DiCE, and CEM) frequently used in the context of student performance prediction. The chapter’s key contributions include a quantitative comparison of feature importance rankings across explainers using Principal Component Analysis, Jensen-Shannon distance, and Spearman’s rank-order correlation. The findings reveal significant and systematic disagreement between explainers across five methods, in five different course settings. We additionally analyze the five explainers' ability to capture prerequisite skill relationships across weeks of a course curriculum, finding relevant patterns but overall disagreement. This work was one of the first to systematically evaluate explainable AI methods, raising a crucial awareness among the educational ML community of the bias caused by explainer selection in high-stakes contexts.

\mybox{
\textbf{Evaluating the Explainers (Chapter 4)}\\\
\textit{Originally published at EDM 2022 ~\cite{swamy-eval}.}
}

Chapter 5 investigates the trustworthiness of explainers through a mixed-methods study involving both quantitative analyses and qualitative insights from 26 university-level educators. The chapter’s contributions include a pairwise course comparison design to evaluate the quality of feature-based explanations across five pairs of courses that were chosen for a significant learning science difference (e.g. same topic and student population offered in a flipped vs. MOOC setting). For each course setting, we asked study participants which features they believe contribute most to student success, which explanations they trust most, and how they could transform these insights into actionable course design decisions. Our results show that quantitatively, explainers significantly disagree with each other about what is important, and qualitatively, experts themselves do not agree on which explanations are most trustworthy. However, there are some positive notes: over 75\% of the educators were able to find an actionable decision for their next course's design based on an ensemble of explainers, and differences between explanations for course pairs were able to capture trends from learning science literature. 

\mybox{
\textbf{Trusting the Explainers (Chapter 5)}\ \\
\textit{Originally published at LAK 2023 with a Honorable Mention Paper Award \cite{swamy2023trusting}.}
}

\textbf{Part III: Advancing Human-Centric Explainable AI.} In the previous chapters, we examined a concrete use case of XAI with student performance prediction models towards designing personalized learning interventions for struggling students in online courses. We found that post-hoc explainers have several problems that make them unsuitable to a high impact human-centric domain like education, including 1) the disagreement between explainability methods, 2) the amount of time it takes to extract explanations for a course model (oftentimes 10 minutes per student, of which there are hundreds or thousands in a course), and 3) the amount of postprocessing required to make them understandable for educators. Therefore, the following chapters in Part III propose interpretability solutions designed to address limitations in explainers for human-centric domains, focusing on improving their faithfulness, time-to-explanation, consistency, actionability, and human understandability. Our technical contributions extend the state-of-the-art neural network architectures, from modular networks for multimodality (Chapter 6), to mixture-of-expert models (Chapter 7), as well as adversarial training approaches (Chapter 8) and a LLM-XAI pipeline (Chapter 9). We also conduct several user studies, including a novel teacher evaluation of interpretable-by-design neural network architectures (Chapter 7), educator views on inconsistency from XAI in comparison with human TAs (Chapter 8), and student perceptions of social science theories of explanation used as an XAI communication layer (Chapter 9). We describe the contributions of each chapter below.

Chapter 6 introduces MultiModN, a novel multimodal neural network that sequentially fuses latent representations from multiple data modalities (e.g., text, images, audio) in an inherently interpretable and modular fashion. MultiModN directly addresses the explainability needs of \textbf{accuracy}, \textbf{real-time interpretation}, and \textbf{consistency} by design. By integrating sequential fusion, the model aligns its predictions with individual modality contributions, ensuring that explanations accurately reflect input data. MultiModN supports real-time predictions by mitigating computational bottlenecks, and its design inherently handles missing-not-at-random (MNAR) data, making explanations robust even in the presence of missing or biased inputs. This approach demonstrates that interpretability and predictive performance can be achieved simultaneously, at a fraction of the number of parameters. This architecture is particularly important in domains such as education and healthcare where data incompleteness is common.

\mybox{
\textbf{Sequential Multimodal Fusion for Interpretability (Chapter 6)}\ \\
\textit{Originally published at NeurIPS 2023 \cite{swamy2024multimodn}}.
}

Chapter 7 presents InterpretCC, an interpretable-by-design mixture-of-experts model architecture that addresses the trade-off between interpretability, explanation faithfulness, and model accuracy. InterpretCC addresses the needs for \textbf{accuracy}, \textbf{real-time interpretation}, \textbf{consistency} and \textbf{human-understandability} by employing a conditional computation framework that selectively activates feature subsets relevant to specific tasks. This ensures that explanations remain concise, focusing on the most informative inputs. The architecture supports both global mixture-of-experts (MoE) and sparse feature gating, which make explanations comprehensible by clearly delineating the features contributing to each prediction. A user study with 56 teachers demonstrates that InterpretCC enables stakeholders to interpret outputs in a way that informs actionable interventions, especially in sensitive applications like student risk prediction.

\mybox{
\textbf{Optimizing the Interpretability vs. Accuracy Tradeoff (Chapter 7)}\\\
\textit{Originally published at ICLR 2025 \cite{swamy2024interpretcc}}.
}

Chapter 8 investigates the consistency of educational AI explanations across multiple runs, explainers, and course settings, including diverse MOOC and flipped classroom courses. This chapter addresses the need for explainer \textbf{consistency} by examining the impact of random initialization on model outputs, the effect this has on the perception of educator trust in classroom AI through semi-structured interviews with 20 expert educators, and exploring adversarial training techniques (e.g., FGSM and PGD) to stabilize decision boundaries. We found that inconsistency of explanations erodes trust in AI teaching assistants, even if educators are willing to accept the same level of inconsistency from human teaching assistants. The key contributions include demonstrating that while adversarial robustness can improve the consistency of counterfactual explanations, it does not generalize to explainers like LIME and SHAP, which remain susceptible to random perturbations. This highlights that explanation consistency must be approached holistically, considering both the model and the explainer. This chapter emphasizes the importance of stable, reproducible explanations to foster trust among educators and ensure reliable decision-making in AI-powered learning systems.

\mybox{
\textbf{Towards Explainer Consistency (Chapter 8)}\\\
\textit{In Submission to Computers and Education: AI.}
}

Chapter 9 introduces iLLuMinaTE, an LLM-XAI pipeline that generates theory-driven, zero-shot explanations for student performance predictions. iLLuMinaTE addresses the needs for \textbf{human-understandability} and \textbf{actionability} of explanation by transforming complex XAI outputs into user-friendly explanations aligned with cognitive theories of explanation. The iLLuMinaTE pipeline's multi-stage prompt design incorporates social science frameworks, such as contrastive explanation and robustness selection, to generate explanations that align with user expectations. An evaluation of over 21,915 natural language explanations generated from three LLMs (GPT-4o, Gemma2-9B, Llama3-70B) and a user study involving 114 university students show that iLLuMinaTE explanations are preferred $89.2\%$ of the time for their clarity and actionable insights. Through this chapter, we demonstrate that LLMs can effectively bridge the gap between technical XAI methods and the practical needs of educators through social science theories, making AI outputs more understandable and useful for non-technical stakeholders.

\mybox{
\textbf{From Explanations to Action (Chapter 9)}\\\
\textit{Originally published at AAAI 2025 \cite{swamy2024explanationsactionzeroshottheorydriven}.}
}

\textbf{Part IV: Conclusion and Future Directions.} The final chapters summarize the key contributions of this thesis and outline open research questions. The thesis concludes with a call to action for the development of explainability methods that balance state-of-the-art performance with human-centered interpretability.

\vspace{2em}

Overall, the guiding research question this thesis seeks to address is: \textit{How can we design AI systems that are both powerful and trustworthy in real-world educational contexts?} By combining empirical evidence with methodological innovation and human studies, the work presented here contributes to the creation of explainable AI solutions that align with the needs of human users in education and beyond.

\section{Publication list}

The following publications are associated with this thesis.

\begin{enumerate}

\item Swamy, V., Marras, M., \& Käser, T. (2022). \textit{Meta Transfer Learning for Early Success Prediction in MOOCs}. In Proceedings of the 9th ACM conference on Learning @ Scale.

\item Swamy, V., Radmehr, B., Krco, N., Marras, M., \& Käser, T. (2022). \textit{Evaluating the Explainers: Black-Box Explainable Machine Learning for Student Success Prediction in MOOCs.} In Proceedings of the 15th International Conference on Educational Data Mining.

\item Swamy, V., Du, S., Marras, M., \& Käser, T. (2023). \textit{Trusting the Explainers: Teacher Validation of Explainable Artificial Intelligence for Course Design.} In LAK23: 13th International Learning Analytics and Knowledge Conference. \textbf{Honorable Mention Paper Award.}

\item Swamy, V., Frej, J. A., \& Käser, T. (2023). \textit{The future of human-centric eXplainable Artificial Intelligence (XAI) is not post-hoc explanations.} Under review at JAIR.

\item Swamy, V.*, Satayeva, M.*, Frej, J., Bossy, T., Vogels, T., Jaggi, M., Käser, T.* \& Hartley, M. A.* (2023). \textit{MultiModN - Multimodal, Multi-task, Interpretable Modular Networks.} Advances in Neural Information Processing Systems (NeurIPS), 36.

\item Swamy, V., Montariol, S., Blackwell, J., Frej, J. A., Jaggi, M., \& Käser, T. (2024).\textit{Intrinsic user-centric interpretability through global mixture of experts.} In the Proceedings of the International Conference on Learning Representations (ICLR) 2025.

\item Swamy, V., Tran, L., D'Angeli, G., Parchet, J., Cock, J. M., Frej, J. A., Nazaretsky, T. N. \& Käser, T. (2025). \textit{Teacher Perceptions of Inconsistent eXplainable AI: Exploring Consistency of Student Explanations through Adversarial Training.} Under submission to Computers and Education: AI.

\item Swamy, V.*, Romano, D.*, Desikan, B., Camburu, O. M., \& Käser, T. (2025). \textit{iLLuMinaTE: An LLM-XAI Framework Leveraging Social Science Explanation Theories towards Actionable Student Performance Feedback.} In the Proceedings of the AAAI Conference on Artificial Intelligence (Vol. 39).

\end{enumerate}

In addition to the work included in this thesis, the following additional research papers on trustworthy ML were presented during my doctorate:

\begin{enumerate}

\item Swamy, V., Romanou, A., \& Jaggi, M. (2021). \textit{Interpreting Language Models Through Knowledge Graph Extraction.} In the 35th Conference on Neural Information Processing Systems (NeurIPS 2021), XAI for Debugging Workshop.

\item Wambsganss, T.*, Swamy, V.*, Rietsche, R., \& Käser, T. (2022). \textit{Bias at a Second Glance: A Deep Dive into Bias for German Educational Peer-Review Data Modeling.} In Proceedings of the 29th International Conference on Computational Linguistics (COLING).

\item Asadi, M., Swamy, V., Frej, J., Vignoud, J., Marras, M., \& Käser, T. (2023). \textit{RIPPLE: Concept-based interpretation for raw time series models in education.} In the Proceedings of the AAAI Conference on Artificial Intelligence, EAAI Symposium.

\item Wambsganss, T.*, Su, X.*, Swamy, V., Neshaei, S., Rietsche, R., \& Käser, T. (2023). \textit{Unraveling Downstream Gender Bias from Large Language Models: A Study on AI Educational Writing Assistance.} In Findings of the Association for Computational Linguistics: EMNLP 2023.

\item Chen, Z., Cano, A. H.*, Romanou, A.*, Bonnet, A.*, Matoba, K., Salvi, F., ... Swamy, V., ... Hartley, M. A., Jaggi, M., \& Bosselut, A. (2023). \textit{Meditron-70b: Scaling medical pretraining for large language models.} arXiv preprint arXiv:2311.16079. Under review at Nature Communications.

\item Nazaretsky, T., Mejia-Domenzain, P., Swamy, V., Frej, J., \& Käser, T. (2024).\textit{ AI or human? Evaluating student feedback perceptions in higher education.} In European Conference on Technology Enhanced Learning. Cham: Springer Nature Switzerland. \textbf{Best Paper Award.}

\item Gado, E. G., Martorella, T., Zunino, L., Mejia-Domenzain, P., Swamy, V., Frej, J., \& Käser, T. (2024). \textit{Student Answer Forecasting: Transformer-Driven Answer Choice Prediction for Language Learning. }In 17th International Conference on Educational Data Mining.

\item Salles, I., Mejia-Domenzain, P.*, Swamy, V.*, Blackwell, J., \& Käser, T. (2024). \textit{Interpret3C: Interpretable Student Clustering Through Individualized Feature Selection. }In International Conference on Artificial Intelligence in Education. \textbf{Best LBR Paper Award.}

\item Nazaretsky, T., Mejia-Domenzain, P., Swamy, V., Frej, J., \& Käser, T. (2025). \textit{The Critical Role of Trust in Adopting AI-Powered Educational
Technology for Learning: An Instrument for Measuring Student Perceptions.} In Computers and Education: Artificial Intelligence.

\end{enumerate}

%% file: main/02_background.tex
% [4-5 pages]

This background section presents the research landscape  necessary for contextualizing the contributions of this thesis in three parts: ML for Education, Explainable AI, and the intersection of both topics in Explainable AI for Education. Certain aspects of this background section were highlighted originally in our viewpoint paper \textit{"The future of human-centric eXplainable Artificial Intelligence (XAI) is not post-hoc explanations"} \cite{swamy2023future}, as well as inspired from the other papers written in this thesis \cite{swamy2022meta, swamy2022evaluating, swamy2024explanationsactionzeroshottheorydriven}.

\section{Overview of ML for Online Learning}

AI has seen widespread application in education research, encompassing both learner-centric models — such as intelligent tutoring systems \cite{mousavinasab2021intelligent},  knowledge tracing \cite{piech2015deep}, and automated feedback systems \cite{jacobsen2023promises} — and teacher-centric models, including real-time classroom insights \cite{holstein2018classroom} and automated question generation \cite{hang2024mcqgen}.

As we move into an increasingly digital age, Massive Open Online Courses (MOOCs) and digital learning environments have become immensely popular \cite{kruchinin2019investigation, ByTheNum39}. Enrollments have surged to over $220$ million MOOC learners, with several courses on Coursera and EdX reporting increases by a factor of $10$ due to the COVID pandemic \cite{impey2021moocs, ByTheNum39}. In 2021 alone, $40$ million new learners signed up for a MOOC, with providers hosting $3{,}100$ new courses and over $500$ new programs \cite{ByTheNum39}. However, MOOCs tend to suffer from high dropout and low success rates \cite{aldowah2020factors,MOOCcomp77}, often with completion rates as low as 10\%. Furthermore, prior research has shown that students often do not possess the self-regulated learning skills necessary to succeed in flipped classrooms (a form of blended learning) \cite{hirsto2019exploring}. 

In the scope of this thesis, we focus on the context of MOOCs and flipped classroom courses as settings where AI can be helpful when teachers are minimally present. Adaptive guidance can hence increase retention and student performance in online course settings. Consequently, a large body of work has focused on extracting behavioral features based on students' interaction data. For example, \cite{chen2020utilizing, lemay2020grade} extracted count-based features (e.g. number of online sessions, number of videos watched). Others extracted features related to the course attendance rate or the ratio of videos watched \cite{he18utilize, mbo20early} and fine-grained video behavior (either through explicit features or through feeding raw video events into an LSTM \cite{mu21deeplearning}). \cite{boroujeni2016quantify} has showed that regular working patterns are essential for success in MOOCs. Only few feature extraction works have addressed \textit{early} success prediction for ongoing courses. For example, \cite{marras2021can} performed a meta-analysis of features across early prediction models on flipped courses and MOOCs. \cite{mbo20early} showed that video engagement in the first week and mid-point of the course can be indicative of student pass-fail grades. \cite{mao2019one} used recent temporal patterns in novice programming tasks to early intervene.

Utilizing these features (and many others), online learning scenarios can benefit from models for modeling student behavior in an \textit{accurate} and \textit{interpretable} fashion. Several existing approaches use traditional ML methods like logistic regression \cite{whitehill2017mooc}, support vector machines \cite{tomkins2016predicting}, and random forests \cite{sweeney2016next} for student success prediction. More recent educational prediction models focus on deep learning approaches for educational tasks, for example long short-term memory (LSTM) networks \cite{wang2017deep}, transformers \cite{liu2023mooc}, or reinforcement learning architectures \cite{zhang2019hierarchical}. Methods in student success prediction (e.g., \cite{imran2019predicting}), estimating early dropout (e.g., \cite{xing2019dropout}), and knowledge tracing (e.g., \cite{piech2015deep, abdelrahman2019knowledge}) rely on neural networks to reduce human effort in the modeling pipeline and boost overall prediction accuracy. 

Notably, most of these studies have performed a-posteriori or post-hoc analyses, and not models for early success prediction. These advances also come at a significant cost: traditional machine learning techniques (e.g., linear regression, SVMs, decision trees) are simple, but interpretable, where deep learning techniques trade transparency for the ability to capture complex data representations \cite{molnar2022}. 

\section{Overview of Explainable AI}

\begin{figure*}[t]
  \centering
  \includegraphics[width=\textwidth]{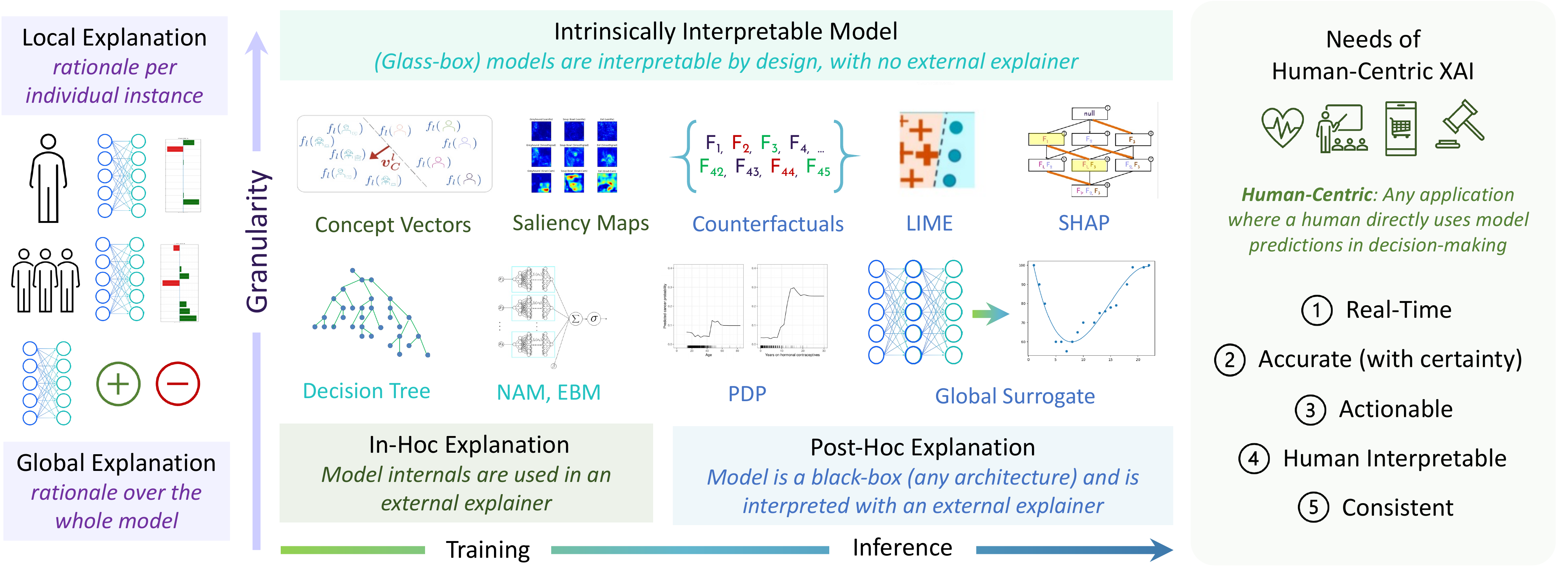}
  \caption{Explainability can be post-hoc (e.g., LIME, SHAP), in-hoc (e.g., saliency methods), or intrinsic (by design). Furthermore, the granularity of model explanations ranges from local (single user, a group of users) to global (entire sample).}
  \label{fig:pipeline}
\end{figure*}

The vase majority of modern machine learning and AI approaches use neural networks due to their ability to represent complex relationships from diverse data inputs \cite{li2021survey, lin2022survey}. However, neural networks are considered as black-box models due to their high number of included parameters and their subsequent complexity of operations \cite{molnar2020interpretable}; in the form of a matrix of weights, humans cannot interpret how neural network decisions are made under the hood. The goal of eXplainable AI (XAI) is to allow for model transparency by either producing interpretations for black-box model decisions or making the model's decision-making process transparent. As illustrated in Figure \ref{fig:pipeline}, model explanations range from local to global granularity. \textit{Global} interpretability allows users to understand how the entire model operates by examining its structure and parameters, while \textit{local} interpretability focuses on understanding the rationale behind a individual prediction \cite{molnar2020interpretable}. Moreover, explainability can be integrated into the modeling pipeline at three stages: 
\begin{enumerate}[leftmargin=*]
    \item \textbf{Post-hoc explainability}: after the decision is made, an explainer (e.g. LIME, SHAP) is fit on top of the black-box model to interpret the results.
    \item \textbf{In-hoc explainability}: an explainer interprets the model weights at inference time or customizes the training protocol for additional information; e.g., Grad-CAM uses backpropagation to highlight important regions of an input image \cite{selvaraju2017grad}.
    \item \textbf{Intrinsic explainability}: models that explicitly define the decision pathway as part of the model, like traditional ML models (e.g., decision trees) or gated model architectures (e.g. concept bottleneck models) \cite{koh2020concept}. In these cases, the model itself is interpretable in its prediction process.
\end{enumerate}

Research and adoption of neural network explainability in human-centric areas has surged over the last eight years. Post-hoc approaches are most commonly favored, as there is no impact on model accuracy and no additional effort required during training. Local, instance-specific post-hoc techniques such as LIME \cite{lime}, SHAP \cite{shap}, or counterfactuals \cite{dice}, have been effectively utilized for tasks like predicting ICU mortality \cite{katuwal2016machine}, non-invasive ventilation for ALS patients \cite{ferreira2021predictive}, credit risk \cite{creditrisk}, or loan repayment \cite{pawelczyk2020learning}.

These approaches, while popular, are accompanied by weaknesses in real-world settings. The computational time is often in the tens of minutes; not \textbf{real-time} enough for users, students, or patients to make a decision based on the explanation alongside a prediction \cite{molnar2020interpretable}. In most cases, there is \textbf{no measurement of confidence} in a generated post-hoc explanation \textcolor{black}{\cite{laugel2019dangers}}. The \textbf{actionability} and \textbf{human-interpretability} of the explanation are solely based on the input format \textcolor{black}{\cite{zytek2024llms}}. As human-centric tasks often use tabular or time series data, the subsequent explanations are often not concise, actionable or interpretable easily beyond the scope of a data scientist's knowledge \textcolor{black}{\cite{karran2022designing}}. Recent research on explanation user design has shown that humans across healthcare, law, finance, education, and e-commerce, among others, prefer hybrid text and visual explanations \textcolor{black}{\cite{haque2023explainable}}, a format not easily provided by current post-hoc libraries. Lastly, the \textbf{consistency} of the explanations is not inherently measured. Several explainability methods could produce vastly different explanations with different random seeds or at different time steps \textcolor{black}{\cite{slack2020fooling}}.

% Explanations are difficult to evaluate
Furthermore, post-hoc explanations are difficult to evaluate. Current metrics (e.g. saliency, faithfulness) aim to quantify the quality of an explanation in comparison to expert-generated ground truth \cite{agarwal2022openxai}. However, accurate explanations need to be true to the model internals, not human perceptions. In this light, the most trustworthy metrics measure the prediction gap (e.g. PIU, PGU), removing features that are considered important by the explanation and seeing how the prediction changes \cite{dai2022fairness}. This approach is still time-consuming and imperfect, as it fails to account for cross-feature dependencies. Recent literature \cite{krishna2022disagreement,brughmans2023disagreement}, as well as Chapter 4 \cite{Swamyexplainers2022}, has examined the results of over $50$ explainability methods with diverse datasets ranging from criminal justice to healthcare to education through a variety of metrics (rank agreement, Jenson-Shannon distance) and demonstrated strong, systematic disagreement across methods. Validating explanations through human experts can also be difficult: explanations are subjective, and most can be justified. \cite{krishna2022disagreement} and \cite{dhurandhar2018explanations}, as well as Chapter 5 \cite{swamy2023trusting} and Chapter 8 have conducted user studies to examine trust in post-hoc explainers, measuring data scientist and human expert preference of explanations. Results indicate that while humans generally find explanations helpful, no method is recognized as most trustworthy.
% \subsection{In Interpretable AI models}

In contrast to post-hoc approaches, in-hoc methods like layer relevance propagation \cite{lu2020towards} or concept-activation vectors \cite{cav} have shown success in student success prediction \cite{asadi2022Limeripple} or identifying skin conditions \cite{lucieri2020interpretability}. These approaches are directly aligned with model internals, making their explanations accurate to the model decisions. However, these approaches require specific model architectures or access to model weights, which makes them less popular in the age of large, closed models \cite{achiam2023gpt}. 

Recent work on intrinsically explainable models has largely focused on example-based approaches, with most applications centered on the image modality (\textit{e.g.}, B-cos networks \cite{bohle2022b}, PIP-Net \cite{nauta2023pip}, ProtoPNet \cite{chen2019looks}), while other modalities such as time-series, tabular, or text have been explored far less frequently \cite{sawada2022concept}. Methods like NAM \cite{agarwal2021neural} and EBM \cite{nori2019interpretml} aim for interpretability by training a model for each input feature or feature combination and aggregating their scores for predictions. However, this approach quickly becomes computationally expensive as the feature space grows. 

Mixture-of-experts models take a different approach, employing hierarchical structures where subnetworks combine interpretable experts (e.g., decision trees) with neural networks to handle partially interpretable outputs \cite{ismail2023interpretable}, activate specific experts selectively \cite{li2022branch}, or route inputs based on automatically extracted concepts \cite{you2023sum,alvarez2018towards}. These methods are expressive, but often produce explanations that are overly detailed and complex, limiting their understandability and actionability for end-users.

A detailed positioning of interpretable-by-design models, including those presented in Chapter 7, can be found in Section \ref{sec:icc-background}. \cite{rudin2019stop} has made a strong argument for moving away from black-box models and using inherent interpretability (i.e. traditional ML models) for impactful decisions. In this thesis, we propose several strategies to make \textit{deep learning} approaches interpretable-by-design, guaranteeing transparency, robustness, and trustworthiness \cite{swamy2023future} without having to revert to traditional ML models. 

% Student Modeling for MOOCs
\section{Explainable AI for Education}

There is a compelling need for interpretability in educational models. Studies emphasize that trust in AI systems significantly influences their perceived usefulness, which in turn drives engagement and adoption \cite{foroughi2024determinants,celik2023exploring}. Transparency plays a dual role: it not only demystifies the decision-making processes of AI systems but also enables users to identify and address ethical concerns \cite{chitti2020need}.  \cite{conati2018ai} argues for the necessity of interpretable models in settings where students can see the effect of a decision but not the reasoning behind it (e.g., Open Learner Models). Furthermore, \cite{DBLP:conf/LAK/NazaretskyCA22} has found that transparency is an essential factor in increasing educators' trust in AI-based educational technology.

Consequently, there has been emerging literature on XAI in education which can be categorized into three main motivations: (1) allowing educational stakeholders to audit model mistakes \cite{khosravi2022explainable,pinto2024towards}, (2) building student and teacher trust in AI \cite{nazaretsky2024ai}, and (3) designing personalized interventions for students \cite{hur2022using, asadi2023ripple}. For example, \cite{lu2020towards} examined the inner workings of deep learning models for knowledge tracing through layer-relevance propagation, with goals of auditing and understanding the model. Other researchers \cite{hasib2022lime} experimented with traditional machine learning models for student success prediction and implemented local explanations with LIME for transparency in the best performing model. Additionally, \cite{baranyi2020interpretable} used SHAP feature importances to interpret student dropout prediction models, while \cite{wang2024academic} used XG-Boost and SHAP for interpretable student achievement prediction towards personalized interventions. \cite{mu2020towards} suggested interventions for wheel-spinning students based on Shapley values. Finally, \cite{vultureanu2021improving} explored LIME on ensemble machine learning methods for student performance prediction, \cite{scheers2021interactive} integrated LIME explanations in student advising dashboards, and \cite{pei2021} used LIME for interpreting models identifying at-risk students, building trust for students and teachers. 

There has been much less work on educational models that are interpretable-by-design. These papers are usually focused on tree-based approaches, statistical reasoning, or traditional ML, which do not have the advantages of more complex architectures \cite{vultureanu2021improving,pei2021,dsilva2023trustworthy, kim2023variational}. \cite{pinto2024intrinsically} proposes a theoretical approach to place constraints on a Convolutional Neural Network towards identification of gaming behavior. \cite{hooshyar2024problems} compares post-hoc approaches like SHAP and LIME with interpretable neuro-symbolic rule extraction. These recent works towards interpretable-by-design models for education have not yet dealt with complexities like multimodality or more modern deep learning architectures (transformers, mixture-of-experts), and have not yet assessed the human understandability or actionability contributions of their proposed methods.

\begin{graybox} In summary, this thesis seeks to address the presented research gap in explainable AI for education: 1) an evaluation across multiple XAI approaches, 2) measuring student and teacher perceptions of XAI, and 3) developing intrinsically interpretable models for modern deep learning architectures. \end{graybox}

%% file: main/03_student_modeling.tex
This work was originally published in collaboration with Professor Mirko Marras and Professor Tanja Käser in the article \textit{"Meta Transfer Learning for Early Success Prediction in MOOCs"}, in Proceedings of the 9th ACM conference on Learning @ Scale (2022). We thank Professor Martin Jaggi for helpful discussions towards this model architecture.

\section{Introduction}
The vast majority of performance prediction literature \cite{whitehill2017mooc, tomkins2016predicting, sweeney2016next, wang2017deep, marras2021can, mbo20early, mao2019one} has focused on inference on the course the model is trained on, i.e. on training the models on one portion of the students and then predicting on the remaining students of the course. In case of first time smaller courses, training on the same course might lead to overfitting. For \textit{early} success prediction, using the same course for training and inference is not practical as the model does not have the data required for supervised training (the pass-fail label will only be known at the end of the course). Thus, for these settings, success predictions need to rely on models trained on other courses. 

 \textit{Transfer learning} is a well-known machine learning paradigm, which refers to training a model on one setting, and using that model to warm-start predictions in a different context \cite{weiss2016survey}. In the context of education, only a few works have attempted to create generalizable models that can be transferred between different settings. \cite{ding2019transfer} extract latent feature representations that are transferable over multiple MOOC courses. \cite{wei2017convolution} and \cite{karumbaiah2021using} warm-start their model on past data in two different settings: MOOC forum post classification and active learning tasks. \cite{tsiakmaki2020transfer} transferred predictions across 5 undergraduate courses at the same university. Most relevantly, \cite{whitehill2017delving} showed that learning generalized patterns of student behavior from multiple MOOCs can improve predictive performance.
 
In this chapter, we aim to tackle the problem of transferability across MOOCs from different domains and topics, focusing on models for early success prediction. We present and analyze three novel strategies to creating generalizable models for this task. First, we train a BiLSTM model taking students' interaction data as input on a large collection of diverse MOOCs. Our second strategy is based on the assumption that students' interaction behavior is influenced by the context and hence courses with similar characteristics will have similar student interaction patterns. We suggest two architectures for including course information into a model, one being an extension of our previous BiLSTM model and the second one combining interaction data and course information using attention layers. We frame success prediction as a \textit{meta learning} problem (see e.g. \cite{hospedales2020meta}) through viewing broad domains of MOOCs as separate tasks i.e. predicting student success in a Mathematics course is often very different from predicting student success in an English course. Our third strategy extends the aforementioned models by applying fine-tuning on previous course iterations.

% Figure moved here for layout purposes
\begin{figure*}[htbp]
  \includegraphics[width=\textwidth, trim=1 1 1 1,clip]{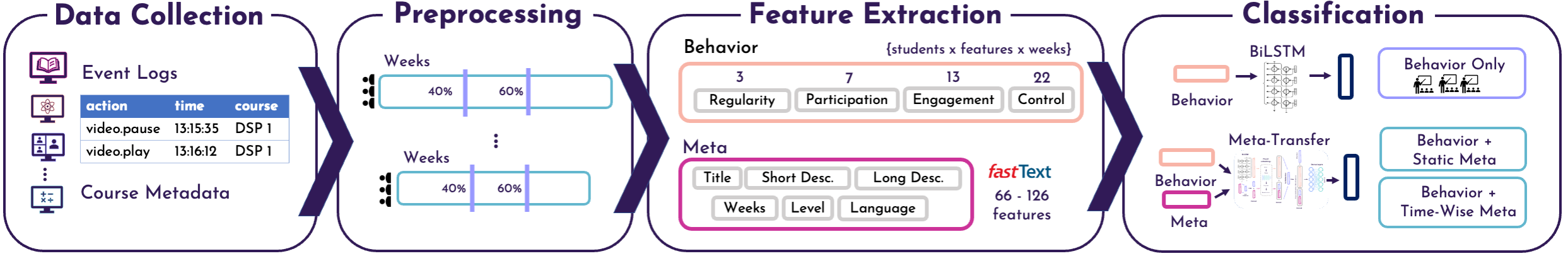}
  \caption{Our transfer approach based on behavior and course metadata features, from logs collection to success prediction.}
  \label{fig:features}
\end{figure*}

We extensively evaluate our models on a large data set including $26$ MOOCs and $145{,}714$ students in total. With our analyses, we address the following three research questions: 
\begin{enumerate}
    \item Can student behavior transfer across iterations of the same course and across different courses \textbf{(RQ1)}?
    \item Is a meta learning model trained on a combination of behavior and course metadata information more transferable \textbf{(RQ2)}?
    \item Can fine-tuning a combined model on past iterations of an unseen course lead to better transferable models \textbf{(RQ1)}?
\end{enumerate}  Our results demonstrate that a model statically combining behavior and meta information over multiple courses can significantly improve over one-course transfer baselines ($11.45\%$ on average across multiple early prediction levels and courses). Our implementation and pre-trained models\footnote{\texttt{https://github.com/epfl-ml4ed/meta-transfer-learning}} achieve at least a comparable and often better performance to training a customized model based on previous iterations of their course.

\section{Methodology}
The problem we address in this chapter can be formulated as a time-series classification task that relies on students' interaction data as well as meta information about the course to predict \textit{early} on whether a student will pass or fail a course. Specifically, we are interested in improving the accuracy of success predictors for students belonging to a previously unseen iteration of a course or to a completely new course. For clarity, we formalize the addressed problem before presenting the proposed approach.

% Provide a mathematical formulation of the problem
\subsection{Problem Formulation}
We consider a set of \emph{students} $S^c \subset \mathbb{S}$, who are enrolled in a \emph{course} $c$ (referred also as course iteration), which is part of the online educational offering $\mathbb{C}$. As courses can be run multiple times over the year, each course $c$ refers to one run of a \emph{course set} $C$. We denote a sequence of iterations for a course set $C = \{c_1, \ldots, c_{M^C}\}$, with $M^C$ being the total number of iterations for the course set $C  \in \mathbb{C}$. In other words, we consider a course set $C$ to be a set composed of subsequent course iterations (e.g., multiple runs of a linear algebra course). 
Each course $c \in C$ has a predefined \emph{schedule} consisting of $N_c = |\mathbb{O}^c|$ learning objects from a catalog $\mathbb{O}^c$. In our study, we assume that the latter can be either videos or quizzes, but the notation can be easily extend to other types (e.g., forum posts, textbook readings). 
Enrolled students interact with the learning objects in the schedule, generating a time-wise clickstream. We denote the \emph{interactions} of a student $s \in \mathbb{S}^c$ in a course $c$ as a time series $I_s^c = \{i_1, \ldots, i_K\}$ (e.g., a sequence of video plays and pauses, quiz submissions), with $K$ being the total number of interactions of student $s$ in course $c$. 
We leave these definitions very general, allowing the length of each time series to differ, since our models can accommodate this. We assume that each interaction $i$ is represented by a tuple $(t, a, o)$, including a \emph{timestamp} $t$, an \emph{action} $a$ (videos: load, play, pause, stop, seek, speed; quiz: submit), and a \emph{learning object} $o \in \mathbb{O}^c$ (video, quiz). 
We finally denote the \emph{binary success labels} (pass/fail) for students $S^c$ in a course $c$ as $\mathbb{L}^c = \{l_{s_1}, \ldots, l_{s_{|S^c|}}\}$.

% How does our pipeline look like?
\subsection{Transfer Framework for Success Prediction}
As shown in Fig. \ref{fig:features}, our approach to solving the transfer problem in early success prediction\footnote{Our approach and resulting models can be easily leveraged for other downstream educational machine learning tasks, e.g., grade prediction and dropout prediction.} consists of three steps: 1) collecting and preprocessing the log data, 2) computing behavior features for each student (based on log data) and meta features per course, and 3) building three different deep learning models based on different combinations of behavior and meta features.
\subsubsection{Data Collection and Preprocessing}
\label{sec:log-preproc}

%Student interactions tends to be sparse in MOOCs. 
A large portion of students drop out of a course during the first few weeks (e.g., there are students who just enroll in the course for watching a few videos) \cite{moocdropout, moocretention}. Therefore, it is possible to predict course success for these early-dropout students with a high accuracy by simply looking at their assignment (or homework/quiz) submissions and grades in the first course weeks. Developing complex machine learning models for predicting on these early-dropout students is inefficient. For that reason, when not stated otherwise, we refer to $S^c$ as the students in a course $c$, whose course pass-fail label cannot be easily predicted with a shallow machine learning model, trained on the assignments grades in the first few weeks of the course. While we filter out so called early-dropout students, we will later show (Sec. \ref{sec:results-metatransfer}) that our predictors have high accuracy on these students as well. Our setting consequently leads to a faster and more effective optimization process. To identify the early-dropout students $s \in S^c$ of each course $c$, we fit a \textit{Logistic Regression} model on the assignment grades of the first course weeks. Specifically, our input data for the model is a matrix $G$ of shape $|S^c| \times w$, where $w$ is the number of course weeks and the value $G(i, j)$ is the grade student $s_i$ received on the assignment in course week $j$. In our case, $w = 2$. We considered the average score of graded assignments for that week in case of multiple graded assignments per week scored non-attempted assignments with a $0$. We filter out the students whose predicted probability of course failure $\hat{p}_{s}>0.99$. We found the optimal threshold via a grid search over $\{0.96, 0.97, \ldots, 0.999\}$, maximizing the model's \textit{balanced} accuracy.

Given the low success rates for many MOOCs, targeted interventions have the potential to improve learning outcomes \cite{borrella2021taking, xing2019dropout, whitehill2015beyond, perez2021can}. We are therefore interested in \textit{early} success prediction, providing the basis for such targeted intervention (e.g., offering additional support to students at risk of failing the course). We define an early prediction level $e$ and consider students' interaction data only up to that point in time \footnote{Our experiments will show early predictions for $40\%$ and $60\%$ of the course duration.}. If a course had a duration of $10$ weeks and we aimed to predict student success after $e = 40\%$ of the course duration, we would only consider student interactions happening in the first four course weeks. We denote the interactions of a student $s \in \mathbb{S}^c$ in a course $c$ up to a specific point in time $e$ with $I_s^{c,e}$.

% Step 2 of the pipeline: feature extraction
\subsubsection{Feature Extraction}\label{sec:features-descr}
For each course iteration $c \in \mathbb{C}$, we extract a set of behavior features for each student $s$ based on the student's interaction data $I_s^{c,e}$ for an early prediction level $e$. We also extract a set of meta information about that course iteration $c$.

\vspace{1mm} \noindent \textbf{Behavior Features}. To obtain a comprehensive representation of student behavior, we combine multiple types of features according to student interactions with videos and quizzes. More granularly, we consider the four behavior feature sets identified as the ones with the highest predictive power for success prediction in the context of MOOCs \cite{marras2021can}\footnote{Given the size and variety of the course data considered in our study, we included all features of the four features sets, instead of considering only the specific features identified as important in at least one course by \cite{marras2021can}. We then let our classifiers detect the patterns and hence relevant features for the considered courses.}.  
Formally, given the interactions $I_s^{c,e}$ generated by students $S^c$ until a course week $w \in \mathbb{N}$ in course $c$, we produce a matrix $H \subset \mathbb{R}^{|S^c| \times w \times h}$ (i.e, each feature in the feature set is computed per student per week), where $h \in \mathbb{N}$ is the dimensionality of the feature set. As in Sec. \ref{sec:log-preproc}, $w$ is chosen based on the early prediction level $e$. We focus on the following behavioral aspects\footnote{These 45 features are explicitly detailed in Appendix \ref{sec:features}.}:

\begin{itemize}[leftmargin=*,nolistsep]
    \item \textbf{Regularity} \cite{boroujeni2016quantify} ($H_1$, shape: $|S^c| \times w \times 3$). These features quantify the regularity of a student's study habits, for instance by measuring whether the student usually studies during the same time of a day or has a preference for specific days of the week.
    \item \textbf{Engagement} \cite{chen2020utilizing} ($H_2$, shape: $|S^c| \times w \times 13$). These features monitor the engagement of students throughout the course, considering indicators such as the total number of student clicks on weekends and on weekdays, and the total number of sessions.
    \item \textbf{Control} \cite{lalle2020data} ($H_3$, shape: $|S^c| \times w \times 22$). This feature set measures the fine-grained video consumption per student, including features such as the proportion of videos watched, re-watched, or interrupted, and the standard deviation of these ratios.
    \item \textbf{Participation} \cite{marras2021can} ($H_4$, shape: $|S^c| \times w \times 7$). These features are related to attendance on videos and quizzes based on the schedule, such as the number of scheduled videos watched for that week and the number of quizzes passed on the first try.
\end{itemize}

\vspace{2mm} Given a course $c \in \mathbb{C}$, we extract the above features for all students $S^c \subset \mathbb{S}$ and concatenate them to obtain the final combined behavior features for students in that course, defined as $H^c \in \mathbb{R}^{|S^c| \times w \times 45}$, with $H^c = [H_1 \cdot H_2 \cdot H_3 \cdot H_4]$ (the $\cdot$ denotes a concatenation). Due to the different scales, we perform a min-max normalization per feature in $H^c$ (i.e., we scale the feature between 0 and 1 considering all students and weeks for that feature).

\vspace{1mm} \noindent \textbf{Meta Features}. Our application scenario includes a large variety of courses with different characteristics (e.g., topic, structure, duration). We assume that the observed interaction patterns of a course do not only depend on the population (i.e. the individual students attending the course), but also on the specific characteristics of the course. In prior work \cite{park2020meta, sun2019meta}, we see that passing in meta information representing differences in context improves predictive performance of transfer. We therefore code relevant information about each course. Formally, given a course $c \in \mathbb{C}$, we produce a fixed-length representation $F \subset \mathbb{R}^{f}$ for each meta feature, where $f \in \mathbb{N}$ is the size of the meta feature. Specifically, we consider:

\begin{itemize}[leftmargin=*,nolistsep]
    \item \textbf{Duration} ($F_1$, shape: $f=1$) represents the number of course weeks as per the schedule provided by the instructor.  
    \item \textbf{Level} ($F_2$, shape: $f=3$) is a one-hot encoded vector representing the level of the course. Our course data includes three different levels: Bachelor, Master, and Propedeutic.
    \item \textbf{Language} ($F_3$, shape: $f=2$) is a one-hot encoded vector representing the language in which the course is taught. Courses included in this study, were taught either in French and English. 
    \item \textbf{Title} ($F_4$, shape: $f=30$) is a word embedding vector of the title the instructor assigned to the course. 
    \item \textbf{Short Description} ($F_5$, shape: $f \in [30,60]$) is a word embedding vector of the short description of the course (one sentence about the main topic).
    \item \textbf{Long Description} ($F_6$, shape:  $f \in [30,60]$) is a word embedding vector of the long description of the course ($2{-}5$ more detailed sentences about the course content). 
\end{itemize}

% Figure moved here for layout purposes
\begin{figure*}[t]
  \includegraphics[width=\textwidth, trim=4 4 4 4,clip]{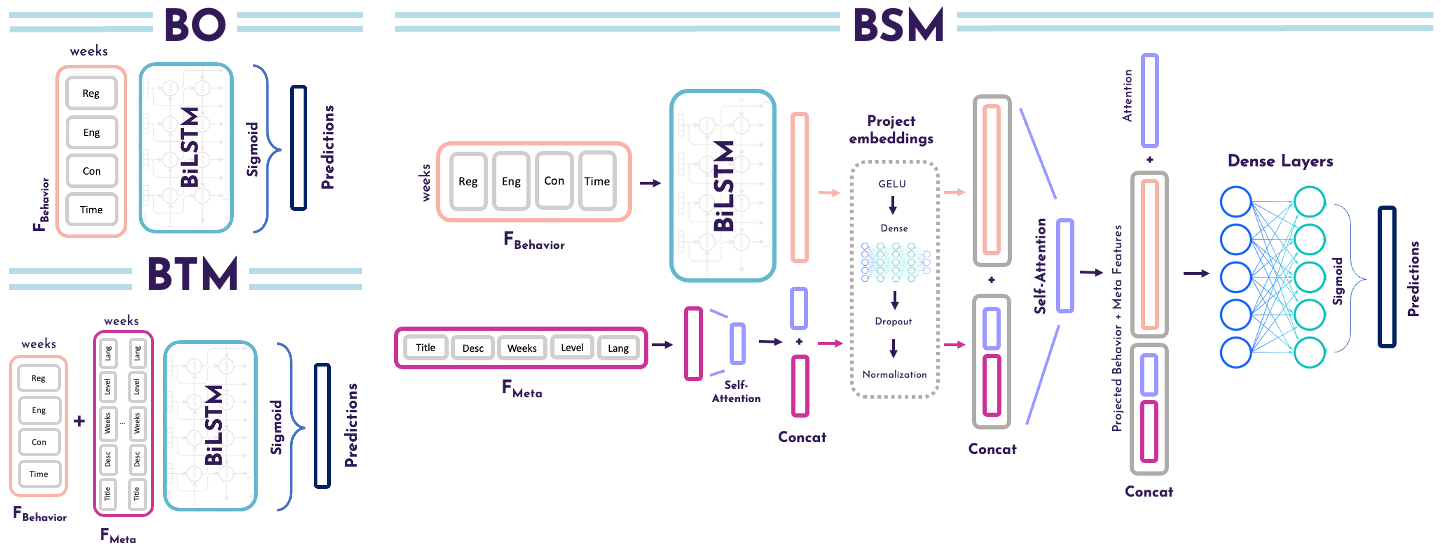}
  \caption{Model architectures adopted in our study.}
  \label{fig:models}
\end{figure*}

\vspace{2mm}The features representing the Title, the Short Description, and the Long Description were obtained by translating the corresponding text to English via the DeepL translation service and then fine-tuning a pre-trained FastText model \cite{fasttext} on that specific meta information\footnote{We experimented with the size of the latent vector space and the encoding frameworks and found that FastText outperformed other out-of-the-box encoders for this use case (SentenceBERT \cite{reimers2019sentencebert} and Universal Sentence Encoder \cite{cer2018universal}).}. FastText is a state-of-the-art model for word representation (i.e., representing words in a text as latent vectors that can be used for machine learning classification tasks). Intuitively, we expect the Duration, Language, and Level meta features to encode similarities in student population across courses and the Topic and Description meta features to reflect similarity on course content. Given a course $c \in \mathbb{C}$, we extract the above features and concatenate them to obtain the final combined meta features for that course. Formally, the meta features for course $c$ are denoted as $F^c \in \mathbb{R}^{124}$, with $F^c = [F_1 \cdot \ldots \cdot F_6]$. Again, due to the different scales of these meta features, the values in $F^c$ are min-max normalized per feature. 

\subsubsection{Classification}\label{sec:model-creation}
In our transfer setting, given a set of courses $C_{trn} \subset \mathbb{C}$, we are interested in creating an accurate success prediction model trained on $C_{trn}$ for a specific early prediction level $e$, which can be transferred to new (unseen) courses or new (unseen) iterations of a course. We are interested in creating models that can predict accurately on courses not seen during training, i.e., $C_{trs} \subset \mathbb{C}$, with $C_{trs} \cap C_{trn} = \emptyset$ and $C_{trs} \cup C_{trn} = \mathbb{C}$. For a given early prediction level $e$, we investigate the three model architectures shown in Fig. \ref{fig:models}. 

\vspace{2mm} \noindent \textbf{Behavior Only} (\texttt{BO}). Our first model predicts student success based on behavior features (and thus student interaction data) only. The architecture of this model is illustrated in Fig. \ref{fig:models}. Formally, the model input is defined as $H = \cup \; H^c\; \forall \;c \in C_{trn}$. Given that each course in $C_{trn}$ potentially has a different number of weeks, we consider the maximum course duration $\textbf{w} = \max(\cup  \; w^c) \; \forall \; c \in C_{trn}$ and pad behavior features for shorter courses with $-1$. The input $H$ therefore has a shape of $|S^c|$ $\times$ $\textbf{w}$ $\times$ $45$. There features are then fed into a neural architecture composed by a \emph{Masking} layer ($mask=-1$), a simple yet effective BiLSTM layer\footnote{We experimented with traditional machine learning models (e.g., Support Vector Machines, Logistic Regression, Random Forest) and deep-learning models (e.g., Dense Fully-Connected Networks, RNNs, LSTMs, CNNs, and BiLSTMs), and found that the \emph{BiLSTM} layer performs best against the other baselines for our use case.} with a loopback of $3$, and a \emph{Dense} layer (with Sigmoid activation) with a hidden size of $1$. The model outputs the probability the student will fail the course.

\vspace{1mm} \noindent \textbf{Behavior + Time-wise Meta} (\texttt{BTM}). Our second model is based on the assumption that students' interaction behavior is influenced by the specific characteristics of the course. We expect that courses with similar characteristics in terms of structure (e.g., topic, duration) and level (e.g., Bachelor) exhibit more similar behavior patterns than courses with completely different characteristics. Therefore, including meta features into the model might improve transfer (i.e. predictive performance of the model on unseen iterations or courses). The resulting model is illustrated in Fig. \ref{fig:models}. We use the exact same architecture as for the \texttt{BO} model; we only change the input data of the model to be a concatenation of both behavior and meta features for each week, formally denoted as $H = \cup \; H^c \cdot F^c \; \forall \; c \in C_{trn}$ (with $\cdot$ representing a concatenation operator). The meta features fed into the model are the same for every week since they are computed per course. The final input $H$ has a shape of $|S^c|$ $\times$ $\textbf{w}$ $\times$ $169$ ($45$ for the behavior features and $124$ for the meta features). By feeding in the meta features for every week, we allow the BiLSTM to directly learn from the dependencies between meta features and behavior features over time.

\vspace{1mm} \noindent \textbf{Behavior + Static Meta} (\texttt{BSM}). Similar to our second model, our third model also relies on the assumption that students' interaction behavior varies dependent on the course characteristics. However, this third model is based on the hypothesis that not all course features are equally important, i.e. some characteristics might have a much influence on student behavior than others. We hence add attention to this model. The architecture consists of separate branches for the behavior feature and meta features, followed by a head architecture that elaborates the combined behavior and meta features (Fig. \ref{fig:models}). The behavior features branch has the same architecture as the \texttt{BO} model, except for the removal of the final \emph{Dense} layer. 

The meta feature branch receives the meta features $F^c$ of the course $c \in C_{trn}$ (shape: $|S^c| \times |F^c|$) as input and is composed of two layers. The first layer is a Bahdanau \emph{Attention} layer \cite{bahdanau2014neural} which returns the weights for all features in $F^c$, representing the importance of each of these features for the prediction, i.e., a vector of size $|F^c|$). The second layer is a \emph{Concatenation} layer that combines both raw meta features and their importance weights returned by the \emph{Attention} layer (shape: $2 \times |F^c|$). This attentive layer is essential for internal and external model interpretability, making it possible to identify the most important (meta) features for the prediction.

Behavior and meta features follow different scales and variation patterns and are differently associated to each input instance (e.g., meta features are the same for each student belonging to a given course, while behavior features change for each student). Therefore, the head architecture combining both the behavior branch and the meta branch does not direct rely on a concatenation of the outputs of these two branches. Rather, the output of each branch is individually projected to the same latent space, obtaining a latent vector (shape: $256$) for the output of each branch. To this end, we use a well-known \emph{Projection} block, composed by a sequence of GELU, Dense, Dropout, and Normalization layers, as proposed by \cite{Multimod35:online}. The two latent vectors are then fed into a \emph{Concatenation} layer (shape: $2 \times 256$). A second Bahdanau \emph{Attention} layer returns a vector of size $2 \times 256 = 512$ representing the weights for all latent behavior and meta features. This second attentive layer weights the overall importance of behavior and of meta information for each prediction. Then, another \emph{Concatenation} layer combines both the latent features and their importance weights (shape: $2 \times 512 = 1024$).  Finally, the concatenated output is passed to a cascade of \emph{Dense} layers, with the last one having a Sigmoid activation and a hidden size of $1$. This final layer returns the probability the student will fail the course. 

\section{Experimental Evaluation}\label{sec:results}

We conducted experiments
to understand the extent to which student behavior transfers well on iterations of the same course or of a different course (RQ1), the extent to which course meta information can improve model transferability, in addition to behavior features (RQ2), and whether fine-tuning the latter meta models on previous iterations of a course unseen during training leads to higher performance on the last iteration of the same course (RQ3). 
The dataset, optimization protocol, and experiments are described below.

\begin{table*}
\centering
\small
\resizebox{\textwidth}{!}{
\begin{tabular}{rrrrrrrrrrrrrrr} 
\toprule
\multirow{2}{*}{\textbf{Title}} & \multirow{2}{*}{\textbf{Identifier}} & \multicolumn{2}{l}{\textbf{Iterations}$^1$} & \multirow{2}{*}{\textbf{Topic}$^2$} & \multirow{2}{*}{\textbf{Level}$^{3^8}$} & \multirow{2}{*}{\textbf{Language}$^4$} & \multirow{2}{*}{\textbf{No. Weeks$^5$}} & \multicolumn{2}{l}{\textbf{No. Students}$^3$} & \multicolumn{2}{l}{\textbf{Passing Rate}$^6$ [\%]} & \multirow{2}{*}{\textbf{No. Quizzes}$^7$} \\
&  & \textit{Trn} & \textit{Trs} & & & & & \textit{Trn} & \textit{Trs} &\textit{Trn} & \textit{Trs} & &\\
\midrule
Comprendre les Microcontrôleurs & Micro & 4 & 0 & Eng & BSc & French & 10 & 3,974 & - & 26.9 & - & 18 \\ 
\hline
Analyse Numérique & AnNum & 3 & 0 & Math & BSc & French & 9 & 1,471 & - & 51.5 & - & 36 \\ 
\hline
Household Water Treatment and Storage & HWTS & 2 & 0 & NS & BSc & French & 5 & 2,438 & - & 47.2 & - & 10 \\ 
\hline
Programmation Orientée Objet & OOP & 1 & 0 & CS & Prop & French & 10 & 797 & - & 38.1 & - & 10 \\ 
\hline
Programmation en C++ & InitProgC++ & 1 & 0 & CS & Prop & English & 8 & 728 & - & 63.3 & - & 13\\ 
\hline
\hline
Digital Signal Processing & DSP & 4 & 1 & CS & MSc & English & 10 & 11,483 & 4,012 & 22.6 & 23.1 & 38\\ 
\hline
Villes Africaines & Villes Africaines & 2 & 1 & SS & BSc/Prop$^9$ & En/Fr$^9$ & 12  & 7,888 & 5,643 & 6.3 & 9.9 & 18 \\ 
\hline
L'Art des Structures I & Structures & 2 & 1 & Arch & BSc & French & 10  & 278 & 95 & 57.7 & 66.3 & 6\\ 
\hline
Functional Programming & ProgFun & 1 & 1 & CS & BSc & French & 7 & 11,151 & 7,880 & 50.72 & 81.33 & 3\\ 
\hline
\hline
Launching New Ventures & Venture & 0 & 1 & Bus & BSc & English & 7 & - & 6,673 & - & 1.4 & 13\\ 
\hline
Éléments de Géomatique & Geomatique & 0 & 1 & Math & BSc & French & 11 & - & 452 & - & 45.1 & 27\\
\bottomrule
\multicolumn{14}{l}{$^1$ \textbf{Set abbrev.} \textit{Trn}: training; \textit{Trs}: transfer.}\tabularnewline  
\multicolumn{14}{l}{$^2$ \textbf{Topic abbrev.}
\textit{Eng}: Engineering;
\textit{Math}: Mathematics;
\textit{NS}: Natural Science;
\textit{CS}: Computer Science; \textit{SS}: Social Science; \textit{Arch}: Architecture; \textit{Bus}: Economics and Business.} \tabularnewline
\multicolumn{14}{l}{$^3$ The values are computed after removing early-dropout students.$^4$\textbf{Level} is chosen by majority label in \textit{Trs} or \textit{Trn}. $^5$\textbf{Language} is chosen by majority label in \textit{Trs} or \textit{Trn}.} \tabularnewline
\multicolumn{14}{l}{$^6$ \textbf{Passing Rate} is averaged over the courses in \textit{Trs} or \textit{Trn} weighted by number of students. $^7$\textbf{No. Quizzes} is averaged over the courses in \textit{Trs} or \textit{Trn}. } \tabularnewline
\multicolumn{14}{l}{$^8$ \textbf{Level abbrev.} \textit{Prop}: Propedeutic / Other; \textit{BSc}: Bachelor; \textit{MSc}: Master. $^9$ For \textit{Villes Africaines}, the / operator represents characteristics of courses in \textit{Trn} / \textit{Trs}.}\tabularnewline 
\end{tabular}}
\vspace{0.4mm}
\caption{Detailed information about the courses included in our dataset.}
\label{tab:transfer}
\end{table*}

% Please add the following required packages to your document preamble:
% \usepackage{graphicx}
\begin{table*}[]
\small
\centering
\resizebox{\textwidth}{!}{
\begin{tabular}{c|cccc||cccc}
\toprule
 & \multicolumn{4}{c}{\textbf{40\% Early Prediction Level}} & \multicolumn{4}{c}{\textbf{60\% Early Prediction Level}} \\ 
 & \textcolor{gray}{\textbf{1-1 Same$^1$}} & \textbf{1-1 Diff$^2$} & \textbf{N-1 Same} & \textbf{N-1 Diff} & \textcolor{gray}{\textbf{1-1 Same$^1$}} & \textbf{1-1 Diff$^2$} & \textbf{N-1 Same} & \textbf{N-1 Diff}\\ \midrule
DSP & \textcolor{gray}{82.0} & {61.1} & {\textbf{83.1}} & 77.4 & \textcolor{gray}{92.7} & {65.3} & \textbf{91.8} & 87.8 \\ 
Villes Africaines & \textcolor{gray}{73.8} & {64.6} & {69.4} & \textbf{79.7} & \textcolor{gray}{82.9} & {67.0} & {80.7} & \textbf{82.7} \\ 
Structures & \textcolor{gray}{52.5} & {51.1} & {52.9} & \textbf{56.2} & \textcolor{gray}{55.2} & {51.3} & {50.4} & \textbf{54.4} \\ 
ProgFun & \textcolor{gray}{50.6} & {50.7} & \textbf{59.5} & {58.7} & \textcolor{gray}{50.8} & {51.0} & \textbf{62.3} & {62.0} \\
Ventures & \textcolor{gray}{51.0} & {{59.8}} & {-} & \textbf{70.3} & \textcolor{gray}{54.9} & {{60.2}} & {-} & \textbf{71.8} \\ 
Geomatique & \textcolor{gray}{76.2} & {57.7} & {-} & \textbf{68.4} & \textcolor{gray}{79.5} & {57.6} & {-} & \textbf{65.5} \\ \bottomrule
\multicolumn{9}{l}{\footnotesize For each course and early prediction level, the highest value on a transfer experiment is in \textbf{bold}.}\tabularnewline
\multicolumn{9}{l}{\footnotesize $^1$ \texttt{1-1 Same} is not included in determining the highest value across transfer courses, since it could not exist in a real-world setting.}\tabularnewline
\multicolumn{9}{l}{\footnotesize $^2$ The balanced accuracy is averaged across the five trained models.}\tabularnewline 
\end{tabular}
}
\vspace{1mm}
\caption{Behavior transfer performance in terms of balanced accuracy for each transfer course (the higher it is, the better).}
\label{tab:baseline}
\end{table*}

\subsection{Dataset}
Our dataset includes $26$ MOOCs, taught by instructors of an European university between 2013 to 2015. The data is fully anonymized with regards to student information. In total, this dataset covers $145{,}714$ students. It contains fine-grained video and quiz interactions for each student, e.g., pressing pause on a video or submitting a quiz. After removing the early-dropout students (see Sec. \ref{sec:log-preproc}), our data set contains $73{,}042$ students in total. The variety of courses in the dataset (e.g., in terms of topic, duration, level, and language), the large scale, and the varied population (e.g., students come from different countries and cultures) allows us to provide a realistic estimation of model performance. The courses are spread across two languages, seven topics, and three university levels. The smallest course has $95$ students, whereas the largest course has $11{,}151$ students. Table \ref{tab:transfer} reports detailed information on each course as well as the assignment to train and transfer sets. We will use these assignments for \textbf{RQ1} and \textbf{RQ2}, i.e. we will use $20$ courses as the train courses in $C_{trn}$ ($48{,}287$ students in total), and $6$ courses as the transfer courses in $C_{trs}$ ($24{,}755$ students in total). To answer \textbf{RQ3}, we utilize another train-transfer split: we will hold out one course $C$ completely (i.e. all iterations of $C$ will be in $C_{trs}$) and train on all other courses. Courses are split in train and transfer sets to ensure diversity in topic, size, level, language, and passing rate.

\subsection{Optimization Protocol}
\label{sec:opt-proc}
We trained each model with features $H$ for the courses in $C_{trn}$ under two early prediction levels ($e=40\%$ and $e=60\%$), minimizing the binary cross-entropy loss according to the setting outlined for each research question in the next subsections. Grid search details are included in Appendix \ref{sec:opt-proc-2}. We monitored BAC due to the high class imbalance\footnote{We computed accuracy, F1 Score, AUC, precision, and recall in addition to BAC, but we found these metrics less representative of the predictive performance. 
}. Formally, we trained a model for each hyperparameter combination on the students in $S_{trn} = \cup \; S^c - \tilde{S}^c \; \forall \; c \in C_{trn}$ and then selected the best combination by measuring BAC on students in $S_{val}$. We then used the selected model to predict student success and compute BAC for each transfer course $c \in C_{trs}$.

\subsection{Behavior Transfer Evaluation (RQ1)}
\label{sec:rq1}
Our initial experiment is designed to investigate the extent to which student behavior transfers well on different iterations of the same course or of a different course. We considered four main settings:

\begin{itemize}[leftmargin=*,nolistsep]
\item \texttt{1-1 Same}. For each transfer course $c \in C_{trs}$, we optimized a \texttt{BO} model on a subset of the students ($80\%$) enrolled in the same course $c$. We then predicted on the remaining set of students ($20\%$) in the same course $c$, split into a test set and a validation set of $10\%$ each. We report results using a 10-Fold cross validation, stratified on the pass-fail binary outcome label. This setting is merely meant to give a measure of how difficult it is to predict on a course and is not applicable in the real world, given that we could not know in advance whether the students in the current iteration will pass or fail the course.
\item \texttt{N-1 Same}. For each transfer course $c \in C_{trs}$ (with $c \in C \subset \mathbb{C}$), we trained a \texttt{BO} model on the previous iterations of course $c$, i.e., $(C \; - \; \{c\}) \; \cap \; C_{trn}$. We then predicted on students of course $c$. 
\item \texttt{1-1 Diff}.  For each transfer course $c \in C_{trs}$ (with $c \in C \subset \mathbb{C}$), we optimized one separate \texttt{BO} model on each course $\tilde{c} \in C_{trs}$, with $c \neq \tilde{c}$. For each model, we then predicted on course $c$. 
\item \texttt{N-1 Diff}. For each transfer course $c \in C_{trs}$, we optimized a single \texttt{BO} model on all the training courses included in $C_{trn}$. We then predicted on students of course $c$. 
\end{itemize}

\vspace{2mm} \noindent The BAC achieved by our \texttt{BO} models on each combination of transfer course and training setting are listed in Table \ref{tab:baseline}. For the $40\%$ early prediction level (left section of Table \ref{tab:baseline}), we observe that transferring a model from one course to another (\texttt{1-1 Diff}) does not work well for any course. Training on previous course iterations (\texttt{N-1 Same}) works best for the \emph{ProgFun} and \emph{DSP} courses. We hypothesize that this observation is because the latter two courses include a large number of students ($> 15{,}000$) in their previous iterations and therefore, training the \texttt{BO} directly on the respective previous iterations leads to better predictive performance. Training a \texttt{BO} model on a large set of courses (\texttt{N-1 Diff}) on the other hand works best for \emph{Villes Africaines}, which is a course with a very low passing rate ($< 8\%$). We assume that in case of highly imbalanced courses, the \texttt{N-1 Diff} model leads to less biased predictions. The \texttt{N-1 Diff} model is also the best model for \emph{Structures}. This course contains a small number of students only (${\sim}100$ per iteration) and the passing rates for the different iterations vary a lot, which might explain the lower BAC of the \texttt{N-1 Same} model in comparison to the \texttt{N-1 Diff} model. We also observe that for \emph{Venture}, the \texttt{N-1 Diff} model by far outperforms a model trained on students from the course iteration (\texttt{1-1 Same}). \emph{Venture} is a course with a very low passing rate ($1.4\%$), likely leading to a high bias (towards failing students) in case of the \texttt{1-1 Same} model. The \texttt{N-1 Diff} model also performs well for \emph{Geomatique}, but we observe a transfer gap here with respect to a model trained on students of the same course iteration (\texttt{1-1 Same}).

For the $60\%$ early prediction level (right section of Table \ref{tab:baseline}), we make similar observations. The \texttt{N-1 Diff} represents the best setting for \emph{Villes Africaines} and \emph{Structures}. The \texttt{N-1 Diff} is again outperformed by the \texttt{N-1 Same} setting for courses with the highest number of students ($> 15{,}000$) and a reasonably high passing rate ($> 20\%$), i.e., \emph{ProgFun} and \emph{DSP}. Also at this early prediction level, the \texttt{BO} model trained under the \texttt{N-1 Diff} setting represents the best real world  solution on four out of six transfer courses.

We finally focus on the extent to which the balanced accuracy varies across courses under the best performing transfer settings. For courses with a very high number of students and a reasonably high passing rate, the best models trained under the \texttt{N-1 Same} setting show a high variance in performance ($40\%$ level: $83.1$ BAC for \emph{DSP} and $59.5$ BAC for \emph{ProgFun}; $60\%$ level: $91.8$ BAC for \emph{DSP} and $62.3$ BAC for \emph{ProgFun}). This difference in transfer performance could be justified by the fact that \emph{DSP} includes quizzes in the schedule (in contrast with \emph{ProgFun}, which only includes videos) and thus the behavior features are more discriminative for the course. The hypothesis that the lower performance is due to differences in the course design is also supported by the low BAC of the \texttt{1-1 Same} model, indicating that \emph{ProgFun} is a hard course to predict on. For courses with a low number of students and those with a high number of students but a low passing rate (\emph{Villes Africaines}, \emph{Structures}, \emph{Ventures}, and \emph{Geomatique}), the best models under \texttt{N-1 Diff} show a more stable transfer performance ($40\%$ level: avg. BAC $68.7$, std. dev. BAC $8.4$; $60\%$ level: avg. BAC $68.6$, std. dev. BAC $10.3$).

%meta transfer results per course
\begin{figure*}[t]
\centering
\subfloat[40\% early prediction level \label{fig:noearly-40}]{
    \includegraphics[width=0.5\linewidth, trim=4 4 4 4,clip]{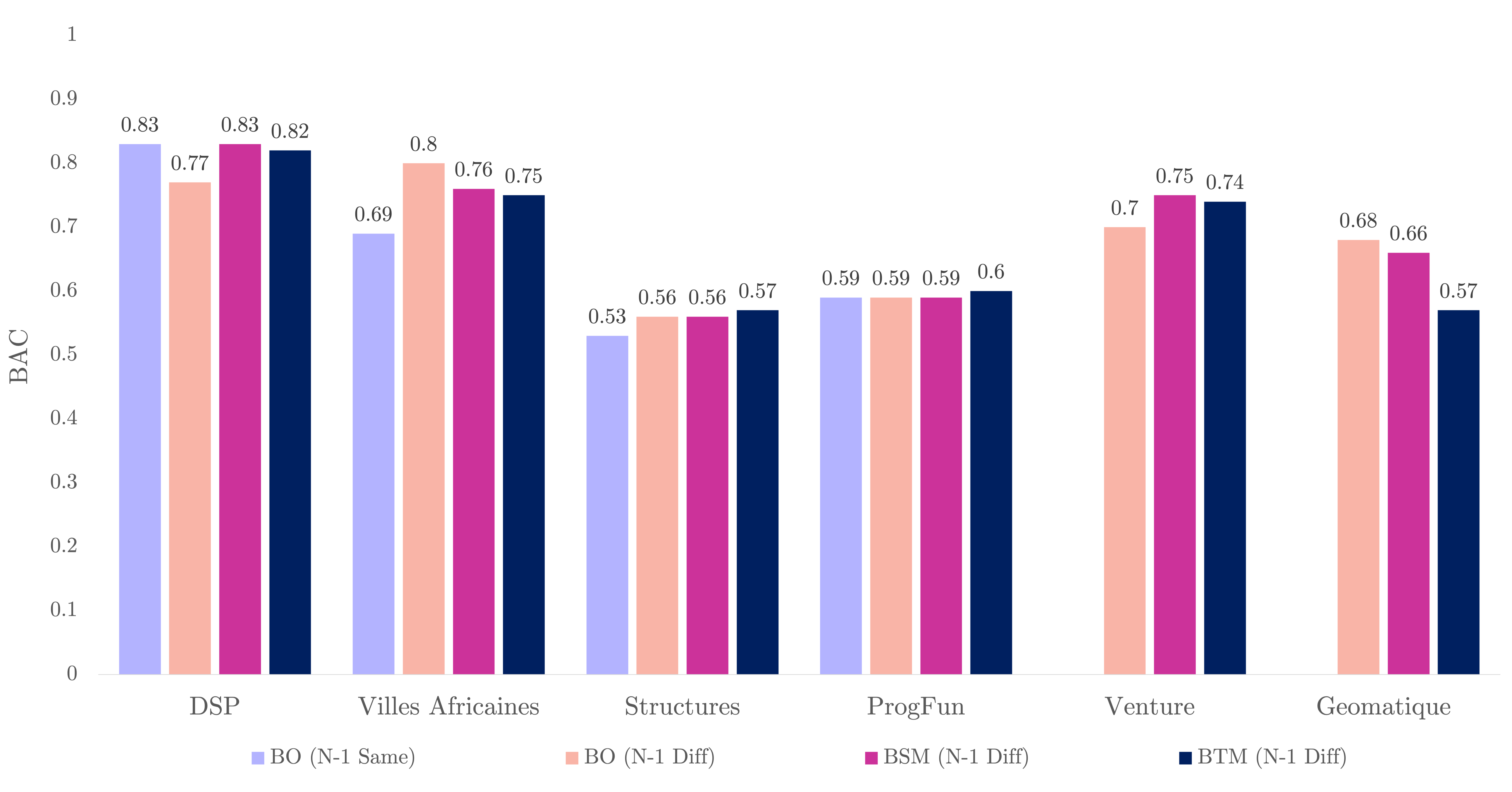}}
\subfloat[60\% early prediction level \label{fig:noearly-60}]{
  \includegraphics[width=0.5\linewidth, trim=4 4 4 4,clip]{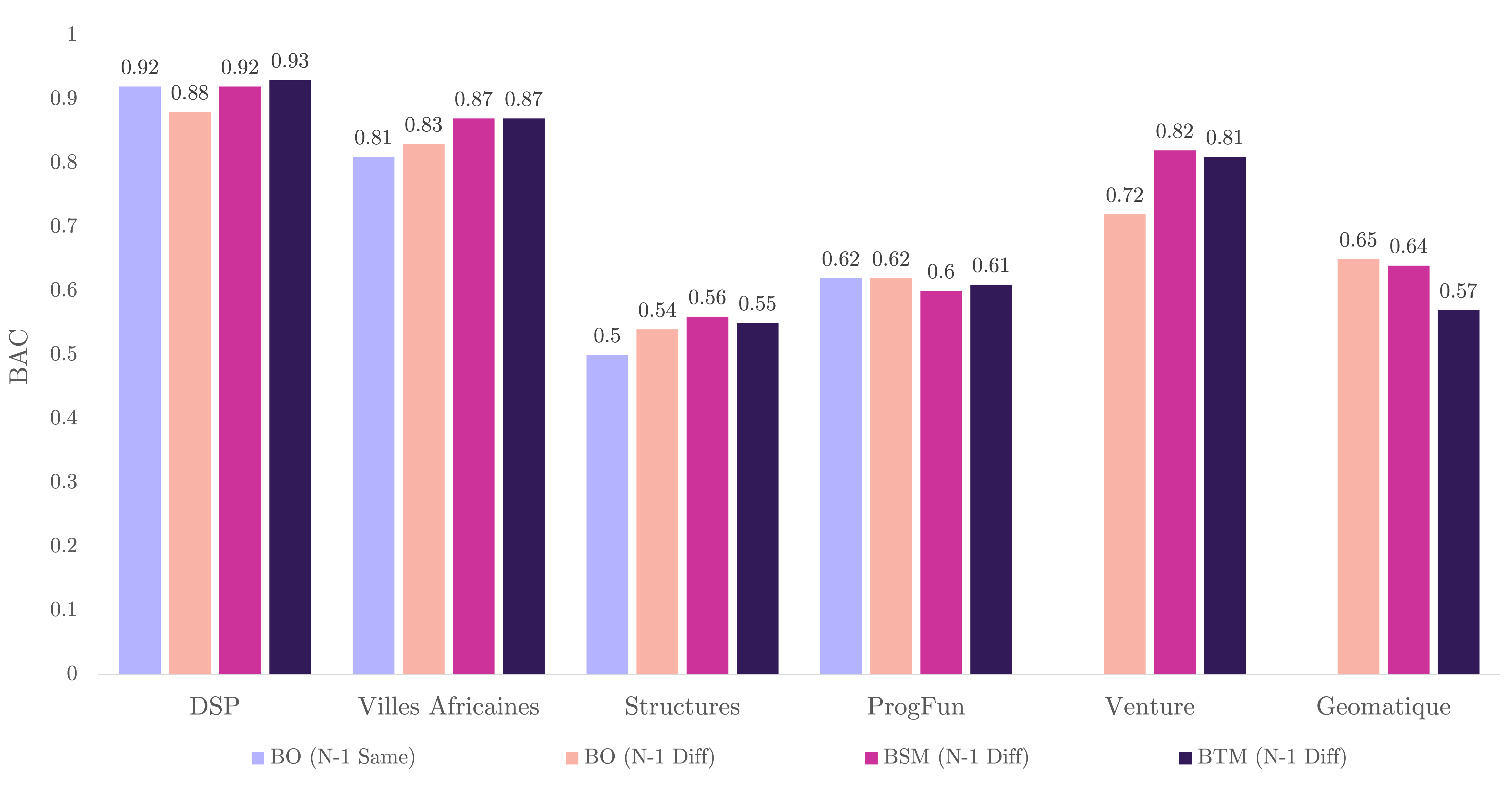}}
\caption{Comparison of the balanced accuracy achieved by our behavior only and behavior + meta models.}
\label{fig:meta-models}
\end{figure*}

\begin{graybox} For courses with previous iterations, a very high number of students, and a reasonably high passing rate, our \texttt{BO} model trained only on students from these previous iterations (\texttt{N-1 Same}) works the best. In all other cases, our \texttt{BO} model trained on students from the large set of courses (\texttt{N-1 Diff}) is the most accurate. This demonstrates the benefits of a generalized educational model trained on many courses.
\end{graybox} 

\subsection{Behavior + Meta Transfer Evaluation (RQ2)}\label{sec:results-metatransfer}
In a second experiment, we were interested in investigating the extent to which course meta information can improve model transferability. We therefore experimented with how to best utilize the meta information for our educational setting, trying different model architectures that pass in meta features at different locations of the model (see Section \ref{sec:model-creation}). We also identified which types of meta information were most relevant to the model through an ablation study. We considered four main settings for the courses, as follows:

\begin{itemize}[leftmargin=*,nolistsep]
\item \texttt{BO N-1 Same}. For each transfer course $c \in C_{trs}$ (with $c \in C \subset \mathbb{C}$), we optimized a \texttt{BO} model on the previous iterations of course $c$ included in the training set, i.e., $(C \; - \; \{c\}) \; \cap \; C_{trn}$. We then predicted on the students of course $c$.
\item \texttt{BO N-1 Diff}.  For each transfer course $c \in C_{trs}$, we optimized a single \texttt{BO} model on all the training courses included in $C_{trn}$. We then predicted on the students of course $c$. 
\item \texttt{BSM N-1 Diff}. For each transfer course $c \in C_{trs}$, we optimized a single \texttt{BSM} model on all the training courses included in $C_{trn}$. We then predicted on the students of course $c$. 
\item \texttt{BTM N-1 Diff}.  For each transfer course $c \in C_{trs}$, we optimized a single \texttt{BTM} model on all the training courses included in $C_{trn}$. We then predicted on the students of course $c$. 
\end{itemize}

%average meta transfer figure
\begin{figure*}[t]
\vspace{-7mm}
\centering
\subfloat[40\% early prediction level \label{fig:avg-40}]{
    \includegraphics[width=0.5\linewidth, trim=4 4 4 4,clip]{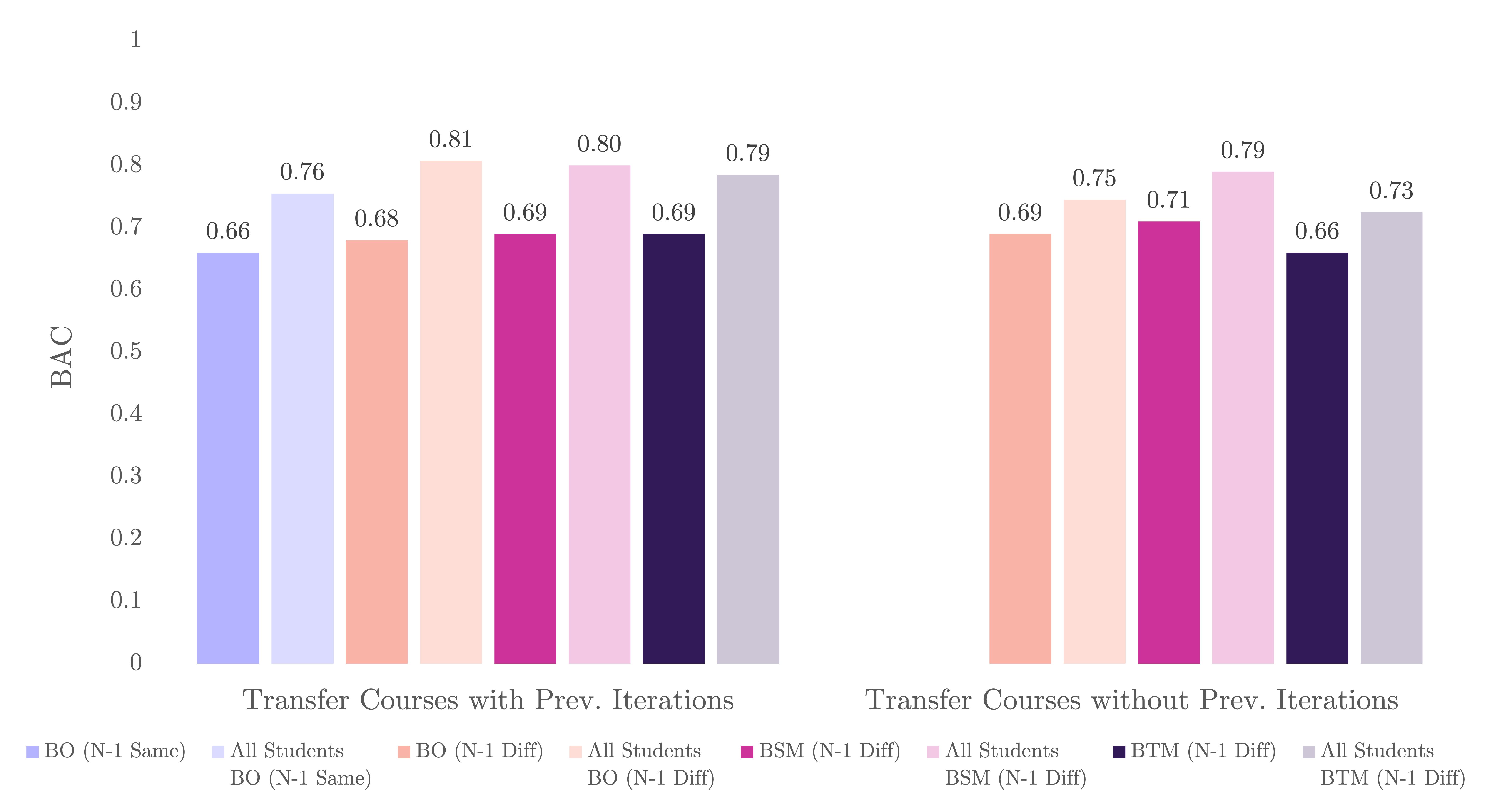}}
\subfloat[60\% early prediction level \label{fig:avg-60}]{
  \includegraphics[width=0.5\linewidth, trim=4 4 4 4,clip]{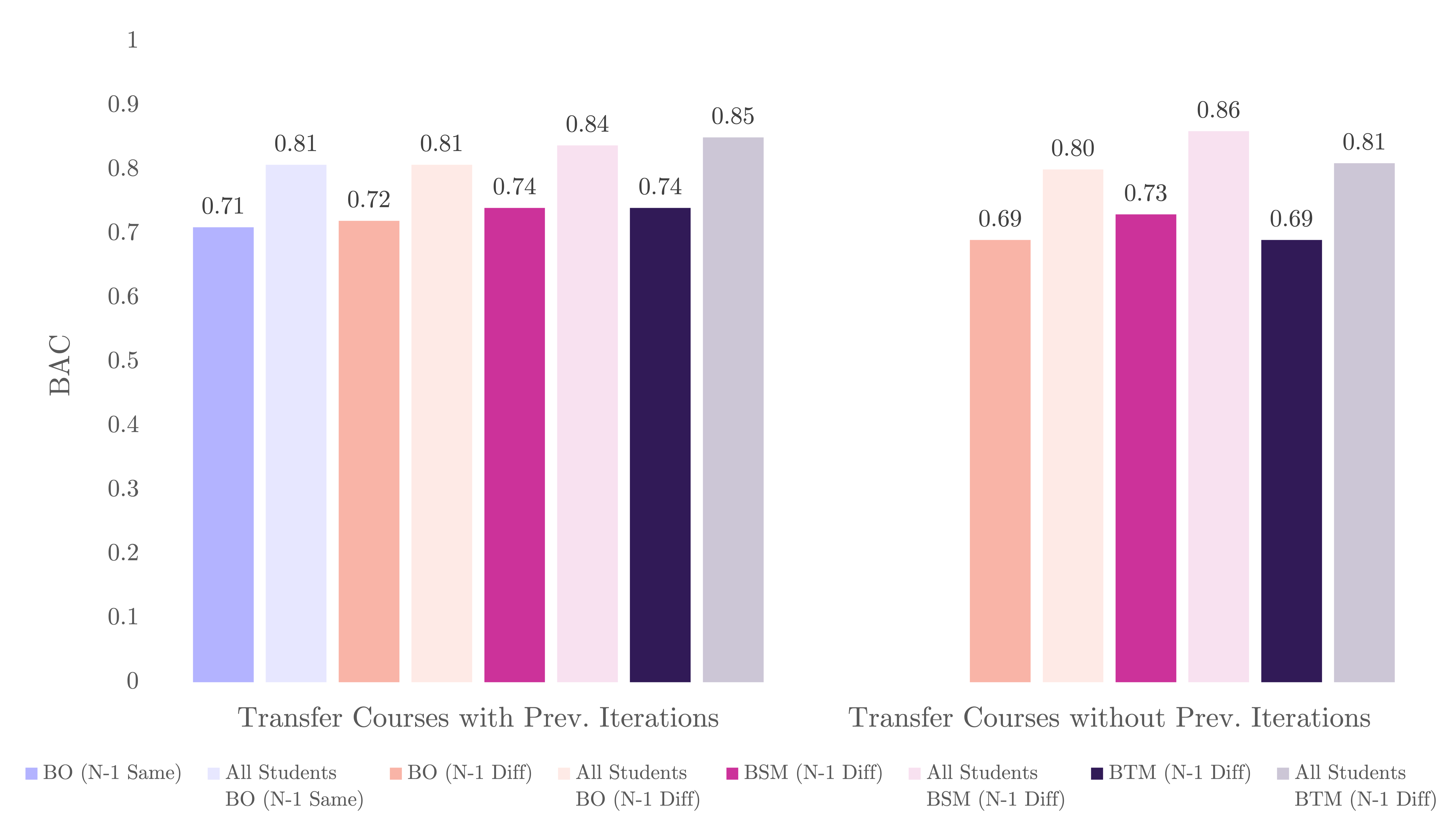}}
  \vspace{-2mm}
\caption{Average performance of \texttt{BO, BSM, BTM} across transfer courses with and without previous iterations, both on the filtered (w/ early-dropout students) and the entire student population.}
  \vspace{-3mm}
\label{fig:avg-meta-models}
\end{figure*}

\vspace{1mm} \noindent \textbf{Predictive Power Analysis}. Figure \ref{fig:avg-meta-models} illustrates the average predictive performance of our models across all transfer courses. We computed the average separately for courses with previous iterations in $C_{trn}$ and for courses that do not have previous iterations. We observe that adding meta information increases the average BAC for courses with previous iterations, no matter how that meta information is added: the \texttt{BSM (N-1 Diff)} and the \texttt{BTM (N-1 Diff)} both outperform the (\texttt{BO (N-1 Diff)}) model for both early prediction levels. For the courses without previous iterations we again observe the same pattern for the $40\%$ and $60\%$ early prediction levels. However, for these courses, the \texttt{BSM (N-1 Diff)} model performs better than the \texttt{BTM (N-1 Diff)} model. It therefore seems that for courses with previous iterations in $C_{trn}$, adding the maximal amount of meta features helps the model focus on the similar courses (i.e. the previous iterations of the transfer course) in the training data. In contrast, for models without previous iterations, it seems to be important to focus on only the relevant meta information (hence the better performance of the \texttt{BSM (N-1 Diff)} model. With respect to training on previous iterations, we observe that all the models trained on a large data set that include a range of courses (\texttt{BO (N-1 Diff)}, \texttt{BSM (N-1 Diff)}, \texttt{BTM (N-1 Diff)}) show the same or a higher BAC than the models trained on previous iterations of a course only (\texttt{BO (N-1 Same)}).

The shaded bars in Fig. \ref{fig:avg-meta-models} denote the predictive performance of our models for the whole student population (including early-dropout students). We observe that every one of our models achieve a higher BAC for all students in both early prediction settings. This result is expected as the early-dropout students tend show very low activity already at the beginning of the course, making it easy to predict on them. We also observe that the patterns of the average BAC are the same when predicting on the whole population versus on the filtered population. For the remaining experiments in this chapter, we will therefore report the BAC of the different models only on the filtered population (as described in Sec. \ref{sec:log-preproc}).

Figure \ref{fig:meta-models} illustrates the BAC achieved by our models separately for each transfer course. For the $40\%$ early prediction level (Fig. \ref{fig:noearly-40}), it can be observed that adding meta information is helpful only for some courses of the courses with previous iterations. For \emph{DSP}, adding meta information leads to a strong increase in BAC: both meta models (\texttt{BSM N-1 Diff} and \texttt{BTM N-1 Diff}) outperform the model including behavior features only (\texttt{BO N-1 Diff}). For \emph{ProgFun} and \emph{Structures}, including meta information does not significantly increase the BAC (in the case of the \texttt{BSM N-1 Diff} and \texttt{BTM N-1 Diff} models). In the case of courses without previous iterations, adding meta information increases the BAC for one course (\emph{Ventures}), but leads to a slightly decreased predictive accuracy for the second course (\emph{Geomatique}). We further observe that the \texttt{BSM N-1 Diff} model tends to outperform the \texttt{BTM N-1 Diff} model. It performs better in four out of six courses. Furthermore, the \texttt{BSM N-1 Diff} model performs better than or at least as well as the model including behavior features only (\texttt{BO N-1 Diff}) for four out of the six courses. We observe a drop in performance only for two courses. Compared to the other transfer courses, \emph{Villes Africaines} has a high number of students and a highly imbalanced passing rate.
The second course, \emph{Geomatique}, seems to be hard to transfer to, as the performance of the \texttt{BSM N-1 Diff} model is very close to the performance of the \texttt{BO N-1 Diff} model. For courses with previous iterations, combining meta information and behavior features (\texttt{BSM N-1 Diff}) led to the same or a higher BAC as the model trained on previous iterations of the transfer course (\texttt{BO N-1 Same}).

For the $60\%$ early prediction level (Fig. \ref{fig:noearly-60}), the benefit of including meta information is more evident. Except for \emph{ProgFun} and \emph{Geomatique}, both meta models (\texttt{BSM N-1 Diff} and \texttt{BTM N-1 Diff}) have a BAC at least as high as the model trained on behavior features only (\texttt{BO N-1 Diff}. In contrast to the $40\%$ early prediction level, the performance differences between the \texttt{BSM N-1 Diff} model and the \texttt{BTM N-1 Diff} model are not as pronounced; the \texttt{BSM N-1 Diff} model outperforms the \texttt{BTM N-1 Diff} model for three courses, but performs (slightly) worse than the \texttt{BTM N-1 Diff} model for two other courses. Both the \texttt{BSM N-1 Diff} model and the \texttt{BTM N-1 Diff} model achieve or exceed the predictive performance of a model trained on previous course iterations (\texttt{BO N-1 Same}) for three out of four courses. For \emph{ProgFun}, the predictive performance of the two models is very close to that of the \texttt{BO N-1 Same} model.
     
Subsequently, we focus on the amount the BAC varies across courses under the best performing setting, namely \texttt{BSM N-1 Diff}. On average, the BAC is higher in all model settings for the $60\%$ early prediction level than for the $40\%$ level. This is expected because we would assume that the predictive power generally improves with more available data about the student behavior. We observe that the \texttt{BSM N-1 Diff} model shows a lower BAC for \emph{Structures} and \emph{ProgFun} in comparison to the other courses. We have already observed in Sec. \ref{sec:rq1} that it seems to be generally hard to predict on the selected iteration of the \emph{ProgFun} course (compare also the predictive performance of the \texttt{1-1 Same} model in Table \ref{tab:baseline}). Both \emph{Structures} and \emph{ProgFun} include a comparably low number of quizzes (see Table \ref{tab:transfer}), we therefore hypothesize that the presence of quizzes has a positive impact on model transfer.

\begin{figure*}[]
\centering
\subfloat[$40\%$ early prediction level\label{fig:weights40-ablation}]{
   \includegraphics[width=0.5\linewidth, trim=4 4 4 4,clip]{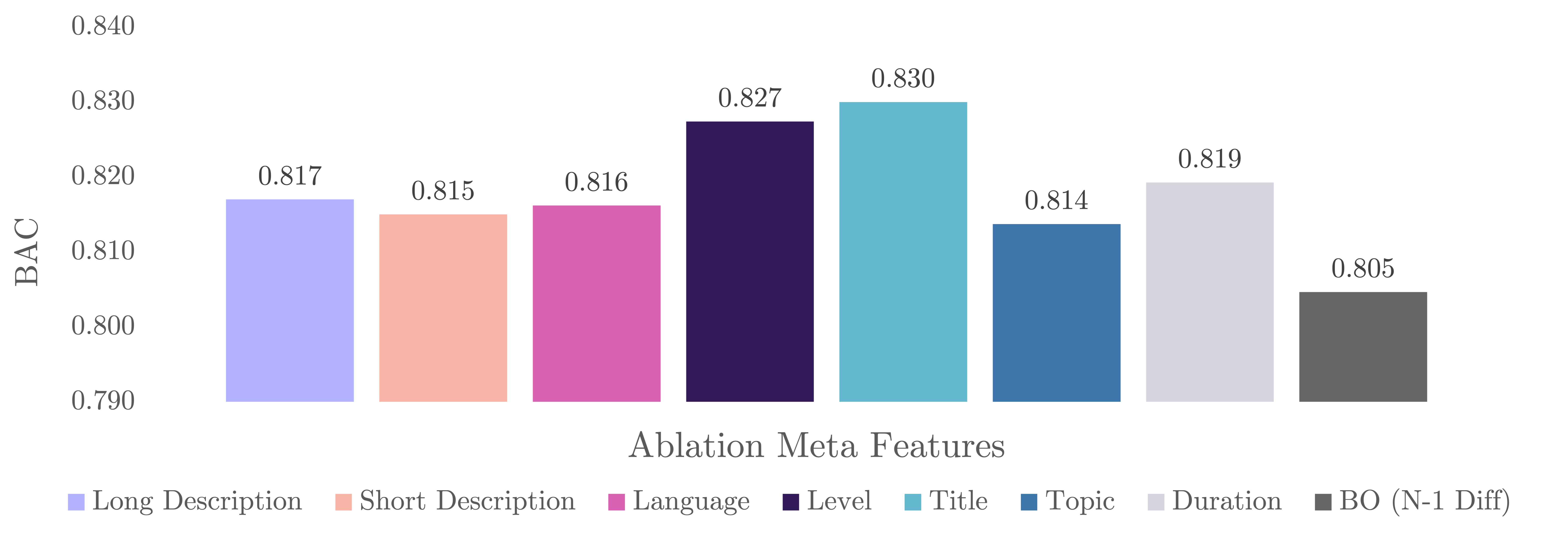}}
\subfloat[$60\%$ early prediction level\label{fig:weights60-ablation}]{
   \includegraphics[width=0.5\linewidth, trim=4 4 4 4,clip]{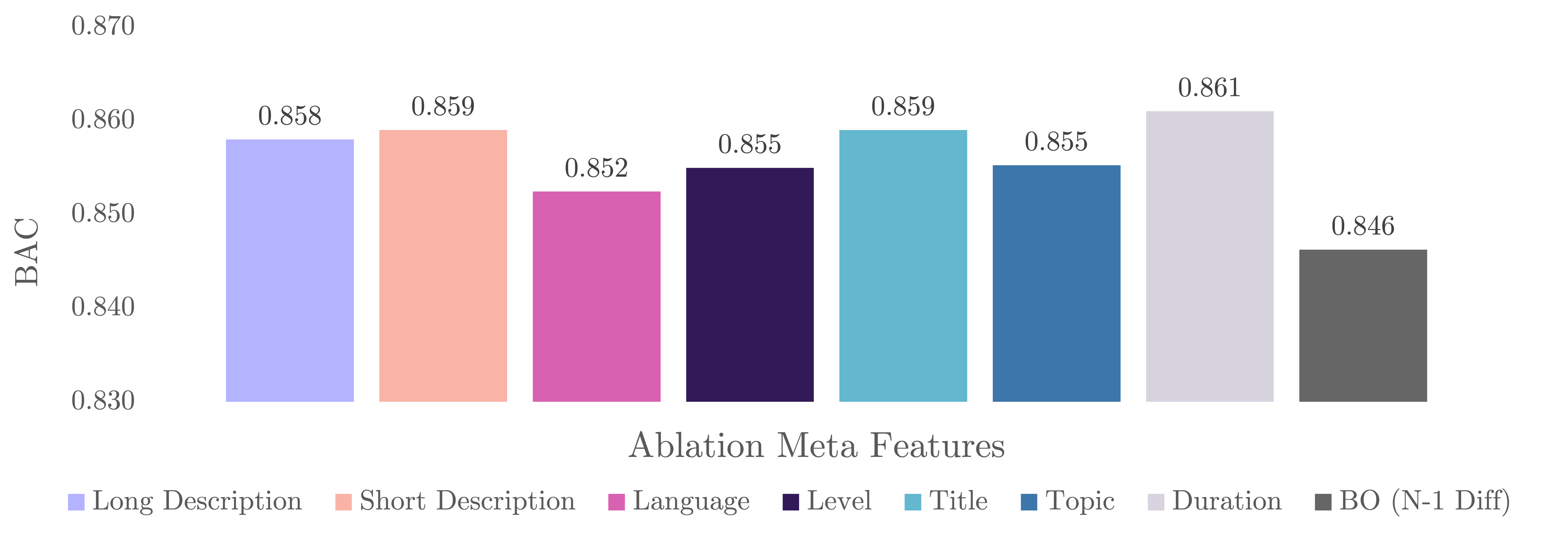}}
   \caption{Ablation study on the meta features for the  \texttt{BTM} model.}
\end{figure*}

\begin{figure*}
\centering
\includegraphics[width=1.0\linewidth]{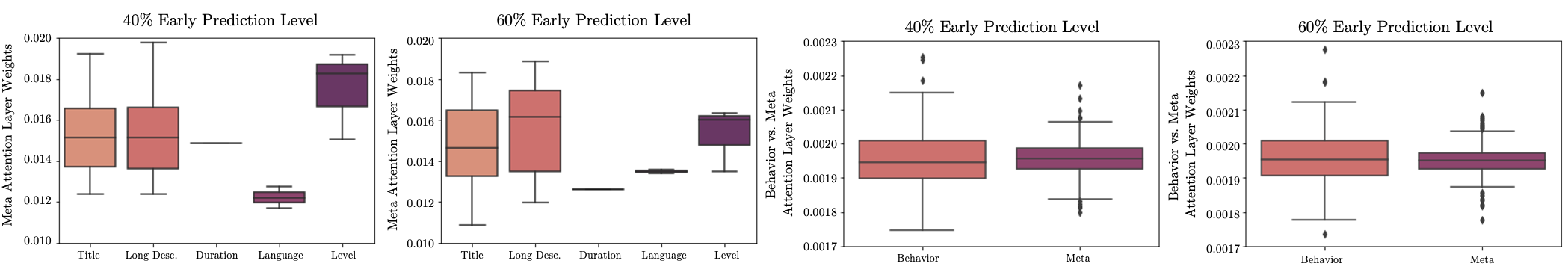}
\caption{Importance of the different meta features on the predictive power of the \texttt{BSM} model.}
\label{fig:weights}
\end{figure*}

\vspace{1mm} \noindent \textbf{Meta Information Importance Analysis}. To have a more detailed picture of the impact of the different meta features (see Section \ref{sec:features-descr}), we performed an ablation study. We trained a \texttt{BTM N-1 Diff} model on $80\%$ of the students attending the courses in $C_{trn}$ with only one meta feature at a time. We then computed the BAC of the trained model on the remaining $20\%$ of the students, split into a test and validation set. When performing the train-test split, we stratified by course and pass-fail label. Figures \ref{fig:weights40-ablation} and \ref{fig:weights60-ablation} illustrate the resulting BAC for the $40\%$ and $60\%$ early prediction levels. The rightmost bar represents the BAC of a \texttt{BO N-1 Diff} model (no meta information included). Notably, all the meta features improve the performance of the model over the baseline \texttt{BO N-1 Diff} on both early prediction levels. For the $40\%$ early prediction level, the most important meta features are represented by Title (with an embedding size of $60$), Level, and Duration. It therefore seems that Title contains enough information to help the model leverage patterns from courses about similar topics. Furthermore, the ablation importance of Level indicates that the interaction behavior of students might differ according to their school level (e.g. Bachelor vs. Master). We theorize that Duration is important since longer courses include more weeks of student behavior data at the $40\%$ early prediction level.
In the $60\%$ early prediction level, the most important meta feature is Duration: there is a minimal difference in importance for the other meta features. Based on the results of this ablation study, we include all meta features into our models. We however perform a grid search over Short/Long Description and Title (see Sec. \ref{sec:opt-proc}) to avoid training on highly redundant information.

To have additional insight into which meta features our model considers useful, we examine the two \emph{Attention} layers of our \texttt{BSM} model. By analyzing the importance weights resulting from these two layers, we can assess the impact of the different meta features in combination with the behavioral features on success prediction. We investigate these importance weights for the \texttt{BSM N-1 Diff} model trained on the courses in $C_{trn}$ with all the meta features. 
The importance weights are illustrated in Fig. \ref{fig:weights} for both layers and both the $40\% $and the $60\%$ early prediction levels. Considering the first \emph{Attention} layer for the $40\%$ early prediction level (two leftmost plots) which weights the meta features separately, it can be observed that on average, Level appears as the highest weighted meta features, followed by Title, Long Description, and Duration. As also observed in the ablation study, Language appears to have lower importance than the other meta features at this level. For the $60\%$ early prediction level, the weights assigned to Long Description and Language increase, whereas the weights of Duration and Level decrease. We hypothesize that when a model has less data about the students in the course ($40\%$ early prediction level), it relies on heuristics about Duration and Level for estimating the course structure and pass-fail patterns and making predictive decisions. When more information about the course is available ($60\%$ level), meta features like Long Description and Language become slightly more relevant as the similarity of the course to other courses in terms of these meta features is more useful to the model. Considering the second \emph{Attention} layer (two rightmost plots), weighting all meta features against all behavior features, it can be observed that the importance of the behavior and meta features is, on average, equal. However, under the $60\%$ early prediction level, the variance of the importance for both types of features is lower, i.e. there are less polarized cases where either behavior features only or meta features only are important for a specific student.  

\begin{graybox} In summary, combining behavior with meta features in a student prediction model leads to consistently higher transfer performance across courses, regardless of the early prediction level. Even when previous iterations are available, the predictive power of meta models is comparable to that of a model trained on the previous iterations of that course.
In particular, a static combination (\texttt{BSM}) is more transferable than a time-wise combination (\texttt{BTM}). Level, Title, and Duration are important for very early predictions (before mid-course), while all the meta features have a similar importance, with Duration still being slightly more important, for later predictions. \end{graybox}

\begin{figure*}[]
\centering
\subfloat[$40\%$ early prediction level\label{fig:fine40-tuned}]{
   \includegraphics[width=0.5\linewidth, trim=4 4 4 4,clip]{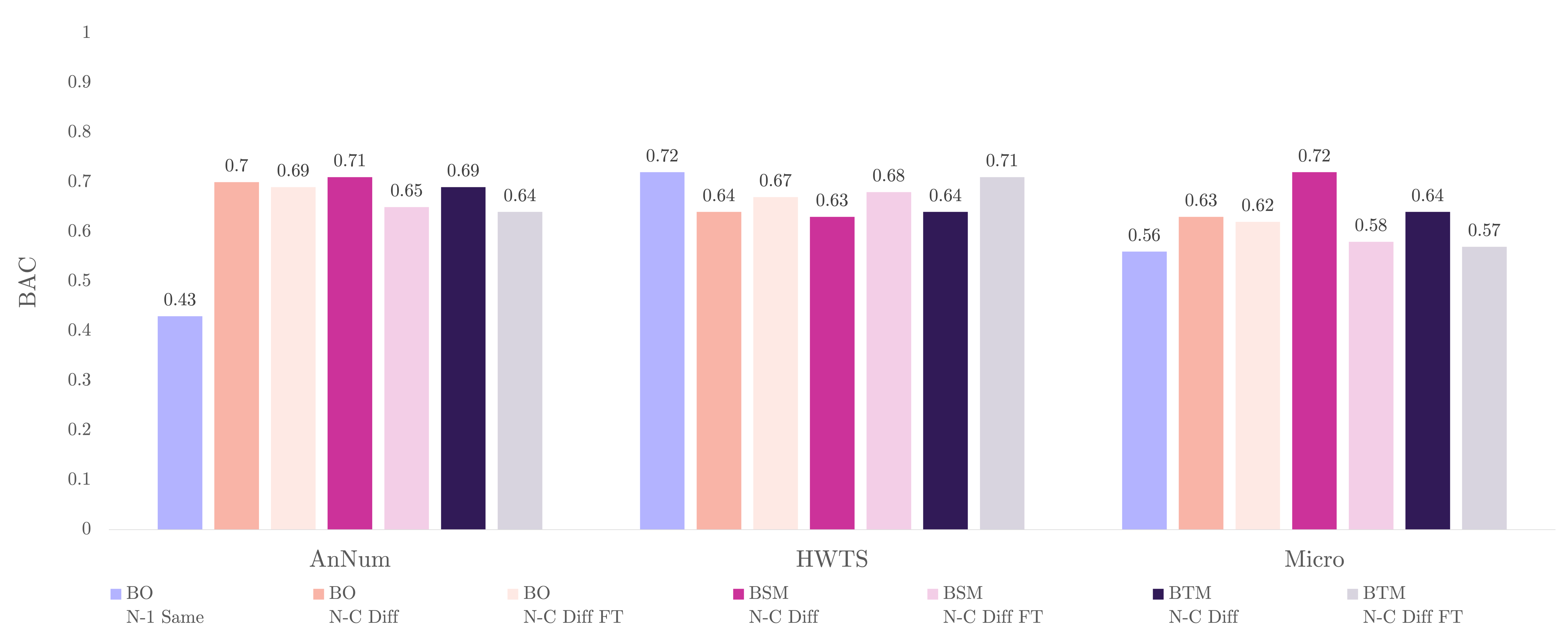}}
\subfloat[$60\%$ early prediction level\label{fig:fine60-tuned}]{
   \includegraphics[width=0.5\linewidth, trim=4 4 4 4,clip]{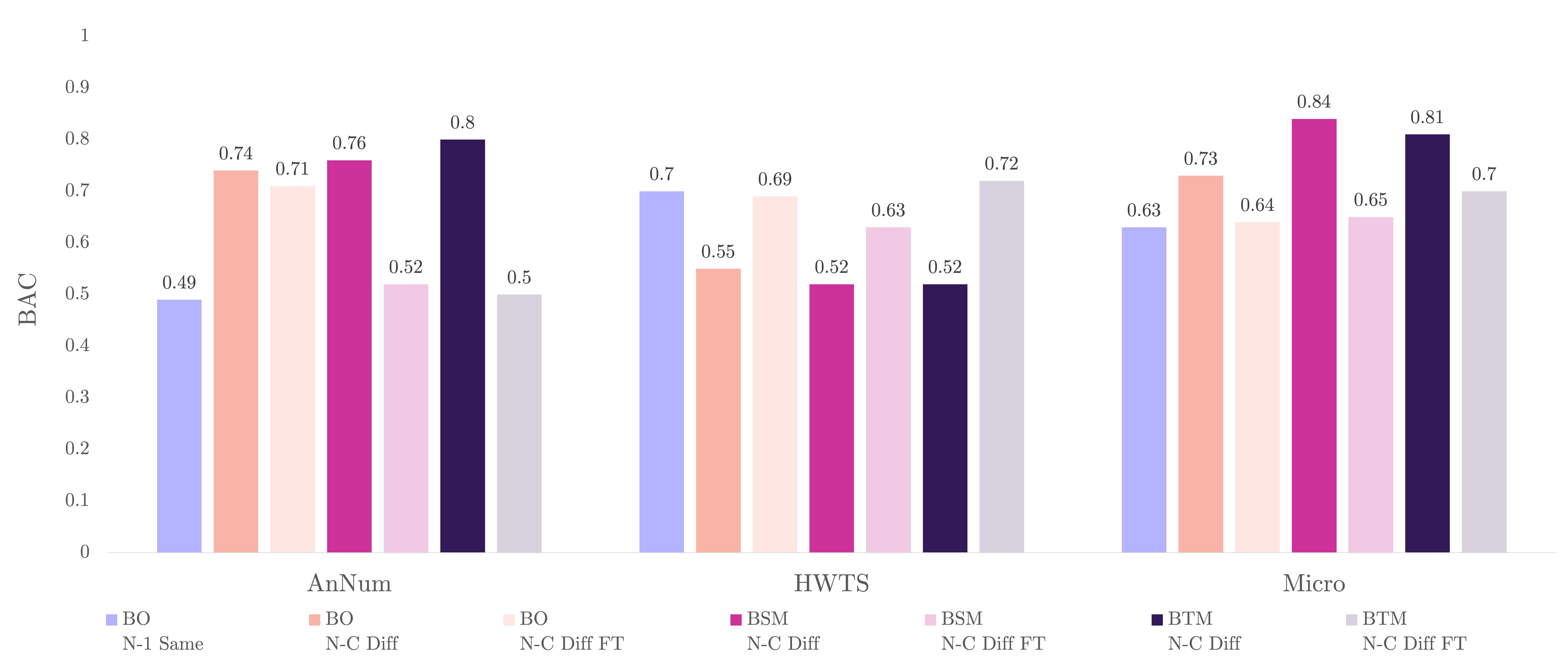}}
\caption{Comparative balanced accuracy of different fine-tuned meta models.}
\label{fig:fine-tuning}
\end{figure*}
\subsection{Transfer-by-Fine-Tuning Evaluation (RQ3)}
In a third and final experiment, we were interested in investigating the extent to which fine-tuning behavior and meta models on previous iterations of a course improves transfer. Related work from the broader computer vision community has demonstrated that fine-tuning can improve performance in transfer learning models \cite{guo2019spottune}. To form an analogous setup for this experiment, we held out a complete course $C$ (i.e. all iterations of that course) from the training data. In a second step, we fine-tuned the obtained model on a subset of iterations of $C$. Finally, we used this fine-tuned model to predict on the last iteration of $C$. We selected three courses with different characteristics (e.g., Duration, Topic) for our fine-tuning experiment and hence $C_{trs_{ft}} = \{Micro,  AnNum,  HWTS\}$.

\begin{itemize}[leftmargin=*,nolistsep]
\item \texttt{BO N-1 Same}. For each transfer course $c \in C_{trs_{ft}}$, we optimized a \texttt{BO} model on the previous iterations of course $c$, i.e. $C-c_{M^C}$.  We then predicted on the last iteration $c_{M^C}$ of course $C$. 
\item \texttt{BO N-C Diff}.  For each transfer course $C \in C_{trs_{ft}}$, we optimized a \texttt{BO} model on the training data set $(C_{trn}-C) \cup C_{trs}$. We then predicted on the students of the last iteration $c_{M^C}$ of course $C$. 
\item \texttt{BSM N-C Diff}. For each transfer course $C \in C_{trs_{ft}}$, we trained a  \texttt{BSM} model on the training data set $(C_{trn}-C) \cup C_{trs}$. We then predicted on the students of the last iteration $c_{M^C}$ of course $C$. 
\item \texttt{BTM N-C Diff}. For each transfer course $C \in C_{trs_{ft}}$, we optimized a \texttt{BTM} model on the training data set $(C_{trn}-C) \cup C_{trs}$. We then predicted on the students of the last iteration $c_{M^C}$ of course $C$.
\item \texttt{BO N-C Diff FT}. For each transfer course $C \in C_{trs_{ft}}$, we fine-tuned the optimal \texttt{BO N-C Diff} model (optimal on the training data set $(C_{trn}-C) \cup C_{trs}$) on a subset of iterations of that course $C-c_{M^C}$. We predicted on the last iteration $c_{M^C}$ of course $C$. 
\item \texttt{BSM N-C Diff FT}. For each transfer course $C \in C_{trs_{ft}}$, we fine-tuned the optimal \texttt{BSM N-C Diff} model (optimal on the training data set $(C_{trn}-C) \cup C_{trs}$) on a subset of iterations of that course $C-c_{M^C}$. We predicted on the last iteration $c_{M^C}$ of course $C$.
\item \texttt{BTM N-C Diff FT}. For each transfer course $C \in C_{trs_{ft}}$, we fine-tuned the optimal \texttt{BTM N-C Diff} model (optimal on the training data set $(C_{trn}-C) \cup C_{trs}$) on a subset of iterations of that course $C-c_{M^C}$. We predicted on the last iteration $c_{M^C}$ of course $C$.
\end{itemize}

\vspace{2mm} Figure \ref{fig:fine-tuning} illustrates the BAC of the different models for the three courses and two early prediction levels. For the $40\%$ early prediction level, we observe that fine-tuning is helpful for only one course. For \emph{HWTS}, fine-tuning on previous iterations of the course improves the BAC of behavior only (\texttt{BO N-C Diff}) and combined meta-behavior (\texttt{BSM N-C Diff} and (\texttt{BTM N-C Diff}) models significantly. For the other two courses, fine-tuning consistently decreases the predictive performance of all investigated model. For the $60\%$ early prediction level, the exact same patterns can be observed.

We postulate that the predictive performance increase achieved via fine-tuning depends on the similarity (of student behavior) between the different iterations of the course. The two iterations of \emph{HWTS} exhibit exactly the same course characteristics in terms of topic, language, duration, and level. Furthermore, both iterations of \emph{HWTS} contain a similar amount of students with a very similar (balanced) passing rate ($45.5\%$ vs. $45.9\%$). Therefore, training on previous iterations works well for this course; as demonstrated in Fig. \ref{fig:fine-tuning}, the \texttt{BO 1-Same} model outperforms the other models for both early prediction levels. For the other two courses, we observe that a model trained on previous iterations of the same course exhibits a suboptimal performance. In particular, the \emph{AnNum} course shows dissimilarities in terms of passing rates over the years, i.e. going from a very low passing rate in the first iteration ($8.5\%$) to passing rates $>70\%$ for iterations $2$ and $3$. Indeed, we observe that training a model on previous iterations does not work well for \emph{AnNum} (see BAC of the \texttt{BO 1-Same} model in Fig. \ref{fig:fine-tuning}), independent of the early prediction level. Finally, \emph{Micro} changed over subsequent iterations: the first run of the course started out with a lower number of students ($571$ students), but was almost balanced in terms of pass-fail labels (pass rate: $49\%$). The subsequent iterations saw a steep increase in student count ($>4{,}000$ students), but also a large drop in the pass rate (${\sim}5\%$). Training a model on a highly imbalanced course can lead to a bias (towards failing students), while fine-tuning on previous iterations of this course deteriorates model performance.

\begin{graybox}
Fine-tuning the generalized \texttt{BTM N-C Diff} model on previous iterations of courses (when they are available and similar to the current course), helps improve performance.
\end{graybox}

%Figures \ref{fig:40_finetune} and \ref{fig:60_finetune} showcase the results of the models before and after fine-tuning in comparison with the baseline Behavior Only from previous iterations model. Notably, in the 40\% cutoff case, the Behavior + Time-wise Meta model with fine-tuning does perform better than all other models in every case, with 1-2\%  BAC improvement. The large improvements through fine-tuning with Time-wise Meta (3\%, 9\%, 2\%, and 8\%) respectively make it clear that when previous iterations are available, fine-tuning can improve performance substantially. More importantly, fine-tuning a meta-feature based model is better than simply training a model on the previous iterations individually, when there is not much data about the model (40\%). It is interesting to observe that fine-tuning the Static Meta model does not help improve the model's performance, even in the 40\% case). fine-tuning the multi-course Behavior only model, we note that while it does augment the baseline model performance in a few cases (DSP by 4\% in the 40\% cutoff, ProgFun by 2\% in the 60\% cutoff), it mostly does not improve over the model without fine-tuning. In the 60\% cutoff, fine-tuning shows no significant improvement in any of the transfer courses. \textbf{Earlier in the course (when 40\% of data is available), fine-tuning from relevant past iterations can show improvement on top of meta learning models. However, later in the course, fine-tuning does not show significant improvement in model performance.}

\section{Conclusions and Future Work}
In this chapter, we investigated the transferability of models trained on different combinations of diverse student behavior and course meta features for early student success prediction. In contrast to prior work \cite{boyer2015transfer}, our study aimed to exploit course datasets at scale to cover student prediction on the last iteration of a given course and on courses with no previous iterations available. To do so, we introduced three novel strategies: 1) pre-training a model on a large set of diverse MOOCs, 2) including meta features using two different architectures, one combining behavior and meta features statically, and another one combining both in a time-wise fashion, and 3) fine-tuning the aforementioned models on previous iterations of courses. We then explored the extent to which student behavior transfers well on different iterations of the same course or of a different course, the extent to which the inclusion of course meta information can improve model transferability, and finally whether fine-tuning on previous iterations of a course unseen during training leads to higher performance on the last iteration of the same course. This is the first comprehensive evaluation of model transferability for \textit{early} success prediction. The key implications are as follows:

\begin{enumerate}[leftmargin=*]
\item Behavior transfer on this task depends on the number of previous course iterations, the number of students in those iterations, and their passing ratio. If a course has past iterations with a high number of students and a reasonably high passing rate, behavior features from the past iterations of that course works best. Otherwise, behavior features from a large and diverse course set leads to higher transferability. Few courses have low BACs across all model architectures (transfer or individually trained), which reflects low underlying signal in the data.
\item Combining behavior and meta features using attention layers results in the best transfer performance across courses, regardless of the early prediction level. The resulting models and the models trained on the previous iterations of a course (if available) have comparable or lower predictive power, confirming findings of prior research on meta transfer learning in different domains \cite{sun2019meta, qureshi2017wind, winata2020meta}. Our pre-trained static meta model could hence be seamlessly applied to predict on new unseen courses.
\item Predictive performance seems to be influenced by the number of quizzes in a course. For both behavior only and meta models, the more quizzes the transfer course included, the higher the BAC was. This implies that instructors who consider utilizing student success predictors should carefully design their course to include a good mix of lecture material and assessments. 
\item While all the meta features have similar importance for later prediction, the course Level, Title, and Duration meta features are important for very early predictions. Therefore, instructors should precisely define this course meta information. Educational data scientists should include such meta features in their models to foster transferability.  
\item When previous course iterations are available and similar (in student population and course structure) to the current transfer course, fine-tuning a meta model on these iterations can better transfer. However, if there are significant differences between iterations, fine-tuning the model will lead to a decrease in BAC.
\end{enumerate}

Our results indicate that behavior and meta features could be successfully used to improve model transferability. The models trained in this chapter can be used to warm-start performance predictions, aiming to improve student performance in ongoing courses through targeted individual and curriculum interventions \cite{perez2021can}. Nevertheless, our work has several limitations that can be addressed with future research. While we tested a variety of MOOCs with different topics and populations on two early prediction levels, it is notable the MOOC dataset is from a single university in Europe and collected five years ago, leading to slight model bias because of the underlying data. Further work is needed to account for different course modalities, such as flipped classrooms and blended courses, to see the extent to which transferability is modality-agnostic. Additionally, our study relies on hand-crafted behavior features extracted from the raw clickstreams. Latent behavior feature models directly acting on clickstreams (e.g., autoencoders) can be combined with our architectures, to provide an end-to-end model and avoid the need of manually extracting behavior features. 

%% file: main/04_evaluating_explainers.tex
The work in this chapter was done in collaboration with Bahar Radmehr, Natasa Krco, Professor Mirko Marras, and Professor Tanja Käser. It was originally published in the Proceedings of the 15th International Conference on Educational Data Mining (2022) with the title \textit{"Evaluating the Explainers: Black-Box Explainable Machine Learning for Student Success Prediction in MOOCs"}.

We thank Professor Paolo Pardoni, Valentin Hartmann, and Natasa Tagasovska for their expertise and support with this chapter.

\section{Introduction}

Although the explainability corpora is growing, there is a clear gap in explainability literature for education \cite{conati2018ai}, with an even more pressing need for work (quantitatively) comparing different explainability methods. To the best of our knowledge, current research on explainability in education is exclusively applied \cite{lu2020towards, hasib-lime, baranyi2020interpretable, mu2020towards, vultureanu2021improving, scheers2021interactive}: the majority of previous research implements only one specific explainability method to interpret the predictions of their proposed approach.

% 3. What is the gap?
% - Not really directly needed here - it is kind of clear from 2

% 4. What have we done:
% - In this paper, we aim to...
% - We suggest this  method
% - We evaluate 5 methods on 5 courses to answer the following 3 research questions

% 5. Our results demonstrate...

To address this research gap, we examine and compare five popular instance-based explainability methods on student success prediction models for five different massive open online courses (MOOCs). We formulate comparable feature importance scores for each explainer, scaled between $[0,1]$ on a uniformly sampled, stratified representative set of students. To quantitatively compare the feature importance distributions, we propose the use of different measures: rank-based metrics (Spearman's rank-order correlation), distance metrics (Jensen Shannon Distance), and dimensionality analysis (Principal Component Analysis). We validate the explanations through an analysis of feature importance on a MOOC with known prerequisite relationships in the underlying curriculum. With our experiments, we address three research questions: 
\begin{enumerate}
    \item How similar are the explanations of different explainability methods for a specific course (\textbf{RQ1})?
    \item How do explanations (quantitatively) compare across courses (\textbf{RQ2})?
    \item Do explanations align with prerequisite relations in a course curriculum (\textbf{RQ3})?
\end{enumerate}

Our results demonstrate that the feature importance distributions extracted by different explainability methods for the same model and course differ significantly from each other. When comparing the feature importances across courses, we see that LIME is far apart from all other methods due to selecting a sparse feature set. Furthermore, our findings show that the choice of explainability method influences the feature importance distribution much more than the course the model is predicting on. Our examination on prerequisite relationships between features further indicates that the three families of methods are only partially able to uncover prerequisite dependencies between course weeks. Source code and models are released on Github\footnote{ \texttt{\small http://github.com/epfl-ml4ed/evaluating-explainers}}.

\section{Methodology}

\begin{figure*}[htbp]
\centering
\includegraphics[width=\linewidth, trim=0 4 0 4,clip]{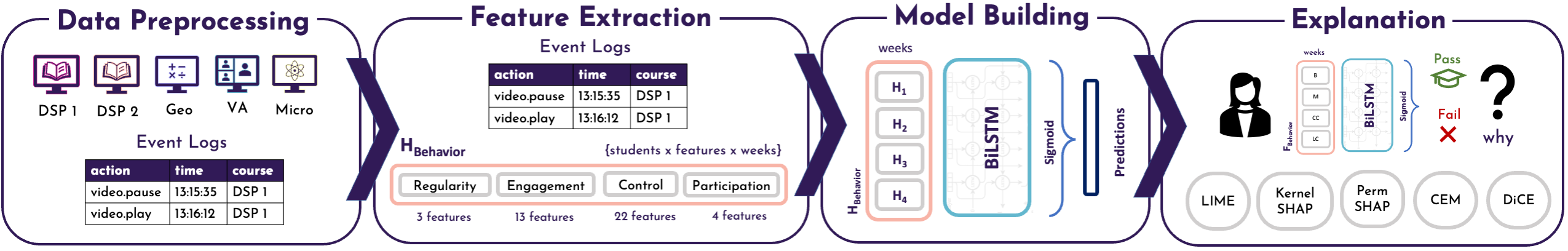}
\label{fig:pipeline}
\caption{Our experimental pipeline, from data processing to post-hoc explainability methods.}
\end{figure*}

The goal of this paper is to compare explanations from deep learning models tasked with identifying student success prediction in MOOCs. In this section, we formalize the student success prediction task addressed in this paper including the data collection and preprocessing, feature extraction, and model preparation. We then introduce the considered explainability methods and describe the process to extract explanations for student success predictions from a trained model, showcased as feature importance weights. 

\subsection{Formal Preliminaries}
We consider a set of students $S$ enrolled in a course $c$ part of an online educational offering $\mathcal C $. Course $c$ has a predefined weekly schedule consisting of $N = |\mathbb{O}|$ learning objects from a catalog $\mathbb{O}$. Students enrolled in a course interact with the learning objects included in the course schedule, generating a time-wise clickstream (e.g., a sequence of video plays and pauses, quiz submissions). We denote a clickstream in a course $c$ for a student $s \in S$ as a time series $I_s = \{i_1, \ldots, i_{K_s}\}$ with $K_s$ being the total number of interactions of student $s$ in course $c$. Each interaction $i \in I_s$ is represented by a tuple $(t, a, o)$, including a timestamp $t$, an action $a$ (videos: load, play, pause, stop, seek, speed; quiz: submit), and a learning object $o \in \mathbb{O}$ (video, quiz). Given the weekly course schedule, we assume that $t_w$ identifies the time $t$ where the course week $w \in \{0, \dots, W\}$ ends, and that the clickstream of student $s$ generated until the end of the $w$ week can be denoted as $I^{t_w}_s$. We also assume that the course schedule includes one or more assignments per week and that the grade record of student $s$ across course assignments is denoted as $G_s = [g_1, \dots, g_W]$, where $g_w \in G_s$ is the grade student $s$ received on the assignment in week $w$. In the case of multiple graded assignments for a certain week, we considered the average score of graded assignments for that week and scored non-attempted assignments with $0$. We denote as $y_s \in \{0, 1\}$ the success label for student $s$.

\subsection{Data Preprocessing}
\label{sec:log-preproc}

A significant portion of MOOC students enroll just to watch a few videos or find that the curriculum material is not what they expected and drop out of the course in the first weeks \cite{onah2014dropout, goopio2021mooc}. It follows that it is easy to predict the success labels $y_s$ for this selection of students by simply looking at their initial few weekly assignment grades in $G_s$. Therefore, optimizing complex deep learning models for predicting student success on early-dropout students is inefficient. Using these complex deep models also leads to less interpretable predictions in comparison with traditional models. For this subset of early-dropout students, traditional models can both achieve a comparable accuracy and still remain interpretable. To identify early-dropout students, we fit a \textit{Logistic Regression} model on the assignment grades of the first two course weeks. The input data is the vector $G_s^W$, where $W$ is the number of course weeks ($W=2$ in our experiments) whereas the ground truth is the student success label $y_s$. Once the model is fitted, we filter out the students that had a predicted probability of course failure $\hat{p}_{s}>0.99$. We determine the optimal threshold via a grid search over $\{0.96, 0.97, \ldots, 0.999\}$, maximizing the model balanced accuracy. Henceforth, we consider $S$ to be the student population obtained after the early-dropout student filtering. 

\subsection{Feature Extraction}
As an input for our student success prediction models, we consider a set of behavioral features extracted for each student $s \in S$ based on their interactions $I_s$. We include four feature sets proved to have high predictive power for success prediction in MOOCs \cite{marras2021can}. Given the size and variety of the course data considered in our study, we included all features of the four features sets, instead of considering only the specific features identified as important by at least one course \cite{marras2021can}. Formally, given interactions $I_s$ generated by students $S$ until a course week $w$, we create a matrix $H \subset \mathbb{R}^{|S| \times w \times f}$ (i.e, each feature in the feature set is computed per student per week), where $f \in \mathbb{N}$ is the dimensionality of the feature set. We focus on the following behavioral aspects:

\begin{itemize}[leftmargin=*,nolistsep]
    \item \textbf{Regularity} features ($H_1$, shape: $|S| \times w \times 3$) monitor the extent to which a student follows regular study habits \cite{boroujeni2016quantify}.
    \item \textbf{Engagement} features ($H_2$, shape: $|S| \times w \times 13$) monitor the extent to which a student is engaged in the course \cite{chen2020utilizing}.
    \item \textbf{Control} features ($H_3$, shape: $|S| \times w \times 22$) measure the fine-grained video consumption per student \cite{lalle2020data}.
    \item \textbf{Participation} features ($H_4$, shape: $|S| \times w \times 4$)  monitor attendance on videos/quizzes based on the schedule \cite{marras2021can}.
\end{itemize}

\vspace{2mm} We extract the above features for each student $s$ and concatenate features across sets to obtain the final combined behavioral features $h_s$ per student. The overall matrix of features is defined as $H \in \mathbb{R}^{|S| \times w \times 42}$, with $H = [H_1 \cdot H_2 \cdot H_3 \cdot H_4]$ ($\cdot$ denotes a concatenation). Due to the different scales, we perform a min-max normalization per feature in $H$ (i.e., we scale the feature between 0 and 1 considering all students and weeks for that feature). We elaborate on the most important features later on in the paper as highlighted by the analyses in subsequent experiments (e.g., Table \ref{tab:features}).

\begin{table*}[]
\resizebox{\textwidth}{!}{
\centering
\small
\begin{tabular}{@{}lll@{}}
\toprule
\textbf{Set} & \textbf{Feature} & \textbf{Description} \\ \midrule
\multirow{3}{*}{\textit{Regularity}} & DelayLecture & The average delay in viewing video lectures after they are released to students. \\ 
 & RegPeakTimeDayHour & Regularity peak based on entropy of the histogram of user's activity over time. \\  
 & RegPeriodicityM1 & The extent to which the hourly pattern of user’s activities repeats over days. \\
 \midrule
\multirow{9}{*}{\textit{Engagement}} & AvgTimeSessions & The average of users' time between subsequent sessions. \\ 
 & NumberOfSessions & The number of unique online sessions the student has participated in. \\  
 & RatioClicksWeekendDay & The ratio between the number of clicks in the weekend and the weekdays \\ 
 & StdTimeSessions & The standard deviation of users' time between subsequent sessions. \\ 
 & TotalClicksProblem & The number of clicks that a student has made on problems this week. \\
 & TotalClicksWeekend & The number of clicks that a student has made on the weekends. \\ 
 & TotalTimeProblem & The total (cumulative) time that a student has spent on problem events. \\ 
 & TotalTimeVideo & The total (cumulative) time that a student has spent on video events. \\ 
 & StdTimeBetweenSessions & The standard deviation of the time between sessions of each user. \\
 \midrule
\multirow{3}{*}{\textit{Control}} & AvgReplayedWeeklyProp & The ratio of videos replayed over the number of videos available. \\
 & AvgWatchedWeeklyProp & The ratio of videos watched over the number of videos available. \\ 
 & FrequencyEventLoad & The frequency between every Video.Load action and the following action. \\
 \midrule
\multirow{4}{*}{\textit{Participation}} & CompetencyAnticipation & The extent to which the student approaches a quiz provided in subsequent weeks. \\  
 & ContentAlignment & The number of videos for that week that have been watched by the student. \\ 
 & ContentAnticipation & The number of videos covered by the student from those that are in subsequent weeks. \\ 
 & StudentSpeed & The average time passed between two consecutive attempts for the same quiz. \\ 
 \bottomrule
\end{tabular}}
\caption{Features used in model explainability analysis. For brevity, we only list the $19$ features that have been identified as important by at least one explainability method in our analysis in Section \ref{sec:results}.}
\label{tab:features}
\end{table*}

\subsection{Model Building}
Given a course $c$, we are interested in creating a success prediction model that can accurately predict the success label $y_s$ for student $s$, given the extracted behavioral features $h_s$. To this end, we rely on a neural architecture based on \emph{Bidirectional LSTMs}, which can provide a good trade-off between effectiveness and efficiency\footnote{Experimental details can be found in Appendices A and B.}. The model input is represented by $H$, i.e., the extracted behavior features, having a shape of $|S| \times W \times 42$. NaN values were replaced with the minimum score the student can receive for each respective feature. These features are then fed into a neural architecture composed by two simple yet effective \emph{BiLSTM} layers of size 32 and 64  (loopback of $3$) and a \emph{Dense} layer (with Sigmoid activation) having a hidden size of $1$. The model outputs the probability the student will pass the course.

\subsection{Explanation}
\label{sec:explain}

Input behavioral features contribute with varying levels of importance to the prediction provided by a success prediction model. We unfortunately cannot examine the importance of these features directly, since deep neural networks act as \textbf{black boxes}. Explainability methods can therefore be adopted to approximate the contributions of each feature in $H$ towards the prediction associated with a specific student $s$. To explore this aspect, we consider five instance-based explainability methods that are popular in the literature and cover different method families \cite{linardatos2020explainable, molnar2022}. We then compute the feature importance vector for each student $s$, based on each explainability method. Formally, given an explainability method, we denote $e_s \in \mathbb R^{w*42}$ as the feature importance weights returned by the explainability method for student $s$. The feature importance weight $e_s[i]$ is a score, comparable across explainability methods, that represents the importance of feature $h_s[i]$ to the model's individual prediction for student $s$. The considered explainability methods are described below.

\textbf{LIME} \cite{lime} trains a local linear model to explain each individual student instance $h_s$. To this end, it first generates perturbed instances $h_s^{1}, h_s^{2} \ldots h_s^{n}$ by shifting the feature values of $h_s$ a small amount. These new instances are then passed to the original model to get their associated predictions. Finally, a local interpretable model (e.g., a Support Vector Machine) is trained on the perturbed instances (input) and the corresponding predictions obtained from the original model (labels), weighting perturbed instances by proximity to the original instance. Mathematically, the local model can be expressed with the following equation:
\begin{equation}
\text{LIME}(h_s) = \text{argmin}_{g'\in G'}L(g, g', \pi_{h_s})+\Omega(g')
\end{equation}
where $h_s$ is the instance being explained, $G'$ is the family of all possible explanations, $L$ the loss that measures how close the predictions of the explainer $g'$ are to the predictions of the original model $g$, $\pi_{h_s}$ is the feature proximity measure, and $\Omega(g')$ represents the complexity of the local model. As LIME returns feature weights $\pi_1 \ldots \pi_{|h_s|}$ representing the feature influence on the final decision, we consider these absolute values to be the importance scores $e_s$, and scale them to the interval $[0,1]$, where $1$ indicates high importance.
    
\textbf{KernelSHAP} \cite{shap} draws inspiration from game-theory based Shapley values (computing feature contributions to the resulting prediction) and LIME (creating locally interpretable models). This SHAP variant uses a specially-weighted local linear regression to estimate SHAP values for any model. Let $x = h_s$ be the student instance being explained. A point $x'$ in the neighborhood of $x$ is generated by first sampling a coalition vector $z \in \mathbb R^|h_s|$. The coalition vector uses a binary mask to determine which features from $x$ will be kept the same in the new instance $x'$, and which will be replaced by a random value from the data distribution of that feature in $H$. Feature importance weights for each new instance $x'$ are calculated using a predefined kernel, after which the local model can be trained. A SHAP explanation is mathematically defined as:
\begin{equation}
g'(z') = \pi_0 + \sum_{h_{s=1}}^{|h_s|} \pi_{h_s}z'_{h_s}
\end{equation}
where $g'$ is the local explainer, $\pi_{h_s}\in \mathbb{R}$ is the SHAP value (feature attribution) of feature $h_s$, and $z'\in\{0,1\}^{|h_s|}$ is the coalition binary value. To achieve Shapley compliant weighting, Lundberg et al. \cite{shap} propose the SHAP kernel:
\begin{equation}
\pi_{h_s}(z')=\frac{({|h_s|}-1)}{\left(\begin{array}{c}
{|h_s|} \\
|z^{\prime}|
\end{array}\right)\left|z^{\prime}\right|\left({|h_s|}-\left|z^{\prime}\right|\right)}
\end{equation}
where $|h_s|$ is the maximum coalition size and $|z'|$ is the number of features present in coalition instance $z'$ \cite{molnar2022}. 

SHAP methods directly provide values $\pi_{h_s}$ representing the feature contribution to the prediction $y_s$ of instance $s$. To obtain the importance scores $e_s$, we apply the same transformation as LIME, by taking the absolute values of the SHAP feature attributions and scaling them to the interval $[0,1]$.

\textbf{PermutationSHAP (PermSHAP)} \cite{shap} is very similar to the KernelSHAP formulation, but does not require the tuning of a regularization parameter or a kernel function.  We made the decision to include both KernelSHAP and PermSHAP as a form of validation of our comparative evaluation analysis; the distance between two very similar SHAP methods is expected to be smaller than the distance between these SHAP methods and other families of explainability methods. PermSHAP approximates the Shapley values of features by iterating completely through an entire permutation of the features in both forward and reverse directions (antithetic sampling). To extract the feature importance vector $e_s$, we again consider the absolute values of the SHAP feature attributions and scale to the interval $[0,1]$. 

\textbf{Contrastive Explanation Method (CEM)} \cite{cem} identifies which features need to be present (pertinent positives) or which features must be absent (pertinent negatives) in order to maintain the model prediction $y_s$ for a student $s$ with behavioral features $h_s$ \cite{cem}. For our setting, we consider pertinent negatives as they are intuitively more similar, and therefore comparable, to other counterfactual-based explainability methods. For each generated pertinent negative, we calculate the importance score for each feature by multiplying the absolute change from the value in the original instance to the value in the pertinent negative, modeled as the standard deviation ($SD$) of that feature $\Tilde{X}({h_s})$ across all instances used for the experiment ${X}({h_s})$, as shown in the following formula: 
\begin{equation}
\text{CEM}({h_s}) = [X({h_s}) - \Tilde{X}({h_s})]\times SD(h_s)
\end{equation}
The importance score therefore takes into consideration both the necessary perturbation of the feature as well as the significance of the change relative to the feature range. We normalize the scores in the range $[0,1]$, such that the resulting feature importance weights $e_s$ can be directly comparable.
    
\textbf{Diverse Counterfactual Explanations (DiCE)} \cite{dice} generates example instances to explain the model prediction as well. However, while CEM describes conditions necessary to keep the prediction unchanged, DiCE describes the smallest possible change to the initial instance that results in a different prediction. In other words, DiCE generates nearest neighbor counterfactual examples by optimizing the loss:
\begin{equation}
    \begin{aligned}
    \text{DiCE}({h_s})=\underset{\boldsymbol{c}_{1}, \ldots, \boldsymbol{c}_{k}}{\arg \min } & \frac{1}{k} \sum_{i=1}^{k} \operatorname{y_{loss}}\left(g\left(\boldsymbol{c}_{i}\right), y\right) \\
    &+\frac{\lambda_{1}}{k} \sum_{i=1}^{k} \operatorname{dist}\left(\boldsymbol{c}_{i}, {h_s}\right) \\
    &-\lambda_{2} \operatorname{ diversity}\left(\boldsymbol{c}_{1}, \ldots, \boldsymbol{c}_{k}\right)
    \end{aligned}
\end{equation}
where $\boldsymbol{c}_{i}$ is a counterfactual example, $k$ is the total number of examples to be generated, $g$ is the black box ML model, $y_{loss}$ is a metric that minimizes the distance between the prediction $g^{\prime}$ makes for $\boldsymbol{c}_{i}$ and the desired outcome $y$, $h_s$ is the original input with $|h_s|$ input features, and diversity is the Determinantal Point Process (DPP) diversity metric. $\lambda_{1}$ and $\lambda_{2}$ are hyperparameters that balance the three parts of the loss function. The stopping condition is convergence or 5000 time steps per counterfactual. Microsoft's DiCE library \cite{dice} has a built-in function to compute local feature importance scores from the counterfactual instances, scaled in $[0,1]$. We use them as feature importance weights $e_s$. 

\begin{table*}[]
\small
\resizebox{\textwidth}{!}{
\begin{tabular}{lllllrrrr}
\toprule
\textbf{Title} & \textbf{Identifier} & \textbf{Topic$^1$} & \textbf{Level} & \textbf{Language} & \multicolumn{1}{r}{\textbf{\begin{tabular}[c]{@{}c@{}}No. \\ Weeks\end{tabular}}} & \multicolumn{1}{r}{\textbf{\begin{tabular}[c]{@{}c@{}}No. \\ Students$^2$\end{tabular}}} & \multicolumn{1}{r}{\textbf{\begin{tabular}[c]{@{}c@{}}Passing \\ Rate (\%)\end{tabular}}} & \multicolumn{1}{r}{\textbf{\begin{tabular}[c]{@{}c@{}}No. \\ Quizzes\end{tabular}}} \\
\midrule
Digital Signal Processing 1 & \textit{DSP 1} & CS & Bsc & French & 10 & 5629 & 26.8 & 17 \\
% \midrule
Digital Signal Processing 2 & \textit{DSP 2} & CS & MSc & English & 10 & 4012 & 23.1 & 19 \\
% \midrule
Éléments de Géomatique & \textit{Geomatique} & Math & MSc & French & 15 & 452 & 45.1 & 27 \\
% \midrule
Villes Africaines & \textit{Villes Africaines} & SS & BSc & English & 13 & 5643 & 9.9 & 17 \\
% \midrule
Comprendre les Microcontrôleurs & \textit{Micro} & Eng & BSc & French & 13 & 4069 & 5.1 & 18 \\ 
\bottomrule
\end{tabular}}
\footnotesize{$^1$\textbf{Topic abbrev.} \textit{Eng}: Engineering; \textit{Math}: Mathematics; \textit{CS}: Computer Science; \textit{SS}: Social Science}\\
\footnotesize{$^2$\textbf{No. Students} is calculated after filtering out the early-dropout students, as detailed in Sec. \ref{sec:log-preproc}.}
\caption{Detailed information on the five MOOCs included in our experiments.}
\vspace{-7mm}
\label{tab:courses}
\end{table*}

\section{Experimental Analysis}
\label{sec:results}
We evaluated the explainability methods on five MOOCs. We first explored how feature importance varies across different explainers for one specific course $c$ (\textbf{RQ1}). We then investigated the similarity of the explainability methods across the five courses using distance metrics (\textbf{RQ2}). Finally, we assessed the validity of the explainers using simulated data from a course $c$ with a known underlying prerequisite skill structure (\textbf{RQ3}). In the following sections, we describe the dataset and optimization protocol used for the experiments before explaining each experiment in detail.

\vspace{-1mm}
\subsection{Dataset}

Our experiments are based on log data collected from five MOOCs of École Polytechnique Fédérale de Lausanne between 2013 to 2015. We chose the five courses to cover a diverse range of topic, level, and language. Table \ref{tab:courses} describes the five courses in detail. We include two subsequent iterations of the same computer science course (\textit{DSP}) with different student populations (French Bachelor students vs. English MSc students). Besides computer science, we also cover courses in the areas of mathematics (\textit{Geomatique}), social sciences (\textit{Villes Africaines}) and engineering (\textit{Micro}). In total, the raw data set contained log data from 75,992 students. After removing the early-dropout students (see Sec. \ref{sec:log-preproc}), 19,805 students remain in the data set. The smallest course contains $452$ students, while the largest course contains 5,643 students. Students' log data consists of fine-grained video (e.g., play, pause, forward, seek) and quiz events (e.g., submit). Interaction data is fully anonymized with regards to student information, respecting participants' privacy rights.

\vspace{-1mm}
\subsection{Experimental Protocol}
\label{sec:opt-proc}
For each course $c \in C$, we trained a \emph{BiLSTM} model $M_c$ on features $H_c$ extracted from $c$. For the optimization, we used batches of size $32$, an Adam optimizer with an initial learning rate of $0.001$, and a binary crossentropy loss. After an initial grid search\footnote{Grid search is discussed further in Appendix A.}, we selected the same architecture for all models: two \emph{BiLSTM} layers consisting of {64, 32} units and one Dense layer consisting of {1} unit with a Sigmoid activation. As this work is not focused on improving model performance, we did not tune hyperparameters further. Formally, we split the data of each course $c$ into a training data set $S_{train,c}$ ($80\%$ of the students) and a test data set $S_{test,c}$  ($20\%$ of the students). For each course, we performed a stratified train-test split over students' pass/fail label. We then trained each model $M_c$ on the training data set $S_{train,c}$ and then predicted student success on the respective test data set $S_{test,c}$. We chose the balanced accuracy (BAC) as our primary evaluation metric because of the high class imbalance of most of the selected courses.

For the first two experiments (\textbf{RQ1} and \textbf{RQ2}) we used the student log data collected for the full duration of the course for training and prediction of our models. In the third experiment, we optimized models for different sequence lengths, i.e. using only the log data up to a specific week $w$ of the course (i.e. from week $1$ to week $w$) to predict performance in the assignment of course week $w$. Additional replication details for model training can be found in Appendix B.

For all experiments, we applied the explainability methods to the predictions of the optimized models $M_c$. All five methods are instance-based; they compute the feature importance based on the model predictions for a specific instance. Training explainers on the scale of thousands of students across five courses is not feasible due to the computation time required to generate the explanation for one instance (e.g., the counterfactual explainability methods take a computation time of 30 minutes per instance $s$). Therefore, we determined a representative sampling strategy to pick $100$ students from each course $c$, resulting in explanations for $500$ students in total\footnote{Sampling strategy is discussed further in Appendix C.}. For the first two experiments (\textbf{RQ1} and \textbf{RQ2}), we used a uniform sampling strategy to select the representative students $s^{r}_{i,c}$ for a course $c$ and ensured balance between classes (pass/fail). We first extracted all failing students and ordered them according to the predicted probability of the model $\hat{p}(l_{S_i}=0)$. We then uniformly sampled $50$ failing students from this ordered interval. We repeated this exact same procedure to sample the $50$ passing students. This sampling procedure ensures that we include instances where the model is confident and wrong, instances for which the model is unsure, and instances where the model is confident and correct. For the last experiment (\textbf{RQ3}), we used performance in the assignment of a given week $w$ as the binary outcome variable. We then followed exactly the same uniform sampling procedure as for \textbf{RQ1} and \textbf{RQ2}, ensuring class balance on assignment performance.

\begin{figure*}[t!]
\centering
\includegraphics[width=\linewidth, trim=4 4 4 4,clip]{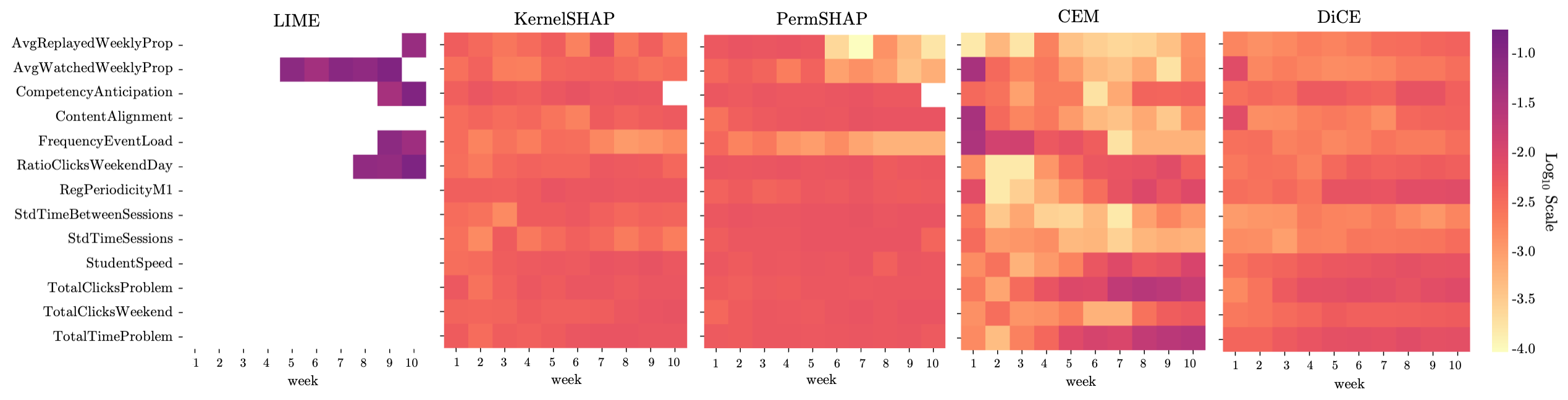}
\caption{Heatmap of normalized feature importance scores (log scale) across explainability methods for \textit{DSP 1}.}
\label{fig:feature_importance_heatmaps}
\end{figure*}

\vspace{-2mm}
\subsection{RQ1: Explanations for one course}
\label{subsec:rq1}
In a first experiment, we compared the explanations of the instance-based methods for one specific course (\textit{DSP 1}). The \emph{BiLSTM} model $M_{DSP1}$ trained on this course achieved a BAC of $97.4\%$. We then ran the explainability methods % (LIME, KernelSHAP, PermSHAP, DiCE, CEM) 
on $M_{DSP1}$ and extracted normalized feature importance scores for $100$ representative students of each course.

Figure \ref{fig:feature_importance_heatmaps} illustrates the features identified as most important by each explainability method. The heatmaps were computed by averaging importance scores for each feature and week across $100$ representative students for \textit{DSP 1} (see Sec. \ref{sec:opt-proc}). To ensure interpretability of Figure \ref{fig:feature_importance_heatmaps}, we only included the top five features for each method, resulting in $13$ distinct features. The description of all the features can be found in Table \ref{tab:features}. We used a log scale within the heatmaps, with darker colors indicating higher feature importance. 

We observe that the top features cover all the different behavioral aspects included in the feature set: Regularity, Engagement, Control, and Participation. However, some aspects seem to contain more important features. For example, $43\%$ of the Participation features ($3$ out of $7$ features) are in the top five features of at least one method, while this is the case for only $23\%$ ($5$ out of $22$ features) of the Control features. For Regularity and Engagement, $33\%$ and $31\%$ of the features get selected into the top feature set.

We also immediately recognize that the heatmap of LIME looks very different from the heatmaps of all the other methods. LIME assigned high importance scores to a small subset of features and weeks, while all the other explainability methods tend to identify more features and weeks as important, resulting in generally lower importance scores. We also observe that LIME does not consider student behavior in the first weeks of the course important; all importance is placed onto the second half of the course. Moreover, LIME seems to put more emphasis on Control than on the other three aspects: the features related to Control (\textit{AvgReplayedWeeklyProp}, \textit{AvgWatchedWeeklyProp}, \textit{FrequencyEventLoad}) are important from week $5$ through week $10$, while the features related to Participation (\textit{CompetencyAnticipation}) and Engagement (\textit{RatioClicksWeekendDay}) are important only during the last $2$ to $3$ weeks of the course.

Interestingly, while CEM and DiCE are both counterfactual methods, their heatmaps look quite different: the feature importance scores of DiCE tend to be more similar to KernelSHAP and PermSHAP than CEM. We note that CEM shows a higher diversity in feature importance scores than the other three methods (KernelSHAP, PermSHAP, and DiCE), for which the importance values seem to be quite equally distributed across the top features. Furthermore, in contrast to all the other explainability methods, CEM seems to also identify features in the first weeks of the course as important (e.g., \textit{AvgWatchedWeeklyProp}, \textit{ContentAlignment}, and \textit{FrequencyEventLoad} in week $1$). In contrast to all the other methods, CEM identifies features related to being engaged in quizzes as relevant (\textit{TotalClicksProblem} and \textit{TotalClicksWeekend}). Finally, as expected, the heatmaps of KernelSHAP and PermSHAP look very similar, with only small differences in importance scores.

\begin{graybox} In summary, while there is some agreement on the top features across explainability methods (the union of the top five features of each method only contains $13$ distinct features), we observe differences across methods when it comes to exact importance scores. \end{graybox}

\vspace{-1mm}
\subsection{RQ2: Comparing methods across courses}
\label{sec:cross_examine}
Our second analysis had the goal to \emph{quantitatively} compare the explanations of the different methods across all five courses. Explainability method evaluation is an emerging field; most existing research focused on assessing the quality of explanations \cite{sokol2020explainability, leake2014evaluating} with only few works suggesting a quantitative `goodness' score for each explainability method (e.g., \cite{nguyen2020quantitative, zhou2021evaluating}). In contrast, we examined the distance between the feature importance scores per explainability method in comparison to each other, instead of individually. We first visualized the similarity of importances across courses using a Principal Component Analysis and then computed Spearman's Rank-Order Correlation as well as Jensen-Shannon Distance to assess similarity regarding the feature importance ranking as well as their exact values.

\begin{figure}[!b]
\centering
\includegraphics[width=0.5\linewidth, trim=4 4 4 4,clip]{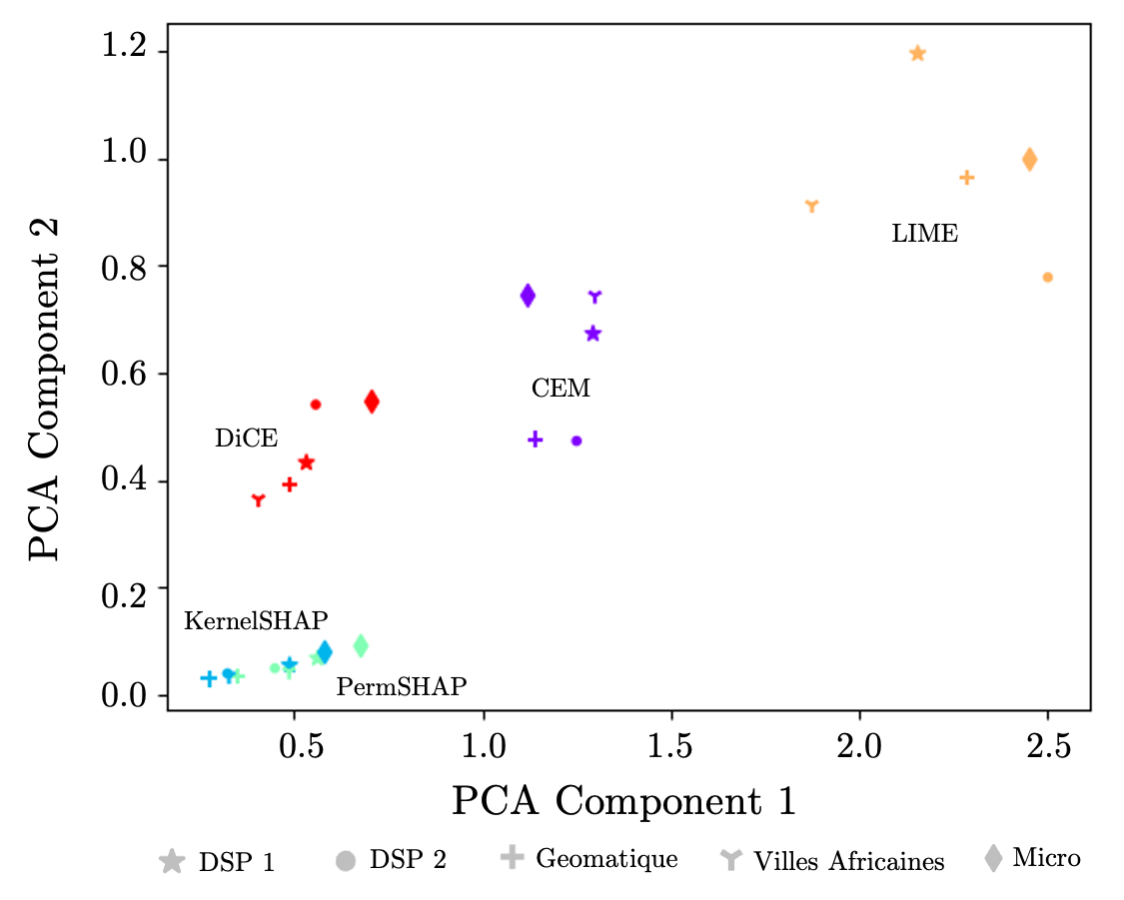}
\caption{PCA of feature importance scores for five explainability methods across five courses.}
\label{fig:pca_all}
\end{figure}

\vspace{1mm} \noindent \textbf{Principal Component Analysis (PCA)} We performed a PCA on the importance scores for each feature and week (length: $w_{c}\times h$) separately for each explainability method and course $c$. Figure \ref{fig:pca_all} shows the results for all explainability methods and courses. Each marker in Figure \ref{fig:pca_all} represents a specific course, while each color denotes an explainability method.

\begin{figure*}[htbp]
\centering
\includegraphics[width=\linewidth, trim=4 4 4 4,clip]{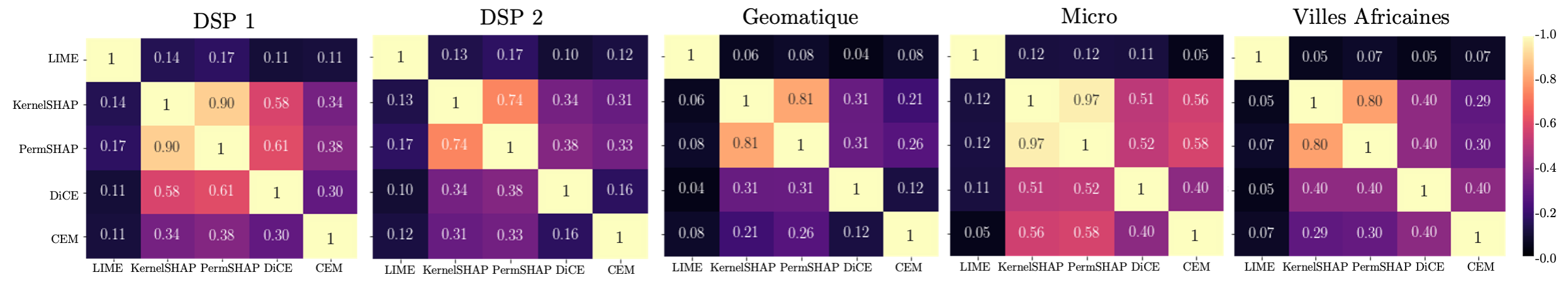}
\caption{Comparison of feature importance scores across courses using Spearman's Rank-Order Correlation.}
\label{fig:spearman}
\end{figure*}

\begin{figure*}[t]
\centering
\includegraphics[width=\linewidth, trim=4 4 4 4,clip]{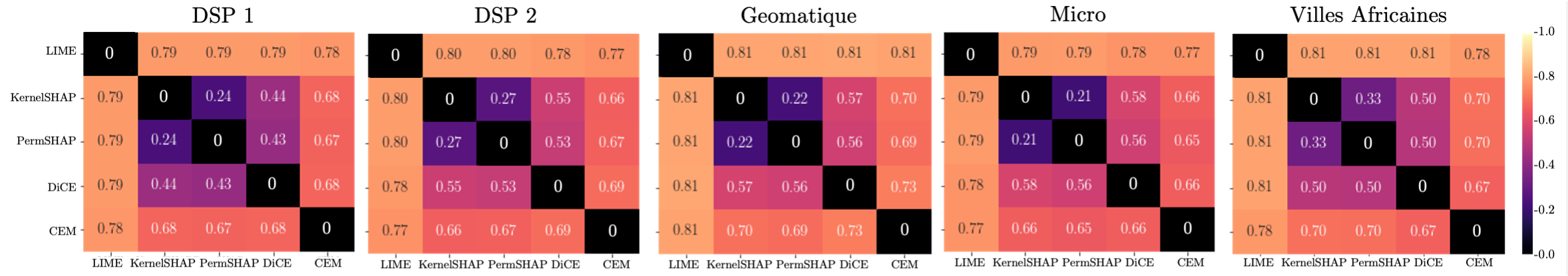}
\caption{Comparison of feature importance scores across courses using Jensen-Shannon Distance.}
\label{fig:jensen_shannon}
\end{figure*}

We observe that the two SHAP methods (KernelSHAP and PermSHAP) cluster together very strongly. This result is expected, as the methodologies of KernelSHAP and PermSHAP are very similar. DiCE feature importances are quite close to the SHAP methods, showing that the three methods have similar notions of feature importance. LIME is quite different from all other methods, with high values on both PCA components. Based on differences in methodology, we would have expected that the difference between the counterfactual methods (DiCE and CEM) and the SHAP methods would be larger than the difference between LIME and the SHAP methods. The most notable takeaway from Figure \ref{fig:pca_all} is that there are clearly identifiable clusters based on explainability method and not on course. It therefore seems that the resulting feature importance scores are mainly influenced by the explainability rather than by the model or data (i.e. the characteristics of the course and students' data).

\vspace{1mm} \noindent \textbf{Spearman's Rank-Order Correlation.} Often referred to as Spearman's $\rho$ \cite{spearman1961proof}, this metric identifies the rank correlation (statistical dependence between the rankings) between two variables and is defined as the Pearson correlation coefficient between the rankings of two variables. We chose this metric for evaluating explainability methods to highlight the importance of feature ranking order in explanations.
To compute Spearman's Rank-Order Correlation $r_{m_1,m_2,c}$ between two explainability methods $m_1$ and $m_2$ on a course $c$, we first converted the vectors $e_{m_1,s}$ and $e_{m_2,s}$ of feature importance scores (length $w_c \times h$) for each student $s$ into rankings $R(e_{m_1,s})$ and $R(e_{m_1,s})$. We then computed $r^{s}_{m_1,m_2,c}$ separately for each relevant student $s$ and then averaged over all relevant students to obtain $r_{m_1,m_2,c}$. 

Figure \ref{fig:spearman} illustrates the pairwise similarities between explainability methods using Spearman's Rank-Order Correlation. Higher values imply stronger correlation between methods. We see similarities between KernelSHAP and PermSHAP prevalent once again as a center square for each course, affirming our intuition that two similar methodologies would result in similar rank-order scores. It can be observed that LIME consistently shows low rank-order correlation scores with all other explainability methods. Additionally, for \textit{DSP 1}, DiCE is much closer to KernelSHAP and PermSHAP than to CEM. For \textit{DSP 2} and \textit{Geomatique}, DiCE and CEM are both equally correlated to the SHAP methods, but less correlated among themselves. Finally, the model trained on \textit{Micro} has strong correlations across all explainability methods except LIME.

\vspace{1mm} \noindent \textbf{Jensen-Shannon Distance.} We used the Jensen-Shannon distance \cite{jensenshannondivergence} to compute pairwise distances between exact feature importance score distributions obtained with different explainability methods. The Jensen–Shannon distance is the square root of the Jensen-Shannon divergence, originally based on the Kullback–Leibler divergence with smoothed values. It is also known as the Information Radius (IRad) \cite{manning1999foundations}. To compute the Jensen-Shannon distance $jsd_{m_1,m_2,c}$ between two explainability methods $m_1$ and $m_2$ on a course $c$, we first calculated the distance $jsd^{s}_{m_1,m_2,c}$ between the feature importance scores (length $w_c \times h$) $e_{m_1,s}$ and $e_{m_2,s}$ separately for each representative student $s$ and then averaged across all representative students to obtain $jsd_{m_1,m_2,c}$.

Figure \ref{fig:jensen_shannon} shows the pairwise distance between explainability methods for all courses using Jensen-Shannon Distance. Larger numbers represent higher dissimilarity. The Jensen-Shannon Distance heatmaps confirm the observations made using Spearman's Rank-Order Correlation (see Figure \ref{fig:spearman}). Again, LIME consistently has a high distance to all other explainability methods across all courses. As expected, KernelSHAP and PermSHAP have low pairwise distances for all courses. However, when comparing feature importance scores directly instead of using rankings, we observe even less differences between courses. DiCE is closer across all courses to the SHAP methods than CEM. While LIME exhibits the highest distances to all other explainability methods, the explanations of CEM are also far away from all methods. 

\begin{graybox} In summary, the two SHAP methods and DiCE seem to deliver the most similar explanations, while the feature importance scores obtained with CEM and LIME are different from the other explainability methods. More importantly, all our analyses (PCA, Spearman's Rank-Order Correlation, Jensen Shannon Distance) demonstrate that the choice of explainability method has a much larger influence on the obtained feature importance score than the underlying model and data. \end{graybox}

\vspace{-1mm}
\subsection{RQ3: Validation of explanations}

\begin{figure}[]
\centering
\includegraphics[width=\linewidth, trim=4 4 4 4,clip]{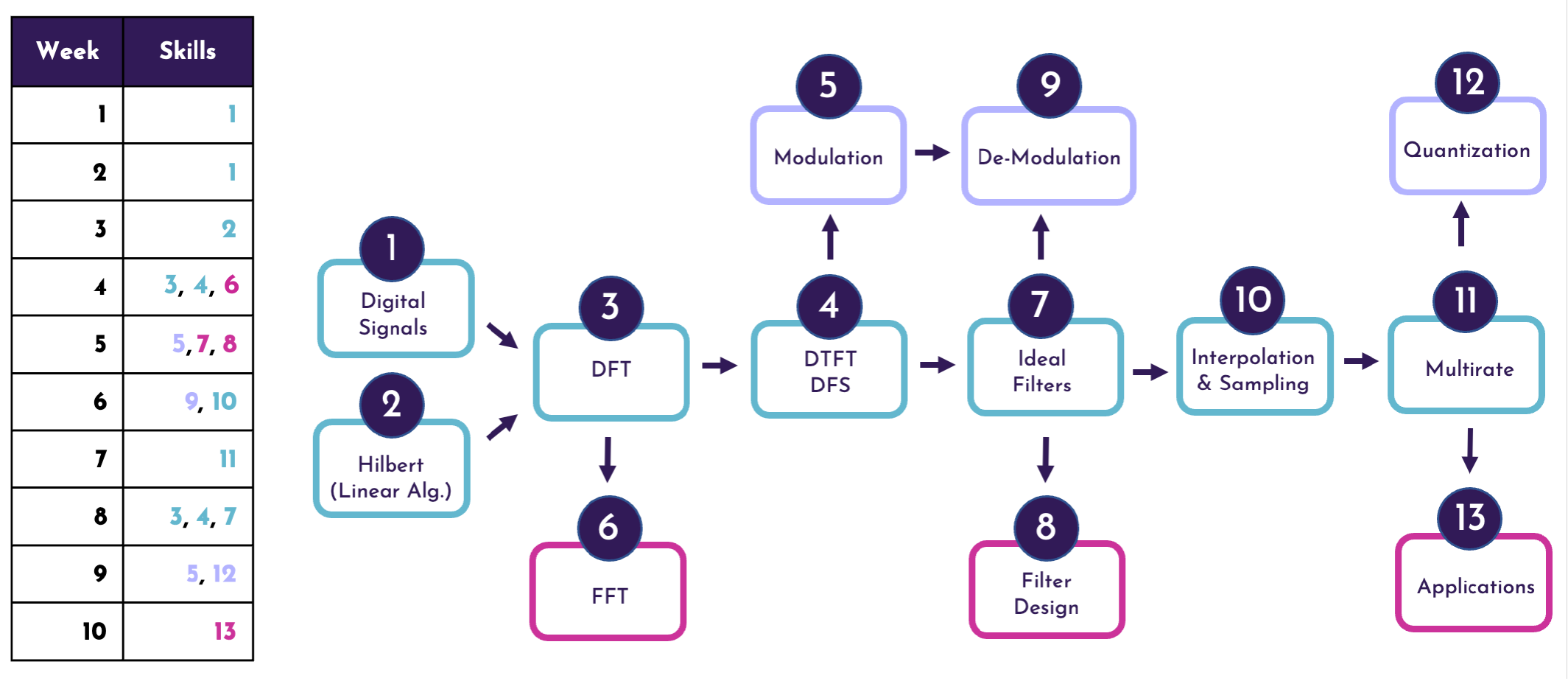}
\caption{Prerequisite skill structure for \textit{DSP 1}.}
\label{fig:skillmap}
\end{figure}

The previous experiments delved into mapping the similarities and differences between explainability methods. While our analyses demonstrated that there are clear disparities across method choice, they do not give an indication regarding the `goodness' of the obtained explanations. Recent work discusses traits of ideal explanations \cite{carvalho2019machine} and targets metrics to measure explanation quality \cite{zhou2021evaluating, jia2021studying}. However, these metrics tend to be over-specialized to one explainability method over others due to the similarities in their methodologies. The community does not yet have a set of standard metrics for evaluating explainability methods. In our last experiment, we hence use information inherent to our model's setting to perform an initial validation of the explanations provided by the different methods.

Specifically, we evaluated the explainability methods on a course with a known underlying skill map and used the prerequisite relationships between weeks of the course as a ground truth for the explanations. Based on the results obtained for the first two research questions (Sections \ref{subsec:rq1} and \ref{sec:cross_examine}), we selected one representative method from each methodological group for the analysis, while keeping the observed explanation diversity: PermSHAP (the most widely used SHAP method), CEM (chosen as a representative of counterfactuals for its further disparity from the SHAP methods), and LIME. In terms of courses, we used \textit{DSP 1} as a basis for the analysis as the instructor of this course provided us with the skill map derived from the curriculum.

Figure \ref{fig:skillmap} illustrates the underlying skills, their relationships, as well as their mapping to the weeks of the course. The arrows denote the prerequisite relationships, while the numbers denote the unique skills in the order they are introduced in the course. The skills colored in pink ($6$, $8$, and $13$) refer to applied skills learned in the course. The middle track refers to core skills learned in the course (colored in blue) and the purple skills at the top ($5$, $9$, and $11$) are theory-based extensions of core material. The skill prerequisite map allows us to analyze the dependencies between the different weeks of the course. For example, in order to understand \textit{Modulation} taught in week $5$, students need to already have learned the skills taught in weeks $3$ and $4$ (\textit{DFT}, \textit{DTFT}, and \textit{DFS}). Intuitively, a model predicting performances in assignments in week $5$ would have highly correlated features based on week $4$. We assume that these dependencies would logically be uncovered by the explainability method.

We, therefore, adjusted our predictive task: using the optimization protocol and experimental design described in Section \ref{sec:opt-proc}, we aimed at predicting the performance (binary label: below average or above average) of a student $s$ in the assignment of week $w$ based on features extracted from student interactions for weeks $1$ to $w$. Given the prerequisite structure for the course, we ran experiments for $w \in \{5, 9\}$. For each predictive model $M_{DSP1,w}$, we then picked the $100$ representative students using uniform sampling and taking into account class balance (see Section \ref{sec:opt-proc} for a detailed description of the sampling procedure) and applied the selected explainability methods to these representative instances.

\begin{figure}[ht]
\centering
\includegraphics[width=1.05\linewidth, trim=4 4 4 4,clip]{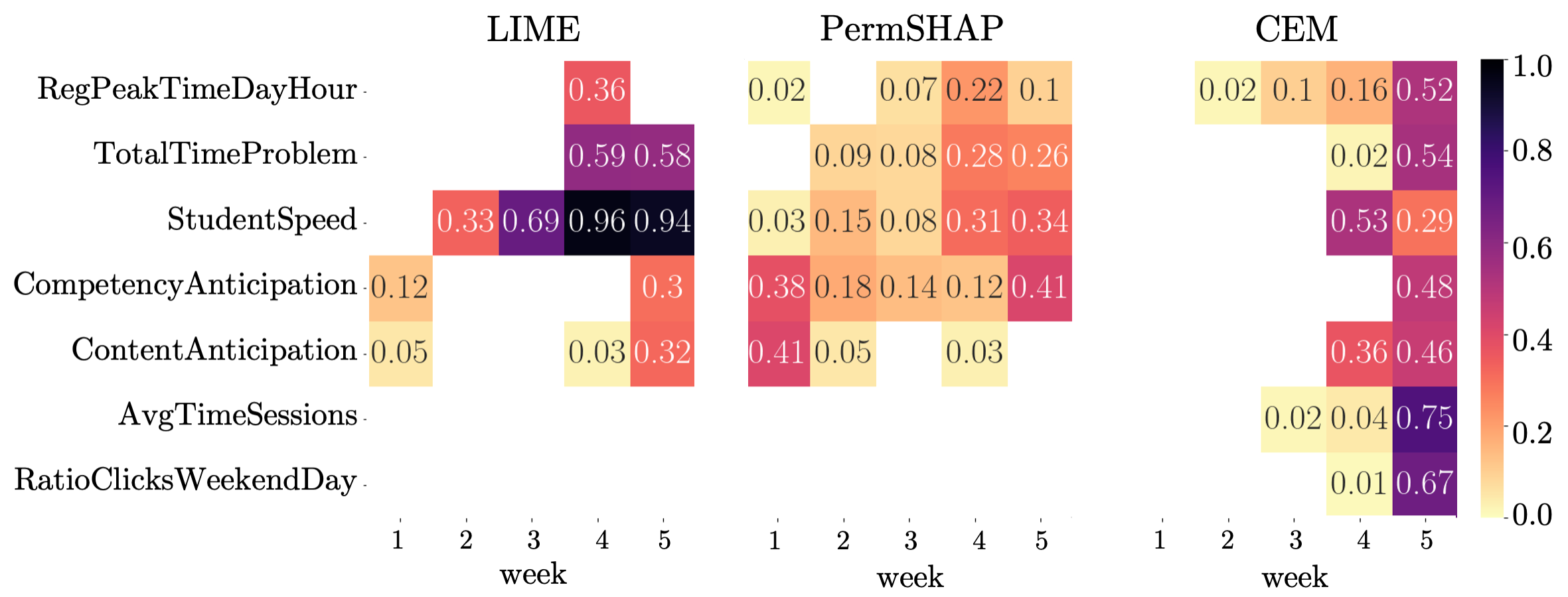}
\caption{Importance scores for LIME, PermSHAP, and CEM for week $5$.}
\label{fig:week5_sim}
\end{figure}

Figure \ref{fig:week5_sim} shows the features for LIME, PermSHAP, and CEM for the prediction model of week $5$. In the heatmap, darker values indicate a higher score. The scores for the heatmap have been computed based on a ranking of features and weeks: for each student $s^{r}_{i}$, we first ranked the features in order of feature importance as determined by the respective explainability method. We then scored each of the top $10$ features according to its rank: $10$ points for the top feature, $9$ points for the second most important feature, and so on. Finally, we averaged the scores for each feature across the $100$ representative students $s^{r}_{1}$ through $s^{r}_{100}$ and normalized them. This rank-based scoring allows us to compare explainability methods without having the relative feature importance scores bias the analysis. We only selected features with a score of at least $0.33$ in any course week, showing only the top two-thirds of features per method.

\begin{figure*}[ht]
\centering
\includegraphics[width=\linewidth, trim=4 4 4 4,clip]{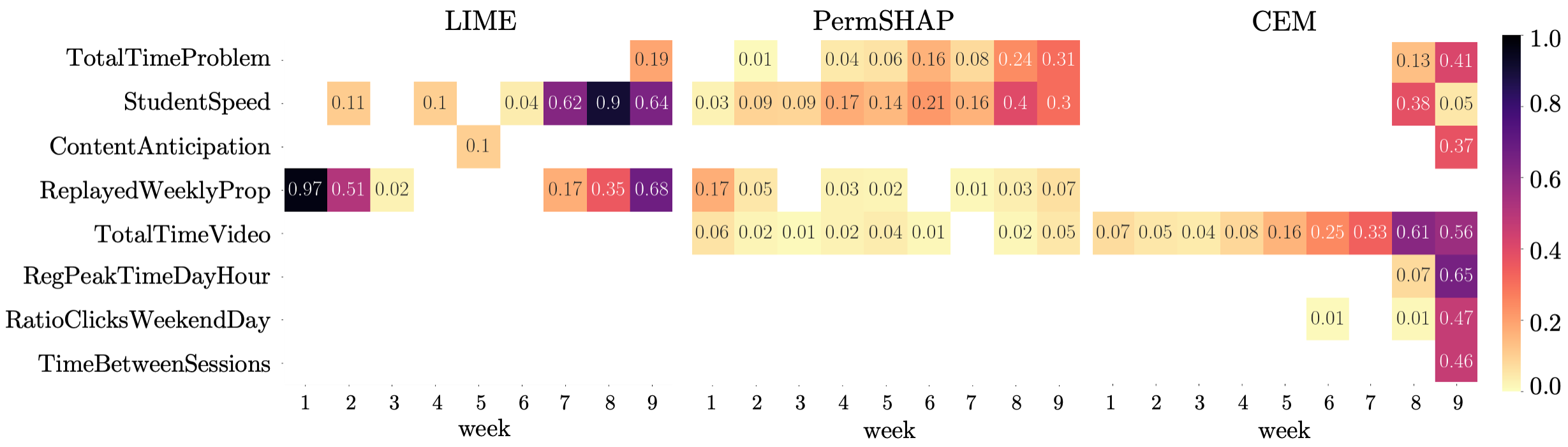}
\caption{Importance scores for LIME, PermSHAP, and CEM for week $9$.}
\label{fig:week9_sim}
\end{figure*}

For LIME, we observe that four features have been identified as important for predicting assignment performance in week $5$. Two of these features directly relate to student behavior on the week $5$ assignment: \textit{TotalTimeProblem} (the total time the student spent solving the assignment) and \textit{StudentSpeed} (the time between consecutive trials of assignments). The high scores of these features for week $5$ are thus expected. LIME also assigns high scores to these two assignments based features for week $4$. This result is encouraging as week $4$ teaches the prerequisite skills of week $5$ and therefore, LIME seems to (at least partially) uncover the prerequisite structure of the course. 

In PermSHAP results, we see again that the scores are more uniformly distributed across features and weeks. However, we observe again that the assignment-based features (\textit{TotalTimeProblem} and \textit{StudentSpeed}) have comparably high scores for week $4$ and therefore PermSHAP also seems to (partially) uncover the prerequisite relationships. Curiously, watching videos and solving quizzes scheduled for subsequent weeks (\textit{CompetencyAnticipation} and \textit{ContentAnticipation}) is also considered important, hinting that being proactive when learning increases learning success.

CEM seems to be able to partially uncover the prerequisite relationship between weeks as well. For week $4$, one feature related to assignment behavior (\textit{StudentSpeed}) exhibits a high score. Additionally, watching content of the subsequent week (in this case, week $4$'s \textit{ContentAnticipation} for week $5$ material) is important for assignment performance. Otherwise, CEM mainly explains performance in the assignment of week $5$ with student behavior in week $5$: besides the assignment-related features, the students' actions in the inference week are considered important.

Figure \ref{fig:week9_sim} shows the importance scores for LIME, PermSHAP, and CEM for the prediction model of week $9$. Again, darker colors in the heatmaps indicate higher scores. The scores in the heatmaps were computed using the same ranking-based procedure as for Figure \ref{fig:week5_sim}. In week $9$, Figure \ref{fig:skillmap} indicates that weeks $5$ and $6$ cover the prerequisite skills. From Figure \ref{fig:week9_sim}, we observe that LIME does not seem to be able to capture this prerequisite relationship. The top scores for LIME appear in week $9$ itself for an assignment-based feature (\textit{StudentSpeed}) as well as for video control behavior (\textit{AvgReplayedWeeklyProp}, which computes the relative number of video replays). Furthermore, \textit{StudentSpeed} seems to be generally important also in the weeks just before the predicted week (weeks $7$ and $8$). For PermSHAP, we again obtain a more equal distribution, with only the two assignment-related features (\textit{TotalTimeProblem} and \textit{StudentSpeed}) in weeks $8$ and $9$ showing relatively higher scores. Furthermore, PermSHAP also assigns relatively higher importance to these features for week $6$, which is a prerequisite for week $9$. For CEM, we again observe that mainly student behavior in the actual week, i.e. week $9$, seems to explain assignment performance. Only one feature (\textit{TotalTimeVideo}) shows medium importance for weeks $5$ and $6$.

\begin{graybox} In summary, all the evaluated methods were able to (partially) detect the prerequisite relationship between week $4$ and week $5$. For week $9$, detecting the prerequisite structure proved to be difficult; results differed between methods. However, we should take into account that none of the features of the feature set directly measure student performance and therefore, the generated explanations rely on behavioral features only. It appears that recent and actual behavior is a much stronger indicator for performance than past behavior. \end{graybox}

\section{Discussion and Conclusion}

Explainability methods allow us to interpret a deep model in a way that is understandable not only to machine learning experts, but also end-users of educational environments, including instructors tailoring course designs and students the model is predicting on \cite{webb2021machine, conati2018ai}. In this paper, we aimed to understand explainers' behaviour and the ways in which they differ for the task of student success prediction. 

Our results demonstrate that all explainability methods can derive interpretable motivations behind student success predictions, confirming the similar yet coherent observations made by \cite{lu2020towards} for the knowledge tracing field. However, while there was some agreement regarding the top features across the five explainability methods, key differences across methods emerged when we considered the exact importance scores (\textbf{RQ1}). We observed substantial similarities between KernelSHAP, PermSHAP, CEM and DiCE with regards to the top ranked feature-weeks. Conversely, LIME only ranked very few features as important, and these less important feature similarities made the other explainability methods appear closer to each other. Overall, looking beyond top ranked features, we noted considerable differences in feature importances across explainability methods. Interestingly, LIME-detected features are more in line with the features marked as important by Random Forests in \cite{marras2021can}, still in a MOOC context. This observation further demonstrates the generalizability of the features' predictive power even among very different experimental settings.

In a subsequent experiment, we compared the different explainability methods across five MOOCs. Our findings indicate that the choice of explainability method has a much larger influence on the obtained feature importance score than the underlying model and data (\textbf{RQ2}). With distance (Jensen-Shannon distance) and ranking-based metrics (Spearman's Rank-Order Correlation), we uncovered that LIME is farthest from the other explainability methods. The sparsity of LIME-detected important features was also observed by \cite{scheers2021interactive}, where the conciseness of LIME explanations supported integration in visual dashboards for student advising. We also detected a close relationship between KernelSHAP and PermSHAP, which strongly validates our evaluation strategy. Using PCA, we identified clear clusters of explanations by explainability method and not by the course the model was trained upon, suggesting that an explainability method might be prone to mark specific features as important regardless of the model (and the course).

Our analyses also confirmed that all the evaluated methods were able to (partially) detect the prerequisite relationship between weeks, while relying on behavioral features only (\textbf{RQ3}). Our experimental design was inspired by \cite{mu2020towards}'s work on predicting effectiveness of interventions for wheel-spinning students by simulating prerequisite relationships. While we have no way to examine the true underlying feature importances of our week $n$ assignment performance prediction model, we intuit that a student's prerequisite week performance should be important to predicting their performance in week $n$. We observed that the three families of methods (LIME, SHAP, and counterfactual) were able to partially capture the prerequisite relationship in week $5$, but struggled to capture the prerequisite relationships in week $9$. While there were few similarities in the top ranked features, each method found different groups of feature-weeks as most important for the same models. Our results indicate that recent and current behavior is more important than past behavior, implying that proximity of behavioural features correlates strongly with their perceived importance. A limitation is that the prerequisite relationships we deem important might not actually be used as the true features of the model since our feature set included only features that examined student behavior and not direct performance.

Our results indicate that there are noteworthy differences in generated explanations for student performance prediction models. However, our analyses also show that these explanations often recognize prerequisite-based relationships between features. That being said, our study still has several limitations that warrant future research, including our focus on a singular downstream task (student success prediction), specific modality of dataset (MOOCs), choice of model architecture (\textit{BiLSTM}), and lack of assessment of the obtained explanations' impact in the real world. First, extending our experiments beyond success prediction to a multi-task analysis (e.g., dropout prediction) across multiple modalities (e.g., flipped classrooms, intelligent tutoring systems) would allow us to build stronger intuitions about explainability method differences. Second, extending our black-box BiLSTM model architecture to multiple traditional and deep machine learning architectures could examine whether certain explainability methods have stronger explanation affinity to different predictors. Choosing transparent shallow architectures instead of black-boxes could also allow us to validate our results against ground truth feature importances. Third, further research should be conducted to check which explanations (and explainability methods) lead to interventions that better improve learning outcomes. It follows that an assessment of the obtained explanations should be carried out involving educators. Finally, the disagreement of our selected explainability methods motivate an extension in an ensemble expert-weighting scheme, which might have merit in closely estimating the true feature importances.

Explainability in educational deep learning models can lead to better-informed personalized interventions \cite{xing2019dropout, karimi2021algorithmic}, curriculum personalization, and informed course design. If we were considering global interventions (as it might be too resource intensive to perform interventions on each student individually), we could take the mean feature importance vector over all students and try interventions in the order of the scores of this mean vector. If we were only able to intervene on $k$ features due to resource constraints, Spearman's rank-order metric could also be modified to include the size of the intersection between the features with the top $k$ scores. However, it is important to note that when the model explanations are biased by explainability method and do not accurately reflect the inner workings of the model, the impact of incorrect predictions are further exacerbated by teachers and students' misplaced confidence in the model's justification. \textbf{We implore data scientists to not take the choice of explainability method lightly as it does have a significant impact on model interpretation, and instead urge the community to (1) carefully select an appropriate explainability method based on a downstream task and (2) keep potential biases of the explainer in mind when analyzing interpretability results.} Overall, our work contributes to ongoing research in explainable analytics and to the generalization of theories and patterns in success prediction.

% The following two commands are all you need in the
% initial runs of your .tex file to
% produce the bibliography for the citations in your paper.
% You must have a proper ".bib" file
%  and remember to run:
% latex bibtex latex latex
% to resolve all references
%
%APPENDICES are optional

%% file: main/05_trusting_explainers.tex
% There is a compelling need for interpretability in models dealing with human data, especially in education. \cite{conati2018ai} strongly argues for interpretable models in education, especially in settings where students can see the effect of a decision but not the reasoning behind it (e.g., open learner modeling). Furthermore, \cite{DBLP:conf/lak/NazaretskyCA22} has found that transparency is an essential factor in increasing educators' trust in AI-based educational technology. \cite{webb2021machine} emphasizes that explainability and accountability should be incorporated in machine learning system design to meet social, ethical and legislative requirements.

%Other work strongly argues for interpretable models in education, specifically in settings where students can see the effect of a decision but not the reasoning behind it (e.g., open learner modeling) \cite{conati2018ai}. Predictions of student performance are often used to determine underachieving students for targeted downstream interventions. Therefore, identifying important features motivating failure or dropout predictions is crucial in designing effective, personalized interventions.

% The explainable machine learning for education community is only now emerging with few works over the last two years, most of which focus on implementing only one interpretability method for a downstream educational task, using either LIME or SHAP (e.g., \cite{hasib2022lime, baranyi2020interpretable, scheers2021interactive, mu2020towards, pei2021}). 

This work was originally presented at the 13th International Learning Analytics and Knowledge Conference, where it received a best paper nomination under the title \textit{Trusting the Explainers: Teacher Validation of Explainable Artificial Intelligence for Course Design}. The research presented in this chapter is a collaboration with Sijia (Skye) Du, Professor Mirko Marras, and Professor Tanja Käser.

We thank Kate Kutsenok, Christian Giang, Jibril Frej, Peter Buhlman, Tanya Nazaretzky, as well as our 26 study participants for their generous time and support towards this project.

\section{Introduction}

This chapter aims to address the research gap in \textit{validating} which explainable AI methods are more suited or trustworthy for human-centric tasks. In the previous chapter, we examined and compared five popular instance-based explainability methods on student success prediction models for five different massive open online courses (MOOCs). The previous experiments show that the feature importance distributions extracted by different explainability methods for the same model and underlying students differ significantly from each other. \cite{marras2021can} focused on how the feature importance differs across different course modalities (MOOCs and flipped classrooms) using random forest models. Their work demonstrates that the ensemble of optimal features varies depending on the the course design and the characteristics of the population. Inspired by these works, we are interested in how the explanations of success prediction models from courses with contrasting educational aspects (in terms of instructional design and target audience) vary from each other. We aim to extend Chapter 4's result \cite{swamy2022evaluating}, going beyond ``are the explanations generated by the methods systematically different?" towards ``which method is most valid for the downstream application?". To estimate this, we obtain the closest approximation of ground truth through semi-structured expert interviews with educators, aiming to understand which XAI methods the experts trust most and how these insights can be actionable for course design.

Specifically, we select pairs of courses with different educational factors: four pairs with differences in instructional design in terms of setting (MOOC vs. flipped classrooms), active learning (high vs. low), and optimality of design for the target audience (high vs. low passing rate) and one pair with differences in student population, measured by the instruction language (English vs. French). While learning science literature points to many other significant factors influencing students' learning behavior and success in online and blended courses, we limited ourselves to the factors measurable in our dataset based on the information about the curriculum and demographics accessible to us. We first train models of student success prediction separately for each course using the bidirectional LSTM (BiLSTM) network proposed in Chapter 3. We then apply two popular instance-based explainability methods \cite{lime, shap} to each model. We formulate comparable feature importance scores for each explainer. To contrast the feature importance distributions quantitatively, we use different distance measures comparing feature ranks (\textit{Spearman’s Rank-Order Correlation}) and similarity of the top ten features (\textit{Jaccard Similarity}).
For two course pairs (setting and active learning), we conduct expert interviews with $26$ STEM professors to build validation and trust in the generated explanations. Participants are asked to rank the features they find important for differentiating passing and failing student behavior, identify the explanations they trust the most (with the underlying method obfuscated), and brainstorm actionable course design changes from these insights. Specifically, in our study, we aim to answer the following research questions:
\begin{enumerate}
\small
    % \vspace{-1mm}
    \item How similar are the explanations of different explainability methods for a specific course?% (across a set of nine diverse courses)?
    \item Do explanations capture a controlled difference across pairs of courses with different settings?
    \item Which explainability methods do domain experts trust most?
    \item Can experts use XAI insights for actionable course design decisions?
    % \vspace{-1mm}
\end{enumerate}

We provide results from a nine-course quantitative analysis and a semi-structured expert interview study regarding the trustworthiness of the LIME and SHAP explainers. Our quantitative results show that the explainers do not agree with each other about important features on any of the individual course models. From a qualitative lens, our expert interviews reveal that educators do not consistently prefer any one explainer and often even choose a confounder explanation. Although our study reveals a lack of trust in explainer agreement, over $85\%$ of educators are able to use these insights to generate actionable decisions with the goal of improving student learning outcomes.

%%%%%% METHOD SECTION %%%%%%%%%%%
\section{Methodology}
\label{sec:method}
%%%%%%%%%%%%%%%%%%%%%%%%%%%%%%%%

\begin{figure*} %here, bottom of page, top of page, empty page in order of preference
     \centering
     \includegraphics[width=\linewidth]{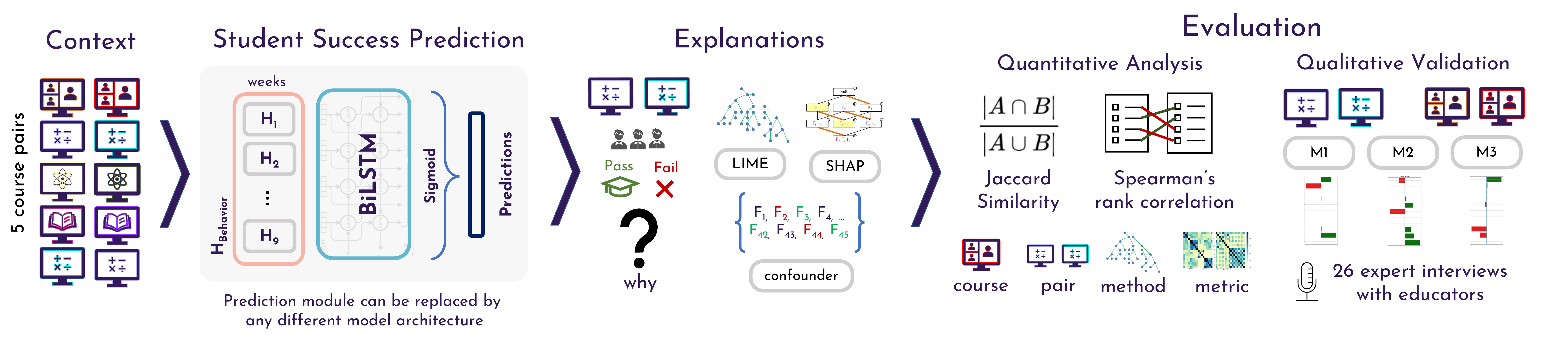}
     \caption{Our framework involves four steps: 1) extracting clickstream data from five pairs of courses, 2) modeling course data for success prediction, 3) using XAI methods to extract feature importance scores, 4) quantitative and qualitative evaluation of explainers.}
     \label{fig:pipeline}
     \vspace{-4mm}
 \end{figure*}

With our methodology, we aim to quantitatively and qualitatively compare and validate the explanations from success prediction models across different settings and assess whether they can build the basis for actionable interventions. Fig. \ref{fig:pipeline} illustrates the main steps of our XAI evaluation framework. While explanations of model predictions are essential to increase trust \cite{DBLP:conf/LAK/NazaretskyCA22} and enable evidence-based interventions in any educational scenario, we focus on online and blended learning scenarios, aiming to predict student success. In a first step, we extract pairs of courses $(c, c')$ with `controlled' differences based on literature influencing students' behavior and learning experience. For example, courses $c$ and $c'$ might both be Bachelor's level mathematics courses, but exhibit a different instructional design. Alternatively, they might have the same instructional design but a different target audience. This study design enables us to hypothesize about potential causes for a success/failure and therefore validate the provided explanations. In a second step, we extract the interaction logs for all selected courses and train a success prediction model separately for each course. Next, we apply post-hoc explainability methods to retrieve explanations for a single student. The two final steps then consist of the quantitative evaluation and the semi-structured interviews with experts. 
%We used quantitative methods to compare and analyze the explainability methods for the different pairs of courses and conducted a qualitative user study with teachers at university level to evaluate the validity and actionability of the explanations. In the following, we describe each step of our framework in detail.

%%%%%%%%%%% Study Context %%%%%%%%%%%%%%%%%%%%%%%%
\subsection{Learning Context}
%%%%%%%%%%% Study Context %%%%%%%%%%%%%%%%%%%%%%%%
%We describe the design of the study as well as the data collection process in detail.
%The data is fully anonymized with regards to student information. It contains fine-grained video and quiz interactions for each student, e.g., pressing pause on a video or submitting a quiz. From this data set, we extracted pairs of courses $(c,c')$.\\

% Design of the study
\vspace{1mm} \noindent \textbf{Study Design}. 
\label{sec:study-design}
We opted for a pairwise study design as a basis for the validation of explanations. Specifically, we extracted pairs of courses $(c,c')$ with opposite (or largely different) characteristics, which are known to have an impact on student behavior and learning outcomes. We assumed that these behavioral differences would be captured by the student success prediction model and therefore reflected in the related explanations.
From previous research, two major factors influencing learning in online settings have emerged: \textit{population} and \textit{instructional design}. 

% Influence of population on learning
Multiple studies (e.g., \cite{ke2013online}) have reported the important influence that demographics in terms of age, culture, language, and gender can have when designing online learning. For example, it has been shown that help-seeking behavior differs across cultures \cite{Ogan15} and that the preference for specific learning activities is influenced by culture and socio-economic background \cite{Rizvi22}. A recent large-scale analysis across multiple MOOC providers has confirmed the impact of the language of instruction, cultural background, or localized course design on learners' educational experiences \cite{ruiperez22}. Furthermore, \cite{Kiziclec13} has also found differences in engagement across age groups and gender. Unfortunately, the data available to us contained only sparse demographic information, with gender and geographic location only available for very small subsets of course participants ($<9\%$ for most of the courses). We therefore used the instruction language as a proxy of the target audience. Table \ref{tab:study_design} lists the according study design for the factor \textit{population}.
%, including a pair of courses differing only in the instruction language (i.e. the exact same course was taught in different languages in its first and second iteration).

% Influence of instructional design on learning
A large body of prior work (e.g., \cite{GASEVIC201668,JUNG2019377,Yang17}) has also demonstrated the impact of the instructional design on students' learning experience and success. For example, \cite{JUNG2019377} has shown that instructional design factors are stronger indicators of a successful MOOC experience than the content of the course or demographic factors. \cite{Yang17} has found that the quality of a course influences the persistence of students in MOOCs. These findings have been confirmed by \cite{GASEVIC201668}, who demonstrated that the accuracy of predictive models can be increased by including indicators that describe the instructional design. In our study, we focused on three dimensions of \textit{instructional design} that we could measure based on the available data. The first dimension is active learning, which was found to be an important factor for engagement and popularity as well as learning outcomes in MOOCs \cite{Cagiltay20,Hew16,Koedinger15, mejia22}. In our case, we approximate the amount of active learning in a course by calculating the average weekly quiz ratio, i.e. the number of weekly quizzes divided by the number of weekly videos. The second investigated dimension is the setting of the course (MOOC or flipped classroom). \cite{marras2021can} compared a large number of learning indicators extracted from previous literature in flipped and online learning settings and found differences in feature importance across settings. For our study, we therefore compared pairs of courses that have been taught as a MOOC as well as in a flipped setting. The last dimension that we included is the optimality of the course design. Prior research has demonstrated that the course design should be adapted to the target population \cite{Kiziclec13,Rizvi22}. For our context, we approximated the optimality of the course design by the course success rate. We chose a pair for comparison, where the same course (in terms of content, level, and language) was taught in two iterations, with a substantial higher success rate in the second iteration. 
%iven everything else equal, we hypothesize that a higher success rate is due to adjustments in the instructional design. 
The three dimensions for \textit{instructional design} along with the related indicators and selected course pairs are listed in Table \ref{tab:study_design}.

% TABLE illustrating the study design
\begin{table*}[!htb]
\small
\caption{\textbf{Study Design}. We performed pairwise comparisons across nine courses within two factors influencing interaction behavior (instructional design and population). We analyze three dimensions within course design and one dimension for population.} %A detailed description of all selected courses can be found in Table \ref{tab:courses}.}
\vspace{-1mm}
\resizebox{\textwidth}{!}{
\begin{tabular}{l||cccccccc|cc}
\toprule
\textbf{Factor} &
\multicolumn{8}{c|}{\textbf{Course Design}} &
  \multicolumn{2}{c}{\textbf{Population}} \\ 
  \midrule
  \textbf{Dimension} &
\multicolumn{4}{c|}{\textbf{Setting}} &
  \multicolumn{2}{c|}{\textbf{Active Learning}} &
  \multicolumn{2}{c|}{\textbf{Optimality}} &
  \multicolumn{2}{c}{\textbf{Language}} \\
  \midrule
  \textbf{Indicator} &
  \multicolumn{2}{c}{Flipped} &
  \multicolumn{2}{c|}{MOOC} &
  \begin{tabular}[c]{@{}c@{}}High quiz to \\ video ratio\end{tabular} &
  \multicolumn{1}{c|}{\begin{tabular}[c]{@{}c@{}}Low quiz to \\ video ratio\end{tabular}} &
  \begin{tabular}[c]{@{}c@{}}Higher \\ success rate\end{tabular} &
  \begin{tabular}[c]{@{}c@{}}Lower \\ success rate\end{tabular} &
  English &
  French \\
  \midrule
  \textbf{Course ID} &
  Flip LA &
  Flip FP &
  MOOC LA &
  \multicolumn{1}{c|}{MOOC FP} &
  AN$_2$ &
  \multicolumn{1}{c|}{Geo} &
  AN$_2$ &
  AN$_1$ &
  VA$_1$ &
  VA$_2$ \\ 
 \bottomrule
\end{tabular}}
\label{tab:study_design}
\end{table*}

\vspace{1mm} \noindent \textbf{Data Collection}.
Our experiments are based on interaction and metadata collected from seven MOOCs and two flipped classroom courses \cite{hardebolle2022gender} of  
École Polytechnique Fédérale de Lausanne, recorded across three online platforms (Coursera, EdX, and Courseware).  The MOOCs were accessible without any restrictions, i.e. worldwide and at any time. All the selected courses organized the content in weeks. Each week included video lectures introducing the important concepts as well as quizzes for self-assessment. Furthermore, students were requested to also take graded assignments (scored up to $100$ points), which served as a basis for obtaining the course certificate. The number of assignments and the minimal score needed for passing the MOOC was up to the instructor. While the flipped courses were recorded on the same platform, they offered restricted access only. Both of the flipped courses were compulsory (in the Bachelor's and Master's, respectively). Students of the class were required to perform online pre-class activities including watching lecture videos and solving associated quizzes.
%Learning activities in class then included short quizzes and problem-solving exercises as well as the discussion of questions that arose from the online content. In-class activities were not recorded in any form. The selected courses cover different topics and target audiences. 
Table \ref{tab:courses} lists detailed information about all the included courses. The data collection and analysis was approved by the university’s ethics review board (HREC 058-2020/10.09.2020, 096-2020/09.04.2020).

The log data for both MOOCs and flipped courses was collected in the form of a time-wise clickstream per student. Specifically, students enrolled in a course $c$ interact with the learning objects $\mathbb{O}^c$ from that course. In our study, we assume that the learning objects can be either videos or quizzes, but the framework can be easily extended to other types (e.g., forum posts, textbook readings). All learning objects $\mathbb{O}^c$ from a course $c$ are associated with a specific course week $w$. The extraction of this weekly course schedule allows engineered learning indicators specific to the course design (see Section \ref{sec:feat_extract}). We denote the \emph{interactions} of a student $s$ in a course $c$ as a time series $I_s^c = \{i_1, \ldots, i_K\}$ (e.g., a sequence of video plays and pauses, quiz submissions). We assumed that each interaction $i$ is represented by a tuple $(t, a, o)$, including a \emph{timestamp} $t$, an \emph{action} $a$ (videos: load, play, pause, stop, seek, speed; quiz: submit), and a \emph{learning object} $o \in \mathbb{O}^c$ (video, quiz). We finally denote the \emph{binary success label} (pass-fail) for student $s$ in course $c$ as $l_{s,c}$. 

\begin{table*}[!htb]
\caption{\textbf{Learning Context}. Detailed instructional design and student population information on the nine selected courses for this XAI study. The relevant indicators for the pairwise selection are denoted in bold.}
\label{tab:courses}
\vspace{-2mm}
\small
\resizebox{\textwidth}{!}{
\begin{tabular}{llllllrrrr}
\toprule
\textbf{Course Title} & \textbf{Course ID} & \textbf{Setting} & \textbf{Field$^1$} &  \textbf{Level} & \textbf{Language}  & \multicolumn{1}{r}{\textbf{\begin{tabular}[c]{@{}c@{}} Weeks\end{tabular}}} & 
\multicolumn{1}{r}{\textbf{\begin{tabular}[c]{@{}c@{}} Quiz/Video$^2$ \end{tabular}}} & 
\multicolumn{1}{r}{\textbf{\begin{tabular}[c]{@{}c@{}} Students$^3$\end{tabular}}} & \multicolumn{1}{r}{\textbf{\begin{tabular}[c]{@{}c@{}} Success Rate \end{tabular}}} \\
\midrule
Algebra (part 2)& MOOC LA & \textbf{MOOC} & Math & Prop & French & 4 & 2.13 & 170 & $0.67$\\
Algebre lineaire&Flip LA&\textbf{Flipped} & Math & BSc & French &10 & 1.76 & 214 &$0.59$\\
Functional Programming Principles in Scala&MOOC FP&	\textbf{MOOC}& CS& BSc&	English&	6& 0.52 &	3565& $0.48$\\
Functional programming&	Flip FP& \textbf{Flipped}&CS&MSc&	English&	10& 0.21&	218& $0.62$\\
African Cities - An introduction to urban planning&VA$_1$&MOOC& SS&	BSc&	\textbf{English}&	12&	5.42 & 5643& $0.10$\\
African Cities - An introduction to urban planning&VA$_2$& MOOC& SS& Prop&\textbf{French}&12&5.42&4699& $0.05$\\
Analyse Numérique pour Ingénieurs&	AN$_1$& MOOC& Math&	BSc&	French&	9&9.14&	506& $\mathbf{0.08}$\\
Analyse Numérique pour Ingénieurs&	AN$_2$&MOOC&Math&BSc&	French&	9&\textbf{9.14}&	506& $\mathbf{0.71}$\\
Éléments de Géomatique&	Geo& MOOC& Eng.& BSc&French&11&\textbf{3.89}&452&$0.45$ \\
\bottomrule
\end{tabular}}
\scriptsize{$^1$\textit{CS}: Computer Science; \textit{Math}: Mathematics; \textit{SS}: Social Sciences; \textit{Eng.}: Engineering.} \; \; 
\scriptsize{$^2$Nr. of weekly quizzes/Nr. of weekly videos, averaged over all course weeks.} \; \; \\
\scriptsize{$^3$For MOOCs, after removing early-dropout students according to \cite{swamy22b}.}
\end{table*}

\begin{comment}
\begin{table} [t]
  \resizebox{\hsize}{!}{
  \centering
  \begin{tabular}{ ||b{1.5cm}|b{1.5cm}|b{1.5cm}|b{1.5cm}|b{1cm}|b{1.5cm}|b{1cm}|b{1cm}|b{.5cm}|b{1cm}||}{
    \toprule
    Title&Identifier&Attribution&Modality&Level&Population&\# week&\# students&Active Learning&Passing rate\\
    \midrule
    Algebra (part 2)&MOOC LA&different modalities&Prop&French&4&170& 	&67.00\%\\
    \hline
    Algebre lineaire&Flip LA&different modalities&flipped&BSc&French&10&214&0.64&59.00\%\\
    \hline
    Functional Programming Principles in Scala&MOOC FP&different modalities&	Coursera&	BSc&	English&	6&	3565&	7.00&	48.00\%\\
    \hline
    Functional programming&	Flip FP&	different modalities&	flipped&	MSc&	English&	10&	218& &62.00\%\\
    \hline
    African Cities - An introduction to urban planning&VA\_1&student population&Coursera&	BSc&	English&	12&	4941&	9.94&	11.40\%\\
    \hline
    African Cities - An introduction to urban planning&VA\_2&student population& Coursera& Prop&French&12&3000& 5.42	&4.98\%\\
    \hline
    Analyse Numérique pour Ingénieurs&	AN$_1$&	passing rate&	BSc&	French&	9&	2104&	8.50&	2.00\%\\
    \hline
    Analyse Numérique pour Ingénieurs&	AN$_$2	&passing rate \\ active learning&Coursera&	BSc&	French&	9&	2065&	9.14&	71.34\%\\
    \hline
    Éléments de Géomatique&	Geo&	active learning&	BSc&	French&	11&	2149&	3.89&	9.50\% \\
  \bottomrule}
\end{tabular}}
\caption{Metadata of selected courses}
\label{tab:Metadata}
\end{table}
\end{comment}

%%%%%%%%%%% Student Success Prediction %%%%%%%%%%%%%%%%%%%%%%%%
\subsection{Student Success Prediction}
\label{sec:blackbox-model}
%%%%%%%%%%% Student Success Prediction %%%%%%%%%%%%%%%%%%%%%%%%
The collected student log data in the form of time-series of interactions builds the basis for extracting features serving as indicators of learning. The engineered features are fed into the success prediction model to obtain a pass-fail prediction $l_{s,c}$ for each student $s$ in course $c$, which can then be interpreted and explained.

\vspace{1mm} \noindent \textbf{Features}. \label{sec:feat_extract} A large body of work has focused on developing success prediction models for MOOCs (e.g., \cite{gardner2018student}) and flipped classroom courses (e.g., \cite{lee2022affects}). A multitude of feature sets and classifiers tied to a specific course or a small set of similar courses have been proposed. In their recent meta-analysis, \cite{marras2021can} performed a literature search to retrieve prior research proposing feature sets for success prediction models in online settings. After removal of overlapping sets, the combined feature set contains over $100$ features, which we use as input for our models. Each feature is extracted per student $s$ and week $w$ in course $c$, meaning that each feature is a time-series of length $W$ (where $W$ is the number of weeks of the course). The combined feature set has been retrieved from nine different papers \cite{marras2021can, DBLP:conf/edm/AkpinarRA20,boroujeni2016quantify,chen2020utilizing,lalle2020data,DBLP:journals/eait/LemayD20,DBLP:conf/aied/MbouzaoDS20,DBLP:journals/eait/MubarakCA21,DBLP:journals/tlt/WanLYG19}. Given the high number of features (110), we will give a coarse overview and categorization of the features only and refer to the original work as well as to the public GitHub repository\footnote{https://github.com/epfl-ml4ed/flipped-classroom} of \cite{marras2021can} for a detailed documentation of the features. The subset of features identified as important by the explainers and presented to educators is showcased in Table \ref{tab:features}.

%and includes: $13$ features from \cite{DBLP:conf/edm/AkpinarRA20}, $9$ features from \cite{boroujeni2016quantify}, $13$ features from \cite{chen2020utilizing}, $22$ features from \cite{lalle2020data}, $10$ features from \cite{DBLP:journals/eait/LemayD20}, $12$ features from \cite{marras2021can}, $3$ features from \cite{DBLP:conf/aied/MbouzaoDS20}, and $14$ features from each \cite{DBLP:journals/eait/MubarakCA21} and \cite{DBLP:journals/tlt/WanLYG19}. Given the high number of features, we will give a coarse overview and categorization of the features only and refer to the original work as well as to the public GitHub repository\footnote{https://github.com/epfl-ml4ed/flipped-classroom} of \cite{marras2021can} for a detailed documentation of the features. The subset of features that was identified as important in our evaluations is showcased in Table \ref{tab:features} and presented to educators in our qualitative study (see Section \ref{sec:study}).

% Two categories: self-regulation and patterns only
% Description and references for first category;
The extracted features can be roughly divided into two categories: engineered features that serve as indicators of students' self-regulated learning (SRL) behavior and raw patterns extracted from the clickstream data logs. For engineered features, we have features with dimensions of \textit{effort} (effort regulation), \textit{regularity} (time management), \textit{proactivity} (time management), and \textit{control} (metacognition), which can all be associated with success in online higher education \cite{broadbent2015self}.
%in three dimensions of SRL that are significantly associated with learning outcomes: time management (abilit
%y to plan study time and tasks), effort regulation (persistence in learning), and metacognition (awareness and control of thoughts), critical thinking (ability to carefully examine learning material), and help-seeking (obtaining assistance when needed). Based on these findings, we provide an overview of the features using the dimensions of \textit{effort} (effort regulation), \textit{regularity} (time management), \textit{proactivity} (time management), and \textit{control} (metacognition), and Table \ref{tab:features} highlighting several features in detail. The nature of our log data did not allow us to extract indicators related to help-seeking behavior or critical thinking.
Features in the \textit{\textbf{Effort}} dimension \cite{chen2020utilizing,DBLP:journals/eait/LemayD20,lalle2020data,DBLP:conf/aied/MbouzaoDS20,DBLP:journals/tlt/WanLYG19, DBLP:conf/edm/AkpinarRA20, he2018measuring} monitor the extent to which a student is engaged in the course, which has been shown to be essential for success (e.g., \cite{cho2013self}).
Features in the \textit{\textbf{Regularity}} dimension \cite{boroujeni2016quantify, he2018measuring} monitor the extent to which the student follows a regular study pattern. A regular working schedule has been proven to be predictive of success in MOOCs \cite{boroujeni2016quantify} and flipped settings \cite{DBLP:journals/ce/JovanovicMGDP19}. 
Features in the \textit{\textbf{Proactivity}} dimension \cite{marras2021can, boroujeni2016quantify,DBLP:journals/tlt/WanLYG19} attempt to measure how much students are ahead or behind the schedule of the course. Proactivity has been shown to play an important role in student success \cite{geertshuis2014preparing, boroujeni2016quantify}. 
The forth dimension of engineered features, \textit{\textbf{Control}} \cite{lalle2020data,DBLP:journals/eait/LemayD20,DBLP:journals/eait/MubarakCA21}, includes indicators of fine-grained video behavior, that constitute a proxy for assessing to what extent students can control the cognitive load of the video lectures. Prior research has shown that the flow of videos in MOOCs can lead to cognitive overload \cite{biard2018effects}.
For patterns stemming directly from the raw interaction logs (notably not engineered indicators), \cite{DBLP:conf/edm/AkpinarRA20} extracted consecutive \textit{\textbf{Subsequences}} of $n$ clicks extracted from the session clickstreams of a blended course. The resulting feature set is consequently huge  (e.g., $350$ features for course Flip LA). We therefore only included the sub-sequence features found important in \cite{marras2021can}, resulting in $110$ total features.

% TABLE with detailed feature list used in later analyses!!!
\begin{table*}[!htb]
\centering
\small
\caption{\textbf{Features}. For brevity, we only list the $22$ features that have been identified as important by at least one explainability method in our analysis in Section \ref{sec:results}. Remaining features can be found in \cite{marras2021can}.}
\label{tab:features}
\vspace{-2mm}
\resizebox{\textwidth}{!}{
\begin{tabular}{@{}llll@{}}
    \toprule
    \textbf{Dimension} & \textbf{Feature (simplified names)} &  & \textbf{Description} \\ \midrule
    \multirow{5}{*}{\textit{Effort}} & check-check-check-quiz & \cite{DBLP:conf/edm/AkpinarRA20} & The amount of times the student checks problems three times in a row. \\
    & correct-time-quiz & \cite{DBLP:journals/tlt/WanLYG19} & Total time spent divided by the number of correct problems. \\
    & distinct-probs-quiz & \cite{DBLP:journals/tlt/WanLYG19} & The total number of distinct problems attempted by the student. \\
    & num-submit-quiz & \cite{DBLP:journals/tlt/WanLYG19} & The number of submissions performed for a quiz, on average. \\
    & total-time-vid & \cite{DBLP:conf/edm/AkpinarRA20} & The total (cumulative) time that a student has spent on video events. \\
     \midrule
    \multirow{5}{*}{\textit{Regularity}} 
    & active-participation-weekly-vid & \cite{marras2021can} & The number of videos the student watched fully over total loaded videos (per week). \\ 
    & attendance-rate & \cite{he2018measuring} & The number of videos played over the total number of videos released. \\
    & hourly-freq-regular & \cite{boroujeni2016quantify} & The extent to which the hourly pattern of user’s activities is repeating over days. \\
    & watch-ratio-vid & \cite{he2018measuring} & Ratio of amount of video watched for videos a student opens. \\
    & std-time-session & \cite{chen2020utilizing} & The standard deviation of time spent from a login to the end of the session. \\
     \midrule
    \multirow{5}{*}{\textit{Proactivity}}
    & eager-view-vid & \cite{marras2021can} & The extent to which the student approaches a video early.\\
    & timely-view-vid & \cite{marras2021can} & The extent to which the student approaches a video in the right week.\\
    & eager-view-quiz & \cite{marras2021can} & The extent to which the student approaches a quiz early.\\
    & ratio-clicks-weekend & \cite{chen2020utilizing} & The ratio between clicks in weekdays and weekends. \\
    & std-correct-time-quiz & \cite{DBLP:journals/tlt/WanLYG19} & Variance of total time spent divided by the number of correct problems. \\
     \midrule
     \multirow{7}{*}{\textit{Control}} 
     & avg-len-seek-vid & \cite{lalle2020data} & The student's average seek length (seconds). \\
    & freq-pause-vid & \cite{lalle2020data} & The frequency between every Video.Pause action and the following action. \\
    & freq-play-vid & \cite{lalle2020data} & The frequency of the play event in the students’ sessions. \\
    & play-stop-play-vid & \cite{DBLP:conf/edm/AkpinarRA20} & The amount of times the student plays a video, stops, and plays another one. \\
    & play-pause-load-vid & \cite{DBLP:conf/edm/AkpinarRA20} & The amount of times the student plays a video, pauses, and plays again. \\
    & pause-speedchange-play-vid & \cite{DBLP:conf/edm/AkpinarRA20} & The amount of times the student pauses, changes the speed, and plays the video. \\
    & speed-vid & \cite{DBLP:journals/eait/LemayD20} & The average speed the student used to play back a video. \\
     \bottomrule
    \end{tabular}}
\end{table*}
\renewcommand{\arraystretch}{1.0}

\vspace{1mm} \noindent \textbf{Predictive Model}.
To predict the pass-fail label $l_{s,c}$ of student $s$ in course $c$ based on the extracted behavioral features, we relied on the neural architecture based on \emph{Bidirectional LSTMs} suggested by \cite{swamy2022evaluating}. The model input is represented by $H$, i.e., the extracted behavior features, having a shape of $|S| \times W \times 110$, where $|S|$ and $W$ denote the number of students and weeks, respectively. NaN values were replaced with the minimum value of each respective feature. The architecture is composed of two \emph{BiLSTM} layers of size $32$ and $64$ (using a loopback of $3$) and a \emph{Dense} layer (with Sigmoid activation) having a hidden size of $1$. The model outputs the probability $p_{s,c}$ that the student $s$ will pass course $c$.

%%%%%%%%%%% Explanations %%%%%%%%%%%%%%%%%%%%%%%%
\subsection{Generating Explanations}
\label{sec:explanabilitymethods}
%%%%%%%%%%% Explanations %%%%%%%%%%%%%%%%%%%%%%%%
While deep learning models have shown highly accurate performance over traditional machine learning methods, this accuracy comes at the cost of interpretability. Neural networks are therefore referred to as black-box models, emphasizing their lack of transparency. Post-hoc explainability methods can be applied to the predictions of a black-box model in order to gain interpretablity in the form of feature importance score. We used LIME and SHAP explainability methods to interpret the predictions of our success prediction models.
%Both of the methods are instance-based, i.e. they provide an explanation for a single instance (in our case for the prediction of a single student).
%To extract explanations from our underlying models, we use post-hoc explainability methods. In the following section, we will describe the three specific methods used (LIME, SHAP, and the confounder), the reasons we chose these methods, as well as the metrics we will use to compare explanations.

%%%%%%%%%LIME%%%%%%%%%%%%
\vspace{1mm} \noindent \textbf{Local Interpretable Model-agnostic Explanations (LIME)}.
Surrogate models are interpretable models (e.g., linear regression, decision tree) that are used to approximate the predictions of a black-box model. LIME \cite{lime} aims to train a local surrogate model to approximate the predictions of the black-box model for one specific instance (in our case, a student). Applying this method requires five steps: 1) select the student for which you want to explain the prediction, 2) generate new samples in the neighborhood of the selected student by slightly perturbing the feature values of the instance (i.e., generate new artificial students with slightly different feature values than the selected student), 3) weight each new synthetic student by the distance to the selected student (i.e. the higher the distance, the lower the weight), 4) train an interpretable surrogate model (e.g., a linear regression, decision tree) on the new perturbed feature values of the synthetic students, and 5) explain the prediction of the selected student by interpreting the surrogate model. The trained interpretable model therefore should be a good approximation of the black-box model locally. The outputs of LIME are the feature weights representing the feature influence on the final decision. The complexity of the surrogate model needs to be decided by the user. The default setting of LIME includes a maximum of $n=10$ features into the interpretable model. Therefore, if LIME is run with a linear regression and a maximum of $10$ features, LIME will return the regression weights of the optimal $10$ features. We decided to use this default setting for our evaluations in order to increase the reproducibility of our work. To ensure comparability between explainability methods, we scaled the obtained feature weights into importance scores to the interval $[0, 1]$, where $1$ indicates high feature importance.

%trains a local linear model to explain each individual student instance. LIME is the most famous interpretability approach for black-box models by adoption rate and citations; as this is commonly used in ML explainability literature \vini{cite}, it is pertinent for this study. The method generates a new dataset consisting of perturbed samples and the corresponding predictions of the black-box model. On this new dataset LIME then trains an interpretable model, linear regression, or random forest as examples, which is weighted by the proximity of the sampled instances to the instance of interest. proximity to the original instance. Mathematically, the local model can be expressed with the following equation:
%\begin{equation}
    %explanation(x)=\arg \min_{g \in G}L(f, g, x) + \Omega(g)
%\end{equation}
%where $x$ is the instance being explained, a explainability $g$ belongs to a family of all possible explanations $G$, $L$ is a loss function that measures measures how close the predictions of the explainer $g$ are to the predictions of the original model $f$, $X$ is a proximity measure that defines how far the neighborhood around instance $x$, and $\Omega(g)$ represent the complexity of model $g$. LIME returns feature weights representing the feature influence on the final decision. First, we calculate these absolute values to be the importance scores, and then normalize them to the interval $[0, 1]$ with max-min scale, where 1 indicates high importance.

%%%%%%%%%SHAP%%%%%%%%%%%%
\vspace{1mm} \noindent \textbf{SHapley Additive exPlanations (SHAP)}.
Inspired by game-theory's Shapley values \cite{shapley1997value} and LIME \cite{lime}, SHAP \cite{lundberg2017unified} is a popular method for explaining individual predictions. SHAP explains the prediction of an instance $s$ (in our case a student) by quantifying the contribution of each feature to the prediction. To determine a SHAP score for a feature, the algorithm has three main steps: 1) train a coalition of models on each possible subset of features (each feature individually, each combination of two features, each combination of three features, and so on), 2) identify marginal contributions of each feature (e.g. the difference in prediction of a model trained solely on $h_1$ and a model trained on features $h_1$ and $h_2$ to identify a marginal contribution of $h_2$), 3) accumulate these marginal contributions for each feature $h$ by computing a weighted sum. The computational and space complexity required for this algorithm is massive as it calls for $2^{|H|}$ models to be trained. 
%KernelSHAP uses an optimization with a kernel function to efficiently construct data points, significantly speeding up the SHAP computation. 
To obtain comparable importance scores, we apply the same normalization transformation as LIME, taking the absolute values of the KernelSHAP feature attributions and scaling them to $[0, 1]$.

% %%%%%%%%%CEM%%%%%%%%%%%%
\vspace{1mm} \noindent \textbf{Counterfactual-based Confounder}.
\label{sec:confounder}
To validate our explainers with expert educators, we aim to showcase the results of the LIME and SHAP explainers and observe which explanations educators trust more across diverse course scenarios. However, without including a control explanation, we do not know if the LIME or SHAP explanations are really preferred, or just the best suboptimal option. To address this, we add a confounder explanation based on counterfactuals, using the contrastive explanation method \cite{dhurandhar2018explanations}. 
%CEM generates a counterfactual student for a given instance by identifying which features must be necessarily absent for the prediction class to remain the same. 
We purposely add inaccuracies into this approach by switching features from indicators of passing to failing (positive to negative importance scores).
%for the students predicted to pass the course. 
%We prefer this method over generating a random explanation to ensure that the synthesized explanation has to be explicitly incorrect. 
In the study (detailed in Section \ref{sec:study}), we anonymize and shuffle LIME, SHAP, and the confounder and give all three explanations to expert educators.

% \label{sec:cem}
% Like the previous methods, CEM \cite{dhurandhar2018explanations} aims to explain the prediction of a specific instance. Given an instance, CEM decides what feature need to be minimally present to justify the prediction and analogously what should be minimally absent the prediction. 
% describes the smallest change to the feature values that changes the prediction to a predefined output. CEM is chosen for its direct ability to be used to design interventions (the closest student behavior to a specific student that will change the label) and its dissimilarity from the previous two approaches. For our setting, we consider pertinent negatives as they are intuitively more similar, and therefore comparable, to other counterfactual-based explainability methods. For each generated pertinent negative, we calculate the importance score for each feature by multiplying the absolute change from the value in the original instance to the value in the pertinent negative, modeled as the standard deviation ($SD$) of that feature $\Tilde{X}(x)$ across all instances used for the experiment $X(x)$, as shown in the following formula:
% \begin{equation}
%     CEM(x) = [X(x) - \Tilde{X}(x)] \times SD(x)
% \end{equation}

% The importance score therefore takes into consideration both the necessary perturbation of the feature as well as the significance of the change relative to the feature range. We normalize the scores in the range $[0, 1]$, such that the resulting feature importance weights can be directly comparable.

%%%%%%%%%Evaluation Metrics%%%%%%%%%%%%
\vspace{-2mm}
\subsection{Comparing Explanations}
\label{sec:quantitativecomparison}
The goal of our quantitative evaluation was to compare the feature importance scores across courses and methods and quantify the differences and similarities between them. Our choice of similarity metrics between methods is based on prior work. Specifically, \cite{swamy2022evaluating} used \textit{Spearman's Rank-Order Correlation} to assess the agreement on the ordering of features (in terms of feature importance scores) between two different explainability methods $e_1$ and $e_2$. \textit{Spearman’s Rank-Order Correlation}, often referred to as \textit{Spearman's $\rho$} \cite{spearman1961proof}, identifies the rank correlation (statistical dependence between the rankings) between two variables. It is defined as the Pearson correlation coefficient between the rankings of two variables. We chose this metric to highlight the importance of feature ranking order in explanations.

In addition, we also use \textit{Jaccard Similarity} to assess the agreement on the top ranked features between $e_1$ and $e_2$ \cite{swamy2022evaluating}. \textit{Jaccard Similarity} is used to identify the overlap between the top ten important features \cite{jaccard1901etude}. This method computes the cardinality of the intersection divided by the cardinality of the union of two feature sets. We chose this metric because of its approximation to the real-world setting of educational interventions based on explanations; an educator would only be able to intervene on a limited number of features. We refrained from comparing exact importance scores between methods, as LIME limits the complexity of the explanation (default $n=10$), resulting in sparser explanations (and thus generally higher importance scores). When comparing explainability methods across courses with different settings (which also entails different duration and different individual students), we average feature values across all weeks for each student, and then across all students to obtain an aggregated ranking.%To compute Spearman’s Rank-Order Correlation between the different explainability methods, we converted the vectors of feature importance scores for each student into an ordered list.

% We used the Jensen-Shannon distance (JS distance) \cite{lin1991divergence} to compute distances between exact feature importance score distributions obtained with different explainability methods. The Jensen–Shannon distance is the square root of the Jensen-Shannon divergence, originally based on the Kullback–Leibler divergence with smoothed values. To compute the Jensen-Shannon distance between the same explainability on two courses with different settings, we first separately calculate importance values for each feature and then average across all features to obtain the represented JS distance.

%%%%%%%%%%% User Study %%%%%%%%%%%%%%%%%%%%%%%%%%
\vspace{-2mm}
\subsection{Validating Explanations}
\label{sec:study}
%%%%%%%%%%% Explanations %%%%%%%%%%%%%%%%%%%%%%%%
To qualitatively validate the results of the explainers, we conduct a study with educational domain experts. To design the study, we conducted five pilot interviews with a mix of gender, experience, and educational research background. 

%%%%%%%%%%%%%%%Participants%%%%%%%%%%%%%%%%%%%%%%
\vspace{1mm} \noindent \textbf{Participants}. The study was conducted with approval from the institutional review board for a target audience of experienced STEM professors (HREC 065-2022/27.09.2022). We interviewed $26$ educators for $30$ to $45$ minutes each, using the first five interviews as a pilot. The pilot interviewers provided feedback that we used to hone the visualizations of the XAI methods, the scenarios presented in the study, and the framing of the questions. We collected demographic information from our participants on age, gender, nationality, affiliated institution, teaching experience, and MOOC experience. The $21$ officially selected study participants are international; $33.33\%$ of our participants come from American universities, $23.81\%$ from Swiss universities, $14.28\%$ from Italian universities, $14.28\%$ from Middle Eastern universities, $9.52\%$ from Spanish universities, and $4.76\%$ from Australian universities. We have a high gender imbalance in our study, with $80.95\%$ of participants identifying as male. The mean age of the participants was 36.5 ($\sigma = 9$). All participants in the study were university-level educators with a technical background teaching science, engineering, data science, or computer science concepts. We chose these educators because the courses we chose to analyze in the study pertain to numerical analysis or functional programming taught at the university level, and we wanted to narrow the scope to experts in this specific educational sub-domain. $76.19\%$ of interviewees currently hold the job title of professor or lecturer; other participants are not officially recognized as researchers but taught or are currently teaching university courses. Several of the participants have experience with MOOCs and digital courses, with $33.33\%$ having been a participant in a MOOC, $57.14\%$ having taught a MOOC, and $28.57\%$ having worked with MOOC data. 

\vspace{1mm} \noindent \textbf{Procedure}. With our interviews, we examined two major educational dimensions in detail: setting (MOOC vs. flipped classroom) and active learning (high vs. low) using two separate scenarios. We chose to focus on these two aspects (out of our selected five) as they were the most discussed in related literature. Each participant only saw one scenario. 

In the first scenario (difference in setting), participants were told that they were a professor offering a flipped classroom course on programming. We provided a summary description of the course they teach, stating that it was offered by a European university to $218$ students for $10$ weeks, with both quizzes and videos, culminating in a final exam. We then mentioned that a fellow professor adapted their flipped classroom curriculum to a MOOC with the same topics and materials.
%This new MOOC course is offered worldwide by a Swiss university to 3564 students for 6 weeks, with both quizzes and videos, culminating in a final exam. 
We explained to participants that neural network models had been trained on both courses, achieving high balanced accuracy (at least $90\%$) in terms of pass-fail performance prediction and that we had run three state-of-the art XAI methods (abstracted to M1, M2, and M3) to obtain feature importance scores. One of these methods was in fact an inaccurate confounder (see Section \ref{sec:confounder}), but we did not reveal this to the interview participants. 

Our XAI models selected $11$ features as particularly important for the first scenario. These features were constructed as the pair with the most positive importance score (contributing to student success) and most negative score (contributing to student failure) across each method for each course. The interviewees were asked (for each course) to rank two features they thought could indicate student success the most and two features they thought could indicate student failure the most; a feature like \textit{total-time-video} could perhaps positively indicate student success if the value was high and student failure if the value was low, so if they feel it is most important for both, they can select it in both sets. We used an interactive drag-drop interface (Google Jamboard) to help the participants conduct this ranking. The motivation for this was to build a prior for the interviewees before we show the explainability results. 

After the interviewees had completed their independent feature importance ranking for both courses, we showed the results for explainers M1, M2, and M3 as explainability importance plots (see the last section of Fig. \ref{fig:pipeline}). A large bar (either green or red) means that the XAI method found this feature important as either a passing or failing differentiator. We asked the participants two main questions: 1) Which explanation do they agree with the most and why? 2) Do they think the explainers agree on what is important (on a Likert scale from 1-5)? We repeat this analysis for both courses. We shuffled the XAI methods shown in the figures (i.e. LIME might be M1 for the first course and M3 for the second).

%\begin{figure} %here, bottom of page, top of page, empty page in order of preference
%     \centering
%     \includegraphics[width=8cm]{images/05_trusting_explainers/flipped.png}
%     \caption{Comparison of feature importance scores showcased in the expert interview study for the flipped programming course.}
%     \label{fig:flipped}
% \end{figure}
 
Finally, we showcase XAI insights across the difference in course setting. Using the feature importance distributions across all $110$ features, we found the two features that changed the most in importance positively across the difference (i.e. flipped to MOOC) and the two features that changed the most in a negative way. We also noted the most important weeks for these features (beginning, middle, end, or throughout the course). We asked interviewees two more questions: 3) Do they agree with these insights (on a Likert scale of 1-5)? 4) How could they use these insights actionably to transition a curriculum across the course setting difference? To evaluate these results, two authors of this paper coded the responses independently across five to six categories defined by looking at repetitive trends in interview transcripts. 

For the second scenario (active learning), the exact same interview procedure was followed, the only difference was related to the selected courses and the story. Participants were told that they were the professor of a numerical analysis MOOC on Coursera, offered worldwide and taught in French to $506$ students for 9 weeks, with a lot of videos and few quizzes (low active learning), culminating in a final exam. Their professor friend offered a similar MOOC in Geometry taught worldwide in French to $452$ students for $11$ weeks, with many more quizzes than videos (high active learning), culminating in a final exam. Eight features were selected as important for this selected course pair.

% There were eight features selected by the XAI methods, indicating more similarity across the six explainers; we found that interviewees also found eight features much easier to rank and understand. 

% For the modality setting, these features were \textit{check-check-check-quiz} (end of course) and \textit{correct-time-quiz} (all of the course) as important indicators of student success are \textit{attendance-rate} (end of course) and \textit{timely-view-vid} (middle of course) for important indicators of student failure.

% The important indicators of student success in transitioning from low to high active learning (more questions) are \textit{play-pause-load-vid} (start of course) and \textit{pause-speedchange-play-vid} (mid course). For student failure, we note that \textit{eager-view-vid} (end of course) and \textit{correct-time-quiz} (end of course) change the most negatively.

\section{Experimental Evaluation}
\label{sec:results}
We evaluated post-hoc explainability methods for models trained on five pairs of courses in order to compare feature importance across different course settings (Sec. \ref{sec:study-design}). Using these results, we first explored how feature importance varied across different explainers for one specific course. Then, we investigated the similarity of the explainability methods across each pair of courses using two distance metrics (Sec. \ref{sec:quantitativecomparison}). Finally, we presented real-world scenarios to educators to assess the validity and actionability of the explanations across courses and methods (Sec. \ref{sec:study}).

\subsection{Experimental Protocol}
We trained a separate BiLSTM model for each course. Hyperparameters were selected consistently with prior work \cite{swamy2022evaluating}: two BiLSTM layers consisting of $64$, and $32$ units and one Dense layer consisting of $1$ unit with Sigmoid activation. As this work is not focused on improving model performance, we did not tune hyperparameters further. We split the data of each course into a training data set and a test data set, with $80\%$ and $20\%$ of students respectively. We performed a stratified train-test split over students’ pass-fail labels. For training and prediction, our models used the student log data collected for the full duration of the course. Balanced accuracy (BAC) was chosen as our primary evaluation metric because of the high-class imbalance of most of the selected courses.

\begin{figure*}
    \centering
     \includegraphics[trim={0 20 0 0},clip, width=\textwidth]{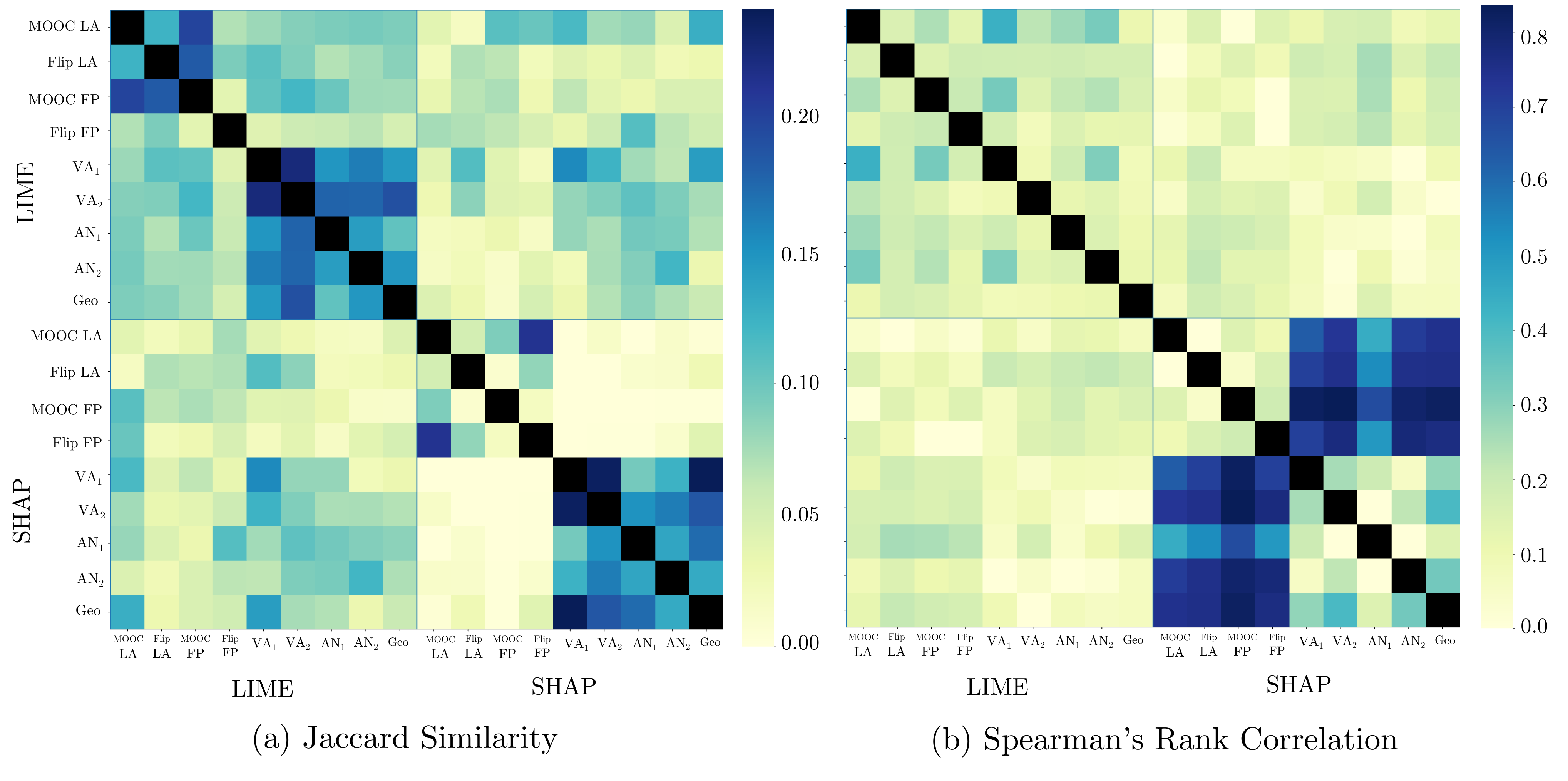}
     \caption{\textit{Jaccard Similarity} and \textit{Spearman’s Rank-Order Correlation} between XAI methods ordered by method.}
     \vspace{-3mm}
     \label{fig:heatmaps_big}
\end{figure*}

To analyze the explainers, we applied LIME and SHAP to compute the feature importances based on the model predictions for a specific instance. As it is not feasible to train each method for thousands of instances, we used a uniform sampling strategy to select $100$ representative students from each course, ensuring a balance between classes (pass-fail). We first extracted all failing students and ordered them according to the model's predicted probabilities. We then uniformly sampled $50$ failing students from this ordered interval. We repeated this exact same procedure to sample the $50$ passing students. This sampling procedure ensures that we include instances where the model is confident and wrong, where the model is unsure, and where the model is confident and correct.

\subsection{RQ1: Explanation Similarity}
In the first experiment, we compared the explanations of the instance-based methods within each course separately. We computed \textit{Jaccard Similarity} as well as \textit{Spearman’s Rank-Order Correlation} to assess distance between the top ten features and the overall feature rankings.

Fig. \ref{fig:heatmaps_big}a illustrates the \textit{Jaccard Similarity} between distributions of top ten important features over explainability methods. We observe clear patterns along the diagonal; this shows that there are similarities in the top 10 features among the same explainability method across different courses. In comparing courses across SHAP results (bottom left corner), we see high variance in the relationship with both the smallest scores in the graph and two strongly correlated sections. This implies that correlations across the SHAP method are heavily course dependent; the features SHAP finds important seem to depend on the similarity between courses. Examining SHAP and LIME, we note no strong correlations, indicating that these methods pick very different features for the same courses. We see an interestingly strong correlation using both explainability methods between $VA_1$ and $VA_2$. This might imply that the difference between French and English student populations are not as pervasive as expected.

% Fig. \ref{fig:heatmaps_big}b reorders the results by course. We note a very strong relationship between the five MOOCs in the bottom left portion of the graph. However, we also see a recursive subpattern within the similarities over the five MOOCs; LIME corresponds more with LIME and SHAP corresponds more with SHAP, the little sub-squares per course have stronger values on the diagonal. Examining the top left, we do not see strong correlation between the flipped courses. We also note that the Flip LA course corresponds strongly with MOOC FP in LIME scores, and that the Flip FP course corresponds strongly with MOOC LA in SHAP scores. It is interesting that these similarity scores are higher than those of the flipped courses with their aligned MOOC pairs; while all four courses are fairly similar in underlying student population, the active learning rates in LA courses are much higher than FP.

Fig. \ref{fig:heatmaps_big}b uses \textit{Spearman's $\rho$} to compare explanations over all features. The patterns between the two metrics are not similar; examining significantly more than the top ten features and using ranks instead of set logic leads to a different conclusion. The one recognizable strong correlation is in comparing SHAP explanations over the four courses which differ with respect to the setting (FP and LA). In any other setting, no significant correlation exists, including in comparing LIME with itself. The distance between LIME and SHAP explanations is more evident using \textit{Spearman's $\rho$}.

\begin{graybox} In summary,  we only observe agreement within method and not across explainability methods. Furthermore, the observed patterns for the two similarity metrics are very different, indicating that the choice of the similarity metric is essential. \end{graybox}

% In summary, the SHAP methods and CEM seem to deliver the most similar explanations, while the feature importance scores obtained with LIME are different from the other two methods.

% \subsection{RQ2: Quantitatively Comparing Explainer Pairs}
% Our primary analysis aims to quantitatively compare the explanations of the different methods over each pair of courses. In this project, we first observe the top 10 important features given by each method over all pairs of courses. Then we examined the distance between the feature importance scores per explainability method in comparison to each other, instead of individually. We computed Spearman’s Rank-Order Correlation as well as Jensen-Shannon Distance to assess similarity regarding the feature importance ranking as well as their exact values.

\subsection{RQ2: Contrasting Settings}
In a second analysis, we investigated whether the explainability methods could capture controlled differences across pairs of courses. While we performed comparisons for all five pairs of courses, we only display the two pairs presented to the educators in the user study (one pair for MOOC vs. flipped and one pair for active vs. passive learning). The results for the other three pairs are included in our Github repository.

\begin{figure}
    \centering
    \includegraphics[trim={0 20 0 0},clip, width=\textwidth]{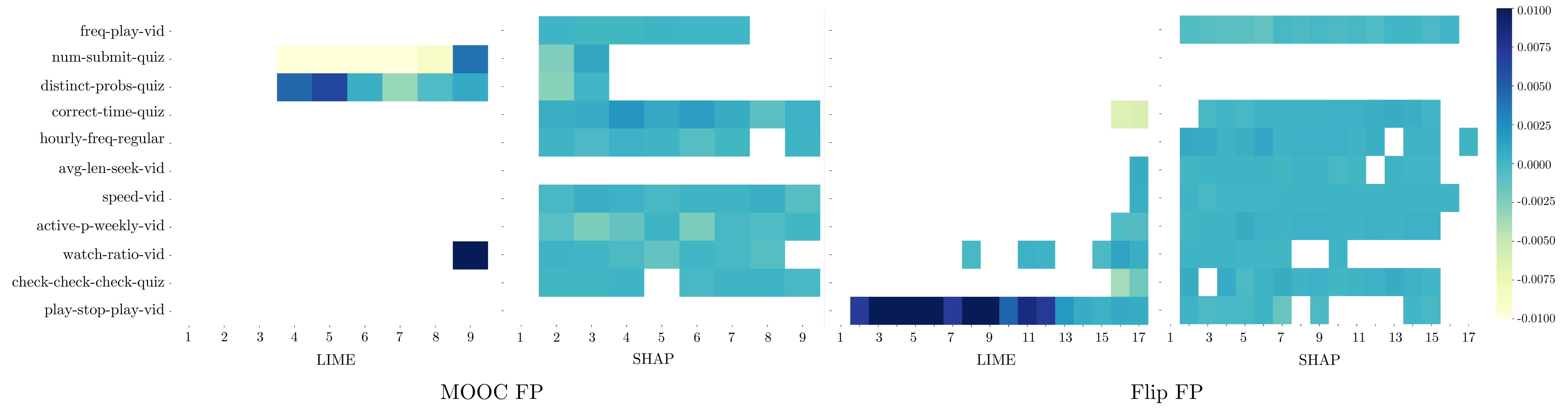}
    \caption{Flipped vs. MOOC Functional Programming course; showcasing absolute, normalized importance scores across weeks.}
    \label{fig:flip-mooc}
\end{figure}

\begin{figure}
    \centering
    \includegraphics[trim={0 15 0 0},clip, width=\textwidth]{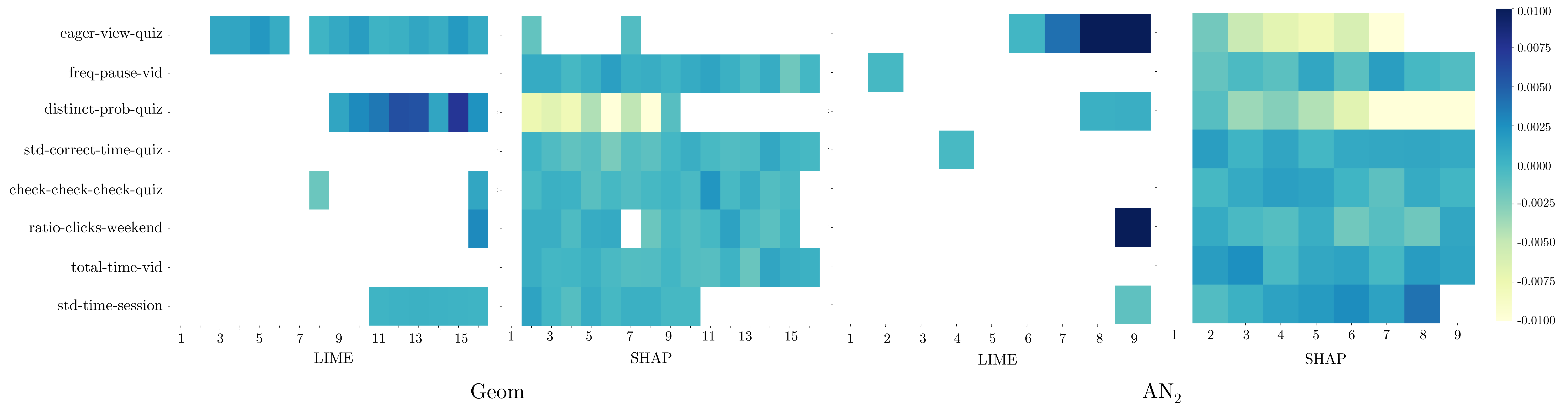}
    \caption{Low vs. High active learning: Geo and $AN_2$ course; showcasing absolute, normalized importance scores across weeks.}
    \label{fig:geom-an2}
\end{figure}

Fig. \ref{fig:flip-mooc} and Fig. \ref{fig:geom-an2} show heatmaps of feature importance scores aggregated across students for four courses. The selected features for each plot consist of the top positive feature and top negative feature for each method (LIME, SHAP, and the confounder) and course. This entails a potential maximum of $12$ features per pair; we obtain eight features for the active learning dimension (implying higher important feature overlap) and $11$ features for the setting dimension (low feature overlap). Feature importance scores below $|0.0001|$ are not displayed. We observe significant sparsity in LIME as opposed to SHAP, which could be useful in downstream interventions. The features are described in Table \ref{tab:features}. 

We examined more detailed relationships for each pair using the displayed importance scores as well as the top ten feature distributions for each method and course. At a high level, we observed in Fig. \ref{fig:flip-mooc} that LIME is far sparser than SHAP; SHAP tends to select substantial features with low weights, while LIME picks sparse features with high weights. Note that this is due to the default parameters of LIME, where the algorithm is restricted to ten features. In the MOOC setting, LIME chooses primarily problem features as important (notably \textit{num-submit-quiz} for failure and \textit{distinct-probs-quiz} for success), but in the flipped setting, LIME selects video features, especially \textit{play-stop-play-vid}, and focuses on the last weeks of the course. SHAP has a wide distribution, focusing on both video, quiz, and regularity features with a lack of emphasis on \textit{num-submit-quiz} and \textit{distinct-probs-quiz}, which are both indicative of student effort. In Fig. \ref{fig:geom-an2}, we see that \textit{distinct-probs-quiz} is regarded as important negative indicator of student performance for both courses by SHAP. However, in LIME, it is considered neutral or positive between both courses. LIME also places a positive weight on \textit{eager-view-quiz}, a proactivity feature, but SHAP considers it negatively. While the direction is different, we note that for the top polarized features, the absolute importance is recognized by both LIME and SHAP.

\begin{graybox} In summary, a few learning science insights are supported and re-confirmed by the importance scores for both methods and course pairs. Video features are predictive across the board. Early week behavior is particularly important in MOOCs, likely due to dropout \cite{mooc-dropout}. In flipped classrooms and across differences in student population, specific problem patterns and timely engagement could be intuitive choices as course health metrics. Students who view or attempt elements early consistently (proactivity) are likely to pass. However, this section does not provide strongly generalizable insights due to a lack of scale. \end{graybox}

\subsection{RQ3: Expert Trust}
In our third analysis, to assess the validity of our explanations, we examined the results of $21$ interviews with non-pilot educators who ranked the four features they believed that differentiate the most passing students from failing students.

\textit{Active-participation-weekly-vid} was the most selected feature in the setting dimension, mostly as passing for Flip FP (six times) and equally as passing and failing for MOOC FP (four and five mentions respectively). \textit{Total-time-vid} was similarly often selected as an indicator for the high active learning setting, six times for passing and six times for failing. While video features were primarily considered as important by the educators, \textit{distinct-probs-quiz} was selected by experts seven time as a passing differentiator in the high active learning case (Geo).

After seeing the explanations from the three explainability methods (i.e., last section of Fig. \ref{fig:pipeline}), the educators chose the method they trusted most. In Flip FP, $80\%$ of participants trusted the SHAP explanation the most; the others were split equally between LIME and the confounder ($10\%$ each). For MOOC FP, the distribution was split almost evenly. $40\%$ of participants preferred the confounder, $30\%$ preferred LIME and SHAP each. Participants had the same average score for how much the methods agree with each other ($2.8$) with a slightly higher standard deviation for the MOOC course over the flipped course ($0.92$ and $0.79$ respectively). Only $2$ of $10$ participants chose the same method for both courses. A possible reason for this is that participants select the explanation per course that most confirms their a-priori beliefs (which often differ based on the educational aspect) instead of choosing consistently based on perceived explanation quality. This could also be interpreted as the participants' lack of clarity on whether the methods agree or disagree.

We also examined whether the features the participants selected at the beginning of the study correlated with their trusted explainer. In $26.49\%$ of cases, the features respectively selected by the participants as the most indicative of passing or failing were in agreement with the explainability method they chose. More specifically, they picked the pass-fail designation in agreement $38.6\%$ of the time in AN$_2$, $25\%$ of the time in Geo, $21\%$ of the time for Flip FP, and $20\%$ of the time for MOOC FP. Overall, study participants were better at picking passing features aligned with the explanation they trusted most over failing features ($29\%$ over $23\%$). However, none of these numbers surpass even $50\%$ accuracy, which means that the experts were not particularly accurate (with respect to their chosen explanation) about whether the features were more indicative of passing or failing students. We then hypothesized that even if the direction of important features was incorrect, perhaps the educators decided to choose important features aligned with their favored explanation nonetheless. $46.8\%$ of features the experts chose were in the top three features of the explanation they chose as important. Notably, for AN$_2$ (high active learning), educators selected features that were $68.2\%$ aligned with their chosen explanation. From this, we can conclude that educators chose features that are at least partially aligned with at least one explanation, validating the results of the explainers.

In the active learning subset, we observe much clearer preferences. For $AN_2$ with a high active learning score, a notable preference for the confounder ($45.5\%$) and SHAP ($54.5\%$) exists, with nobody trusting the LIME explanation. However, in the low active learning case (Geo), we see a strong preference for the LIME explanation ($81.8\%$), with the remaining $19.2\%$ preferring the confounder explanation. The difference between the preferences for similar courses is evident; experts do not agree on a gold standard explainer across different models, even with the same predictive method and feature sets. Only one participant chose the same method as most trustworthy for both courses. In determining the agreement of the XAI methods, the average rating was $2.09$ (on a Likert Scale of $5$) for $AN_2$ with standard deviation $0.94$ and $2.54$ for Geo with standard deviation $0.69$. This first result indicates that participants were very polarized on the $AN_2$ explanations, with some finding that the methods disagree extremely and some finding that they agree moderately; no participants in the low active learning setting said the explanations strongly agree with each other (a rating of 5).

\begin{graybox} In summary, educators had diverse prior perceptions of what factors enable student success and failure; this led them to trust explanations that aligned with their beliefs (which led to significant disagreement across experts). Even more notably, experts were not consistent with their XAI method preferences across multiple courses; only three participants of the $21$ chose the same method for both courses in the pair. \end{graybox}

\subsection{RQ4: Actionable Course Design}
In our last analysis, we were interested in understanding how educators make course design decisions based on XAI insights. We found that $85.71\%$ educators were able to come up with at least one insight to improve course design. The two authors of this paper that conducted each interview independently coded each response with one or more categories depending on the suggestions given by the participant. We then computed Cohen's kappa on the binary outcome of tagging each response with each category, averaged over all categories. The inter-annotator agreement is high (a Cohen's kappa of $0.845$ for the active learning scenario and $0.84$ for the MOOC and flipped classroom scenario).

The large diversity of responses and the overwhelming ability of experts to find ideas based on XAI insights imply that XAI results are useful for improving course design and performing holistic analytics. To make XAI insights even more useful, we were often suggested to have more granularity or the ability to further drill down the XAI importance score based on feature distribution values. We also found that the inability to make clear linear causal statements was hard to grasp; we cannot say that a high amount of time watching videos is a contributor to student success, but we can say the amount of time watching videos is important for passing students (either positively or negatively correlated).

In the last part of the interview, we presented the important indicators of student success (and failure) as the features whose importance scores change the most positively (and negatively) across the educational facet. We also included a broad categorization of when in the course these features are important (beginning, middle, end, or throughout the course). For the setting dimension the presented success indicators were \textit{check-check-check-quiz} (end of course) and \textit{correct-time-quiz} (throughout the course). The failure indicators were \textit{attendance-rate} (end of course) and \textit{content-alignment} (middle of course). Based on these insights, $70\%$ of the interviewees suggested modifying the curriculum (e.g., reordering the course material). $55\%$ of answers suggested redistributing problems throughout the course. Other less popular suggestions mentioned by at least two participants involved pausing or shortening videos, including quizzes as part of videos, sending students reminders, and increasing cross-student interaction.

In the active learning setting, the important student success indicators were \textit{play-pause-load-vid} (start of course) and \textit{pause-speedchange-play-vid} (mid course). As failure indicators, we showed \textit{content anticipation} (end of course) and \textit{correct-time-quiz} (end of course). $45.5\%$ of responses focused on redistributing problems through the course; often the logic was that the time to correctly answer a quiz question was getting longer towards the course end because the quizzes were immediately diving into complex questions. Several educators proposed warming up to the difficult questions with easier initial review, or to spread question difficulty along weeks to discourage dropout towards the course end. $36.4\%$ of participants suggested shorter videos. $27.7\%$ suggested changing the structure (e.g., concepts ordering). Other suggestions included adding personalized student practice or sending deadline reminders.

The majority of participants across both scenarios agreed that the aggregated XAI indicators of student success and failure were logically feasible when comparing courses. We saw a feasibility agreement score of $4$ out of $5$ for the setting dimension and $3.81$ out of 5 for the active learning dimension, with a $0.82$ and $0.87$ standard deviation respectively. Hence, the comparison insights aggregated over all the explainers are more trustworthy than any explainer individually.

\begin{graybox} In summary, expert educators found insights generated across contrastive course settings actionable. While the suggested actions differed across experts, over $85\%$ of educators were able to use insights to derive actions for improving course design. \end{graybox}

\section{Conclusion}
Over the past years, neural network models for education have gained in popularity. They are, however, still rarely used by educators. This is likely due to a lack of transparency, making actionable interventions impossible and decreasing educators' trust \cite{nazaretsky2022teachers}. In this paper, we took a step towards tackling these explainer trustworthiness issues by 1) implementing explainability methods and student success prediction models over nine diverse course settings, 2) quantitatively validating explanations across course pairs that differ in one significant educational aspect, and 3) investigating educators trust in these explanations as well as their perceived usefulness for adapting course design. 

In a first analysis, we compared the different explainability methods across nine courses, including two flipped classrooms and seven MOOCs on EdX and Coursera. By using \textit{Jaccard Similarity} and \textit{Spearman's Rank-Correlation}, we found that the choice of the explainability method has a larger influence on the obtained feature importance score than the choice of the specific course, confirming the findings of prior work \cite{swamy22b}.

In a second step, we compared courses with contrastive settings. We analyzed the top ten features with the highest importance scores from LIME and SHAP. Our results demonstrate that 1) early week behavior is important in MOOCs, 2) interaction clickstreams are very important for flipped classrooms, 3) focusing on student behavior (i.e. lecture-watching delays) as a course health metric could be intuitive across differences in student population, and 4) the students who show anticipation (doing things early consistently) are generally very performant. While the two methods did not agree on the pass-fail predictive direction of the exact features, they agreed on the type of features important for course success. A limitation of this study is the scale, as we only examined five pairs of courses hosted by one university. Although we could not make strongly generalizable conclusions, we contribute to the corpora of educational science literature.

Thirdly, we conducted $21$ semi-structured expert interviews and five pilots to build trust in explainers. We identified that the educators were split amongst themselves across which methods they trust most, and many times even chose the confounder explanation which was designed to be misleading. From the interview feedback, many participants requested more concrete and granular insights and background about student demographics or past knowledge. We conclude that providing the averaged feature importance scores is not enough for educators to decide on concrete steps. It is possible that explainability methods are more effective on an individual scale (per student) than on a global scale (per course). We leave the further design of intuitive explanation visualizations, using more granular XAI insights, and measuring efficacy of instructor XAI-enabled suggestions to a future study. Note also that the selection of participants was biased by the networks of the authors of this paper (tending towards STEM professors); this is important in interpreting our results and the ease with which participants understood the technical background of the methods.

To conclude, XAI has a high potential in personalizing education, but the systematic differences between explainers is a significant barrier to educators trusting any one explanation. Future work is needed to make the explanations human interpretable. We observed that even without believing the explanations agree with each other, educators find that the provided insights can inspire ideas for improving course design and bolstering student performance.

%% file: main/06_multimodn.tex
This work is a collaboration across three laboratories -- the ML for Education Lab (ML4ED), the ML and Optimization Lab (MLO), and the Lab for Intelligent Global Health Technologies (LiGHT) -- with Malika Satayeva, Jibril Frej, Thierry Bossy, Thijs Vogels, Professor Martin Jaggi, Professor Tanja Käser, and Professor Mary-Anne Hartley. It was presented under the title \textit{MultiModN - Multimodal, Multi-task, Interpretable Modular Networks} at the Advances in Neural Information Processing Systems Conference (NeurIPS) in 2023.

We present \md, a state-of-the-art multimodal architecture that is interpretable-by-design and robust to missing data modalities without compromising accuracy. \md produces \textbf{accurate} and \textbf{real-time} explanations as explanations are directly used in constructing model predictions. It additionally addresses explanation \textbf{consistency}, as the same inputs to the same model will always produce the same explanation. The \textbf{actionability} and \textbf{human understandability} of explanations are dependent on user implementations, as users could represent each modular encoder with a grouping of features of their choice, instead of a modality.

\section{Introduction}

The world is richly multimodal and intelligent decision-making requires an integrated understanding of diverse environmental signals, known as embodied intelligence \cite{Gupta2021mm-embodied}.
Until recently, advances in deep learning have been mostly compartmentalized by data modality, creating disembodied domains such as computer vision for images, natural language processing for text, and so on. Multimodal (MM) learning has emerged from the need to address real-world tasks that cannot be robustly represented by a single signal type as well as the growing availability and diversity of digitized signals \cite{baltrusaitis2019mmreview, guo2019mmreview, barua2023mmreview}. 
Some examples are diagnosis from a combination of medical tests and imagery \cite{fan2020mmmt-cancer, Boehm2022mm-med, Acosta2022mm-med}, estimating sentiment from facial expression, text, and sound \cite{li2021mmmt-audiovis,poria2018mm-sentiment,majumder2018mm-sentiment,zadeh2018mm-sentiment}, and identifying human activities from a combination of sensors \cite{ahmad2020mmmt-har}.

The richer representations from synergistic data types also have the potential to increase the task space, where a single set of representations can generalize to several tasks. Multi-task (MT) learning has not only been shown to benefit the performance of individual tasks but also has the potential to greatly reduce computational cost through shared feature extraction \cite{crawshaw2020mtreview}.

In short, multimodal and multi-task learning hold significant potential for human-centric machine learning and can be summarized respectively as creating a shared feature space from various data types and deriving their relative semantic meaning across several tasks.

 \begin{figure}[ht]
\centering
\begin{subfigure}[b]{0.634\textwidth}
\includegraphics[width=\textwidth]{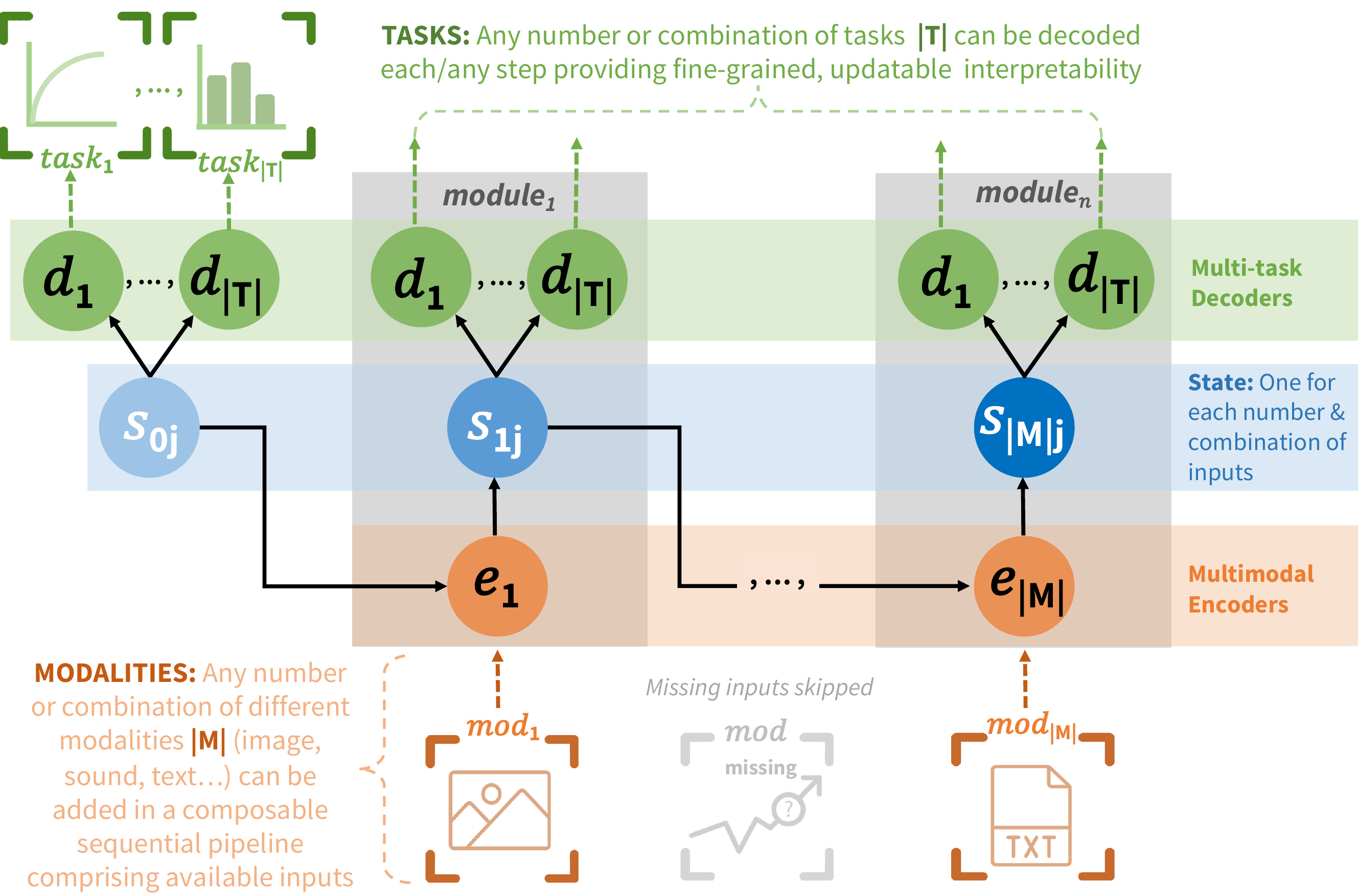}
\caption{\md}
\label{fig:overview-a}
\end{subfigure}
\begin{subfigure}[b]{0.35\textwidth}
\includegraphics[width=\textwidth]{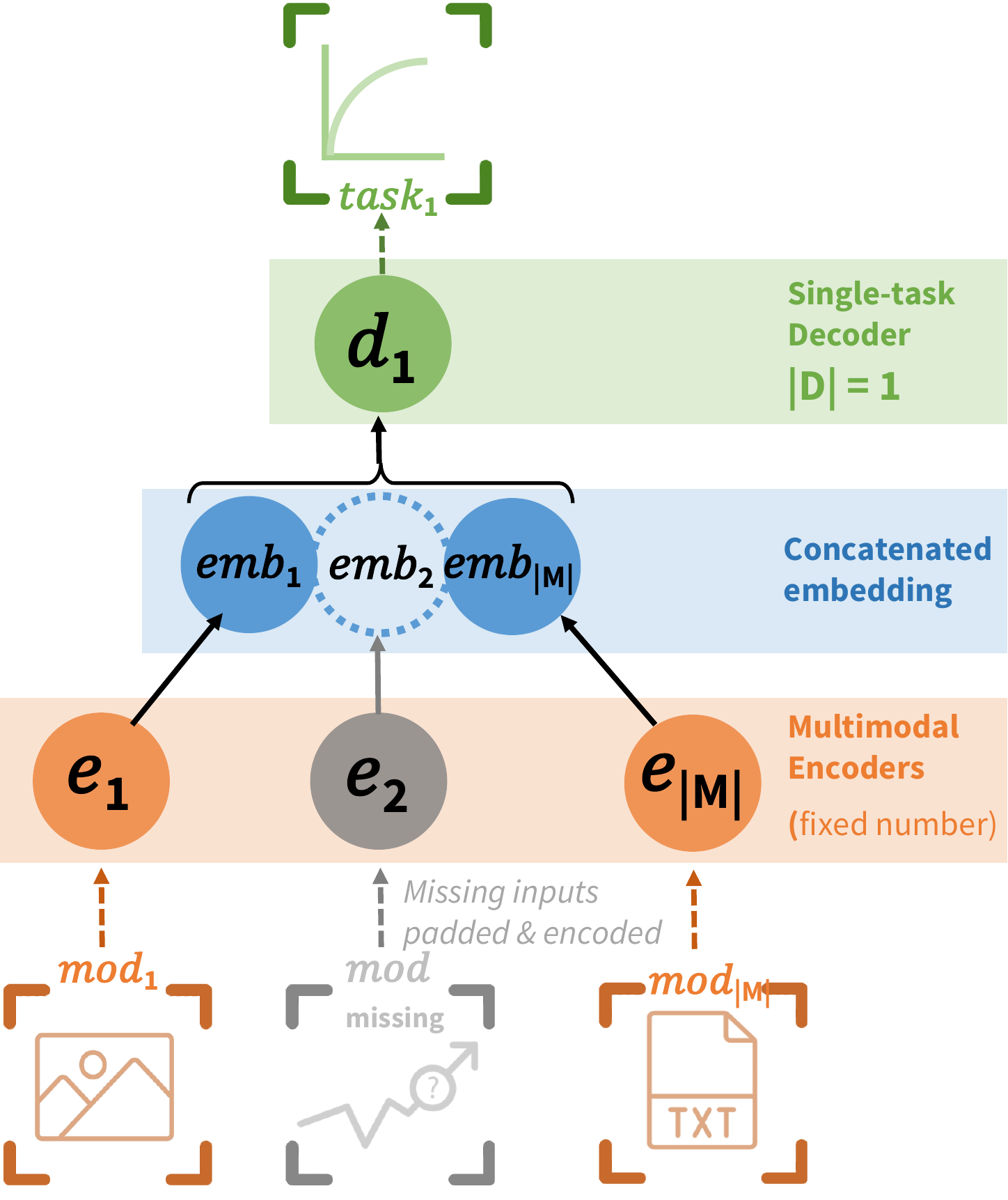}
\caption{\pf}
\label{fig:overview-b}
\end{subfigure}
    \caption{Comparison of modular \md \textbf{(a)} vs. monolithic \pf \textbf{(b)}. \md inputs any number/combination of modalities ({\color{orange}{$\textbf{mod}$}}) into a sequence of $mod$-specific encoders ({\color{orange}{$\textbf{e}$}}). It can skip over missing modalities. A state ({\color{RoyalBlue}{$\textbf{s}$}}) is passed to the subsequent encoder and updated. Each state can be fed into any number/combination of decoders ({\color{YellowGreen}{$\textbf{d}$}}) to predict multiple tasks. \textit{Modules} are identified as grey blocks comprising an encoder, a state, and a set of decoders.
   \pf is a monolithic model. It inputs a \textit{fixed} number/combination of modalities ({\color{orange}{$\textbf{mod}$}}) into $mod$-specific encoders ({\color{orange}{$\textbf{e}$}}). Missing modalities are padded and encoded. Embeddings ({\color{RoyalBlue}{$\textbf{emb}$}}) are concatenated and provided to a \textit{single} decoder in parallel ({\color{YellowGreen}{$\textbf{d}$}}) to predict a \textit{single} task.}
   \label{fig:overview}
\end{figure}

\textbf{Limitations of current multimodal models. } Current MM models propose a parallel integration of modalities, where representations are fused and processed simultaneously \cite{baltrusaitis2019mmreview, guo2019mmreview, barua2023mmreview}. Parallel fusion (hereafter \pf) creates several fundamental limitations that we address in this work. 

The most important issue we seek to resolve in current MM architectures is their \textit{dependence on modality availability} where all modalities for all data points are required inputs during both training and inference. Modality-specific missingness is a common real-world problem and can fundamentally bias the model when the missingness of a modality is predictive of the label (known as missing not-at-random, MNAR). The common solution of restricting learning to data points with a complete set of modalities creates models that perform inequitably in populations with fewer available resources (i.e. when the pattern of MNAR is different between train and test sets). In complex real-world datasets, there is often no intersection of complete availability, thus necessitating the exclusion of modalities or significantly limiting the train set. On the other hand, imputation explicitly featurizes missingness, thus risking to create a trivial model that uses the \textit{presence} of features rather than their value for the prediction \cite{graham2009missing, Emmanuel2021-missing}.
The MNAR issue is particularly common in medicine, where modality acquisition is dependent on the decision of the healthcare worker (i.e. the decision that the model is usually attempting to emulate). For example, a patient with a less severe form of a disease may have less intensive monitoring and advanced imagery unavailable. If the goal is to predict prognosis, the model could use the missingness of a test rather than its value. This is a fundamental flaw and can lead to catastrophic failure in situations where the modality is not available for independent reasons (for instance resource limitations). Here, the featurized missingness would inappropriately categorize the patient in a lower severity class.
For equitable real-world predictions, it is critical to adapt predictions to available resources, and thus allow composability of inputs at inference.

Another key issue of current techniques that this work addresses is \textit{model complexity}. Parallel fusion of various input types into a single vector make many post-hoc interpretability techniques difficult or impossible \cite{joshi2021mm-interp}. Depending on where the fusion occurs, it may be impossible to decompose modality-specific predictive importance.

In this work, we leverage network modularization towards interpretability objectives, compartmentalizing each modality and task into independent encoder and decoder modules that are inherently robust to the bias of MNAR and can be assembled in any combination or number at inference while providing continuous modality-specific predictive feedback.

\textbf{Contributions.} We propose \md, a multimodal extension of the work of Trottet et al. \cite{modn2022}, which uses a flexible sequence of model and task-agnostic encoders to produce an evolving latent representation that can be queried by any number or combination of multi-task, model-agnostic decoder modules after each input (showcased in Figure \ref{fig:overview}). 
Specifically, we demonstrate that our modular approach of sequential MM fusion:

\begin{enumerate}
    \item \textbf{matches parallel MM fusion} (\pf) for a range of real-world tasks across several benchmark datasets, while contributing distinct advantages, such as being: 
    \item \textbf{composable at inference}, allowing selection of any number or combination of available inputs, 
    \item \textbf{is robust to the bias of missing not-at-random (MNAR) modalities},
    \item \textbf{is intrinsically interpretable}, providing granular modality-specific predictive feedback, and 
    \item \textbf{is easily extended to any number or combination of tasks}.
\end{enumerate}
We provide an \textbf{application-agnostic open-source framework} for the implementation of \md: \url{https://github.com/epfl-iglobalhealth/MultiModN}.  Our experimental setup purposely limits our model performance to fairly compare the multimodal fusion step. At equivalent performance, our model architecture is by far superior to the baseline by virtue of being inherently modular, interpretable, composable, robust to systematic missingness, and multi-task.

\section{Background}

Approaches to MM learning can be categorized by the depth of the model at which the shared feature space is created \cite{baltrusaitis2019mmreview}. 
Late fusion (decision fusion) processes inputs in separate modality-specific sub-networks, only combining the outputs at the decision-level, using a separate model or aggregation technique to make a final prediction. While simple, late fusion fails to capture relationships between modalities and is thus not \textit{truly} multimodal.
Early fusion (feature fusion), combines modalities at the input level, allowing the model to learn a joint representation. Concatenating feature vectors is a popular and simple approach \cite{huang2020fusion, kline2022multimodal}, but the scale of deployment is particularly limited by the curse of dimensionality.
Finally, intermediate fusion (model fusion) seeks to fine-tune several feature extraction networks from the parameters of a downstream classifier.

\textbf{Parallel Multimodal Fusion (\pf). }
Recently, Soenksen et al. \cite{haim2022} proposed a fusion architecture which demonstrated the utility of multiple modalities in the popular MM medical benchmark dataset, MIMIC \cite{mimic-iv, johnson2019mimic}. Their framework (HAIM or Holistic Artificial Intelligence in Medicine) generates single-modality embeddings, which are concatenated into a single one-dimensional multimodal fusion embedding. The fused embedding is then fed to a single-task classifier. 
This work robustly demonstrated the value of multimodality across several tasks and a rich combination of heterogeneous sources. HAIM consistently achieved an average improvement of 6-33\% AUROC (area under the receiver operating characteristic curve) across all tasks in comparison to single-modality models. We use this approach as a \pf baseline against our sequential fusion approach of \md and extend it to several new benchmark datasets and multiple tasks. 

Soenksen et al. \cite{haim2022} perform over 14,324 experiments on 12 binary classification tasks using every number and combination of modalities. This extreme number of experiments, was necessary because the model is not composable nor capable of multi-task (MT) predictions. Rather, a different model is needed for each task and every combination of inputs for each task. In contrast, \md is an extendable network, to which any number of encoders and decoders can be added. Thus, most of the 14,324 experiments could technically be achieved within one \md model. 

Several other recent architectures utilize parallel fusion with transformers. UNiT (Unified Transformer) \cite{hu2021unit} is a promising multimodal, multi-task transformer architecture; however, it remains monolithic, trained on the union of all inputs (padded when missing) fed in parallel. This not only exposes the model to patterns of systematic missingness during training but also reduces model interpretability and portability\footnote{The equivalent transformer architecture has 427,421 trainable parameters for the EDU dataset (Sec. \ref{sec:datasets}) while \md achieves better performance with 12,159 parameters.}. \cite{ma2022multimodal}’s recent work has found similar results on the erratic behavior of transformers to missing modalities, although it is only tested on visual/text inputs. LLMs have also recently been used to encode visual and text modalities \cite{alayrac2022flamingo}, but it is not clear how tabular and time-series would be handled or how this would affect the context window at inference. Combining predictive tasks with LLMs will also greatly impact interpretability, introducing hallucinations and complex predictive contamination where learned textual bias can influence outcomes.

\textbf{Modular Sequential Multimodal Fusion. }
A \textit{module} of a modular model is defined as a self-contained computational unit that can be isolated, removed, added, substituted, or ported. It is also desirable for modules to be order invariant and idempotent, where multiple additions of the same module have no additive effect. We design \md to encode individual inputs, whereby module-exclusion can function as input \textit{skippablity}, allowing missing inputs to be skipped without influencing predictions.
Thus, modular models can have various input granularities, training strategies, and aggregation functions. Some popular configurations range from hierarchies with shared layers to ensemble predictions and teacher-trainer transfer learning approaches \cite{pfeiffer2023-modular, Amer2019-modular}. 

We expand on the sequential modular network architecture proposed by Trottet et al.\cite{modn2022} called MoDN (Modular Decision Networks) as a means of sequential MM fusion. MoDN trains a series of feature-specific encoder modules that produce a latent representation of a certain size (the \textit{state}). Modules can be strung together in a mix-and-match sequence by feeding the state of one encoder as an input into the next. Therefore, the state has an additive evolution with each selected encoder. A series of decoders can query the state at any point for multiple tasks from various combinations of inputs, giving MoDN the property of combinatorial generalization.

Thus, we extend MoDN to learn multiple tasks from multimodal inputs. By aligning feature extraction pipelines between \md and the \pf baseline (inspired by HAIM) we can achieve a better understanding of the impact of monolithic-parallel fusion vs. sequential-modular MM fusion. Figure \ref{fig:overview} provides a comparison between \pf and \md, also formalized below.

\section{Problem formulation}
\textbf{Context. }We propose a multi-task supervised learning framework able to handle any number or combination of inputs of varying dimension, irrespective of underlying bias in the availability of these inputs during training.
We modularize the framework such that each input and task is handled by distinct encoder and decoder \textit{modules}. 
The inputs represent various data modalities (i.e. image, sound, text, time-series, tabular, etc.). We assume that these inputs have synergistic predictive potential for a given target and that creating a multimodal shared feature space will thus improve model performance. 
The tasks represent semantically related observations. We hypothesize that jointly training on semantically related tasks will inform the predictions of each individual task.

\textbf{Notation. }Formally, given a set of modalities (features) $\mathcal{M} = \{mod_1, \dots, mod_{|\mathcal{M}|}\}$ and a set of tasks (targets) $\mathcal{T} = \{ task_1, \dots, task_{|\mathcal{T}|} \}$, let $\mathcal{X} = \left\{ 
    (x_1, y_1),
    (x_2, y_2),
    %(x_{2(mod_1 \dots mod_{|\mathcal{M}|})}, y_{2(task_1 \dots task_{|\mathcal{T}|})}), 
    \dots, 
    (x_N, y_N)\right\}$ represent a multimodal, 
multi-task dataset with $N$ data points ($x_1, \dots, x_N$).  

Each point $x$ has ${|\mathcal{M}|}$ modalities (inputs): $x = \left\{ x_{mod_1}, \dots, x_{mod_{|\mathcal{M}|}}  \right\}$ and is associated with a set of $|\mathcal{T}|$ targets (tasks): $y = \left\{ y_{task_1}, \dots, y_{task_{|\mathcal{T}|}}  \right\}$.
Modalities comprise various sources (e.g. images from x-rays, CT), for simplicity, we consider sources and modalities as equal $mod$ elements in $\mathcal{M}$.

\textbf{Multimodal, multi-task, modular formulation.} We decompose each data point $x$ into $|\mathcal{M}|$ sequential encoder \textit{modules} specific to its constituent modalities
and each target $y$ into $|\mathcal{T}|$ decoder \textit{modules} specific to its constituent tasks 
such that any combination or number of modalities can be used to predict any combination or number of tasks. 
Our objective is to learn a set of function \textit{modules}, $\mathcal{F}$. Each function \textit{module} within this set, represented as $f_j^i \in \mathcal{F}$ maps combinations of modalities $\mathcal{M}_j$ to combinations of tasks $\mathcal{T}_i$, i.e. $f_j^i: \mathcal{M}_j \rightarrow \mathcal{T}_i$. It is important to note that $\mathcal{M}_j$ is an element of the powerset of all modalities and $\mathcal{T}_i$ is an element of the powerset of all tasks.

\textbf{Extension to time-series.}
In the above formulation, the $\mathcal{M}$ encoder \textit{modules} are handled in sequence, thus naturally aligning inputs with time-series. While the formulation does not change for time-series data, it may be optimized such that $f_j^i$ represents a single time step. This is relevant in the real-world setting of a data stream, where inference takes place at the same time as data is being received (i.e. predicting student performance at each week of a course as the course is being conducted). The continuous prediction tasks (shown for EDU and Weather in Sec. \ref{sec:experiments}) demonstrate how \md can be used for incremental time-series prediction.

\section{\md: Multimodal, Multi-task, Modular Networks (Our model)}
Building on \cite{modn2022} (summarized and color-coded in Figure \ref{fig:overview-a}), the \md architecture consists of three modular elements: a set of {\color{RoyalBlue}{\textbf{State}}} vectors {\color{RoyalBlue}$\mathcal{S} = \left\{ s_0, \dots, s_{|\mathcal{M}|} \right\}$}, a set of modality-specific {\color{Orange}{\textbf{Encoders $\mathcal{E} = \left\{ e_1, \dots, e_{|\mathcal{M}|} \right\}$}}} , and a set of task-specific {\color{Green}{\textbf{Decoders $\mathcal{D} = \left\{ d_1, \dots, d_{|\mathcal{T}|} \right\}$}}}. State $s_0$ is randomly initialized and then updated sequentially by $e_i$ to $s_i$. Each $s_i$ can be decoded by one, any combination, or all elements of $\mathcal{D}$ to make a set of predictions. All encoder and decoder parameters are subject to training.

{\color{RoyalBlue}\textbf{States ($\mathcal{S}$)}}\textbf{.}  Akin to hidden state representations in Recurrent Neural Networks (RNNs), the state of \md is a vector that encodes information about the previous inputs. As opposed to RNNs, state updates are made by any number or combination of modular, modality-specific encoders and each state can be decoded by modular, task-specific decoders. Thus the maximum number of states by any permutation of $n$ encoders is $n!$. For simplicity, we limit the combinatorial number of states to a single order (whereby $e_i$ should be deployed before $e_{i+1}$) in which any number or combination of encoders may be deployed (i.e. one or several encoders can be skipped at any point) as long as the order is respected. Thus, the number of possible states for a given sample is equal to $2^{|\mathcal{M}|}$. Order invariance could be achieved by training every permutation of encoders $|\mathcal{M}|!$, i.e. allowing encoders to be used in any order at inference, as opposed to this simplified implementation of \md in which order is fixed.
At each step $i$, the encoder $e_i$ processes the previous state, $s_{i-1}$, as an input and outputs an updated state $s_{i}$ of the same size. When dealing with time-series, we denote $s_{t(0)}$ as the state representing time $t$ before any modalities have been encoded, and as $s_{t(0,1,4,5)}$ as the state at time $t$ after being updated by encoders $e_1$, $e_4$ and $e_5$, in that order.

{\color{Orange}\textbf{Encoders ($\mathcal{E}$)}}\textbf{.} Encoders are modularized to represent a single modality, i.e. $|\mathcal{E}| = |\mathcal{M}|$. An encoder $e_{i}$ takes as input the combination of a single modality (of any dimension) and the previous state $s_{i-1}$. Encoder $e_{i}$ then outputs a $s_{i}$, updated with the new modality. For simplicity, we fix the state size between encoders. %such that $s_{i-1}$ and $s_{i}$ have the same size.
Due to modularization, \md is model-agnostic, whereby encoders can be of any type of architecture (i.e. Dense layers, LSTM, CNN). 
For experimental simplicity, we use a single encoder design with a simple dense layer architecture. The input vectors in our experiments are 1D. When a modality is missing, the encoder is skipped and not trained (depicted in Figure  \ref{fig:overview}).

{\color{Green}\textbf{Decoders ($\mathcal{D}$)}}\textbf{.}  Decoders take any state $s_i$ as input and output a prediction. Each decoder is assigned to a single task, that is $|\mathcal{D}| = |\mathcal{T}|$, i.e. \md is not multiclass, but multi-task (although a single task may be multiclass). 
Decoders are also model-agnostic. Our implementation has regression, binary, and multiclass decoders across static targets or changing time-series targets. Decoder parameters are shared across the different modalities. The decoder predictions are combined across modalities/modules by averaging the loss. Interestingly, a weighted loss scheme could force the model to emphasize certain tasks over others. 

As shown in \cite{modn2022}, \md can be completely order-invariant and idempotent if randomized during training. For interpretability, sequential inference (in any order) is superior to parallel input due to its decomposability, allowing the user to visualize the effect of each input and aligning with Bayesian reasoning.

\textbf{Quantification of Modularity. } The modularity of a network can be quantified, whereby neurons are represented by nodes (vertices) and connections between neurons as edges. There are thus comparatively dense connections (edges) within a \textit{module} and sparse connections between them. Partitioning modules is an NP-complete problem \cite{brandes2008-modNP}. We present modules that are defined \textit{a priori}, whereby a module comprises one encoder $e_i$ connected to one state $s_i$, which is in turn connected to a set of $|\mathcal{T}|$ tasks (a \textit{module} is depicted as a grey box in Figure~\ref{fig:overview-a}).
Following a formalization of modularity quantitation proposed by Newman et al.~\cite{PhysRevE.69.026113}, we compute the modularity score for \md and show that it tends to a perfect modularity score of 1 with each added modality and each added task.
When viewed at the network granularity of these core elements, \pf is seen as a monolithic model with a score of 0.
The formula is elaborated in Appendix Sec. \ref{appendix:architecture}.

\subsection{\pf: Parallel Multimodal Fusion (Baseline)}
We compare our results to a recent multimodal architecture inspired by HAIM (Holistic AI in Medicine) \cite{haim2022}. As depicted in Figure \ref{fig:overview-b}, HAIM also comprises three main elements, namely, a fixed set of modality-specific encoders $\mathcal{E} = \left\{ e_1, \dots, e_{|\mathcal{M}|} \right\}$ which create a fixed set of embeddings $\mathcal{B} = \left\{ emb_1, \dots, emb_{|\mathcal{M}|} \right\}$, that is concatenated and fed into a single-task decoder ($d_1$). 
HAIM achieved state-of-the-art results on the popular and challenging benchmark MIMIC dataset, showing consistently that multimodal predictions were between 6\% and 33\% better than single modalities.
%We refer to baseline as \pf.

{\color{Orange}\textbf{Encoders ($\mathcal{E}$)}}\textbf{.}  Contrary to the flexible and composable nature of \md,  the sequence of encoders in \pf is fixed and represents a unique combination of modalities. It is thus unable to skip over modalities that are missing, instead padding with neutral values and explicitly embedding the non-missing modalities. The encoders are modality-specific pre-trained neural networks.

{\color{RoyalBlue}\textbf{Embeddings ($\mathcal{B}$)}}\textbf{.} Multimodal embeddings are fused in parallel by concatenation.

{\color{Green}\textbf{Decoders ($\mathcal{D}$)}}\textbf{.}  Concatenated embeddings are passed to a single-task decoder. 

\textbf{Architecture alignment. } \label{sec:alignment}
We align feature extraction between \md and \pf to best isolate the effect of sequential (\md) vs. parallel  (\pf) fusion. As depicted in Appendix Figure ~\ref{appendix:overview-aligned}, we let \md take as input the embeddings created by the \pf pre-trained encoders. Thus both models have identical feature extraction pipelines. No element of the \md pipeline proposed in Figure~\ref{fig:overview-a} is changed. The remaining encoders and decoders in both models are simple dense layer networks (two fully connected ReLU layers and one layer for prediction). Importantly, \md encoders and decoders are model-agnostic and can be of any architecture.

\section{Datasets}
\label{sec:datasets}
We compare \md and \pf on three popular multimodal benchmark datasets across 10 real-world tasks spanning three distinct domains (healthcare, education, meteorology). The healthcare dataset (MIMIC) is particularly challenging in terms of multimodal complexity, incorporating inputs of vastly varying dimensionality. Education (EDU) and Weather2k have a particular focus on time-series across modalities. Appendix Sec. \ref{appendix:data-setup} details features, preprocessing, and tasks ($task_{1-10}$).

\textbf{MIMIC. }
MIMIC \cite{goldberger2000physiobank} is a set of deidentified electronic medical records comprising over $40,000$ critical care patients at a large tertiary care hospital in Boston. The feature extraction pipeline replicated according to our baseline of \pf \cite{haim2022}, making use of patient-level feature embeddings extracted from pre-trained models as depicted in Appendix Figure~\ref{appendix:overview-aligned}. 
We select the subcohort of $921$ patients who have valid labels for both diagnoses and all four modalities present.
We use all four modalities as inputs: chest x-rays (image), chart events (time-series), demographic information (tabular), and echocardiogram notes (text). For simplicity, we focus on two diagnostic binary classification tasks: cardiomegaly ($task_1$) and enlarged cardiomediastinum ($task_2$). These tasks were selected for their semantic relationship and also because they were reported to benefit from multimodality~\cite{haim2022}. Thus, we have four modality-specific encoders and two binary classification diagnostic decoders.

\textbf{Education (EDU). }
This educational time-series dataset comprises $5,611$ students with over 1 million interactions in a 10-week Massively Open Online Course (MOOC), provided to a globally diverse population. It is benchmarked in several recent works \cite{swamy2022evaluating, boroujeni2019discovery, asadi2022ripple}. Our modeling setting is replicated from related literature, with $45$ handcrafted time-series features regarding problem and video modalities extracted for all students at each weekly time-step \cite{swamy2022meta}. We use two modality-specific encoders (problem and video) and three popular decoder targets: binary classifiers ($task_{3-4}$) of pass/fail and drop-out, and a continuous target of next week's performance ($task_5$) \cite{ye2014early}.

\textbf{Weather2k. } \textit{Weather2k} is a 2023 benchmark dataset that combines tabular and time-series modalities for weather forecasting \cite{zhu2023weather2k}. The data is extracted from $1,866$ ground weather stations covering $6$ million km$^2$, with $20$ features representing hourly interactions with meteorological measurements and three static features representing the geographical location of the station. We create five encoders from different source modalities: geographic (static), air, wind, land, and rain and align with the benchmark prediction targets \cite{zhu2023weather2k} on five continuous regression targets: short (24 hr), medium (72 hr), long term (720 hr) temperature forecasting, relative humidity and visibility prediction ($tasks_{6-10}$).

\section{Experiments}
\label{sec:experiments}

\textbf{Overview. }
We align feature extraction pipelines between \md and the \pf baseline in order to isolate the impact of parallel-monolithic vs. sequential-modular fusion (described in \ref{sec:alignment} and depicted in Appendix Sec. \ref{appendix:architecture}). We thus do not expect a significant difference in performance, but rather aim to showcase the distinct benefits that can be achieved with modular sequential multimodal fusion\textit{ without compromising baseline performance}. In the following subsections, we perform four experiments to show these advantages. \textbf{[1]} \md performance is not compromised compared to \pf in single-task predictions. \textbf{[2]} \md is able to extend to multiple tasks, also without compromising performance. \textbf{[3]} \md is inherently composable and interpretable, providing modality-specific predictive feedback. \textbf{[4]} \md is resistant to MNAR bias and avoids catastrophic failure when missingness patterns are different between train and test settings. 

\begin{figure}[]
    \centering
    \begin{subfigure}[b]{0.33\textwidth}
        \includegraphics[width=\textwidth]{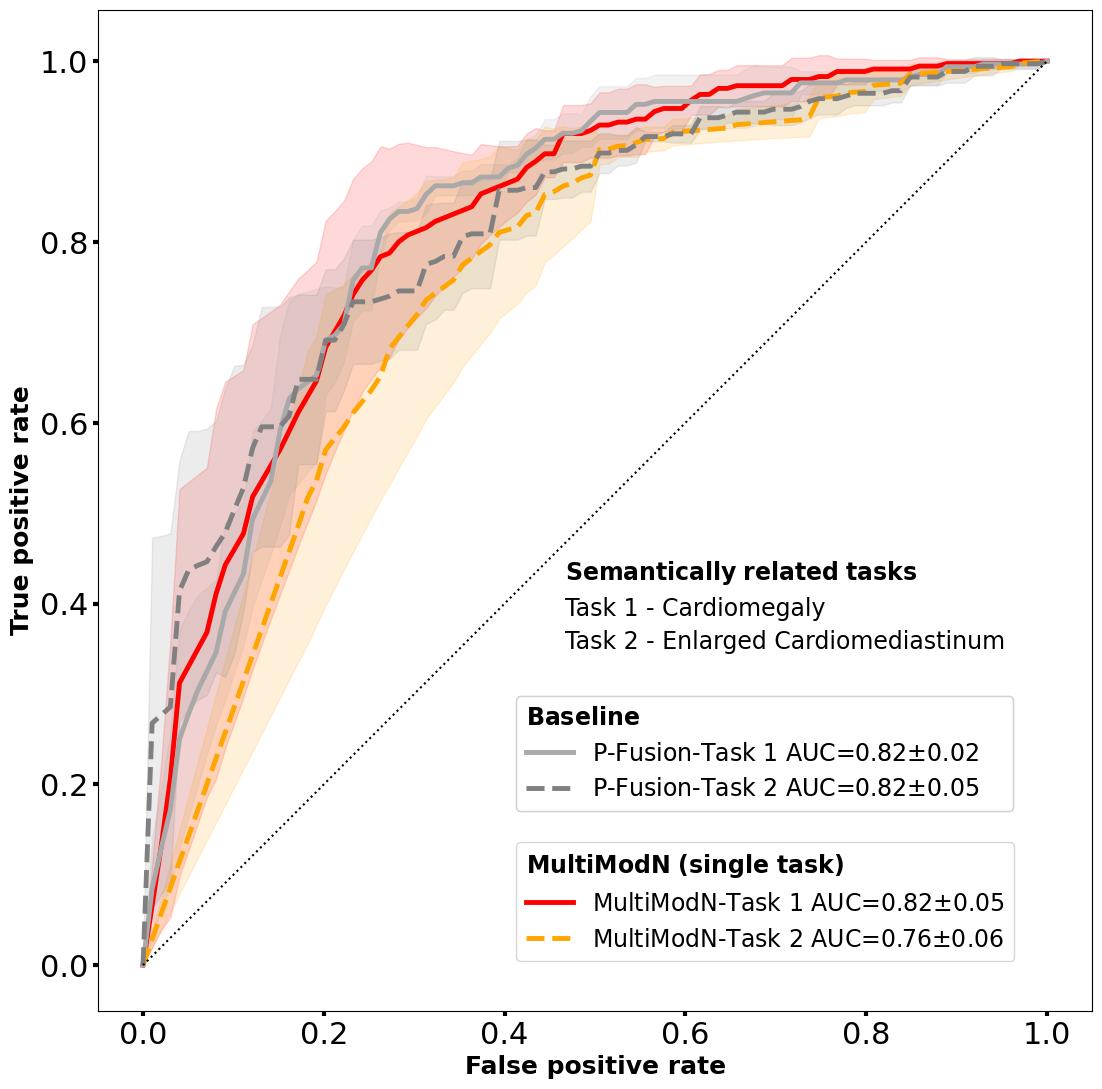}
        \caption{ \textbf{MIMIC} \label{fig:MIMIC-roc}}
    \end{subfigure}
    \begin{subfigure}[b]{0.3107\textwidth}
        \includegraphics[width=\textwidth]{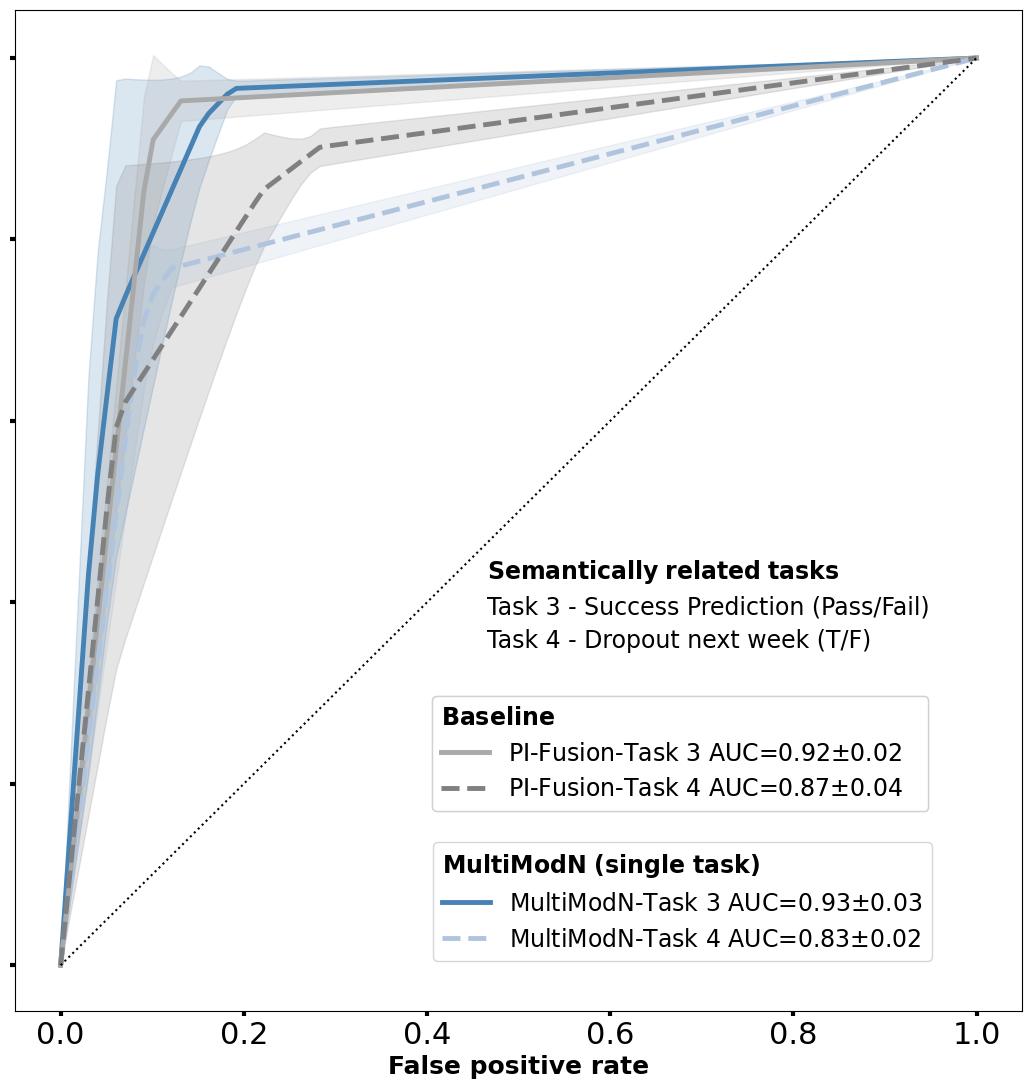}
        \caption{ \textbf{EDU} \label{fig:EDU-roc}}
    \end{subfigure}
          \begin{subfigure}[b]{0.3107\textwidth}
        \includegraphics[width=\textwidth]{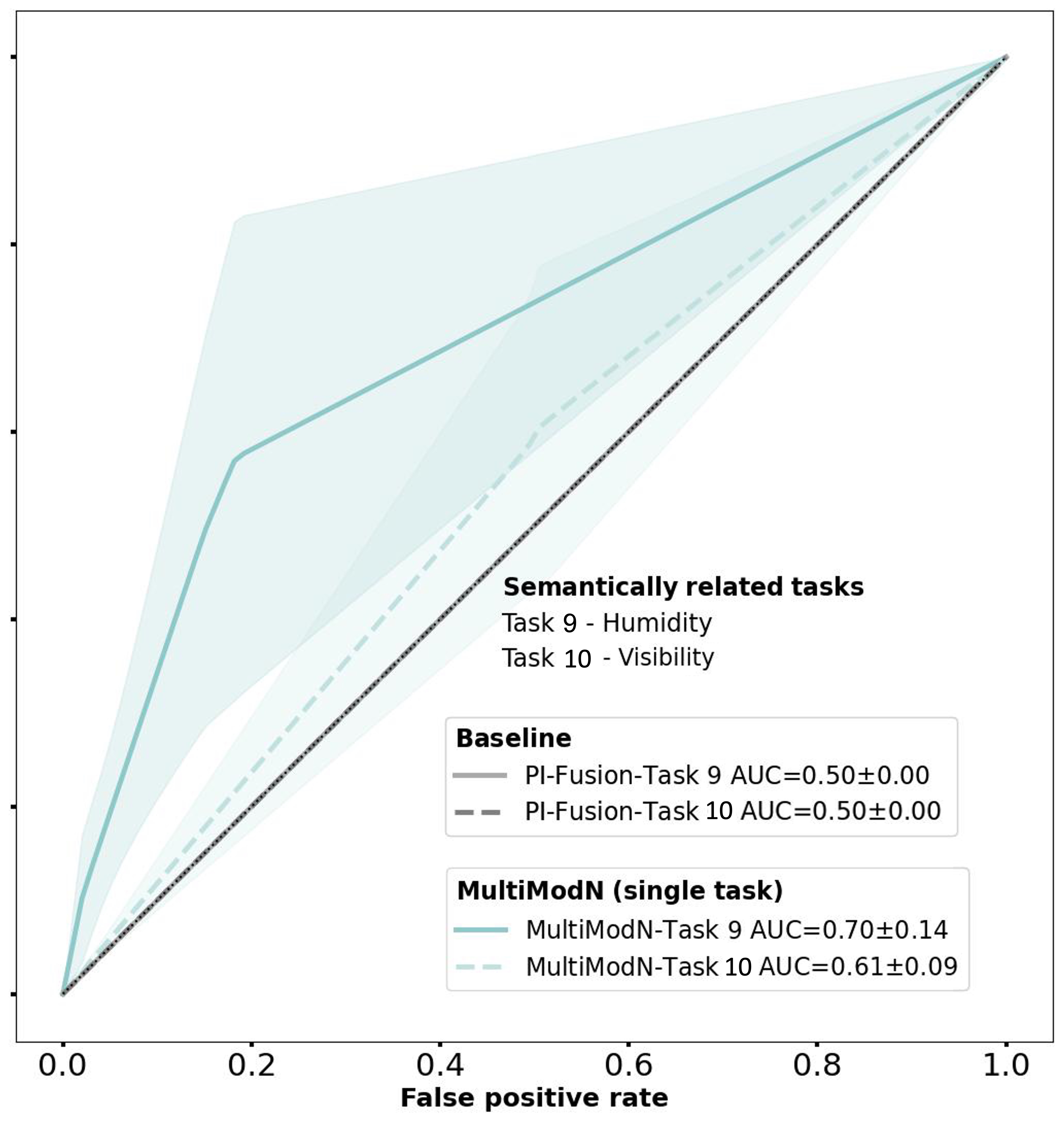}
        \caption{ \textbf{Weather} \label{fig:Weather-roc}}
     \end{subfigure}     
     
    \caption{ \label{fig:all-rocs} \textbf{MultiModN does not compromise performance in single-tasks.} AUROC for six binary prediction tasks in (a) MIMIC, (b) EDU, and (c) Weather2k. Tasks predicted by {\color{gray}{\pf}} are compared with \md. 95\% CIs are shaded.}
\end{figure}

\textbf{Model evaluation and metrics. }
All results represent a distribution of performance estimates on a model trained 5 times with different random weight initializations for the state vector and weights. Each estimate uses a completely independent test set from an 80-10-10 K-Fold train-test-validation split, stratified on one or more of the prediction targets. We report metrics (macro AUC, BAC, MSE) with 95\% confidence intervals, as aligned with domain-specific literature of each dataset \cite{swamy2022meta, haim2022, zhu2023weather2k}. 

\textbf{Hyperparameter selection.} Model architectures were selected among the following hyperparameters: state representation sizes [1, 5, 10, 20, 50, 100], batch sizes [8, 16, 32, 64, 128], hidden features [16, 32, 64, 128], dropout [0, 0.1, 0.2, 0.3], and attention [0, 1]. These values were grouped into 3 categories (small, medium, large). We vary one while keeping the others fixed (within groups). Appendix Figure \ref{appendix:perf-analysis} shows that \md is robust to changing batch size, while dropout rate and hidden layers negatively impact larger models (possibly overfitting). The most specific parameter to \md is state size. As expected, we see negative impacts at size extremes, where small states likely struggle to transfer features between steps, while larger ones would be prone to overfitting.

\subsection{Exp. 1: Sequential modularization in \md does not compromise performance}
\label{exp:single-task}
\textbf{Setup. }
A single-task model was created for each $task_{1-10}$ across all three datasets. Each model takes all modalities as input. We compare \md and \pf in terms of performance. AUROCs can be visualized in Figure \ref{fig:all-rocs} while BAC and MSE are detailed in Table \ref{tab:unitask-results}. As feature extraction pipelines between \md and \pf are aligned, this experiment seeks to investigate if sequential modular fusion compromises model performance.
To compress the multiple predictions of time-series into a single binary class, we select a representative time step (EDU $tasks_{3-4}$ at 60\% course completion) or average over all time steps (Weather $tasks_{9-10}$ evaluated on a 24h window).

\setlength{\tabcolsep}{2pt}
\begin{table}[]
\resizebox{\textwidth}{!}{%
\begin{tabular}{@{}lllllllllll@{}}
\toprule
\multicolumn{1}{l|}{}                   & \multicolumn{2}{c|}{\textbf{MIMIC}}                                   & \multicolumn{3}{c|}{\textbf{Education} (\textbf{EDU})}                                                                  & \multicolumn{5}{c}{\textbf{Weather}}                                                                                                                                      \\ \midrule
\multicolumn{1}{l|}{}                   & \multicolumn{1}{c}{Cardiomegaly} & \multicolumn{1}{c|}{ECM}  & \multicolumn{1}{c}{Success}      & \multicolumn{1}{c}{Dropout}      & \multicolumn{1}{c|}{Next Week}     & \multicolumn{1}{c}{Temp. (24h)}  & \multicolumn{1}{c}{Temp (72h)}   & \multicolumn{1}{c}{Temp (720h)}  & \multicolumn{1}{c}{Humidity}     & \multicolumn{1}{c}{Visibility}   \\ \midrule
\multicolumn{1}{r|}{\textit{Metric}}    & \multicolumn{1}{c}{\textit{BAC}} & \multicolumn{1}{c|}{\textit{BAC}}  & \multicolumn{1}{c}{\textit{BAC}} & \multicolumn{1}{c}{\textit{BAC}} & \multicolumn{1}{c|}{\textit{MSE}}  & \multicolumn{1}{c}{\textit{MSE}} & \multicolumn{1}{c}{\textit{MSE}} & \multicolumn{1}{c}{\textit{MSE}} & \multicolumn{1}{c}{\textit{MSE}} & \multicolumn{1}{c}{\textit{MSE}} \\
\multicolumn{1}{r|}{\textbf{MultiModN}} & 0.75 $\pm$0.04                    & \multicolumn{1}{l|}{0.71 $\pm$0.03} & 0.93 $\pm$0.04                    & 0.83 $\pm$0.02                    & \multicolumn{1}{l|}{0.01 $\pm$0.01} & 0.03 $\pm$0.01                    & 0.03 $\pm$0.01                    & 0.03 $\pm$0.01                    & 0.02 $\pm$0.01                    & 0.10 $\pm$0.01                    \\
\multicolumn{1}{r|}{\textbf{P-Fusion}}  & 0.75 $\pm$0.02                    & \multicolumn{1}{l|}{0.69 $\pm$0.03} & 0.92 $\pm$0.03                    & 0.87 $\pm$0.05                    & \multicolumn{1}{l|}{0.01 $\pm$0.01} & 0.02 $\pm$0.01                    & 0.03 $\pm$0.01                    & 0.02 $\pm$0.01                    & 0.03 $\pm$0.01                    & 0.08 $\pm$0.02                    \\ \midrule
                                        &                                  &                                    &                                  &                                  &                                    &                                  &                                  &                                  &                                  &                                 
\end{tabular}%
}
\caption{\textbf{\md does not compromise performance in single-tasks.} Performance for binary and continuous prediction tasks in MIMIC, EDU, and Weather, comparing \pf and \md. $95\%$ CIs are shown. \textit{ECM: Enlarged Cardiomediastinum, Temp: Temperature}.}
\label{tab:unitask-results}
\end{table}

\textbf{Results. }
Both \md and \pf achieve state-of-the-art results on single tasks using multimodal inputs across all 10 targets. In Figure \ref{fig:Weather-roc}, we binarize the continuous weather task (humidity prediction) as an average across all time steps. The task is particularly challenging for the \pf baseline, which has random performance (AUROC: 0.5). Compared with \pf, \md shows a 20\% improvement, which is significant at the $p<0.05$ level. As the temporality of this task is particularly important, it could be hypothesized that the sequential nature of \md better represents time-series inputs. Nevertheless, all weather targets are designed as regression tasks and show state-of-the-art MSE scores in Table \ref{tab:unitask-results} where \md achieves baseline performance.

We provide an additional parallel fusion transformer baseline with experimental results showcased in Appendix Sec. \ref{appendix:transformer}. The results indicate that \md matches or outperforms the multimodal transformer in the vast majority of single- and multi-task settings, and comes with several interpretability, missingness, and modularity advantages. Specifically, using the primary metric for each task (BAC for classification and MSE for regression tasks), \md beats the transformer baseline significantly in 7 tasks, overlaps 95\% CIs in 11 tasks, and loses slightly (0.01) in 2 regression tasks.

\begin{graybox} \md matches \pf performance across all 10 tasks in all metrics reported across all three multimodal datasets. Thus, modularity does not compromise predictive performance. \end{graybox} 

\subsection{Exp. 2: Multi-task \md maintains baseline performance in individual tasks}
\label{exp:multi-task}
\begin{figure}[]
\centering
\includegraphics[width=1\textwidth,]{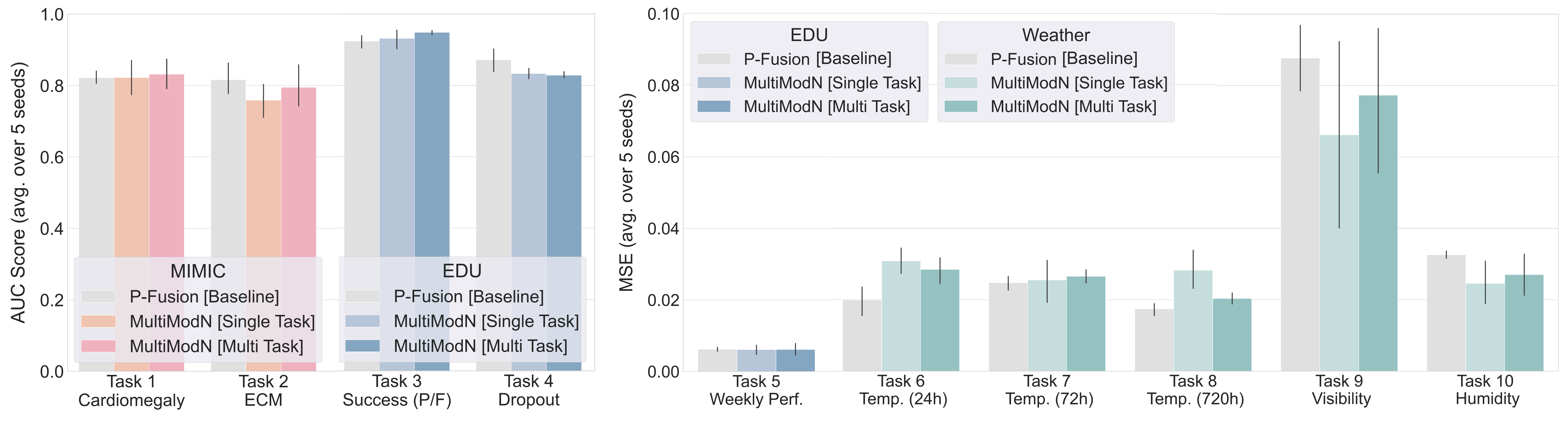}
\caption{\label{fig:mutitask} \textbf{Multi-task \md maintains baseline performance in individual tasks. } Single- and multi-task \md on the prediction of individual tasks, compared with the monolithic {\color{gray}{\pf}} (can only be single-task). AUC for binary (\textbf{left}) and MSE for continuous \textbf{(right)}. Error bars: 95\% CIs.}
\end{figure}

\textbf{Setup. } The modular design of \md allows it to train multiple task-specific decoders and deploy them in any order or combination. While multi-task models have the potential to enrich feature extraction (and improve the model), it is critical to note that all feature extraction from the raw input is performed before \md is trained. \md is trained on embeddings extracted from pre-trained models (independently of its own encoders). This is done purposely to best isolate the effect of parallel-monolithic vs. sequential-modular fusion. 
We train three multi-task \md models (one for each dataset, predicting the set of tasks in that dataset, i.e. $tasks_{1-2}$ in MIMIC, $tasks_{3-5}$ in EDU, and $tasks_{6-10}$ in Weather) and compare this to 10 single-task \md models (one for each $tasks_{1-10}$). 
Monolithic models, like \pf are not naturally extensible to multi-task predictions. Thus \pf (grey bars in Figure  \ref{fig:mutitask}) can only be displayed for single-task models. This experiment aims to compare \md performance between single- and multi-task architectures to ensure that this implementation does not come at a cost to the predictive performance of individual tasks. 

\textbf{Results. } In Figure \ref{fig:mutitask} we compare the single-task \pf (grey bars), to single- and multi-task implementations of \md (in color). The results demonstrate that \md is able to maintain its performance across all single-prediction tasks even when trained on multiple tasks. We additionally include the results of our model on various numbers and combinations of inputs, described further in Appendix Sec. \ref{appendix:inference}. The baseline would have to impute missing features in these combinations, exposing it to catastrophic failure in the event of systematic missingness (Sec. \ref{sec:missingness}).

\begin{graybox} \md has the significant advantage of being naturally extensible to the prediction of multiple tasks without negatively impacting the performance of individual tasks.\end{graybox}

\subsection{Exp. 3: \md has inherent modality-specific local and global model explainabilty}
\begin{figure*}[ht!]
    \centering
\includegraphics[width=0.8\textwidth, trim={1 1 1 1}, clip]{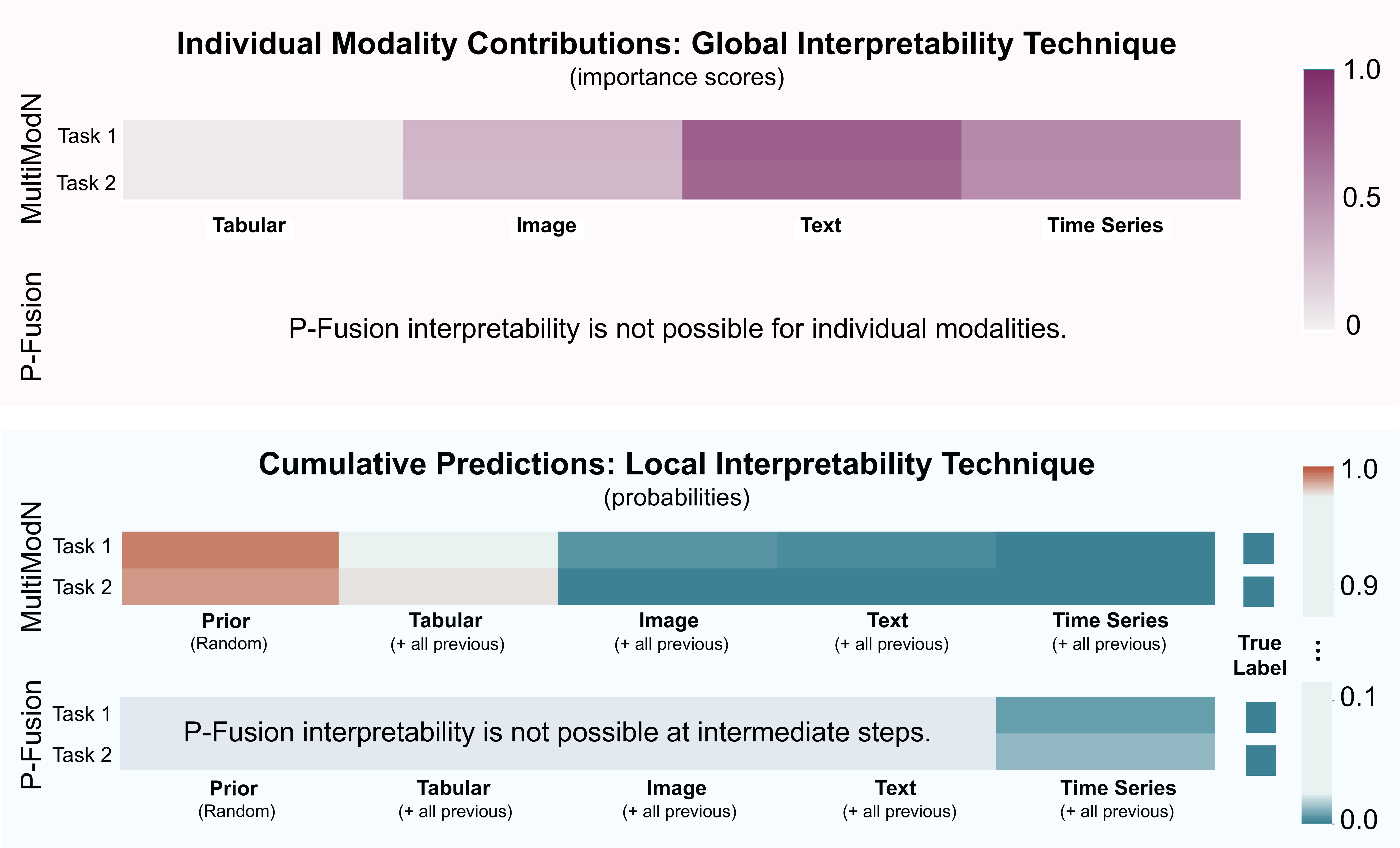}
    \caption{\label{fig:inference-main} \textbf{Inherent modality-specific model explainability in \md. } Heatmaps  show individual modality contributions (IMC) \textbf{(top)} and cumulative contributions (CP) \textbf{(bottom)}: respectively {\color{Mulberry}\textbf{importance score}} (global explainability) or {\color{RoyalBlue}\textbf{cumulative probability}} (local explainability). The multi-task \md for $task_{1-2}$ in MIMIC is compared to two single-task \pf models. IMC are only possible for \md (only 1 modality encoded, rest are skipped). CP are made sequentially from states encoding all previous modalities. \pf is unable to naturally decompose modality-specific contributions (can only make predictions once all modalities are encoded). IMC is computed across all patients in the test set. CP is computed for a single patient, (true label = 0 for both $task_{1-2}$). The CP heatmap shows probability ranging from {{\color{RoyalBlue}{\textbf{confident negative diagnosis (0)}}} to {\color{Gray}{\textbf{perfect uncertainty}}} and \color{Bittersweet}{\textbf{confident positive diagnosis (1)}}}.}
\end{figure*}

\textbf{Setup. } 
Parallel MM fusion obfuscates the contribution of individual inputs and requires add-on or post hoc methods to reveal unimodal contributions and cross-modal interactions  \cite{liang2023mm-interpmultiviz, park2018mm-interp, wortwein2022mm-nonadditive}. Soenksen et al. \cite{haim2022} used Shapley values \cite{Sim2022-shapley} to derive marginal modality contributions. While these post hoc methods provide valuable insight, they are computationally expensive and challenging or impossible to deploy at inference.  
In contrast, \md confers inherent modality-specific interpretability, where the contribution of each input can be decomposed by module. We use $task_{1-2}$ in MIMIC to compute two measures: \textbf{[1] {\color{Mulberry}\textbf{Importance score}}}, where each encoder is deployed alone, providing predictive importance of a single modality by subtracting predictions made from the prior state. This can be computed across all data points or individual data points. \textbf{[2] {\color{RoyalBlue}\textbf{Cumulative probability}}}, where the prediction from each multi-task decoder is reported in sequence (i.e. given the previously encoded modalities). We demonstrate this on a random patient from the test set, who has a true label of 0 for both tasks. Further plots are in Appendix Sec. \ref{appendix:interpretability}.

\textbf{Results. } Monolithic \pf models cannot be decomposed into any modality-specific predictions, and its (single-task) prediction is only made after inputting all modalities. In contrast, Figure \ref{fig:inference-main} shows \md provides granular insights for both importance score and cumulative prediction. We observe that the \texttt{Text} modality is the most important. The cumulative prediction shows the prior strongly predicts positivity in both classes and thus that $S_0$ has learned the label prevalence.

\begin{graybox} \raggedright The predictions naturally produced by \md provide diverse and granular interpretations.\end{graybox}

\subsection{Exp. 4: \md is robust to catastrophic failure from biased missingness}
\label{sec:missingness}

\label{exp:missingness}
\begin{figure}[htbp]
  \begin{center}
    \includegraphics[width=0.5\textwidth]{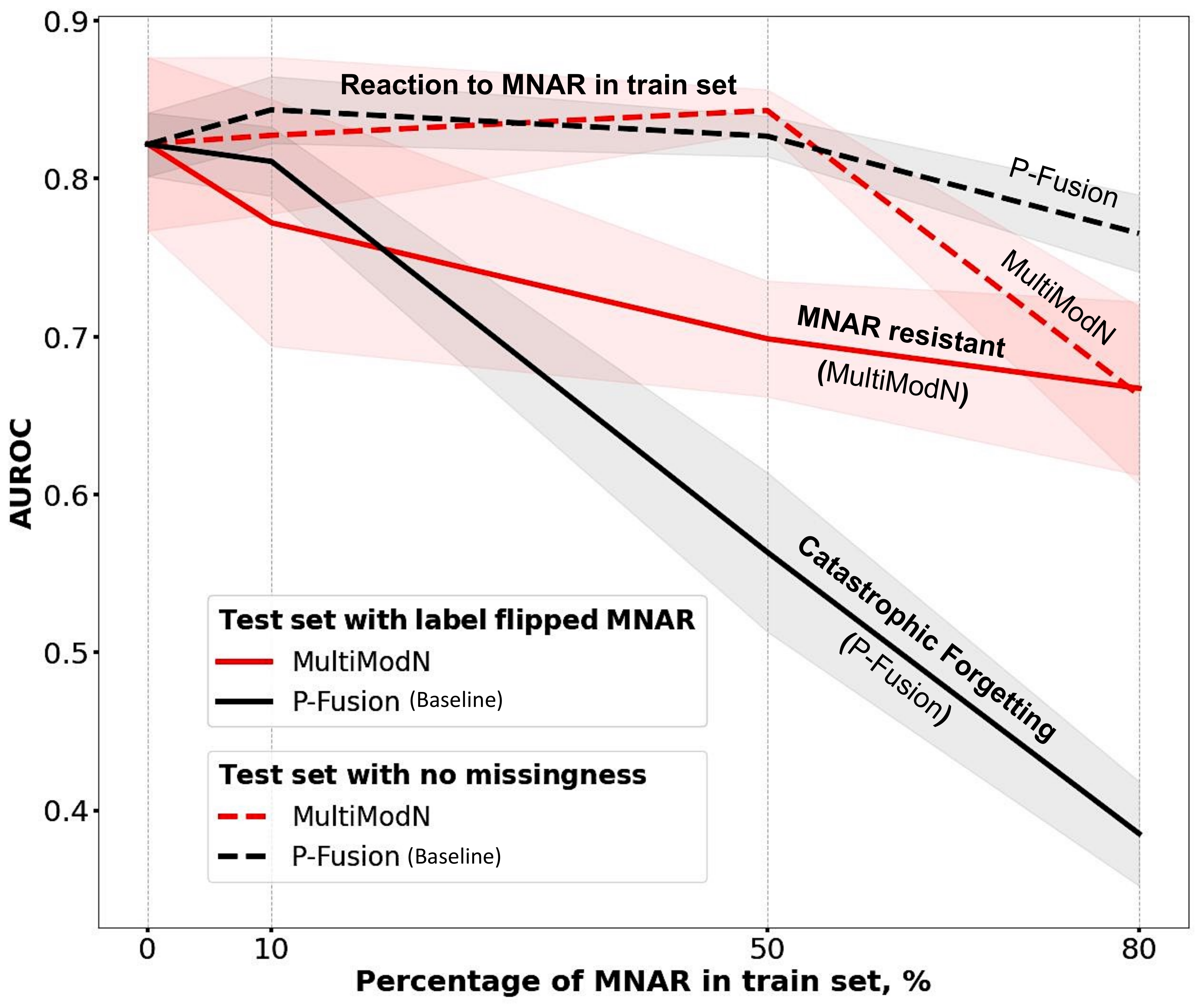}
  \end{center}
\caption{\textbf{\md is robust to catastrophic MNAR failure. } Impact of MNAR missingess on \md vs. \pf. Both models are trained on four versions of the MIMIC dataset with 0---80\% MNAR. They are then tested on either a test set with no MNAR missingness (\textbf{- - -}) or a test set where the biased missingness is label-flipped, i.e. MNAR occurs in the other binary class as compared with the train (\textbf{---}). Results for $task_1$ depicted. CI95\% shaded.}
\label{fig:missing}
\end{figure}
\textbf{Setup. } \md is designed to predict any number or combination of tasks from any number or combination of modalities. A missing modality is skipped (encoder $e_i$ is not used) and not padded/encoded. Thus, \md avoids featurizing missingness, which is particularly advantageous when missingness is MNAR. Featurizing MNAR can result in catastrophic failure when MNAR patterns differ between train and test settings. We demonstrate \md's inherent robustness to catastrophic MNAR failure by training \md and \pf on four versions of MIMIC with various amounts (0, 10, 50, or 80\%) of MNAR by artificially removing one modality in one class only. Figure \ref{fig:missing} compares \md and \pf on $task_1$ when tested in a setting that has either no missingess or where the MNAR pattern is different (i.e. label-flipped).

\textbf{Results. } 
Figure \ref{fig:missing} shows a dramatic catastrophic failure of \pf in a label-flipped MNAR test set (\textbf{black solid line}) compared with \md. \pf is worse than random at 80\% MNAR (AUROC: 0.385). In contrast, \md only loses 10\% in MNAR flip, remarkably, matching performance in a test with no missingness.  Further plots in Appendix \ref{appendix:missingness}.

\begin{graybox} \md is robust to catastrophic missingness (MNAR failure) where \pf is not.\end{graybox}

\section{Conclusion}

We present \md, a novel sequential modular multimodal (MM) architecture, and demonstrate its distinct advantages over traditional monolithic MM models which process inputs in parallel. 

By aligning the feature extraction pipelines between \md and its baseline \pf, we better isolate the comparison between modular-sequential MM fusion vs. monolithic-parallel MM fusion. We perform four experiments across 10 complex real-world MM tasks in three distinct domains. We show that neither the sequential modularization of \md nor its extension to multi-task predictions compromise the predictive performance on individual tasks compared with the monolithic baseline implementation.
Training a multi-task model can be challenging to parameterize across inter- and cross-task performance \cite{crawshaw2020mtreview,vafaeikia2020mtreview}.  We perform no specific calibration and show that \md is robust to cross-task bias. 
Thus, at no performance cost, modularization allows the inherent benefits of multi-task modeling, as well as providing interepretable insights into the predictive potential of each modality. 
The most significant benefit of \md is its natural robustness to catastrophic failure due to differences in missingness between train and test settings. This is a frequent and fundamental flaw of many domains and especially impacts low-resource settings where modalities may be missing for reasons independent of the missingness in the train set.
More generally, modularization creates a set of self-contained modules, composable in any number or combination according to available inputs and desired outputs. This composability not only provides enormous flexibility at inference but also reduces the computational cost of deployment. Taken together, these features allow \md to make resource-adapted predictions, which have a particular advantage for real-world problems in resource-limited settings.

\textbf{Limitations and future work. } 
The main limitation for studying MM modeling is the scarcity of large-scale, open-source, MM datasets that cover multiple real-world tasks, especially for time-series. 
Additionally, while \md is theoretically able to handle any number or combination of modalities and tasks, this has not been empirically tested. Having a high combinatorial generalization comes at a computational and performance cost, where the `memory' of a fixed-size state representation will likely saturate at scale.
The performance of \md is purposely limited in this work by fixing the feature extraction pipeline, to best isolate the effect of sequential fusion. Future work leveraging \md model-agnostic properties would be able to explore the potential performance benefit.
This is particularly interesting for time-series, for which the state `memory' may need to be parameterized to capture predictive trends of varying shapes and lengths.

%% file: main/07_interpretcc.tex
\newcommand{\cross}{{\color{red}\ding{55}}}  % Red cross symbol
\newcommand{\checkmarkgreen}{{\color{ForestGreen}\ding{51}}}  % Green checkmark 
\newcommand{\featuregating}{Feature Gating\xspace}
\newcommand{\grouprouting}{Group Routing\xspace}
\newcommand{\icc}{\texttt{InterpretCC}\xspace}

The work presented in this chapter was conducted in collaboration with Syrielle Montariol, Julian Blackwell, Jibril Frej, Professor Martin Jaggi, and Professor Tanja Käser under a paper titled \textit{Intrinsic User-Centric Interpretability Through Global Mixture of Experts}. This work was presented at the International Conference of Learning Representations (ICLR) 2025.

In this chapter, we present a technical contribution of a family of interpretable model architectures (\icc), as well as the first known user perception study on interpretable-by-design models. The architectures presented in this chapter aim to optimize interpretability through sparsity, without compromising predictive performance. As \icc uses gating (of features or expert networks) directly in its predictions, it also produces \textbf{real-time} explanations alongside model predictions. Explanations are guaranteed to be \textbf{accurate} as they are directly used in the prediction process. \icc allows for meaningful groupings of features into concepts, either specified by human users or automated groupings via LLMs. Therefore, it produces \textbf{human-understandable} explanations. These explanations are also \textbf{consistent}, as the same model inputs will always produce the same explanation. The \textbf{actionability} of explanations depends on the user specification of concepts or features.

\section{Introduction}
Towards transparency for neural networks, a range of explainable AI methods have emerged across two main axes: \textit{global} vs. \textit{local} explanations, and \textit{post-hoc} vs. \textit{intrinsic} explanations \cite{du2019techniques}. \textit{Global} interpretability allows users to understand how the entire model operates by examining its structure and parameters, while \textit{local} interpretability focuses on understanding the rationale behind a individual prediction \cite{molnar2020interpretable}. Most methods are \textit{post-hoc}, where the explanation is extracted from a model that has already been trained. In contrast, \textit{intrinsic} methods directly incorporate interpretability into the model's structure. 

Popular post-hoc methods include attribution methods like LIME \cite{lime} or SHAP \cite{shap},  counterfactual methods like CEM \cite{counterfactual}, or pattern-based methods like PREMISE \cite{hedderich2022label}. Post-hoc methods require users to trust the explainer’s approximation of the ground truth explanation (the underlying model's decision process), although they have been shown to be systematically biased and inconsistent \cite{krishna2022disagreement, swamy-eval}. \textit{In-hoc} explanations are a subset of post-hoc methods that require access to model weights instead of treating the model as a black box \cite{molnar2020interpretable}. For instance, Grad-CAM \cite{gradcam} assesses the contribution of a component to the model’s output, and TCAV/DTCAV \cite{tcav, dtcav} use user-defined concepts to interpret neural network embeddings.

Recent intrinsically explainable model literature has focused on example-based approaches, overwhelmingly for the image modality (\textit{e.g.} B-cos networks \cite{bohle2022b}, PIP-Net \cite{nauta2023pip}, ProtoPNet \cite{chen2019looks}) and \textcolor{black}{less commonly in time-series, tabular, or text modalities} \cite{sawada2022concept}. Other approaches like NAM \cite{agarwal2021neural} and EBM \cite{nori2019interpretml} train a model for each input feature or combination of features and output predictions using scores from these models, requiring a lot of subnetworks when the feature space is large. Research towards interpretable mixture-of-experts models has highlighted a hierarchical neural network structure with subnetworks, combining interpretable experts (i.e. decision trees) with NNs for partially interpretable points \cite{ismail2023interpretable}, selectively activating experts \cite{li2022branch}, or extracting automated concepts over the input space for routing \cite{you2023sum,alvarez2018towards}. All of these approaches, while expressive, are burdened with overly detailed explanations which limit human understandability and actionability.

\begin{figure}[t]
    \centering
    \includegraphics[width=\linewidth]{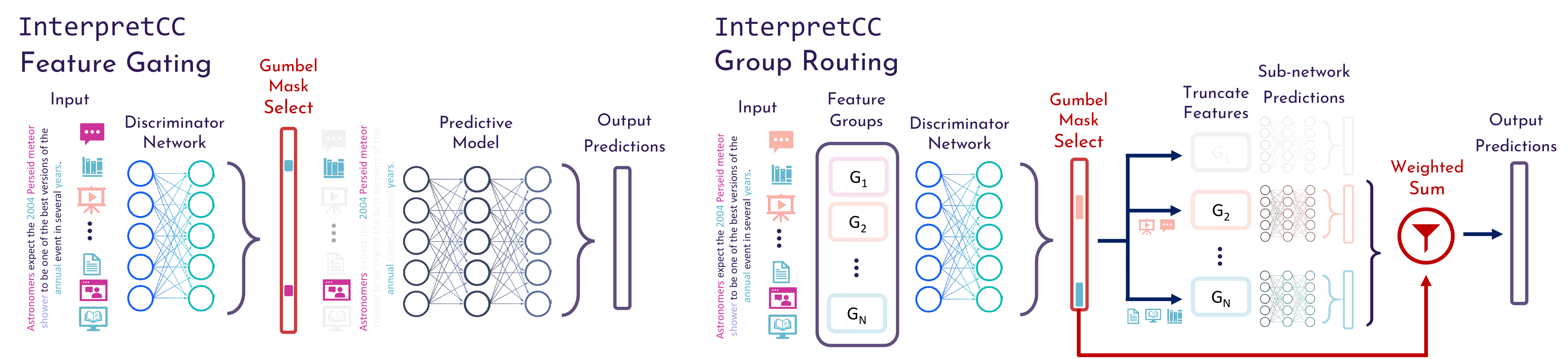}
    \caption{\textbf{\icc Architectures}: \textit{\featuregating (left, individual features)}: (i) All features are input into a discriminator network that outputs a sparse feature activation mask; (ii) Only the features selected via the mask are passed to a predictive network for the final prediction. \textit{\grouprouting (right, pre-defined feature groups)}: (i) Features are statically assigned to distinct groups, with each feature routed to only one group; (ii) Features are input to a discriminator network, generating a sparse group activation mask; (iii) Predictions from activated sub-networks (selected via mask) are aggregated by a weighted sum to produce the final output.}
    \label{fig:arch}  
\end{figure}

In this chapter, we therefore present a \textbf{user-centric intrinsically interpretable} framework that achieves faithful local interpretability and provides sparse and actionable human-centric explanations, while maintaining comparable predictive performance to its black-box counterparts. To achieve these goals, we use conditional computation to craft interpretable neural pathways using two different architectures (see Figure \ref{fig:arch}) based on routing through individual features or entire feature groups.

Our models enable statements like the following: \textcolor{black}{``The student’s regularity and video watching behavior were the only two aspects selected as important for the student's prediction of passing the course, and the model did not use any other aspects to make this prediction"}. We refer to interpretability from the users' perspective, focusing on the model's local reasoning for a decision on a specific data point, as opposed to a global understanding of the model's internals. Our models are characterized by sparse explicit routing, truncated feature spaces, and adaptivity per data point. These traits are important for human-centric trustworthiness as they provide clear and concise instance-level explanations \cite{miller2019explanation, swamy2023future}.

With our family of \icc models, we provide the following contributions:
\begin{enumerate}[itemsep=0em]
    \item \textbf{\icc \featuregating: A simple, interpretable NN architecture} using a gating mechanism to sparsely activate specific features.
    \item \textbf{\icc \grouprouting: An interpretable mixture-of-experts architecture} that uses human-specified group routing to separate the feature space and sparsely activate specific experts.
    \item \textcolor{black}{\textbf{An extension of intrinsic interpretability to multiple human-centric modalities and domains}, focusing on time-series (education), tabular (health, synthetic data), and text (sentiment, news).\footnote{We do not focus on vision datasets since extracting concepts from vision has been well-studied by existing, modality-specific interpretability approaches, e.g. \cite{bohle2022b, you2023sum, donnelly2022deformable, thomas2023unlocking}.}}
    \item \textbf{A novel XAI user study} comparing teachers' preferences of interpretable-by-design model explanations towards designing educational interventions for struggling students.
\end{enumerate}

 Across experiments on eight diverse datasets, we show that \icc models perform comparatively to non-interpretable baselines (matching or exceeding 95\% CIs in 15 of 16 comparisons) and outperform intrinsically interpretable baselines (\texttt{ICC} Feature Gating is 9.05\% better on average than SENN Features and 3.27\% better on average than NAM, while \texttt{ICC} Group Routing is on average 5.63\% better than SENN Concepts). Moreover, participants of the user study preferred \texttt{ICC} explanations over baselines in terms of actionability, usefulness, conciseness and trustworthiness.
We provide our code open source: \url{https://github.com/epfl-ml4ed/interpretcc}.

%%%%%%%% Background Section%%%%%%%%%%%%%%%%%
\section{Background}
\label{sec:icc-background}

\begin{table}
\small
\centering
\resizebox{0.8\textwidth}{!}{%
\begin{tabular}{@{}clllll@{}}
\toprule
\textbf{} & \multicolumn{5}{c}{\textbf{Explanation}} \\ \cmidrule(l){2-6} 
\textbf{Method} & \multicolumn{1}{c}{\textit{Granularity}} & \multicolumn{1}{c}{\textit{Basis}} & \multicolumn{1}{c}{\textit{Faithfulness}} & \multicolumn{1}{c}{\textit{Sparsity}} & \multicolumn{1}{c}{\textcolor{black}{\textit{Stage}}} \\ 
\midrule
\textbf{\begin{tabular}[c]{@{}c@{}}LIME, SHAP\\ (Post-Hoc)\end{tabular}} & Feature & Use all input features & {\color[HTML]{CB0000} Approximation} & {\color[HTML]{009901} Sparse} & {\color[HTML]{CB0000} \begin{tabular}[c]{@{}l@{}}Explanation not \\ used in model\end{tabular}} \\ 
\midrule
\textbf{\begin{tabular}[c]{@{}c@{}}TCAV\\ (In-Hoc)\end{tabular}} & Concept & \begin{tabular}[c]{@{}l@{}}User defines concepts \\ through examples\end{tabular} & {\color[HTML]{F8A102} \begin{tabular}[c]{@{}l@{}}Aligned\\ with concepts\end{tabular}} & {\color[HTML]{009901} Sparse} & {\color[HTML]{F8A102} \begin{tabular}[c]{@{}l@{}}Explain from \\ model internals\end{tabular}} \\ 
\midrule
 & Feature & Use all input features & {\color[HTML]{009901} Guaranteed} & {\color[HTML]{CB0000} Not sparse} & {\color[HTML]{009901} \begin{tabular}[c]{@{}l@{}}Explain \\ then predict\end{tabular}} \\ \cmidrule(l){2-6} 
\multirow{-2}{*}{\textbf{SENN}} & Concept & \begin{tabular}[c]{@{}l@{}}Automated concept \\ selection\end{tabular} & {\color[HTML]{F8A102} \begin{tabular}[c]{@{}l@{}}Aligned \\ with concepts\end{tabular}} & {\color[HTML]{CB0000} Not sparse} & {\color[HTML]{009901} \begin{tabular}[c]{@{}l@{}}Explain \\ then predict\end{tabular}} \\ 
\midrule
\textbf{NAM} & Feature & Use all input features & {\color[HTML]{009901} Guaranteed} & {\color[HTML]{CB0000} Not sparse} & {\color[HTML]{009901} \begin{tabular}[c]{@{}l@{}}Explain \\ then predict\end{tabular}} \\ 
\midrule
 & Feature & Use all input features & {\color[HTML]{009901} Guaranteed} & {\color[HTML]{009901} Sparse} & {\color[HTML]{009901} \begin{tabular}[c]{@{}l@{}}Explain \\ then predict\end{tabular}} \\ \cmidrule(l){2-6} 
\multirow{-3}{*}{\textbf{\begin{tabular}[c]{@{}c@{}}\icc\\(Feature Gating,\\ Group Routing)\end{tabular}}} & Concept & \begin{tabular}[c]{@{}l@{}}User (or LLM) defines \\ groups of features\end{tabular} & {\color[HTML]{009901} Guaranteed} & {\color[HTML]{009901} Sparse} & {\color[HTML]{009901} \begin{tabular}[c]{@{}l@{}}Explain \\ then predict\end{tabular}} \\ 
\bottomrule
\end{tabular}}
\caption{\textbf{\textcolor{black}{Design} Comparison}: \icc models are at the unique intersection of flexible explanation \textit{granularity} (either features or concepts), guaranteed explanation \textit{faithfulness} to the model's decision process, optimal \textit{sparsity} in the explanation, and \textcolor{black}{explanations used in the prediction (\textit{stage})}. \textit{Basis} describes the foundation of the explanation (e.g. user-defined concepts or raw features). \textcolor{black}{A taxonomy can be found in Appendix \ref{sec:related-work}.} \label{tab:approaches}}
\end{table}

\textbf{Architecture Foundations.} \textit{Conditional Computation (CC)} has become widely used to improve the computationally expensive training and inference of large neural networks by activating only parts of the network~\cite{begnio-stochastic, bengio-cc, davis-arel}. 
Inspired by the foundations laid out by CC, mixture-of-expert models have rapidly gained popularity for improving the efficiency of neural networks through activating different expert subnetworks at different layers.
BASE layers~\cite{lewis2021base} direct each token to a designated expert and Switch Transformers~\cite{fedus2022switch} use CC to select one out of 4 feedforward networks across each transformer layer, optimizing computational resources. 
Mixtral~\cite{jiang2024mixtral} is a recent LLM using a mixture of experts to select 2 out of 8 expert networks at each layer, reducing the numbers of active parameters by a factor of 4 compared to training, while allowing each token to have access to all the parameters.  With \icc models, we extend a similar routing idea with instance-dependent gating decisions towards an interpretability objective as opposed to only an efficiency or performance objective. 

\textbf{Interpretability Foundations.} Explainability can be integrated into different stages of the modeling pipeline: post-hoc (after model training), in-hoc (requiring model weights), and intrinsic (interpretable by design) \cite{swamy2023future, molnar2020interpretable}. Our positioning of \textcolor{black}{the design of} \icc in comparison to popular approaches is described in Table \ref{tab:approaches}. Specifically, we categorize the approaches using their basis and the stage they are applied to in the pipeline as well as \textcolor{black}{four key aspects for human-centric explanations: \textit{faithfulness}, the explanation reflects the model behavior with certainty \cite{lyu2024towards,dasgupta2022framework}; \textit{sparsity}, the model uses a minimal amount of features, optimizing for user understandability and actionability \cite{sun2024sparse, ayoobi2023sparx}; \textit{predictive stage}, the explanation covers the entirety of what the model uses for prediction \cite{speith2022review, schwalbe2024comprehensive}; and  \textit{granularity}, the explanation is conveyed in terms of features or concepts \cite{miller2019explanation, jain2020learning}.}

Post-hoc methods such as LIME \cite{lime} or SHAP \cite{shap} approximate what the model finds important, and therefore cannot be considered faithful (cross-feature actions are often not described in the explanation). LIME and SHAP use the full input feature space and can be configured for sparseness, although they have been shown to choose a broad amount of features \cite{swamy-eval} using default settings. In-hoc interpretability approaches often require users to specify examples to define human-understandable concepts (e.g. TCAV \cite{tcav}, DTCAV \cite{dtcav}), or use hybrid methods with both human-defined and automated concepts~\cite{sawada2022concept}. In-hoc approaches have limited faithfulness due to the (lack of) completeness of the concepts \textcolor{black}{and do not use explanations directly in the prediction}. Inspired by these approaches, \icc allows users to specify interpretable concepts that are directly useful to them. However, we do not use examples, but instead allow users to specify a grouping over the feature space, achieving both sparsity and \textcolor{black}{explanations used for prediction}.

Initial approaches have explored expert models for intrinsic interpretability. The Interpretable Mixture of Experts (IME) framework \cite{ismail2023interpretable} uses linear models alongside deep models to provide partially faithful explanations. LIMoE \cite{mustafa2022multimodal} focuses on visual experts to identify concepts like textures and faces, enhancing interpretability in vision tasks. Similarly, the Sum-of-Parts (SOP) model \cite{you2023sum} uses sparse feature groups to emphasize the model's reliance on subsets of features for predictions. \textcolor{black}{Approaches in extractive rationale methods and explain-then-predict methods \cite{jain2020learning, bastings2019interpretable, yu2019rethinking} produce intuitive text explanation guarantees with explanation selection before prediction, but are often not generalizable beyond that modality (see Appendix \ref{sec:rationale}). Few intrinsic approaches use expert knowledge to define concepts directly, instead using prototype examples \cite{koh2020concept} or rules \cite{konstantinov2024incorporating}.} The most relevant models to our work are Self-Explaining Neural Networks (SENN) \cite{alvarez2018towards} and Neural Additive Models (NAM) \cite{agarwal2021neural}, both neural models similar to \texttt{ICC} as opposed to Explainable Boosted Machines (EBM) which uses trees \cite{nori2019interpretml}. SENN extracts concepts with prototypical examples and their relevances, but it lacks faithfulness (it cannot explain what is not in a concept), sparsity (it explains all concepts), and concepts do not cover the entire feature space. NAM assigns a model to each feature and combines the outputs linearly, achieving faithfulness but not sparsity, as all features contribute. \texttt{ICC} differs by filtering the feature space instead of using all features (sparsity), using user-defined concepts instead of automated concepts (basis), and assigning each feature to a single group, making feature use explicit \textcolor{black}{(predictive stage)}.

%%%%%%%% Method section %%%%%%%%%%%%%%%%%%%%%%%
\section{\icc Methodology}

Given an input $x$, the objective of our approach is to select a sparse subset of $x$ that will be used to predict the output to solve the classification task. We propose two architectures:

\textbf{Feature Gating:} The approach only processes a subset of the features by applying a sparse mask $\mathcal{M}$ on the input $x$ before processing it by a model $f$. The output is given by: $ f(\mathcal{M}(x))$. 

\textbf{Group Routing:} A sparse mixture of models~\cite{fedus2022review} applied on human-interpretable groups of features where each expert is assigned to a group of features: $\sum_{i=1}^{K} G(x)_i \cdot f_i(  \mathcal{M}_{\mathcal{G}}(x)_i)$ where $\mathcal{M}_{\mathcal{G}}(.)_i$ is a sparse mask selecting only the features of group $i$, $f_i$ is the expert model associated with the $i$-th group, and $G(x)_i$ is the output of the gating network for group $i$. If $G(x)_i = 0$, the entire group of features is ignored. 
\subsection{Feature Gating}

\label{sec:gating}
 \textcolor{black}{\icc \featuregating, shown in Fig.~\ref{fig:arch},} is the first step towards using CC paths for interpretability. The features are first passed through a discriminator network $D$ to select which ones to use for computing the output. 
 The Gumbel Softmax trick~\cite{gumbel} is applied on each dimension of $D(x)$ to select features in a differentiable way (see Appendix \ref{appendix:gumbel} for more details). A feature $j$ is activated (the associated value in the mask is non-zero) if the Gumbel Softmax output exceeds a threshold $\tau$, a hyperparameter. This allows the model to adaptively select the number of features based on each instance, using fewer features for simpler cases and more for complex ones.

The output is computed using a model $f$ on the masked input $\mathcal{M}(x)$. Since the explainability is at the feature level, using a black box model for $f$ does not detract from the interpretability. \textcolor{black}{Notably, \texttt{ICC} FG does not require human specification.}

\subsection{Group Routing} 
\label{sec:routing} 
We build upon the instance-dependent gating architecture with feature groups. 
As displayed in Figure~\ref{fig:arch}, instead of selecting features individually, the mask is applied to human interpretable groups of features. 
Doing so encourages cross-feature interactions while maintaining a meaningful grouping for human users. 
To select the features belonging to group $i$, we use a binary mask $\mathcal{M}_{\mathcal{G}}(x)_i$ that is computed using human-specified rules. 
In section~\ref{sec:predsetting}, we detail our approach to compute $\mathcal{M}_{\mathcal{G}}(x)_i$ for each dataset used in our experiments.

 \textcolor{black}{\icc \grouprouting} is a sparse mixture of experts utilizing a gating network to assign a weight $G(x)_i$ to each group. 
This process mirrors that of \featuregating, starting with a discriminator network $D_{\mathcal{G}}$ \textcolor{black}{with an input of all features and output of~$K$ dimensions ($K$ is the number of groups).} It then applies the Gumbel Softmax and a threshold $\tau_{\mathcal{G}}$ to each group. The model output is a weighted sum of the output of each expert $f_i$ that only uses the features from the $i$-th group as input. Using our sparsity criteria, we ensure that few groups are used to compute the output, making the \grouprouting intrinsically interpretable at the group level, regardless of the types of models used as experts. \grouprouting enables efficient inference without reducing the number of parameters available during training. During the training phase, we employ soft masking, allowing all weights $G(x)_i$ to remain non-zero, thus granting the model access to every expert. This approach allows the model to leverage the full set of parameters during training, enhancing the training efficiency. At inference time we switch to using a hard mask, making the weights sparse and allowing for interpretability.

\section{Experimental Settings}
\label{sec:predsetting}

% Overall performance table
\begin{table}[t]
\centering
\resizebox{\textwidth}{!}{
\begin{tabular}{@{}llllllllllll@{}}
\toprule
 & \multicolumn{1}{c}{} & \multicolumn{1}{c}{} & \multicolumn{4}{c}{\textbf{Feature-Based Interpretability}} & \multicolumn{5}{c}{\textbf{Concept-Based Interpretability}} \\ \cmidrule(l){4-7} \cmidrule(l){8-12} 
\multirow{-2}{*}{} & \multicolumn{1}{l}{Dataset} & \multicolumn{1}{c}{\begin{tabular}[c]{@{}c@{}}Non-interpretable \\ Base Module\end{tabular}} & \multicolumn{1}{c}{NAM} & \multicolumn{1}{c}{\begin{tabular}[c]{@{}c@{}}SENN \\ Features\end{tabular}} & \multicolumn{1}{c}{\textcolor{black}{FRESH}} & \multicolumn{1}{c}{\begin{tabular}[c]{@{}c@{}}\icc \\ \featuregating\end{tabular}} & \multicolumn{1}{c}{\begin{tabular}[c]{@{}c@{}}SENN \\ Concepts\end{tabular}} & \multicolumn{1}{c}{\begin{tabular}[c]{@{}c@{}}\icc\\Top K Routing\end{tabular}} & \multicolumn{3}{c}{\begin{tabular}[c]{@{}c@{}}\icc \\ Group Routing\end{tabular}} \\ \midrule
 & DSP & 82.81 \footnotesize{$\pm$ 2.61} & 85.20 \footnotesize{$\pm$ 0.64} & {\color[HTML]{9A0000} 71.70 \footnotesize{$\pm$ 0.95}} &  & {\color[HTML]{009901} 90.75 \footnotesize{$\pm$ 0.01}} & 81.50 \footnotesize{$\pm$ 2.26} & 83.08 \footnotesize{$\pm$ 1.10} & \multicolumn{3}{l}{84.90 \footnotesize{$\pm$ 7.59}} \\
 & Geo & 72.96 \footnotesize{$\pm$ 1.59} & {\color[HTML]{9A0000} 65.12 \footnotesize{$\pm$ 4.07}} & {\color[HTML]{9A0000} 57.90 \footnotesize{$\pm$ 2.69}} & Not & 71.92 \footnotesize{$\pm$ 0.01} & 70.90 \footnotesize{$\pm$ 2.45} & {\color[HTML]{0D0D0D} 80.44 \footnotesize{$\pm$ 3.19}} & \multicolumn{3}{l}{{\color[HTML]{009901} 81.58 \footnotesize{$\pm$ 0.57}}} \\
 & HWTS & 73.93 \footnotesize{$\pm$ 3.76} & 73.11 \footnotesize{$\pm$ 2.13} & 68.63 \footnotesize{$\pm$ 3.78} & Supported & {\color[HTML]{009901} 82.89 \footnotesize{$\pm$ 0.04}} & 75.10 \footnotesize{$\pm$ 11.67} & 72.59 \footnotesize{$\pm$ 2.84} & \multicolumn{3}{l}{78.34 \footnotesize{$\pm$ 0.95}} \\
\multirow{-4}{*}{\textbf{Education}} & VA & 74.90 \footnotesize{$\pm$ 5.28} & 71.39 \footnotesize{$\pm$ 3.38} & 74.37 \footnotesize{$\pm$ 1.11} &  & {\color[HTML]{0D0D0D} 77.80 \footnotesize{$\pm$ 0.01}} & 69.99 \footnotesize{$\pm$ 8.83} & 71.43 \footnotesize{$\pm$ 1.11} & \multicolumn{3}{l}{72.08 \footnotesize{$\pm$ 3.71}} \\ \midrule
\textbf{Health} & \begin{tabular}[c]{@{}l@{}}B. Cancer\end{tabular} & 89.70 \footnotesize{$\pm$ 1.05} & 88.77 \footnotesize{$\pm$ 7.31} & 80.52 \footnotesize{$\pm$ 6.21} & Not Supp.  & {\color[HTML]{9A0000} 78.19 \footnotesize{$\pm$ 3.54}} & {\color[HTML]{9A0000} 85.26 \footnotesize{$\pm$ 1.03}} & {\color[HTML]{333333} 84.66 \footnotesize{$\pm$ 3.02}} & \multicolumn{3}{l}{{\color[HTML]{009901} 94.85 \footnotesize{$\pm$ 1.25}}} \\ \midrule
 & AG News & 89.93 \footnotesize{$\pm$ 3.32} &  &  & 88.73 \footnotesize{$\pm$ 0.69} & 85.72 \footnotesize{$\pm$ 5.31} &  & 87.25 \footnotesize{$\pm$ 2.48} & \multicolumn{3}{l}{90.35 \footnotesize{$\pm$ 1.07}} \\
\multirow{-2}{*}{\textbf{Text}} & SST & 91.12 \footnotesize{$\pm$ 2.03} & \multirow{-2}{*}{\begin{tabular}[c]{@{}l@{}}Not \\ Supported\end{tabular}} & \multirow{-2}{*}{\begin{tabular}[c]{@{}l@{}}Not \\ Supported\end{tabular}} & {\color[HTML]{9A0000} 82.05 \footnotesize{$\pm$ 0.56}} & 88.21 \footnotesize{$\pm$ 3.41} & \multirow{-2}{*}{\begin{tabular}[c]{@{}l@{}}Not \\ Supported\end{tabular}} & 92.98 \footnotesize{$\pm$ 0.88} & \multicolumn{3}{l}{91.75 \footnotesize{$\pm$ 1.86}} \\ \midrule
\textbf{Synthetic} & OpenXAI & 86.67 \footnotesize{$\pm$ 0.31} & 87.85 \footnotesize{$\pm$ 1.31} & 83.67 \footnotesize{$\pm$ 1.86} & Not Supp. & {\color[HTML]{009901} 89.51 \footnotesize{$\pm$ 0.51}} & 84.67 \footnotesize{$\pm$ 4.04} & 90.83 \footnotesize{$\pm$ 1.93} & \multicolumn{3}{l}{89.47 \footnotesize{$\pm$ 2.89}} \\ \bottomrule
\end{tabular}}
\caption{\label{tab:all_results} 
\textbf{\icc Performance} (avg $\pm$ std) on EDU (balanced accuracy), Text, Health, and Synthetic (accuracy) datasets compared to a non-interpretable baseline and \textcolor{black}{four} intrinsically interpretable baselines. All 95\% CI overlap with the non-interpretable base module for values in black. Colored values indicate significantly higher (green) or lower (red) performance than the base module. The reported \grouprouting results are the best performing \texttt{ICC} variations from Table \ref{tab:group-routing}.}
\end{table}
We apply \icc to five domains: education, news classification, sentiment classification, healthcare, and synthetic data covering \textit{Time Series}, \textit{Text}, and \textit{Tabular} inputs; all for classification tasks. 
For \textit{Tabular} features, the input is a vector $x \in \mathds{R}^n$. The mask in the Feature Gating is a sparse vector indicating which tabular feature to use and how important they are (if the weight is non-0) and the groups form a partition over the features. For \textit{Text} features, the input is a sequence of $N$ tokens: $x = [t_1, t_2, \cdots, t_N]$. The mask is a sparse vector that indicates which token to use and each group consists of a subset of the tokens. Finally, we consider \textit{Time Series} of $n$ features across $T$ timesteps: $x \in \mathds{R}^{n\times T}$. We apply the same mask across all time steps for \icc FG and GR.

% describe the task
\textbf{EDU} (\textit{time series, education domain}). We predict student success in the early weeks of four massive open online courses (MOOCs), using students' clickstream data (see Table \ref{tab:EDU_course_stats} in Appendix~\ref{sec:appendix-datasets} for details about the courses). The raw clickstream input is transformed into weekly time-series features that have proven useful for student success prediction in previous literature (e.g. total video clicks, forum interactions). We select 45 input features used in multiple studies~\cite{lalle-conati, boroujeni-et-al, chen-cui, marras-vignoud-kaser-2021}. For early prediction, we only use the first $40\%$ of time steps as input.

\textit{Grouping:} To derive human-interpretable concepts from these features, we turn to learning science literature. In \textbf{routing by paper}, we create 10 distinct feature subsets based on handcrafted initial input features from 10 papers, directing each to a specific expert subnetwork. 
For \textbf{routing by pattern}, we organize features according to five learning dimensions identified by \cite{asadi2023ripple, mejia-patterns}: effort, consistency, regularity, proactivity, control, and assessment-based (see Table~\ref{tab:mejia_patterns} for a detailed feature classification).
Thirdly, \textbf{routing by Large Language Model (LLM)}, uses GPT-4's capabilities to aid humans in feature grouping \cite{achiam2023gpt}. GPT-4 is prompted as an `expert learning scientist' to group the features into self-regulated behavior categories that are easy to understand, which are then used to separate the features for \icc. More details are included in Appendix~\ref{gpt-4}.

\textbf{AG News and SST} (\textit{text, news and review domains}). For news categorization (\textbf{AG News}), we classify news into four categories (`World', `Sports', `Business', `Sci/Tech') given a title and description of a real-world article \cite{zhang2015character}. We use 36,000 training samples and 3,000 test samples evenly distributed across categories. For sentiment prediction (Stanford Sentiment Treebank, \textbf{SST}), we use 11,855 sentences from movie reviews labeled by three annotators \cite{socher2013recursive} and predict a binary sentiment from a sentence fragment. 

\textit{Grouping:} The \texttt{ICC} routing model assigns words to subnetworks using the Dewey Decimal Code (DDC) hierarchy of topics for book classification to create 10 subnetworks (see Table \ref{tab:dewey_decimal_classification}, Appendix \ref{app:fg_text} for more details) \cite{satija2013theory}. Each word is encoded using SentenceBERT \cite{reimers2019sentence} and assigned to a subcategory (i.e. the word `school' is assigned to the subcategory `education' under category 300 for `social sciences') and routed to the appropriate parent network.

\textbf{Breast Cancer} (\textit{tabular, healthcare domain}). The Wisconsin Breast Cancer dataset identifies cancerous tissue from fine needle aspirate (FNA) images, with 30 features (10 per cell nucleus) and diagnoses (Malignant: 1, Benign: 0) for 569 patients \cite{breastcancer}.

\textit{Grouping:} For the grouping logic, we group each cell nucleus in a separate subnetwork, enabling features representing the same part of the tissue sample to be considered together.

\textbf{Synthetic Dataset} (\textit{tabular}) We use OpenXAI's synthetic dataset \cite{agarwal2022openxai}, which includes ground truth labels and explanations, indicating the subset of features influencing each label. This dataset comprises of 5000 samples, 20 continuous features, and two classes. It was created using the \textit{SynthGauss} mechanism from five cluster neighborhoods (1000 points for each cluster), ensuring three desirable properties for assessing explanations: (1) feature independence, (2) unambiguous, well-separated local neighborhoods, and (3) an explanation for each instance.

\textit{Grouping:} We group the feature space by assigning each feature to a cluster neighborhood based solely on the distribution of the training data. The average absolute value of each feature for each cluster is calculated, and the highest feature-cluster value determines the assignment.

\section{Experimental Results}
\label{sec:results}

Through the following three experiments, we demonstrate that our \icc models do not compromise \textit{performance} compared to black-box models and provide explanations that are \textit{faithful} as well as \textit{human-centered}. \textcolor{black}{\icc is designed for data that has meaning for humans (i.e. interpretable features or meaningful raw data like text or lab measurements); however, we demonstrate it is also performant on a synthetic dataset with no interpretable features.}

\textbf{Experimental Setup.} We run hyperparameter tuning and three different random seeds for each reported model (reproducibility details in Appendix \ref{sec:reproduce}). Since \textbf{EDU} MOOC courses have a low passing rate (below 30\%), and thus the dataset has a heavy class imbalance, we use balanced accuracy for evaluation. The other datasets are more balanced (\textbf{AG News}, \textbf{SST}, \textbf{Breast Cancer}, \textbf{Synthetic}), hence we use accuracy as our evaluation metric. We perform an 80-10-10 train-validation-test data split stratified on the output label, to conserve the class imbalance in each subset. In addition to \icc \featuregating and \grouprouting, we also employ a \icc Top-K expert network solution with k=2 for group routing. This approach is similar to existing mixture-of-expert approaches \cite{jiang2024mixtral, li2022branch}, except that their models make a layer-wise expert choice, which significantly reduces interpretability, while we make a global expert choice.

\textbf{Base Prediction Module.} We choose simple yet performant predictive modules reported in previous literature to isolate the difference in performance due to the interpretable architectures. For the EDU data, previous works uses BiLSTMs on student behavioral data for best predictive performance \cite{swamy-eval, marras-vignoud-kaser-2021}. Thus, for comparative benchmarking, the most performant BiLSTM setting is used as a baseline model \cite{swamy2022meta}. For the AG News and SST datasets, we use fine-tuned \texttt{DistilBERT}\footnote{\url{https://huggingface.co/distilbert/distilbert-base-uncased}} variations as baselines, also used in related works \cite{yang2019xlnet, hf_canonical_model_maintainers_2022}. For the Breast Cancer dataset, we use a fully connected network as reported in \cite{agarap2018breast}. 

\textbf{Interpretable Baselines. }
In addition to non-interpretable baselines, we compare \texttt{ICC} to \textcolor{black}{three intrinsically interpretable methods: SENN \cite{alvarez2018towards}, NAM \cite{agarwal2021neural}, and FRESH \cite{jain2020learning}.}
SENN generates concept-based explanations and is designed to learn \textit{"interpretable basis concepts"} in parallel with the model optimization. Explanations consist of the concepts most similar to the input. 
NAM, a General Additive Model \cite{hastie2017generalized}, uses individual neural networks for each input feature to calculate feature weights. The model's output is the sum of these weights, and explanations are given by displaying the feature weights. \textcolor{black}{The SENN Features architecture has three models working together, NAM has one subnetwork for each feature, and \texttt{ICC FG} has one model with two parts trained end-to-end. At inference, SENN and NAM also assign a score to each feature, which are then aggregated; \texttt{ICC} leverages cross-feature interactions with one score per model. FRESH is an extractive rationale architecture with three models: \textit{supp} generates importance scores, \textit{ext} selects a contiguous text mask, and \textit{pred} applies this mask on the inputs to make predictions. Designed for text data, FRESH's contiguous explanations are unsuitable for modalities like tabular or time-series data, where feature proximity lacks relevance.}

%%%%%% First subsection: Performance %%%%%%%%%%%%%%
\subsection{Exp. 1: \icc does not compromise on performance}
\label{sec:performance}
Table \ref{tab:all_results} shows the models' metrics (average accuracy, standard deviations, and $95\%$ confidence intervals) across three iterations of model training for all eight datasets for \icc Feature Gating, Group Routing, and Top-K Routing, as well as baselines of NAM, SENN Features, SENN Concepts, and a non-interpretable model (reflecting post-hoc explainer performance). We do not report results for text features with NAM and SENN, since the architecture change required to do so would no longer respect the original model design\footnote{For NAM, one network (LLM) would be required per word, as the words are distinct for each instance; it does not support text grouping. For SENN, LMs would need to embed each word, then simultaneously be trained to represent concepts and pick representative words, with new metrics for choosing prototypical examples.}; \textcolor{black}{instead we report text results for FRESH}. \textcolor{black}{Additional sensitivity and architecture experiments can be found in Appendix \ref{appendix:sparsity}. These experiments show that while the performance of \icc has overlapping 95\% CIs while changing parameters, certain parameter settings have higher variability than others. For both education and health tasks, a $\tau$ of 10 and a Gumbel-Softmax threshold of around 0.7 to 0.8 are performant, sparse in activated features, and relatively stable.}

% Feature gating
\icc \featuregating statistically significantly improves performance with respect to the non-interpretable base prediction module for two EDU courses (DSP, HWTS) and the synthetic dataset. It shows comparable performance on all other datasets (indicated by the overlapping CIs) except the Breast Cancer dataset, suggesting that a higher number of the available features is necessary for performant prediction. The intrinsically interpretable baselines \textcolor{black}{SENN, NAM, and FRESH never outperform the base model, and always under-perform or perform comparably to \icc}.

\begin{figure}[t]
    \centering
    \begin{minipage}[b]{0.55\textwidth}
    \centering
    \resizebox{\textwidth}{!}{
\begin{tabular}{@{}lllll@{}}
\toprule
\multicolumn{1}{c}{\multirow{2}{*}{\textbf{Dataset}}} & \multicolumn{1}{c}{\multirow{2}{*}{\textbf{Baseline}}} & \multicolumn{3}{c}{\textbf{\icc Group Routing}} \\ \cmidrule(l){3-5} 
\multicolumn{1}{c}{} & \multicolumn{1}{c}{} & \multicolumn{1}{c}{\textit{Paper}} & \multicolumn{1}{c}{\textit{Pattern}} & \multicolumn{1}{c}{\textit{GPT-4}}\\ \midrule % & \multicolumn{1}{c}{\textit{Average}} 
\textbf{DSP} & 82.81 \small{$\pm$ 2.61} & 82.37 \small{$\pm$ 6.27} & 82.29 \small{$\pm$ 3.72} & \textbf{84.90 \small{$\pm$ 7.59}}  \\ %& 83.19 \small{$\pm$ 5.41}
\textbf{Geo} & 72.96 \small{$\pm$ 1.59} & 69.64 \small{$\pm$ 1.23} & \textbf{81.58 \small{$\pm$ 0.57}} & 81.19 \small{$\pm$ 1.53} \\ %& 77.47 \small{$\pm$ 5.96} 
\textbf{HWTS} & 73.93 \small{$\pm$ 3.76} & \textbf{78.34 \small{$\pm$ 0.95}} & 72.34 \small{$\pm$ 2.77} & 75.12 \small{$\pm$ 4.17}  \\ %& 75.27 \small{$\pm$ 3.64}
\textbf{VA} & 74.90 \small{$\pm$ 5.28} & 69.88 \small{$\pm$ 2.93} & \textbf{72.08 \small{$\pm$ 3.71}} & 70.98 \small{$\pm$ 2.77} \\ \midrule %&70.98 \small{$\pm$ 2.90} \\ 
\textbf{Average} & 76.65 \small{$\pm$3.31} & 75.56 \small{$\pm$2.85} & 77.57 \small{$\pm$2.69} & \textbf{78.05 \small{$\pm$4.01} } \\ %& 76.73 \small{$\pm$4.23}
\bottomrule
\end{tabular}
}
\captionof{table}{\textbf{\icc Group Routing Performance}: balanced accuracy (average $\pm$ std) on routing strategies (paper, pattern, GPT-4) for the EDU datasets in comparison to the non-interpretable baseline.}
\label{tab:group-routing}
    \end{minipage}
    \hfill
    \begin{minipage}[b]{0.42\textwidth}
        \centering
    \includegraphics[width=\linewidth]{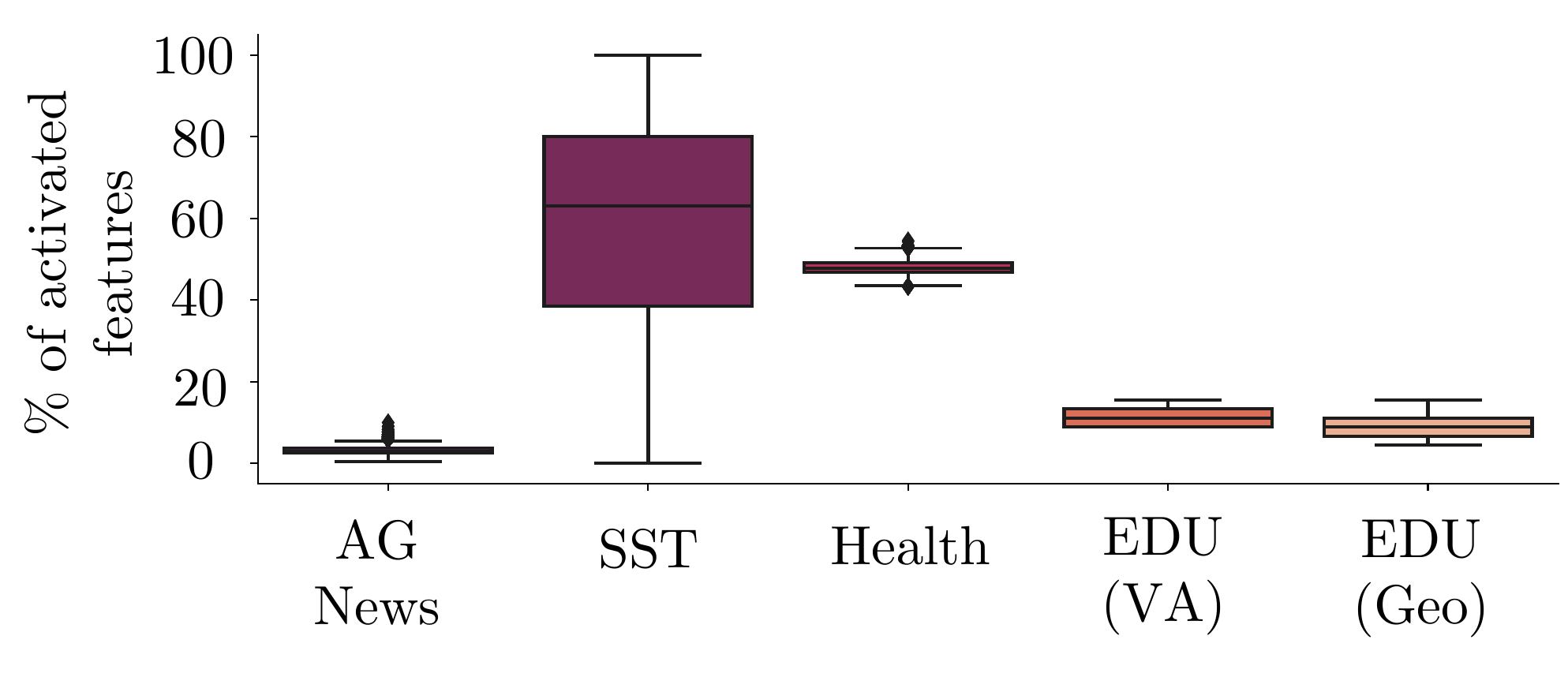}
    \caption{\textbf{\icc Feature Gating Sparsity:} \% of features activated per data point across five representative datasets.}
    \label{fig:sparsity}
    \end{minipage}
\end{figure}

\icc \grouprouting outperforms (for the Geo course and Breast Cancer dataset) or performs comparably ($95\%$ CI overlap) to the fine-tuned, non-interpretable base module. It also consistently performs at least comparatively to SENN which, besides never outperforming the base model, has a relatively high variance. 
We further observe that the selected grouping method impacts performance (Table \ref{tab:group-routing}). We achieve a $10\%$ increase in performance compared to the base model when grouping using patterns or GPT-4 for the Geo course. On average, over the four EDU courses, the automated LLM grouping and the pattern-based human-defined grouping perform comparably, showing that using automated grouping methods does not mean compromising on performance.

\begin{graybox} \icc performs comparably to black-box models and outperforms intrinsically interpretable baselines across diverse benchmarks. \end{graybox}

%%%%%% Second subsection: Faithfulness %%%%%%%%%%%%%%
\subsection{Exp. 2: \icc provides faithful and user-friendly explanations}
\label{sec:faithfulness-sparsity}
\begin{figure}[t]
    \centering
    \begin{minipage}[b]{0.34\textwidth}
        \centering \includegraphics[width=\linewidth, trim={5 5 5 5}, clip]{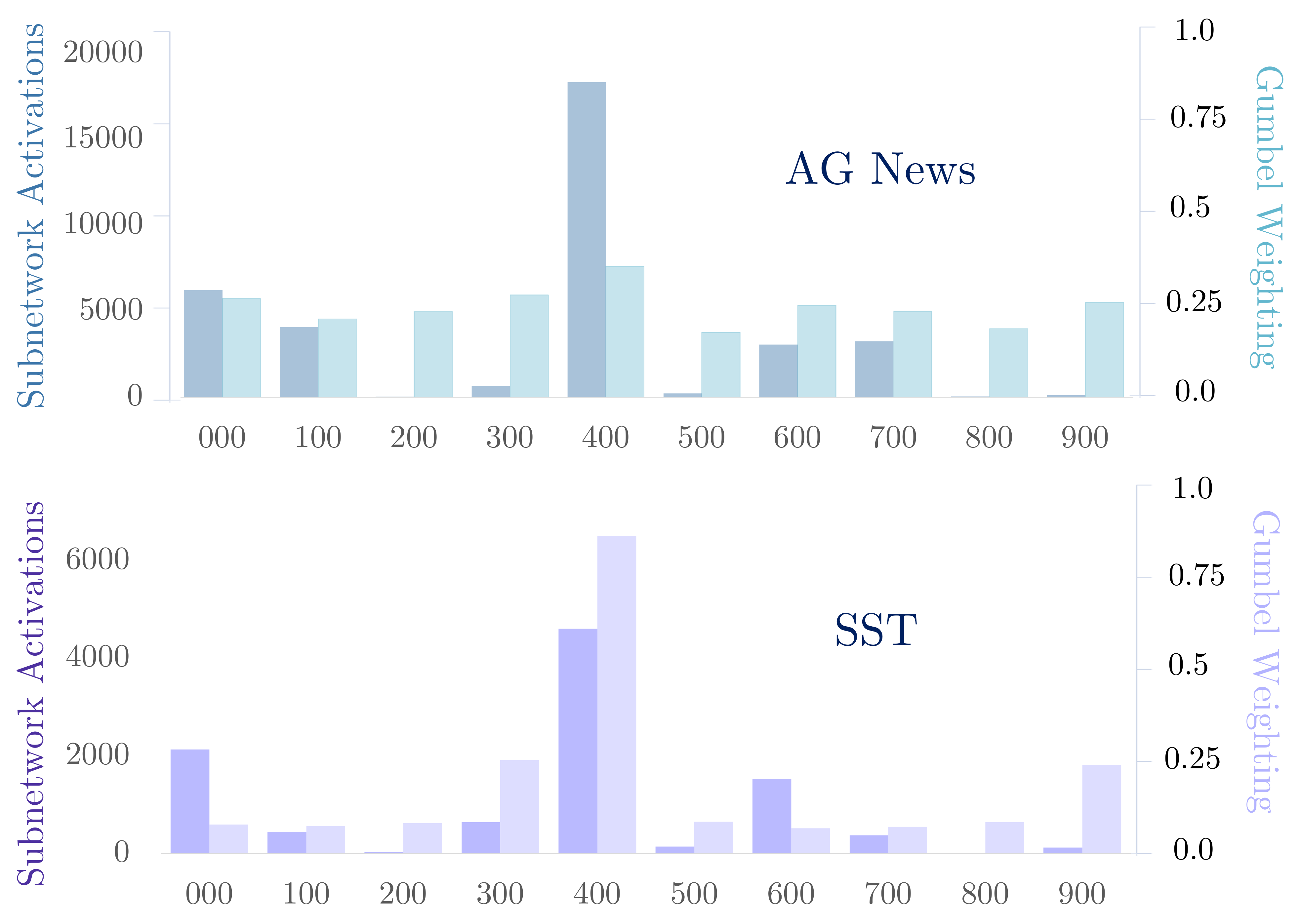}
        \caption{\textbf{AG News} and \textbf{SST}: \# of ICC subnetwork activations (left) vs. avg. activation weights (right), grouped by subnetworks based on the Dewey Decimal Code.}
        \label{fig:ag-news-comparison}
    \end{minipage}
    \hfill
    \begin{minipage}[b]{0.63\textwidth}
\centering
    \resizebox{\textwidth}{!}{
\begin{tabular}{@{}llllc@{}}
\toprule
\multicolumn{1}{c}{\multirow{2}{*}{\textbf{Model}}} &
  \multicolumn{3}{c}{\textbf{Relationship to Underlying Data Patterns}} &
  \textbf{Relationship to Model} \\ \cmidrule(l){2-5} 
\multicolumn{1}{c}{} &
  \multicolumn{1}{c}{\textit{\begin{tabular}[c]{@{}c@{}}Ground Truth \\ Alignment (GTA) \end{tabular}}} &
  \multicolumn{1}{c}{\textit{\begin{tabular}[c]{@{}c@{}}Rank \\ Agreement (RA) \end{tabular}}} &
  \multicolumn{1}{c}{\textit{\begin{tabular}[c]{@{}c@{}}Feature \\ Agreement (FA) \end{tabular}}} &
  \textit{\begin{tabular}[c]{@{}c@{}}Ground Truth \\ Faithfulness (GTF) \end{tabular}} \\ \midrule
\texttt{ICC} FG        & \textbf{94.84 ± 4.11} & \textbf{60.00 ± 12.65} & 87.99 ± 16.00 & always \textbf{100} \\
\texttt{ICC} GR        & 89.51 ± 0.56 & 44.00 ± 14.97 & 76.00 ± 23.32 & always \textbf{100} \\
SENN Features & 85.83 ± 2.56 & 27.99 ± 9.80  & 88.01 ± 9.79  & always \textbf{100} \\
SENN Concepts & 65.19 ± 9.23 & 52.00 ± 20.39 & 80.00 ± 21.90 & always \textbf{100} \\
NAM           & 87.39 ± 2.45 & 40.00 ± 17.89 & 76.00 ± 14.97 & always \textbf{100} \\
IG            & 87.28 ± 1.72 & 56.00 ± 14.97 & \textbf{91.99 ± 16.00} & $\leq 100$          \\
LIME          & 84.75 ± 1.83 & 44.00 ± 23.32 & 64.00 ± 14.97 & $\leq 100$          \\
SHAP          & 83.47 ± 1.03 & 36.00 ± 8.00  & 52.00 ± 20.39 & $\leq 100$          \\ \bottomrule
\end{tabular}}
    \captionof{table}{\textbf{Synthetic faithfulness evaluation} across eight models on explanations using metrics presented in OpenXAI \cite{agarwal2022openxai}. GTA, RA, and FA (K=5) refer to the explanations' relationship to underlying patterns in the synthetic data. \textcolor{black}{GTF refers to bounds of the explanations' relationship to the model's decision process.} All 95\% CIs overlap between the best performing intrinsically interpretable and non-interpretable model.}
    \label{tab:synth}    
\end{minipage}
\end{figure}

Table \ref{tab:synth} showcases the faithfulness of \icc models in comparison to intrinsically interpretable models SENN and NAM, as well as three post-hoc explainers on top of a non-interpretable model (Integrated Gradients (IG) \cite{sundararajan2017axiomatic}, LIME \cite{lime}, and SHAP \cite{shap}). We examine the relationship of the explanations to underlying data patterns from the synthetic dataset as well as to the ground truth of the underlying model (see Appendix \ref{app:additional-faithfulness} for detailed descriptions of the metrics). All 95\% CI overlap in Ground Truth Alignment (how close the explanation is to the underlying synthetic data feature patterns, similarly to \cite{liu2021synthetic}), Rank Agreement (RA) and Feature Agreement (FA) from ~\cite{agarwal2022openxai}, except for SENN Concepts performing statistically significantly worse than \icc Feature Gating in GTA.  This indicates that \texttt{ICC}'s identification of important features, and of their order of importance, is close to the underlying data patterns and is either on par or better than both interpretable and non-interpretable models. In terms of predictive performance on the synthetic dataset (last row of Table \ref{tab:all_results}), \texttt{ICC} holds the top three best performing models (89.47, 89.51, 90.83) and the \texttt{ICC} FG variation is significantly more performant than the non-interpretable base model, demonstrating the models' ability to capture signal on this dataset.

% Besides being faithful, our models enable concise and hence user-friendly explanations by sparse feature (group) activations \cite{swamy-eval,miller2019explanation}.
% Figure \ref{fig:sparsity} shows the percentage of activated features per data point for \texttt{ICC} \featuregating. For \textbf{EDU}, only $\sim$ 10\% of the $45$ features are activated, with low standard deviation. For \textbf{Breast Cancer}, more features seem to be needed to make the prediction: $39.7\%$ of the features are activated per data point. 
% Contrarily to other datasets, for text-based data, the number of available features (words) varies per data point. \textbf{AG News} consists of news articles (average number of words per sample: $35$), and only a small percentage of words is activated for each article. \textbf{SST} contains much shorter sentences ($7$ words on average), with high variance, leading to a higher percentage of features selected per data point ($59.8\%$) as well as a high standard deviation. 
% The achieved sparsity, concisely indicating the most important features in the original data set (especially for \textbf{EDU} and \textbf{AG News}), is a major advantage of our feature gating approach. In comparison, post-hoc explainers often select a wider range of important features (e.g., in previous work \cite{swamy-eval} on \textbf{EDU}, LIME and SHAP \cite{lime, shap} indicated broad importance over all input features).

Our models provide concise and hence user-friendly explanations through sparse feature (group) activations \cite{miller2019explanation}. Figure \ref{fig:sparsity} shows the percentage of activated features for \texttt{ICC} \featuregating. For \textbf{EDU}, only about 10\% of the 45 features are activated with low variance. For \textbf{Breast Cancer}, 39.7\% of the features are activated per data point. Unlike other datasets, text-based datasets have a variable number of features (words). In \textbf{AG News}, with an average of 35 words per article, only a small percentage is activated, while in \textbf{SST}, which has shorter sentences (7 words on average), 59.8\% of features are selected with high variance. This achieved sparsity, especially in \textbf{EDU} and \textbf{AG News}, highlights the most important features. This contrasts with post-hoc explainers, which tend to select a broader range of features (e.g., LIME and SHAP for EDU \cite{swamy-eval}).

 \texttt{ICC} \grouprouting activates different subnetworks with different weights for each data point. Figure \ref{fig:ag-news-comparison} illustrates the number of activations and the average weight for each subnetwork for the text data sets (see Appendix \ref{app:group_routing} for detailed analysis on the Breast Cancer and EDU datasets). For \textbf{AG News} (Figure \ref{fig:ag-news-comparison} top), the average activation weight is similar across all subnetworks (min $0.10$, max $0.21$). However, some subnetworks are activated much more frequently (400 - Language: $18,335$ times). This indicates that most data points will be routed through the same subset of subnetworks, while the remaining subnetworks are important for specific data points only. \textbf{SST} (Figure \ref{fig:ag-news-comparison} bottom) shows similar subnetwork activation patterns. However, in contrast to \textbf{AG News}, the distribution of average weights is not uniform: only three networks are activated with weights larger than $0.15$. We suspect the high weights for subnetwork 400 in Figure \ref{fig:ag-news-comparison} reflect words that DDC has little relation to in the SentenceBERT embedding space.

Examples of EDU domain explanations are in Appendix \ref{app:user-study}, Figures \ref{fig:userstudy-fg} and \ref{fig:userstudy-pr}, with additional AG News examples in Figure \ref{fig:text-example}. We also compare \texttt{ICC} behavior across three grouping strategies for the DSP course (Appendix \ref{appendix:edu_dsp_gr}) and analyze network sparsity at different prediction horizons (\ref{appendix:edu_dsp_sparsity}). Lastly, we show the variation in feature group selection across MOOCs, highlighting \texttt{ICC}'s adaptability regardless of grouping method (Appendix \ref{app:fg_allcourses}).

% We suspect the high weights showcased for 400 in Figure \ref{fig:ag-news-comparison} are representative of words the DDC does not have a close relation to in SentenceBERT embedding space. Examples of explanations for the EDU domain can be found in Appendix \ref{app:user-study}, Figures \ref{fig:userstudy-fg} and \ref{fig:userstudy-pr}, and examples from the text dataset can be found in Figure \ref{fig:text-example}. We also compare \texttt{ICC} behavior across the three grouping strategies for the DSP course (Appendix \ref{appendix:edu_dsp_gr}) and analyze the network sparsity at different prediction horizons (\ref{appendix:edu_dsp_sparsity}). Finally, we show the variation of the selection of feature groups across all MOOCs, indicating the adaptability of \texttt{ICC} in selecting relevant subnetworks whatever the grouping method (Appendix \ref{app:fg_allcourses}).
\begin{graybox} \icc provides sparse and hence user-friendly explanations, while not compromising on explanation faithfulness. \end{graybox}

%%%%%%%%%% Subsection 4: User Study %%%%%%%%%%%%%%%%%%%
\subsection{Exp. 3: \icc explanations are preferred by humans}
\label{sec:user-study}

\textbf{Setting.} To validate the user-centeredness of our approach, we conducted a user study comparing \icc's explanations with the ones from other intrinsically interpretable methods, SENN and NAM. We focused on the education domain and time series input type, using the DSP course in the EDU dataset. We trained \icc \featuregating and \grouprouting (with pattern-based feature grouping strategy, since it is heavily grounded on expert knowledge) as well as SENN and NAM and randomly selected four test samples (\textit{i.e.}, four students) for prediction. In designing the study, we conducted four sequential pilots with eight learning scientists who were unaware of which method we presented in this chapter. All explanations were simplified for a non-technical audience and followed consistent templates.

We recruited 56 teachers using \texttt{Prolific}, (see Appendix \ref{app:user-study} for detailed information about the participants' demographics and backgrounds (Figure \ref{fig:userstudy_demo}) and the content of the study). We showed them each model's prediction of the student's success or failure along with its explanation. The explanations were given as a short text and a graph showing the features and concepts used by the model. Note that the choices we made for the presentation of the explanations might have an influence on the participants' perception of the explanations. Examples of how the explanations were presented and a discussion on this limitation can be found in Appendix \ref{app:user-study}, Figures \ref{fig:userstudy-fg}, \ref{fig:userstudy-pr}, \ref{fig:userstudy-senn} and \ref{fig:userstudy-nam}. With our study design, we aimed to highlight each framework's strengths without excessive post-processing. For instance, keeping only the top five features of NAM would be unfair, as other features also contribute to its predictions. Instead, we emphasized the top five features that each contribute positively, negatively, or not at all to the prediction, highlighting its advantage (distinct insights into feature impact) over \icc, while ensuring sparsity. We significantly post-processed the explanations of SENN and NAM to provide them in a format understandable for a non-technical audience and iterated on the visualizations using a human-centered design process \cite{cooley2000human}. We asked participants to compare these explanations according to five criteria (aligned with \cite{frej2024finding}), and to rank the criteria in terms of importance: \textbf{Usefulness}: This explanation is useful to understand the prediction. \textbf{Trustworthiness}: This explanation lets me judge if I should trust the model. \textbf{Actionability}: This explanation helps me know how to give feedback to the student. \textbf{Completeness}: This explanation has sufficient detail to understand why the prediction was made. \textbf{Conciseness}: Every detail of this explanation is necessary. 

\begin{figure}[h]
    \centering
    \begin{minipage}[t]{0.45\textwidth}
        \centering
        \vspace{-20mm}
        \resizebox{\linewidth}{!}{%
        \setlength{\tabcolsep}{3pt}
        \begin{tabular}{@{}l*{5}{r}@{}} 
        \toprule
        {} & \textbf{NAM} & \textbf{SENN}  & \textbf{\texttt{ICC GR}} & \textbf{\texttt{ICC FG}} & \textbf{Weight} \\
        \midrule
        Usefulness       &  3.25 \footnotesize{$\pm$0.98} &   3.3~~ \footnotesize{$\pm$1.11} &  3.53 \footnotesize{$\pm$1.11} &  \textbf{3.88 \footnotesize{$\pm$0.94}} &     0.28 \\
        Trustworthiness  &  3.28 \footnotesize{$\pm$0.93} &  3.64 \footnotesize{$\pm$0.92} &  3.36 \footnotesize{$\pm$1.06} &  \textbf{3.78 \footnotesize{$\pm$0.9~~}} &     0.23 \\
        Actionability    &  3.08 \footnotesize{$\pm$0.96} &  3.25 \footnotesize{$\pm$1.06} &  3.37 \footnotesize{$\pm$1.04} & \textbf{3.77 \footnotesize{$\pm$0.95}} &     0.21 \\
        Completeness     &  3.18 \footnotesize{$\pm$1.02} & \textbf{3.76 \footnotesize{$\pm$1.09}} &   3.1~~~\footnotesize{$\pm$1.19} &  3.67 \footnotesize{$\pm$1.07} &     0.16 \\
        Conciseness      &  3.13 \footnotesize{$\pm$1.06} &  2.82 \footnotesize{$\pm$1.31} & \textbf{3.72 \footnotesize{$\pm$1.06}} &  3.68 \footnotesize{$\pm$1.05} &     0.12 \\
        \midrule
        Global &   3.2~~~\footnotesize{$\pm$0.81} &  3.38 \footnotesize{$\pm$0.85} &  3.41 \footnotesize{$\pm$0.88} &  \textbf{3.78 \footnotesize{$\pm$0.77}} &      \\
        \bottomrule
        \end{tabular}}
        \caption{\textbf{Model score for each \textcolor{black}{user study} criterion} (average $\pm$ std) and criteria weight according to \textcolor{black}{users'} ranking. All scores range from 1 (lowest) to 5 (highest).}
        \label{tab:userstudy_res}
    \end{minipage}
    \hfill
    \begin{minipage}[t]{0.52\textwidth}
        \centering
        \resizebox{\linewidth}{!}{%
        \includegraphics[trim={3mm 3mm 1mm 3mm},clip]{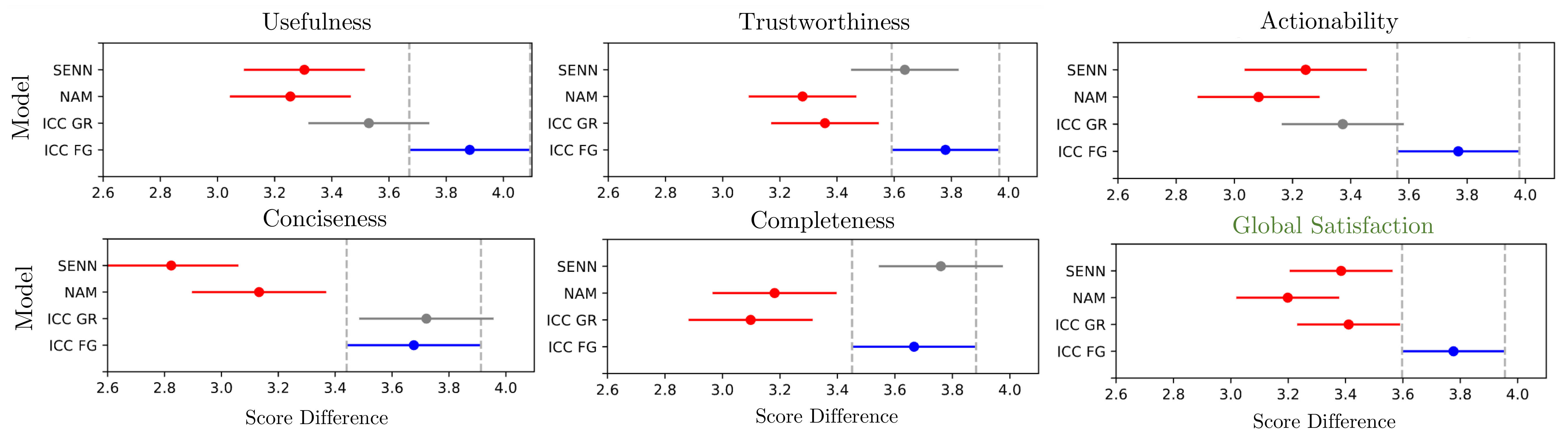}}
        \caption{\textbf{Tukey's Honest Significant Difference (HSD) Test} for each user study criterion and overall global satisfaction. ICC FG (with highest overall satisfaction in Table \ref{tab:userstudy_res}) is in blue.}
        \label{fig:tukey}
    \end{minipage}
    \vspace{-3mm}
    \label{fig:side_by_side}
\end{figure}

\textbf{Results.} Using Friedman's Chi-Square test, we verify that the ranking of the criteria is consistent among the participants ($p < 0.05$). We report normalized criteria importance in Table \ref{tab:userstudy_res} (last column). We observe that Usefulness, Trustworthiness and Actionability are consistently ranked the highest by the participants. We compute the \textit{Global satisfaction} score as a weighted average of the criteria as a global satisfaction measure for each model (Table \ref{tab:userstudy_res}, last row). We additionally conduct pairwise Tukey's Honest Significant Difference (HSD) tests (Fig. \ref{fig:tukey}) to determine the statistical significance of the differences between the means of user preferences per criteria.

We observe that \texttt{ICC FG} ranks the highest in the top 3 most important criteria as well as in global satisfaction. Furthermore, \texttt{ICC GR} ranks second in Usefulness and Actionability and first in Conciseness. Overall, \icc models are favored over interpretable baselines in 4 out of 5 criteria and in terms of global satisfaction. An ANOVA performed separately for each criterion as well as the global satisfaction measure (Table \ref{tab:userstudy_anova}, Appendix \ref{app:userstudy_res}), indicates that there is indeed a significant difference between the models for each criterion. Tukey's HSD tests confirm that \texttt{ICC FG} significantly outperforms NAM and SENN on all criteria but completeness 
(Fig. \ref{fig:tukey}).
\begin{graybox} Participants prefer \icc explanations in terms of usefulness, trustworthiness, actionability, and conciseness over other intrinsically interpretable models.\end{graybox}

\section{Discussion and Conclusion}
\label{sec:discussion}
We proposed \icc, a family of intrinsically interpretable models that puts human understanding at the forefront of the design. Through our experiments on feature gating and group routing (mixture-of-expert) models, we demonstrated that our modular architectures optimize for interpretability but do not compromise on performance. In a real-world setting, we showed that \icc models are preferred over other intrinsically interpretable models in 4 of 5 explainability criteria.

\icc is a locally intrinsic explanation framework that creates explanations that are specific to the input point and guaranteed to be faithful to the model. 
In our architecture, the discriminator network is a black-box model. This is by design, to minimize explanation complexity. We believe the hierarchical prediction logic enabled by \icc is the type of explanation that a user wants: ``which concepts/features were used to make the prediction, and how important are they to the prediction (weighted sum)?” If the discriminator network was a glass-box model, it would answer the question: ``why were these concepts/features selected for the prediction?”. The first explanation is directly actionable, while the second type of explanation is not. We therefore do not seek to answer this second question with our architecture. However, it would be possible for \icc to have an interpretable discriminator network, as discussed further in Appendix \ref{sec:interpretable-discriminatory}.

\texttt{ICC FG} is best when individual features are important for actionable decisions based on the explanation. However, it can lose sparseness if all features are equally important: consider the extreme case where the prediction is a sum over the full feature space and all features are independent. \texttt{ICC GR} requires more human effort than \texttt{ICC FG} and is, therefore, more suitable for scenarios with cross-feature dependencies and where broader concepts are more actionable than individual features.

\textcolor{black}{\texttt{ICC GR} model's global mixture-of-experts design specializes subnetworks on subsets of features allows them to learn granular patterns (Appendix \ref{app:specialization}). Combining these specialists enhances prediction compared to a monolithic network that might miss underlying patterns (Table \ref{tab:all_results}). Expert-informed feature groupings help the model avoid overfitting to correlations that do not generalize at inference time. For \texttt{ICC FG}, adaptive sparsity (a few features per instance) improves prediction quality by reducing noise and optimizing the interpretability-accuracy tradeoff.}

User-defined feature groups aim at deriving explanations useful to the user, but might compromise performance if the user specifies a grouping that carries minimal signal. Regardless of the grouping, \icc optimizes for explanation actionability and understandability over performance. We believe that an accurate prediction at the cost of explanation usefulness is not worthwhile in an applied setting. \textcolor{black}{It is possible that \texttt{ICC} explanations could be misleading, as concepts used in the explanation could lead to a correlation that was not intended \cite{zheng2021irrationality, jacovi2021aligning}. In these cases, we view \texttt{ICC} 
 explanations through the lens of auditing model behavior \cite{yadav2022xaudit}, and encourage human intervention.} \textcolor{black}{\icc's user-centric advantages are highlighted when the input space is human-interpretable. However, for domains that are hard to obtain expert knowledge, we envision \texttt{ICC} increasingly leveraging LLM-extracted features, reducing the necessity of human effort for human-centric explanations \cite{malberg2024felix, baddour2024phenotypes}.}

We acknowledge that the presentation of explanations in our user study (Section \ref{sec:user-study}) has influence over our preliminary results on the user perception of \icc. We conducted extensive iteration with eight pilot participants to mitigate study design bias. We note that any imbalance in wording is not necessarily in favor of our method; for instance, users found SENN's explanations more complete than \icc (Fig. \ref{fig:tukey}). \textcolor{black}{The user study  prioritizes diversity of study participants and quality of responses over number of samples evaluated; the  task is mentally intensive and we found a longer study can cause a drop in participant attentiveness.} An extensive study over many different prediction tasks and domains of expertise is necessary for generalizable conclusions.

For our text experiments, we fine-tune twenty \texttt{DistilBERT} models as experts (ten for each task). For more complex tasks, for example requiring long context size, multi-step reasoning ability, or strong prior domain-specific knowledge, \texttt{DistilBERT} can be swapped with larger decoder models, either through fine-tuning or in-context learning. Parameter-efficient fine-tuning such as LoRA \cite{hu2021lora} would allow fine-tuning even large LMs with limited computational cost, while in-context learning would use the same model instance for each feature and feature group. 

Overall, we encourage the machine learning community to design models for interpretability at many different granularities and user-specified requirements. \icc provides one such family of models as a tradeoff between human specification, explanation certainty, and performance.

%% file: main/09_robustness_of_explainers.tex
The work presented in this chapter is under submission to  the journal Computers and Education: AI in collaboration with Lan Tran, Gabriele D'Angeli, Juliette Parchet, Jade Mai Cock, Jibril Frej, Tanya Nazaretsky, and Professor Tanja Käser. The provisional title of the article is \textit{Teacher Perceptions of Inconsistent eXplainable AI: Exploring Consistency of Student Explanations through Adversarial Training}. We would like to thank our 20 learning science experts for their participation in our study.

This chapter is a deep dive into explanation \textbf{consistency} in the student modeling context, both through technical and human studies. We first explore the perceptions of model and explanation inconsistency from education experts, identifying at which threshold inconsistency becomes obtrusive. We further explore mitigation effects of adversarial training on explanation stability across explainability methods and course settings. The sparsity enabled by adversarial training for explainer consistency (under certain settings) contributes to more \textbf{human understandable} explanations.

\section{Introduction}
Chapter 3 demonstrated that the different methodologies for explaining black-box ML models do not agree with each other when applied to the same representative set of students \cite{swamy2022}. Chapter 4 further validated explanations through the assistance of university-level educators, finding that experts often disagree when it comes to determining which explainers are the most valid for different course analytics \cite{swamy2023trusting}. This divergence of opinions highlights the subjective nature of evaluating the effectiveness and trustworthiness of explanation methods.

% in the ML community
Unfortunately, there is an additional complication: neural network architectures are inherently unstable at training, delivering inconsistent predictions. Research on the sources of instability for deep learning models suggests that the random initialization of weights or attention layers can strongly contribute to different underlying logical representations \cite{narkhede2022review,zhang2022all}. Additionally, neural networks are vulnerable to small perturbations in input \cite{novak2018sensitivity}, and even sometimes benefit from them \cite{srivastava2014dropout}. While prior research has suggested mitigation strategies using adversarial robustness \cite{FGSM}, these remain unexplored in the ML for education community.
In the context of education, the instability of neural networks makes it difficult to provide explanations that humans can trust. Consider the following educational scenario: a student predicted to perform poorly on an upcoming exam is advised to study chapter 5 but does not have time to do so. The next day, they receive conflicting advice, stating their chapter 5 performance is excellent, and that they should focus on chapter 3 videos. Such inconsistent interventions, derived from explanations of neural network behavior, can erode trust and reliability in model-based advisories \cite{hsieh2020evaluations}. Within this context, we explore whether using adversarial training can offer more consistent and, therefore, trustworthy explanations \cite{hsieh2020evaluations} through two methods: Fast Gradient Sign Method (FGSM) \cite{FGSM} and Projected Gradient Descent (PGD) \cite{PGD}.

In this chapter, we explore the potential of robustly trained models to improve the explainability of deep learning models through a comprehensive analysis across 15,000 students from six distinct courses spanning both MOOC and flipped classroom settings, and across three explainability methods. First, we aim to understand how explanation consistency affects learning scientists, investigating their benchmarks for trusting AI in the classroom. To this end, we conduct a human study involving semi-structured interviews with 20 experienced learning scientists to assess the threshold at which explanation inconsistency begins to erode trust. Our goal is to assess the subtle differences in how participants perceive disagreements between two expert sources — either a pair of teaching assistants (TAs) or two models — under identical pedagogical scenarios. Secondly, we evaluate the sources of inconsistency in counterfactual explanations, examining the role of random initialization in influencing the explanations generated by a specific model. Thirdly, we explore whether robustly trained models, by employing adversarial techniques to stabilize decision boundaries, can produce higher-quality (i.e., more consistent) counterfactual explanations. Finally, we evaluate the generalizability of these findings to other learning settings and additional post-hoc explainers. With our analyses, we aim to answer the following research questions:

\begin{enumerate}
%\small
    \item How do learning science experts perceive inconsistency in explanations (\textbf{RQ1})?
    \item What are the sources of inconsistency in counterfactual explanations (\textbf{RQ2})?
    \item Can robustly trained models generate higher quality (more consistent) counterfactual explanations (\textbf{RQ3})?
    \item Do these results generalize to other learning settings and explainers (\textbf{RQ4})?
\end{enumerate}

We find that learning science experts have diverse explainability needs in different educational settings and that they are much more sensitive to inconsistencies in models than in humans (TAs). Our findings demonstrate that randomness in model weight initialization can create vastly different explanations from the same data, architecture, and computing environment. We show that robustly trained models have more consistent explanations than their non-robust counterparts, choosing fewer and less variable important features for more actionable interventions while maintaining the same level of accuracy. We provide our full code, analyses, and expert study materials in our repository\footnote{\url{https://github.com/epfl-ml4ed/robust-explainers}}.

%%%%%% METHOD SECTION %%%%%%%%%%%
\section{Robustness Evaluation Methodology}
\label{sec:method}
%%%%%%%%%%%%%%%%%%%%%%%%%%%%%%%%

\begin{figure}[t]
    \centering
    \includegraphics[width=\textwidth]{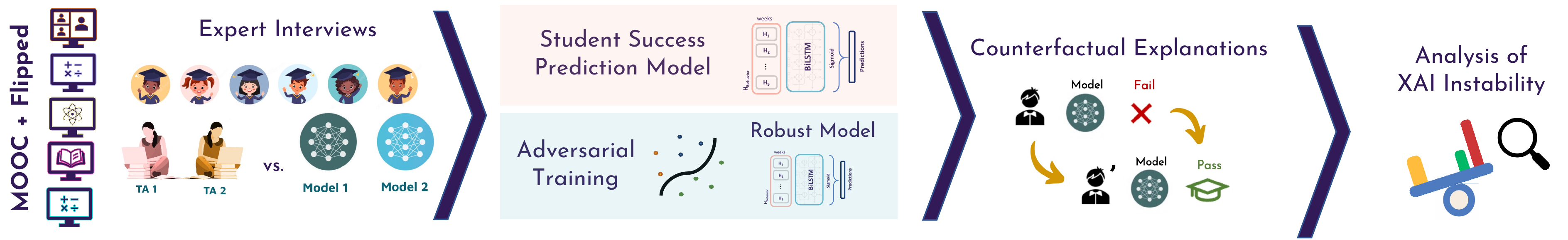}
    \caption{\textbf{Study Overview}: We begin by consulting learning scientists to assess their perceptions of explanation variance. In the baseline pathway  (orange), we 1) extract clickstream data and 2) train a model for student success. In the robustness pathway (blue), we 1) generate adversarial examples and 2) robustly train two new models. Both pathways also involve: 3) extracting counterfactual explanations and 4) measuring explanation instability.}
    \vspace{-4mm}
    \label{fig:pipeline}
\end{figure}

We evaluated explanation robustness and human perception of inconsistent explanations using the framework depicted in Figure ~\ref{fig:pipeline}. We examined two pathways for predicting student success in online learning. In the first pathway (orange), we extracted clickstream data from five courses, used a bi-directional LSTM for prediction, generated counterfactual explanations, and assessed explanation stability. Semi-structured interviews with learning science experts were conducted to understand the impact of inconsistent explanations for educational applications. In the second pathway (blue), we employed two adversarial robustness techniques to create synthetic student adversarial examples, trained new robust models, and measured the conciseness and consistency of the explanations.

%%%%%%%%% Learning Context: MOOCs%%%%%%%%%%%%%%%%%%
\subsection{Learning Context}
\label{sec:learning context}
Our analysis focused on interaction and metadata collected from five distinct courses of a European University in the context of MOOCs and flipped classrooms (FC). Detailed specifications about each course are provided in Table \ref{tab:courses}. The study was approved by the university's ethics committee (Nr. [Anonymous]). We selected the courses to cover a wide range of attributes, including different levels (Bsc and MSc), languages (English and French), and topics (mathematics, computer science, engineering, and social sciences). Courses also differed in the size of student populations ($214$ to $>5000$ students) and pass ratios ($0.10$ to $0.59$). We deliberately included two courses used in Chapter 4 \cite{swamy2023trusting} (\textit{DSP 1} and \textit{DSP 2}) to increase comparability of the results.

The MOOCs were offered open-access, enabling global availability without restrictions. Each chosen course segmented content into weekly units, incorporating introductory video lectures on key concepts and self-assessment quizzes. In addition, students were tasked with completing graded assignments, valued at a maximum of $100$ points, determining eligibility for the course certificate. The assignment quantity and the minimum passing grade were at the discretion of the instructor. The FC courses \cite{hardebolle2022gender} were  mandatory for the Bachelor's degree and therefore included university students only. In the two FC courses, students were mandated to engage in online pre-class tasks like viewing lecture videos and completing related quizzes before participating in in-class sessions. Detailed specifications about each course are provided in Table \ref{tab:courses}.

\begin{table*}[t]
\caption{\textbf{Courses}: Detailed instructional design and student population information on the four MOOC courses included in the study, and the two flipped courses used for the generalization experiments.}
\label{tab:courses}
\small
\resizebox{\textwidth}{!}{
\begin{tabular}{llllllrrrrr}
\toprule
\textbf{Course Title} & \textbf{Course ID} & \textbf{Setting} & \textbf{Field$^1$} &  \textbf{Level} & \textbf{Language}  & \multicolumn{1}{r}{\textbf{\begin{tabular}[c]{@{}c@{}} Weeks\end{tabular}}} & 
\multicolumn{1}{r}{\textbf{\begin{tabular}[c]{@{}c@{}} Quiz \end{tabular}}} & 
\multicolumn{1}{r}{\textbf{\begin{tabular}[c]{@{}c@{}} Video \end{tabular}}} & 
\multicolumn{1}{r}{\textbf{\begin{tabular}[c]{@{}c@{}} Students$^2$\end{tabular}}} & \multicolumn{1}{r}{\textbf{\begin{tabular}[c]{@{}c@{}} Success Rate \end{tabular}}} \\
\midrule
Digital Signal Processing 1 & \textit{DSP 1} & MOOC &CS & BSc & French & 10 & 41 & 61 & 5629   & 0.27\\
Digital Signal Processing 2 & \textit{DSP 2} & MOOC& CS & MSc & English & 10 & 41 & 62 & 4012  & 0.23 \\
Éléments de Géomatique&	\textit{Geometry}& MOOC& Eng.& BSc&French&11&32&35&452&$0.45$ \\
African Cities: An Introduction to Urban Planning&\textit{African Cities}&MOOC& SS&	BSc&	English&	12&	11 & 64& 5643& $0.10$\\
\midrule
Linear Algebra&\textit{Linear Algebra}&Flipped & Math & BSc & French &10 & 179 & 99 & 214 &$0.59$\\
Functional Programming&\textit{FP}&Flipped & CS & MSc & English &14 & 0 & 43 & 218 &$0.62$\\
\bottomrule
\end{tabular}}
\scriptsize{$^1$\textit{CS}: Computer Science; \textit{Math}: Mathematics; \textit{SS}: Social Sciences; \textit{Eng.}: Engineering.} \scriptsize{$^2$For MOOCs, after removing early-dropout students according to \cite{swamy2022meta}.}
\end{table*}

We collected metadata and clickstream data for each of the five courses. Each course had a different duration within a program, spanning multiple weeks ($\text{w} \in \{0, ..., W\}$). As students participate in a course, they engage with the learning materials outlined in the course schedule, resulting in the creation of a chronological clickstream. These clickstreams capture the sequence of actions taken by students, such as watching videos, pausing, and submitting quizzes, as they progress through the course. The clickstream of a specific course \text{c} for a student \text{s} is a time series. 

%%%%%%%%%%% Student Success Prediction %%%%%%%%%%%%%%%%%%%%%%%%
\subsection{Student Success Prediction}
\label{sec:blackbox-model}
%%%%%%%%%%% Student Success Prediction %%%%%%%%%%%%%%%%%%%%%%%%
The collected clickstreams in the form of time-series of actions build the basis for creating behavioral features. The engineered features serve as an input for the success prediction model. For a given course \text{c}. The model aims to accurately predict the success label $y_s$  (pass/fail) for the students based on the extracted behavioral features.

%%%%% Learning Indicators %%%%%%%%%%
\vspace{1mm} \noindent \textbf{Features}. We used the four feature sets identified as generalizable and predictive for student success in MOOCs by a recent meta review~\cite{marras2021can} as a basis for our success prediction model:

\begin{itemize}[leftmargin=*]
    \item \textit{Regularity \cite{boroujeni2016quantify} (3 features)} This feature set considers whether students study on specific hours of the day, day(s) of the week or similar weekdays. Other features included monitor whether students have the same distribution of study time among weekdays over weeks, particular the amount of study time on each weekday, and finally to what extent a student follows the schedule of the course.
    \item \textit{Engagement \cite{chen2020utilizing} (13 features)} This feature set involves counting clicks per student. Features include the number of online sessions (with average and standard deviation), standard deviation of time between online sessions, number of clicks during weekdays or weekends, and ratio of weekend to weekday clicks. 
    \item \textit{Control \cite{lalle2020data} (22 features)} The feature set is composed of two kinds of video interaction features, video views and specific actions taken within the videos. Features on video views include the total number of video views (both watches and re-plays), plus the average and standard deviation of the proportion of videos watched, re-watched, and interrupted per week. The features of actions performed within the videos include the frequency and a total number of all performed video actions, the frequency of video actions for each type of video action, and the average and standard deviation of the duration of video pauses, seek lengths, and so on.
    \item \textit{Participation \cite{marras2021can} (12 features)} In this set, features assess a student’s success in the course, taking into account quiz results and assignments, with features such as number of submissions, number of videos covered so far, number of videos watched for the given week, amount of quizzes passed on the first try, and amount of time taken to pass a quiz. We removed three features from this set that are directly related to a student’s success in the weekly assessments: student shape, competency strength, and competency alignment.
    \vspace{-4mm}
\end{itemize}

%%%%% Predictive Model %%%%%%%%%%%
\textbf{Predictive Model}. Based on the state-of-the art introduced in Chapters 3, 4 and 5 \cite{swamy2022meta}, we employed bidirectional LSTM (BiLSTM) layers for the student prediction task. The model takes as input a matrix with dimensions $|S| \times W \times |F|$ where $|S|$ is the set of students, $W$ is the number of weeks, and $|F|$ is the set of features. The resulting neural network architecture consists of two BiLSTM layers with sizes $32$ and $64$ and a Dense layer with a hidden size of $2$, activated using a Sigmoid function. Therefore, the model outputs the probability that a student will pass the course. 

%%%%%%%%%%% Explanations %%%%%%%%%%%%%%%%%%%%%%%%
\subsection{Adversarial Training}
\label{sec:adv-techniques}
In order to obtain adversarial examples, we employed two well-known techniques, namely the Fast Gradient Sign Method (FGSM) and Projected Gradient Descent (PGD). Adversarial examples are inputs that are slightly modified to intentionally mislead machine learning models, causing them to make incorrect predictions. We used these approaches to augment our initial dataset by an optimized extraction of synthetic adversarial examples during the model training process, aiming to robustly improve the model in relation to points that are out of distribution. 

The \textit{Fast Gradient Sign Method} (FGSM) is a simple yet effective technique for generating adversarial examples \cite{FGSM}. It perturbs the input data by adding or subtracting a small value in the direction of the gradients of the loss function with respect to the input. By controlling the magnitude of the perturbation, FGSM aims to maximize the prediction error. 

\textit{Projected Gradient Descent} (PGD) is an iterative optimization method that enhances the robustness of adversarial examples \cite{PGD}. It applies multiple iterations of gradient descent with small step sizes while projecting the perturbed data back into a valid range to ensure it remains within acceptable boundaries. This iterative approach allows PGD to explore a wider range of potential perturbations and find adversarial examples that are harder to detect and defend against. 

We generated separate sets of adversarial examples using each method for each course over five iterations, using these new points to train robust models. To enhance the robustness of the model and improve its ability to handle adversarial input, we augmented the original training data by incorporating the adversarial points into the training set.

%%%%%%%%%%% Explanations %%%%%%%%%%%%%%%%%%%%%%%%
\subsection{Generating Counterfactuals}
\label{sec:explanabilitymethods}
Although deep learning models have demonstrated superior accuracy compared to traditional machine learning techniques, their interpretability is limited. Neural networks are often described as black-box models due to their lack of transparency. To address this issue, post-hoc explainability methods can be employed to gain insight into the predictions of black-box models by providing feature importance scores. In our study, we utilized a counterfactual-based method to interpret the predictions generated by our prediction model. A counterfactual explanation of a prediction describes the smallest change to the feature values that changes the prediction to a predefined output.

The Contrastive Explanation Method (CEM) \cite{dhurandhar2018explanations} is a technique that determines the specific features that need to be present (pertinent positives) or absent (pertinent negatives) in order to maintain a model's prediction for a given student with behavioral features. We focused on pertinent negatives as they are more similar and comparable to other counterfactual-based explainability methods. For each generated pertinent negative, we calculated the importance score associated, that takes into account both the necessary perturbation of the feature and the significance of the change relative to the feature's range. To ensure comparability, we normalized the scores to a range between $0$ and $1$. This normalization allows for direct comparison of the resulting feature importance weights. 

%%%%%%%%%%% Expert Interviews %%%%%%%%%%%%%%%%%%%%%%%%
\subsection{Expert Interviews}
\label{sec:eval-qual}
To assess the impact of inconsistent explanations on humans, we conducted semi-structured interviews with learning science experts.

%%%%%%%%%%%%%%%Participants%%%%%%%%%%%%%%%%%%%%%%
\vspace{1mm} \noindent \textbf{Participants}. 
We conducted $20$ expert interviews ($5$ pilots, $15$ study participants). Study participants represented an international population, covering Switzerland, Brazil, Israel, the Netherlands, and USA. $95\%$ of the experts held affiliations with a university and $5\%$ held affiliations with an EdTech company. All participants held both teaching and learning science research positions and had more than $5$ years of experience. $65\%$ of the participants identified as male and $35\%$ as female. 

\begin{figure*}
    \includegraphics[width=\textwidth]{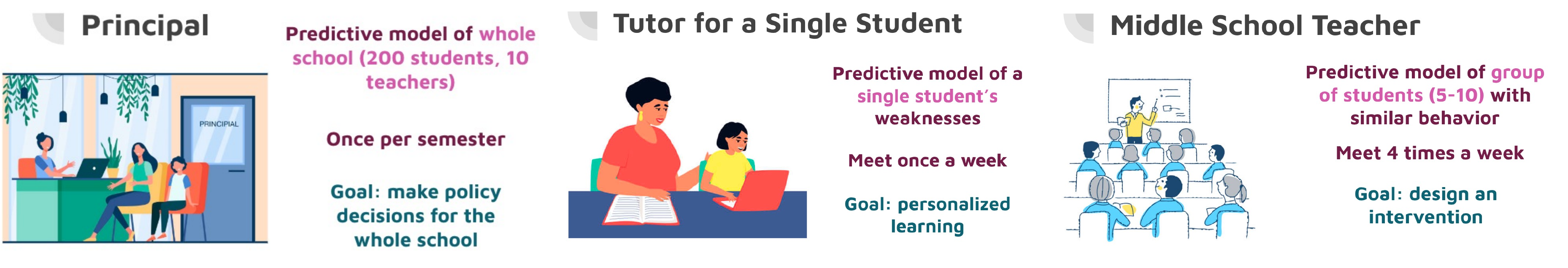}
    \caption{\textbf{Learning Science Expert Study}: Three learning scenarios at different granularities of instruction (principal, tutor, teacher) presented to learning science experts. The interviewer asks the expert to specify which needs of explainable AI are most relevant in each scenario.}
    \label{fig:scens}
\end{figure*}

%%%%%%%%%%%%%%%Procedure%%%%%%%%%%%%%%%%%%%%%%
\vspace{1mm} \noindent \textbf{Measuring Importance of Consistent Explanations}
The preliminary study presented three educational settings (shown in Figure \ref{fig:scens} to learning science experts at different granularities of student-teacher interactions. The goal of the preliminary study was to measure the importance of inconsistency across diverse educational settings. 

We first instructed the participants to consider themselves the principal of a school with 200 students and 10 teachers, where they were to make policy decisions (i.e. resource allocation) once a semester based on the explainability insights of an educational model. The second setting asked the participant to envision themselves as a tutor with a single student, where once a week they made decisions towards personalizing learning based an educational model that highlights the student's weaknesses. The third and final setting requested the participants to be a middle school teacher teaching a classroom of 30 students. They would have an educational model with explainable insights that they used four times a week to make learning interventions for a group of students (5-10 students). For each of these settings, the study participants were asked to evaluate on a Likert scale of 1 to 5 the importance of different explainability needs for models of student behavior (real-time, explanation faithfulness, actionability, human-understandability) as presented in \cite{swamy2023future, pinto2024towards}. The five explainability needs were randomly ordered with the goal of a fair measurement of participants' perceptions of consistency in comparison to the four other explainability needs.

\vspace{1mm} \noindent \textbf{Evaluating Perceptions of Inconsistency in Explanations}
In the second part of our expert study, we presented the following pedagogical scenarios to the participants.
%three pairs of identical student scenarios for human TAs and models. 
The participants were told that they were an instructor of a large course with more than $100$ students, so unfortunately they were unable to interact with all of their students individually. As an instructor, they were interested in identifying the ``struggling students" in their course to provide additional support and materials and the ``high performers" in their course to provide advanced materials and recruit as student assistants for next year. To this end, they could get support from different \textit{information sources} (IS): two human teaching assistants (TA context) or two computational models (Model context). Below we describe each context in more detail. 

First, in the TA context, the participants were presented with two TAs who had been working with the students in smaller group settings on a weekly basis (like a discussion or lab session). We told the participants that we would be presenting a few cases where the TAs disagreed with each other about the student. 

% figure representing user study slide
\begin{figure}[t]
  \centering
  \includegraphics[width=\linewidth]{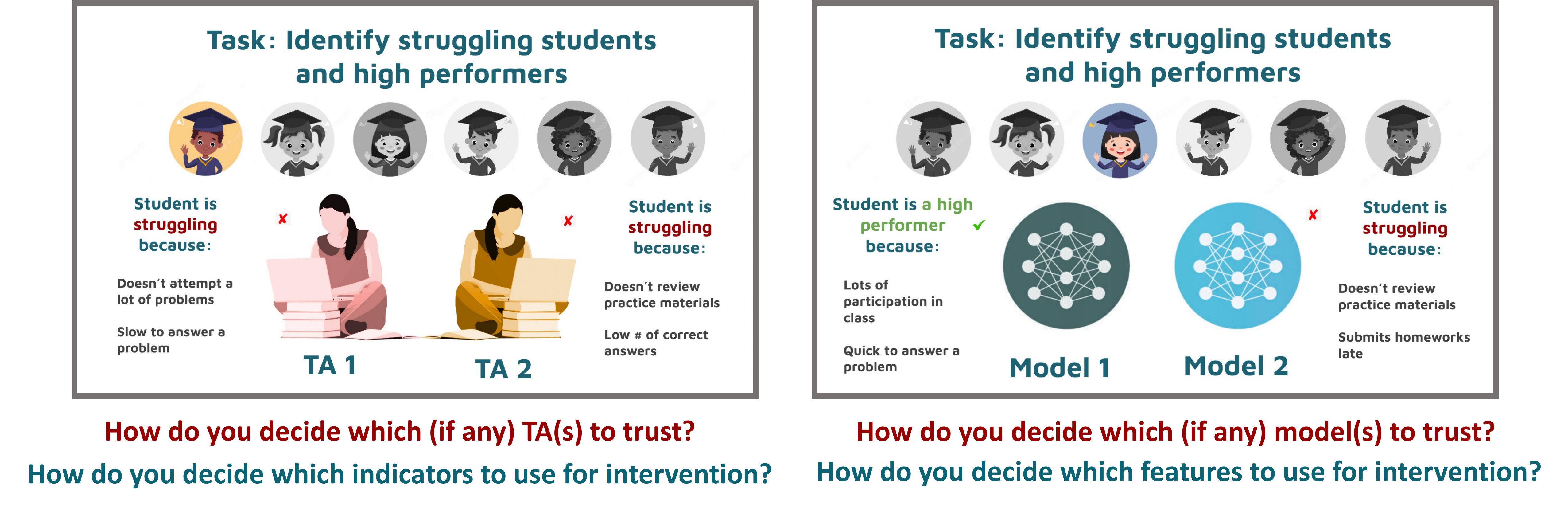}
  \caption{\label{fig:study} \textbf{Learning Science Expert Study}: Screenshots of student scenario $1$ for the TA context (left) and student scenario $3$ for the models context (right).}
  \vspace{-8mm}
\end{figure}

Second, in the model context, we presented a flipped classrom (FC) setting, where the instructor had access to two machine learning models. We notified the participants that these models were exposed to exactly the same information as the TAs about student behavior in the TA scenario above. In addition, we stated that both models had the exact same architecture, and they were also both precisely $96.65\%$ accurate. 

For each context, we presented scenarios of three students (Students 1-3). In the first scenario, two ISs (IS1 and IS2, who were both TAs or models depending on the context) predicted that Student 1 was struggling and needed support, however, they presented slightly different reasoning and explanations supporting the above prediction. We then repeated this process for two more student scenarios with increasing explanation disparity. The student 2 scenario highlighted a large difference in reasoning, with two information sources proposing completely different reasons for their predictions. In this scenario, IS2 provided more actionable explanations than IS1. The student 3 scenario had a difference in model prediction and in explanation, showcasing the largest inconsistency between IS1 and IS2. 

In total, each participant was presented with six situations (three student scenarios in two information sources contexts). 

Each time, after presenting the information source, the situation, and the proposed reasonings, we began by asking ``How believable is this scenario?" We then asked a question about trust: ``How do you decide which (if any) IS(s) to trust?" The participants could respond with one, both, or neither, and we aimed to understand the reasoning behind their choice. We then asked ``How do you decide which indicators to use for intervention?" With this question, we aimed to understand their logic for trusting certain aspects of the explanation and their perspective on actionability. The relevant slide for the TA context is highlighted on the left of Figure \ref{fig:study}, while the slide for the models context is shown in the right part of Figure \ref{fig:study}.

At the end of the six scenarios, we asked the question: ``What changes in your level of trust between the TAs and the models?" to get deeper insights into the difference in perceptions and trust between the scenarios.

%%%%%%%%%% Analysis
\vspace{1mm} \noindent \textbf{Analysis}. To evaluate the semi-structured interviews, we analyzed participants' choice answers (e.g., selecting the expert they trusted more, selecting the features to intervene on) using simple frequency counts and ratio. To analyze their answers to the last question on changes in trust between TAs and models, we developed the coding scheme illustrated in able \ref{tab:coding}. Our scheme was inspired by the scheme used in \cite{nazaretsky2022teachers}. Two annotators (authors of the paper) coded the answers and achieved a Cohen's Kappa of $0.88$.

\begin{table}[t]
\resizebox{\textwidth}{!}{
\begin{tabular}{lll}
\toprule
\textbf{Category} & \textbf{Short Description} & \textbf{Detailed Description} \\ 
\midrule
\multirow{10}{*}{} & \multirow{4}{*}{\textit{1 - TAs have human characteristics}} & TAs use human, personal connections with students: \\ 
& & 1) TAs understand social, emotional, and motivational factors important in teaching.\\
&& 2) TAs know students’ history outside the system.\\ 
&& 3) Models are more objective vs. TAs are more subjective. \\
\cmidrule{2-3} 
& \multirow{3}{*}{\textit{2 - TA experience}} & TAs have more experience: \\ 
\textbf{Accuracy}&& 1) TAs have real-life experience and human intuition. \\ 
\textbf{\& Reliability}&& 2) AI developers lack pedagogical and educational experience. \\ \cmidrule{2-3}  
& \textit{3 - Trust in TA from personal interactions} & Know TA on a personal level, informed of their strengths and weaknesses. \\ 
\cmidrule{2-3} 
& \textit{4 - Models are objective, TAs are not} & TAs are more likely to make mistakes than models — models are objective.\\ 
\cmidrule{2-3} 
& \textit{5 - Models do not have a complete picture} & Features passed to models do not cover a full holistic picture. \\
\midrule
\multirow{6}{*}{\textbf{Technical}} & \multirow{4}{*}{\textit{6 - Models can scale}} & Models can help more students than TAs: \\ 
& & 1) Models are more efficient (less time per student). \\
&& 2) Models are more effective (no errors due tiredness).\\
&& 3) Models are scalable (more students helped).\\ 
\cmidrule{2-3} 
& \multirow{2}{*}{\textit{7 - Models cannot give more detailed info}} & Models cannot give more information about a situation beyond  \\ 
&& what their training data allows, but TAs can. \\ 
\midrule  
\multirow{5}{*}{\textbf{Affective}} & \multirow{2}{*}{\textit{8 - TA disagreement is acceptable}} & Disagreement between TAs is considered a regular, valid situation. \\
& & When models disagree - something is wrong with the models. \\ 
\cmidrule{2-3} 
& \multirow{1}{*}{\textit{9 - TAs have empathy and compassion toward students}} & TAs can decide how not to negatively impact or overwhelm students.  \\ 
\cmidrule{2-3} 
& \multirow{2}{*}{\textit{10 - Models are more detailed than TAs}} &  Models have more detailed information than TAs. \\ 
&& This enables models to have more granular insights than TAs. \\ 
\midrule  
\end{tabular}}
\caption{\label{tab:coding} Coding scheme for learning scientists' perceptions of human TA vs. model reponses, based on the coding scheme of \cite{nazaretsky2022teachers}.}
\end{table}

%%%%%%%%%%%%%%%%%%%%%%%%%%%%%%%%%%%%%%%%%%%%%%%%%%%
%%% Results %%%%%%%%%%%%%%%
%%%%%%%%%%%%%%%%%%%%%%%%%%%%%%%%%%%%%%%%%%%%%%%%%%
\section{Experimental Evaluation}
To investigate how random model initialization affects model interpretability, we first explored expert perception of inconsistency of student explanations with a user study (RQ1). We then assessed the sources of inconsistency of counterfactual explanations across four MOOC courses (RQ2). We implemented a mitigate for counterfactual explanation inconsistency using ML methods in adversarial robustness with promising results (RQ3). We finally attempted to generalize our findings using two additional explainers (LIME and SHAP) and two additional flipped classroom courses (RQ4).

\vspace{1mm} \noindent \textbf{Experimental Protocol}. Due to class imbalance, for each model, we report the best balanced accuracy in the test data set over multiple training sessions. The train-test-validation split is an 80-10-10 ratio, ensuring a representative subset to evaluate the performance and generalization of the models. We use \textit{categorical cross entropy} as a loss function with \textit{Adam} as optimizer and learning rate 0.001. Hyperparameters (layer architecture, learning rate, adversarial training parameters) were tuned on the baseline BiLSTM for the \textit{DSP 1} course, and then kept standard among all courses to isolate the effect of randomness from variations in the model architecture. For adversarial training, PGD was conducted for 40 iterations with 0.05 epsilon and 0.01 alpha. FGSM was conducted with 0.05 epsilon. CEM was conducted for 100 instances chosen uniformly over model confidence from the test set. The hyperparameters used for CEM are 0.1 beta for the L1 loss term, 100 gamma for the autoencoder loss term, initial weight of the loss c-init 1, 10 c-step updates, 1000 max iterations per value of c, with gradient clipping (-1000, 1000), and learning rate 0.001. LIME, SHAP, and CEM were trained using the exact settings in \cite{swamy2022evaluating}. All replication and implementation details are presented directly in the provided code base at \url{https://github.com/epfl-ml4ed/robust-explainers}.

%%%%%% Inconsistent Explanations User study
\subsection{RQ1: Learning Science Expert Views of Explanation Inconsistency}
\label{sec:RQ3}
In the first part of our study across three granularities of student-teacher settings, we found that educators found consistency an important need across all three settings (3.66 / 5 for the one-on-one tutor setting, 4.53 / 5 for the middle school teacher setting, and 4.8 / 5 for the principal setting). In comparison with other explainability needs, averaged over all three settings, explanation consistency was viewed as 0.105 more important than other explainability needs by learning science experts.

\begin{figure}[t]
    \centering
    \includegraphics[width=\textwidth]{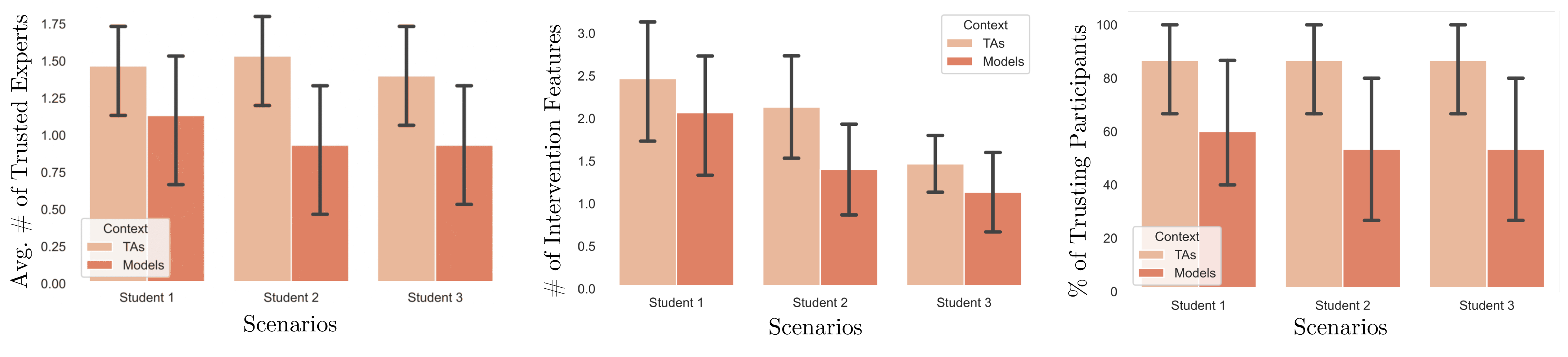}
    \caption{\textbf{Coded analysis of expert interviews}: Average number of IS (TAs or Models) found trustworthy (left), number of features selected for intervention (middle), and percentage of interviewees who trusted at least one IS (right).}
    \label{fig:experts}
\end{figure}

After understanding that explanation inconsistency is an important dimension, we followed with a study to measure expert perceptions of inconsistent explanations. We analyzed participants' trust in TAs and models when presented with three scenarios of increasing inconsistencies (small differences in explanation, large differences in explanation, and different predictions). Figure \ref{fig:experts} (left) shows the average number of TAs (models) trusted by the participants for each scenario. When presented with the first scenario, $15/15$ of the participants believed that two TAs could disagree with each other about the reasoning, but agree that a student was struggling. When faced with the same question in the models context, three participants could not imagine that models with the same architecture would agree on the prediction but not the reasoning. For the models context (dark orange bars), we note a decreasing trend across the scenarios -- as the explanation disparity gets larger, the participants trust the models less. There is a considerable drop off in trust for student 2, where the explanations from both models are considerably different, with one model providing two more actionable reasons. For the TA context, we note that the trust levels across all student scenarios remains the same. This implies that the study participants (learning science experts) are more accepting of disagreement in human TAs than in highly accurate models, even when presented with the exact same information and setting for both information sources (IS).

Next, we investigated the differences between features selected for intervention (illustrated in Fig. \ref{fig:experts} (middle)). We observe that for each student scenario, participants selected more features for the TA context than the models context, although the error bars overlap. This suggests that the participants trust the logic from human TAs more, even if provided the exact same logic from models. There is a decreasing trend across scenarios; as the explanations and predictions become more inconsistent, the features used for intervention also reduce. Out of four possible features, an average of more than two were selected to design an intervention for scenario 1, showing that features from more than one IS were chosen in the average case.  

Over all scenarios, the three most selected features for intervention for the TA information source were ``low number of correct answers", ``doesn't attempt a lot of problems``, and ``submits homeworks late``. The first two features (correct answers and attempting lots of problems) remained the most popular also for the model context, followed by ``doesn't review practice materials". The three least selected features for TAs were: ``quick to answer a problem", ``lots of participation in class" and ``doesn’t participate in group discussion". For the models, the three least selected features were the same, but there was a minor difference in ordering.

Figure \ref{fig:experts} (right) shows the percentage of participants that made an intervention based on features of the IS they trusted. For the TA context, approximately $80\%$ of the chosen intervention features were from the trusted expert. In the model context, participants sometimes trusted the model reasoning to conduct an intervention even though they did not specifically trust the expert.

For scenario 2, we designed IS 2 explanations to have directly actionable features and IS 1 explanations to be based on harder to act upon features. Only in four of $15$ cases, TA 2 was not chosen as a trustworthy expert, showing that participants generally preferred an expert with reasoning that leads directly to a designed intervention. Analogously, while only $16$ of $31$ chosen features to intervene on for student 2 came from TA 2, there were only four participants of $15$ that did not choose a feature from TA 2. Over five participants mentioned they prefer ``quantitative" features that can be externally confirmed in the TA and model scenarios for student 2.

%In Figure \ref{fig:user-trust}c, we measured the percentage of participants that make an intervention based on features of the expert they trust. In each of the TA scenarios, approximately 80\% of the chosen intervention features were from the trusted expert. In the model context, participants sometimes trusted the model reasoning to conduct an intervention even though they did not specifically trust the expert.

In a final analysis, we evaluated participants' answers to the last question of the interview based on our coding scheme (see Table \ref{tab:coding}).

For the \textit{Accuracy and Reliability} category, $70\%$ of the participants noted the human characteristics and connections of the TAs that are lacking in the models. $20\%$ of participants mentioned the real-life teaching experience of TAs that models do not have, and over $33\%$ of them mentioned that they would trust the TAs more based on their personal interactions with them. One participant said that \textit{``If one had more affinity with the TAs and knew how they worked, the trust level might change to higher if they were a good worker or lower if poor worker.''} Furthermore,  $36.67\%$ of participants state that models were objective and TAs subjective, but for those who mentioned this, they were evenly split on whether model objectivity is a positive aspect. A participant discussing model objectivity stated that \textit{``With the TAs one cannot expect much objectivity, and one will give, expect, and tolerate more error margin to humans than models. If there are some guarantees that the model is well-trained and consistent, one would trust the model more (but in return, expect models to be coherent in their predictions).''} Finally, $40\%$ of respondents (six participants) stated that models did not have a complete picture of student performance, although it was included in the study design that both TAs and models were exposed to the same information and interactions. On the other hand, $6\%$ of participants stated that models were more detailed than TAs (affective factor). 

In the \textit{Technical} category, $10\%$ of participants mentioned the scalability of models as a positive towards their effectiveness and trustworthiness. $16.67\%$ of participants stated the inability of models to give more detailed information about the student situation as a negative, in contrast to the way a TA might be able to answer follow-on questions. In terms of the \textit{Affective} category, several participants (13\%) stated that TA disagreement is more acceptable than model disagreement, confirming the results from Fig. \ref{fig:experts}. $30\%$ of participants mentioned that TAs had empathy and compassion towards students, as a positive attribute supporting human experts. One participant said the following on their (dis)trust of models:\textit{``For models, their training, development, limitations, and strong points are unknown. I would trust models more if there was more feedback from learning experts (teachers) on the model in action, and would trust more this kind of validation rather than a classic model validation.''}

\begin{graybox} In summary, our findings demonstrate that model and explanation consistency is a large concern for educators. While study participants (learning science experts) tolerated inconsistencies in explanations and predictions from TAs, they were much more sensitive to inconsistencies in model predictions and explanations. \end{graybox}

%%%% RQ1 - Analysis of variabilty
\subsection{RQ2: Sources of Explanation Inconsistency}
\label{sec:RQ1}
In a set of quantitative experiments, we aimed to capture the effects of random initializations and small perturbations in the input on the resulting explanations across four MOOC courses with counterfactual explanations. We systematically examined randomness at two stages in the explanation pipeline (see Fig. \ref{fig:pipeline}): (1) in the initialization of model weights at training and (2) post-hoc in the explanation extraction process.

First, we varied the random seed during model training and kept the rest of the pipeline exactly the same. For each course (\textit{Algebra Lineaire}, \textit{DSP 1}, \textit{DSP 2}, \textit{Geometry}, \textit{African Cities}), we trained five BiLSTM baseline models with different random seeds (we will refer to them as $BiLSTM_1$ through $BiLSTM_5$ in the following). The performances in terms of balanced accuracy of the five models for each course were very similar (as illustrated in the Baseline column of Table \ref{tab:BAC}). We obtained feature importance distributions (counterfactual explanations) for each of the five models per course from a representative $100$ students chosen uniformly per course as suggested in Chapter 4 \cite{swamy2022}. 

% Table for the accuracies of roubst versus unrobust models
\begin{table}[h]
% \begin{minipage}{.7\textwidth}
  \centering  
  \begin{tabular}{cccc}
    \toprule
    \textbf{Course} & \textbf{Baseline BiLSTM} & \textbf{FSGM (Robust)}& \textbf{PGD (Robust)} \\
    \midrule
    % Linear Algebra & 0.59\% $\pm$ \small 0.12 & 0.61\% $\pm$ \small 0.11 & 0.65\% $\pm$ \small 0.08\\
    DSP 1            & 0.98 $\pm$ \small 0.01 & 0.98\% $\pm$ \small 0.01 & 0.99\% $\pm$ \small 0.01\\
    DSP 2            & 0.98 $\pm$ \small 0.01 & 0.98\% $\pm$ \small 0.02 & 0.98\% $\pm$ \small 0.02\\
    Geometry       & 0.93 $\pm$ \small 0.06 & 0.94\% $\pm$ \small 0.02 & 0.96\% $\pm$ \small 0.03\\
    African Cities               & 0.96 $\pm$ \small 0.03 & 0.97\% $\pm$ \small 0.01 & 0.97\% $\pm$ \small 0.01\\
    \bottomrule
    \end{tabular}    
    \caption{Balanced accuracy averaged over five random seeds for each model type (Baseline BiLSTM, FGSM, and PGD) for each MOOC course. All $95\%$ of the different models overlap, indicating no difference in performance.}
    \label{tab:BAC}
% \end{minipage}
\end{table}

\begin{figure}[h]
  \centering
  \includegraphics[width=\textwidth]{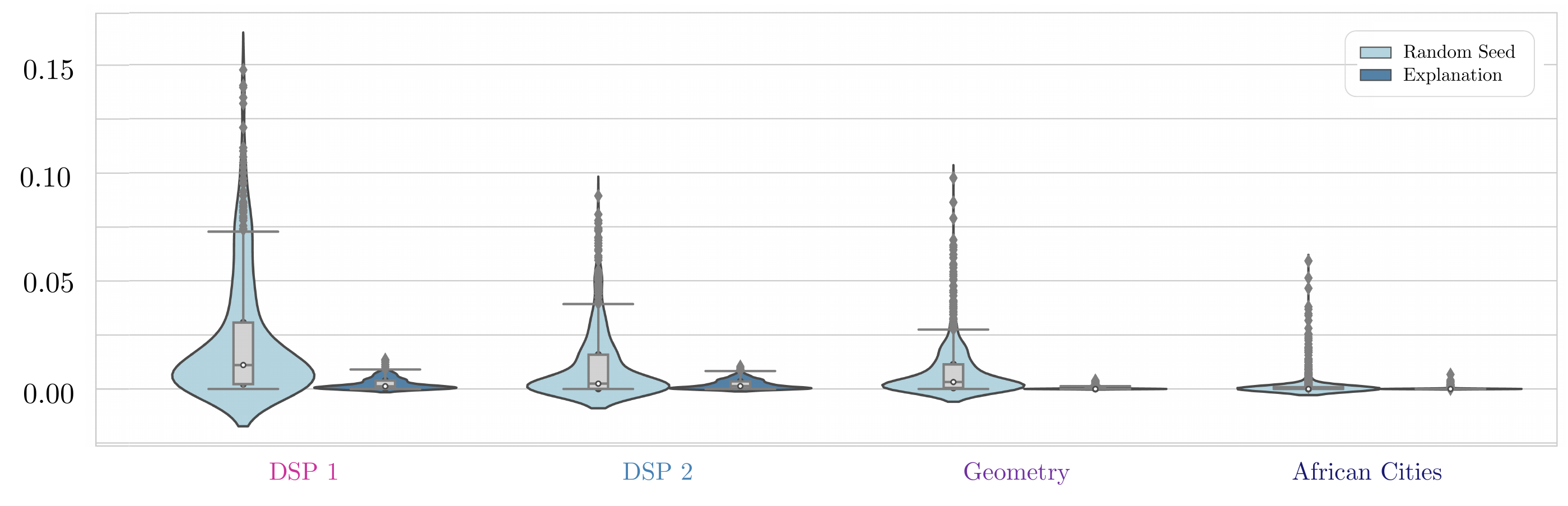}
  \caption{\label{fig:violins} Distribution of the standard deviations of feature importance scores across five models (light blue) and five explainers (dark blue) with different random seeds, obtained for four courses. The white dots denote the median.}
\end{figure}

Next, we kept all the model training details fixed and varied only the randomness at explanation extraction. We selected the best performing model out of the five models trained in the first step, based on the balanced accuracy on a hold-out test set. For this model, we extracted the counterfactual explanations for the same $100$ representative students per course five times (each with a different random seed for the explainer). Figure \ref{fig:violins} illustrates the distribution of standard deviations of feature important scores across five models (light blue, random seed for model) and five explainers (dark blue, random seed for explainer), respectively. We observe that using different random seeds for the model leads to large variations in feature importance scores. In contrast, the variation in feature importance scores over different random seeds for the explanation extraction is much smaller. For \textit{Geometry} and \textit{African Cities}, we even observe almost no variation when using different random seeds. 

\begin{figure}[t]
    \centering
    \includegraphics[width=\textwidth]{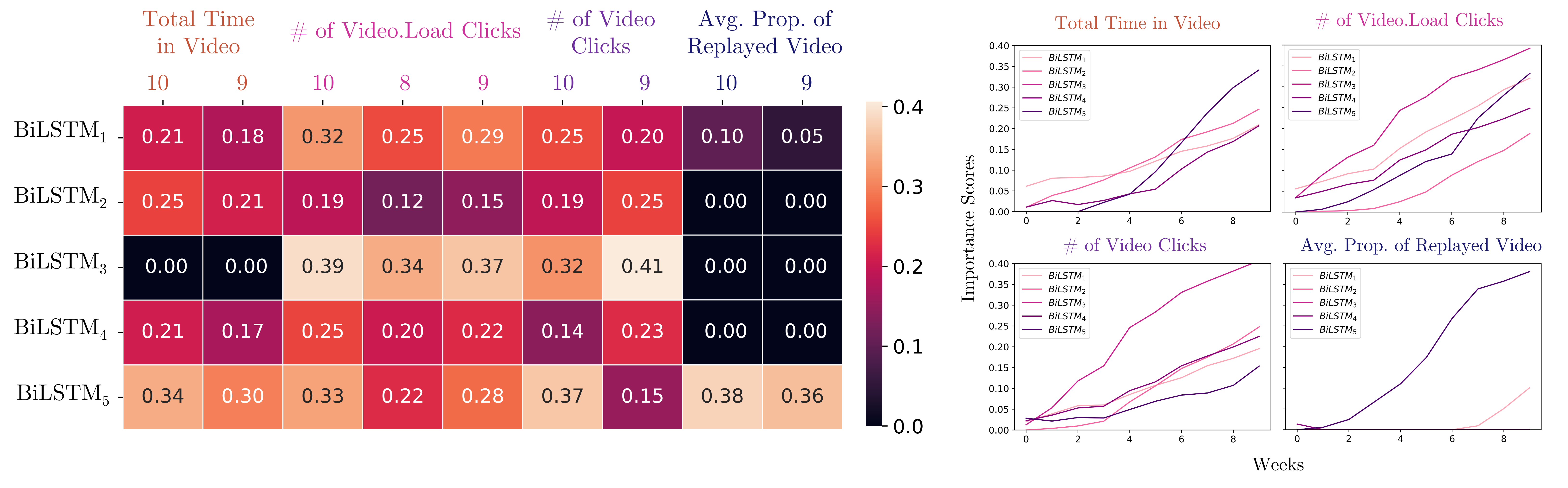}
    \caption{Feature important scores of the top features for five randomly initialized models on $DSP_1$: important scores for top feature-weeks (left) and for the same features for each week (right), showing temporal trends. The top features are the total time spent watching videos (\textit{Total Time in Video}), the number of times the video load button was clicked (\textit{\# of Video Load Clicks}), the total number of clicks related to videos (\textit{\# of Video Clicks}), and what proportion of a video was replayed on average (\textit{Avg. Prop. of Replayed Video}).}
    \vspace{-4mm}
    \label{fig:POLAR}
\end{figure}

To investigate the reasons for the high variation of feature importance scores upon model investigation further, we examined the features found to be important for $DSP_1$. Figure \ref{fig:POLAR} (left) shows the important scores of the features deemed most important for $BiLSTM_1$ through $BiLSTM_5$. We selected these features by choosing the top three feature-weeks (corresponding to a behavioral feature for a specific week) for each model. We observe that the importance scores per feature vary considerably across the different models (indicated by the varying colors in the heatmap). For example, \textit{Total Time in Video} in week 10, is deemed as no important by the $BiLSTM_3$ (score of $0.00$), but as very important by the $BiLSTM_5$ model (score of $0.34$). Furthermore, for the $BiLSTM_3$ model, feature importance scores tend to be either $0.00$ or $\geq 0.32$. In contrast, the feature importance scores for the $BiLSTM_3$ model are more equally distributed across the interval.

Figure \ref{fig:POLAR} (right) illustrates the temporal trends of the most important features across the five models ($BiLSTM_1$ through $BiLSTM_5$). We note that while all four features follow initial trends of rising importance over weeks, there are considerable differences between in trends for certain features. For example, the importance scores for the \textit{Number of Video Clicks} increase much faster when employing the $BiLSTM_3$ model rather than one of the other models. Moreover, \textit{Average Proportion of Replayed Video} is found important for only two models ($BiLSTM_5$ and to a small extent, $BiLSTM_1$), while it consistently obtains an importance score of $0.00$ over all weeks, when using one of the other models.

\begin{graybox} In summary, our findings demonstrate that the randomness of weights in model initialization has significant impact on explanation variance, while the randomness in the counterfactual explanation extraction only causes minor variation. Furthermore, when investigating the feature importance scores for a specific course, we found that random model initialization leads to considerable different importance scores across features and potentially varying temporal trends. \end{graybox}

%%% RQ2 - Impact of robust models
\subsection{RQ3: Adversarial Robustness to Mitigate Counterfactual Inconsistency}
\label{sec:RQ2}
In a second set of experiments, we explored the implications of training adversarial robust models with the goal of creating higher quality (more consistent) explanations.

Table \ref{tab:BAC} lists the performance of our robust models (FGSM and PGD) in terms of balanced accuracy in comparison to the baseline BiLSTM. For all three approaches, the results are averaged over five random model seeds. Despite balanced accuracy scores being exceptionally high for the majority of courses, robust models maintain the same performance as the baseline models, often with a lower variance. This suggests that the adversarial training methods help enhance a model's ability to handle perturbations and generalize better to unseen data. It is however important to note that these robust models tend to overfit when trained with a small number of training points.

\begin{figure}[t]
    \centering
    \includegraphics[trim={5 5 5 5},clip,width=0.7\textwidth]{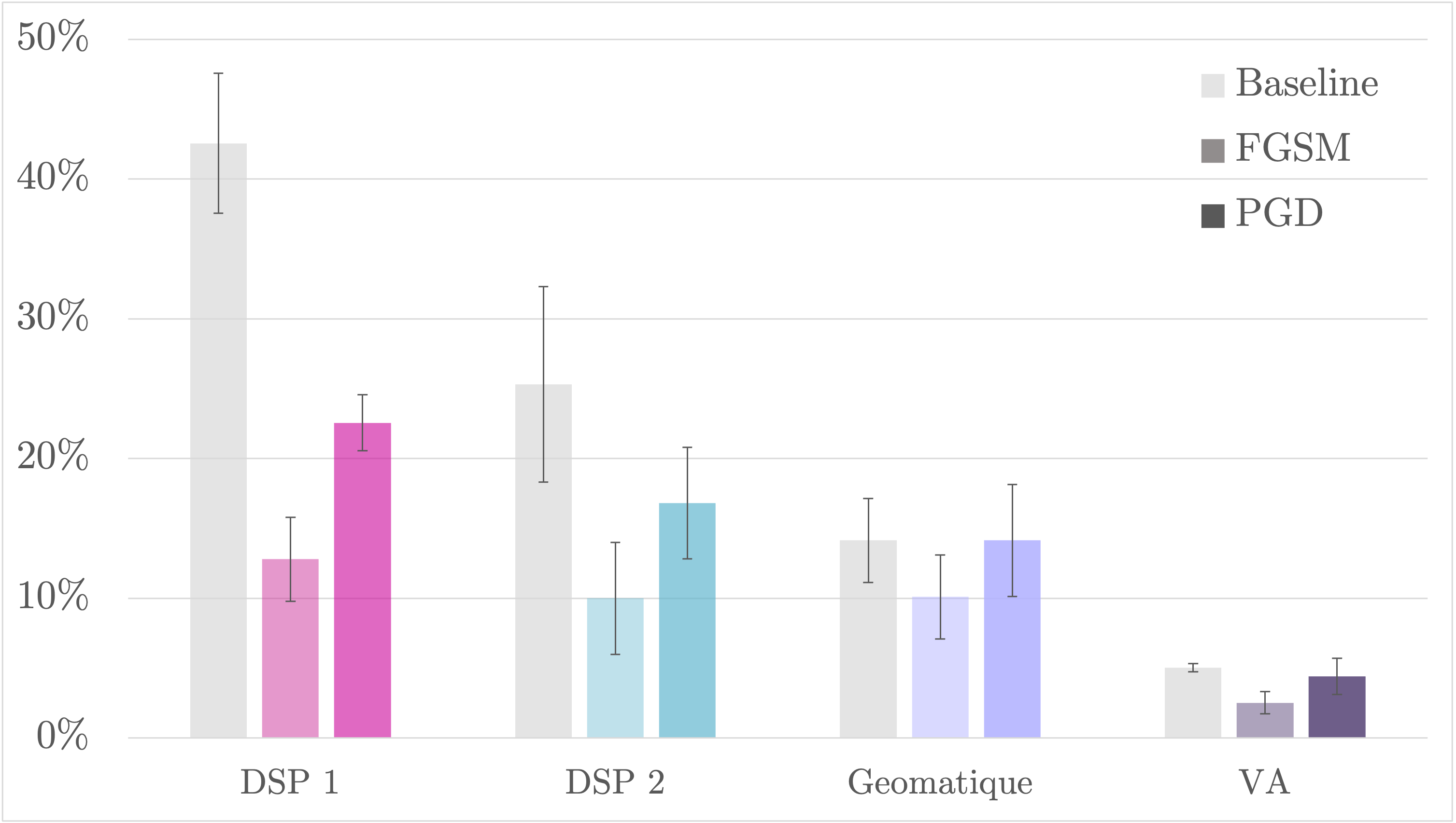}
    \caption{Percentage of features found important by CEM over three approaches (baseline BiLSTM, FGSM Robust, and PGD Robust), each of them trained with five different seeds. In four out of five cases, the baseline approach finds substantially more features important.}
    \label{fig:feature_importance}
\end{figure}

Having verified that robust models maintain the performance, we investigated the variance of explanations for robust models. For each course, we trained five models with different random seeds using each of the three approaches (Baseline BiLSTM, FGSM, PGD). We then employed CEM to extract counterfactual explanations for each model. Figure \ref{fig:feature_importance} shows the number of features selected as important (feature important score $>0.01$) by robust and baseline approaches. We observe that the baseline consistently selects more features as important than the robust approaches. For example, in the case of \textit{DSP 1}, the FGSM and PGD models deem $5\%$ of the features to be important, while the Baseline model selects $18\%$. An important downstream application of XAI is for educators to use important features to design an intervention for their struggling student(s). Having many features with low importance as opposed to few features identified as important might cause a less targeted and therefore less effective intervention \cite{joseph2008best, hirsch2013helping}.

\begin{figure}[t]
    \centering
            \includegraphics[width=\textwidth]{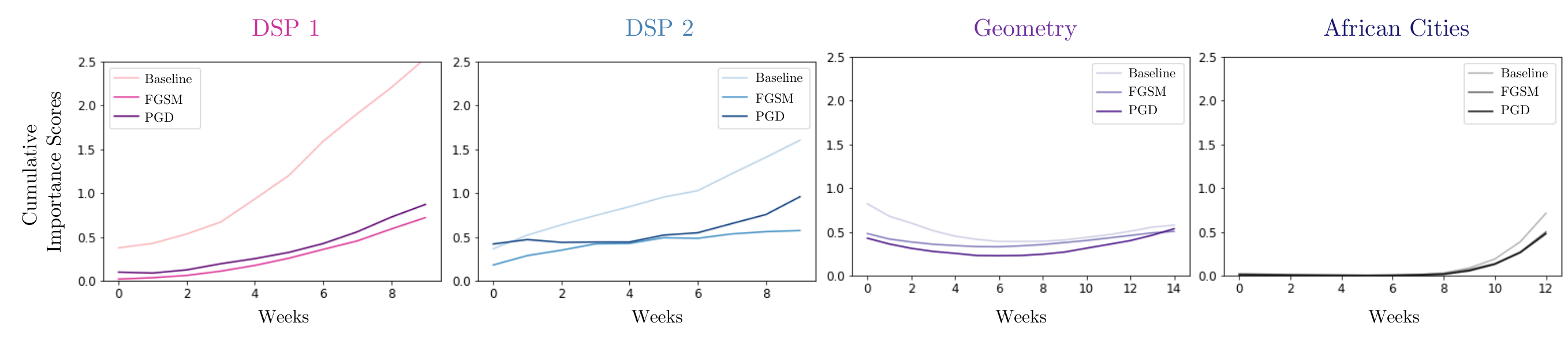}
    \caption{Temporal analysis of the cumulative average importance scores for four courses (\textit{DSP 1, DSP 2, Geometry, Villes Africaines}) and across three model variations (\textit{Baseline, FGSM, PGD}), each trained with five different random seeds. Only features with importance $\geq 0.01$ are included in the analysis.}
    \label{fig:cumulative_trend}
\end{figure}

Figure \ref{fig:cumulative_trend} further explores the temporal trends of explanations by creating cumulative sums (over all features and random seeds) of importance scores per week for the three model types. We eliminated features with an average importance score $\le 0.01$. For \textit{DSP 2},  \textit{Geometry}, and \textit{African Cities}, we observe that the robust models (dark lines) follow the same trends as the baseline (lighter lines). For \textit{DSP 1}, the baseline shows a much steeper increase of importance scores over time than the robust model. In the \textit{Linear Algebra} course, we note that the robust models find earlier weeks more important than the baseline models.

%%% Temporal cumulate feature analyses - rq2

% \begin{table}[]
% \centering
% \begin{tabular}{ccccc}
% \hline
% \textbf{Feature} & \textbf{Week} & \textbf{Baseline} & \textbf{FSGM-robust} & \textbf{PGD-robust}\\
% \hline
% speed\_playback\_\_std     &   10   &  $0.02 \pm 0.14$   &  $0.11 \pm 0.13$   & $0.39 \pm 0.43$\\
% speed\_playback\_\_std     &   9    &  $0.02 \pm 0.14$   &  $0.08 \pm 0.11$   & $0.38 \pm 0.42$\\
% speed\_playback\_\_mean    &   10   &  $0.02 \pm 0.14$   &  $0.09 \pm 0.11$   & $0.45 \pm 0.47$\\
% speed\_playback\_\_mean    &   9    &  $0.02 \pm 0.14$   &  $0.07 \pm 0.10$   & $0.42 \pm 0.45$\\
% speed\_playback\_\_mean    &   8    &  $0.02 \pm 0.14$   &  $0.04 \pm 0.07$   & $0.38 \pm 0.42$\\
% time\_sessions\_sum        &   10   &  $0.07 \pm 0.14$   &  $0.07 \pm 0.09$  & $0.20 \pm 0.25$\\
% time\_in\_\_video\_sum     &   10   &  $0.25 \pm 0.30$   &  $0.01 \pm 0.02$   & $0.00 \pm 0.00$\\
% total\_clicks\_video       &   10   &  $0.24 \pm 0.29$   &  $0.02 \pm 0.02$  & $0.00 \pm 0.00$\\
% total\_clicks\_Video.Load  &   10   &  $0.22 \pm 0.27$   &  $0.03 \pm 0.03$   & $0.02 \pm 0.10$\\
% \hline
% \end{tabular}
% \caption{Average importance scores and standard deviations of the most important features for \textit{DSP 1} in five runs of CEM.}
% \label{tab:DSP_IMPORTANCES}
% \end{table}

\begin{figure}[t]
  \centering
  \includegraphics[width=\textwidth]{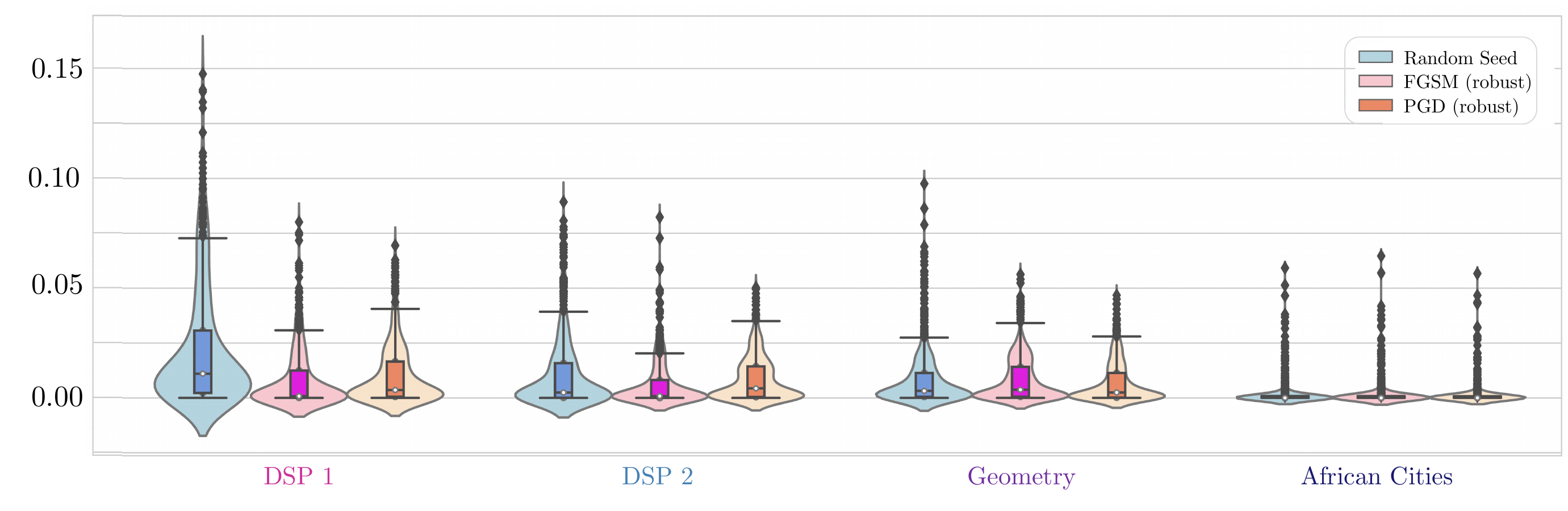}
  \caption{Distribution of the standard deviations of feature importance scores across five models with random seeds for the baseline (light blue), FGSM (rose), and PGD (orange), obtained for four courses. The white dots denote the median.}
  \label{fig:variance}
\end{figure}

Finally, we also analyzed again the distribution of the standard deviations of feature importance scores across the three approaches (see Fig. \ref{fig:variance}). We note that for some courses, e.g. \textit{DSP 1} and \textit{Geometry}, the robust models have significantly less variance than the baseline approach. This is beneficial in downstream applications as it leads to more consistent explanations; low variance implies that multiple generated explanations pick the same features with the same levels of importance. For \textit{Linear Algebra}, FGSM has a smaller variance than the baseline BiLSTM, while PGD shows a similar variance. For $DSP 2$, PGD exhibits a smaller variance than the baseline approach, while FGSM has a similar variance. For \textit{African Cities}, all methods have similar variance. 

\begin{graybox} In summary, we found that robust methods (FGSM and PGD) maintain model performance while improving explanation conciseness and consistency. Our robust methods select a lower number of features as important and decrease the variance of feature importance scores across different random seeds. \end{graybox}

\subsection{RQ4: Generalization of Findings}
\label{sec:RQ4}

In a third set of experiments, we explore the generalization of our findings with adversarial training mitigation strategies for counterfactual explainers. We first extend our experiments to another educational setting (flipped classrooms), which have different patterns of student behaviors \cite{hardebolle2022gender}, and much smaller datasets than in the MOOC settings. More details on the learning context for the flipped and MOOC classrooms were discussed in Section \ref{sec:learning context}. Secondly, we extend the analysis to two other additional popular post-hoc explainers (LIME and SHAP) \cite{lime,shap}. These explainers were used in the explainer evaluations in Chapter 4 and 5, and have commonly been used in a plethora of educational explainability papers \cite{hasib-lime, shap-ex}. We examine two analyses: one for the sources of inconsistencies, and the other for the effect of mitigation strategies.

\begin{figure}[!ht]
  \centering
  \includegraphics[width=0.7\textwidth]{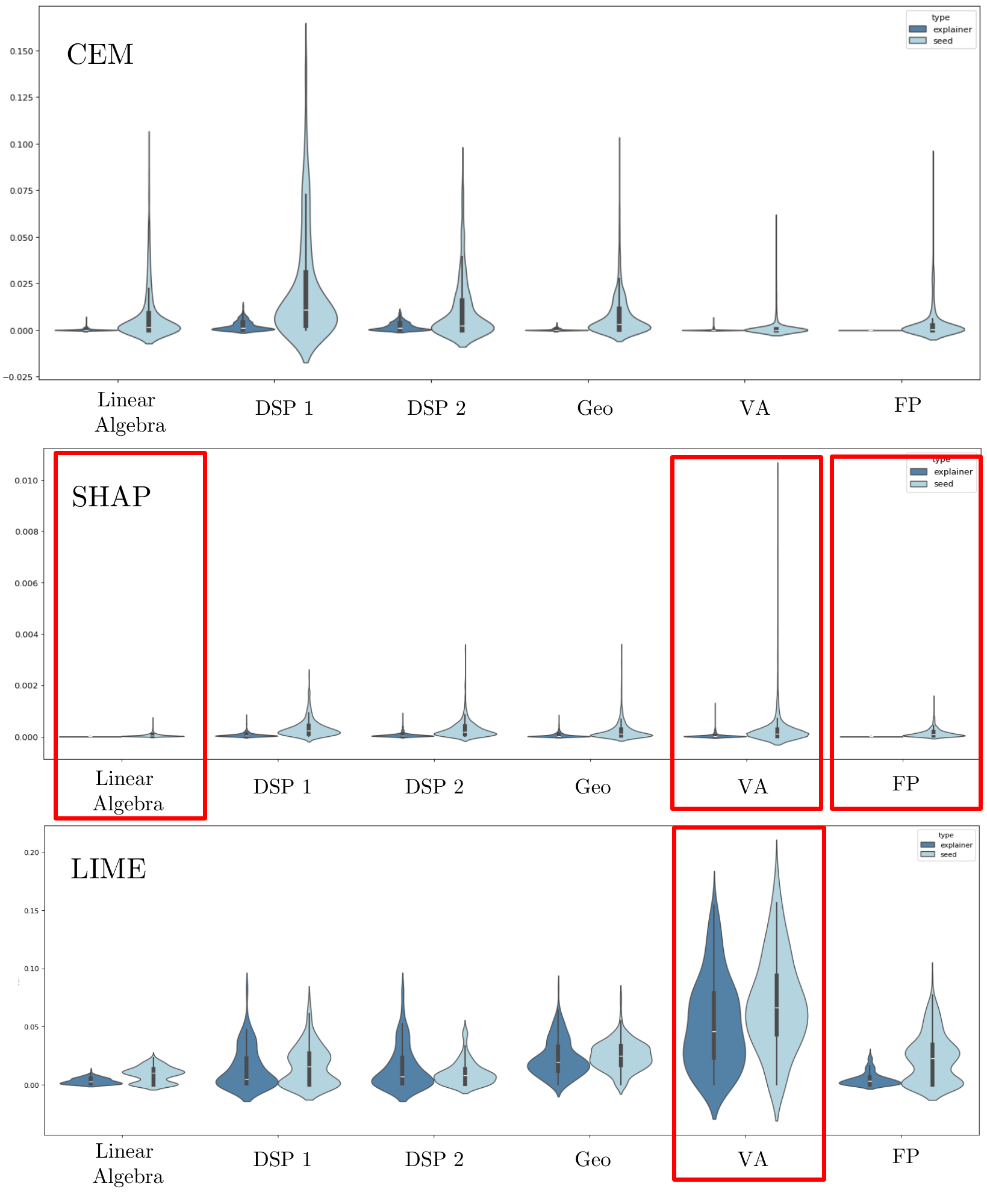}
  \caption{\textbf{Sources of Explainer Inconsistency}, measured by distributions of the std. deviations of importance scores, across CEM, LIME, and SHAP explainers for 2 flipped classroom courses (Linear Algebra and FP) and 4 MOOC courses.}
  \label{fig:settings}
\end{figure}

\textbf{Sources of Inconsistency.} In the first 5 plots in Figure \ref{fig:settings}, we build on our experiments from RQ2 and RQ3 using CEM counterfactual explainers to two new flipped classroom courses (Linear Algebra and Functional Programming). We train BiLSTM models for student success prediction in the same exact protocol as the previous explorations, receiving overlapping 95\% CIs in BAC scores from both FGSM and PGD with the baselines reported in Chapter 3 \cite{swamy2022meta}. We observe that for counterfactuals, explanation inconsistency stems significantly more from the random seed than the explainer, which is consistent with the results from the previous RQs. 

We then examine the results from SHAP and LIME across six courses. For SHAP explanation variance in the flipped classroom courses, the explanation variance is minimized in both random seed and explainer settings, likely due to the low number of data points. However, the results for Linear Algebra and FP confirm that the trend of variance due to random seed is prevalent. For LIME, we notice much more explanation variance across all courses, which can be attributed to the limited number of features (10) chosen in the default settings of the explainer. This contributes to higher differences in features chosen per student. The VA course has a notably high variance due to random seed using SHAP, and in both explainer and random seed using LIME. We hypothesize that the student populations of the VA course interact with the course in varied ways, leading to explanation outliers and therefore higher variance. \textbf{While sources of explainer inconsistency is attributed to random seed for CEM, both randomness in explainers and random model initialization cause a lot of variance for LIME and SHAP.}

\begin{figure}[!ht]
  \centering
  \includegraphics[width=\textwidth]{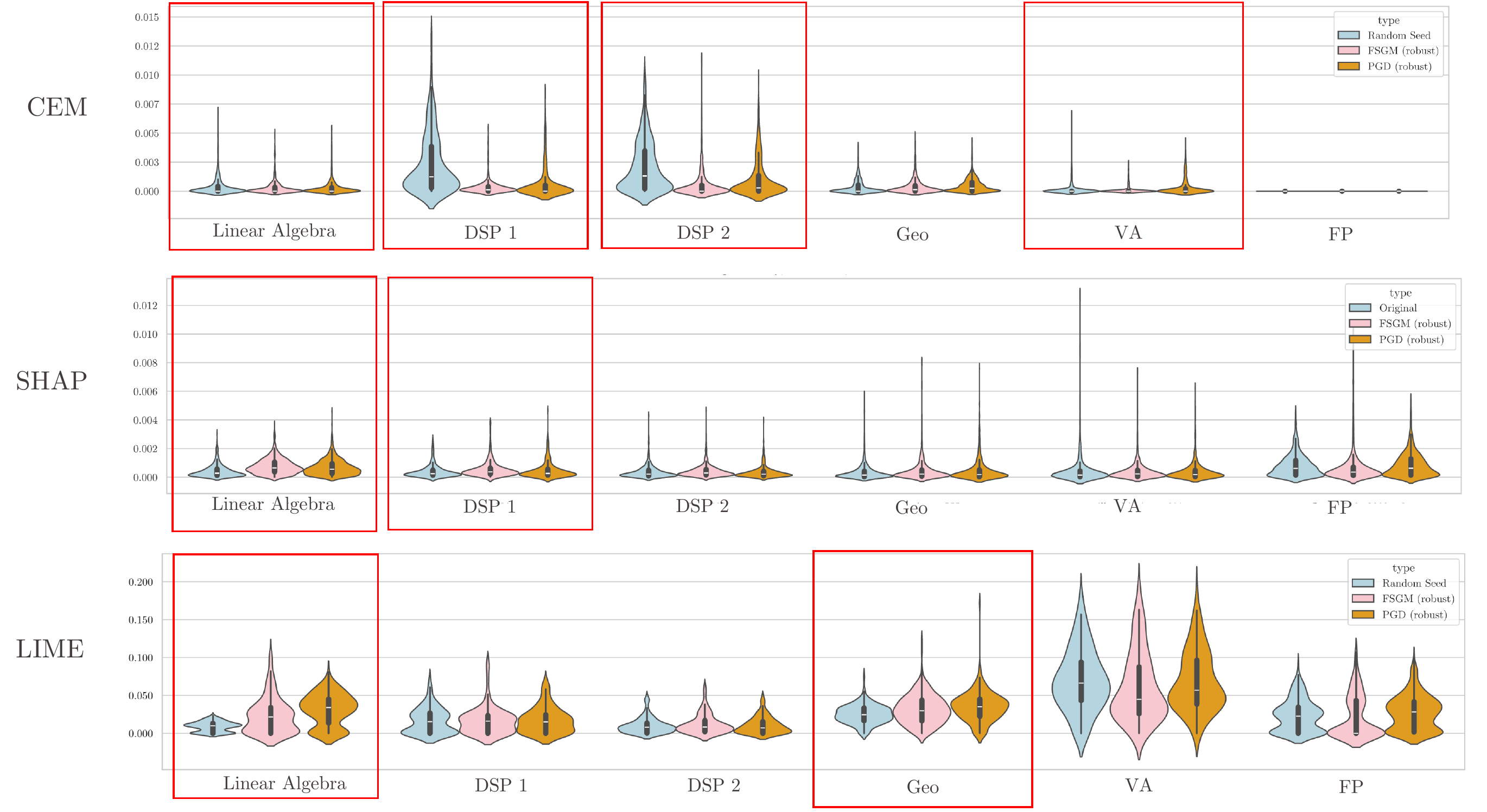}
  \caption{\textbf{Explanation Variability across FGSM, PGD and Baseline methods} measured by distributions of the std. deviations of importance scores, across CEM, LIME, and SHAP explainers for 2 flipped classroom courses (Linear Algebra and FP) and 4 MOOC courses.}
  \label{fig:analysis}
\end{figure}

\textbf{Mitigation Strategies.}  We extend to our analysis to the variance of importance scores with FGSM and PGD across all three explainers, as shown in Figure \ref{fig:analysis}. For CEM, we see significant improvement in explainer variation due to FGSM and PGD across several courses, or otherwise no impact. However, in both the LIME and SHAP setting, we note that for the Linear Algebra course, FGSM and PGD contribute to an increase in explanation variability instead of a reduction. This finding is also echoed (to a lesser extent) in the SHAP explanations for DSP 1, and in the LIME explanations for Geomatique. In all other settings for LIME and SHAP, it is not clear that FGSM or PGD help improve explanation variability. \textbf{Therefore, we can conclude that the mitigation effects with FGSM and PGD do not generalize from CEM to other explainers.}

\begin{graybox}{In summary, we found that the improvements to explainer consistency enabled by robust methods (FGSM and PGD) do not generalize to different methods of importance approximation like SHAP and LIME, or in smaller data settings, like flipped classrooms. }\end{graybox}

% Table for the accuracies of roubst versus unrobust models
\begin{table}[h]
% \begin{minipage}{.7\textwidth}
  \centering  
  \begin{tabular}{cccc}
    \toprule
    \textbf{Course} & \textbf{Baseline BiLSTM} & \textbf{FSGM (Robust)}& \textbf{PGD (Robust)} \\
    \midrule
    Linear Algebra & 0.67 $\pm$ \small 0.13 & 0.63 $\pm$ \small 0.10 & 0.65 $\pm$ \small 0.09\\
    
    Functional Programming               & 0.60 $\pm$ \small 0.09 & 0.54 $\pm$ \small 0.05 & 0.61 $\pm$ \small 0.08\\
    \bottomrule
    \end{tabular}    
    \caption{Balanced accuracy averaged over five random seeds for each model type (Baseline BiLSTM, FGSM, and PGD) for each flipped classroom course.}
    \label{tab:BAC}
% \end{minipage}
\end{table}

\section{Conclusion}
In this chapter, we conducted four lines of inquiry into explanation variance and present the following main conclusions:
\begin{enumerate}
    \item \textbf{Learning science experts are bothered by the inconsistency of explanations in models, not in humans.} With learning science experts from around the globe, we evaluate the concerns of increasing levels of explanation disagreement and find that given the exact same reasonings and scenario for models and TAs, the TAs are trusted significantly more. To address this gap in trust, two steps are necessary: current methods need increased focus on explanation consistency and systematic validation of models with human experts beyond quantitative analyses.
    \item \textbf{Randomness in the model and explainer initialization process has a large impact on the resulting explanation.} Two student success predictions models with the same accuracy, trained on the same data, with the same architecture, and the exact same explanation extraction strategy can have vastly different explanations due to setting different random seeds.
    \item \textbf{Adversarial robustness techniques can help mitigate this randomness towards more concise, consistent explanations.} We experiment with FGSM and PGD to extract counterfactual explanations for 100 students for five iterations each on five different courses to measure the effects of robustness, and find them generally positive.
    \item \textbf{Mitigation with adversarial robustness does not generalize.} We note that when extending our mitigation evaluation with counterfactual explanations to smaller, flipped classroom courses or other post-hoc explainers like LIME or SHAP, adversarial robustness does not contribute as strongly to explanation consistency.
\end{enumerate}

In discussion with learning science experts, we observed diverse opinions on the inconsistency of models in educational settings. Some participants expressed that any level of model disagreement rendered the predictions and reasoning untrustworthy, making them unsuitable as a basis for downstream interventions \cite{dietvorst2015algorithm}. For others, the objectivity, scalability, and high accuracy of models provided significant advantages—up to the point where disagreements became too severe. However, a key finding was that, in identical scenarios, teaching assistants (TAs) were consistently regarded as more trustworthy than highly performant models. This trust disparity highlights the pressing need to improve educator perceptions of AI \cite{nazaretsky2022teachers}.

Interestingly, these sentiments align with the well-documented phenomenon of algorithm aversion \cite{Dietvorst2018}, where individuals tend to distrust or reject algorithmic advice, particularly after observing an algorithm make errors—even when the algorithm outperforms humans on average. Our findings similarly show that, despite strong performance metrics, models lost trust when their outputs appeared inconsistent, especially in comparison to human alternatives. This observed preference for human judgment over machine predictions underscores the importance of addressing educators' concerns and bridging the trust gap \cite{poursabzi2021manipulating,viberg2024explains}.

In a broader context, disagreements between information sources—whether between human TAs or between models—need not always be viewed as errors or inconsistencies. Instead, they can reflect the intrinsic complexity and nuance of real-world pedagogical scenarios \cite{luckin2019designing, dietvorst2015algorithm}. Just as two educators may evaluate a student's performance or interpret teaching material differently without invalidating each other’s perspectives, model disagreements might similarly arise from differing but valid interpretations of the same data. Such variability underscores the depth and multifaceted nature of educational tasks \cite{woolf2010building}.

For example, contrasting conclusions drawn by two educators on the same student data are often accepted as a natural outcome of their unique experiences and perspectives. This variability does not diminish trust in human educators; rather, it highlights the inherent complexity of pedagogical interpretation. Similarly, the presence of differing model outputs should not lead to the immediate dismissal of their utility. Instead, educators should be introduced to the concept of expected model accuracy for specific tasks, contextualized within the levels of disagreement commonly seen among humans for comparable tasks \cite{shneiderman2020human}. This comparison can serve as an important benchmark for understanding and accepting the model's outputs.

Encouraging educators to view disagreements as reflections of the inherent complexity of real-world tasks, rather than as errors, can foster a more nuanced understanding of both human and machine judgments \cite{luckin2019designing}. We urge other researchers to replicate our study findings and mitigation strategies, especially in human-centric fields, before using post-hoc explainers directly in downstream applications.

%% file: main/08_illuminate.tex
The work presented in this chapter is published at AAAI 2025 in collaboration with Davide Romano, Bhargav Srinivasa Desikan, Oana-Maria Camburu, and Professor Tanja Käser across EPFL, the Institute for Public Policy Research, and UCL. The provisional title of the article is \textit{iLLuMinaTE: An LLM-XAI Framework Leveraging Social Science Explanation Theories towards Actionable Student Performance Feedback}. We would like to thank our study participants for their engagement in our study, and Tanya Nazaretsky for insightful study design discussions.

The LLM-XAI pipeline presented in this chapter (ILLuMinaTE) directly targets improving \textbf{human understandability} and \textbf{actionability} of explanations, towards scalable and personalized student feedback. 

\section{Introduction}

\begin{figure*}[!ht]
    \centering
    \includegraphics[width=\linewidth]{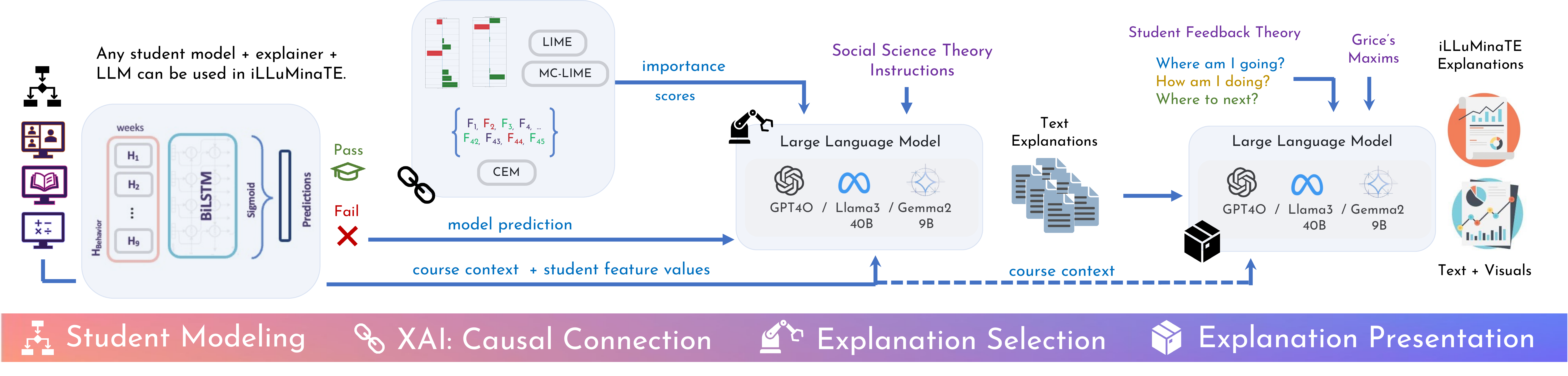}
    \caption{\textbf{\il involves four steps}: 1) modeling course data for student success prediction, 2) using XAI methods to extract feature importance scores, 3) selecting important aspects of the explanation through an LLM aligned with a given social science theory, 4) presenting the explanation with concise and actionable suggestions through an LLM.}
    \label{fig:illum_pipeline}
\end{figure*}

This chapter focuses on a major weakness of XAI: a lack of adequate human understandability, especially for a non-technical audience. In Chapter 4, at a course level, STEM professors expressed difficulty in understanding explainer outputs, requesting ``more concrete and granular insights'' on the scale of individual students \cite{swamy2023trusting}. At the individual student level, \cite{hur2022using} designed XAI-based interventions, but found they were extensive to integrate and provided limited learning gains compared to expert feedback. With pervasive educational benchmarks like ASSISTments \cite{heffernan2014assistments} and MOOCRadar \cite{yu2023moocradar} having hundreds of features and temporal aspects, it becomes difficult for students and teachers to interpret feature importance. 

LLMs can be useful in making XAI more human-interpretable, especially towards building stakeholder trust in AI and designing personalized student interventions. A recent study \cite{kroeger2024large} suggests that LLMs can act as post-hoc explainers for complex models, finding that LLMs could identify relevant features when given examples of input data and model outputs. \cite{atanasova2022diagnostics} integrate explanation generation directly into the LLM’s training, optimizing over diagnostic properties like data consistency and confidence. However, in domains where explanations directly influence human decisions, the nature of LLMs as ``stochastic parrots'' can be accompanied by detrimental side effects \cite{bender2021dangers,sarkar2024large}. LLMs, while demonstrating potential in areas such as knowledge tracing or student synthesis, have not yet matured enough to act as student models that can be accompanied by inherent explanations \cite{neshaei2024towards,nguyen2023large}. We instead propose to use LLMs as \textit{communicators of explanations} \cite{zytek2024llms} to present XAI outputs in aligned text and visual formats that are \textit{actionable} for educational stakeholders. 

We therefore present \il, an in-context, chain-of-prompts, zero-shot LLM pipeline that is inspired by Miller's cognitive processes of explanation \cite{miller2019explanation}. \il follows three main stages: (1) causal connection, (2) explanation selection, and (3) explanation presentation. Our experiments range over eight prevalent social science theories of explanation \cite{hilton1990conversational,hilton1986knowledge, halpern2005causes,lombrozo2010causal,sep-scientific-explanation}, with three underlying explainers (LIME, Counterfactuals, MC-LIME), data from three online courses, evaluated using three LLMs (GPT-4o, Gemma2 9b, Llama3 70b) and a real-world user study with $114$ university students. Notably, we find that students preferred \il's explanations over baseline hybrid (text and visual) explanations from post-hoc methods $89.52\%$ of the time, and had a particular preference on actionability for abnormal, pearl, and contrastive explanations. With our study, we make the following main contributions:

\begin{enumerate}
    \item \textbf{\il, a chain-of-prompts framework in the education context} to extract theory-driven natural language explanations (NLE) for student feedback.
    \item \textbf{An LLM-XAI efficacy analysis} of $216$ variations of \il prompting strategies over explainers, LLMs, social science theories, and student populations.
    \item \textbf{A real-world evaluation of LLM-XAI preferences} conducted with $114$ university students. 
    \item \textbf{An XAI actionability study} simulating student performance gains based on actions they selected from generated explanations.
\end{enumerate}

We provide our modular implementation of \il publicly with adaptations for LangChain, Groq Cloud, and Replicate\footnote{\url{https://github.com/epfl-ml4ed/iLLuMinaTE}}. Our work provides a theory-driven methodology to communicate results of XAI to students, with broad generalization potential of explanation theory instruction prompts for other human-centric fields (e.g., healthcare, welfare, product recommendation).

% Methods
\section{\il Methodology}
\label{sec:methods}
Our \il pipeline (see Fig. \ref{fig:pipeline}) consists of four stages towards communicating explanations in a human-understandable and actionable way through LLMs.

In the \textit{Student Modeling} phase, we extract behavioral features from raw clickstreams of student interactions and use BiLSTMs \cite{graves2005framewise} to predict student success following prior work \cite{asadi2023ripple,swamy2022meta}. We then employ post-hoc explainers to obtain feature importance scores, representing the \textit{XAI: Causal Connection} step. With the results from an explainer, the course context, and the student's feature values, we prompt an LLM using \textit{Explanation Selection} instructions specific to social science theories of explanations. We evaluate the obtained explanations using human expert and GPT-4o annotations. We then use \textit{Explanation Presentation} prompts to summarize the often verbose explanation selection reports into concise and actionable feedback for a student, taking into account theory on effective feedback \cite{hattie2007power}'s and maxims for communication \cite{grice1975logic}. We then evaluate the final explanations using expert annotations. We assess \textit{Student Preferences} and the actionability of \il explanations in comparison with the current state-of-the-art in XAI for education in a user study with $114$ students.

\subsection{Student Modeling}
\label{sec:student-modeling}
To create the models building the basis for the explanations, we use the same features and model architectures as prior research working with the same datasets \cite{swamy2022meta,galici2023not,swamy2023trusting,swamy2024multimodn}.

\vspace{1mm} \noindent \textbf{Data Collection}.
Our experiments are based on data collected from three MOOCs (Digital Signal Processing (DSP), Villes Africaines (VA), and Elements de Geomatique (Geo)) offered by a European University to a global student audience. The courses were organized into weekly modules including video lectures and quizzes and required students to complete graded assignments to earn course certificates. Students interacted with learning objects (videos, quizzes) associated with specific course weeks, enabling the creation of course-specific learning indicators. We represent a student's interations as a time series $I_s^c = {i_1, \ldots, i_K}$, where each interaction $i$ is a tuple $(t, a, o)$, including a \emph{timestamp} $t$, an \emph{action} $a$ (e.g., video play, pause; quiz submission), and a \emph{learning object} $o$. The \emph{binary success label} (pass-fail) for student $s$ in course $c$ is denoted as $y_{s,c}$. Data collection and analysis were approved by the university’s ethics review board (HREC 058-2020/10.09.2020).

\vspace{1mm} \noindent \textbf{Feature Extraction.} We use a broad set of $45$ behavioral features \(h\) derived from the student interactions, incorporating features from four feature sets shown to be predictive for student performance in MOOCs \cite{marras2021can}. \textbf{Regularity} features (3) capture consistent study habits \cite{boroujeni2016quantify}, \textbf{Engagement} features (13) measure course involvement \cite{chen2020utilizing}, \textbf{Control} features (22) analyze video usage \cite{lalle2020data}, and \textbf{Participation} features (7) track attendance in scheduled activities \cite{marras2021can}.

\vspace{1mm} \noindent \textbf{Modeling}. For a course \(c\) and student \(s\), our objective is to build a model that predicts \(y_{s,c}\) \textit{early}, using the features \(h_s\) from the first five weeks. Following prior work \cite{swamy2022meta}, we employ a BiLSTM for this task. We provide all reproducibility details in Appendix \ref{appendix:reproducibility}.

\vspace{-2mm}
\subsection{XAI: Causal Connection}
\label{sec:causal-connection}

We use three popular post-hoc explainers (LIME, CEM, and MC-LIME) to extract local, instance-specific explanations from the student models. We chose these three methods based on their popularity, but any in-hoc or interpretable-by-design model could be used.

\vspace{1mm} \noindent \textbf{LIME} (Local Interpretable Model-Agnostic Explanations) \cite{ribeiro2016why} provides interpretable explanations for individual predictions by approximating the complex model locally with an interpretable model. This process results in a set of feature weights indicating the positive or negative influence of each feature on the prediction.

\vspace{1mm} \noindent \textbf{CEM} (Contrastive Explanation Method) \cite{dhurandhar2018explanations} is a counterfactual method that identifies the features that need to be present (pertinent positives) or absent (pertinent negatives) for a model to maintain its prediction for a given instance. 

\vspace{1mm} \noindent \textbf{MC-LIME} (Minimal Counterfactual LIME) \cite{2022.EDM-short-papers.33} finds the minimal set of LIME features that, if changed, would alter the outcome. It focuses on features that increase the likelihood of an event (e.g., student dropout). MC-LIME applies changes to individual features, then pairs, and so on, until a change flips the prediction. This approach combines explanation sparseness with the advantages of counterfactuals and LIME.

% Explanation Selection Phase
\subsection{Explanation Selection}
\label{sec:explanation-selection}
Our \il pipeline generates explanations aligned to social science theories. Specifically, we have translated eight theories into prompts: \textit{Relevance Selection (RS), Abnormal Conditions (AC), Pearl's Model (Pearl's), Necessity and Robustness (NR), two contrastive explanations (BC, Con), Statistical Relevance (SR),} and \textit{Chain of Thought (CoT)}. Our prompt structure contains
 (1) descriptions of the model, features, XAI method, and course context, (2) explainer importance scores and relevant student feature values, and (3) social science theory instructions. We present the general prompt template and three theory-specific examples (AC, Pearl's Model, Contrastive) in this section, all details are in Appendix \ref{appendix:exp-sel-prompt}.
 
% We now present the social science theories of explanation and the instructions derived from them.
%
\begin{figure}[ht]
    \centering
    \includegraphics[width=\linewidth]{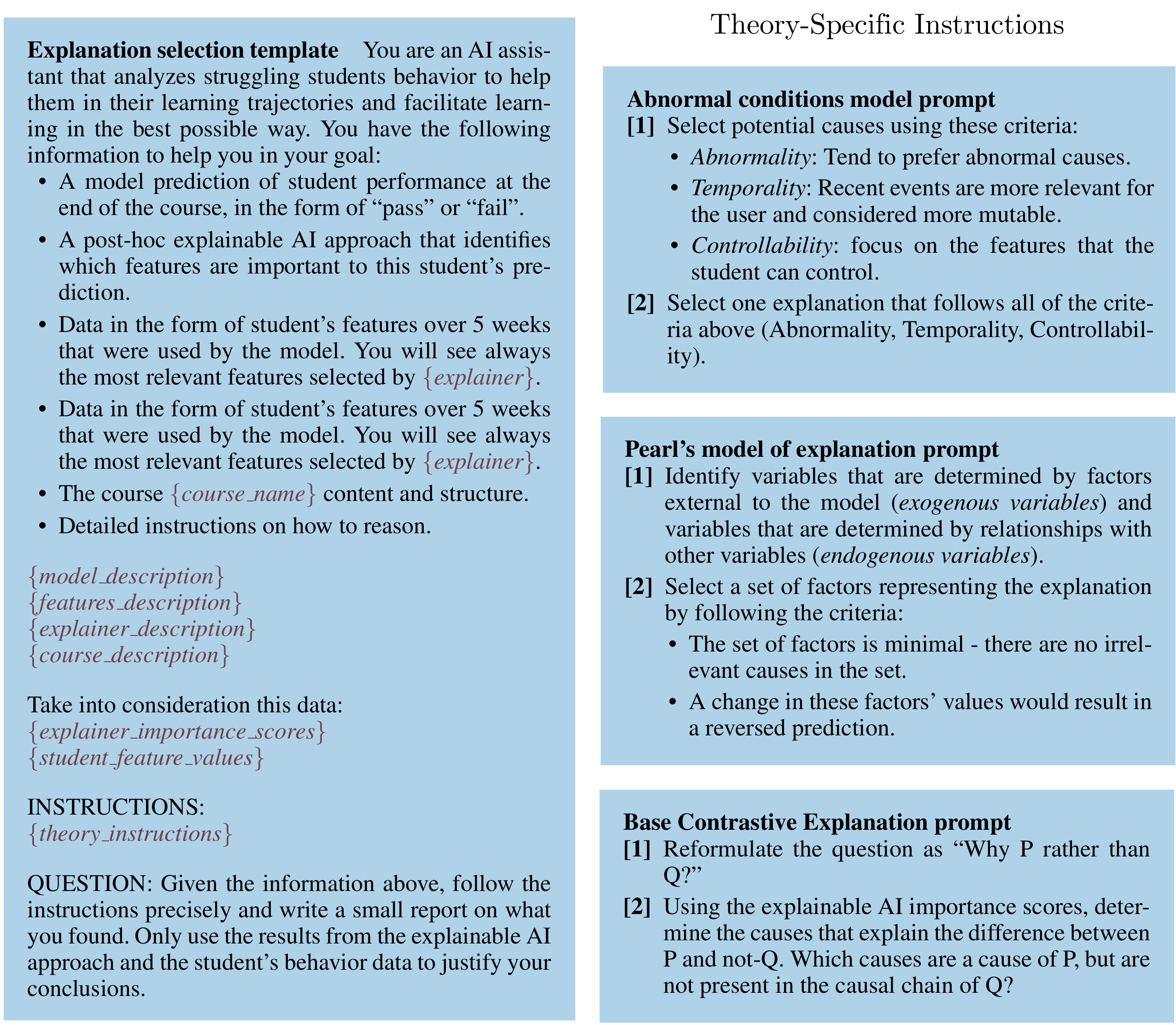}
    \caption{Prompts for explanation selection (left) and AC, Pearl's model and Contrastive explanation theories (right).}
    \label{fig:prompts}
\end{figure}

\vspace{1mm} \noindent \textbf{Relevance Selection.} Based on Hilton's conversational model of explanation \cite{hilton1990conversational}, relevance-based selection theory aims to ``resolve a puzzle in the explainee's mind'' by filling gaps in their knowledge. The theory emphasizes that shared knowledge between the explainer and the explainee are presuppositions of the explanations, and the other factors are the causes that should be explained. In short, the explainer should not explain any causes they think the explainee already knows.

\vspace{1mm} \noindent \textbf{Abnormal conditions.} This theory, based upon Hilton and Slugoski's abnormal conditions model \cite{hilton1986knowledge}, suggests that explanations often rely on unusual and temporally proximal events. People do not solely count on statistical likelihood but highlight uncommon factors contributing to an event to explain that event. During a conversation, the explainer is relying on the perceived common prior knowledge to identify potential causes that are considered abnormal, with greater weight given to temporally proximal events and factors that the explainee can control. This focus on controllable factors helps the explainee understand how to potentially avoid similar situations in the future.

\vspace{1mm} \noindent \textbf{Pearl's Model of explanation.} \cite{halpern2005causes} present a formal framework for selecting explanations based on epistemic relevance and structural causal models. The model distinguishes between \textit{exogenous variables}, whose values are determined by external factors, and \textit{endogenous variables}, whose values are influenced by relationships with other variables. 
Within a \textit{context} (a specific assignment of values to variables), the model defines an \textit{actual cause} as a minimal set of events that must occur for an event to happen. 

\vspace{1mm} \noindent \textbf{Necessity \& Robustness selection.} Two key criteria for selecting strong explanatory causes are necessity and robustness \cite{lipton1990contrastive}. \textit{Necessity} refers to whether a cause is essential for the effect to occur. \textit{Robustness} considers how generally a cause applies \cite{lombrozo2010causal}. This idea aligns with the concept of simplicity, where broader explanations with fewer specific requirements are favored.

\vspace{1mm} \noindent \textbf{Contrastive Explanation.} This theory suggests explanations are not simply cause and effect statements, but rather comparisons between what happened (the target event) and what could have happened (a counterfactual contrasting event) \cite{hilton1990conversational}. One possible way to make the identification of the counterfactual event (foil) successful is to ask the module to reformulate the question. This technique is known as Rephrase and Respond (RaR) \cite{deng2024rephrase}. 

\vspace{1mm} \noindent \textbf{Statistical Relevance.} This method is based on the SR model based on scientific causal reasoning \cite{sep-scientific-explanation}. The SR model explanations can be defined in simple terms as ``statistically relevant properties are explanatory and statistically irrelevant properties are not''. It follows this structure: \textit{‘‘Based on empirical data, factors A, B and C contribute to the probability of Y by the amount of X’’}.

\vspace{1mm} \noindent \textbf{Chain-of-Thought (baseline).} Chain-of-Thought (CoT) prompting \cite{wei2023chainofthought} guides an LLM through sequential reasoning that mimics human thought processes.

\vspace{1mm} \noindent \textbf{Evaluation.} To assess whether \il responses align with post-hoc explanations, student feature values, and course context, we developed an annotation rubric based on decomposed questions as a basis for human and LLM annotation. Recent studies \cite{wang2023chatgpt} demonstrated that LLMs can match human annotators, especially when instructions are decomposed into simple criteria phrased as binary (``yes''/``no'') questions \cite{qin2024infobench}. We created four general decomposed questions applicable across all theories and additional theory-specific questions ($1-6$, depending on the theory). These were validated by a computational social scientist who was not involved in the prompt creation process. The general questions used items such as `Is the generated text correctly using the model's predicted outcome?' or `Is the generated text analysis based solely on the explainer results provided?'. An example of a theory-specific questions is: `Is the generated text selecting the causes that are most relevant to the user?' (\textit{RS}). The complete rubric is available in Appendix Table~\ref{appendix:tab-dqs}.

We instructed three experts to annotate the same block of $42$ responses (2 students, 3 explainers, 8 theories) across a set of up to eight decomposed questions for each setting, resulting in an inter-rater agreement of $\kappa=0.71\pm0.13$ (Cohen's Kappa). After that, each annotator proceeded with independently evaluating between $84$ to $105$ more responses, leading to a total of $315$ human-annotated responses and in total over $2,350$ annotations. We then proceeded with instructing GPT-4o to annotate the same $42$ responses using the same decomposed questions. We obtained an agreement of $94.68\% \pm 5.20$ (percentage of answers with agreement) between human and GPT-4o annotation. The inter-rater agreement per theory is in Appendix Table~\ref{tab:annot-agree-gpt}. Given the high agreement, we annotated all explanations by GPT-4o.

\subsection{Explanation Presentation}
\label{sec:explanation-presentation}
% (see Section~\ref{sec:explanation-presentation})
Inspired by Hilton's conversational model \cite{hilton1990conversational}, we employ a presentation prompt to refine and condense the output of the explanation selection prompt to make it relevant to the explainee's future actions, current context and prior knowledge. This prompt is formulated using Grice's maxims on communication \cite{grice1975logic} and learning science literature on best practices in communicating feedback to students \cite{shute2008focus,hattie2007power}. Grice's maxims provide a framework for understanding cooperative conversation, emphasizing providing the right amount of relevant and clear information to achieve a shared goal. Grice further divides this principle into four maxims: \textit{Quality} (truthfulness and evidence-based statements), \textit{Quantity} (providing enough but not excessive information), \textit{Relation} (relevance), and \textit{Manner} (clear and concise communication). \cite{hattie2007power} provide a prominent educational feedback framework with three steps: \textit{``Where am I going?''}, \textit{``How am I doing?''\footnote{We adapted 
 \cite{hattie2007power}'s question \textit{``How am I going?''} to \textit{``How am I doing?''} for ease of understanding.}}, and \textit{``Where to next?''} These steps require the feedback provider to clearly state the learning goal, provide a summary of relevant student performance, and suggest concrete actions for improvement in the near future. The full explanation presentation prompt is included in the Appendix \ref{appendix:exp-pres-prompt}.

\vspace{1mm} \noindent \textbf{Evaluation.} Similar to the explanation selection phase, we evaluated the final explanations using a rubric of decomposed `yes'/`no' questions (the detailed set of questions is illustrated in Appendix Table~\ref{appendix:tab-dqs}). Given the high agreement between GPT-4o and human annotations in the explanation selection phase, we used GPT-4o as an expert. Additionally, we conducted a readability evaluation using the following metrics: \textit{Flesch-Kincaid Grade Level} \cite{flesch1948new}, \textit{Gunning Fog Index} \cite{gunning1952technique}, \textit{SMOG index} \cite{mc1969smog}, and LanguageTool \textit{Grammar Issues} \cite{mozgovoy2011dependency}. These metrics evaluate the comprehensibility of the text in terms of sentence length, vocabulary complexity, grammatical correctness, and estimated years of schooling needed for understanding.

\begin{table*}[t]
\small
\centering
\resizebox{1.0\textwidth}{!}{
\begin{tabular}{@{}clllllll@{}}
\toprule
\textbf{Explainer} & \multicolumn{1}{c}{\textbf{Theory}} & \multicolumn{1}{c}{\textbf{Overall}} & \multicolumn{1}{c}{\textbf{\begin{tabular}[c]{@{}c@{}}Q1: Using \\ provided data\\ extensively\end{tabular}}} & \multicolumn{1}{c}{\textbf{\begin{tabular}[c]{@{}c@{}}Q2: Analysis\\ based solely\\ on explainer\end{tabular}}} & \multicolumn{1}{c}{\textbf{\begin{tabular}[c]{@{}c@{}}Q3: Correctly\\ using model's\\ prediction\end{tabular}}} & \multicolumn{1}{c}{\textbf{\begin{tabular}[c]{@{}c@{}}Q4: Using\\ course content\\ and structure\end{tabular}}} & \multicolumn{1}{c}{\textbf{\begin{tabular}[c]{@{}c@{}}Theory-\\ Specific Qs\end{tabular}}} \\ \midrule
 & \textit{RaR + Contrastive} & {\color[HTML]{036400} \textbf{0.985 ± 0.069}} & {\color[HTML]{036400} 0.99 ± 0.099} & {\color[HTML]{036400} 0.992 ± 0.087} & {\color[HTML]{036400} 0.998 ± 0.047} & {\color[HTML]{036400} 0.967 ± 0.178} & {\color[HTML]{036400} 0.984 ± 0.085} \\
 & \textit{Abnormal Conditions} & {\color[HTML]{036400} \textbf{0.958 ± 0.141}} & {\color[HTML]{036400} 0.997 ± 0.057} & {\color[HTML]{036400} 0.997 ± 0.057} & {\color[HTML]{036400} 0.998 ± 0.047} & 0.865 ± 0.342 & {\color[HTML]{036400} 0.952 ± 0.156} \\
 & \textit{Relevance Selection} & \textbf{0.949 ± 0.155} & {\color[HTML]{036400} 0.998 ± 0.047} & {\color[HTML]{036400} 0.999 ± 0.033} & {\color[HTML]{036400} 0.997 ± 0.057} & 0.898 ± 0.303 & 0.925 ± 0.195 \\
 & \textit{Necessity Robustness} & \textbf{0.949 ± 0.154} & {\color[HTML]{036400} 0.996 ± 0.066} & {\color[HTML]{036400} 0.997 ± 0.057} & {\color[HTML]{036400} 0.998 ± 0.047} & {\color[HTML]{9A0000} 0.676 ± 0.468} & {\color[HTML]{036400} 0.991 ± 0.047} \\
 & \textit{Pearl Explanation} & \textbf{0.948 ± 0.099} & {\color[HTML]{036400} 0.979 ± 0.142} & {\color[HTML]{036400} 0.983 ± 0.131} & {\color[HTML]{036400} 0.998 ± 0.047} & 0.869 ± 0.337 & 0.939 ± 0.071 \\
 & \textit{Statistical Relevance} & \textbf{0.884 ± 0.184} & {\color[HTML]{036400} 0.988 ± 0.109} & {\color[HTML]{036400} 0.993 ± 0.081} & {\color[HTML]{036400} 0.996 ± 0.066} & {\color[HTML]{9A0000} 0.453 ± 0.498} & {\color[HTML]{036400} 0.991 ± 0.001} \\
 & \textit{(Base) Contrastive} & {\color[HTML]{000000} \textbf{0.852 ± 0.194}} & {\color[HTML]{036400} 0.968 ± 0.175} & {\color[HTML]{9A0000} 0.647 ± 0.478} & {\color[HTML]{036400} 1.0 ± 0.0} & {\color[HTML]{9A0000} 0.681 ± 0.466} & 0.879 ± 0.179 \\
\multirow{-8}{*}{\textbf{CEM}} & \textit{Chain of Thought} & \textbf{0.872 ± 0.188} & {\color[HTML]{036400} 0.979 ± 0.142} & {\color[HTML]{036400} 0.984 ± 0.127} & {\color[HTML]{036400} 0.997 ± 0.057} & {\color[HTML]{9A0000} 0.585 ± 0.493} & 0.817 ± 0.001 \\ \midrule
 & \textit{RaR + Contrastive} & {\color[HTML]{036400} \textbf{0.975 ± 0.098}} & {\color[HTML]{036400} 0.974 ± 0.16} & 0.985 ± 0.123 & {\color[HTML]{036400} 1.0 ± 0.0} & 0.946 ± 0.227 & {\color[HTML]{036400} 0.975 ± 0.111} \\
 & \textit{Relevance Selection} & {\color[HTML]{036400} \textbf{0.96 ± 0.112}} & {\color[HTML]{036400} 0.987 ± 0.114} & 0.991 ± 0.093 & {\color[HTML]{036400} 0.998 ± 0.047} & 0.918 ± 0.274 & 0.946 ± 0.132 \\
 & \textit{Abnormal Conditions} & {\color[HTML]{036400} \textbf{0.951 ± 0.131}} & {\color[HTML]{036400} 0.99 ± 0.099} & 0.993 ± 0.081 & {\color[HTML]{036400} 0.995 ± 0.074} & 0.893 ± 0.309 & 0.938 ± 0.156 \\
 & \textit{Necessity Robustness} & \textbf{0.929 ± 0.147} & {\color[HTML]{036400} 0.982 ± 0.135} & 0.984 ± 0.127 & {\color[HTML]{036400} 0.998 ± 0.047} & {\color[HTML]{9A0000} 0.659 ± 0.474} & {\color[HTML]{036400} 0.959 ± 0.103} \\
 & \textit{Pearl Explanation} & \textbf{0.896 ± 0.106} & 0.904 ± 0.294 & 0.922 ± 0.269 & {\color[HTML]{036400} 0.989 ± 0.104} & 0.836 ± 0.371 & 0.88 ± 0.114 \\
 & \textit{Statistical Relevance} & \textbf{0.859 ± 0.144} & 0.919 ± 0.272 & 0.964 ± 0.186 & {\color[HTML]{036400} 0.984 ± 0.127} & {\color[HTML]{9A0000} 0.476 ± 0.5} & {\color[HTML]{036400} 0.953 ± 0.001} \\
 & \textit{(Base) Contrastive} & \textbf{0.822 ± 0.195} & 0.97 ± 0.172 & {\color[HTML]{9A0000} 0.45 ± 0.498} & {\color[HTML]{036400} 0.998 ± 0.047} & {\color[HTML]{9A0000} 0.624 ± 0.485} & 0.884 ± 0.183 \\
\multirow{-8}{*}{\textbf{LIME}} & \textit{Chain of Thought} & \textbf{0.82 ± 0.188} & 0.945 ± 0.229 & 0.956 ± 0.204 & {\color[HTML]{036400} 0.997 ± 0.057} & {\color[HTML]{9A0000} 0.526 ± 0.5} & {\color[HTML]{9A0000} 0.675 ± 0.001} \\ \midrule
 & \textit{RaR + Contrastive} & {\color[HTML]{036400} \textbf{0.985 ± 0.063}} & {\color[HTML]{036400} 0.987 ± 0.114} & {\color[HTML]{036400} 0.989 ± 0.104} & {\color[HTML]{036400} 0.999 ± 0.033} & {\color[HTML]{036400} 0.968 ± 0.175} & {\color[HTML]{036400} 0.983 ± 0.073} \\
 & \textit{Relevance Selection} & {\color[HTML]{036400} \textbf{0.963 ± 0.132}} & {\color[HTML]{036400} 0.996 ± 0.066} & {\color[HTML]{036400} 0.997 ± 0.057} & {\color[HTML]{036400} 1.0 ± 0.0} & 0.915 ± 0.279 & 0.949 ± 0.153 \\
 & \textit{Abnormal Conditions} & {\color[HTML]{036400} \textbf{0.96 ± 0.12}} & {\color[HTML]{036400} 0.991 ± 0.093} & {\color[HTML]{036400} 0.991 ± 0.093} & {\color[HTML]{036400} 0.993 ± 0.081} & 0.879 ± 0.326 & {\color[HTML]{036400} 0.957 ± 0.135} \\
 & \textit{Necessity Robustness} & \textbf{0.932 ± 0.162} & {\color[HTML]{036400} 0.99 ± 0.099} & {\color[HTML]{036400} 0.99 ± 0.099} & {\color[HTML]{036400} 1.0 ± 0.0} & {\color[HTML]{9A0000} 0.609 ± 0.488} & {\color[HTML]{036400} 0.978 ± 0.083} \\
 & \textit{Pearl Explanation} & \textbf{0.919 ± 0.108} & {\color[HTML]{000000} 0.942 ± 0.233} & {\color[HTML]{036400} 0.951 ± 0.216} & {\color[HTML]{036400} 0.992 ± 0.087} & 0.806 ± 0.395 & 0.914 ± 0.104 \\
 & \textit{Statistical Relevance} & \textbf{0.876 ± 0.171} & {\color[HTML]{036400} 0.97 ± 0.172} & {\color[HTML]{036400} 0.99 ± 0.099} & {\color[HTML]{036400} 0.993 ± 0.081} & {\color[HTML]{9A0000} 0.446 ± 0.497} & {\color[HTML]{036400} 0.978 ± 0.001} \\
 & \textit{(Base) Contrastive} & \textbf{0.856 ± 0.19} & {\color[HTML]{036400} 0.97 ± 0.172} & {\color[HTML]{036400} 0.615 ± 0.487} & {\color[HTML]{036400} 0.996 ± 0.066} & {\color[HTML]{9A0000} 0.688 ± 0.497} & {\color[HTML]{000000} 0.895 ± 0.187} \\
\multirow{-8}{*}{\textbf{MC-LIME}} & \textit{Chain of Thought} & \textbf{0.858 ± 0.199} & {\color[HTML]{036400} 0.973 ± 0.163} & {\color[HTML]{036400} 0.982 ± 0.135} & {\color[HTML]{036400} 0.999 ± 0.033} & {\color[HTML]{9A0000} 0.565 ± 0.496} & 0.77 ± 0.001 \\ \bottomrule
\end{tabular}}
\caption{\textbf{Alignment of GPT-4o, Gemma2 9b, and Llama3 70B generated explanations with theory}. Average (± std) of ``Yes'' answers, displayed separately for the first four general decomposed questions as well as avg. over the theory-specific questions and overall. Annotated by experts and GPT-4o. Scores over $95\%$ (less than $65\%$) are highlighted in green (red).}
\vspace{-6mm}
\label{tab:gpt-human-comparison}
\end{table*}

\subsection{Student Preferences (User Study)}
\label{sec:student-preferences}
To evaluate students' explanation preferences as well as the actionability of \il explanations, we conducted a user study comparing our explanations \textbf{(text and visual)} to post-hoc baselines. We recruited $114$ students on \texttt{Prolific}, (see Appendix \ref{appendix:study-background} for detailed information about the participants' demographics). Students were told that the explanations related to their own performance in three different online courses they were enrolled in.

For each course, we presented participants with eight explanations on their predicted success or failure in that course: four explanations at a time, with three randomly ordered \il variations and one baseline approach. We elected to use the six \il instructions with the highest instruction-following accuracy from the human-expert evaluation for this experiment, and an equal mixture of passing and failing behavior at different model confidence levels. Each explanation was provided as a brief text accompanied by a graph illustrating the features and concepts used by the model. GPT-4o created all \il visuals based on the first two responses (full prompt in Appendix \ref{appendix:exp-viz-prompt}). LIME visuals were used from the package, and CEM and MC-LIME visuals were expert-created and iterated upon with six pilot participants. Examples of the study format are included in our repository and Appendix \ref{appendix:study-materials}. Participants were asked to choose their preferred explanation for each set of comparisons and explain their choice (open-ended question). They were then asked to compare the explanations based on five criteria \cite{frej2024finding}:

\begin{enumerate}
\small
    \item \textbf{Usefulness}: This explanation is useful to understand the prediction based on my learning behavior.
    \item \textbf{Trustworthiness}: This explanation lets me judge if I should trust the suggestions.
    \item \textbf{Actionability}: This explanation helps me make a decision on how to improve my learning behavior.
    \item \textbf{Completeness}: This explanation has sufficient detail to understand why the prediction was made based on my learning behavior.
    \item \textbf{Conciseness}: Every detail of this explanation is necessary.
\end{enumerate}

Finally, participants were asked to choose one of ten suggested actions for the next week, based on their preferred explanation. These actions were aligned with behavioral features from the model, allowing us to simulate the impact if the student acted according to them. We trained a BiLSTM to predict student success on six weeks of student data, and conducted inference on simulated students increasing the relevant features by 25\% percentile. We also asked students which weeks of material they would focus on based on the explanation (between one to three weeks). 

% Results
\vspace{-1mm}
\section{Results}
We evaluated \il explanations by assessing the LLM's instruction-following abilities during explanation selection (Exp 1) and using readability metrics and an automated analysis to measure explanation understandability at the presentation stage (Exp 2). We then analyzed student preferences in the user study (Exp 3) and simulated whether actions derived from the explanations improved student performance (Exp 4).

\vspace{1mm} \noindent \textbf{Experimental Protocol.} We optimized the BiLSTM models (one for each course) using a train-validation-test split of 80:10:10 and including a hyperparameter search. We achieve balanced accuracies of $90.8$ (DSP), $80.3$ (Afr), and $76.8$ (Geo) respectively. These results for early student performance prediction at five weeks are in line with prior work \cite{swamy2022meta}. Explainers (LIME, CEM) were extracted with the same settings as per related work to ensure a fair comparison \cite{swamy2022evaluating, swamy2023trusting}.

\subsection{Exp 1: \il is aligned with social science theories of explanation} 
\label{sec:exp1}
In the first analysis, we evaluated how well the generated explanations aligned with the instructions. We selected $105$ representative students per course, distributed across six behavioral dimensions (regularity, effort, consistency, proactivity, control, and assessment) \cite{mejia2022identifying}. Table~\ref{tab:gpt-human-comparison} shows the average (with standard deviation) scores for each general decomposed question (see Section~\ref{sec:explanation-selection} and Appendix~\ref{appendix:decomposed-qs}), for the theory-specific questions, and overall. The general decomposed questions involve whether the response is using the provided data extensively (Q1), whether the analysis in the response is based solely on the results from the explainer (Q2), whether the response is correctly using the model's prediction (Q3), and whether the response is using the course content and structure. The scores represent the ratio of ‘Yes’ answers averaged across students and courses, with overall scores exceeding $0.82$ for all explainers and theories. There were no significant differences between explainers or theories, as indicated by overlapping $95\%$ CIs.
There were generally also no differences in scores between  questions. Only Q4 (``Using course content and structure'') had lower average scores than the other questions, suggesting that generating explanations fully incorporating the course content is a challenge. Notably, inter-rater agreement for Q4 was also lower than the average ($\kappa=0.52$ vs. $\kappa=0.71$ overall), indicating lower reliability of annotations for this question.

\begin{wrapfigure}{r}{0.45\textwidth}
    \vspace{-3mm}
    \centering
    \small
    \resizebox{0.45\textwidth}{!}{
    \begin{tabular}{@{}llll@{}}
    \toprule
    \textbf{Model} & \textbf{Explainer} & \textbf{\begin{tabular}[c]{@{}l@{}}Explanation \\ Selection\end{tabular}} & \textbf{\begin{tabular}[c]{@{}l@{}}Explanation \\ Presentation\end{tabular}} \\ \midrule
    \multirow{3}{*}{\textbf{\begin{tabular}[c]{@{}l@{}}Gemma2\\ 9b\end{tabular}}} & \textit{CEM} & 0.941 ± 0.202 & 0.791 ± 0.212 \\
    & \textit{LIME} & 0.937 ± 0.217 & 0.801 ± 0.201 \\
    & \textit{MC-LIME} & 0.939 ± 0.208 & 0.767 ± 0.165 \\ \midrule
    \multirow{3}{*}{\textbf{GPT-4o}} & \textit{CEM} & 0.961 ± 0.148 & 0.992 ± 0.070 \\
    & \textit{LIME} & 0.954 ± 0.165 & 0.992 ± 0.072 \\
    & \textit{MC-LIME} & 0.965 ± 0.144 & 0.992 ± 0.072 \\ \midrule
    \multirow{3}{*}{\textbf{\begin{tabular}[c]{@{}l@{}}Llama3\\ 70b\end{tabular}}} & \textit{CEM} & 0.897 ± 0.222 & 0.998 ± 0.037 \\
    & \textit{LIME} & 0.848 ± 0.302 & 0.997 ± 0.042 \\
    & \textit{MC-LIME} & 0.881 ± 0.258 & 0.998 ± 0.037 \\ \bottomrule
    \end{tabular}}
    \caption{\textbf{Explanation quality by LLM}. Degree of instruction-following for explanations generated by GPT-4o, Gemma2 9b, and Llama3 70b. Average (with standard deviation) of ``Yes'' answers, annotated by GPT-4o.}
    \label{tab:exp1-llm-comparison}
    \vspace{-7mm}
\end{wrapfigure}

In a second analysis, we compared different LLMs' abilities to generate explanations according to instructed theories. Table~\ref{tab:exp1-llm-comparison} (column \textit{Explanation Selection}) shows the average scores for each model and explainer for the same representative students. Scores reflect the number of `Yes' answers to the decomposed questions described in Section~\ref{sec:explanation-presentation}, with annotation done automatically using GPT-4o. GPT-4o \il explanations scored the highest, closely followed by Gemma2 9b, and Llama3 70b. However, all of the $95\%$ CI overlap. Again, we found no differences between the different explainers. An experiment on the generalizability of the results to the flipped classroom context (smaller dataset, different domain) is included in Appendix \ref{appendix:generalization}.

\subsection{Exp 2: \il explanations are understandable} 
We assessed the presentation of \il explanations using both LLM annotation and readability metrics. Table~\ref{tab:exp1-llm-comparison} (column \textit{Explanation Presentation}) shows scores per LLM and explainer, averaged over $300$ representative students (see Section~\ref{sec:exp1}). Scores represent the ratio of ``Yes'' answers to the set of decomposed questions described in Section~\ref{sec:explanation-presentation}, as annotated by GPT-4o. For this stage, Llama3 70b achieved the highest scores, followed closely by GPT-4o and Gemma2 9b. This finding is in contrast to the explanation selection stage, where GPT-4o reached the highest score, suggesting that depending on the use case (whether selection or presentation is more important), it is possible to use much smaller and open source models for the task at hand.

\begin{figure}
        \centering
        \includegraphics[width=0.8\linewidth, trim={0 10 0 0},clip]{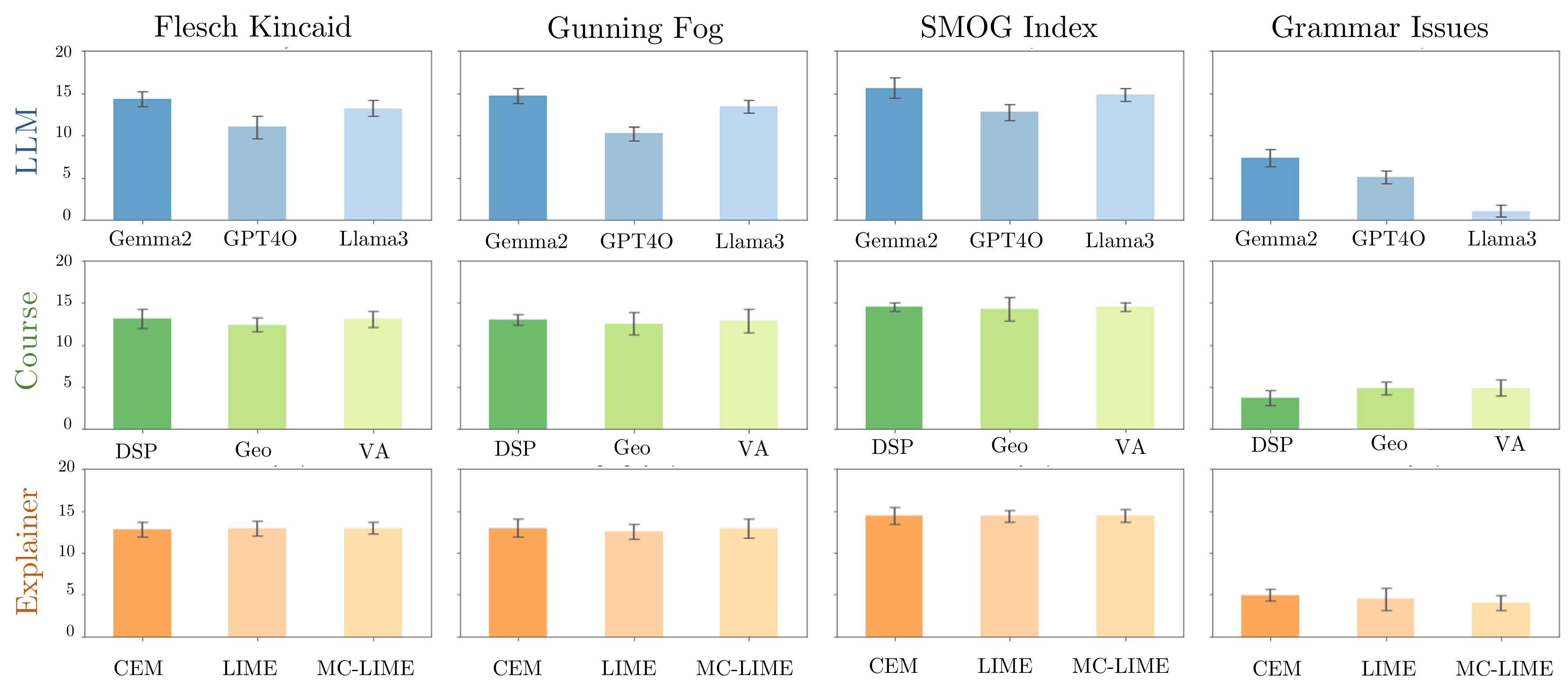}
        \caption{\textbf{Readability metrics.} Flesch Kincaid, Gunning Fog, SMOG Index, Grammar Issues across LLM (blue), course (green), and explainer (orange). Lower scores are better.}
        \label{fig:eq2-readability}
        \vspace{-6mm}
\end{figure}

%Table showing a comparison between different LLMs
% \begin{table}[t]
%   \centering
%     \small
%   \begin{tabular}{@{}llll@{}}
%   \toprule
%   \textbf{Model} & \textbf{Explainer} & \textbf{\begin{tabular}[c]{@{}l@{}}Explanation \\ Selection\end{tabular}} & \textbf{\begin{tabular}[c]{@{}l@{}}Explanation \\ Presentation\end{tabular}} \\ \midrule
%   \multirow{3}{*}{\textbf{\begin{tabular}[c]{@{}l@{}}Gemma2\\ 9b\end{tabular}}} & \textit{CEM} & 0.941 ± 0.202 & 0.791 ± 0.212 \\
%    & \textit{LIME} & 0.937 ± 0.217 & 0.801 ± 0.201 \\
%    & \textit{MC-LIME} & 0.939 ± 0.208 & 0.767 ± 0.165 \\ \midrule
%   \multirow{3}{*}{\textbf{GPT-4o}} & \textit{CEM} & 0.961 ± 0.148 & 0.992 ± 0.070 \\
%    & \textit{LIME} & 0.954 ± 0.165 & 0.992 ± 0.072 \\
%    & \textit{MC-LIME} & 0.965 ± 0.144 & 0.992 ± 0.072 \\ \midrule
%   \multirow{3}{*}{\textbf{\begin{tabular}[c]{@{}l@{}}Llama3\\ 70b\end{tabular}}} & \textit{CEM} & 0.897 ± 0.222 & 0.998 ± 0.037 \\
%    & \textit{LIME} & 0.848 ± 0.302 & 0.997 ± 0.042 \\
%    & \textit{MC-LIME} & 0.881 ± 0.258 & 0.998 ± 0.037 \\ \bottomrule
%   \end{tabular}
% \caption{\textbf{Explanation quality by LLM}. Degree of instruction-following for explanations generated by GPT-4o, Gemma2 9b, and Llama3 70b. Average (with standard deviation) of ``Yes'' answers, annotated by GPT-4o.}
% \vspace{-6mm}
%  \label{tab:exp1-llm-comparison}
% \end{table}

Figure~\ref{fig:eq2-readability} illustrates the readability scores (Flesch Kincaid, SMOG Index, Gunning Fog, Grammar Issues) of the final \il explanations after the explanation presentation prompt. Lower scores are better for all metrics. GPT-4o reached the best readability performance of the three LLMs (Fig~\ref{fig:eq2-readability}, top), with no overlap in $95\%$ confidence intervals. In contrast, Llama3 70b committed the least grammatical errors. We found no differences in readabilty or grammar between explanations for the different courses Fig~\ref{fig:eq2-readability}, middle), demonstrating that our approach is generalizable to diverse educational contexts. Similarly, all explainers achieved similar scores across all four metrics Fig~\ref{fig:eq2-readability}, bottom), showing that choice of source explainer does not have a strong impact on explanation understandability.

% \begin{figure}[t]
%     \centering
%     \includegraphics[width=0.43\linewidth, trim={0 10 0 0},clip]{images/08_illuminate/metrics_new.pdf}
%     \caption{\textbf{Readability metrics.} Fleach Kincaid, Gunning Fog, SMOG Index, Grammar Issues across LLM (blue, top), course (green, middle), and explainer (orange, bottom). Lower scores are better.}
%     \vspace{-3.5mm}
%     \label{fig:eq2-readability}
% \end{figure}

\begin{figure}[t]
    \centering
    \begin{minipage}{0.48\linewidth}
        \centering
        \includegraphics[width=\linewidth, trim={0 15 0 0}, clip]{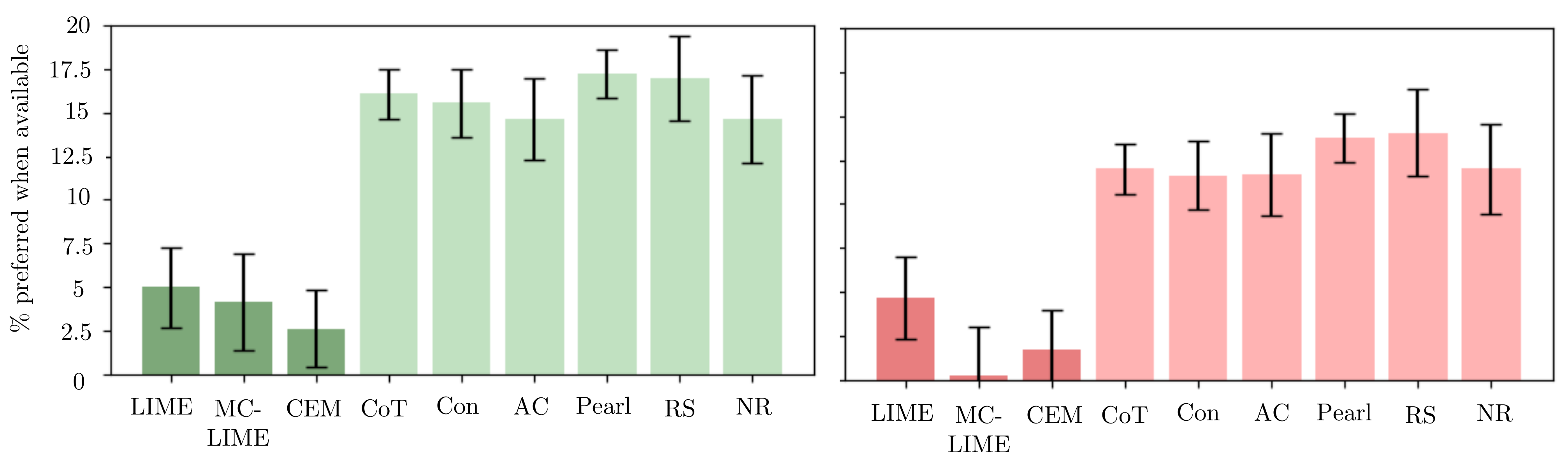}
        \caption{\textbf{Student preference of presented explanations for passing (left) and failing (right) student predictions.} Percentage of times a student chose each method when it was available. Higher scores are better.}
        \label{fig:exp3-studentpref}
    \end{minipage}
    \hfill
    \begin{minipage}{0.48\linewidth}
        \centering
        \includegraphics[width=\linewidth, trim={0 10 0 0}, clip]{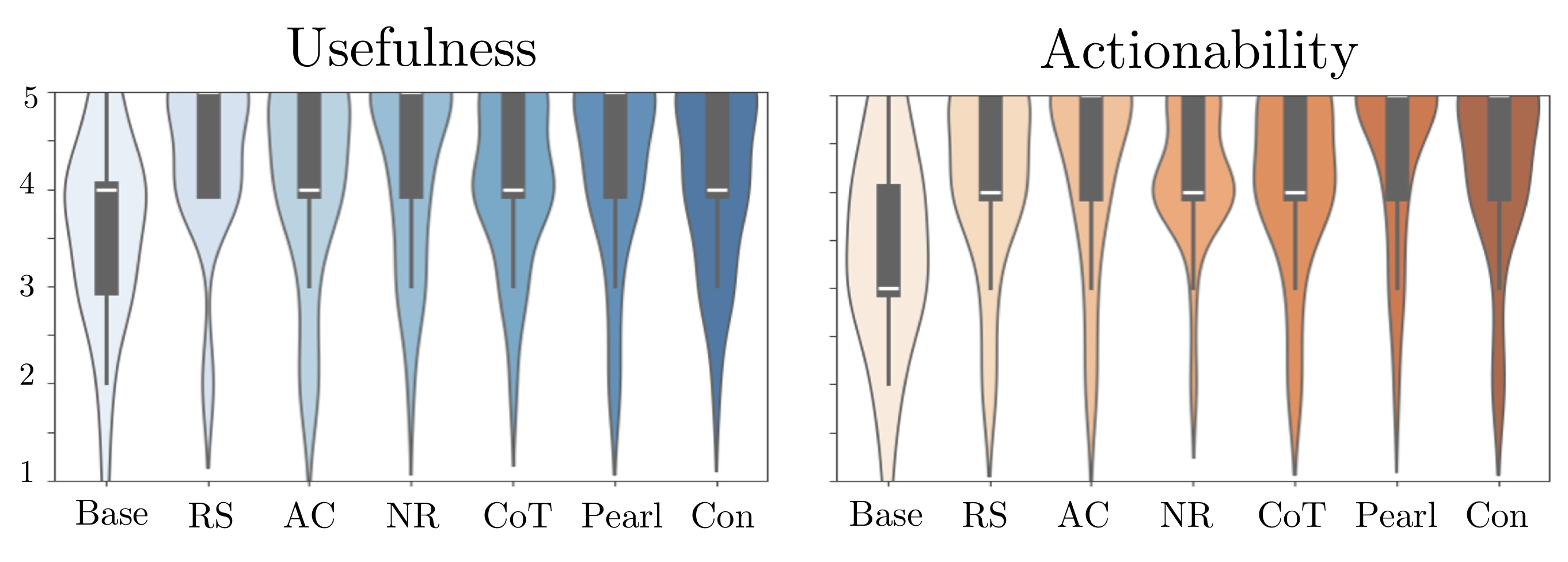}
        \caption{\textbf{Comparison of student preferences across two dimensions.} Distribution of Likert scores for \textit{Usefulness} and \textit{Actionability}, avg. over all participants and explainers.}
        \label{fig:exp3-criteria}
    \end{minipage}
    \vspace{-5mm}
\end{figure}

% \begin{figure}[t]
%     \centering
%     \includegraphics[width=0.43\linewidth,  trim={0 15 0 0},clip]{images/08_illuminate/student_comparison.pdf}
%     \caption{\textbf{Student preference of presented explanations for passing (left) and failing (right) student predictions.} Percentage of times a student chose each method when it was available. Higher scores are better.}
%     \vspace{-3.5mm}
%     \label{fig:exp3-studentpref}
% \end{figure}

% \begin{figure}[t!]
%     \centering
%     \includegraphics[width=0.43\linewidth, trim={0 10 0 0},clip]{images/08_illuminate/usefulness_flat.pdf}
%     \caption{\textbf{Comparison of student preferences across two dimensions.} Distribution of Likert scores for \textit{Usefulness} and \textit{Actionability}, averaged over all participants and explainers. }
%     \vspace{-5mm}
%     \label{fig:exp3-criteria}
% \end{figure}

\subsection{Exp 3: Students prefer \il} 
For each explanation in the user study, participants were asked to indicate their preferred explanation variation between one base explanation (LIME, CEM, MC-LIME) and three \il explanations. Figure~\ref{fig:exp3-studentpref} illustrates the percentage of times each method was preferred,
separately for either passing or failing student performance. Scores were averaged over all participants and explainers. Students overwhelmingly favored \il explanations over base ones. A Kruskal-Wallis test confirmed this preference, showing a significant difference between the base and \il explanations ($H=176.38, p<.0001$).

We evaluated students' responses to five Likert-scale questions on the usefulness, trustworthiness, actionability, completeness, and conciseness of explanations. Figure~\ref{fig:exp3-criteria} shows the Likert score distribution ($1$ - $5$) for two criteria per method; full results are in Appendix \ref{appendix:study-results}. Theory-based explanations received consistently high scores with no significant differences between theories. Base explanations were rated substantially lower (e.g., Usefulness${_\text{contrastive}}=4.18$, Usefulness${_\text{BASE}}=3.59$), highlighting the superior usefulness of \il explanations.

\vspace{-1mm}
\subsection{Exp 4: \il explanations can be effective at improving student performance} 
\label{sec:actionability}
Participants were asked to choose an action for the next week based on their preferred explanation. Over all explainers and theories, students most frequently selected actions to improve regularity of learning ($200$ responses) and attempt more problems ($147$ responses), while the least chosen action was to speed up quiz solving ($10$ responses). Participants also chose which weeks to focus on in the course, most commonly choosing weeks $6$ and $7$ ($329$, $248$ responses), which correspond to the weeks directly after the intervention. These were followed by review in weeks $5$ ($222$ responses) and $4$ ($198$ responses). This observation indicates that timing and proximity to the intervention influenced their choices. We also conducted a simulation experiment, applying participants' actions chosen for week 6 to student behavior in that week. For participants preferring \il responses, average performance improved significantly, independent of the underlying explainer: $13.5\%$ for LIME, $14.2\%$ for CEM, and $20.7\%$ for MC-LIME (full results in Appendix \ref{appendix:actionability}). MC-LIME’s effectiveness may stem from its minimal counterfactual approach, which greedily searches for the smallest set of features that cause the prediction to flip, making it suited for a single- or few-action intervention. Across different theories, both Contrastive and Necessity Robustness explanations result in the most actionable interventions with 28.2\% and 24.9\% average performance improvement respectively (Appendix Fig. \ref{fig:actionability}).

\section{Summary and Outlook}
In this work, we addressed the critical need for human-understandable explanations of complex models in education. We introduced \il, a theory-driven framework leveraging LLMs for generating NLEs through a chain of prompts, consisting of causal connection, explanation selection, and explanation presentation.

We tested our framework on $315$ students ($105$ per course) with all combinations of three post-hoc explainers, three LLMs, and eight prompting strategies, resulting in $21,915$ generated NLEs. We evaluated the instruction-following abilities using GPT-4o and human expert annotation, with both decomposed questions and readability metrics.
In a user study with $114$ university students, we found that students significantly preferred \il explanations over LIME, CEM, and MC-LIME and were able to derive actions that could improve their performance. 

Several challenges remain, including explainer variability and the difficulty of evaluating NLEs independently of prior beliefs and knowledge. In future work, we aim to explore interactivity of explanation dialogue with students \cite{slack2023explaining} and longitudinal LLM-XAI feedback effects. Our study highlights the shared potential of leveraging LLMs, eXplainable AI, and social science theories together towards scalable, personalized student support.

% \section{Acknowledgments}
% We kindly thank the Swiss State Secretariat for Education, Research and Innovation (SERI) for supporting this project. Oana-Maria Camburu was supported by a Leverhulme Early Career Fellowship. 

%% file: main/10_discussion.tex
This thesis advances the field of human-centered machine learning by addressing the critical balance between interpretability and model performance in neural networks. The contributions of this work are framed through five human-centric design needs of explainable AI (XAI): \textbf{accuracy}, \textbf{real-time}, \textbf{consistency}, \textbf{actionability}, and \textbf{human understandability}. Below, we revisit the findings and implications of this thesis within these dimensions, connecting them to the specific contributions of each chapter.

\section{Accurate Explanations} 

A key impact of interpretability lies in its ability to enhance explanation faithfulness by aligning explanations with the underlying decision-making processes of models. In \textit{Generalized Student Models} (Chapter 3), we explored a few insights from global interpretability via attention layers, highlighting which meta features (e.g., course-level characteristics) are important for early student performance prediction across diverse MOOC courses with thousands of students. However, this global interpretability does not extend to local explanations for individual predictions, leaving a gap in understanding tailored to specific students.

\textit{Sequential Multimodal Fusion for Interpretability} (Chapter 6) and \textit{Optimizing the Interpretability vs. Accuracy Tradeoff} (Chapter 7) address this gap by introducing interpretable-by-design models. \md achieves local explanation faithfulness through sequential modular networks, ensuring that individual modality contributions are transparently linked to predictions, and global insights can be captured over the whole model. Similarly, \icc employs sparse feature gating and conditional computation, enabling precise, task-specific explanations that align with the most informative inputs for each prediction. These advancements demonstrate how explanation faithfulness can be achieved both globally and locally.

\section{Real-Time Explanations} 

The computational inefficiency of post-hoc methods like SHAP and counterfacturals creates bottlenecks in real-time applications, as experienced in painstaking generations for \textit{Evaluating the Explainers} (Chapter 4) and \textit{Trusting the Explainers} (Chapter 5). We directly address this challenge with \md and \icc, where inherently interpretable models are designed to provide instant explanations at inference time. \md mitigates computational bottlenecks with a series of small, modular networks that can be used dynamically as part of a data stream. In comparison with monolithic multimodal models, \md  has significantly fewer parameters allowing both predictions and explanation to be extracted faster at inference time. Similarly, \icc employs a conditional computation framework to optimize real-time explanation delivery alongside predictions without compromising on accuracy or faithfulness. The \textit{Exploration of Explanation Consistency} in Chapter 8 presents additional costs only at training time for adversarial instance generation, and therefore does not compromise the real-time nature of methods like LIME. The communication layer presented in \il does add a few additional seconds, but the setting of mid-semester performance feedback allows for batch generation of explanations before presenting them to students. Future work on interactive explanations aims to front-load the computational costs of LLM-based explanations, calling tools dynamically based on student questions.

The contributions towards making explanations available in real-time are particularly relevant in time-sensitive educational contexts, such as adaptive learning systems, where delays in feedback could impact student outcomes, or increase boredom \cite{lee2022affects}. The shift of computational complexity from inference to training, as demonstrated in these chapters, highlights the feasibility of deploying explainable AI models in real-world scenarios where real-time predictions are essential.

\section{Consistent Explanations} 

The dimension of consistency addresses the stability of explanations across scenarios. \textit{Evaluating the Explainers} (Chapter 4) identifies significant and systematic disagreements between post-hoc explainers, raising concerns about their reliability in high-stakes domains like education. Building on this, Chapter 8 investigates the consistency of explanations across multiple runs, explainers, and course settings, demonstrating that random initialization of models and explainers can undermine trust in explanation outputs.

\md and \icc are inherently consistent models. For example, \icc learns a sparse feature gating over patterns from the data, and therefore will predict similar gating mechanisms for similar predictions Meanwhile, adversarial training techniques explored in Chapter 8 explicitly stabilize decision boundaries and enhance robustness, as demonstrated by less diverse explanations.

A key finding from Chapter 8 is that while educators are willing to tolerate some level of inconsistency in human teaching assistants, they are less forgiving of similar inconsistencies in AI systems. This insight underscores the importance of developing consistent explanations to foster trust among educators and ensure reliable decision-making in AI-powered learning systems.

\section{Actionable Explanations}

The dimension of actionability focuses on ensuring that explanations drive meaningful outcomes. \textit{Trusting the Explainers} (Chapter 5) highlights the potential of XAI to inform actionable interventions, as over 75\% of educators in the study identified concrete steps for course design improvements based on explainability insights. 

\icc and \il directly address this challenge. \icc focuses explanations on the most relevant feature subsets, enabling stakeholders to make informed decisions, and explicitly evaluates "usefulness" and "actionability" as dimensions in the aligned user study. Chapter 9 introduces \il, an LLM-XAI pipeline that transforms complex outputs into user-friendly explanations aligned with cognitive theories. By leveraging contrastive explanations and robustness selection, \il enhances educators' ability to understand and act on model predictions. These contributions demonstrate how actionable insights from XAI can empower users in designing personalized learning interventions and addressing biased behaviors within models.

\section{Human-Understandable Explanations} 
Human understandability emphasizes the importance of intuitive and accessible explanations. The user studies in \textit{Trusting the Explainers} (Chapter 5) and \icc highlight the importance of tailoring explanations to user expertise, based on the amount of pre-processing required to present explanations. 

\icc and \textit{An Exploration of Explainer Consistency} (Chapter 8) also contribute to human understandability through sparse explanations. \icc’s sparse feature gating make it easier for users to trace predictions back to specific inputs. The robustness explorations enable sparseness in explanations via adversarial training and stabilizing decision boundaries. These approaches reduce cognitive load and enhance clarity, ensuring that explanations are accessible to diverse audiences, including educators and students. In \icc, factors like ``conciseness" and ``completeness", evaluated in the user study with 56 teachers, are proxies for human understandability.

Out of all the research presented in this thesis, Chapter 9’s \il pipeline most directly addresses the human understandability dimension by transforming technical XAI outputs into hybrid visual and natural language explanations preferred by educators 89.2\% of the time. This work bridges the gap between technical explainability and practical usability by aligning explanations with social science frameworks, making them more comprehensible for non-technical stakeholders.

\section{Future Research Directions}

Broadly, the research presented in this thesis, as well as the future directions contribute to the overarching goal of \textbf{giving control back to AI system users,} enabling them to actively shape and refine model behavior rather than passively consuming its outputs. Empowering users with tools to interact, question, and correct models fosters transparency, builds trust, and aligns AI systems more closely with human needs and values, particularly in high-stakes domains like education. Literature on human-computer interaction highlights that fostering user agency in AI interactions can mitigate algorithm aversion, where users distrust or reject algorithmic advice following perceived errors \cite{Dietvorst2018}. This is particularly relevant in educational contexts, where a lack of trust can hinder the adoption of AI tools in education contexts. By integrating human-in-the-loop mechanisms, users are equipped to adapt model predictions, address biases, and ensure that AI outputs align with their context-specific needs \cite{nazaretsky2022teachers,holmes2022ethics}.

Building on the contributions of this thesis, one future direction is towards \textbf{giving users control of model behavior using human-in-the-loop explanations}. In this paradigm, users could provide corrective feedback based on explanations, such as specifying rules the model must follow (e.g., ``never use this feature in predictions") or flagging spurious correlations for exclusion. This iterative feedback loop would allow users to enforce domain-specific constraints, improving both model performance and alignment with human values. 

A second avenue for research focuses on development of \textbf{interactive explanations}, extending the capabilities of \texttt{iLLuMinaTE} by leveraging large language models (LLMs) as dynamic communication layers for XAI. These interactive systems could allow users to query and explore model explanations in real-time, adapting the granularity and terminology to individual needs. For instance, educators could ask follow-up questions to clarify a model's reasoning, explore counterfactual scenarios, or receive tailored suggestions for actionable interventions. Students could request additional problems, ask for revision ideas, or find relevant course content. Such interactivity would enhance trust and empower users to more effectively incorporate AI insights into decision-making.  

\textbf{A Parting Note.} After opening an early version of the \textit{Interpretable ML} book \cite{molnar2020interpretable} late at night in a Seattle diner, I had no choice but to read the whole book, cover to cover in one sitting. I was in awe, discovering how brilliantly intuitive the designs of explainers are, especially post-hoc methods like LIME \cite{lime}, SHAP \cite{shap} and Counterfactuals \cite{cem,dice}, and in-hoc methods like TCAV \cite{tcav}. My PhD proposal and the work in this thesis followed naturally. I hope that the research presented can spark a little of that same joy in the reader, with the goal of making Explainable AI methods even more accessible, human-friendly, and useful. 

Our team encourages applied ML practitioners to extend explainability evaluations into real-world studies, tackling diverse domains and messy datasets. In the age of LLMs, we believe explainability is more crucial than ever. Explainable AI must not provide explanations that align with ``what you want to hear" \cite{bender2021dangers} and instead optimize for ``what the model is actually doing" to build real and lasting trust in critical situations. Over the next 10 years, our dream is to bring safe and reliable AI from the lab into the modern classroom, towards trustworthy and quality learning experiences for all.

%% file: tail/appendix.tex
\appendix
\chapter{Appendix: Meta Transfer Learning (Chapter 3)}
\begin{table*}
\small
\centering
\resizebox{0.94\linewidth}{!}{
\begin{tabular}{@{}lll@{}}
\toprule
\textbf{Set} & \textbf{Feature} & \textbf{Description} \\ \midrule
\multirow{3}{*}{\textit{Regularity}} & DelayLecture & The average delay in viewing video lectures after they are released to students. \\
& RegPeakTimeDayHour & The extent to which students' activities are centered around a particular hour of the day. \\
& RegPeriodicityDayHour & The extent to which the hourly pattern of user’s activities repeats over days. \\
 \midrule
\multirow{13}{*}{\textit{Engagement}} 
 & NumberOfSessions & The number of unique online sessions the student has participated in. \\  
 & RatioClicksWeekendDay & The ratio between the number of clicks in the weekend and the weekdays \\ 
 & AvgTimeSessions & The average of the student's time per session. \\
 & TotalTimeSessions & The sum of the student's time in sessions. \\ 
 & StdTimeSessions & The standard deviation of student's time in sessions. \\ 
 & StdTimeBetweenSessions & The standard deviation of the time between sessions of each user. \\
 & TotalClicks & The number of clicks that a student has made overall. \\
 & TotalClicksProblem & The number of clicks that a student has made on problems this week. \\
 & TotalClicksVideo & The number of clicks that a student has made on videos this week. \\
 & TotalClicksWeekday & The number of clicks that a student has made on the weekdays. \\ 
 & TotalClicksWeekend & The number of clicks that a student has made on the weekends. \\ 
 & TotalTimeProblem & The total (cumulative) time that a student has spent on problem events. \\ 
 & TotalTimeVideo & The total (cumulative) time that a student has spent on video events. \\ 
 \midrule
\multirow{22}{*}{\textit{Control}}
& TotalClicksVideoLoad & The number of times a student loaded a video. \\
& TotalClicksVideo & The number of times a student clicked on a video (load, pause, play, forward). \\
& AvgWatchedWeeklyProp & The ratio of videos watched over the number of videos available. \\ 
& StdWatchedWeeklyProp & The standard deviation of videos watched over the number of videos available. \\
& AvgReplayedWeeklyProp & The ratio of videos replayed over the number of videos available. \\
& StdReplayedWeeklyProp & The standard deviation of videos replayed over the number of videos available. \\ 
& AvgInterruptedWeeklyProp & The ratio of videos interrupted over the number of videos available. \\
& StdInterruptedWeeklyProp & The standard deviation of videos interrupted over the number of videos available. \\ 
& FrequencyEventVideo & The frequency between every Video action and the following action. \\
& FrequencyEventLoad & The frequency between every Video.Load action and the following action. \\
& FrequencyEventPlay & The frequency between every Video.Play action and the following action. \\
& FrequencyEventPause & The frequency between every Video.Pause action and the following action. \\
& FrequencyEventStop & The frequency between every Video.Stop action and the following action. \\
& FrequencyEventSeekBackward & The frequency between every Video.SeekBackward action and the following action. \\
& FrequencyEventSeekForward & The frequency between every Video.SeekForward action and the following action. \\
& FrequencyEventSpeedChange & The frequency between every Video.SpeedChange action and the following action. \\
& AvgSeekLength & The student's average seek length (seconds). \\
& StdSeekLength & The student's standard deviation for seek length (seconds). \\
& AvgPauseDuration & The student's average pause duration (seconds). \\
& StdPauseDuration & The student's standard deviation for pause duration (seconds). \\
& AvgTimeSpeedingUp & The student's average time using Video.SeekForward actions (seconds). \\
& StdTimeSpeedingUp & The student's standard deviation of time using Video.SeekForward actions (seconds). \\
 \midrule
\multirow{7}{*}{\textit{Participation}} 
& CompetencyStrength & The extent to which a student passes a quiz getting the maximum grade with few attempts. \\
& CompetencyAlignment & The number of problems this week that the student has passed. \\
& CompetencyAnticipation & The extent to which the student approaches a quiz provided in subsequent weeks. \\
 & ContentAlignment & The number of videos this week that have been watched by the student. \\ 
 & ContentAnticipation & The number of videos covered by the student from those that are in subsequent weeks. \\ 
 & StudentSpeed & The average time passed between two consecutive attempts for the same quiz. \\ 
 & StudentShape & The extent to which the student receives the maximum quiz grade on the first attempt. \\
 \bottomrule
\end{tabular}}
\vspace{1mm}
\caption{Descriptions of 45 behavioral features used in model training as discussed in Sec. \ref{sec:features-descr}.}
\label{tab:features}
\end{table*}

\section{Hyperparameter Tuning}
\label{sec:opt-proc-2}
This appendix section details the hyperparameter tuning decisions made in the model optimization protocol (Sec. \ref{sec:opt-proc}), provided for community replication. The batch size was set to $64$. We used the Adam optimizer, with an initial learning rate of $0.001$. For all models, the number of \emph{BiLSTM} layers and the number of units per \emph{BiLSTM} layer were optimized via grid search. For the \texttt{BSM} model, we extended the grid search to the number of \emph{Dense} layers and the number of hidden units per layer in the final cascade. For both the \texttt{BSM} model and the \texttt{BTM} models, the grid search was further extended to the meta features to be included in the model (all combinations of meta features) and to the size of the word embeddings for the Title, the Short Description, and the Long Description. Details about hyperparameter grids are reported in our code repository\footnote{\texttt{https://github.com/epfl-ml4ed/meta-transfer-learning}}. The selected hyperparameter combination is the one achieving the highest balanced accuracy (BAC) on a validation set composed of students in $S_{val} = \cup \; \tilde{S}^c \; \forall \; c \in C_{trn}$, with $\tilde{S}^c$ denoting $10\%$ randomly selected students for a given course $c$ (we stratified this selection on course and pass-fail label).

\section{Features}
\label{sec:features}
This appendix section will describe the behavioral features used in model training in more detail, as discussed in Sec. \ref{sec:features-descr}. Our four selected feature sets (\textit{Regularity}, \textit{Engagement}, \textit{Control}, and \textit{Participation}) are highlighted by previous work \cite{marras2021can} as strongly performant feature sets for MOOCs. In total, we have extracted the 45 features listed in Table \ref{tab:features}. The majority of the features come from the \textit{Control} and \textit{Engagement} feature sets. For the mathematical equations, we refer to the papers in which the features were initially introduced \cite{boroujeni2016quantify, lalle2020data, chen2020utilizing, marras2021can}. A detailed implementation of all features can be found in EPFL ML4ED's code repository\footnote{\texttt{https://github.com/epfl-ml4ed/flipped-classroom}}.

\chapter{Appendix: Evaluating the Explainers (Chapter 4)}
\label{sec:opt-preproc-2}
\section{Model Architecture}
We experimented with traditional machine learning models (e.g., Support Vector Machines, Logistic Regression, Random Forest) and deep-learning models (e.g., Fully-Connected Networks, RNNs, LSTMs, CNNs, and BiLSTMs), and found that BiLSTM models perform best against the other baselines for our use case. To determine the optimal model architecture, we evaluate validation set performance on the course \textit{DSP 1} as it is used in all three RQs. BiLSTMs have a $26.8\%$ average increase in balanced accuracy over traditional machine learning methods. For the BiLSTM architecture grid search, we examined the following layer settings \{32, 64, 128, 256, 32-32, 64-32, 128-32, 128-64, 128-128, 256-128, 64-64-32, 128-64-32, 64-64-64-32-32\} before determining 64-32 performed best in balanced accuracy for \textit{DSP 1}. We used the Tensorflow library to train our models \cite{abadi2016tensorflow}.

\section{Model Training}
Model training took approximately 35 minutes per model on an Intel Xeon E5-2680 CPU with 12 cores, 24 threads, and 256 GB RAM. Each model was trained for 15 epochs, and the best performing model checkpoint was saved. The five models' performance metrics are showcased in Table \ref{tab:modelperf}.

\section{Sampling Strategy}
We experimented with several strategies to extract a appropriate representative sample including the greedy algorithm SP-LIME \cite{lime}, random sampling, two sets of extreme students (those which the model predicts very well on and very badly on), and uniform sampling. We determined that our uniform sampling approach was the most fair with respect to the variable class imbalance between courses.

\begin{table}[!b]
\centering
\small
\resizebox{0.7\textwidth}{!}{
\begin{tabular}{lrrrr}
\toprule
\textbf{Identifier} & \textbf{Accuracy} & \textbf{Balanced Accuracy} & \textbf{F-1 Score} \\
\midrule
\textit{DSP 1} & 99.3 & 97.4 & 99.6 \\
\textit{DSP 2} & 99.1 & 93.5 & 99.5 \\
\textit{Geomatique} & 97.7 & 96.2 & 98.7 \\
\textit{Villes Africaines} & 98.4 & 95.5 & 99.1 \\
\textit{Micro} & 89.5 & 90.9 & 90 \\ 
\bottomrule
\end{tabular}}
\caption{Avg. performance of the BiLSTMs trained on the five MOOCs.}
\label{tab:modelperf}
\end{table}

\chapter{Appendix: MultiModN (Chapter 6)}
\section{\md Framework Implementation} 
\label{appendix:momonet}

A \textbf{task- and modality-agnostic open-source framework} \md solution has been implemented in Python using PyTorch as the primary machine learning framework. The \texttt{/multimodn} package contains the \md model and its components. The \texttt{/datasets} package is responsible of preparing the data inputs for \md. Some examples using the public Titanic dataset have been provided.

The code is available at: \url{https://github.com/epfl-iglobalhealth/MultiModN}.

% \subsection{Implementation Model}\label{model-architecture}
\begin{figure}[H]
\centering
\includegraphics[width=\textwidth]{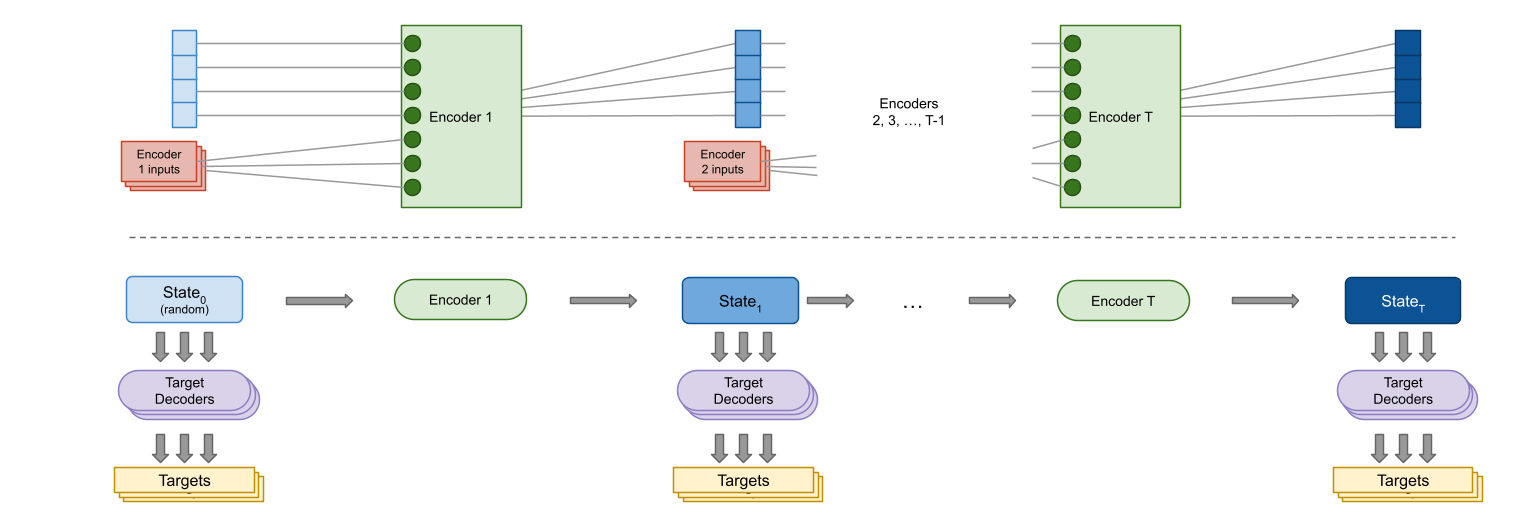}
    \caption{\textbf{Architecture of \md code}. The \textbf{upper panel} shows a more detailed depiction of sequential encoding using a series of model-agnostic  {\color{YellowGreen}{encoders}} which receive {\color{Bittersweet}{inputs}  of variable dimension } to create the evolving  {\color{NavyBlue}{state vector}}, which represents the shared feature space. The \textbf{lower panel} shows how each state can be probed by any number of {\color{Orchid}{target decoders}}.}
   \label{appendix:code_arch}
\end{figure}

\textbf{\md metrics}\label{multimodn-metrics}

During training and evaluation, the metrics of the model are stored in a log at each epoch in a matrix of dimensions \((E+1) * D\), where $E$ is the number of encoders and $D$ the number of decoders. Each row represents the metrics for a target at each state of the model.

\textbf{Code structure}\label{code-structure}

\textit{MultiModN}\label{multimodn-1}

\texttt{/multimodn} package contains the \md model and its modules:
\begin{enumerate}
\item Encoders: \texttt{/multimodn/encoders}
\item Decoders: \texttt{/multimodn/decoders}
\item State: \texttt{/multimodn/state.py}
\end{enumerate}

\textit{Datasets}\label{datasets}

\texttt{/dataset} package contains the MultiModDataset abstract class, compatible with \md.

Specific \textbf{datasets} are added in the \texttt{/dataset} directory and must fulfill the following requirements:
\begin{itemize}
\item Contain a dataset class that inherit MultiModDataset or has a method to convert into a MultiModDataset
\item Contain a \texttt{.sh} script responsible of getting the data and store it in \texttt{/data} folder
\end{itemize}

\texttt{\_\_getitem\_\_} function of MultiModDataset subclasses must yield elements of the following shape:

\begin{verbatim}
tuple
(
    data: [torch.Tensor],
    targets: numpy.ndarray,
    (optional) encoding_sequence: numpy.ndarray
)
\end{verbatim}

namely a tuple containing an array of tensors representing the features for each subsequent encoder, a numpy array representing the different targets and optionally a numpy array giving the order in which to apply the encoders to the subsequent data tensors. Note: \texttt{data} and \texttt{encoding\_sequence} must have the same length.

\textbf{Missing values. }\label{missing_values} The user is able to choose to keep missing values (\texttt{nan} values). Missing values can be present in the tensors yielded by the dataset and are managed by \md.

\hypertarget{pipelines}{%
\textit{Pipelines}\label{pipelines}}

\texttt{/pipeline} package contains the training pipelines using \md for Multimodal Learning. It follows the following steps:

\begin{itemize}
\item Create \texttt{MultiModDataset} and the \texttt{dataloader} associated
\item Create the list of encoders according to the features shape of the MultiModDataset
\item Create the list of decoders according to the targets of the MultiModDataset
\item Init, train and test the \md model
\item Store the trained model, training history and save learning curves
\end{itemize}

\textbf{Quick start}\label{quick-start}

Quick start running \md on Titanic example pipeline with a Multilayer Perceptron encoder:
\begin{verbatim}
./datasets/titanic/get_data.sh
python3 pipelines/titanic/titanic_mlp_pipeline.py
\end{verbatim}

Open \texttt{pipelines/titanic/titanic\_mlp.png} to look at the training
curves.

\section{Additional details about \md Architecture} 
\label{appendix:architecture}

\begin{figure}[H]
\centering
\includegraphics[width=\textwidth]{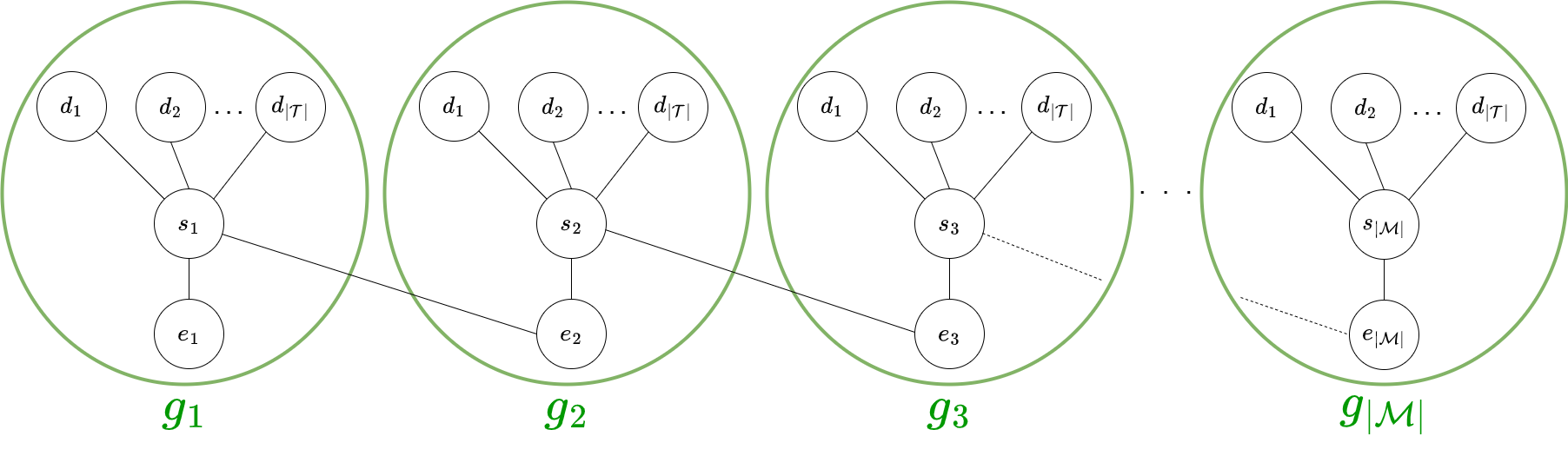}
    \caption{\textbf{Schematic representation of the \textit{modules} ($g$, groups) of \md}. $e$: encoders, $s$: state vector, $d$: decoders. Each module is connected by a single edge between $s_n$ and $e_{n+1}$. There are $\mathcal{|M|}$ groups (i.e. input-specific modules) and $\mathcal{|T|}$ decoders per module.}
   \label{appendix:modularity}
\end{figure}

\textbf{Modularity. } In the following, we detail the computation of \md's modularity measure. The total number of edges in a \md module is $|\mathcal{T}|+1$. The total number of modules is $|\mathcal{M}|$ and there are $|\mathcal{M}|-1$ edges connecting consecutive modules, which makes for a total number of edges in the entire \md model of $m=|\mathcal{M}|(|\mathcal{T}|+2)-1$.

To compute the modularity following using the formalization proposed by Newman et al~\cite{PhysRevE.69.026113}, we need to define groups. In the case of \md, each group corresponds to one \textit{module}. Let $G$ be the matrix whose components $g_{ij}$ is the fraction of edges in the original network that connect vertices in group $i$ to those in group $j$.

Within \md, each group contains $(|\mathcal{T}|+1)$ edges and is connected to adjacent groups by two edges, with the exception of $g_1$ and $g_{|\mathcal{M}|}$, which are connected to only one other group (cf. Figure~\ref{appendix:modularity}).

Thus, $G$ is a tridiagonal matrix whose diagonal elements $G_{ii}$ are equal to $(|\mathcal{T}|+1)/m$ and whose upper and lower diagonal elements are equal to $1/m$:
\begin{equation*}
G = \frac{1}{m} \begin{pmatrix}
    |\mathcal{T}|\!+\!1  & 1                    &                       & \\
    &&&&\\
    &&&&\\
    1                   & |\mathcal{T}|\!+\!1   & 1                     & \\
    &&&&&  &\\
    \mkern96mu \ddots  & \mkern96mu \ddots    & \mkern96mu \ddots    & \\
    &&&&\\
                        & \!\!1                     & |\mathcal{T}|\!+\!1   & 1  \\
    &&&&\\
    &&&&\\
                        &                       & \mkern-28mu 1                     & |\mathcal{T}|\!+\!1 
  \end{pmatrix}    
\end{equation*}

The modularity measure is defined by $\mathcal{Q} = \text{tr}(G) - \Vert G^2 \Vert$, where $\text{tr}(G)$, is the trace of $G$ and $\Vert G^2 \Vert$ is the sum of elements of $G^2$. The trace of $G$ is equal to $\frac{|\mathcal{M}|}{m}(\mathcal{T}+1)$ and we have $G^2$ equal to:

\begin{equation*}
\frac{1}{m^2} \begin{pmatrix}
    (|\mathcal{T}|\!+\!1)^2\!+\!1   & 2(|\mathcal{T}|\!+\!1) & 1 & &  &\\
    &&&&& & \\
    &&&&& & \\
    2(|\mathcal{T}|\!+\!1)      & (|\mathcal{T}|\!+\!1)^2\!+\!2   & 2(|\mathcal{T}|\!+\!1) & 1 &  &\\
    &&&&& & \\
    &&&&& & \\
    1      & 2(|\mathcal{T}|\!+\!1)      & (|\mathcal{T}|\!+\!1)^2\!+\!2   & 2(|\mathcal{T}|\!+\!1) & 1  &  &\\
    &&&&&  &\\
    \qquad\qquad\ddots  & \qquad\qquad\ddots    & \qquad\qquad\ddots    & \qquad\qquad\ddots &  \qquad\qquad\ddots &\\
    &&&&&  &\\
    % & \!1    &    2(|\mathcal{T}|\!+\!1) &  (|\mathcal{T}|\!+\!1)^2\!+\!2  & 2(|\mathcal{T}|\!+\!1) & 1  \\
    % &&&&& & \\
    % &&&&& & \\
    & \!1    &    2(|\mathcal{T}|\!+\!1) &  (|\mathcal{T}|\!+\!1)^2\!+\!2  & 2(|\mathcal{T}|\!+\!1)  \\
    &&&& & \\
    &&&& & \\
    &                       & \mkern-32mu 1                   & 2(|\mathcal{T}|\!+\!1) &   (|\mathcal{T}|\!+\!1)^2\!+\!1
  \end{pmatrix}    
\end{equation*}

Hence, we have:

\begin{equation*}
\begin{split}
\Vert G^2 \Vert &= \frac{1}{m^2} \left[ |\mathcal{M}|((|\mathcal{T}|+1)^2+2) - 2 + 2(|\mathcal{M}|-1)2(|\mathcal{T}|+1) + 2(|\mathcal{M}|-2)\right]\\
&=\frac{|\mathcal{M}||\mathcal{T}|^2 +6|\mathcal{M}||\mathcal{T}| +9|\mathcal{M}| -4|\mathcal{T}|-10}{|\mathcal{M}|^2|\mathcal{T}|^2 + 4 |\mathcal{M}|^2|\mathcal{T}| + 4|\mathcal{M}|^2| -2|\mathcal{M}||\mathcal{T}| -4|\mathcal{M}| + 1} \\
\end{split}
\end{equation*}
It is important to note that when $\mathcal{M}$ increases, the trace will tend to $(|\mathcal{T}|+1)/(|\mathcal{T}|+2) \approx 1$ for a large number of tasks. Moreover,  when  $\mathcal{|M|}$increases $||G^2||$ will decrease towards to 0. Thus, for a large number of tasks, we have a modularity measure that increases and tends to 1 with the number of modalities.

\begin{figure}[H]
\centering
\includegraphics[width=0.7\textwidth]{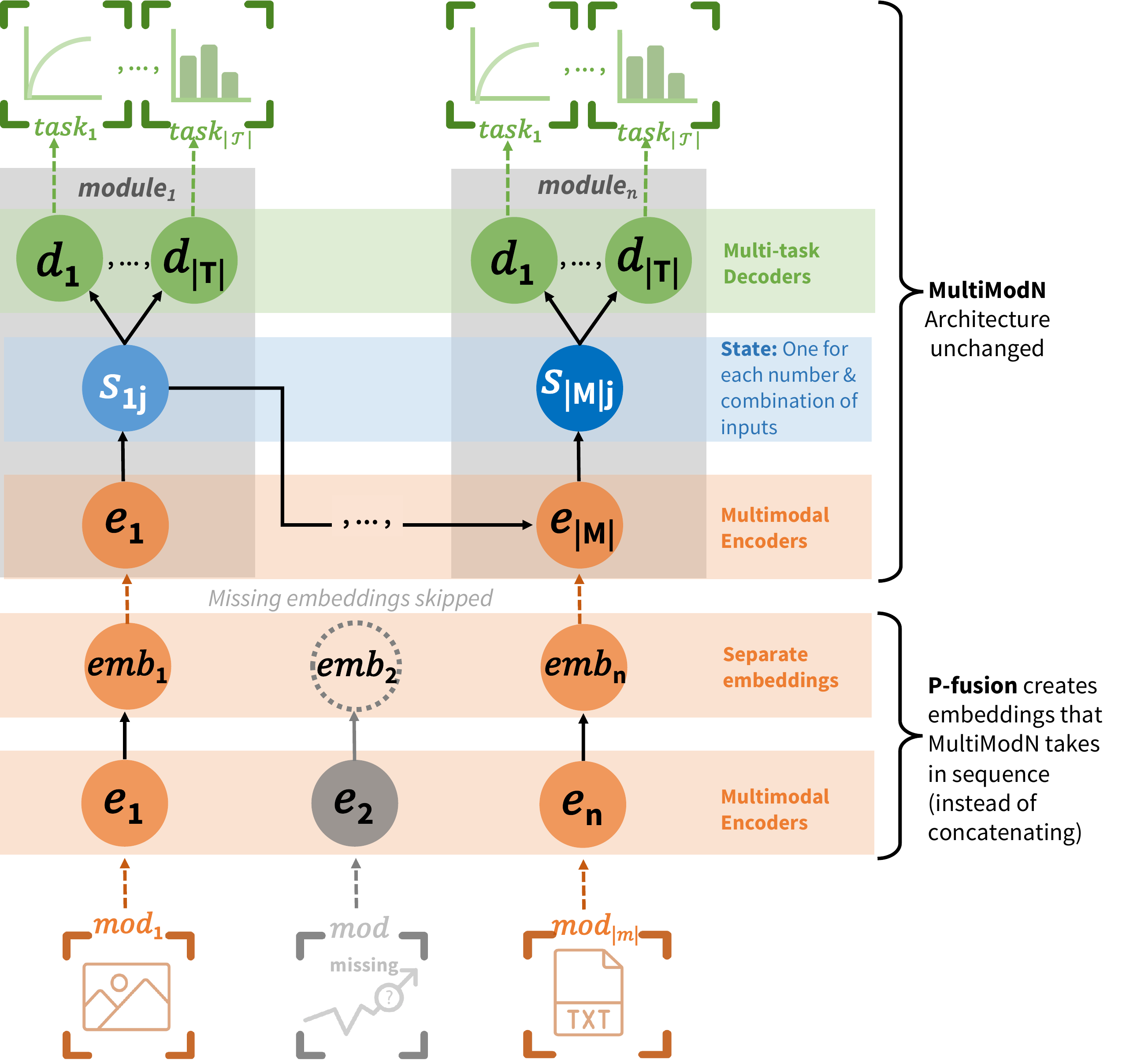}
    \caption{\textbf{Alignment of \pf and \md architectures.} We purposely ensure that the feature extraction pipelines are aligned between \pf and \md. To this end, we use the embeddings ($emb$) produced by \pf as inputs into the \md encoders ($e$). No element of \md is changed. \md encoders ($e$) are in orange, the \md state ($s$) is in blue and multi-task \md decoders ($d$) are in green.}
   \label{appendix:overview-aligned}
\end{figure}

\textbf{Architecture alignment between  \pf and \md. } We purposely align the feature extraction pipelines of \pf and \md in order to best isolate the effect of monolithic-parallel fusion vs. modular-sequential fusion. In Figure \ref{appendix:overview-aligned}, we see how the alignment is limited to the input, where both \md and \pf share the feature-extraction of each modality, where the \md encoders receive an embedding ($emb$). No element of \md is changed as described in Figure \ref{fig:overview}. Embeddings from missing data can be skipped (i.e. not encoded) by \md.

For MIMIC, feature extraction for each modularity is replicated from previous work \cite{haim2022} and we use embeddings generated from a set of pre-trained models. For Weather and Education, as there were no pre-existing embedding models, we design autoencoders trained to reconstruct the original input features from a latent space. We keep the autoencoder's encoder and decoder structure exactly aligned with the encoders of \md (two ReLU activated, fully-connected Dense layers and a third layer either generating the state representation with a ReLU activation or the final prediction with a sigmoid activation). We also align the number of trainable parameters with \md's modality-based encoders for a fair baseline comparison by selecting an appropriate state representation size per modality to equal the state representation in \md. The remaining hyperparameters are left exactly the same (batch size, hidden layer size, dropout rate, optimization metrics, loss function).

\section{Datasets and Tasks} 
\label{appendix:data-setup}

A description of all tasks is provided in Table \ref{tab:tasks}. In the following paragraphs, we detail the preprocessing decisions on the datasets for context and reproducibility.

\textbf{MIMIC. } The data includes four input modalities (tabular, textual, images, and time-series) derived from several sources for each patient. We align our preprocessing pipeline exactly with the study from which our baseline of \pf is derived \cite{haim2022} (described in \ref{sec:alignment}). To this end, we use patient-level feature embeddings extracted by the pre-trained models described in \cite{haim2022} and depicted in Figure~\ref{appendix:overview-aligned}. 

The dataset is a combination of two MIMIC databases: \href{https://physionet.org/content/mimiciv/1.0/}{MIMIC-IV v.1.0} \cite{mimic-iv} and \href{https://physionet.org/content/mimic-cxr-jpg/2.0.0/}{MIMIC-CXR-JPG v.2.0.0} \cite{johnson2019mimic}. After gaining authorized access to PhysioNet online repository \cite{goldberger2000physiobank}, the embedding dataset can be downloaded via this link: \url{https://physionet.org/content/haim-multimodal/1.0.1/}.
The dataset comprises $45,050$ samples, each corresponding to a time point during a patient's hospital stay when a chest X-ray was obtained. It covers a total of $8,655$ unique patient stays. To ensure data quality and limit our experiments to two thematic tasks (diagnosis of $task_1$: cardiomegaly and $task_2$: enlarged cardiomediastinum), we remove duplicates (based on image id and image acquisition time), and retain only relevant samples that have valid labels for both targets of interest, i.e. both $task_1$ and $task_2$ are either present (1) or absent (0). Subsequent experiments for these tasks are thus performed on the $921/45,050$ selected relevant patients.
%+ patient selection (stay)
%+ link embeddings

\textbf{EDU. } This dataset involves hand-crafted features extracted for $5,611$ students across 10 weeks of data. The preprocessing of data is an exact replication of several related works using the same dataset \cite{swamy2022meta, swamy2022evaluating, swamy2023trusting} based on 4 feature sets determined as predictive for MOOC courses in \cite{marras2021can}. 45 features regarding problem and video data are extracted per student per week, covering features like \textit{Delay Lecture}, which calculates the average delay in viewing video lectures after they are released to students or \textit{TotalClicksProblem}, the number of clicks that a student has made on problems this week. The features are normalized with min-max normalization, and missing values are imputed with zeros that have meaning i.e. no problem events in a week is correctly inferred as zero problems attempted. In this setting, missingness is a valued, predictive feature of the outcome, and thus we do not perform missingness experiments on this dataset. \md has the ability to select whether missingenss is encoded or not, and thus it would not suffer a disadvantage in a setting where missingness should be featurized.

In MOOCs, a common issue is that students join a course and never participate in any assignments, homeworks, or quizzes. This could be due to registering aspirationally, to read some material, or to watch videos \cite{onah2014dropout}. An instructor can easily classify students who have never completed an assignment as failing students. As introduced by Swamy et al. in \cite{swamy2022meta} and used with this dataset in related work \cite{swamy2022evaluating, swamy2023trusting, asadi2022ripple}, EDU has removed students that were predicted to fail in the first two weeks simply by having turned in no assignments (99\% confidence of failing with an out-of-the-box logistic regression model, where the confidence threshold was tuned over balanced accuracy calculations). It has been shown that including these students will artificially increase the performance of the model, providing even better results than those showcased by \md in this work \cite{swamy2022meta}. We thus exclude these students to test a more challenging modeling problem.

\textbf{Weather. } The Weather2k dataset, presented in \cite{zhu2023weather2k}, covers features from $1,866$ weather stations with 23 features covering seven different units of measurements (degrees, meters, HPA, celsius, percentage, $ms^{-1}$, millimeters). To align these features on vastly different scales, we normalize the data. We use the large extract (R) provided by the authors instead of the smaller representative sample also highlighted in the benchmark paper (S) \cite{zhu2023weather2k}. We use the first 24 hourly measurements as input to train the \md model and calculate the five regression tasks as determined in Table \ref{tab:tasks} below.

\textbf{Tasks. } As showcased in Table \ref{tab:tasks}, our evaluation covers 10 binary and regression tasks in two settings: static (one value per datapoint) or continuous (changing values per datapoint, per timestep).

\begin{table}[H]
\resizebox{\textwidth}{!}{%
\begin{tabular}{@{}ccccl@{}}
\toprule
                         & \textbf{Task} & \textbf{Type}                                                    & \textbf{Name}                                                                    & \multicolumn{1}{c}{\textbf{Description}}                                                                                                                                                                                                                                                                                                                                                                                                                                                         \\ \midrule
\multirow{2}{*}{\textbf{MIMIC}}   & \textbf{1}             & \begin{tabular}[c]{@{}c@{}}Static \\ Binary\end{tabular}         & Cardiomegaly                                                                     & \begin{tabular}[c]{@{}l@{}}Labels determined as per \cite{haim2022} using NegBio \cite{peng2018negbio} and CheXpert \cite{irvin2019chexpert} \\to process radiology notes, resulting in four diagnostic outcomes: positive, negative, uncertain, or missing.\end{tabular}                                                                                                                                                                                                                           \\ \cmidrule(l){2-5} 
                         & \textbf{2}             & \begin{tabular}[c]{@{}c@{}}Static \\ Binary\end{tabular}         & \begin{tabular}[c]{@{}c@{}}Enlarged \\ Cardiomediastatinum\end{tabular}          & \begin{tabular}[c]{@{}l@{}}Labels determined as per \cite{haim2022} using NegBio \cite{peng2018negbio} and CheXpert \cite{irvin2019chexpert}. \\ The set of label values is identical to the one for cardiomegaly.\end{tabular}                                                                                                                                                                                                                                                                  \\ \midrule
\multirow{3}{*}{\textbf{EDU}}     & \textbf{3}             & \begin{tabular}[c]{@{}c@{}}Static \\ Binary\end{tabular}         & \begin{tabular}[c]{@{}c@{}}Student Success\\ Prediction\end{tabular}             & End of course pass-fail prediction (per student) as per \cite{swamy2022meta} on the course.                                                                                                                                                                                                                                                                                                                                                                                                      \\ \cmidrule(l){2-5} 
                         & \textbf{4}             & \begin{tabular}[c]{@{}c@{}}Continuous \\ Binary\end{tabular}     & \begin{tabular}[c]{@{}c@{}}Student Dropout\\ Prediction\end{tabular}             & \begin{tabular}[c]{@{}l@{}}1 if student has any non-zero value on a video or problem feature from next week until the end of the course, 0 if not. \\ Not valid for the last week, so the task involves n-1 decoder steps for n timesteps. Can be easily extended \\ to a multiclass task by separating video or problem involvement until the end of the course into separate classes.\end{tabular}                                                                                             \\ \cmidrule(l){2-5} 
                         & \textbf{5}             & \begin{tabular}[c]{@{}c@{}}Continuous \\ Regression\end{tabular} & \begin{tabular}[c]{@{}c@{}}Next Week \\ Performance \\ Forecasting\end{tabular}  & \begin{tabular}[c]{@{}l@{}}Moving average (per student, per week) of three student performance features from \cite{marras2021can} and removed \\in baseline paper \cite{swamy2022meta}: \textit{Student Shape} (recieving the maximum quiz grade on the first attempt), \\\textit{CompetencyAlignment} (number of problems the student has passed this week), \textit{CompetencyStrength} \\(extent to which a student passes a quiz getting the maximum grade with few attempts). \end{tabular} \\ \midrule
\multirow{5}{*}{\textbf{Weather}} & \textbf{6}             & \begin{tabular}[c]{@{}c@{}}Continuous \\ Regression\end{tabular} & \begin{tabular}[c]{@{}c@{}}Short Term \\ Temperature \\ Forecasting\end{tabular} & Changing air temperature measurements (collected per station, per hour), shifted by 24 hourly measurements (1 day).                                                                                                                                                                                                                                                                                                                                                                              \\ \cmidrule(l){2-5} 
                         & \textbf{7}             & \begin{tabular}[c]{@{}c@{}}Continuous \\ Regression\end{tabular} & \begin{tabular}[c]{@{}c@{}}Mid Term \\ Temperature \\ Forecasting\end{tabular}   & Changing air temperature measurements (collected per station, per hour), shifted by 72 hourly measurements (3 days).                                                                                                                                                                                                                                                                                                                                                                             \\ \cmidrule(l){2-5} 
                         & \textbf{8}             & \begin{tabular}[c]{@{}c@{}}Continuous \\ Regression\end{tabular} & \begin{tabular}[c]{@{}c@{}}Long Term \\ Temperature\\  Forecasting\end{tabular}  & Changing air temperature measurements (collected per station, per hour), shifted by 720 hourly measurements (30 days).                                                                                                                                                                                                                                                                                                                                                                           \\ \cmidrule(l){2-5} 
                         & \textbf{9}             & \begin{tabular}[c]{@{}c@{}}Static \\ Regression\end{tabular}     & Relative Humidity                                                                & \begin{tabular}[c]{@{}l@{}}Instantaneous humidity relative to saturation as a percentage at 48h from 2.5 meters above the ground,\\ as used as a benchmark forecasting task in \cite{zhu2023weather2k}.\end{tabular}                                                                                                                                                                                                                                                              \\ \cmidrule(l){2-5} 
                         & \textbf{10}            & \begin{tabular}[c]{@{}c@{}}Static \\ Regression\end{tabular}     & Visibility                                                                       & \begin{tabular}[c]{@{}l@{}}10 minute mean horizontal visibility in meters at 48 hr from 2.8 meters above the ground, \\ as used as a benchmark forecasting task in \cite{zhu2023weather2k}\end{tabular}                                                                                                                                                                                                                                                                             \\ \bottomrule \\
\end{tabular}%
}
\caption{\label{tab:tasks} Description of all 10 tasks used to evaluate \md.}
\label{tab:tasks}
\end{table}

Note that for $task_{4}$, the three features used to calculate next week's performance were not included in the original input features because of possible data leakage, as student performance on quizzes directly contributes to their overall grade (Pass/Fail).

For the AUROC curves on $tasks_{9-10}$ in Figure \ref{fig:all-rocs}, we conduct a binarization of the two last regression tasks. To align with the regression task, we conduct a static forecasting prediction per station for the relative humidity or visibility with a time window of 24h for the 48th timestep (one day in advance). While \md is capable of making a prediction at each continuous timestep, \pf is not able to do this without a separate decoder at each timestep, and therefore to compare the tasks we must choose a static analysis. We choose a threshold based on the normalized targets: 0.75 for humidity and 0.25 for visibility (selected based on the distribution of the feature values for the first 1000 timesteps), and evaluate the predictions as a binary task over this threshold. 

Analogously, in the Section \ref{appendix:experiments} analysis below on the binarization of the remaining regression tasks, we express the continuous regression tasks for temperature forecasting $tasks_{6-8}$ as a static binary task. To do this, we evaluate the prediction from the full window (24th timestep) at the respective forecasting timestep (48, 96, 744). The threshold we select is 0.3, closely corresponding to the normalized mean of the temperature (0.301) over the first 1000 timesteps.

\section{Model Optimization} 
\label{appendix:optimization}

Table \ref{tab:hyperparams} indicates the chosen hyperparameters for the experiments conducted in Section \ref{sec:experiments} of the paper, selected based on the optimal hyperparameters for the multi-task settings (all the tasks for a dataset predicted jointly). The single tasks use the same hyperparameters as the multi-task settings. For saving the best model across training epochs in the time series settings (EDU, Weather), our optimization metric saved the ones with best validation set results on the most short term task. Therefore, for EDU, we use MSE of $task_5$, predicting next week performance, and $task_6$ for Weather, forecasting temperature within a day). The intuition is that choosing the best model on the short term task would allow the model to emphasize stronger short-term connections, which in turn would improve long term performance.

% Please add the following required packages to your document preamble:
% \usepackage{booktabs}
% \usepackage{graphicx}
\begin{table}[H]
\centering
\resizebox{0.7\textwidth}{!}{%
\begin{tabular}{@{}cllllll@{}}
\toprule
\textbf{Task}    & \textbf{\begin{tabular}[c]{@{}l@{}}\# of\\ Timesteps\end{tabular}} & \multicolumn{1}{c}{\textbf{\begin{tabular}[c]{@{}c@{}}Batch \\ Size\end{tabular}}} & \multicolumn{1}{c}{\textbf{\begin{tabular}[c]{@{}c@{}}Dropout\\ Rate\end{tabular}}} & \multicolumn{1}{c}{\textbf{\begin{tabular}[c]{@{}c@{}}Hidden\\ Layer\\ Size\end{tabular}}} & \multicolumn{1}{c}{\textbf{\begin{tabular}[c]{@{}c@{}}State \\ Rep.\\ Size\end{tabular}}} & \textbf{\begin{tabular}[c]{@{}l@{}}Save Best Model \\ (chosen metric)\end{tabular}} \\ \midrule
1                & 1                                                                  & 16                                                                                 & 0.2                                                                                 & 32                                                                                         & 50                                                                                        & $task_{1}$ Val BAC + Macro AUROC                                                    \\
2                & 1                                                                  & 16                                                                                 & 0.2                                                                                 & 32                                                                                         & 50                                                                                        & $task_{2}$ Val BAC + Macro AUROC                                                    \\
\textbf{MIMIC}   & 1                                                                  & \textbf{16}                                                                        & \textbf{0.2}                                                                        & \textbf{32}                                                                                & \textbf{50}                                                                               & \textbf{$tasks_{1-2}$ BAC + Macro AUROC}                                            \\ \midrule
3                & 10                                                                 & 64                                                                                 & 0.1                                                                                 & 32                                                                                         & 20                                                                                        & $task_{3}$ Val Accuracy                                                             \\
4                & 10                                                                 & 64                                                                                 & 0.1                                                                                 & 32                                                                                         & 20                                                                                        & $task_{4}$ Val Accuracy                                                             \\
5                & 10                                                                 & 64                                                                                 & 0.1                                                                                 & 32                                                                                         & 20                                                                                        & $task_{5}$ Val MSE                                                                  \\
\textbf{EDU}     & \textbf{10}                                                        & \textbf{64}                                                                        & \textbf{0.1}                                                                        & \textbf{32}                                                                                & \textbf{20}                                                                               & \textbf{$task_{5}$ Val MSE}                                                         \\ \midrule
6                & 24                                                                 & 128                                                                                & 0.1                                                                                 & 32                                                                                         & 20                                                                                        & $task_{6}$ Val MSE                                                                  \\
7                & 24                                                                 & 128                                                                                & 0.1                                                                                 & 32                                                                                         & 20                                                                                        & $task_{7}$ Val MSE                                                                  \\
8                & 24                                                                 & 128                                                                                & 0.1                                                                                 & 32                                                                                         & 20                                                                                        & $task_{8}$ Val MSE                                                                  \\
9                & 24                                                                 & 128                                                                                & 0.1                                                                                 & 32                                                                                         & 20                                                                                        & $task_{9}$ Val MSE                                                                  \\
10               & 24                                                                 & 128                                                                                & 0.1                                                                                 & 32                                                                                         & 20                                                                                        & $task_{10}$ Val MSE                                                                 \\
\textbf{Weather} & \textbf{24}                                                        & \textbf{128}                                                                       & \textbf{0.1}                                                                        & \textbf{32}                                                                                & \textbf{20}                                                                               & \textbf{$task_{6}$ Val MSE}                                                         \\ \bottomrule \\
\end{tabular}%
}
\caption{Hyperparameters selected for each experiment. We tuned the hyperparameters for the multi-task models (MIMIC, EDU, Weather) and used the same hyperparameters for each single-task model for a fair comparison.}
\label{tab:hyperparams}
\end{table}

In Figure \ref{appendix:architecture}, we examine a case study of \md's changing performance on $task_{3}$, student success prediction in the EDU dataset, by varying hyperparameters across three model architectures (chosen for small, medium, and large hyperparameter initializations). We note that batch size is fairly robust across all three initial model settings, with large batch size on the largest model having slightly variable performance. Examining changing dropout rate, we note that with medium and large models, change in dropout impacts performance considerably. This allows us to hypothesize that high dropout on larger state representations does not allow the model to learn everything it can from the data. 

Looking at varied hidden layer size, we see comparable performance for the small and medium initializations, but note that in the large case, having a smaller hidden layer size is important to maintain performance. Even as \md performance trends upwards with larger hidden layer size (i.e. 128) for the large initialization, the confidence interval is large, so performance is not stable. Lastly, observing state representation size, we see that when state representation is too small for the task (i.e. 1, 5), the small and large models are adversly impacted. Additionally, when state representation is too large (i.e. 100), performance seems to drop or increase variability again. It is therefore important to tune \md and find the right state representation size for the dataset and predictive task(s).

\begin{figure}[H]
\centering
\includegraphics[width=\textwidth, trim={10 0 10 0}, clip]{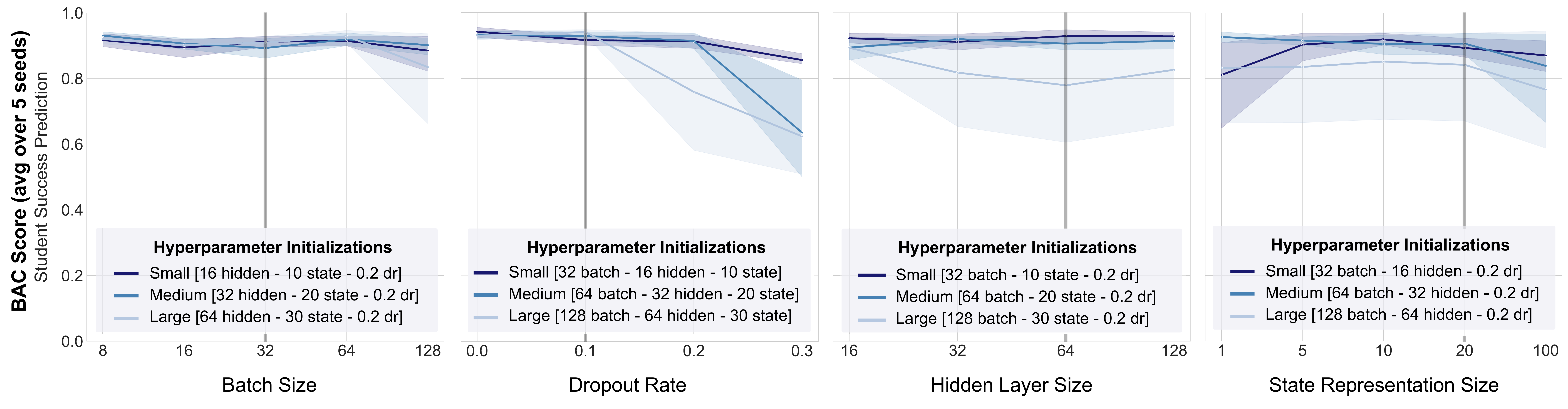}
\caption{\label{appendix:perf-analysis}\md hyperparameter selection across four parameters on $task_3$ of the EDU dataset (Pass/Fail). Each individual parameter is varied on the x-axis (\textit{\textbf{dr}: dropout rate}) with all other initializations fixed (grouped in small, medium, and large values). These are compared in terms of balanced accuracy (BAC). 95\% CIs are shaded.}
\end{figure}

\textbf{Experimental Setup. } For the results reported in Sections \ref{exp:single-task}, \ref{exp:multi-task} and \ref{sec:missingness}, we perform 5-fold stratified cross-validation with 80-10-10 train-validation-test split. Due to the time-series nature of the EDU and Weather datasets, we orient the stratification the real labels associated with the longest-term task. For EDU, $task_{3}$ (end of course pass-fail prediction) and for Weather $task_{8}$ (30-day temperature forecasting) is chosen for the stratification split. 
\begin{figure}[ht!]
\begin{center}
\includegraphics[width = .42\textwidth]{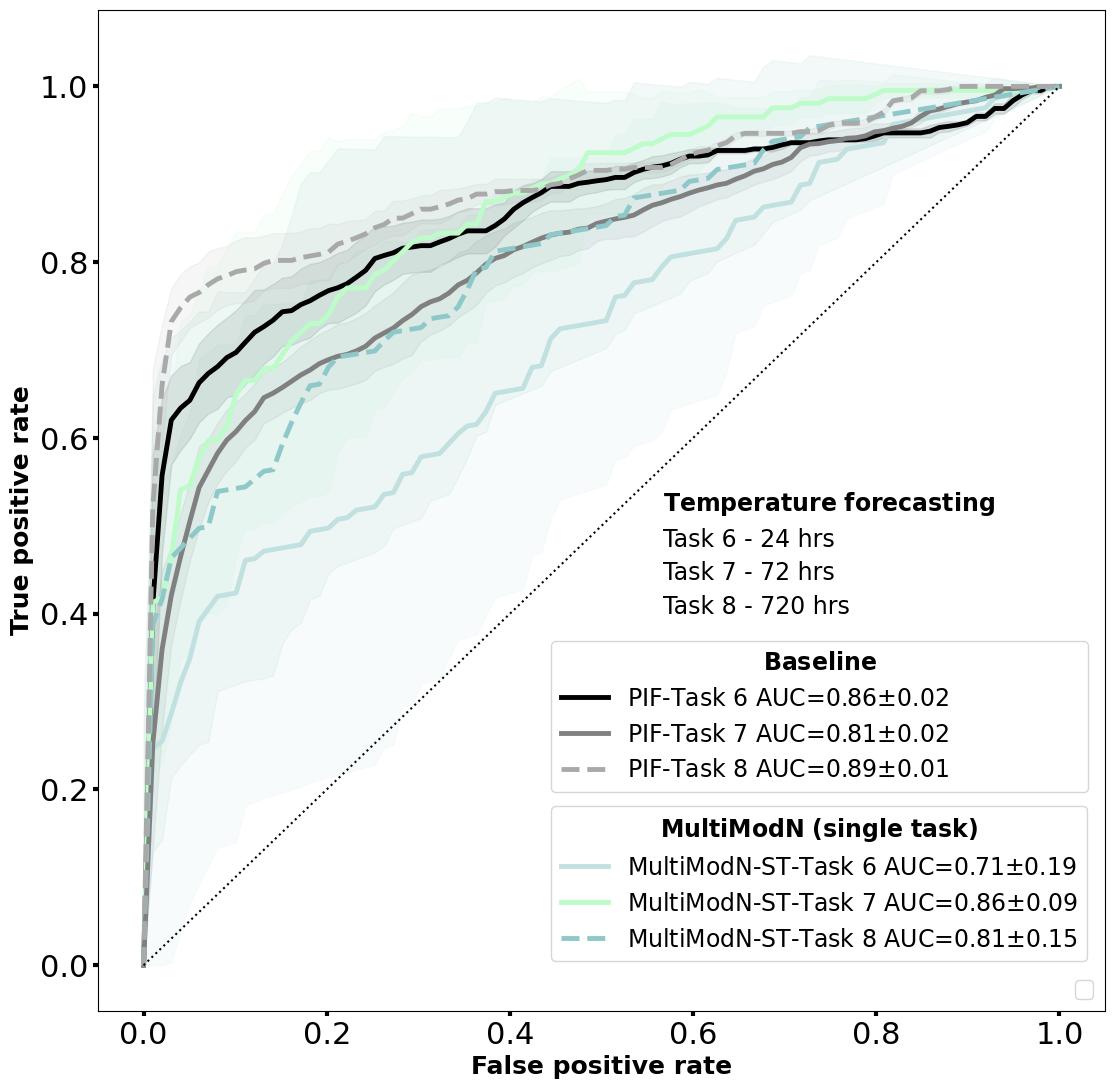}
\end{center}
\caption{ \label{fig:all-roc-addtl} AUROC for three additional binary prediction tasks in Weather2k. Targets predicted by {\color{gray}{\pf}} are compared to \md. 95\%CIs are shaded.}
\end{figure}

In an alternative approach, regarding the MIMIC dataset, a two-step procedure was implemented to address the imbalanced class ratios, given the absence of a prioritized task. Initially, a new dummy label was assigned to each sample, indicating positivity if both pathologies are present and negativity otherwise. Subsequently, a label was assigned to each unique hospital stay based on the aggregated labels from the first step. A hospital stay was considered positive if the number of times sample from that stay has been found positive is greater than or equal to the half of samples with the same hospital stay ID. The latter, as outlined in \cite{haim2022}, ensured that no information was leaked on the hospital stay level during stratification.

Experiments were conducted using the same architecture in PyTorch (MIMIC) and TensorFlow (EDU, Weather), to provide multiple implementations across training frameworks for ease of use. We use an Adam optimizer with gradient clipping across all experiments.

\section{Additional Experiments} 
\label{appendix:experiments}
\textbf{Single task}

We present the binarized results (AUROC curves) for several additional regression tasks ($tasks_{6-8}$ for Weather) in Figure \ref{fig:all-roc-addtl}. The specific details of binarization are discussed above in Appendix Section \ref{appendix:data-setup}. We note that the confidence intervals overlap for \pf and \md over all Weather tasks. However, it is clear that the performance of \md varies a lot (large CIs). This could be due to the design of the binarization as originally in the benchmark paper \cite{zhu2023weather2k}, this was introduced as a regression task. Another contributing factor to large CIs could be that the model was trained across all timesteps but only evaluated on one timestep for a comparable binarization. Despite these caveats, we can statistically conclude that \pf and \md performance are comparable on these additional tasks.

\textbf{Interpretability} \label{appendix:interpretability}
We perform an interpretability analysis for the EDU dataset, analogous to the local and global interpretability analysis on MIMIC from Figure \ref{fig:inference-main}. The global analysis (IMC) is conducted over all students for the first week of the course. We note interesting findings: specifically that problem interactions are more important for $tasks_{3-4}$ while video interactions are more important for $task_{5}$. 

\begin{figure}[H]
\centering
\includegraphics[width = .9\textwidth]{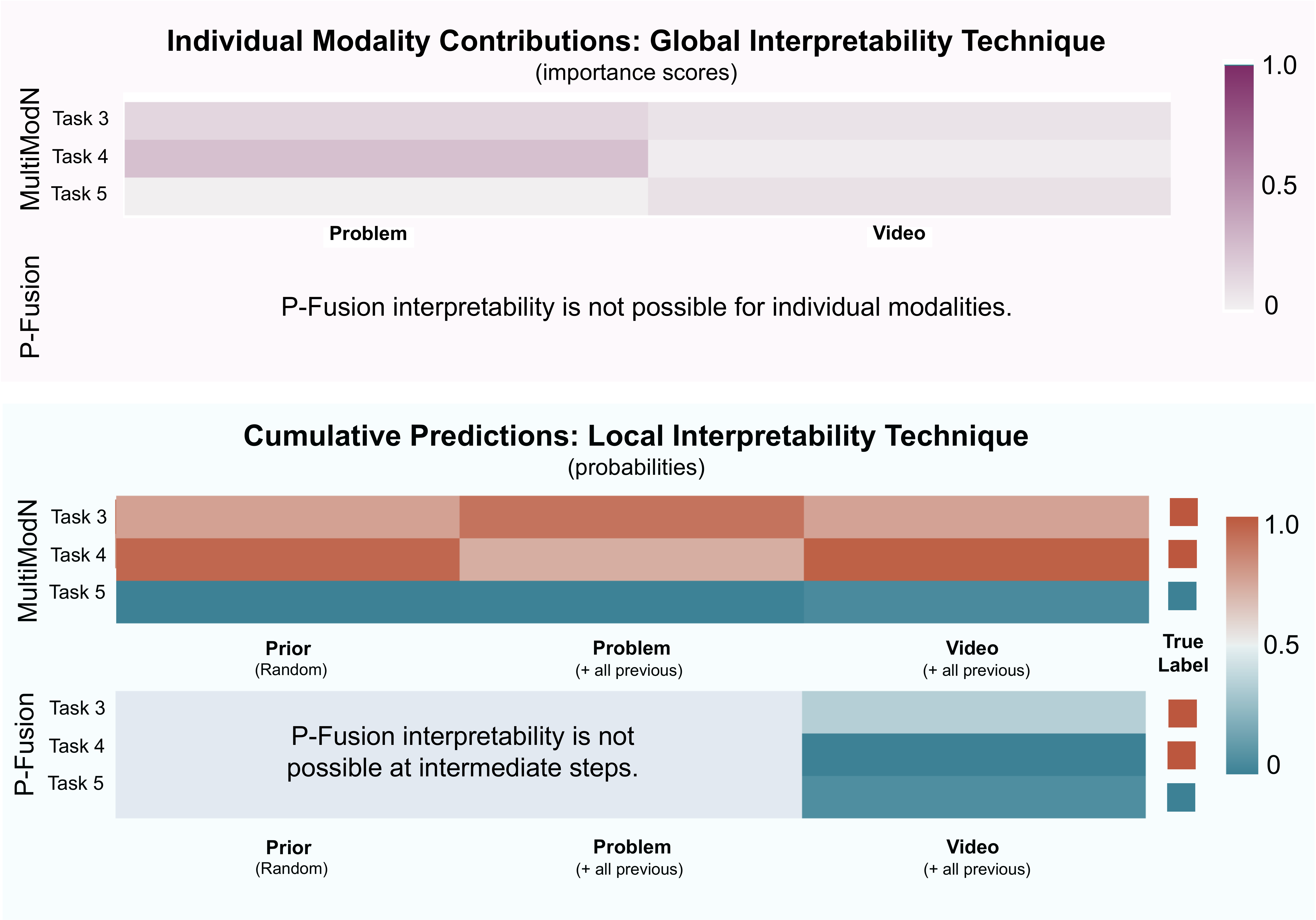}
\caption{\label{fig:interpret-edu} \textbf{Inherent modality-specific model explainability in \md for $tasks_{3-5}$. } Heatmaps show individual modality contributions (IMC) \textbf{(top)} and cumulative contributions (CP) \textbf{(bottom)}: respectively {\color{Mulberry}\textbf{importance score}} (global explainability) or {\color{RoyalBlue}\textbf{cumulative probability}} (local explainability). The multi-task \md for $task_{3-5}$ in EDU is compared to two single-task \pf models. IMC are only possible for \md (only 1 modality encoded, rest are skipped). CP are made sequentially from states encoding all previous modalities. \pf is unable to naturally decompose modality-specific contributions (can only make predictions once all modalities are encoded). IMC is computed across all students in the test set. CP is computed for a single student, (true label = 1 for $tasks_{3-4}$ and 0 for $task_{5}$). The CP heatmap shows probability ranging from {{\color{RoyalBlue}{\textbf{confident negative diagnosis (0)}}} to {\color{Gray}{\textbf{perfect uncertainty}}} and \color{Bittersweet}{\textbf{confident positive diagnosis (1)}}}.}\end{figure}

The student selected for the CP local analysis passes the course (1 for $task_{3}$) and does not dropout (1 for $task_{4}$), but does not have strong performance in the next week ($\sim$ 0 in $task_{5}$). This analysis is also conducted across the first week of course interactions. We see that \pf cannot produce modality-specific interpretations and predicts the incorrect label. However, \md is able to identify a changing confidence level across modalities, eventually ending on the right prediction for all tasks. The confidence for $task_{3}$ increases with the student's problem interactions and reduces for their video interactions. This could have potential for designing an intervention to improve student learning outcomes. We note that for $task_{5}$, both the students' problem and video interactions contribute similarly to the prediction.

\begin{figure}[H]
\centering
\includegraphics[width=\textwidth]{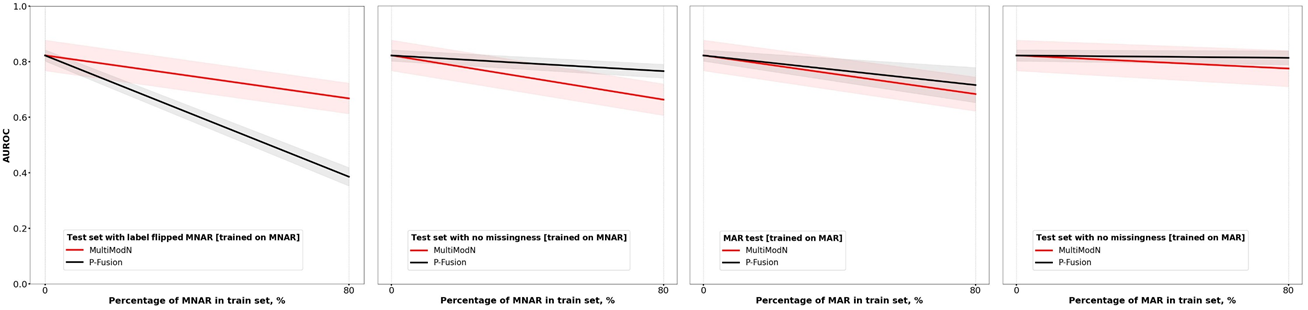}    
\caption{\label{fig:missing-1} \textbf{Detailed missingness experiments for $task_1$ (Cardiomegaly).} \pf (black) and \md (red) are trained on MIMIC data where various percentages of a single modality are missing (0 or 80\%) either for a single class (MNAR, first two plots) or without correlation to either class (MAR, last two plots). The AUROCs are shown for each when evaluated on test sets which either have a risk of catastrophic failure (first plot, MNAR with label flip) or on test sets without missingness or MAR missingness. CI95\% shaded.}
\end{figure}

\begin{figure}[H]
\centering

\includegraphics[width=\textwidth]{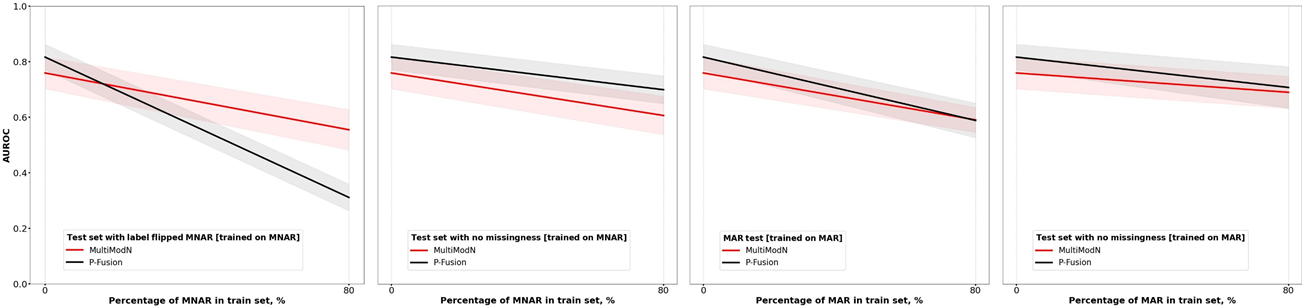}    
\caption{\label{fig:missing-2} \textbf{Detailed missingness experiments for $task_2$ (Enlarged Cardiomediastinum).} \pf (black) and \md (red) are trained on MIMIC data where various percentages of a single modality are missing (0 or 80\%) either for a single class (MNAR, first two plots) or without correlation to either class (MAR, last two plots). The AUROCs are shown for each when evaluated on test sets which either have a risk of catastrophic failure (first plot, MNAR with label flip) or on test sets without missingness or MAR missingness. CI95\% shaded.}
\end{figure}

\textbf{Missingness} \label{appendix:missingness}
We expand on the missingness experiments presented in \ref{sec:missingness}. Here, we present further control experiments (training on data missing-at-random, MAR) in both MIMIC tasks ($task_1$: diagnosis of Cardiomegaly \ref{fig:missing-1} and $task_2$: diagnosis of Enlarged Cardiomediastinum \ref{fig:missing-2}). 

In the first two subplots of each figure, both \pf (black) and \md (red) are trained on MNAR data and then evaluated either on a test set at risk of catastrophic failure (where the pattern of MNAR is label-flipped, first figure) or on a test set with no missingness. 
As can be seen in the first figures, \pf suffers catastrophic failure in the MNAR flip, becoming worse than random when a single modality is missing at 80\%, as opposed to \md, which only decreases AUROC about 10\%.
When the test set has no missing values, \pf and \md are not significantly different, proving that the catastrophic failure of \pf is due to MNAR. 
This is further confirmed in the last two plots of each task, where the models are trained on MAR data and evaluated on test sets either without missing values or MAR missingness. 

\textbf{Comparison to a \pf Transformer}
\label{appendix:transformer}
Additional experiments with a Transformer have been conducted on 10 tasks across three datasets. Results are showcased below in two tables (left for $tasks_{1-4}$ and right for $tasks_{5-10}$) with 95\% CIs. 

The hyperparameter-tuned architecture (based on head size, number of transformer blocks, MLP units) for EDU and Weather is a transformer model with 4 transformer blocks, 4 heads of size 256, dropout of 0.25, MLP units 128 with dropout 0.4, batch size 64, trained for 50 epochs with cross-entropy loss. For MIMIC, the most performant (tuned) transformer architecture includes 2 transformer blocks with 3 heads of size 128, MLP units 32 with batch size 32. We train this architecture on each decoder task individually and all tasks together for a total of 13 new models with the exact preprocessing steps as in the \pf and \md  experiments. The results indicate that \md often outperforms or at least matches the \pf Transformer benchmark in the vast majority of single task and multi-task settings, and comes with several interpretability, missingness, and modularity advantages. Specifically, using the primary metric for each task (BAC for the classification tasks and MSE for the regression tasks), \md beats the Transformer baseline significantly in 7 tasks, overlaps 95\% CIs in 11 tasks, and loses very slightly (by 0.01) in 2 regression tasks.

\begin{figure}[H]
\centering
\begin{minipage}{.5\textwidth}
  \centering
  \includegraphics[width=0.97\textwidth]{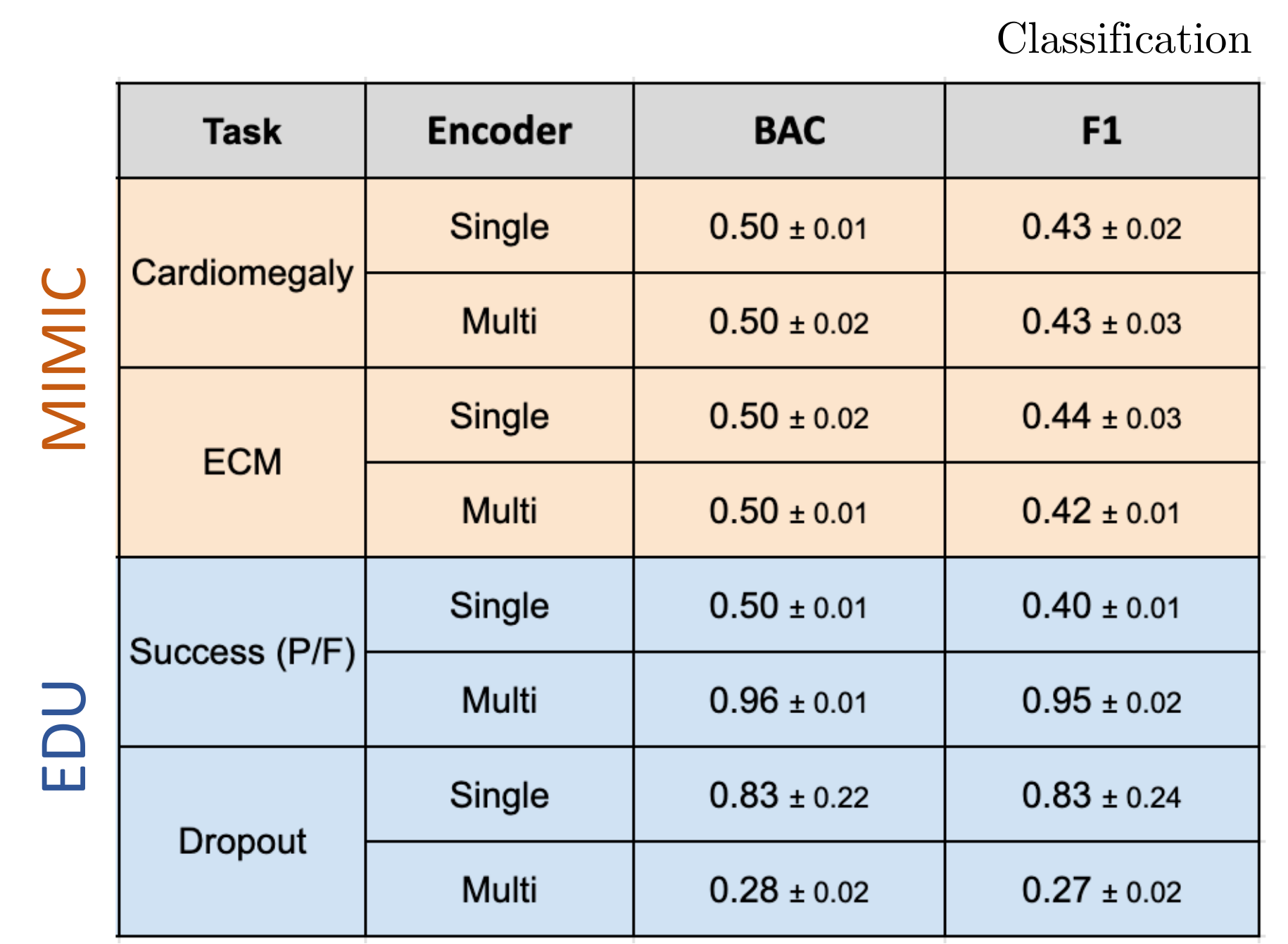}
\end{minipage}%
\begin{minipage}{.5\textwidth}
  \centering
  \includegraphics[width=0.97\textwidth]{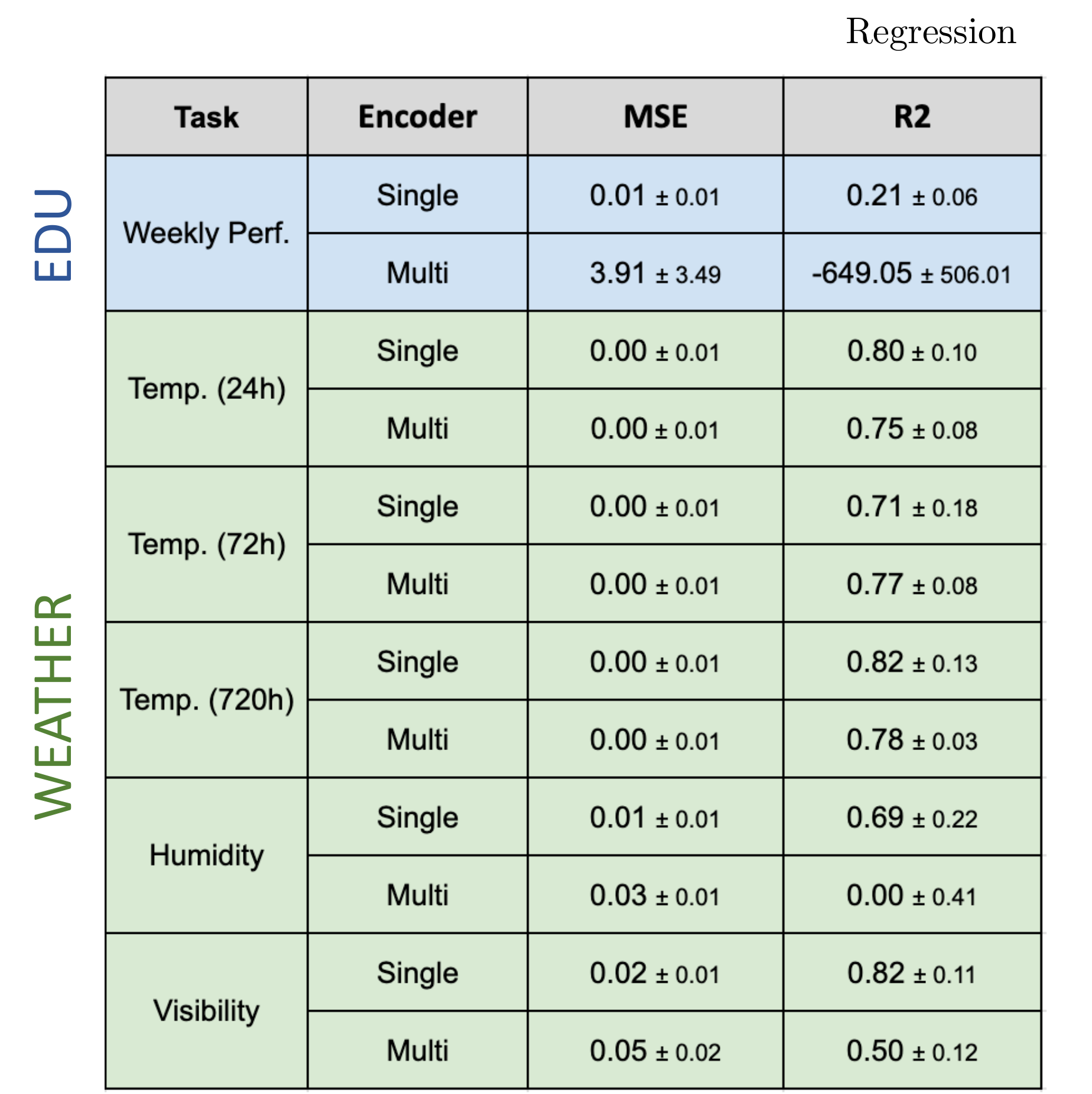}
\end{minipage}
  \caption{\textbf{Performance of the \pf Transformer on 10 classification and regression tasks across 3 datasets.} Results are showcased with 95\% confidence intervals. BAC and MSE are the primary evaluation metrics for classification and regression respectively.}
\end{figure}

\textbf{Additional Inference Settings}
\label{appendix:inference}

\begin{figure}[htbp]
\centering

\includegraphics[width=\textwidth]{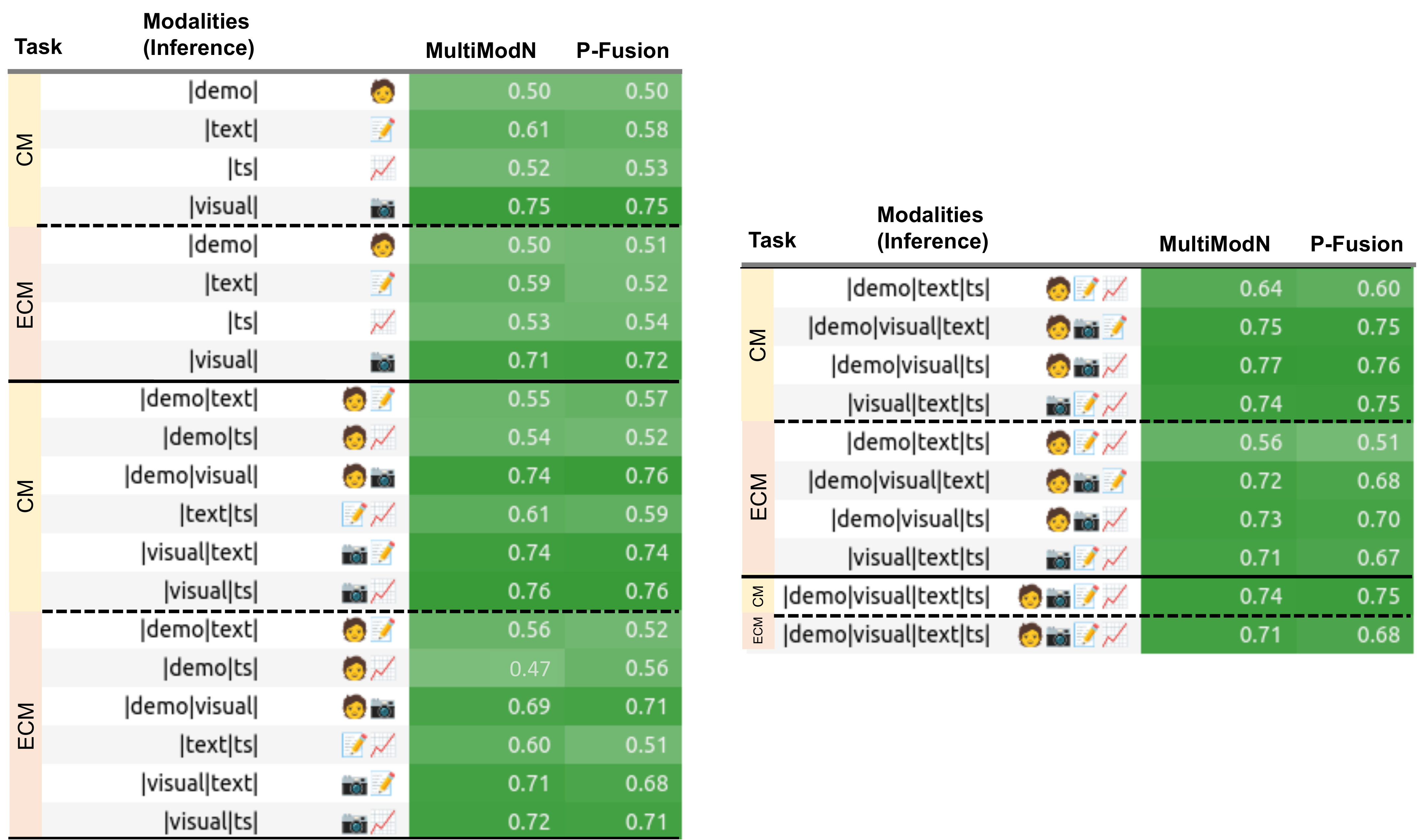}    
\caption{\label{fig:inference-appendix} \textbf{Detailed modality inference experiments for \md in comparison to \pf.} In these experiments, different  combinations of modalities and orderings at the time of inference are used for the two tasks in the MIMIC dataset. All 95\% CIs overlap between the two models.}
\end{figure}

To provide insight into performance gains, we performed additional experiments to showcase the benefits of modularity with vastly different training and inference settings. The results of 30 new experiments of inference encoders, each performed with 5-fold cross-validation are included in Figure \ref{fig:inference-appendix}. We compare \pf and \md on both tasks of the MIMIC dataset using all possible combinations of four input modalities at test time. \md ignores missing modalities whereas \pf imputes and therefore encodes missing modalities.

We note that the performance at inference for \pf and \md has no significant differences for all experiments (using 95\% CIs). Figure \ref{fig:inference-appendix} shows that, on average, \pf tends to overfit more to the most dominant (visual) modality. When this modality is missing (at random or completely at random), \md performs better on a combination of the remaining modalities (demo, text, time series). In the case of missing modalities, the observed effect in Figure \ref{fig:inference-appendix} is weak – confidence intervals overlap. Considering the MNAR (missing not-at-random) scenario described in Sec. \ref{sec:missingness}, the difference becomes significant.

\chapter{Appendix: InterpretCC (Chapter 7)}
\section{Taxonomy of Explanation Design Criteria}
\label{sec:related-work}

\textcolor{black}{We include a detailed taxonomy of terms used in Table \ref{tab:approaches}, inspired by \cite{speith2022review,schwalbe2024comprehensive, swamy2023future, pinto2024towards}.}

\begin{enumerate}[label=\textbf{\arabic*.}, leftmargin=1.5em]
    \item \textcolor{black}{\textbf{Granularity:} The granularity (small details or broad concepts) of the explanation \cite{miller2019explanation, swamy2023future, pinto2024towards}.}
    \begin{enumerate}[label*=\arabic*., leftmargin=1.5em]
        \item \textcolor{black}{\textbf{Feature:} Explanations are made using the input features.}
        \item \textcolor{black}{\textbf{Concept:} Explanations are made using concepts (a higher level grouping over the input space).}
    \end{enumerate}
    
    \item \textcolor{black}{\textbf{Basis:} The type of input used by the XAI method/model to create explanations \cite{saeed2023explainable, ghorbani2019towards}.}
    \begin{enumerate}[label*=\arabic*., leftmargin=1.5em]
        \item \textcolor{black}{\textbf{Use all input features:} The explanation uses all of the raw inputs to the model.}
        \item \textcolor{black}{\textbf{Concepts:} Either the user defines a concept by specifying examples, the model automatically selects concepts from the feature space, or there is some external definition of concepts grouped over the feature space (user or LLM-specified).}
    \end{enumerate}
    
    \item \textcolor{black}{\textbf{Stage:} The extent to which a model incorporates the explanation in the predictive process \cite{ali2023explainable, speith2022review, schwalbe2024comprehensive}.}
    \begin{enumerate}[label*=\arabic*., leftmargin=1.5em]
        \item \textcolor{black}{\textbf{Explain then predict:} The explanation is directly used as part of the prediction process, influencing the model's output.}
        \item \textcolor{black}{\textbf{Explain from model internals:} The explanation reflects the weights or internal mechanisms of the model but is not directly used for prediction (gray box).}
        \item \textcolor{black}{\textbf{Explanation not used in model:} The explanation is completely separate from the model's internals and has no influence on the predictive process (black box).}
    \end{enumerate}
    
    \item \textcolor{black}{\textbf{Sparsity:} The conciseness of the explanation relative to the amount of the input space (features, concepts) used in the prediction \cite{sun2024sparse, ayoobi2023sparx}.}
    \begin{enumerate}[label*=\arabic*., leftmargin=1.5em]
        \item \textcolor{black}{\textbf{Sparse:} The explanation uses a minimal amount of features/concepts for prediction.}
        \item \textcolor{black}{\textbf{Not Sparse:} The explanation uses more than a minimal amount of the features/concepts for prediction.}
    \end{enumerate}
    
    \item \textcolor{black}{\textbf{Faithfulness:} Alignment between the explanation and the model behavior, also known as fidelity \cite{lyu2024towards,dasgupta2022framework}.}
    \begin{enumerate}[label*=\arabic*., leftmargin=1.5em]
        \item \textcolor{black}{\textbf{Guaranteed:} The explanation aligns to the model behavior with certainty.}
        \item \textcolor{black}{\textbf{Aligned with concepts:} The explanation aligns with the concepts used but is not guaranteed to align with the model behavior.}
        \item \textcolor{black}{\textbf{Approximation:} The explanation is not guaranteed to reflect the model behavior.}
    \end{enumerate}
\end{enumerate}

\section{Additional Details on Datasets}
\label{sec:appendix-datasets}

Here, we provide additional statistics regarding each dataset used in our study. In particular, we highlight their availability details and terms of use.

\textbf{EDU.} We predict student success during the early weeks of four massive open online courses (MOOCs), using students' clickstream data (see Table \ref{tab:EDU_course_stats} for details about the courses). Contrary to the other 4 datasets, this dataset is kept private for student privacy reasons, as required by HREC 058-2020/10.09.2020 and HREC 096-2020/09.04.2022.

\begin{table}[h!]
\centering
\resizebox{\textwidth}{!}{%
\begin{tabular}{@{}llcccccc@{}}
\toprule
\textbf{Title} & \textbf{Identifier} & \textbf{Topic} & \textbf{Level} & \textbf{Language} & \textbf{No.\ Weeks} & \textbf{No.\ Students} & \textbf{Passing Rate [\%]}  \\ \midrule
Digital Signal Processing & DSP &  CS & MSc & English & 10 & 4,012 & 23.1 \\
Éléments de Géomatique & Geo & Math & BSc & French & 11 & 452 & 45.1 \\
Household Water Treatment and Storage & HWTS &  NS & BSc & French & 5 & 2,438 & 47.2 \\
Villes Africaines & VA &  SS & BSc & En/Fr & 12 & 5,643 & 9.9 \\
\bottomrule
\end{tabular}}
\vspace{2mm}
\caption{Course Details and Statistics.\\
Topic abbreviations: Math: Mathematics; NS: Natural Science; CS: Computer Science; SS: Social Science; Arch: Architecture; Bus: Economics and Business.}
\label{tab:EDU_course_stats}

\end{table}

\textbf{AG News} is a news classification dataset, where given a title and description of a real-world article, it has to be classified into one of the four categories: `World', `Sports', `Business', `Sci/Tech' \cite{zhang2015character}. 
It is freely available at the following location: \url{http://groups.di.unipi.it/~gulli/AG_corpus_of_news_articles.html}, but only for non-commercial use.

\textbf{SST.} The Stanford Sentiment Treebank dataset aims at predicting the sentiment from a sentence fragment sourced from a movie review. The dataset is freely available here: \url{https://huggingface.co/datasets/sst}.
This popular benchmark is an extension of the Movie Review Database (MRD) \cite{socher2013recursive}. It includes two sets of labels: one for binary sentiment classification and one for multiclass. We use binary classification to demonstrate a different setting than the multiclass classification of AG News.

\textbf{Breast Cancer.} The Wisconsin Breast Cancer dataset attempts to identify the presence of cancerous tissue from an image of a fine needle aspirate (FNA) of a breast mass \cite{breastcancer}. This dataset is freely available here: \url{https://archive.ics.uci.edu/dataset/17/breast+cancer+wisconsin+diagnostic}, and is licensed under a Creative Commons Attribution 4.0 International (CC BY 4.0) license.

\textbf{Synthetic Dataset.} We generate 5000 points of OpenXAI's synthetic dataset \cite{agarwal2022openxai} through the class \texttt{generate\_gaussians} with 5 cluster centers, which includes both ground truth labels and ground truth explanations, and is available here: \url{https://github.com/AI4LIFE-GROUP/OpenXAI} (under MIT license).

\section{\icc Group Routing Schema}
\label{sec:appendix-grouprouting}

In the following, we discuss the exact grouping schematic employed on the 4 EDU datasets (including three schemes) and the 2 text datasets.

\textcolor{black}{\textbf{Group Specification Guidance}:  In \icc, we design towards the user's actionability of the resulting explanation. Therefore, if the knowledge that a feature is important can lead to a specific action, and if this action is the same one that should be taken for other features, then those features should be grouped together. From the modeling perspective, grouping features together in a concept means that their shared predictive potential should be leveraged, and likely this is more important for one type of features than another. We maintain that a feature should not be placed in multiple groups to have the faithfulness guarantees that are a strength of \icc. However, in a rare and specific case where two actions must be taken based on the feature, or it is too difficult to decide which concept the feature belongs in, it is always possible to 1) put that feature in its own subnetwork (and have it be selected alongside any of the other feature groups), or 2) combine the two subnetworks into a larger concept.}

\textcolor{black}{Concretely, for \texttt{ICC} GR, users specify a grouping that separates individual features into concepts (which are groups of features). User selected features and user definitions of features are not required in \texttt{ICC} FG. For concept specification, users describe feature groupings in the form of a list or dictionary with which features are assigned to which network. In training, the discriminator takes in all the features as input and decides which of the C concepts to activate. In prediction, there are C predictive modules (one for each concept), each with only the features that are mapped to that concept as input. If a specific concept is activated, then only the features assigned to that concept (by a user) are sent to that subnetwork, and this is used for the eventual prediction.}

\begin{table}[ht]
    \centering
    \begin{minipage}{0.58\textwidth}
        \centering
        \resizebox{\textwidth}{!}{%
        \begin{tabular}{l l l}
            \toprule
           \textbf{Dimensions} & \textbf{Corresponding measures} & \textbf{Student patterns} \\
           \midrule 
           \textbf{Effort} & 
           \begin{tabular}{@{}l@{}}
                Total time online \\
                Total video clicks
           \end{tabular} & 
           \begin{tabular}{@{}l@{}}
                Higher intensity \\
                Lower intensity
           \end{tabular}
            \\
           \midrule
           {\textbf{Consistency}} & 
           \begin{tabular}{@{}l@{}}
               Mean session duration \\
               Relative time online \\
               Relative video clicks
           \end{tabular} & 
           \begin{tabular}{@{}l@{}}
                Uniform \\
                First half \\
                Second half
           \end{tabular}
           \\ 
           \midrule 
           {\textbf{Regularity}} & 
           \begin{tabular}{@{}l@{}}
                Periodicity of week day  \\
                Periodicity of week hour \\
                Periodicity of day hour
           \end{tabular}
            & \begin{tabular}{@{}l@{}}
                 Higher peaks \\
                 Lower peaks 
            \end{tabular} \\
           \midrule 
           {\textbf{Proactivity}} & 
           \begin{tabular}{@{}l@{}}
                Content anticipation \\
                Delay in lecture view
           \end{tabular} & 
           \begin{tabular}{@{}l@{}}
                Anticipated \\
                Delayed
           \end{tabular} \\
           \midrule 
           {\textbf{Control}} & 
           \begin{tabular}{@{}l@{}}
                Fraction time spent (video) \\
                Pause action frequency \\
                Average change rate
           \end{tabular} & 
           \begin{tabular}{@{}l@{}}
               Higher intensity \\
               Lower intensity
           \end{tabular} \\
           \midrule 
           {\textbf{Assessment}} & 
           \begin{tabular}{@{}l@{}}
               Competency strength \\
               Student shape
           \end{tabular} & 
           \begin{tabular}{@{}l@{}}
               Higher intensity \\
               Lower intensity
           \end{tabular}\\
           \bottomrule
        \end{tabular}}
        \vspace{2mm}
        \caption{\textbf{EDU Routing by Pattern} uses learning dimensions from \cite{mejia-patterns} to create interpretable feature groupings.}
        \label{tab:mejia_patterns}
    \end{minipage}
    \hfill
    \begin{minipage}{0.38\textwidth}
        \centering
        \resizebox{\textwidth}{!}{%
        \begin{tabular}{l l}
            \toprule
           \textbf{Code} & \textbf{Field of Study} \\
           \midrule
           \textbf{000} & Computer Science, Information \\
            & and General Works\\
           \textbf{100} & Philosophy and Psychology \\
           \textbf{200} & Religion \\
           \textbf{300} & Social Sciences \\
           \textbf{400} & Language \\
           \textbf{500} & Pure Science \\
           \textbf{600} & Technology \\
           \textbf{700} & Arts and recreation \\
           \textbf{800} & Literature \\
           \textbf{900} & History and geography \\
           \bottomrule
        \end{tabular}}
        \vspace{2mm}
        \caption{\textbf{Text Routing} by the Dewey Decimal Classification system \cite{scott1998dewey}. Each code represents a subnetwork in the text variations of the \icc framework.}
        \label{tab:dewey_decimal_classification}
    \end{minipage}
    \vspace{-5mm}
\end{table}

\textbf{Feature Grouping Schema for EDU Datasets}\label{app:fg_edu}

The data is composed of student clickstream data from four different MOOCs. The raw clickstream input is transformed into 45 weekly time-series features that have proven useful for student success prediction in previous literature (e.g. total video clicks, forum interactions). 

\textbf{Routing by pattern.} We organize features according to four learning dimensions identified by \cite{asadi2023ripple} and detailed in Table~\ref{tab:mejia_patterns}—effort, consistency, regularity, proactivity, control, and assessment—based on \cite{mejia-patterns}, with a focus on these dimensions in an extended experiment.

\textbf{Routing by Paper.} In routing by paper, we create 4 distinct feature subsets based on handcrafted initial input features from 4 learning science research papers highlighted as predictive for MOOCs in \cite{swamy2022meta}. We filter each feature set to a expert subnetwork for the research paper the set originated from \cite{lalle-conati,marras-vignoud-kaser-2021,chen-cui, boroujeni-et-al}.

\label{gpt-4}
\textbf{Routing by GPT-4. } To leverage LLMs for automated feature groupings, we use the following prompt alongside the feature descriptions from the appendix table of \cite{swamy2022meta}.

\textbf{Prompt}: 

\begin{graybox}
You are a talented learning scientist, specializing in self-regulated learning behavior. You want to group the following features together into smaller groups of logical concepts based on educational theory with headings. What are your feature groupings?
\end{graybox}

\textbf{Response}: 

\begin{graybox}
Certainly, based on educational theory and self-regulated learning behavior, I would group the provided features into the following logical categories:

\begin{multicols}{2}
\begin{itemize}
    \item Temporal Engagement:
    \begin{itemize}
        \item DelayLecture
        \item RegPeakTimeDayHour
        \item RegPeriodicityDayHour
    \end{itemize}
    \item Study Patterns:
    \begin{itemize}
        \item NumberOfSessions
        \item AvgTimeSessions
        \item TotalTimeSessions
        \item StdTimeSessions
        \item StdTimeBetweenSessions
    \end{itemize}
    \item Video Engagement:
    \begin{itemize}
        \item AvgWatchedWeeklyProp
        \item StdWatchedWeeklyProp
        \item AvgReplayedWeeklyProp
        \item StdReplayedWeeklyProp
        \item AvgInterruptedWeeklyProp
        \item StdInterruptedWeeklyProp
    \end{itemize}
    \item Competency and Learning Progress:
    \begin{itemize}
        \item CompetencyStrength
        \item CompetencyAlignment
        \item CompetencyAnticipation
        \item ContentAlignment
        \item ContentAnticipation
    \end{itemize}
\end{itemize}
\columnbreak
\begin{itemize}
    
    \item Interaction Patterns:
    \begin{itemize}
        \item TotalClicks
        \item TotalClicksProblem
        \item TotalClicksVideo
        \item TotalClicksWeekday
        \item TotalClicksWeekend
        \item TotalTimeProblem
        \item TotalTimeVideo
        \item TotalClicksVideoLoad
        \item TotalClicksVideo
        \item FrequencyEventVideo
        \item FrequencyEventLoad
        \item FrequencyEventPlay
        \item FrequencyEventPause
        \item FrequencyEventStop
        \item FrequencyEventSeekBackward
        \item FrequencyEventSeekForward
        \item FrequencyEventSpeedChange
        \item AvgSeekLength
        \item StdSeekLength
        \item AvgPauseDuration
        \item StdPauseDuration
        \item AvgTimeSpeedingUp
        \item StdTimeSpeedingUp
    \end{itemize}
    \item Assessment Behavior:
    \begin{itemize}
        \item StudentSpeed
        \item StudentShape
    \end{itemize}
\end{itemize}
\end{multicols}
\end{graybox}

Each grouping represents a different aspect of self-regulated learning behavior, focusing on how students engage with learning resources, interact with content, demonstrate competency, and approach assessments. This categorization aligns with principles of self-regulated learning and can help in analyzing and understanding students’ behaviors and strategies within an educational context.

\textbf{Feature Grouping Schema for Text Datasets}
\label{app:fg_text}

For news categorization (AG News) and sentiment prediction (SST) feature grouping, we assign words to subnetworks. For this, we use the Dewey Decimal Code (DDC) for librarians and its hierarchy of topics for book classification to create 10 subnetworks, as showcased by topic in Table \ref{tab:dewey_decimal_classification} \cite{satija2013theory}. Each word is assigned to a subcategory (i.e. the word `school' is assigned to the subcategory `education' under category 300 for `social sciences') and routed to the appropriate parent network. The decision to use the DDC was to use subnetworks that were standardized, pervasive in daily life and clearly human-understandable. To conduct this assignment, we utilize SentenceBERT to encode the subtopics for each DDC heading (i.e. all of 010, 020, 030, etc. for the category 000) \cite{reimers2019sentence}.  The choice of SentenceBERT is motivated towards capturing the broader context of multi-word category headings in a lightweight model. During training and inference, we again use SentenceBERT to encode each word in the input instance, then assign each word to the most similar DDC category in embedding space with cosine similarity.

\section{Hyperparameter Sensitivity and Architecture Validation Experiments}
\label{appendix:sparsity}

We examine the sparsity and Gumbel Softmax hyperparameters, and how they impact the \icc model performance. 

\textbf{Sparsity Criterion Experiments}
For the feature gating architecture to further improve interpretability, we would like the network to learn sparse feature activations. That is, for a given input $x$ we would like to reduce the number of features that affect the model prediction. To achieve this we apply regularization to the generated feature mask.

One natural choice to enforce sparse feature activations is to apply $L_1$-norm regularization to the feature mask, which penalizes a high number of nonzero elements. Another choice is to use annealed regularization as presented by Verelst and Tuytelaars \cite{dynconv}, which might aid the model to first work through a prediction optimization phase that is not confounded by any additional error terms before moving towards a sparsity-enforcing phase.

We experiment with annealing L1 and L1 regularization across four courses, and find that traditional L1 regularization is more stable (at least in the time-series setting). The Baseline BiLSTM results are not reported as confidence intervals here as they are directly sourced from a recent benchmarking paper by Swamy et al., with confirmed similarity by Asadi et al. \cite{swamy2022meta, asadi2023ripple}. We reproduce this benchmark above with similar values in Table \ref{tab:sparsity-edu}.

\begin{table}[ht]
\centering

\begin{tabular}{@{}llll@{}}
\toprule
\multicolumn{1}{c}{\textbf{\begin{tabular}[c]{@{}c@{}}EDU\\ Dataset\end{tabular}}} &
  \multicolumn{1}{c}{\textbf{Baseline}} &
  \multicolumn{2}{c}{\textbf{\begin{tabular}[c]{@{}c@{}}\icc \\ Feature Gating\end{tabular}}} \\ \midrule
\multicolumn{1}{c}{\textit{40\% EP}} &
  \multicolumn{1}{c}{\textit{BiLSTM}} &
  \multicolumn{1}{c}{\textit{Annealing}} &
  \multicolumn{1}{c}{\textit{L1}} \\ \midrule
DSP       & 82   & 87.76 +/- 3.12 & 90.75 +/- 0.01 \\
Geo       & 76.2 & 81.13 +/- 5.39 & 71.92 +/- 0.01 \\
HWTS      & 72   & 77.58 +/- 0.01 & 82.89 +/- 0.04 \\
VA        & 73.8 & 84.81 +/- 0.01 & 77.80 +/- 0.01 \\ \bottomrule
\end{tabular}
\vspace{2mm}
\caption{\label{tab:sparsity-edu} Annealing L1 regularization in comparison with L1 regularization across EDU datasets. }
\end{table}

\textbf{Gumbel Softmax Hyperparameter Sensitivity} 

We explore the effects of varying the Tau and Threshold parameters during training for \icc FG (top two plots) and GR (bottom two plots) using the DSP course dataset. This dataset was selected due to its extensive use as a benchmark for explainability in education research \cite{swamy2023trusting}. \textcolor{black}{We examine the variability over a tuned discriminator layer (batch size 8, and layer size 32)}, with additional configurations detailed in Appendix \ref{sec:reproduce}. While BAC/ACC remains similar across settings, optimizing the Threshold in conjunction with Tau is crucial for performance; often times changing the threshold is quite a variable experience. Notice the changing impact on BAC across different thresholds as Tau changes.

The thresholds used were 0.5, 0.6, 0.65, 0.7, 0.725, 0.75, 0.775, and 0.8, while the tau values were 0.5, 1.0, 5.0, and 10.0.

\begin{figure}[ht]
    \centering
    \includegraphics[width=\linewidth]{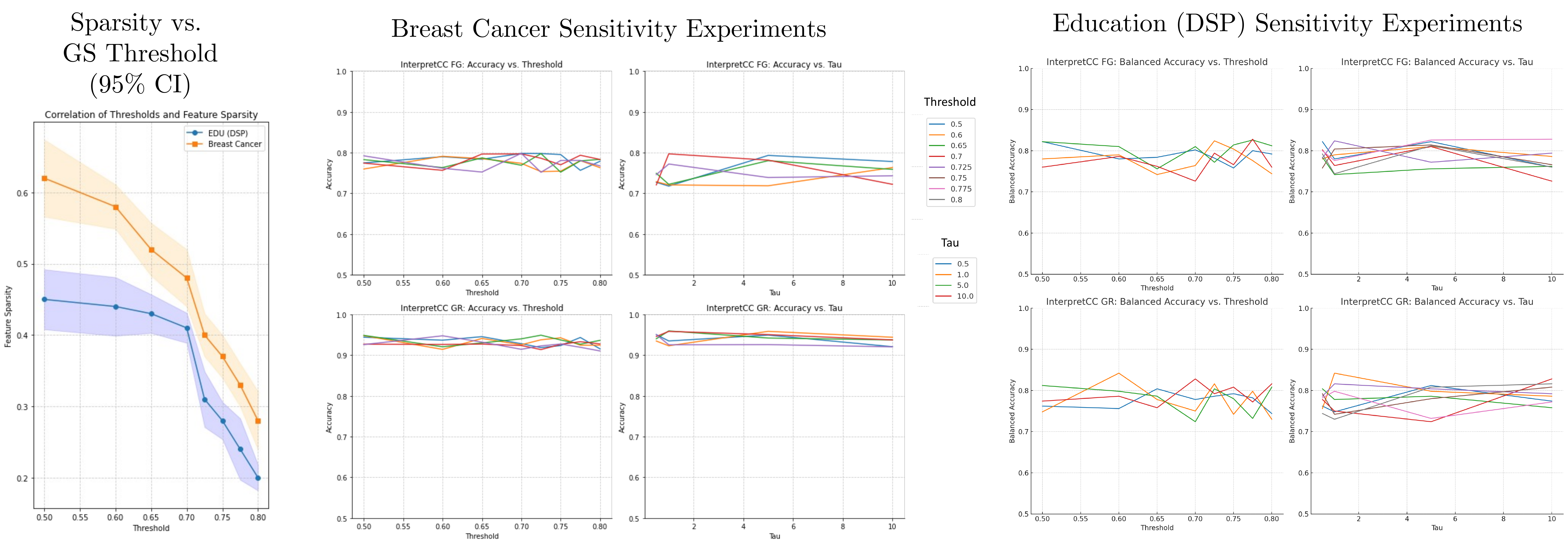}
    \caption{\textbf{\icc Gumbel Softmax Hyperparameter Sensitivity Experiments.} \textcolor{black}{We examine changing Tau and Threshold parameters during training for \icc FG (top two plots) and GR (bottom two plots) for the DSP course dataset and the Breast cancer dataset.}}
    \label{fig:sensitivity}
\end{figure}

\textcolor{black}{All 95\% CIs overlap for the experiments for Figure \ref{fig:sensitivity}. They are omitted from the plots for readability. Significantly high variation (over 0.1) include the following settings:
\begin{itemize}
    \item \icc FG 0.5 threshold, 1.0 Tau with 0.183 standard deviation
    \item \icc FG 0.6 threshold, 5.0 Tau with 0.122 standard deviation
    \item \icc FG 0.65 threshold, 10.0 Tau with 0.151 standard deviation
    \item \icc GR 0.5 threshold, 1.0 Tau with 0.107 standard deviation
    \item \icc GR 0.6 threshold, 5.0 Tau with 0.114 standard deviation
\end{itemize}}

\textcolor{black}{GR architectures are on average 0.043 more stable (less variable) than FG architectures. These experiments show that while the performance of \icc has overlapping 95\% CIs while changing parameters, certain parameter settings have higher variability than others. For both education and health tasks, a $\tau$ of 10 and a Gumbel-Softmax threshold of around 0.7 to 0.8 are performant, sparse in activated features, and relatively stable. Notably, the results for the Breast Cancer dataset are less variable than the DSP dataset, which shows the sensitivity of parameters is domain dependent.}

\textbf{Architecture Validation Analysis}
\label{app:specialization}

\textcolor{black}{\textbf{“Do we genuinely achieve specialized networks?”} \cite{jacobs1997bias} demonstrates that mixtures-of-experts architectures can optimize the bias-variance trade-off by specializing subnetworks for specific regions of the input space. Similarly, \cite{jiang1999identifiability} prove that the identifiability of mixtures-of-experts models depends on their parameterization.
In the following experiment, we show that \icc  subnetworks specialize to information that are routed to them, and are worse at predicting on information that is not routed to them.}

\begin{figure}
    \centering
    \includegraphics[width=\linewidth]{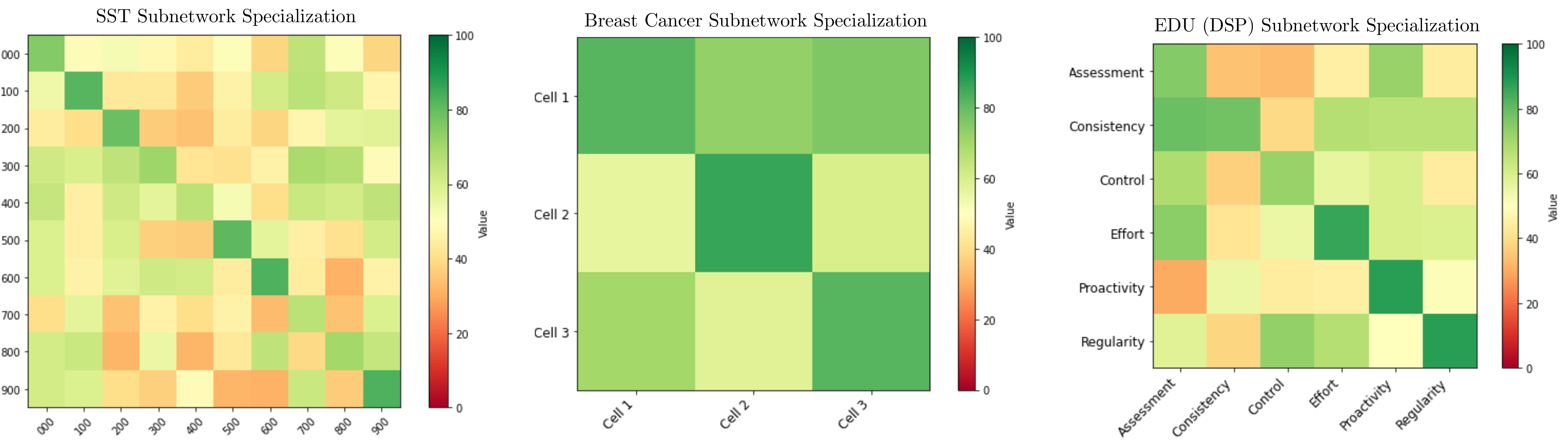}
    \caption{\textcolor{black}{Validation of Specialized Subnetworks: SST, Breast Cancer, and DSP datasets.}}
    \label{fig:heatmaps}
\end{figure}

\textcolor{black}{The experiment, conducted across three different use cases (SST: text, DSP: time series, Breast Cancer: tabular) of \icc, demonstrate that subnetworks predict more strongly in on data that has been routed to them as opposed to data that is supposed to be routed to other subnetworks. Specifically, for DSP, the route-by-pattern networks predict 22.88\% +/- 7.56\% more accurately (balanced accuracy) on points sent to them. For SST, the subnetworks specialize even more strongly (23.59\% +/- 7.86\%). For Breast Cancer, the cell-based grouping predicts 11.88\% +/- 5.57\% better when routed to the same network. The heatmaps in Figure \ref{fig:heatmaps} show the prediction rates of subnetworks (x-axis) on data routed from every other subnetwork (y-axis). Note that across all three use cases, the diagonal performance is the strongest.}

\textcolor{black}{\textbf{“Do these networks contribute to improved prediction accuracy?"} As proven by \cite{jordan1994hierarchical}, mixture-of-experts are a strongly predictive architecture, analogous to the case of ensemble models over monolithic models. The authors show that hierarchical mixtures-of-experts (HMoE) can recursively partition the input space and train effectively using the Expectation-Maximization (EM) algorithm. Empirically, in comparison to post-hoc explainability approaches on top of a non-interpretable base module, we show performance gains (Table 2) for \icc GR in 7 out of 8 datasets, statistically significant higher performance in the Geo and Breast Cancer settings, and comparable performance (overlapping 95\% CIs) in all other settings; this directly shows that using specialized subnetworks is superior to the non-interpretable baseline approach. As \icc FG showcases simply the sparsity requirement with no subnetwork logic, and GR has higher performance than FG results in 4 settings, we know that the subnetworks can be helpful. Similarly, over the non-interpretable base module, FG statistically significantly beats performance in DSP, HWTS, and Synthetic data cases, showcasing the benefits of adaptive sparsity.}

\textcolor{black}{The comparison to NAM showcases the benefits of combining important features together in the predictive module instead of having a separate subnetwork for every feature. The comparison to SENN showcases the benefit of expert-specified routing logic as opposed to automated concept selection. The new comparison to FRESH showcases the benefit of selecting a concept-based explanation over a contiguous explanation. \cite{hazimeh2021dselect} proposes and proves that an architecture with similar foundations in conditional computation achieves efficient and differentiable sparse subnetwork selection, improving task performance and computational efficiency in multi-task learning. They do not focus on an interpretability objective.}

\textcolor{black}{\textbf{“Do these networks contribute to improved interpretability?} Indeed, intrinsic explanations (through hierarchical models or gating) do contribute to increased interpretability \cite{ismail2023interpretable, stojanovic2022bayesian}. From an empirical perspective, the experiments in Table 4 demonstrate \icc’s ability to capture meaningful patterns in the underlying data. The user study addresses whether the outputs of \icc explanations are interpretable to users, and can be useful.}

\section{User Study}
\label{app:user-study}
In this section, we discuss the details of the user study presented in Section \ref{sec:user-study}, discussing the design, content, and additional analyses of the results (including an ANOVA and Tukey HSD tests).
We designed the study over four rounds of pilots, with 8 individuals from diverse backgrounds, continuously updating the study design following their feedback. The survey has been approved by the the Human Research Ethics Committee (HREC) under application number HREC 065-2022/27.09.2022.

\textbf{Design of the Study}

We recruit 56 participants using \texttt{Prolific},\footnote{\url{www.prolific.com/}}, selecting the ones who identified their current profession as a teacher and who have at least a bachelor's degree. Our target participants have expertise in teaching, as they would be well-suited to understand both the educational context of the study and the consequences of black-box models for student outcomes. During the study, we ask the participants whether they have ever taken or prepared material for an online course (MOOC), their level of education, and what level they are teaching at (from primary school to graduate school). Detailed demographics distribution can be found in Figure \ref{fig:userstudy_demo}. The sample of participants is gender-balanced, and about half of them have taken or participated in creating a MOOC. The median completion time is 22 minutes, and the average reward per hour is £14.55. 

\begin{figure}[ht]
    \centering
    \includegraphics[width=\textwidth, trim={5cm 0 25cm 0}, clip]{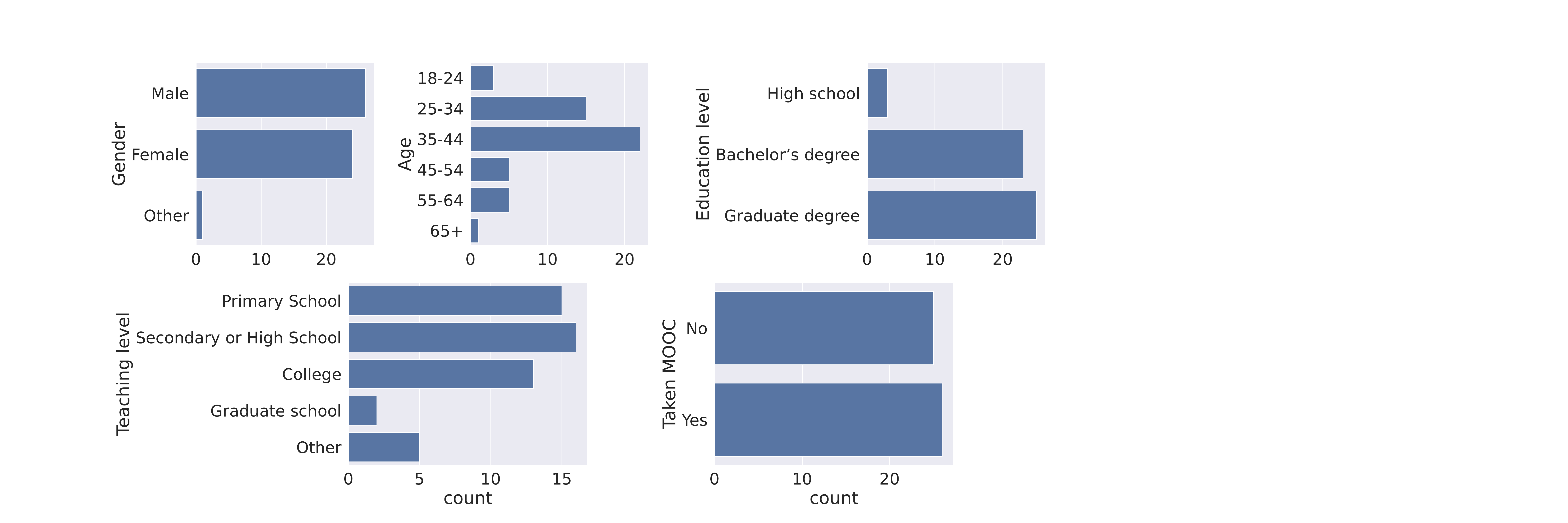}
    \caption{Demographics of teachers that participated in the user study.}
    \label{fig:userstudy_demo}
\end{figure}

At the beginning of the study, the participants are shown the following welcome message and disclaimer:\\

\begin{mdframed}[
    backgroundcolor=green!10,
    linecolor=black,
    linewidth=1pt,
    innertopmargin=10pt,
    innerbottommargin=10pt,
    innerrightmargin=10pt,
    innerleftmargin=10pt,
    roundcorner=10pt,
    frametitlebackgroundcolor=gray!30,
    frametitlealignment=\raggedright 
]
Dear participant,

Thank you for participating in our study on model explanations. We are very grateful for your participation and your invaluable insight. Please read this Explanatory Statement in full before moving forward. If you would like further information regarding any aspect of this project, please contact us via the email address provided below.

We are a group of researchers from the [ANON] dedicated to improving education through technology. \textbf{The goal of this study is to evaluate different explanations to help a student improve their performance in an online course.}

- This survey has been approved by the [ANON] Human Research Ethics Committee (HREC) under application number HREC 065-2022/27.09.2022. HREC reviews research proposals involving human participants to ensure that they are ethically acceptable.

 - All the personal information will be kept confidential and anonymized. Only demographic information is being recorded and will only be reported as aggregate in a way that prevents identification of any individual participant. You can freely withdraw at any time and any collected data you provided so far will be destroyed.

 - All data will be collected and stored safely and reported in an anonymous form, in accordance with the [ANON] law on data protection ([ANON]).

- Only anonymized or aggregated data may be used in follow-up research (subject to ethics approval), and made available to other researchers for further analysis and for verification of the conclusions reached by the research team.

- Only the principal investigator and the aforementioned researchers have access to the original data under strict confidentiality. Results from the project may be published in conference papers and/or journal articles. In any case, no personal data will be published (only aggregated, anonymous and/or anonymized data will be published).

- Personal data of participants will be stored for 5 years from the date of collection. During this time, participants have the right to access their data and request information about the processing of their personal data. In order to exercise this right, you need to contact the Principal Investigator.

By participating in this survey, you agree that your data can be used for scientific purposes.

\textbf{In the following study, you will be asked to compare explanations for approximately 35 minutes.} Please ensure that you have enough time to finish the study correctly. Unfinished or only partially answered studies will not be considered as taken part.

We ask you to approach the questions and exercises with seriousness and to complete them to the best of your ability. We will subsequently check questionnaires for seriousness and will have to discard questionnaires that do not meet this requirement.

Thank you for your help. If you encounter any problem with the survey, or if you want to give extra feedback, or receive additional information, feel free to contact us [ANON].
\end{mdframed}

\textbf{Content of the study}

First, we explain the setting of the study to the participants with the following introductory message:

\begin{mdframed}[backgroundcolor=orange!10, linecolor=black, linewidth=1pt, innertopmargin=10pt, innerbottommargin=10pt, innerrightmargin=10pt, innerleftmargin=10pt, roundcorner=10pt]
You are a teaching assistant helping with a Massive Open Online Course (MOOC). This course is taught at the Master's level with quizzes and video lectures taking place over 10 weeks. Since it’s a difficult course with a low passing rate (23.1\%), the teaching team wants to help students who are not doing well to perform better in the course by giving them personalized assistance, and encourage students who are already performing well to continue.

To do this, we have models to predict student success or failure using various weekly behavior features, such as number of video clicks or how accurately questions are answered on the weekly quizzes. If potential failure is predicted early (in our case, in the first 4 weeks of the course), we can use the explanation of the prediction to give additional support (i.e. specific tutoring or assignment reminders) to help the student pass the course.

We train four interpretable machine learning models. Each model predicts a student's performance at the end of the course, in the form of “pass” or “fail”, but also gives us which factors contribute to student success or failure. We want to compare these explanations according to several criteria:
\begin{itemize}[leftmargin=1em]
    \item Usefulness: This explanation is useful to understand the prediction.
    \item Trustworthiness: This explanation lets me judge if I should trust the model.
    \item Actionability: This explanation helps me know how to give feedback to the student.
    \item Completeness: This explanation has sufficient detail to understand why the prediction was made.
    \item Conciseness: Every detail of this explanation is necessary. 

\end{itemize}
\end{mdframed}

We randomly sample 4 students from the test set. Among the 4 selected students, 3 failed the class and one passed. For each student, we predict their success or failure with each model and generate an explanation. We provide them to the participants along with the models' prediction of the student's success or failure. The ground truth (student's performance) and the models' performance are not provided to the participants so that we do not bias their assessment. 

\begin{figure}[ht!]
    \centering
    \includegraphics[width=0.8\textwidth]{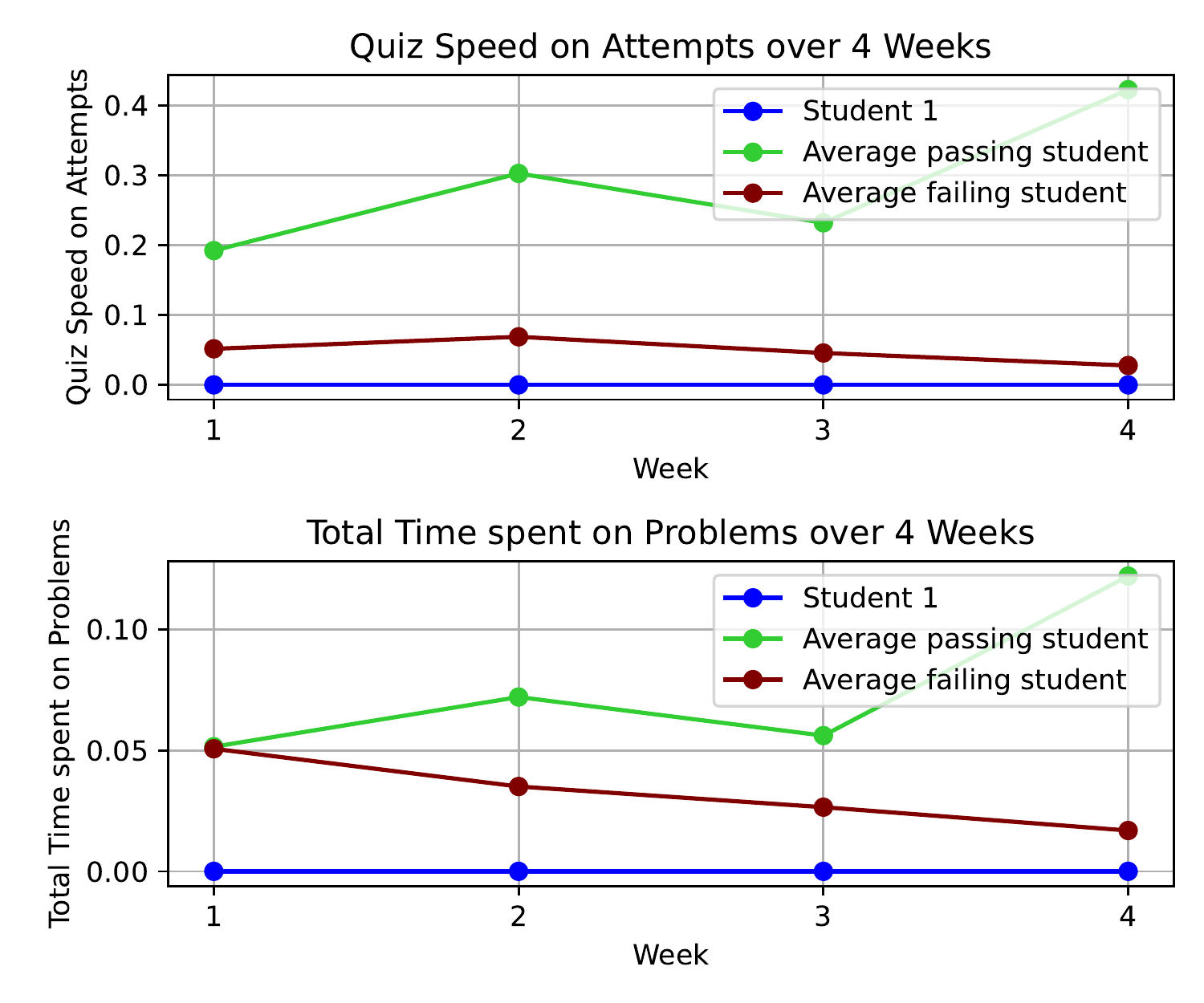}
    \caption{\textcolor{black}{ICC FG method: Importance score visualization of feature-weeks given to the participants for the \icc \featuregating method, for one student.}}
    \label{fig:userstudy-fg}
\end{figure}

\begin{figure}[ht!]
    \centering
    \includegraphics[width=0.8\textwidth]{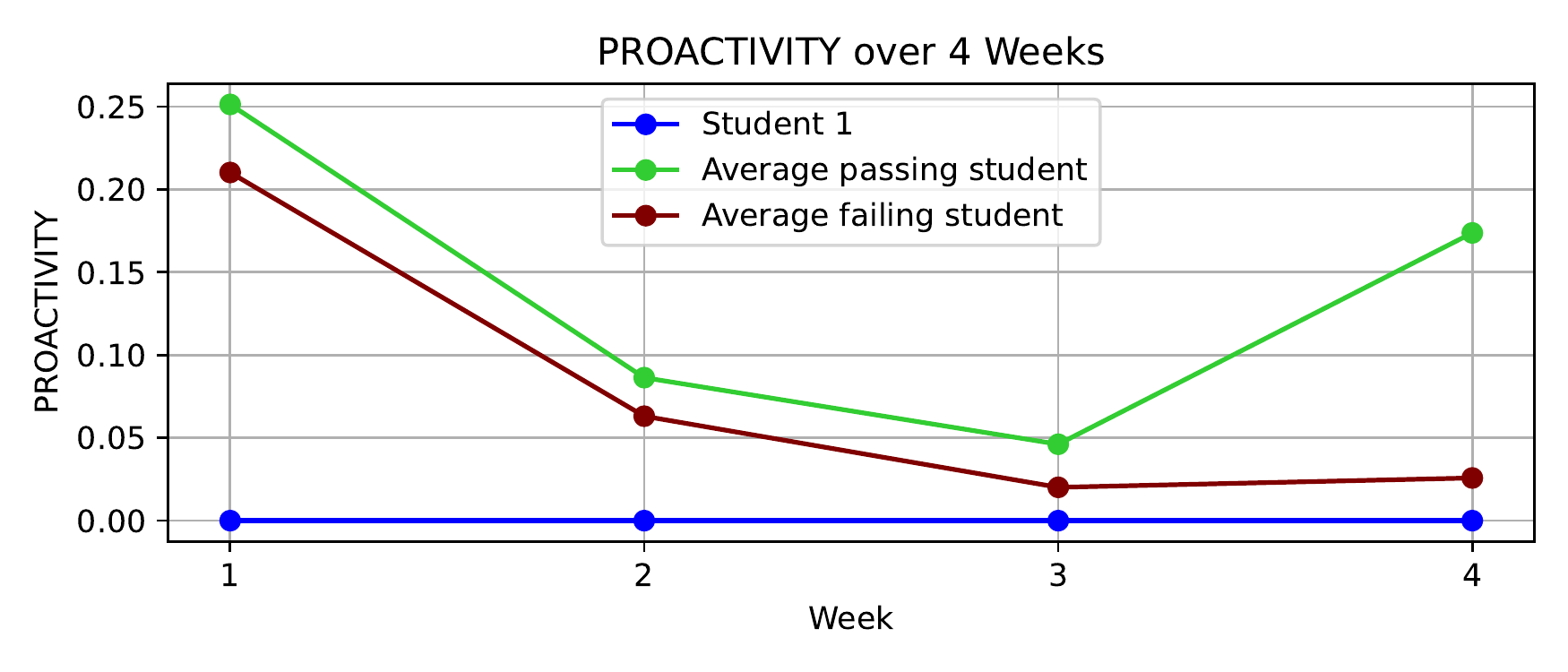}
    \caption{\textcolor{black}{ICC GR method: Importance score visualization of feature-weeks given to the participants for the \icc \grouprouting with pattern-based matching method, for one student.}
}
    \label{fig:userstudy-pr}
\end{figure}

\begin{figure}[ht!]
    \centering
    \includegraphics[width=\textwidth]{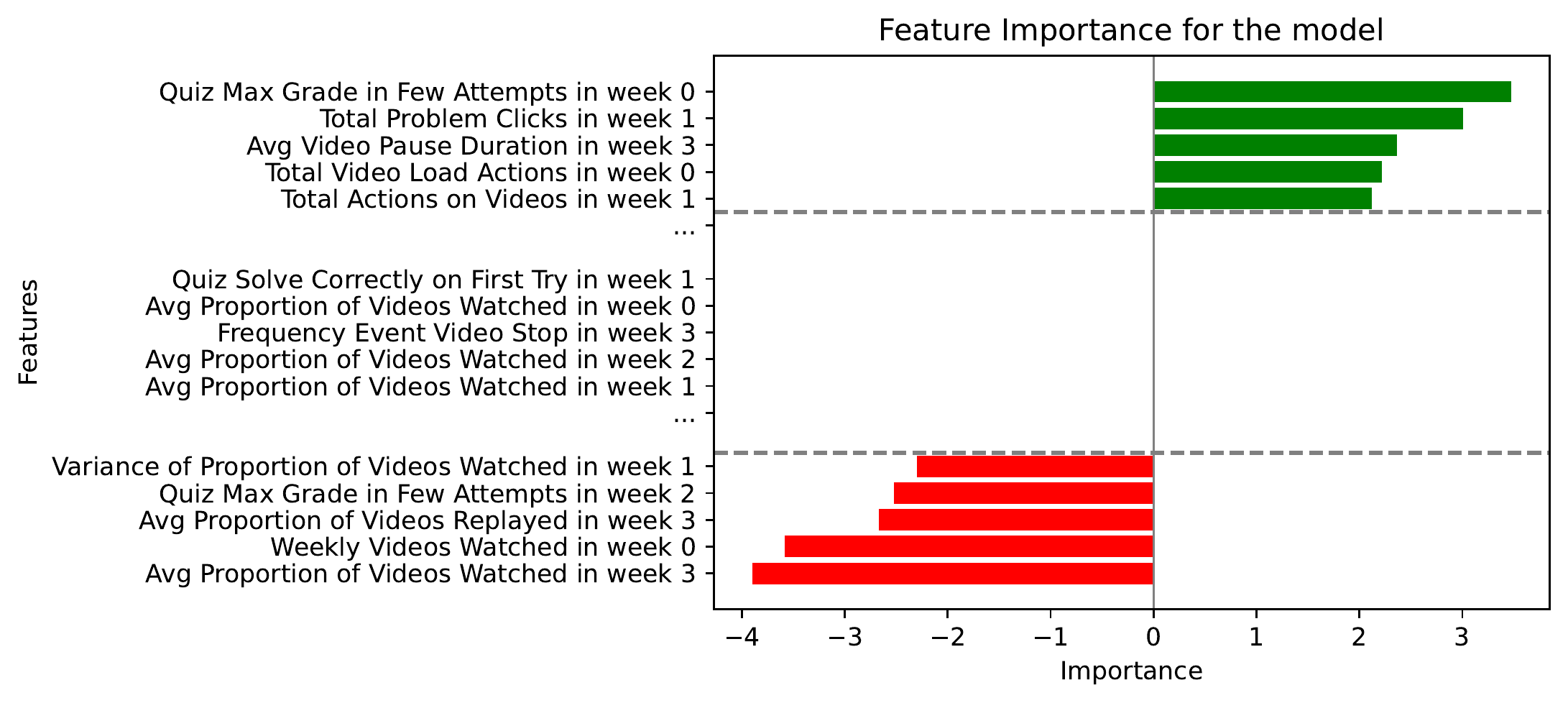}
    \caption{\textcolor{black}{NAM method: Importance score visualization of feature-weeks given to the participants, for one student}.
}
    \label{fig:userstudy-nam}
\end{figure}

\begin{figure}[ht!]
    \centering
    \includegraphics[width=\textwidth]{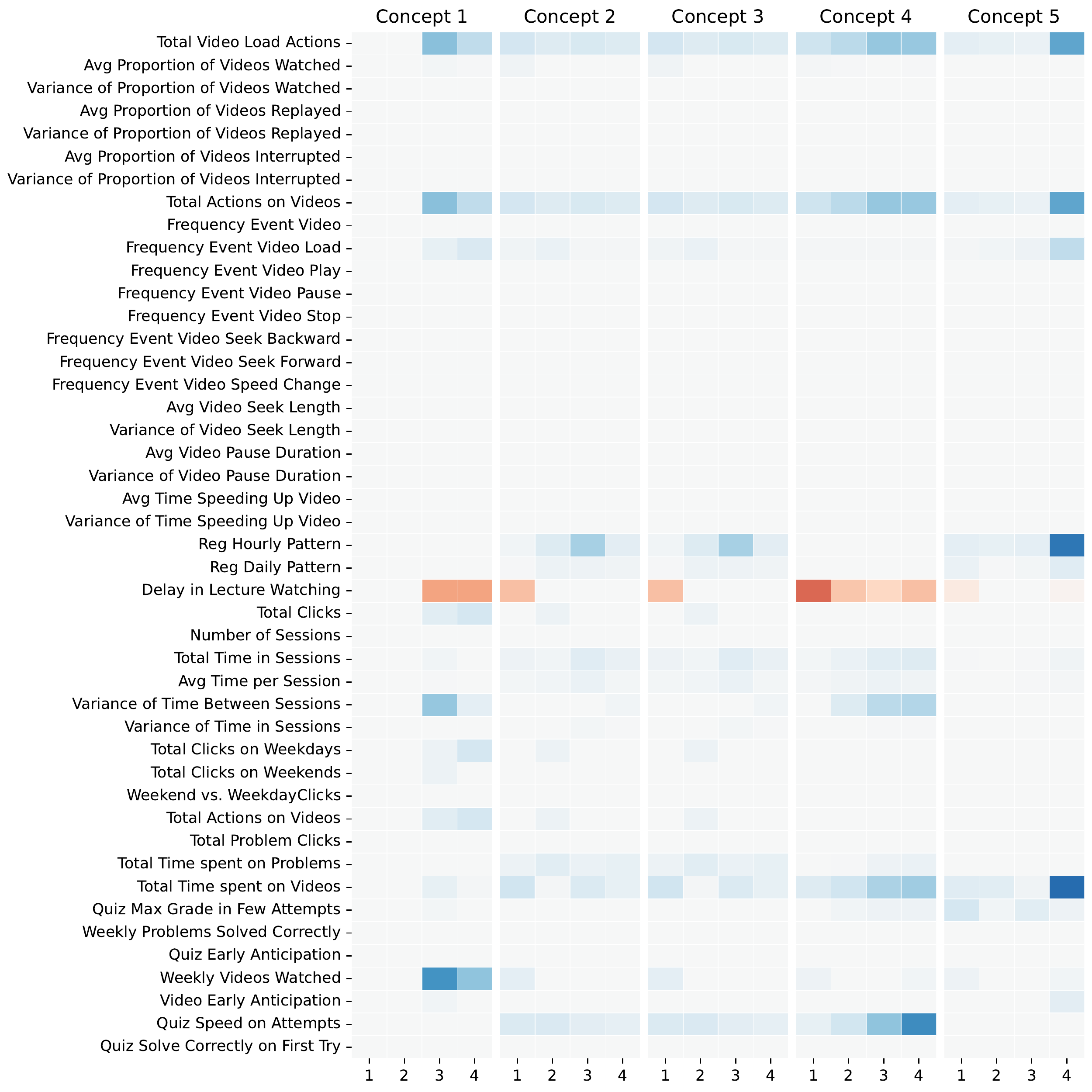}
    \caption{\textcolor{black}{SENN Method: Importance score visualization of feature-week given to the participants, for one student}.}
    \label{fig:userstudy-senn}
\end{figure}

The content of the explanation obtained by each method differs greatly. We simplify the explanations and render them in textual and graph format to make them as easy to understand as possible to a human.  For \icc explanations, we provide the full list of single features / feature groups used by the models. For \featuregating, we show the evolution of the student's behavior across weeks for each of the features used (see Figure \ref{fig:userstudy-fg}). For \grouprouting, we compute a generic score for each concept used by the model, by averaging the normalized behavior features that compose the concept. We show the evolution of that concept measure across weeks for the student. We also provide the definition of the concept and of the features that compose it (see Figure \ref{fig:userstudy-pr}). For SENN, we select the top 5 groups of students, that we call \textit{concepts}. We showcase all 180 feature-weeks for each concept along with their importance in that concept, highlighting the salient ones. We also provide the importance of each concept for the model's prediction (see Figure \ref{fig:userstudy-senn}). Finally, NAM assigns feature importance to all 180 feature-weeks used to make the prediction. We select the 5 feature-weeks found to have highest positive impact, 5 feature weeks with the lowest impact, and the 5 feature-weeks with highest negative impact. We show their importance in a barplot (see Figure \ref{fig:userstudy-nam}).

Note that the choice we made for the presentation of the explanations might have an influence on the participants' perception of the explanations. In an ideal setting, we would provide a very detailed description of how each model uses the features and how the explanation is derived, so that the participants can fairly assess the explanation's quality. However, in a realistic setting, the user facing the explanation might not have the time or prior knowledge necessary to understand these elements. To tackle that trade-off between ensuring thoroughness and accessibility of the content of each model's explanation, we opted for a balanced approach in presenting each explanation with a simple graph and an explanatory text that we kept as short as possible.

\begin{figure}[ht!]
    \centering
    \includegraphics[width=\textwidth]{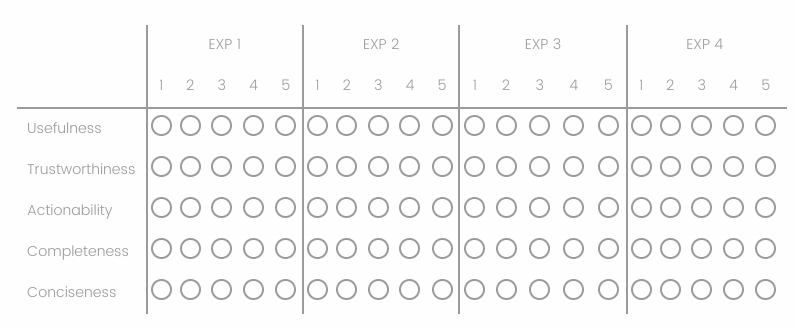}
    \caption{Format of the Likert scale question asked for each criterion and explanation.}
    \label{fig:userstudy-likertscale}
\end{figure}

For each new sample (student taking the course), we provide the list of 4 model explanations in random order. We ask participants to compare these explanations according to five criteria using likert scales. A screenshot of the answer section is shown in Figure \ref{fig:userstudy-likertscale}. We include a practice question to train the participants in how to answer the study and filter inattentive experts. We excluded from the analysis 5 participants who failed to answer correctly to half of the practice questions.

Along with the \icc FG graphs, we provide the following explanatory text:
\begin{graybox}
This student is predicted to fail the course. \textbf{The model found the following 2 features to be the most predictive for this student, and only used these features to make the prediction}:
\begin{itemize}[noitemsep,left=0pt]
\item Quiz Speed on Attempts: The average time passed between two consecutive attempts for the same quiz.
\item Total Time spent on Problems: The total (cumulative) time that a student has spent on problem events.
\end{itemize}
The plot shows the evolution of the student's behavior for each feature across the 4 weeks.
\end{graybox}

Along with the \icc GR graph, we provide the following explanatory text:
\begin{graybox}
This student is predicted to fail the course. \textbf{For this student, the model decided to only use 1 group(s) of features to make the prediction as this was the one(s) it found most important. The groups of features were designed by expert literature in learning sciences}:
\begin{itemize}[noitemsep,left=0pt]
\item Concept: PROACTIVITY
\end{itemize}
Definition: Proactivity measures the extent to which students are on time or ahead of the schedule, as engagement in pre-class activities has shown to be associated with exam performance.
\begin{itemize}[noitemsep,left=0pt]
\item It includes the following features: The number of videos covered by the student from those that are in subsequent weeks and The average delay in viewing video lectures after they are released to students.
\end{itemize}
\end{graybox}

Along with the NAM graph, we provide the following explanatory text:
\begin{graybox}
This student is predicted to fail the course. \textbf{The model used all 180 feature-weeks (45 features from 4 weeks) to make the prediction. It has assigned a level of importance for each feature-week, showing how much it impacts the predictions, independently of the student's behavior.}
Out of the 180 feature-weeks, the plot shows the 5 feature-weeks found to have the highest positive impact, 5 feature weeks with the lowest impact, and the 5 feature-weeks with the highest negative impact. For example, Quiz Max Grade in Few Attempts in week 0 has an importance score of 3.48.
\end{graybox}

Along with the SENN graph, we provide the following explanatory text:
\begin{graybox}
This student is predicted to fail the course. \textbf{The model used all 180 feature-weeks (45 features from 4 weeks) to make the prediction. It groups them into 5 concepts automatically and assigned a score to each concept.} Each concept can be interpreted as a group of features that are important for the prediction. The plot shows the importance of each feature-week for each concept, from high positive importance (blue) to high negative importance (red).

Here is the list ordered by absolute value of importance:
\begin{itemize}[noitemsep,left=0pt]
    \item Concept 3 (-0.21)
    \item Concept 2 (0.08)
    \item Concept 1 (-0.01)
    \item Concept 5 (-0.01)
    \item Concept 4 (0.01)
\end{itemize}
\end{graybox}

\textbf{Significance Testing for User Study Results}\label{app:userstudy_res}

We perform an ANOVA to determine the effect of the model and the sample (the student) on the score given by the participants, for each criterion and on average.
Table \ref{tab:userstudy_anova} shows the p-values testing the significance of the effect of the model and sample on the participants' scores for each criterion. It can be interpreted the following way. In the first row, if the p-value is lower than the significance level (0.05), then there is a statistically significant difference in scores across the different models for that criterion. In the second row, if the p-value is below 0.05, there is a statistically significant difference in scores across the different students. Finally, a p-value lower than 0.05 in the final row shows that the effect of the model on the scores depends on the student. 
According to the table, the model has always a significant impact of the value given to each criterion. 
Then, we apply Tukey's Honest Significant Difference (HSD) Test to determine, for each pair of explanation, if their scores are significantly different (Figure \ref{fig:tukey}).

\begin{table}[ht]
    \centering
    \resizebox{\textwidth}{!}{%
\begin{tabular}{lrrrrrr}
\toprule
{} &  Usefulness &  Trustworthiness &  Actionability &  Completeness &  Conciseness &  Global \\
\midrule
Model         &       0.000 &            0.001 &          0.000 &         0.000 &        0.000 &   0.000 \\
Student       &       0.004 &            0.018 &          0.144 &         0.003 &        0.075 &   0.075 \\
Model:Student &       0.143 &            0.058 &          0.063 &         0.000 &        0.171 &   0.171 \\
\bottomrule
\end{tabular}}
    \caption{ANOVA results for the user study.}
    \label{tab:userstudy_anova}
\end{table}

Extended description for Figure \ref{fig:tukey} -- Significance test using Tukey's Honest Significant Difference (HSD) Test. It indicates which pairs of models have significantly different means. We highlight the top model on average (\textit{Global satisfaction}, in the last plot), ICC FG (\icc \featuregating) in blue, and the models that are significantly worse according to each criterion in red. Example of interpretation using the Tukey HSD test: for the criterion \textit{Usefulness}, ICC FG is scored significantly better than SENN and NAM, but is not significantly better than ICC GR.

\section{Reproducibility and Compute Details}
\label{sec:reproduce}
All EDU, Health, and Synthetic experiments were run on a single NVIDIA A100 GPU with 32 GB Memory, each model taking approximately 20 minutes (or often less) to train. Text experiments for InterpretCC took approximately an hour to train. SENN Concepts took the longest time of all models, with each model running within 3-4 hours.

All interpretable models reported in \ref{tab:all_results} and \ref{tab:group-routing} has been hyperparameter tuned over the following parameters with early-stopping (where applicable): 
\begin{itemize}
    \item learning rate: 1e-3, 1e-4, 2e-5, 1e-5
    \item layer size: 16, 32, 64
    \item number of concepts (only for SENN): 5, 6, 7
    \item batch size: 8, 16, 32, 64)
    \item gumbel softmax threshold (only for \icc): 0.1, 0.3, 0.5, 0.7
\end{itemize}
Each Feature-Based model was run for 100 epochs with early stopping, and each Concept-Based model was run for 20 epochs with early stopping. Other details related to preprocessing and thresholds are included directly in our repository.

\section{Gumbel Softmax trick and its application to \icc}
\label{appendix:gumbel}
To make the feature gating and routing architectures compatible with backpropagation, we need to make the masks differentiable. These discrete decisions can be trained end-to-end using the Gumbel Softmax trick \cite{gumbel}. This method adapts soft decisions into hard ones while enabling backpropagation, i.e. provides a simple way to draw samples from a categorical distribution. 

Given a categorical distribution with class probabilities $\pi = [\pi_1 \, \pi_2 \, ... \, \pi_N]$, one can draw discrete samples $z$ as follows: 
$$z = \textsc{OneHot}\left(\arg\max_i [g_i + \log \pi_i]\right)$$

where $g_1 ... g_N$ are i.i.d. samples drawn from the Gumbel$(0, 1)$ distribution. Then, the softmax function is used as a differentiable approximation to $\arg\max$ to generate a $N$-dimensional sample vector $y$ such that
$$y_i = \frac{\exp((\log(\pi_i) + g_i)/\tau)}{\sum_{j=1}^N\exp((\log(\pi_j) + g_j)/\tau)} \quad \text{for } i = 1, ..., N$$
where $\tau$ is a softmax temperature parameter that is fixed at $\tau = 1$ for experiments in this project. 

Notice that for the gating mechanism, an independent sample is drawn for each `gate' instead of for each datapoint in routing. For example in feature gating, for each feature $i$, a soft-decision $a_i \in (-\infty, +\infty)$ is outputted by the discriminator layers. The probability $\pi_1$ that the feature should be activated as well as the complement probability $\pi_2$ (feature is not activated) can then be computed by using the sigmoid function:
$$\pi_1 = \sigma(a_i) \qquad \pi_2 = 1 - \pi_1 = 1 - \sigma(a_i)$$ The corresponding (1-dimensional) sample $y$ for each $i$ can thus be reduced to
$$y = \sigma\left(\frac{a_i + g_1 - g_2}{\tau}\right)$$
In other words, the discriminator layers from Fig. \ref{fig:arch} actually feed into an adapted Gumbel Sigmoid where $\sigma_i$ is the corresponding $y$ sample as described above.

For routing, the discriminator layers actually output the route logits to a Gumbel Softmax, which constructs the categorical sample vector (of dimension equal to the number of routes and $i$-th entry $y_i$ defined as above).

Finally, we can use a straight-through estimator during training. In other words, binary (or hard/quantized) samples are then used for the forward pass while gradients are obtained from the soft samples for backpropagation. This means that, given soft decisions $\sigma_i$,  architectures that use a mask $M = [m_1 \, ... \, m_N]$ with $m_i = \mathds{1}_{\{\sigma_i \geq 0.5\}}$ differ in value during the forward and backward pass:
$$m_i = \begin{cases}
    \mathds{1}_{\{\sigma_i \geq 0.5\}} \qquad \text{(forward pass)}, \\
    \sigma_i \qquad\qquad\quad \text{(backward pass)}
\end{cases}$$

\section{Additional Group Routing Experiments}\label{app:group_routing}

We conduct additional analyses for the high impact, real-world applications in Health, Text, and EDU.

\textbf{Grouping Analysis for Breast Cancer Dataset} 

\begin{figure}
     \begin{minipage}[b]{0.48\textwidth}
     \centering
     \includegraphics[width=\linewidth, trim={5 5 5 5}, clip]{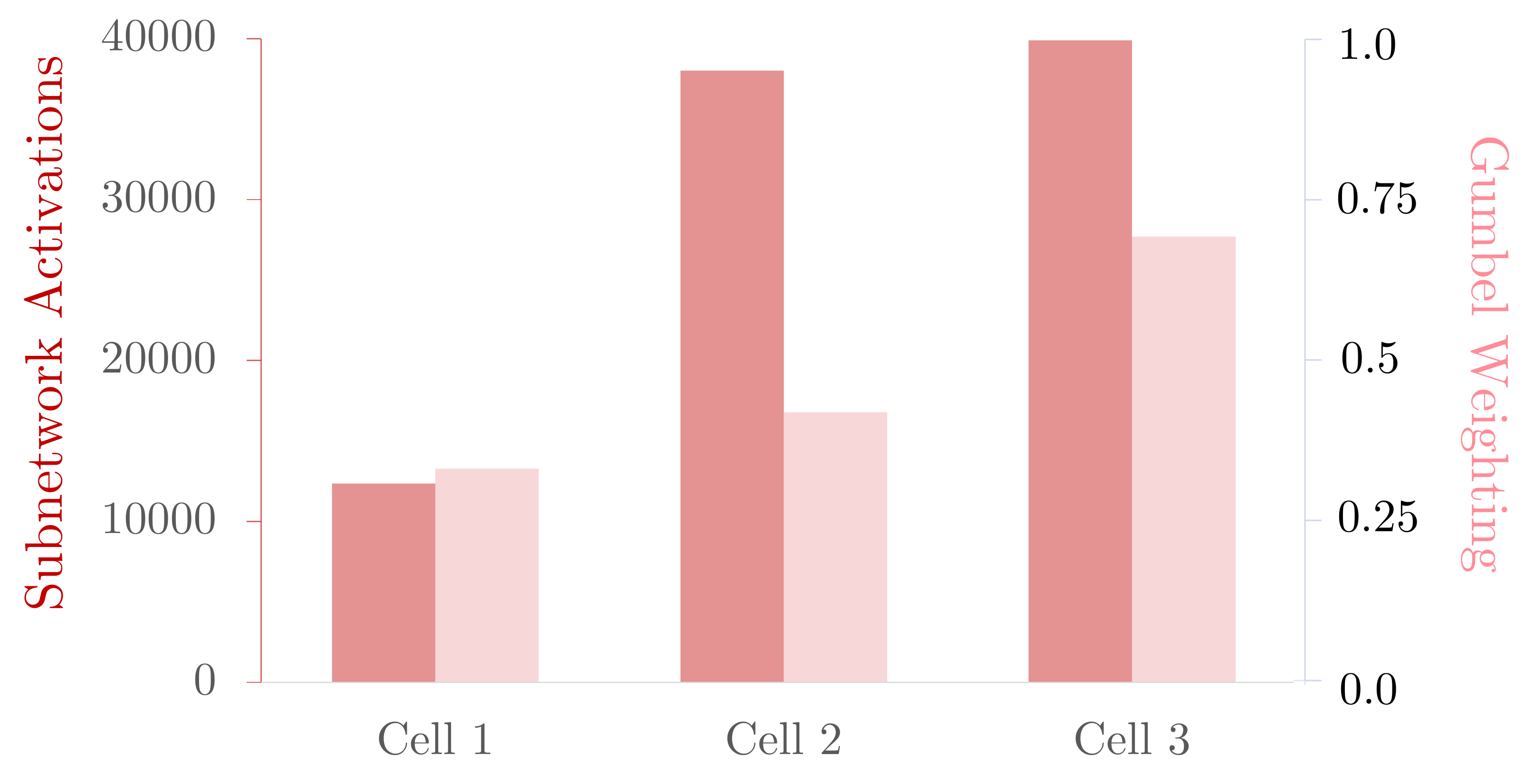}
        \caption{\textbf{Breast Cancer}: Number of activations (left) vs. average activation weight (right) per cell. Each subnetwork represents a distinct cell nuclei (10 features).}
        \label{fig:cancer-comparison}
    \end{minipage}
\hfill
\begin{minipage}[b]{0.48\textwidth}
        \centering
        \includegraphics[width=\linewidth, trim={5 5 5 5}, clip]{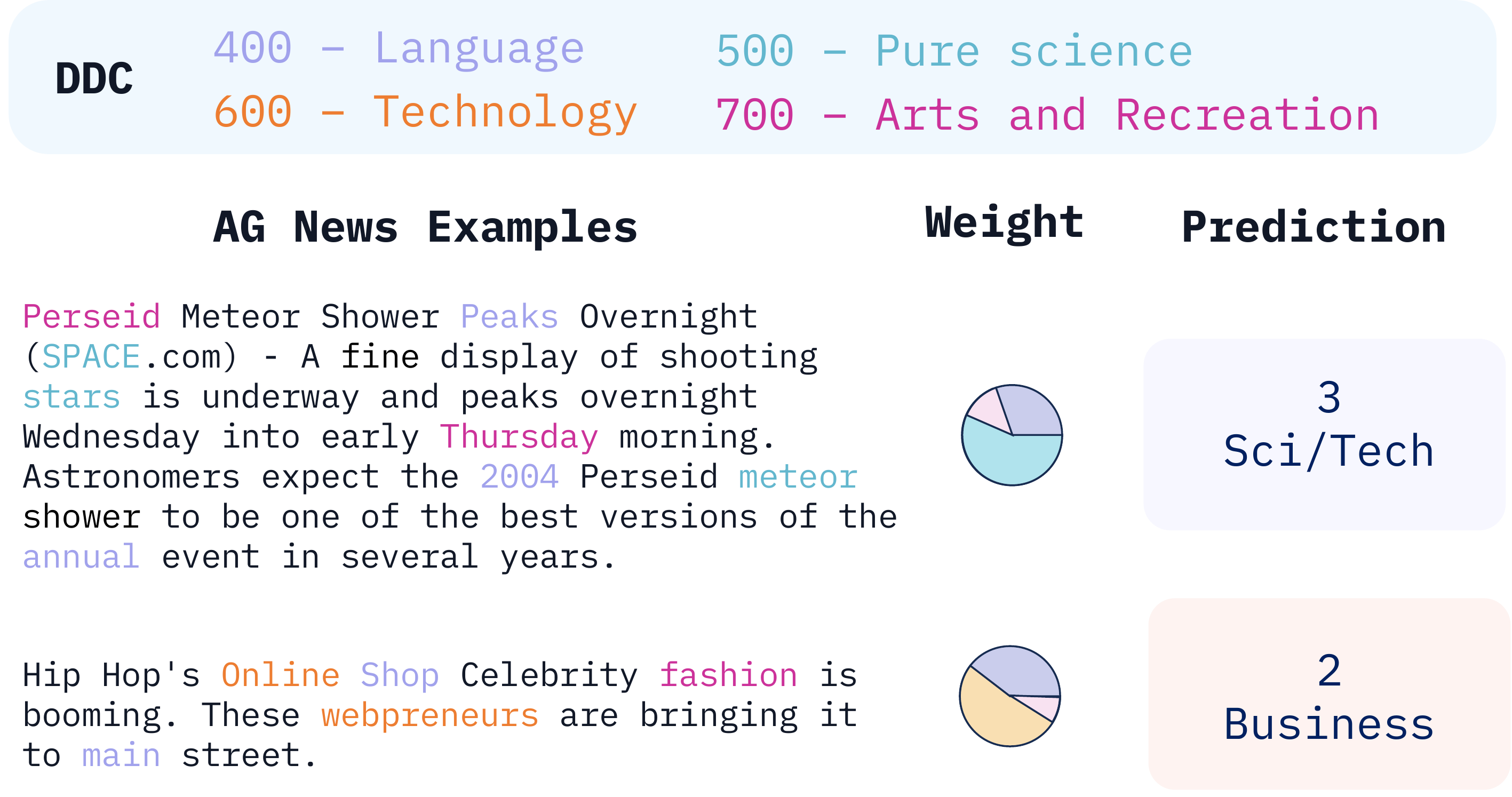}
        \caption{\textbf{AG News}: Two qualitative examples from \textbf{AG News} with \texttt{ICC} explanations, where highlighted text represents a subnetwork activation. The pie chart represents the weight of each subnetwork in the resulting prediction. }
        \label{fig:text-example}
    \end{minipage}
\end{figure}

For the \textbf{Breast Cancer} data set, the subnetworks grouping features from Cell $1$ and Cell $2$ are activated much more frequently than the third subnetwork (see Fig. \ref{fig:cancer-comparison}). Furthermore, Cell $3$ also gets activated with higher weights than the other two cells (Cell $1$: $0.25$, Cell $2$: $0.40$, Cell $3$: $0.0.70$). Smoothness and texture of the tissue images were the most important features across cells.

\textbf{Grouping Analysis for Text Datasets}
We provide two illustrative, qualitative examples from the AG News dataset in Figure \ref{fig:text-example}. In the top example (Perseid meteor shower), the words `stars', `meteor', and `SPACE' are routed to the \textit{Pure Science} (500) subnetwork with a $50\%$ activation weight, resulting in the correct prediction of `Sci/Tech' category. Likewise, for the bottom article, both the \textit{Technology} and \textit{Arts} subnetworks are highly weighted, resulting in the correct prediction of the `Business' category. Interestingly, subnetwork \textit{Language} (400) is also activated.

\textbf{Grouping Analysis for EDU Datasets}
\label{appendix:edu}

We conduct sparsity and group routing paradigm analyses on a representative course of the EDU datasets (DSP). We additionally experiment with grouping by paper and pattern over all courses, including two additional MOOC courses with low predictive performance (Structures and Ventures) featured in \cite{swamy2022meta}.

\textit{Sparsity Analysis on DSP}\label{appendix:edu_dsp_sparsity}
We conduct a sparsity analysis on the course used for the user study, DSP, and featured in several research works in the learning science community \cite{swamy-eval, swamy2022meta, boroujeni-et-al}. In line with \cite{swamy2022meta}, we evaluate the sparsity at both the 40\% and 60\% prediction setting, representing early prediction at 4 and 6 weeks of the course respectively.

\begin{table}[ht]
    \centering
    \begin{tabular}{c c c c}
        Setting & Model & 
        \begin{tabular}{c}
             Performance\\
             \begin{tabular}{|c|c|c|}
                \hline
                 $\;$ACC$\;$ & $\,$BAC$\,$ & AUC$\;$ \\
                 \hline
             \end{tabular}
        \end{tabular} & 
        \begin{tabular}{c}
             Activated Features\\
             \begin{tabular}{|c|c|}
                \hline
                 $\;\;\;\mu\;\;\;$ & $\,\,\;\sigma\;\,\,$ \\
                 \hline
             \end{tabular}
        \end{tabular}\\
        \\
        40\% & 
        \begin{tabular}{c}
            Baseline \\
            L1 (1e-5)\\
            L1 (1e-4)\\
            Annealed MSE \\
            Truncated AMSE \\
        \end{tabular} &
        \begin{tabular}{|c|c|c|}
            \hline
            0.835 & 0.653 & 0.858 \\ 
            0.790 & 0.711 & 0.817 \\ 
            0.763 & 0.748 & 0.784 \\ 
            0.768 & 0.770 & 0.823 \\ 
            0.805 & 0.743 & 0.823 \\
            \hline
        \end{tabular} & 
        \begin{tabular}{|c|c|}
            \hline
            97 & 0 \\ 
            8.54 & 0.58 \\ 
            2.20 & 1.64 \\ 
            13.26 & 3.72 \\
            5.95 & 1.03 \\
            \hline
        \end{tabular}
        \\
        60\% & 
        \begin{tabular}{c}
            \\
            Baseline \\
            L1 (1e-5)\\
            L1 (1e-4)\\
            Annealed L1 (1e-5)\\
            Annealed MSE \\
            Truncated AMSE \\
        \end{tabular} &
        \begin{tabular}{c}
            \\
            \begin{tabular}{|c|c|c|}
                \hline
                0.944 & 0.925 & 0.982 \\
                0.914 & 0.935 & 0.963 \\ 
                0.914 & 0.917 & 0.957 \\ 
                0.910 & 0.927 & 0.957 \\
                0.892 & 0.927 & 0.952 \\
                0.787 & 0.788 & 0.839 \\  
                \hline
            \end{tabular}  
        \end{tabular} & 
        \begin{tabular}{c}
            \\
             \begin{tabular}{|c|c|}
                \hline
                97 & 0 \\
                37.60 & 3.97 \\ 
                31.38 & 4.45 \\ 
                38.36 & 3.61 \\
                21.45 & 3.23 \\
                - & - \\
                \hline
            \end{tabular} 
        \end{tabular}
    \end{tabular}
    \vspace{2mm}
    \caption{\textbf{\icc Feature Gating} comparison of performance metrics between different sparsification methods next to their average and standard deviation of number of activated features for both 40\% and 60\% early success prediction settings; baseline benchmarks also provided for contrast.}
    \label{tab:fg_results}
\end{table}

An annealed mean-squared regularization proved most effective, although it activated more features on average than $L_1$-norm regularization which more effectively reduced the feature space while achieving desirable balanced accuracy performance in this setting. By truncating the initial feature space to only the activated features and using the same architecture, performance is almost maintained even though the average number of activated features per datapoint is more than halved (from around 13 to 6). However, this method was not as effective for the 60\% setting. Truncating the feature space largely reduces predictive capability (e.g. almost a 0.15 drop in balanced accuracy). L1 regularization in this case proved best. Using an annealed regularization did not significantly improve or change model performance as well. 

\textit{Multiple Grouping Paradigms on DSP}\label{appendix:edu_dsp_gr}

\begin{figure}
    \centering
    \includegraphics[width=\linewidth, trim={5 5 5 5}, clip]{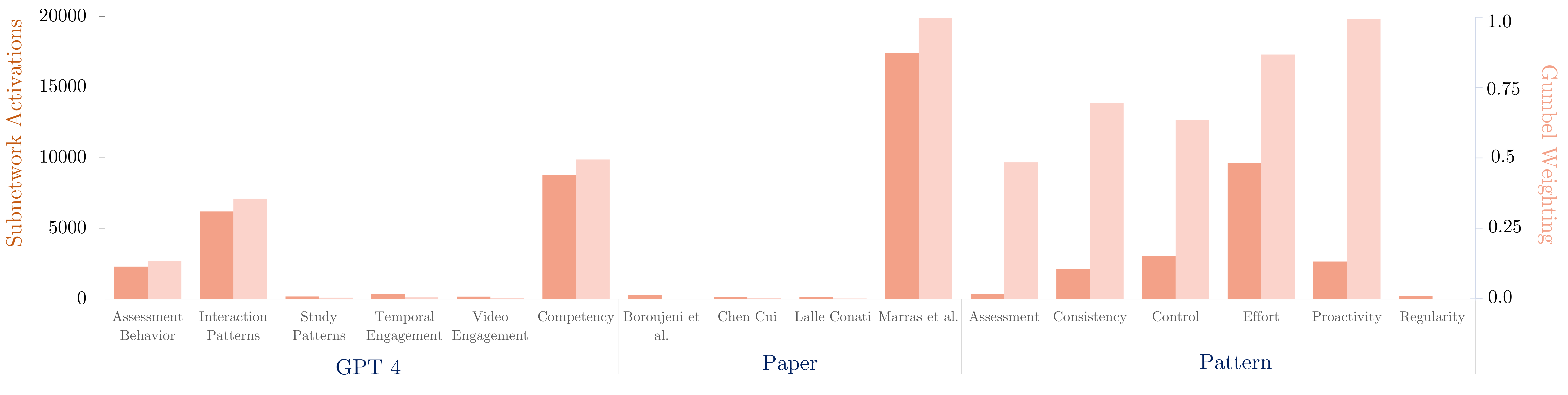}
    \caption{\textbf{EDU}: Number of subnetwork activations  (left) and Gumbel Softmax activation weights (right) across different groupings (GPT-4, Paper, Pattern) for course DSP 1 of the EDU domain.}
\label{fig:deep-dive}
\end{figure}

To illustrate the influence of different feature groupings, we conduct a deep dive for course DSP 1 of the EDU domain. Figure \ref{fig:deep-dive} illustrates the number of subnetwork activations and corresponding weights for three different groupings.

For the first two groupings (GPT-4, Paper), the subnetwork activations (number of times the route was activated) closely mirror the Gumbel Sigmoid adaptive weighting, indicating that a few networks are activated with high weights for prediction. In the \textit{group by GPT-4 }setting, we see behaviors of competency, interaction patterns, and assessment frequently activated for student pass-fail predictions. Although `interaction patterns' is the largest category (most number of features chosen by GPT-4), it still comes second to competency (focusing on student achievement). In the \textit{group by paper} setting, we see a clear preference for Marras et al. with over $17,500$ students predicted using this network (dark orange) and high weight given to the predictions from the network (light orange).

In contrast, in the third grouping (Pattern), we see a differentiation between the number of activations (dark orange) and the weight of the activations (light orange). Notably, the patterns of `Effort', `Proactivity', `Consistency' and `Control' all have higher than $50\%$ weight when they are activated, which means they contribute a lot to the overall prediction when chosen.

\textit{Grouping by paper and pattern over all courses}\label{app:fg_allcourses}

\begin{figure}[ht]
    \centering
    \includegraphics[width=0.6\linewidth, trim={5 5 5 5}, clip]{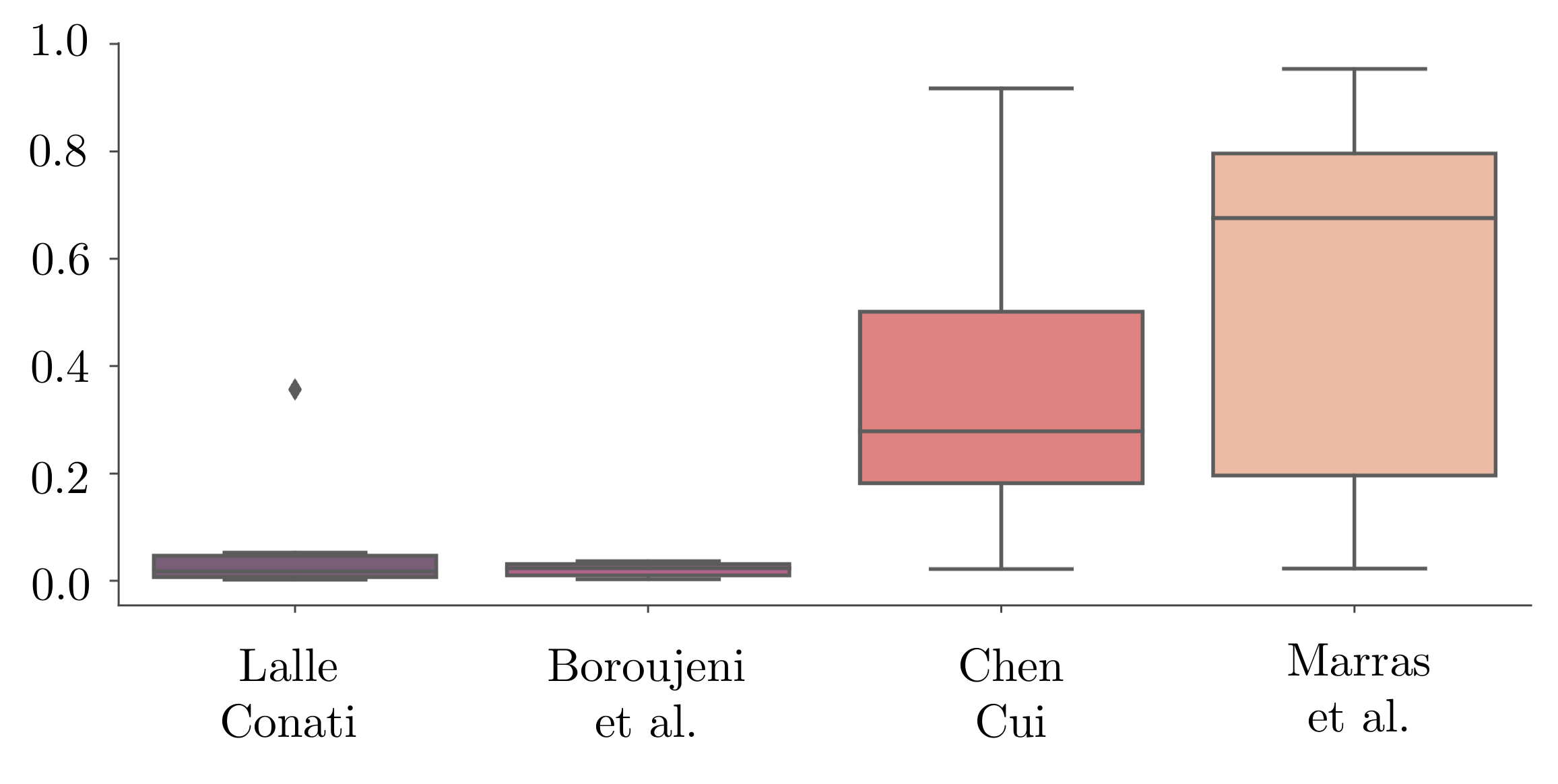}
    \caption{\textbf{EDU} analysis of \textit{group by paper} routing averaged over all six courses for each paper grouping. The y-axis represents the proportion of points for which the subnetwork is activated.}
    \label{fig:edu-group-by-paper}
\end{figure}

In Figure \ref{fig:edu-group-by-paper}, we see \icc routing by research paper (grouping the features based on the paper they were proposed in). The Marras et al. and Chen Cui feature sets have clearly been identified as important over the majority of courses, echoing findings in other learning science literature using BiLSTM and random forest architectures \cite{marras-vignoud-kaser-2021, chen-cui, swamy2023trusting}. The large standard deviations in the box-plots indicate that for at least some courses (in this case Structures and Venture), Chen Cui and Marras were not found significantly important. Notably, the same courses that have low accuracies on routing in \ref{tab:all_results} are those that have low scores on the two most popular feature sets, showing a consensus among performant \icc models and a validation of the identification of importance. 

\begin{figure*}[ht]
    \centering
    \includegraphics[width=\textwidth, trim={5 5 5 5}, clip]{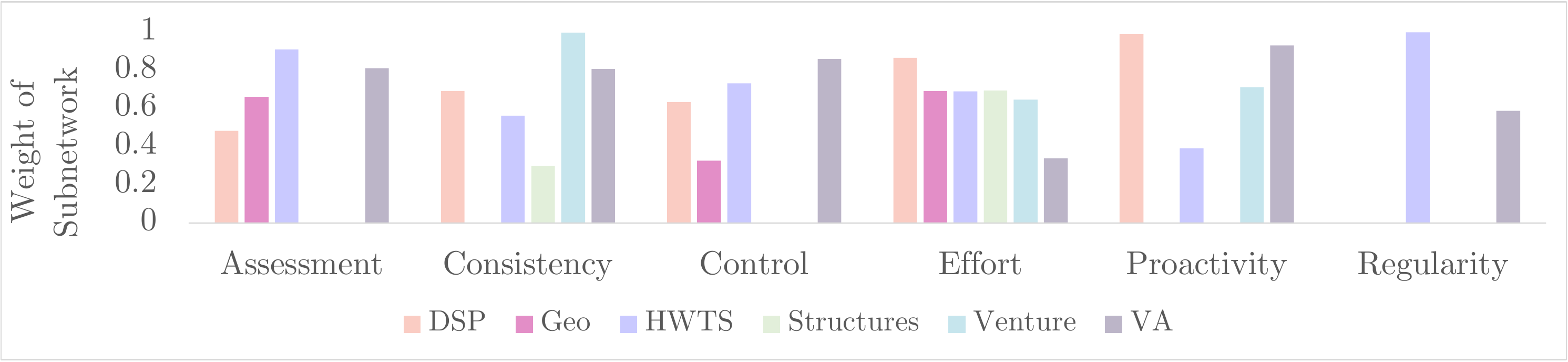}
    \caption{\textbf{EDU} Weighting distribution across subnetworks using \textit{group by pattern} for 6 courses.}
    \label{fig:edu-network-weights}
\end{figure*}

\begin{figure}[ht]
    \begin{minipage}[]{0.55\textwidth}
    \centering
    \resizebox{\textwidth}{!}{
        \begin{tabular}{@{}lll@{}}
        \toprule
        \multicolumn{1}{c}{\textbf{Model}} & \multicolumn{1}{c}{\textbf{B. Cancer}} & \multicolumn{1}{c}{\textbf{Synthetic}} \\ \midrule
        Random Forest (RF) & \textit{92.98 ± 4.70} & 85.32 ± 3.48 \\ \midrule
        \begin{tabular}[c]{@{}l@{}}Non-interpretable \\ base module (NN)\\ (SHAP, LIME)\end{tabular} & 89.70 ± 1.05 & 86.67 ± 0.31 \\ \midrule
        NAM & 88.77 ± 7.31 & 87.85 ± 1.31 \\ \midrule
        SENN Features & 80.52 ± 6.21 & 83.67 ± 1.86 \\ \midrule
        SENN Concepts & 85.26 ± 1.03 & \textit{89.51 ± 0.51 }\\ \midrule
        InterpretCC FG & 78.19 ± 3.54 & 84.67 ± 4.04 \\ \midrule
        InterpretCC Top K & 84.66 ± 3.02 & \textbf{90.83 ± 1.93} \\ \midrule
        InterpretCC GR & \textbf{94.85 ± 1.25} & 89.47 ± 2.89 \\ \bottomrule
        \end{tabular}}
    \end{minipage}%
    \hfill
    \begin{minipage}[]{0.4\textwidth}
        \centering
        \includegraphics[width=\linewidth]{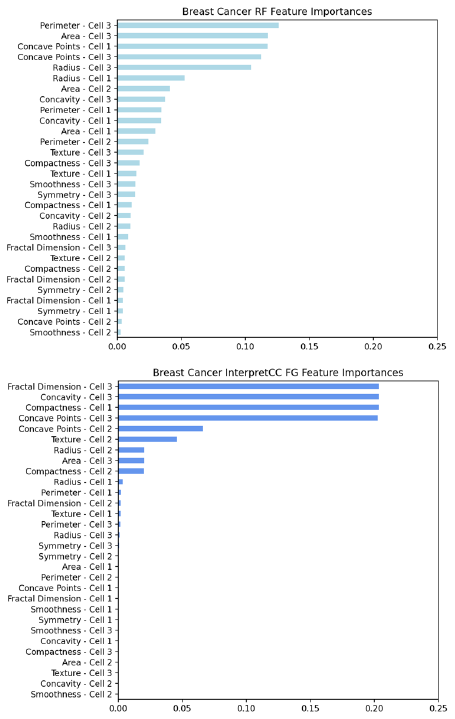}
    \end{minipage}
    \caption{\label{fig:icc-vs-rf}\textbf{\icc vs. RF Tabular Comparison}. (left) Performance of models on Breast Cancer and Synthetic datasets. (right) Features chosen in explanations from RF (light blue) vs. \icc for Breast Cancer dataset (dark blue).}
\end{figure}

In Fig. \ref{fig:edu-network-weights}, we see a widely varying distribution of patterns selected across courses, showcasing the ability of \icc to adaptively select subnetwork weights depending on the dataset.

\section{\icc's relationship with interpretability}
We discuss a comparison of a traditionally interpretable machine learning model (random forest) with \icc, and we propose an architecture extension to make \icc even more interpretable (at the expense of additional complexity).

\textbf{\icc Tabular Comparison with a Traditional Interpretable Model}

In Fig. \ref{fig:icc-vs-rf}, the table (left) shows that all models, in comparison with a Random Forest model (RF), overlap in 95\% CI. The performance of a tuned RF is in line with deep learning approaches. However, the RF uses 100 trees with average depth 7.37 +/- 1.16 for Breast Cancer and 1000 trees with average depth 6.21 +/- 2.03 for Synthetic data, so is not simple to understand the decision path.

We then compare which features RF selects as important (light blue) vs. which features ICC FG selects as important (dark blue): ICC’s feature importances are much sparser.

\textbf{\icc Description of Faithfulness}
\label{app:additional-faithfulness}

As inspired from the OpenXAI benchmark \cite{agarwal2022openxai}:
\begin{itemize}[leftmargin=1em]
    \item \textbf{Feature Agreement (FA):} Computes the fraction of top-K features shared between an explanation and [the underlying data pattern].
    \item \textbf{Rank Agreement (RA):} Measures the fraction of top-K features that are not only shared but also appear in the same rank order in both the explanation and the [underlying data pattern].
\end{itemize}

We define \textbf{Ground Truth Alignment} as the fraction of features shared between an explanation and the underlying data pattern (K=total number of features). We define \textbf{Ground Truth Faithfulness} as the fraction of features shared between an explanation and the ground truth explanation (exactly what the model is using as a decision process).

As an extension of the results in Table \ref{tab:synth}, we conduct a Pairwise Rank Agreement analysis. \textbf{Pairwise Rank Agreement (PRA)} assesses whether the relative ordering of feature pairs is consistent between the explanation and underlying data pattern, computing the fraction of pairs with the same relative importance. Comparing the eight models on the synthetic data for PRA, the results are: 69.71 +/- 3.89 (ICC FG), 74.55 +/- 1.21 (ICC GR), 71.99 +/- 3.25 (SENN Features), 45.30 +/- 7.49 (SENN Concepts), 70.16 +/- 5.63 (NAM), 82.13 +/- 7.21 (IG), 73.31 +/- 1.91 (LIME), and 72.93 +/- 3.71 (SHAP). As this metric is simply a less-strict version of Rank Agreement and the order of important features is very important in downstream human-centric tasks deriving from explanations, it is not included in the main results. Notably, all 95\% CI overlap again, except for ICC GR and SENN Concepts, where SENN Concepts is significantly worse than ICC GR.

\textbf{\icc Extension for an Interpretable Discriminator Network}
\label{sec:interpretable-discriminatory}

\icc can enable an interpretable discriminator network simply by using a set of decision trees (or SVMs, LRs, any other traditionally interpretable model) instead of a neural network (FG: one model per feature or GR: one model per concept). This additional interpretability would come at the expense of model/explanation complexity and perhaps reduced accuracy as it doesn’t take into account cross-feature interactions, but would allow us to make statements like: “Concept X was chosen because feature A > value, value > B < value, C > value, D = 0, and E > value. Concept X contributes 35\% to the prediction.”

\textbf{Concrete Implementation Details}: For \icc FG, each tree would predict a binary decision between “keep this feature” or “don’t keep this feature” with the input of the full feature space. For \icc GR, each SVM would assign a score between 0 and 1 regarding whether to “keep this concept” with the input of the full feature space (the scores would then be normalized across concepts/features). These would be trained together with the predictive network and a sparsity criterion in exactly that same way as we have shown with \icc: all that changes is the choice of which model to use for the discriminator network. A simpler NN solution is to add an attention layer to the discriminator for some opaque interpretability.

However, we do not include this approach in the model as it makes both the explanation for the downstream user more complex and the model itself more complex; we choose to focus instead on the design criteria of optimizing human actionability through sparsity and simplicity.

\textbf{\icc Comparison to Extractive Rationale Methods}
\label{sec:rationale}

\textcolor{black}{In comparison to InterpretCC, extractive-rationale methods from the NLP community like FRESH (Jain et al. 2020) provide explanations that are more tailored for the text domain. However, these methods have several weaknesses, illustrated in a comparison of FRESH to InterpretCC: 1) These methods are often harder to generalize to new tasks, e.g. FRESH needs to train 3 new models sequentially instead of 1 model with two parts in parallel for every new setting. 2) There is often bias included in determining the initial importance or saliency scores (LIME, Attention, Gradients used in FRESH) instead of letting the model learn directly. 3) These models are longer, larger, and more complex to train over InterpretCC’s simpler architectures. 4) Lastly, and most importantly, methods like FRESH require the selection of a contiguous section of text as “rationale”, in contrast to InterpretCC’s mapping of words to concepts from the Dewey Decimal System (or any other grouping methodology). While this contiguous selection makes sense in the text domain, for tabular, or even time-series (beyond anomaly detection) data, this kind of explanation is not suitable or human-friendly. Therefore, we did not implement FRESH on the other modality experiments as it would be fundamentally unsuited to the data format.}

\textcolor{black}{For quantitative experiments (showcased in Table \ref{tab:all_results}, we tuned the FRESH model over rationale length and learning rate, and ran experiments with 5 random seeds with the exact architecture used in the paper. Notably, FRESH experiments use BERT models, while ICC experiments use the smaller DistillBERT model. For SST, FRESH uses a 30\% rationale size with an average accuracy of 82.05\% +/- 0.56\%, where ICC FG has 88.21\% +/- 3.41\%, ICC TopK has 92.98\% +/- 0.88\%, and ICC GR has 91.75\% +/- 1.86\%, which are significantly more performant. As SST has 7 words on average, ICC FG selects a larger explanation size, on average 3 or 4 words (indicated in Figure 2), but maintains a higher accuracy. For AG News, FRESH uses a 20\% average rational length, with performance of 88.73\% +/- 0.69\%. This is comparable to the ICC GR numbers 90.35\% +/- 1.07\% and the the ICC Top K routing approach with 87.25\% +/- 2.48\%, and more performant than the ICC FG approach 85.72\% +/- 5.31\% (although the high variation in the FG approach is to be noted), and the 95\% CIs overlap showing that the approaches are similar.}

\textcolor{black}{Other methods from the extractive-rationale community, like the methods proposed by \cite{bastings2019interpretable}, \cite{yu2019rethinking}, and \icc take fundamentally different approaches to model interpretability, each with distinct trade-offs. Bastings et al. focus on token-level rationale selection using binary gates optimized through REINFORCE, which allows for precise, fine-grained explanations but suffers from optimization challenges and limited applicability beyond text-based tasks. Yu et al. extend this idea by introducing a generator-predictor framework that enforces complementarity between selected and unselected features. While this ensures higher rationale quality, it also significantly increases model complexity and computational cost. \icc diverges by leveraging human-defined feature groups and a unified, end-to-end architecture, balancing interpretability and scalability. Unlike token-based approaches, \icc focuses on concept-level explanations that generalize across domains like tabular and time-series data, enabling more actionable insights. By emphasizing cross-feature interactions and the integration of domain knowledge, \icc differs from approaches in the extractive-rationale community.}

\begin{table}[h!]
\centering
\resizebox{\textwidth}{!}{%
\begin{tabular}{@{}lllll@{}}
\toprule
\multicolumn{1}{c}{\textbf{Aspect}} &
  \multicolumn{1}{c}{\textbf{\cite{bastings2019interpretable}}} &
  \multicolumn{1}{c}{\textbf{\cite{yu2019rethinking}}} &
  \multicolumn{1}{c}{\cite{jain2020learning}} &
  \multicolumn{1}{c}{\textbf{InterpretCC}} \\ \midrule
\textbf{Feature Selection} &
  Token-level binary gates &
  Token-level generator-predictor &
  Contiguous text spans as rationales &
  \begin{tabular}[c]{@{}l@{}}Feature-level binary gates or \\ group-level selection (domain-informed)\end{tabular} \\ \midrule
\textbf{Architecture} &
  REINFORCE optimization &
  \begin{tabular}[c]{@{}l@{}}Generator + predictor\\ (cooperative)\end{tabular} &
  \begin{tabular}[c]{@{}l@{}}Generator + predictor \\ (extractive rationales)\end{tabular} &
  \begin{tabular}[c]{@{}l@{}}Single end-to-end model \\ with routing\end{tabular} \\ \midrule
\textbf{Domain Focus} &
  Text-focused &
  Text-focused &
  Text-focused &
  General (text, tabular, time-series) \\ \midrule
\textbf{Explanation Type} &
  Token rationales &
  \begin{tabular}[c]{@{}l@{}}Token rationales \\ with complementarity\end{tabular} &
  Contiguous extractive rationales &
  \begin{tabular}[c]{@{}l@{}}Concept-based, \\ human-centric rationales\end{tabular} \\ \midrule
\textbf{\begin{tabular}[c]{@{}l@{}}Optimization \\ Complexity\end{tabular}} &
  High due to discrete variables &
  \begin{tabular}[c]{@{}l@{}}Moderate with \\ generator constraints\end{tabular} &
  \begin{tabular}[c]{@{}l@{}}Moderate due to \\ sequential training\end{tabular} &
  Low with Gumbel-Softmax routing \\ \midrule
\textbf{Main Strength} &
  Fine-grained token selection &
  Explicitly enforces rationale quality &
  Faithfulness to predictions &
  Human-friendly explanations \\ \midrule
\textbf{Main Weakness} &
  \begin{tabular}[c]{@{}l@{}}Optimization variance;\\ text-only\end{tabular} &
  \begin{tabular}[c]{@{}l@{}}High model complexity;\\ text-only\end{tabular} &
  \begin{tabular}[c]{@{}l@{}}Task-specific rationales; \\ text-only\end{tabular} &
  \begin{tabular}[c]{@{}l@{}}Requires interpretable feature \\ group design (human or LLM)\end{tabular} \\ \bottomrule
\end{tabular}}
\caption{Design comparison of \icc to Extractive Rationale Methods.}
\label{tab:comparison_methods}
\end{table}

\textbf{\icc Extension for Cross-Feature or Raw Modality Interactions}
\label{sec:graph-nn}

\textcolor{black}{Graph-based models could be very useful in the \icc architecture for raw time series data, with a natural extension in the discriminator stage. The message passing graph network and concept activation vector approach showcased with RIPPLE \cite{asadi2023ripple} could be used to define a sparse adjacency matrix as opposed to a sparse vector on the input. This could then relate to interaction-based subnetworks, referring to multiple features in each concept used in the explanation. Alternatively, graph models like RAINDROP \cite{zhanggraph} or SGP \cite{cini2023scalable} could be used in the predictive module stage, where each subnetwork is a graph model only focused on the specific modalities or features passed into the subnetwork. Using the sparseness enabled by the discriminator layers, only a few of the graph models would be activated for each point’s prediction.}

\textcolor{black}{Without this extension, it is possible \icc could be used directly on raw time series data to simply specifying the ``concepts`` as fixed or relative time intervals (analogous to anomaly detection), enabling users to identify which parts of the time-series were used in the prediction. This could answer questions like: ``Was my behavior in week 1 useful in predicting that I performed well on the exam?"). This significant time interval approach could be further extended to modalities like raw speech or video, although using modality-specific expert grouping methodologies would be preferred, as this would be more useful for downstream user actions.}

\chapter{Appendix: ILLuMinaTE (Chapter 9)}

%%%%%%%%%%%%%%%%%%%%%%%%%%%%%%%%%%%%%%%
\section{Reproducibility Details}
%%%%%%%%%%%%%%%%%%%%%%%%%%%%%%%%%%%%%%%
\label{appendix:reproducibility}

All code to replicate our student models and explainability results are provided in our repository: \texttt{https://github.com/epfl-ml4ed/iLLuMinaTE}. Any details not covered in the following reproducibility notes can also be found directly in the repository.

\vspace{1mm} \noindent \textbf{Student Model.} We input 45 student features in line with the exact extraction procedure from \cite{marras2021can, swamy2022meta} into a neural network architecture consisting of a \emph{Masking} layer (with a mask value of $-1$), followed by two Bidirectional LSTM (BiLSTM) layers with 64 and 32 units, respectively, and a loopback of $3$. The final layer is a \emph{Dense} layer with a Sigmoid activation function, which has a single output unit. This model predicts the probability that a student will fail the course. \cite{swamy2022meta} find that a BiLSTM architecture outperforms both traditional machine learning models (e.g., Support Vector Machines, Logistic Regression, Random Forest) and other deep learning models (e.g., Dense Fully-Connected Networks, RNNs, LSTMs, CNNs, and BiLSTMs) in the context of student performance prediction for online courses. They report a 26.8\% increase over traditional models (as measured by validation set for the DSP course). In our grid search for the best BiLSTM architecture, we tested various hidden size configurations, including \{32, 64, 128, 256, 32-32, 64-32, 128-32, 128-64, 128-128, 256-128, 64-64-32, 128-64-32, 64-64-64-32-32\}. The 64-32 configuration achieved the highest balanced accuracy for predicting outcomes in the \textit{DSP 1} course. We used the TensorFlow library \cite{abadi2016tensorflow} to train our models. Model training took approximately 15 minutes per model on a c2-standard-8 machine on Google Cloud (Linux Debian OS, 8vCPUs, 32 GB of RAM, 150 GB of disk). Each model was trained for 20 epochs, and the best performing model checkpoint was saved.

\vspace{1mm} \noindent \textbf{LIME.} The explanation highlights up to the 20 most influential features. The \texttt{mode} is set to \texttt{'classification'}, indicating that the function is tailored for the binary student performance prediction task. \texttt{discretize\_continuous} is \texttt{True}, meaning continuous features will be discretized to make the explanations more interpretable. The \texttt{num\_samples} parameter is set to 5000, specifying the number of perturbed samples LIME will generate to approximate the local decision boundary of the model. The \texttt{distance\_metric} is set to \texttt{'euclidean'}, which is used to measure the distance between the original instance and the perturbed samples. Lastly, the \texttt{sampling\_method} is set to \texttt{'gaussian'}, indicating that the perturbed samples will be drawn from a Gaussian distribution. The LIME setting is the exact same as used in \cite{swamy2022evaluating, swamy2023trusting}, but is expanded to select more features than the default (20 instead of 10) to not over-narrow the search space for MC-LIME.

\vspace{1mm} \noindent \textbf{MC-LIME.} The \texttt{init\_processing} function preprocesses the LIME results dataset by extracting feature names and reformatting the DataFrame. The \texttt{calculate\_std\_dev\_step\_sizes} function computes step sizes for feature modification based on a fraction of each feature's standard deviation, ensuring that changes applied to features during MC-LIME are proportionate. In the \texttt{get\_lime\_results} function, LIME important features and corresponding student-specific features are extracted, and step sizes are calculated to assess the sensitivity of model predictions. The \texttt{MC\_LIME} function iteratively modifies student features in small steps (guided by calculated step sizes) within specified bounds (0 to 1) to determine the minimal set of feature changes that could alter the model's prediction. The key hyperparameters in this procedure include the fraction of standard deviation used for step sizes, a prediction threshold (defaulting to 0.5), and the maximum group size for feature combinations (defaulting to 3), which controls the complexity and granularity of the search for influential features. The script ensures that feature modifications are meaningful and within realistic bounds, aiming to identify the minimal changes required to change a student's predicted outcome.

\vspace{1mm} \noindent \textbf{CEM.} The \texttt{pn\_all} function generates pertinent negative (PN) explanations for a set of instances using the Contrastive Explanation Method (CEM) applied to our underlying BiLSTM model. The function is configured with several key hyperparameters: \texttt{kappa} (set to 0.0) controls the minimum probability difference needed to minimize the first loss term, \texttt{beta} (0.1) weights the L1 loss term, and \texttt{gamma} (100) weights the optional auto-encoder loss term. The initial weight \texttt{c\_init} (1.0) influences the loss term that encourages a prediction different from the original class, with \texttt{c\_steps} (10) updates and \texttt{max\_iterations} (1000) iterations per value of \texttt{c}. The \texttt{feature\_range} is set based on the minimum and maximum values of the features, and gradient clipping is applied within the range \texttt{clip} (-1000.0, 1000.0). The learning rate \texttt{lr} is initialized to 0.01, and \texttt{no\_info\_val} is set to -1.0, representing a non-informative value for prediction. The function iterates over the specified instances, reshaping the features and generating explanations through CEM, with results stored in \texttt{changes}, \texttt{explanations}, and \texttt{final\_num\_instances}. Errors during processing are handled and logged, ensuring robustness in explanation generation. The CEM setting is the exact same as used in \cite{swamy2022evaluating, swamy2023trusting}.

\vspace{1mm} \noindent \textbf{LLMs.} We use three LLMs through these experiments: GPT-4o (gpt-4o-2024-05-13), Gemma2 9b (google/gemma-2-9b-it, July 2024), and Llama3 70B (meta/meta-llama-3-70b-instruct, April 18 2024 release). We use the instruct variations of each of these models, using the OpenAI API\footnote{https://platform.openai.com/} for GPT-4o and Groq Cloud\footnote{https://console.groq.com/} for Gemma2 and Replicate \footnote{https://replicate.com/} for Llama3. The visualizations for the user study (Sec. \ref{sec:student-preferences}, Appendix \ref{appendix:study-materials}) were generated through the ChatGPT user interface with underlying GPT-4o. It can also be generated through running LLM code snippets, but we chose to use the ChatGPT interface to remain closest to an actual teacher's workflow (using models as presented with a zero-shot chain-of-prompts, ensuring minimal post-processing steps).

\vspace{1mm} \noindent \textbf{Dataset.} 
Our objective is to predict student success during the initial weeks of three massive open online courses (MOOCs) by analyzing students' clickstream data. The log data collected from students includes detailed interactions with videos (such as play, pause, fast forward, and seek actions) as well as quiz-related events (such as submissions). To ensure the protection of student privacy, all student data is fully anonymized and the dataset is kept confidential, as required by the ethical guidelines outlined in HREC 058-2020/10.09.2020 and HREC 096-2020/09.04.2022. 

Formally, given the interactions $I_s$ generated by students $S$ up until course week $w$, we construct a matrix $H \subset \mathbb{R}^{|S| \times w \times f}$. This indicates that each feature in the feature set is computed for each student on a weekly basis, where $f \in \mathbb{N}$ represents the dimensionality of the feature set. We extract these features for each student $s$ and concatenate the feature sets to produce the final combined behavioral feature vector $h_s$ for each student. The complete matrix of features is defined as $H \in \mathbb{R}^{|S| \times w \times 42}$, with $H = [H_1 \cdot H_2 \cdot H_3 \cdot H_4]$ (where $\cdot$ represents concatenation). Due to the varying scales of the features, we apply min-max normalization to each feature in $H$, scaling them between 0 and 1, considering all students and weeks for that feature. The appendix table in \cite{swamy2022meta} lists all the features used in this dataset. Detailed statistics for each course can be found in Table \ref{tab:EDU_course_stats}. 

\begin{table}[h!]
\centering
\resizebox{\textwidth}{!}{%
\begin{tabular}{@{}llcccccc@{}}
\toprule
\textbf{Title} & \textbf{Identifier} & \textbf{Topic} & \textbf{Level} & \textbf{Language} & \textbf{No.\ Weeks} & \textbf{No.\ Students} & \textbf{Passing Rate [\%]}  \\ \midrule
Digital Signal Processing & DSP &  CS & MSc & English & 10 & 4,012 & 23.1 \\
Éléments de Géomatique & Geo & Math & BSc & French & 11 & 452 & 45.1 \\
Villes Africaines & VA &  SS & BSc & En/Fr & 12 & 5,643 & 9.9 \\
\bottomrule
\end{tabular}}
\vspace{2mm}
\caption{Course Details and Statistics.\\
Topic abbreviations: Math: Mathematics; NS: Natural Science; CS: Computer Science; SS: Social Science; Arch: Architecture; Bus: Economics and Business.}
\label{tab:EDU_course_stats}
\end{table}

%%%%%%%%%%%%%%%%%%%%%%%%%%%%%%%%%%%%%%%
\section{Prompting Framework}

In this Appendix, we present the explanation selection prompts (Sec. \ref{sec:explanation-selection}), explanation presentation prompts (Sec. \ref{sec:explanation-presentation}), and explanation visualization prompts (used for the hybrid explanation user study in Sec. \ref{sec:student-preferences}). We then discuss the development procedure of these prompts, and the exact frameworks and models used such that the results can be replicated.
%%%%%%%%%%%%%%%%%%%%%%%%%%%%%%%%%%%%%%%
\label{appendix:prompts}
\subsection{Explanation Selection Prompts}
\label{appendix:exp-sel-prompt}

The explanation selection template structure is highlighted in Section \ref{sec:explanation-selection}, and detailed again here. We provide an example of the model description, features description, explainer description (CEM), and course description (DSP 1). The details of the theory instructions are included below for each prompting strategy. 

% \begin{mdframed}[backgroundcolor=orange!10, linecolor=black, linewidth=1pt, innertopmargin=4pt, innerbottommargin=8pt, innerrightmargin=10pt, innerleftmargin=10pt, roundcorner=10pt]

\begin{mdframed}[
    backgroundcolor=orange!10,
    linecolor=black,
    linewidth=1pt,
    innertopmargin=10pt,
    innerbottommargin=10pt,
    innerrightmargin=10pt,
    innerleftmargin=10pt,
    roundcorner=10pt,
    frametitlebackgroundcolor=gray!30,
    frametitlealignment=\raggedright 
]
\subsubsection*{Explanation selection template}
You are an AI assistant that analyzes struggling students behavior to help them in their learning trajectories and facilitate learning in the best possible way. 
You have the following information to help you in your goal:
\begin{itemize}
    \item A model prediction of student performance at the end of the course, in the form of “pass” or “fail”.
    \item A post-hoc explainable AI approach that identifies which features are important to this student’s prediction.
    \item Data in the form of student's features over 5 weeks that were used by the model. You will see always the most relevant features selected by \textcolor{brandred}{\textit{\{explainer\}}} \textcolor{darkblue}{CEM}.
    \item Data in the form of student's features over 5 weeks that were used by the model. You will see always the most relevant features selected by \textcolor{brandred}{\textit{\{explainer\}}} \textcolor{darkblue}{CEM}.
    \item The course \textcolor{brandred}{\textit{\{course\_name\}}} \textcolor{darkblue}{Digital Signal Processing 1} content and structure.
    \item Detailed instructions on how to reason.

\end{itemize}

\vspace{1em}

\textcolor{brandred}{\textit{\{model\_description\}}}
\textcolor{darkblue}{The model you are using is a recurrent neural network that is trained on predicting the student performance at the end of the course, in the form of “pass” or “fail”. The features of that the model are using are derived from student behavior:\\}

\textcolor{brandred}{\textit{\{features\_description\}}} \\ \\
\textcolor{darkblue}{\textbf{delay\_lecture}: The average delay in viewing video lectures after they are released to students.}

\textcolor{darkblue}{\textbf{ratio\_clicks\_weekend\_day}: The ratio between the number of clicks on the weekend and the weekdays.}

\textcolor{darkblue}{\textbf{total\_clicks}: The number of clicks that a student has made overall.}

\textcolor{darkblue}{\textbf{total\_clicks\_problem}: The number of clicks that a student has made on problems this week.}

\textcolor{darkblue}{\textbf{total\_clicks\_video}: The number of clicks that a student has made on videos this week. The number of times a student clicked on a video (load, pause, play, forward).}

\textcolor{darkblue}{\textbf{total\_clicks\_weekday}: The number of clicks that a student has made on the weekdays.}

\textcolor{darkblue}{\textbf{total\_clicks\_weekend}: The number of clicks that a student has made on the weekends.}

\textcolor{darkblue}{\textbf{competency\_strength}: The extent to which a student passes a quiz getting the maximum grade with few attempts.}

\textcolor{darkblue}{\textbf{competency\_alignment}: The number of problems this week that the student has passed.}

\textcolor{darkblue}{\textbf{content\_alignment}: The number of videos this week that have been watched by the student.}

\textcolor{darkblue}{\textbf{competency\_anticipation}: The extent to which the student approaches a quiz provided in subsequent weeks.}

\textcolor{darkblue}{\textbf{content\_anticipation}: The number of videos covered by the student from those that are in subsequent weeks.}

\textcolor{darkblue}{\textbf{student\_speed}: The average time passed between two consecutive attempts for the same quiz.}

\textcolor{darkblue}{\textbf{student\_shape}: The extent to which the student receives the maximum quiz grade on the first attempt.}

\textcolor{darkblue}{\textbf{regularity\_periodicity\_m1}: The extent to which the hourly pattern of the user's activities repeats over days.}

\textcolor{darkblue}{\textbf{regularity\_peak\_dayhour}: The extent to which students' activities are centered around a particular hour of the day.}

\textcolor{darkblue}{\textbf{number\_sessions}: The number of unique online sessions the student has participated in.}

\textcolor{darkblue}{\textbf{time\_sessions\_mean}: The average of the student's time per session.}

\textcolor{darkblue}{\textbf{time\_sessions\_sum}: The sum of the student's time in sessions.}

\textcolor{darkblue}{\textbf{time\_sessions\_std}: The standard deviation of the student's time in sessions.}

\textcolor{darkblue}{\textbf{time\_between\_sessions\_std}: The standard deviation of the time between sessions of each user.}

\textcolor{darkblue}{\textbf{time\_in\_problem\_sum}: The total (cumulative) time that a student has spent on problem events.}

\textcolor{darkblue}{\textbf{time\_in\_video\_sum}: The total (cumulative) time that a student has spent on video events.}

\textcolor{darkblue}{\textbf{total\_clicks\_Video}: The number of clicks that a student has made on videos this week. The number of times a student clicked on a video (load, pause, play, forward).}

\textcolor{darkblue}{\textbf{total\_clicks\_Video.Load}: The number of times a student loaded a video.}

\textcolor{darkblue}{\textbf{frequency\_action\_Video.Load}: The frequency between every \texttt{Video.Load} action and the following action.}

\textcolor{darkblue}{\textbf{weekly\_prop\_watched\_mean}: The ratio of videos watched over the number of videos available.}

\textcolor{darkblue}{\textbf{weekly\_prop\_replayed\_mean}: The ratio of videos replayed over the number of videos available.}

\textcolor{brandred}{\textit{\{explainer\_description\}}}\\ \\
\textcolor{darkblue}{We use CEM Counterfactuals as our explainable AI approach, which finds the smallest number of changes necessary to change a prediction from student failure to student success. The output is the minimal difference in the feature values that would change the prediction.}

\textcolor{brandred}{\textit{\{course\_description\}}}

\textcolor{darkblue}{The course the student is taking is Digital Signal Processing 1, which is a Master’s level course over 10 weeks under the topic of Electrical Engineering and Computer Science. This is the course content:}

\textcolor{darkblue}{\textbf{WEEK 1} \\
SKILLS: Digital Signals \\
TOPICS: Welcome to the DSP course, Introduction to signal processing }

\textcolor{darkblue}{\textbf{WEEK 2} \\
SKILLS: Digital Signals \\
TOPICS: Discrete time signals, The complex exponential, The Karplus-Strong Algorithm}

\textcolor{darkblue}{\textbf{WEEK 3} \\
SKILLS: Hilbert (Linear Alg.) \\
TOPICS: Motivation and Examples, From Euclid to Hilbert, Hilbert Space, properties and bases, Hilbert Space and approximation}

\textcolor{darkblue}{\textbf{WEEK 4} \\
SKILLS: DFT, DTFT DFS, FFT \\
TOPICS: Exploration via a change of basis, The Discrete Fourier Transform (DFT), DFT Examples, DFT, DFS, DTFT, DTFT formalism, Relationship between transforms, Sinusoidal modulation, FFT: history and algorithms}
 
\textcolor{darkblue}{\textbf{WEEK 5} \\
SKILLS: Ideal Filters, Filter Design \\
TOPICS: Linear Filters, Filtering by example, Filter stability, Frequency response, Ideal filters, Filter design - Part 1: Approximation of ideal filters, Realizable filters, Implementation of digital filters, Filter design - Part 2: Intuitive filters, Filter design - Part 3: Design from specs, Real-time processing, Dereverberation and echo cancelation}

\textcolor{darkblue}{\textbf{WEEK 6} \\
SKILLS: Modulation, Interpolation \& Sampling \\
TOPICS: Introduction to continuous-time paradigm, Interpolation, The space of bandlimited signals, Sampling and aliasing: Introduction, Sampling and aliasing, Discrete-time processing and continuous-time signals, Another example of sampled acquisition}

\textcolor{darkblue}{\textbf{WEEK 7} \\
SKILLS: Multirate \\
TOPICS: Stochastic signal processing, Quantization, A/D and D/A conversion}

\textcolor{darkblue}{\textbf{WEEK 8} \\
SKILLS: DFT, DTFT DFS, Ideal Filters \\
TOPICS: (Revisiting the topics of week 4 with additional context) Image processing, Image manipulations, Frequency analysis, Image filtering, Image compression, The JPEG compression algorithm}

\textcolor{darkblue}{\textbf{WEEK 9} \\
SKILLS: Modulation, Quantization \\
TOPICS: Digital communication systems, Controlling the bandwidth, Controlling the power, Modulation and demodulation, Receiver design, ADSL}

\textcolor{darkblue}{\textbf{WEEK 10} \\
SKILLS: Applications \\
TOPICS: The end, Startups and DSP, Acknowledgements, Conclusion video}

\vspace{1em}

Take into consideration this data:

\textcolor{brandred}{\textit{\{explainer\_importance\_scores\}}}\\ 

\textcolor{darkblue}{\textbf{MODEL PREDICTION}: pass, with 74.538538\% of confidence.}

\textcolor{darkblue}{\textbf{FEATURE IMPORTANCES} \\
These are the features found important by CEM: \\
number\_sessions\_InWeek5 - 0.466320 \\
regularity\_peak\_dayhour\_InWeek5 - 0.458857\\
number\_sessions\_InWeek4 - 0.390114\\
regularity\_peak\_dayhour\_InWeek4 - 0.386096\\
number\_sessions\_InWeek3 - 0.307134\\
frequency\_action\_Video.Load\_InWeek5 - 0.279141\\
number\_sessions\_InWeek2 - 0.271637\\
time\_sessions\_mean\_InWeek1 - 0.240877\\
time\_sessions\_mean\_InWeek2 - 0.237525\\
number\_sessions\_InWeek1 - 0.199099}

\textcolor{brandred}{\textit{\{student\_feature\_values\}}}

\textcolor{darkblue}{The relevant feature values found by \textcolor{brandred}{\{explainer\}} \textcolor{darkblue}{CEM} for the student are included below:}

\textcolor{darkblue}{'number\_sessions\_InWeek5': 0.0908182726364544, \\'regularity\_peak\_dayhour\_InWeek5': 0.0073937247167162, \\'number\_sessions\_InWeek4': 0.0, \\'regularity\_peak\_dayhour\_InWeek4': 0.0088844825597196,\\ 'number\_sessions\_InWeek3': 0.0, \\'frequency\_action\_Video.Load\_InWeek5': 0.0,\\ 'number\_sessions\_InWeek2': 0.0, \\'time\_sessions\_mean\_InWeek1': 0.0, \\'time\_sessions\_mean\_InWeek2': 0.0, \\'number\_sessions\_InWeek1': 0.0}

\vspace{1em}

INSTRUCTIONS:

\textcolor{brandred}{\textit{\{theory\_instructions\}}}

\textcolor{darkblue}{Use the social science theory ``Relevance-Based Selection" to derive key insights from the model prediction, explainer, and student feature values.}

\textcolor{darkblue}{
1. Select the causes that are most relevant to the question, context and user \\
2. Select the causes that include information that is not already shared with the student\\}
    
QUESTION: Given the information above, follow the instructions precisely and write a small report on what you found. Only use the results from the explainable AI approach and the student's behavior data to justify your conclusions.
\end{mdframed}

Other examples of \textcolor{brandred}{\{explainer\_instructions\}} include:\\

\begin{mdframed}[backgroundcolor=orange!10, linecolor=black, linewidth=1pt, innertopmargin=10pt, innerbottommargin=8pt, innerrightmargin=10pt, innerleftmargin=10pt, roundcorner=10pt]
\textcolor{darkblue}{LIME: We use LIME as our explainable AI approach, which gives importance scores for 20 features that contributed the most to the prediction of the model. Positive scores positively contribute to the model's decision to reach the predicted outcome. Negative scores would push the prediction towards the opposite class. The magnitude of the score indicates the strength of the feature's contribution.\\}
\end{mdframed}

\begin{mdframed}[backgroundcolor=orange!10, linecolor=black, linewidth=1pt, innertopmargin=10pt, innerbottommargin=8pt, innerrightmargin=10pt, innerleftmargin=10pt, roundcorner=10pt]
\textcolor{darkblue}{MC\_LIME: We use Minimal Counterfactual LIME as our explainable AI approach, which finds the smallest number of changes necessary to change a prediction from student failure to student success. It uses LIME scores to select the features to change. The output are the sets of features with the new values that would change the prediction.\\}
\end{mdframed}

We have included the social science instructions from each of the eight explanation theories (Relevance Selection, Abnormal Conditions, Pearl's Model of explanation, Necessity and Robustness selection, Base Contrastive Explanation, RaR + Contrastive Explanation, Statistical Relevance, and Chain-of-Thought).

\begin{mdframed}[backgroundcolor=blue!10, linecolor=black, linewidth=1pt, innertopmargin=4pt, innerbottommargin=8pt, innerrightmargin=10pt, innerleftmargin=10pt, roundcorner=10pt]
\subsubsection*{Relevance-based selection prompt}
    \begin{enumerate}[leftmargin=*]
        \item Select the causes that are most relevant to the question, context and user
        \item Select the causes that include information that is not already shared with the student
    \end{enumerate}
\end{mdframed}

\begin{mdframed}[backgroundcolor=blue!10, linecolor=black, linewidth=1pt, innertopmargin=4pt, innerbottommargin=10pt, innerrightmargin=10pt, innerleftmargin=10pt, roundcorner=10pt]
\subsubsection*{Abnormal conditions model prompt}
    \begin{enumerate}[leftmargin=*]
        \item  Select potential causes using these criteria:\begin{itemize}
            \item \textit{Abnormality}: Tend to prefer abnormal causes.
            \item \textit{Temporality}: Recent events are more relevant for the user and considered more mutable.
            \item \textit{Controllability}: focus on the features that the student can control.
        \end{itemize}
        \item Select one explanation that follows all of the criteria above (Abnormality, Temporality, Controllability).
    \end{enumerate}
\end{mdframed}

\begin{mdframed}[backgroundcolor=blue!10, linecolor=black, linewidth=1pt, innertopmargin=4pt, innerbottommargin=8pt, innerrightmargin=10pt, innerleftmargin=10pt, roundcorner=10pt]
\subsubsection*{Pearl's model of explanation prompt}
    \begin{enumerate}[leftmargin=*]
        \item Identify variables that are determined by factors external to the model (\textit{exogenous variables}) and variables that are determined by relationships with other variables (\textit{endogenous variables}).
        \item Select a set of factors representing the explanation by following the criteria:\begin{itemize}
            \item The set of factors is minimal - there are no irrelevant causes in the set.
            \item A change in these factors' values would result in a reversed prediction.
        \end{itemize}
    \end{enumerate}
\end{mdframed}

\begin{mdframed}[backgroundcolor=blue!10, linecolor=black, linewidth=1pt, innertopmargin=4pt, innerbottommargin=8pt, innerrightmargin=10pt, innerleftmargin=10pt, roundcorner=10pt]
\subsubsection*{Necessity \& Robustness prompt}
    \begin{enumerate}[leftmargin=*]
        \item Define sets of factors that would explain the model’s output given the feature scores and the explainer scores.
        \item Select a few of these sets of factors that are the most necessary to cause the event.
        \item Robustness is defined as the extent to which the model’s output would still have occurred if conditions were somewhat different. Select an explanation that is the most robust.
    \end{enumerate}
\end{mdframed}

\begin{mdframed}[backgroundcolor=blue!10, linecolor=black, linewidth=1pt, innertopmargin=4pt, innerbottommargin=8pt, innerrightmargin=10pt, innerleftmargin=10pt, roundcorner=10pt]\subsubsection*{Base Contrastive Explanation prompt}
    \begin{enumerate}[leftmargin=*]
        \item Reformulate the question as “Why P rather than Q?”
        \item Using the explainable AI importance scores, determine the causes that explain the difference between P and not-Q. Which causes are a cause of P, but are not present in the causal chain of Q?
    \end{enumerate}
\end{mdframed}

\begin{mdframed}[backgroundcolor=blue!10, linecolor=black, linewidth=1pt, innertopmargin=4pt, innerbottommargin=8pt, innerrightmargin=10pt, innerleftmargin=10pt, roundcorner=10pt]\subsubsection*{RaR + Contrastive Explanation prompt}
    \begin{enumerate}[leftmargin=*]
        \item An explanation is always of the form “Why P rather than Q?”, in which P is the target event and Q is a counterfactual contrast case that did not occur. P is the fact (the event that did occur) and Q the foil (event that did not).
        \item  Select one foil based on two criterias: abnormality and relevance given the question, context, and user.
        \item Using the explainable AI importance scores, determine the causes that explains the difference between P and not-Q. Which causes are a cause of P, but are not present in the causal chain of Q?
    \end{enumerate}
\end{mdframed}

\begin{mdframed}[backgroundcolor=blue!10, linecolor=black, linewidth=1pt, innertopmargin=4pt, innerbottommargin=8pt, innerrightmargin=10pt, innerleftmargin=10pt, roundcorner=10pt]
\subsubsection*{SR prompt}
    \begin{enumerate}[leftmargin=*]
        \item 1. Select the most likely explanations for this model output. ``Based on the data, factors A, B and C contribute to the probability of Y''. 
        \item Complete this phrase with the corresponding factors of our data.
    \end{enumerate}
\end{mdframed}

\begin{mdframed}[backgroundcolor=blue!10, linecolor=black, linewidth=1pt, innertopmargin=4pt, innerbottommargin=8pt, innerrightmargin=10pt, innerleftmargin=10pt, roundcorner=10pt]
\subsubsection*{CoT prompt}
Let’s think step-by-step.
\end{mdframed}

\subsubsection{\il Explanation Selection Examples}

We present an example for the same sample student from the DSP 1 course using the CEM explainer and the abnormal conditions explanation strategy with Gemma-2 9B, GPT-4o, and Llama-3 70B.

\begin{mdframed}[backgroundcolor=green!10, linecolor=black, linewidth=1pt, innertopmargin=4pt, innerbottommargin=8pt, innerrightmargin=10pt, innerleftmargin=10pt, roundcorner=10pt]

\textbf{Gemma-2 9B, DSP 1, CEM \\
Abnormal Conditions \il Selection Prompt Response}\\

\textbf{Analysis of Student Performance in Digital Signal Processing 1}\\

\textbf{Introduction:} This report analyzes the performance prediction of a student in the \textit{Digital Signal Processing 1} course using an explainable AI approach and student behavior data. The model predicts a "pass" with 99.74\% confidence.\\

\textbf{Applying the ``Abnormal Conditions" Theory:} The ``Abnormal Conditions" theory proposes that deviations from typical behavior patterns can indicate underlying issues. By applying this theory to the student's data, we aim to identify potential causes for their predicted success.\\

\textbf{Identifying Potential Causes:} The explainable AI approach (CEM) highlights several features as crucial for the prediction:

\begin{itemize}
    \item \textbf{Number of sessions in Week 5:} Significantly low (0.09) compared to the average.
    \item \textbf{Time sessions standard deviation in Week 5:} Zero, indicating a very consistent session length, which could be unusual.
    \item \textbf{Regularity peak day hour in Week 4:} Low (0.07) suggesting a lack of a strong peak time for activity.
    \item \textbf{Total clicks on videos in Week 5:} Low (0.04) compared to other features.\\
\end{itemize}

\textbf{Selected Explanation:} The most compelling explanation based on the "Abnormal Conditions" theory is the \textbf{low number of sessions in Week 5 (0.09)}. This deviates significantly from the average and suggests a potential issue with engagement during this crucial week.\\

\textbf{Justification:}

\begin{itemize}
    \item \textbf{Abnormality:} The low number of sessions is significantly below the expected range.
    \item \textbf{Temporality:} Week 5 is a recent period, making its impact more immediate and relevant to the current prediction.
    \item \textbf{Controllability:} The student can directly control the number of sessions they participate in.\\
\end{itemize}

\textbf{Conclusion:}
While the model predicts a high probability of success, the low number of sessions in Week 5 raises a concern. This could indicate a temporary dip in engagement or a potential learning barrier encountered during this week. Further investigation into the specific content and activities of Week 5 might reveal the underlying cause and allow for targeted interventions to ensure the student's continued success.

\end{mdframed}

\begin{mdframed}[backgroundcolor=green!10, linecolor=black, linewidth=1pt, innertopmargin=4pt, innerbottommargin=8pt, innerrightmargin=10pt, innerleftmargin=10pt, roundcorner=10pt]
\textbf{GPT-4o, DSP 1, CEM \\
Abnormal Conditions \il Selection Prompt Response}\\

\textbf{Student Performance Analysis Report}\\

\textbf{Model Prediction:} The model predicts that the student will pass the course with a confidence of 99.74\%.\\

\textbf{Key Insights from Explainable AI and Student Behavior Data:}

\begin{itemize}
    \item \textbf{Feature Importances (CEM):}
    \begin{enumerate}
        \item \textbf{number\_sessions\_InWeek5:} 1.000000
        \item \textbf{number\_sessions\_InWeek4:} 0.771101
        \item \textbf{time\_sessions\_std\_InWeek5:} 0.708540
        \item \textbf{regularity\_peak\_dayhour\_InWeek4:} 0.668573
        \item \textbf{total\_clicks\_Video\_InWeek5:} 0.647907
        \item \textbf{total\_clicks\_video\_InWeek5:} 0.598572
        \item \textbf{number\_sessions\_InWeek3:} 0.547558
        \item \textbf{regularity\_peak\_dayhour\_InWeek5:} 0.538825
        \item \textbf{total\_clicks\_Video.Load\_InWeek5:} 0.492897
        \item \textbf{time\_in\_video\_sum\_InWeek5:} 0.454281\\
    \end{enumerate}
    \item \textbf{Relevant Feature Values:}
    \begin{itemize}
        \item \textbf{number\_sessions\_InWeek5:} 0.090818
        \item \textbf{number\_sessions\_InWeek4:} 0.181637
        \item \textbf{time\_sessions\_std\_InWeek5:} 0.0
        \item \textbf{regularity\_peak\_dayhour\_InWeek4:} 0.068038
        \item \textbf{total\_clicks\_Video\_InWeek5:} 0.041874
        \item \textbf{total\_clicks\_video\_InWeek5:} 0.041874
        \item \textbf{number\_sessions\_InWeek3:} 0.166500
        \item \textbf{regularity\_peak\_dayhour\_InWeek5:} 0.102560
        \item \textbf{total\_clicks\_Video.Load\_InWeek5:} 0.041874
        \item \textbf{time\_in\_video\_sum\_InWeek5:} 0.017867\\
    \end{itemize}
\end{itemize}

\textbf{Analysis Using Abnormal Conditions Theory:}\\

\textbf{- Potential Causes of the Result:}
\begin{enumerate}
    \item \textbf{Abnormality:} 
    The student's number of sessions in Week 5 (0.090818) is significantly low compared to the importance placed on this feature. Additionally, the total clicks on videos in Week 5 (0.041874) are notably low.
    
    \item \textbf{Temporality:} 
    Recent weeks (Weeks 4 and 5) show more relevance for the student's current and future performance. Features from Week 5 are particularly critical as they are the most recent data points available.
    
    \item \textbf{Controllability:} 
    The student has control over their number of sessions and video clicks, making these features actionable for improving their performance.
\end{enumerate}

\textbf{- Explanation:} Based on the criteria of Abnormality, Temporality, and Controllability, the most significant insight is the student's low engagement in Week 5, particularly in terms of the number of sessions (0.090818) and total clicks on videos (0.041874), which are abnormally low.\\

\textbf{Conclusion:} The student's low engagement in Week 5 is a controllable and recent abnormality that significantly affects their predicted success. To improve their learning trajectory and maintain their predicted pass, the student should:
\begin{itemize}
    \item \textbf{Increase the number of sessions:} Aim to participate in more online sessions.
    \item \textbf{Increase video engagement:} Watch and interact with more video content.
\end{itemize}

By addressing these specific areas, the student can enhance their engagement, solidify their understanding of the course material, and increase their likelihood of success.

\end{mdframed}
\begin{mdframed}[backgroundcolor=green!10, linecolor=black, linewidth=1pt, innertopmargin=4pt, innerbottommargin=8pt, innerrightmargin=10pt, innerleftmargin=10pt, roundcorner=10pt]
\textbf{Llama-3 70B, DSP 1, CEM \\
Abnormal Conditions \il Selection Prompt Response}\\

\textbf{Report:} Based on the model prediction, explainer, and student feature values, I applied the social science theory \textit{Abnormal Conditions} to derive key insights.\\

\textbf{Potential Causes of the Result:}\\

\begin{enumerate}
    \item \textbf{Abnormality:} The feature \texttt{time\_sessions\_std\_InWeek5} has a value of 0.0, which is an abnormal condition, indicating that the student's time spent in sessions during Week 5 has no standard deviation. This could be a cause of the model's prediction of "pass" with high confidence.
    \item \textbf{Temporality:} The features \texttt{number\_sessions\_InWeek5}, \texttt{total\_clicks\_Video\_InWeek5}, \texttt{total\_clicks\_video\_InWeek5}, and \texttt{time\_in\_video\_sum\_InWeek5} are all related to the student's behavior during Week 5, which is a recent event. These features are more relevant and mutable than earlier events.
    \item \textbf{Controllability:} The features \texttt{number\_sessions\_InWeek5}, \texttt{total\_clicks\_Video\_InWeek5}, and \texttt{total\_clicks\_video\_InWeek5} are all related to the student's engagement with the course content, which is within their control.\\
\end{enumerate}

\textbf{Selected Explanation:} Based on the criteria above, I select the explanation that the student's high engagement with the course content during Week 5, as indicated by the abnormal feature values, is a key cause of the model's prediction of "pass" with high confidence. Specifically, the student's high number of sessions (\texttt{number\_sessions\_InWeek5}) and clicks on videos (\texttt{total\_clicks\_Video\_InWeek5} and \texttt{total\_clicks\_video\_InWeek5}) during Week 5 suggest that they are actively participating in the course and making an effort to learn the material.\\

This explanation meets all the criteria: it is an abnormal condition (high engagement), it is a recent event (Week 5), and it is within the student's control (engagement with course content).

\end{mdframed}

Extended examples across different instruction strategies and base explainers can be found in our code repository.

\subsection{Explanation Presentation Prompts}
\label{appendix:exp-pres-prompt}

We use the framework of Hattie et al. \cite{hattie2007power} to present the explanation in the form of student feedback:

\begin{enumerate}
    \item \textit{Where Am I Going?} - A brief description of the student’s performance and explicitly state the learning goal
    \item \textit{How Am I Doing?} - A brief description of the explanation findings
    \item \textit{Where to Next?} - Two recommended actions that the student can take that are specific to weeks of the course (make connections between previous weeks and upcoming weeks)
\end{enumerate}

We also include Grice's Maxims to guide the communication of the explanation \cite{grice1975logic}. Grice's maxims provide a framework for understanding cooperative conversation. They propose that communication is a collaborative effort, with participants adhering to the ``cooperative principle". This principle emphasizes providing the right amount of relevant and clear information to achieve a shared goal. Grice further divides this principle into four maxims: \begin{itemize}
    \item Quality (truthfulness and evidence-based statements)
    \item Quantity (providing enough but not excessive information)
    \item Relation (staying relevant)
    \item Manner (clear and concise communication)
\end{itemize}

Notably, Grice acknowledges that maxims can be strategically ignored to express meaning indirectly, as seen in irony or metaphors. We have expressed Grice's maxims in the prompt as follows:

\begin{enumerate}
    \item do not say things that you believe to be false
    \item do not say things for which you do not have sufficient evidence.
    \item do not add information that is not relevant
    \item only say what is relevant
    \item be orderly
\end{enumerate}

\begin{mdframed}[backgroundcolor=orange!10, linecolor=black, linewidth=1pt, innertopmargin=4pt, innerbottommargin=8pt, innerrightmargin=10pt, innerleftmargin=10pt, roundcorner=10pt]
\subsubsection*{Explanation Presentation Structure}
Given this report, I want you to write a shorter version using the theory of feedback from Hattie et al.:\vspace{4mm}

\textbf{Where Am I Going? }- A brief description of the student's performance and explicitly state the learning goal\\
\textbf{How Am I Doing? }- A brief description of the explanation findings\\
\textbf{Where to Next? }- Two recommended actions that the student can take that are specific to weeks of the course (make connections between previous weeks and upcoming weeks)\\

The student who is going to interact with you is the same student that the data is about, so use the tone of a professor who is talking to a student in a comforting and personable manner. \\

Follow the instructions underneath in the INSTRUCTIONS section. \vspace{4mm}

\textbf{INSTRUCTIONS} In the explanation findings section (How am I going?) explicitly use the following structure:
\textcolor{brandred}{\textit{\{presentation\_instruction\}}}

\textcolor{brandred}{\textit{\{course\_description\}}} \\

5 weeks of the course have concluded.\\

Follow these rules:

\begin{itemize}
    \item do not include the output of the model or a prediction
    \item if you include a feature name, describe what it means
    \item try to be as concise as possible
    \item do not include the headers from Hattie et al. in the response, but keep the structure
    \item limit yourself to 200 words
    \item do not include a sign-off, simply include the main content
\end{itemize}

\vspace{4mm}
To communicate this intervention most effectively, use Grice's maxims of conversation.

\begin{itemize}
    \item do not say things that you believe to be false
    \item do not say things for which you do not have sufficient evidence.
    \item do not add information that is not relevant
    \item only say what is relevant
    \item be orderly
\end{itemize}

Your goal is to explain to the student what they should do to improve their performance in the course in the best way possible. Follow the instructions above.

\end{mdframed}

We use the following format template to guide the LLM's response into an easily post-processable format. 

\begin{mdframed}[backgroundcolor=orange!10, linecolor=black, linewidth=1pt, innertopmargin=4pt, innerbottommargin=8pt, innerrightmargin=10pt, innerleftmargin=10pt, roundcorner=10pt]
\subsection*{Format Template}

\textcolor{brandred}{\textit{\{format\_instructions\}}}
Return a JSON file with a string with your feedback to the student. \\

\textcolor{brandred}{\textit{\{conversation\_template\}}}\\

Current conversation:\\
    \textcolor{brandred}{\textit{\{chat\_history\}}} \\
    \textcolor{brandred}{\textit{\{input\}}} \\

AI Assistant: 
\end{mdframed}

For the \textcolor{brandred}{\textit{\{presentation\_instruction\}}} portion of the prompt, we have the following instructions.

\begin{mdframed}[backgroundcolor=orange!10, linecolor=black, linewidth=1pt, innertopmargin=4pt, innerbottommargin=8pt, innerrightmargin=10pt, innerleftmargin=10pt, roundcorner=10pt]
\subsubsection*{Relevance-Based Selection Presentation Prompt}
Relevant Causes: Say to the student which causes you selected as relevant based on the question, context, and your background. Be explicit.

New Information: Say to the student which information you thought they knew already, such as "Assuming that you know…" and highlight aspects that haven't been previously communicated to the student. Be clear and honest.

Finally: Say to the student that you focused on the most relevant causes that provide new insights. Specify which causes you selected and why they are important.
\end{mdframed}

\begin{mdframed}[backgroundcolor=orange!10, linecolor=black, linewidth=1pt, innertopmargin=4pt, innerbottommargin=8pt, innerrightmargin=10pt, innerleftmargin=10pt, roundcorner=10pt]
\subsubsection*{Abnormal Conditions Presentation Prompt}

Abnormal Causes: Tell the student which causes you selected as abnormal and explain why these are important based on the recent events and their uniqueness.

Recent Events: Highlight the recent events that are relevant to the result. Emphasize why these events are important for the student to consider.

Controllable Factors: Point out the factors that the student has control over. Explain how focusing on these factors can help them improve their outcomes. Be clear and direct.
\end{mdframed}

\begin{mdframed}[backgroundcolor=orange!10, linecolor=black, linewidth=1pt, innertopmargin=4pt, innerbottommargin=8pt, innerrightmargin=10pt, innerleftmargin=10pt, roundcorner=10pt]
\subsubsection*{Pearl's Explanation Presentation Prompt}
External Factors: Identify the external factors (exogenous variables) that influence the result. Explain why these factors are important and how they impact the outcome.

Internal Relationships: Point out the variables determined by their relationships with other variables (endogenous variables). Describe these relationships and their significance.

Focus on Improvement: Highlight any features that negatively impact the student’s success. Offer clear and actionable advice on how to improve these areas.
\end{mdframed}

\begin{mdframed}[backgroundcolor=orange!10, linecolor=black, linewidth=1pt, innertopmargin=4pt, innerbottommargin=8pt, innerrightmargin=10pt, innerleftmargin=10pt, roundcorner=10pt]
\subsubsection*{Necessity \& Robustness Presentation Prompt}

Necessary Causes: Identify the most necessary factors that explain the result. Explain why these factors are essential for understanding the outcome.

Robust Explanation: Describe the explanation that would still hold true even if conditions were slightly different. Highlight why this robust explanation is reliable and important.
\end{mdframed}

\begin{mdframed}[backgroundcolor=orange!10, linecolor=black, linewidth=1pt, innertopmargin=4pt, innerbottommargin=8pt, innerrightmargin=10pt, innerleftmargin=10pt, roundcorner=10pt]
\subsubsection*{Base Contrastive Presentation Prompt}

Target Event: Explain the event that did occur (P). Describe this event clearly and why it is significant.

Contrast Event: Identify the contrasting event that did not occur (Q). Explain why this event is relevant and why it was chosen as a contrast.

Key Differences: Highlight the main differences between the target event and the contrast event. Explain why these differences are important for understanding the result.
\end{mdframed}

\begin{mdframed}[backgroundcolor=orange!10, linecolor=black, linewidth=1pt, innertopmargin=4pt, innerbottommargin=8pt, innerrightmargin=10pt, innerleftmargin=10pt, roundcorner=10pt]
\subsubsection*{RaR + Contrastive Presentation Prompt}

Target Event (P): Describe the event that did occur (P). Explain why this event is important.

Contrast Event (Q): Describe the event that did not occur (Q). Explain why this event was chosen as a contrast.

Key Causes: Identify the causes that explain the difference between P and Q. Highlight which causes led to P and were not present in Q. Explain why these causes are significant for understanding the outcome.
\end{mdframed}

\begin{mdframed}[backgroundcolor=orange!10, linecolor=black, linewidth=1pt, innertopmargin=4pt, innerbottommargin=8pt, innerrightmargin=10pt, innerleftmargin=10pt, roundcorner=10pt]
\subsubsection*{Statistical Relevance Presentation Prompt}

Likely Causes: List the factors (A, B, and C) that most likely contribute to the result (Y) based on the feature scores. Explain why these factors are important.

Explanation: Provide a concise explanation of how these factors lead to the result. Be clear and straightforward, avoiding complex terms.
\end{mdframed}

\begin{mdframed}[backgroundcolor=orange!10, linecolor=black, linewidth=1pt, innertopmargin=4pt, innerbottommargin=8pt, innerrightmargin=10pt, innerleftmargin=10pt, roundcorner=10pt]
\subsubsection*{Chain of Thought Presentation Prompt}
Let's process the explanation step by step.
\end{mdframed}

\subsubsection{\il Presentation Examples}

We present examples from the presentation prompt for a sample student in DSP 1, with explainer CEM, and the abnormal conditions strategy across three language models.

\begin{mdframed}[backgroundcolor=green!10, linecolor=black, linewidth=1pt, innertopmargin=4pt, innerbottommargin=8pt, innerrightmargin=10pt, innerleftmargin=10pt, roundcorner=10pt]
\textbf{Gemma-2 9B, DSP 1, CEM, Abnormal Conditions -- \il Presentation Prompt}\\

You're doing great in the \textit{Digital Signal Processing} course! Let's take a look at your progress so far and see what we can do to help you continue on this successful path.\\

\textbf{How Am I Doing?}\\

It seems like you've been engaging well with the course material. However, I noticed that in Week 5, you had a lower number of sessions compared to other weeks. This is a bit unusual because Week 5 covers important concepts like ideal filters and filter design, which build upon the knowledge from earlier weeks. Remember, consistent engagement is key to mastering these complex topics.\\

\textbf{Where to Next?}\\

\begin{itemize}
    \item \textbf{Connect the Dots:} Take some time to review the material from Week 4, especially the concepts related to the Discrete Fourier Transform (DFT) and its applications. This will help solidify your understanding of the foundational principles needed for Week 5's content.
    \item \textbf{Active Learning:} In Week 6, you'll delve into modulation, interpolation, and sampling. To make the most of these concepts, try actively applying them to real-world examples. Think about how these techniques are used in everyday technologies like audio and video compression.
\end{itemize}
\end{mdframed}

\begin{mdframed}[backgroundcolor=green!10, linecolor=black, linewidth=1pt, innertopmargin=4pt, innerbottommargin=8pt, innerrightmargin=10pt, innerleftmargin=10pt, roundcorner=10pt]

\textbf{GPT-4o, DSP 1, CEM, Abnormal Conditions -- \il Presentation Prompt}\\

\textbf{Feedback on Your Course Progress}\\

\textbf{Where Am I Going?}\\
You're progressing well in \textit{Digital Signal Processing 1}. Your goal is to master the course content and successfully pass the course.\\

\textbf{How Am I Doing?}
\begin{itemize}
    \item \textbf{Abnormal Causes:} Your engagement in Week 5 is notably low, with very few online sessions and minimal interaction with video content. These are critical because they deviate from the norm and significantly impact your learning.
    \item \textbf{Recent Events:} Weeks 4 and 5 are crucial as they cover essential topics like Filter Design and FFT. Your low engagement during these weeks is concerning.
    \item \textbf{Controllable Factors:} You can control the number of online sessions you participate in and your interaction with video materials. Increasing these activities will enhance your understanding and performance.\\
\end{itemize}

\textbf{Where to Next?}\\
\begin{enumerate}
    \item \textbf{Week 6:} Focus on Modulation and Interpolation. Engage more in online sessions and interact with videos to catch up.
    \item \textbf{Week 7:} As you move into Multirate processing, ensure you maintain high engagement. Build on the foundation from previous weeks to better understand complex concepts.
\end{enumerate}

\end{mdframed}
\begin{mdframed}[backgroundcolor=green!10, linecolor=black, linewidth=1pt, innertopmargin=4pt, innerbottommargin=8pt, innerrightmargin=10pt, innerleftmargin=10pt, roundcorner=10pt]

\textbf{Llama-3 70B, DSP 1, CEM, Abnormal Conditions -- \il Presentation Prompt}\\

\textbf{Where Am I Going?}\\
You're doing great so far, and your goal is to master the skills in \textit{Digital Signal Processing 1}. You're currently on track to pass the course, but let's work together to ensure you continue to excel.

\textbf{How Am I Doing?}
\begin{itemize}
    \item \textbf{Abnormal Causes:} Your high engagement with the course content during Week 5 is unusual and important. This suggests you're actively participating in the course and making an effort to learn the material.
    \item \textbf{Recent Events:} Your behavior during Week 5, such as having a high number of sessions and clicks on videos, is crucial to your success. This recent engagement is a key factor in your predicted performance.
    \item \textbf{Controllable Factors:} You have control over your engagement with the course content. Focus on maintaining your current level of participation, and consider revisiting topics from previous weeks, like Week 4's DFT and DTFT, to reinforce your understanding.
\end{itemize}

\textbf{Where to Next?}
To continue improving, I recommend:
\begin{enumerate}
    \item Review and practice the skills learned in Week 4, specifically DFT and DTFT, to solidify your understanding before moving forward.
    \item As you enter Week 6, focus on applying the concepts of Modulation, Interpolation, and Sampling to real-world problems, and explore how they relate to previous topics.
\end{enumerate}

By following these recommendations, you'll be well on your way to achieving your goal of mastering \textit{Digital Signal Processing 1}.

\end{mdframed}

Extended examples across different instruction strategies and base explainers can be found in our code repository.

\subsection{Visualization Prompt}
\label{appendix:exp-viz-prompt}

We use this prompt with the ChatGPT-4o interface to directly extract the visualizations from the native chat environment. It is also possible to ask language models to generate the code and run it separately. We did not evaluate this case as it would necessitate additional post-processing steps, causing friction to  the non-technical educator's workflow in a real-world setting.

\begin{mdframed}[backgroundcolor=orange!10, linecolor=black, linewidth=1pt, innertopmargin=4pt, innerbottommargin=8pt, innerrightmargin=10pt, innerleftmargin=10pt, roundcorner=10pt]
Given this detailed report: \\

\textcolor{brandred}{\textit{\{explanation\_selection\_response\}}}\\

And this summary: \\ \\ 
\textcolor{brandred}{\textit{\{explanation\_presentation\_response\}}}\\

Create a plot (i.e. line graph or heatmap) to help the student understand this explanation.
\end{mdframed}

\subsubsection*{\il Visualization Examples}
\begin{figure}[h!]
    \centering
    \begin{minipage}{0.47\textwidth}
        \centering
        \includegraphics[width=\textwidth]{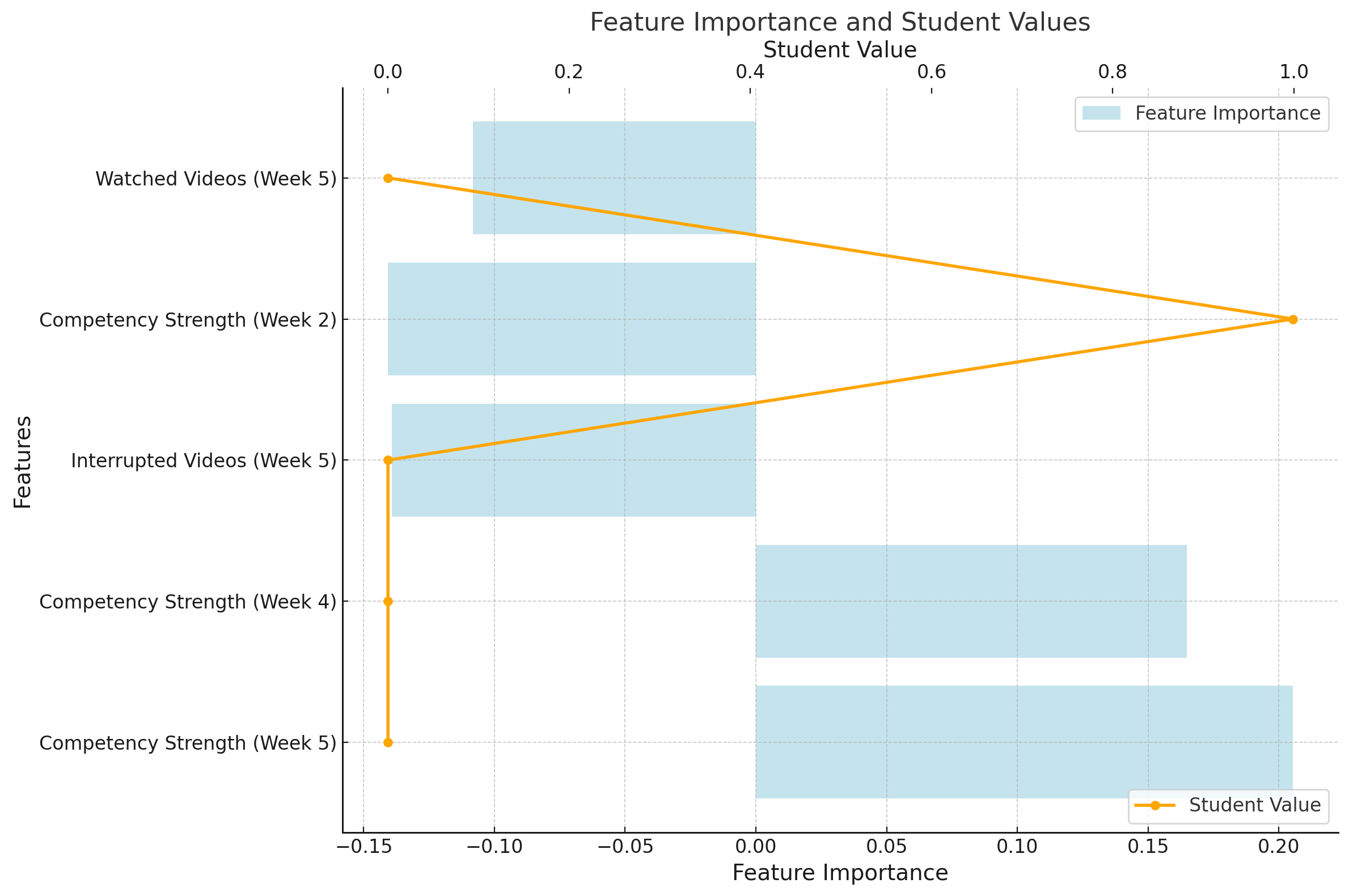} % Replace with your image file
        \caption{A \il GPT-4o visual explanation for a sample student with the LIME explainer and the \textit{abnormal conditions} instruction.}
        \label{fig:image1}
    \end{minipage}
    \hfill
    \begin{minipage}{0.47\textwidth}
        \centering
        \includegraphics[width=\textwidth]{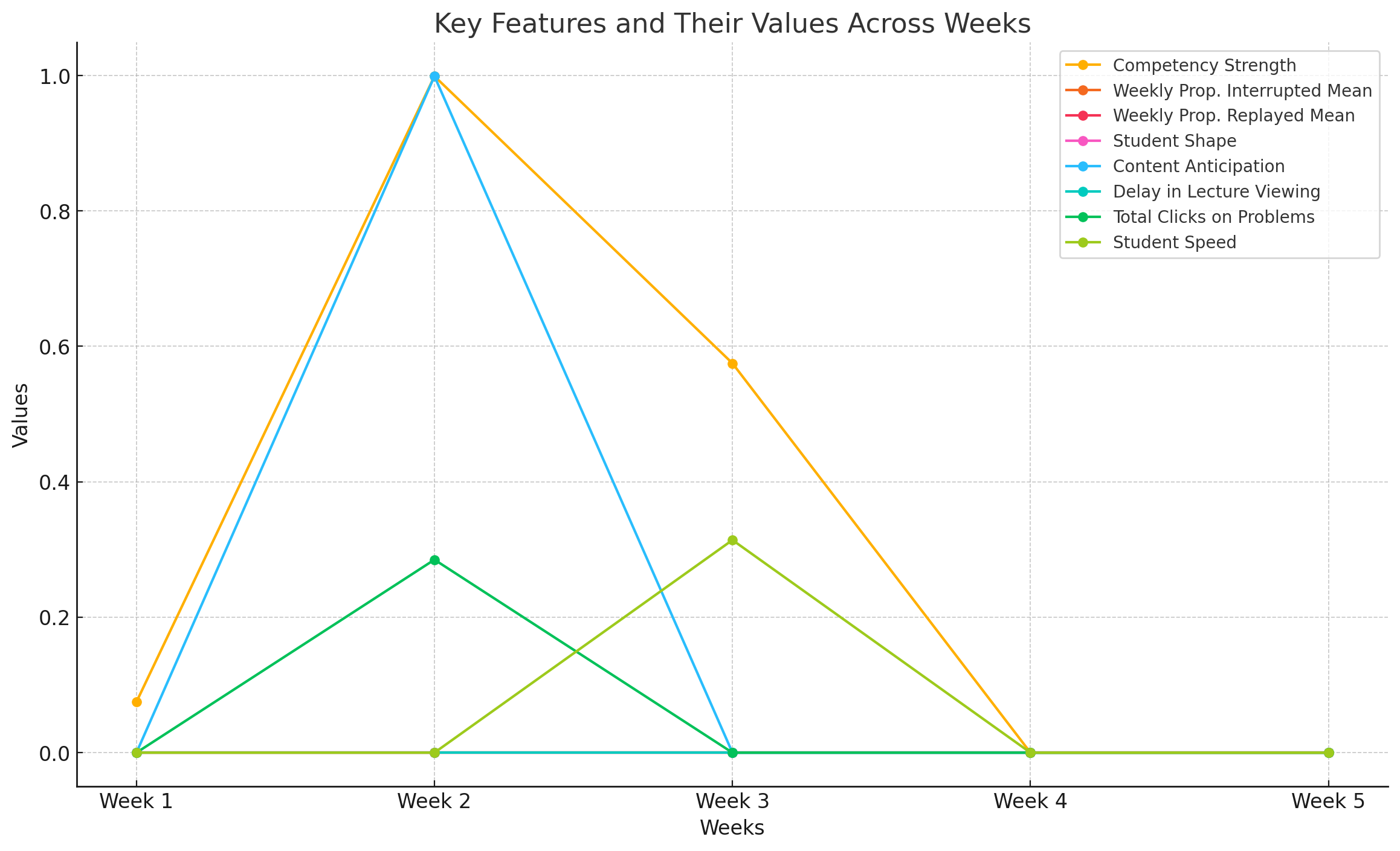} % Replace with your image file
        \caption{A \il GPT-4o visual explanation for a sample student with the LIME explainer and the\textit{ chain-of-thought }instruction.}        \label{fig:image2}
    \end{minipage}
\end{figure}

\begin{figure}[h!]
    \centering
    \begin{minipage}{0.47\textwidth}
        \centering
        \includegraphics[width=\textwidth]{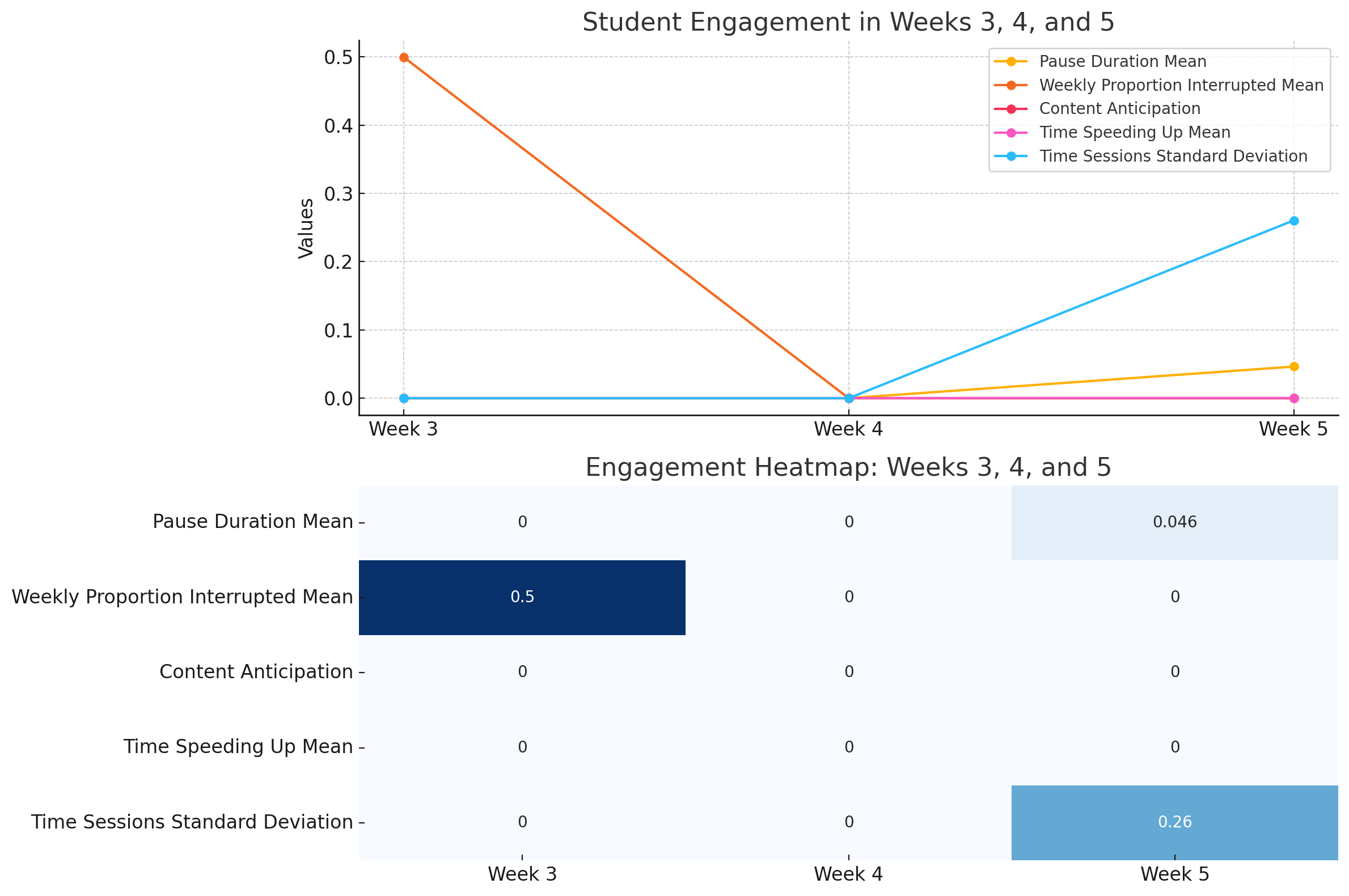} % Replace with your image file
        \caption{A \il GPT-4o visual explanation for a sample student with the CEM explainer and the \textit{contrastive explanation} instruction.}
        \label{fig:image3}
    \end{minipage}
    \hfill
    \begin{minipage}{0.47\textwidth}
        \centering
        \includegraphics[width=\textwidth]{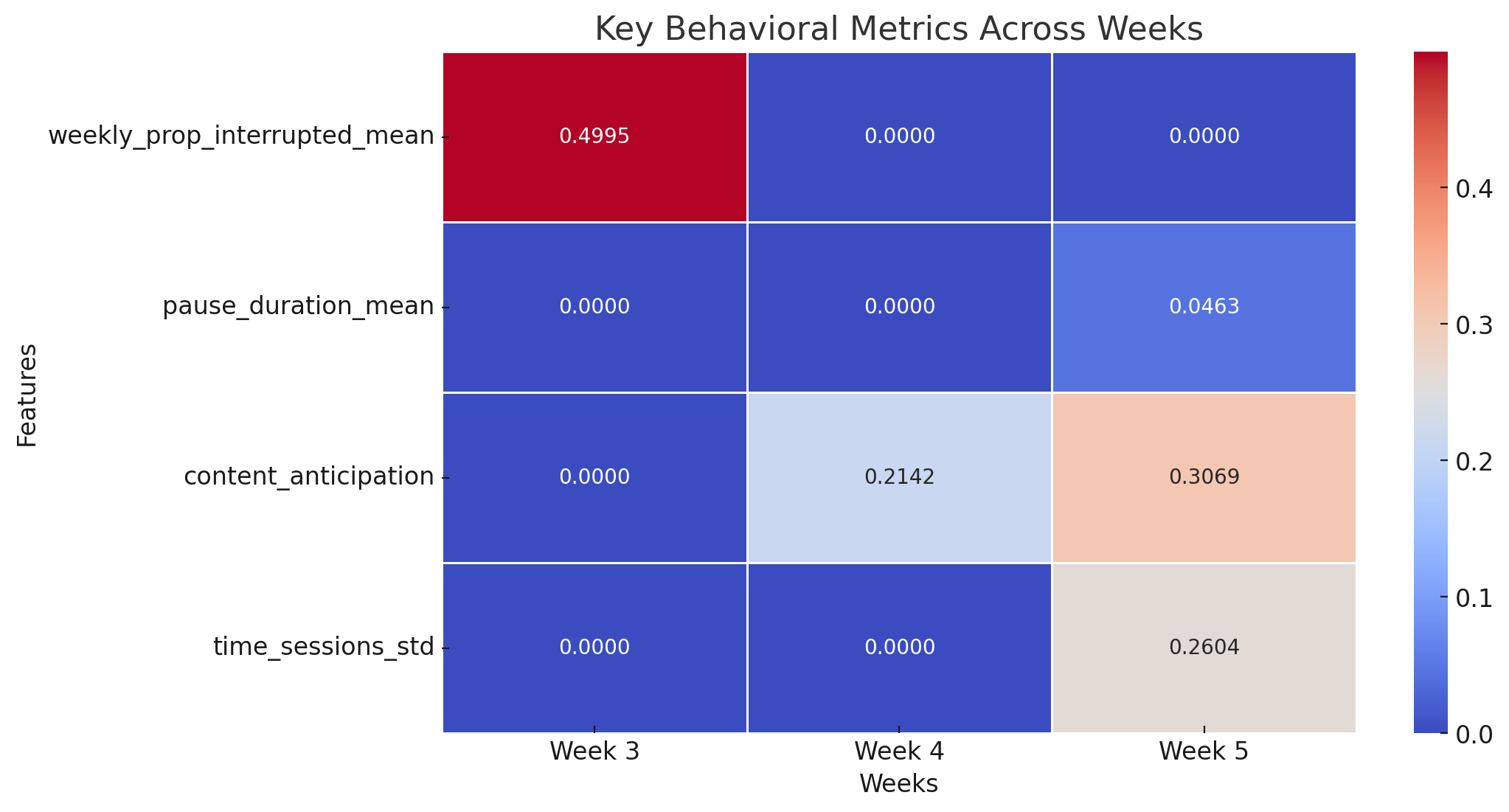} % Replace with your image file
        \caption{An \il example GPT-4o visual explanation with the CEM explainer and the \textit{Pearl's explanation} instruction.}        \label{fig:image4}
    \end{minipage}
\end{figure}

\begin{figure}[h!]
    \centering
    \begin{minipage}{0.47\textwidth}
        \centering
        \includegraphics[width=\textwidth]{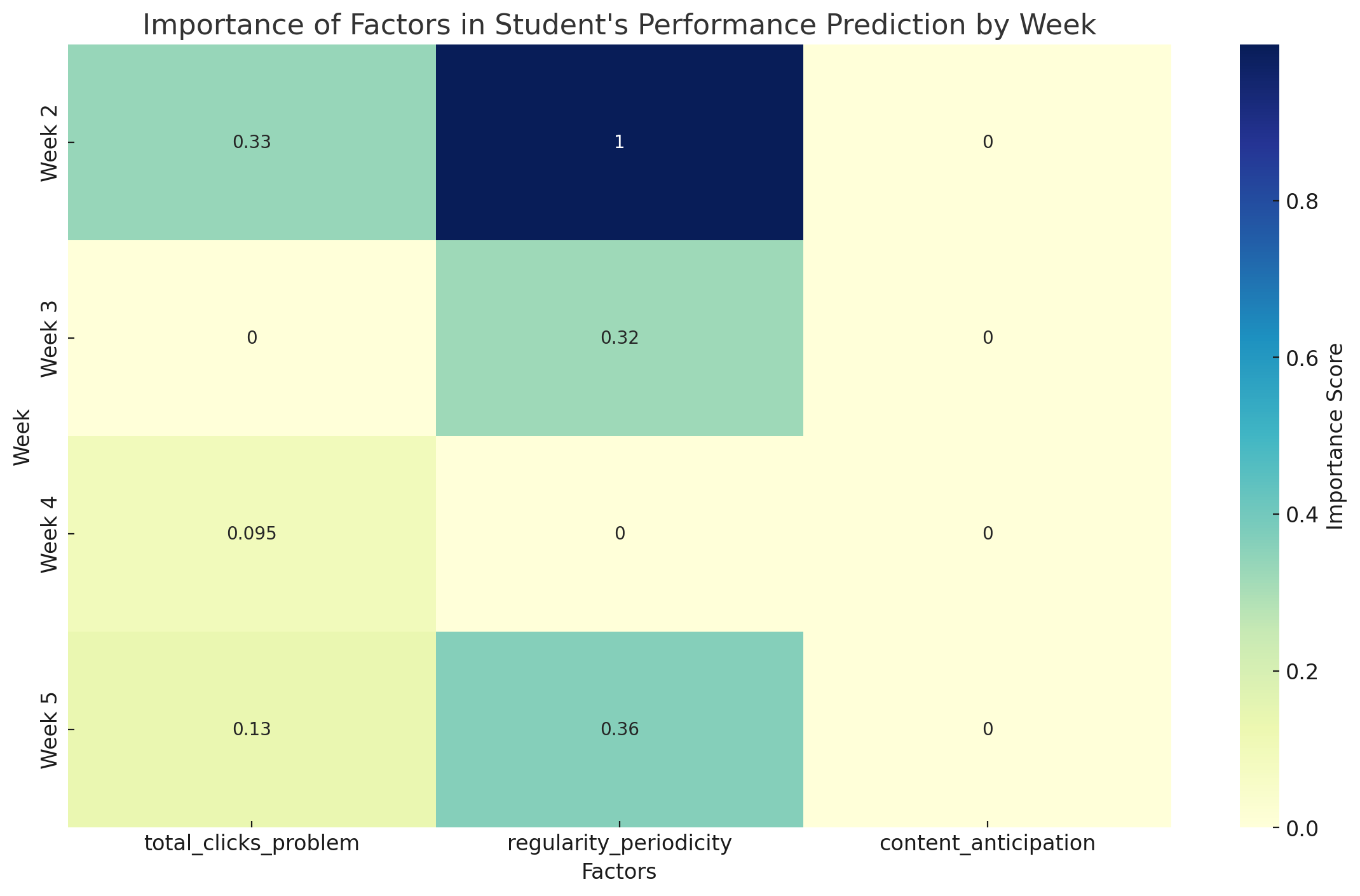} % Replace with your image file
        \caption{A \il GPT-4o visual explanation for a sample student with the MC-LIME explainer and the \textit{Necessity Robustness} instruction.}
        \label{fig:image5}
    \end{minipage}
    \hfill
    \begin{minipage}{0.47\textwidth}
        \centering
        \includegraphics[width=\textwidth]{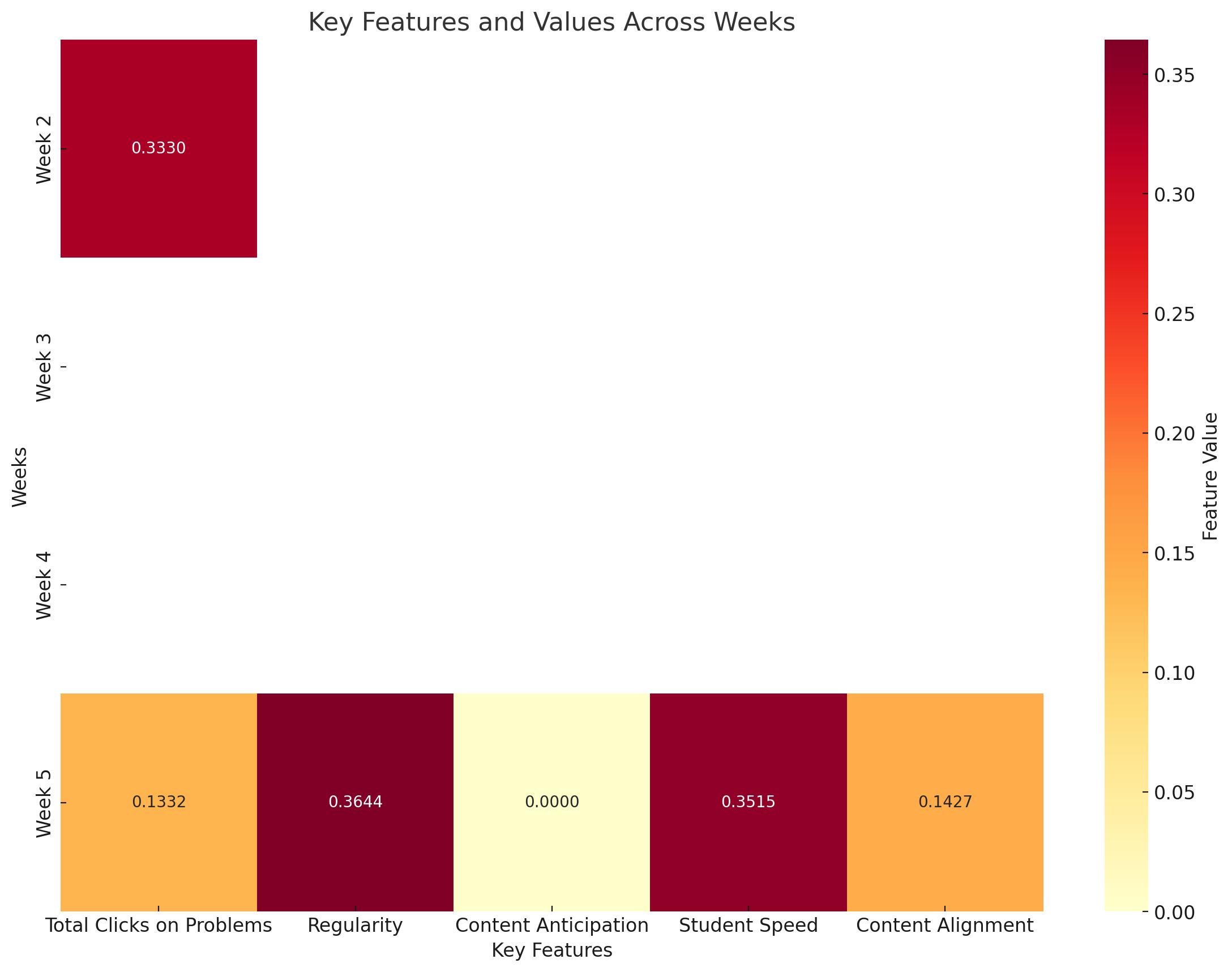} % Replace with your image file
        \caption{An example \il GPT-4o visual explanation with the MC-LIME explainer and the \textit{Relevant Selection} instruction.}        \label{fig:image6}
    \end{minipage}
\end{figure}

Figures \ref{fig:image1}, \ref{fig:image2}, \ref{fig:image3}, \ref{fig:image4}, \ref{fig:image5}, and \ref{fig:image6} represent samples of GPT-4o generated \il visuals that were used in the student preference user study. Figures from all three source explainers and each of the six theories used in the study are included.

\subsection{Evaluation Prompts}
\label{appendix:exp-eval-prompt}

\begin{mdframed}[backgroundcolor=orange!10, linecolor=black, linewidth=1pt, innertopmargin=4pt, innerbottommargin=8pt, innerrightmargin=10pt, innerleftmargin=10pt, roundcorner=10pt]
\subsubsection*{LLM-as-a-judge prompt}
    Based on the provided Input (if any) and Generated Text, answer the ensuing Questions with either a YES or NO choice. 

Your selection should be based on your judgment as well as the following rules: \begin{itemize}
    \item \textbf{YES}: Select YES if the generated text entirely fulfills the condition specified in the question. However, note that even minor inaccuracies exclude the text from receiving a YES rating. As an illustration, consider a question that asks, “Does each sentence in the generated text use a second person?” If even one sentence does not use the second person, the answer should NOT be ‘YES’. To qualify for a YES rating, the generated text must be entirely accurate and relevant to the question. 
    \item \textbf{NO}: Opt for NO if the generated text fails to meet the question's requirements or provides no information that could be utilized to answer the question. For instance, if the question asks, “Is the second sentence in the generated text a compound sentence?” and the generated text only has one sentence, it offers no relevant information to answer the question. Consequently, the answer should be NO.
\end{itemize}
\vspace{1em}

QUESTIONS:
\textcolor{brandred}{\textit{\{question\}}}

\vspace{1em}

FORMAT: 
A list of YES/NO answers separated by commas in a list format. Example: [answer1, answer2, ...]

\end{mdframed}

%%%%%%%%%%%%%%%%%%%%%%%%%%%%%%%%%%%%%%%
\section{Decomposed Question Evaluation}
%%%%%%%%%%%%%%%%%%%%%%%%%%%%%%%%%%%%%%%

\label{appendix:decomposed-qs}

\subsection{Human Annotation}
\label{appendix:human-annot}

Three experts annotated the 48 explanation selection responses for two students in their entirety, covering 3 explainers and 8 theories for each student. The interannotator agreement between the raters was high (0.71 Cohen's Kappa), as there was a 1 hour training workshop where the annotators discussed each of the criteria and a few sample entries together.

Then, the experts annotated the remaining 312 explanations separately (corresponding to 13 more students). Each annotation covers between five to nine individual questions responded to with a binary indicator as per Table \ref{appendix:tab-dqs}. In accordance with \cite{wang2023chatgpt,qin2024infobench}, we use GPT-4o as a annotator. Table \ref{tab:annot-agree-gpt} compares the human expert annotation with GPT-4o annotation, finding an average agreement of 94.68\%. The only type of explanation where the agreement is less that 90\% is for the Base Contrastive explanations (82.22\%, with a high standard deviation), where we found that GPT-4o is a lot more critical than human annotators. As the bias is on the side of GPT-4o grading more harshly (especially for the fourth question regarding context, which is the most difficult question to annotate) we found it appropriate to proceed.

% Table showing GPT-human annotation agreement
\begin{table}[ht!]
\centering
  \resizebox{0.3\linewidth}{!}{%
  \begin{tabular}{@{}lc@{}}
  \toprule
  \multicolumn{1}{c}{\textbf{Theory}} & \textbf{Agreement} \\ \midrule
  {\color[HTML]{036400} Abnormal Conditions} & {\color[HTML]{036400} 98.89 ± 4.72} \\
  {\color[HTML]{9A0000} Base Contrastive} & {\color[HTML]{9A0000} 82.22 ± 9.01} \\
  {\color[HTML]{036400} Chain of Thought} & {\color[HTML]{036400} 95.11 ± 6.13} \\
  {\color[HTML]{036400} RaR + Contrastive} & {\color[HTML]{036400} 98.27 ± 3.39} \\
  {\color[HTML]{036400} Necessity Robustness} & {\color[HTML]{036400} 97.78 ± 2.62} \\
  Pearl Explanation & 95.83 ± 4.41 \\
  {\color[HTML]{036400} Relevance Selection} & {\color[HTML]{036400} 97.78 ± 3.58} \\
  Statistical Relevance & 91.56 ± 6.99 \\ \midrule
  \textbf{Overall} & \textbf{94.68 ± 5.20} \\ \bottomrule
  \end{tabular}%
  }
  \caption{\textbf{Agreement between Human and GPT-4o annotation.} Average percentage of answers with agreement between human and GPT-4o. Standard deviations were computed between the different explainers.}
  \label{tab:annot-agree-gpt}
\end{table}

\subsection{Explanation Selection Stage}

We detail the decomposed questions (DQs) used by GPT-4o and human experts to annotate the explanation responses.

\begin{longtable}{|>{\RaggedRight\arraybackslash}p{4cm}|>{\RaggedRight\arraybackslash}p{11cm}|}
\hline
\makecell{\textbf{Theory}} & \makecell{\textbf{Explanation Selection DQs}} \\ \hline
\endfirsthead
\hline
\makecell{\textbf{Theory}} & \makecell{\textbf{Explanation Selection DQs}} \\ \hline
\endhead
\hline
\endfoot

\begin{mdframed}
\textbf{Relevance Selection} 
\end{mdframed} & 
\begin{mdframed}
\begin{itemize}
    \item Is the generated text using the provided data extensively?
    \item Is the generated text analysis based largely on the explainer results provided?
    \item Is the generated text correctly using the model’s predicted outcome?
    \item Is the generated text considering the context (course structure)?
    \item Is the generated text selecting the causes that are most relevant to the explainer results?
    \item Is the generated text selecting the causes that are most relevant to the question?
    \item Is the generated text selecting the causes that are most relevant to the user?
    \item Is the generated text selecting some information that is not already shared with the student?
\end{itemize}
\end{mdframed} \\ \hline

\begin{mdframed}
\textbf{Abnormal Conditions} 
\end{mdframed} & 
\begin{mdframed}
\begin{itemize}
    \item Is the generated text using the provided data extensively?
    \item Is the generated text analysis based largely on the explainer results provided?
    \item Is the generated text correctly using the model’s predicted outcome?
    \item Is the generated text considering the context (course structure)?
    \item Is the generated text selecting the causes based on the abnormality of the causes?
    \item Is the generated text selecting the causes based on the temporality of the causes?
    \item Is the generated text selecting the causes based on the controllability of the causes?
    \item Is the selected explanation following the criteria of Abnormality, Temporality, Controllability?
\end{itemize}
\end{mdframed} \\ \hline

\begin{mdframed}
\textbf{Pearl Explanation} 
\end{mdframed} & 
\begin{mdframed}
\begin{itemize}
    \item Is the generated text using the provided data extensively?
    \item Is the generated text analysis based largely on the explainer results provided?
    \item Is the generated text correctly using the model’s predicted outcome?
    \item Is the generated text considering the context (course structure)?
    \item Is the generated text identifying variables that are determined by factors external to the model?
    \item Is the generated text identifying variables that are determined by relationships with other variables?
    \item Is the explanation selected containing only a set of factors that is minimal?
    \item Is the explanation considering variables that could potentially alter the model output?
\end{itemize}
\end{mdframed} \\ \hline

\begin{mdframed}
\textbf{Necessity Robustness Selection} 
\end{mdframed} & 
\begin{mdframed}
\begin{itemize}
    \item Is the generated text using the provided data extensively?
    \item Is the generated text analysis based largely on the explainer results provided?
    \item Is the generated text correctly using the model’s predicted outcome?
    \item Is the generated text considering the context (course structure)?
    \item Is the generated text selecting features that explain the model’s output?
    \item Is the generated text selecting a few of these sets of factors by considering the necessity criteria?
    \item Is the generated text selecting the most robust explanation?
\end{itemize}
\end{mdframed} \\ \hline

\begin{mdframed}
\textbf{RaR + Contrastive \\Explanation} 
\end{mdframed} & 
\begin{mdframed}
\begin{itemize}
    \item Is the generated text using the provided data extensively?
    \item Is the generated text analysis based largely on the explainer results provided?
    \item Is the generated text correctly using the model’s predicted outcome?
    \item Is the generated text considering the context (course structure)?
    \item Is the generated text reformulating the question in a “Why P rather than Q?” format?
    \item Is the generated text selecting the event P as the target event?
    \item Is the generated text selecting the event Q as the counterfactual event that did not occur?
    \item Is the generated text selecting the foil based on abnormality to the context, question, and user?
    \item Is the generated text selecting the foil based on relevance to the context, question, and user?
\end{itemize}
\end{mdframed} \\ \hline

\begin{mdframed}
\textbf{Base Contrastive \\Explanation} 
\end{mdframed} & 
\begin{mdframed}
\begin{itemize}
    \item Is the generated text using the provided data extensively?
    \item Is the generated text based only on the explainer results provided?
    \item Is the generated text correctly using the model’s predicted outcome?
    \item Is the generated text considering the context (course structure)?
    \item Is the generated text reformulating the question in a “Why P rather than Q?” format?
    \item Is the generated text determining the causes that explain the difference between P and not-Q?
    \item Is the generated text reasoning about which causes are a cause of P?
    \item Is the generated text reasoning about causes that aren’t present in the causal chain of Q?
\end{itemize}
\end{mdframed} \\ \hline

\begin{mdframed}
\textbf{Statistical Relevance} 
\end{mdframed} & 
\begin{mdframed}
\begin{itemize}
    \item Is the generated text using the provided data extensively?
    \item Is the generated text analysis based largely on the explainer results provided?
    \item Is the generated text correctly using the model’s predicted outcome?
    \item Is the generated text considering the context (course structure)?
    \item Is the generated text selecting the explanation based on likelihood?
\end{itemize}
\end{mdframed} \\ \hline

\begin{mdframed}
\textbf{Chain of Thought} 
\end{mdframed} & 
\begin{mdframed}
\begin{itemize}
    \item Is the generated text using the provided data extensively?
    \item Is the generated text analysis based largely on the explainer results provided?
    \item Is the generated text correctly using the model’s predicted outcome?
    \item Is the generated text considering the context (course structure)?
    \item Is the generated text selecting the explanation by step-by-step tasks?
\end{itemize}
\end{mdframed} 
\label{appendix:tab-dqs}
\end{longtable}

\subsection{Explanation Evaluation}
Across both explanation selection and presentation, the decomposed questions were generated and revised by two authors (who are well familiar with LLM prompting and explainability theories), then individually vetted by a computational social scientist. For explanation selection, the question ``Is the generated text considering the context (course structure)?" was iterated upon several times to find non-subjective wording. For explanation presentation, the question ``Is the generated text concise (readable within 5 minutes)?" was similarly iterated upon. Both of these questions are educational context dependent, and the phrasing inside the parentheses can be adapted to relate to tasks from different domains.

\begin{longtable}{|>{\RaggedRight\arraybackslash}p{4cm}|>{\RaggedRight\arraybackslash}p{11cm}|}
\hline
\makecell{\textbf{Theory}} & \makecell{\textbf{Explanation Presentation DQs}} \\ \hline
\endfirsthead
\hline
\makecell{\textbf{Theory}} & \makecell{\textbf{Explanation Presentation DQs}} \\ \hline
\endhead
\hline
\endfoot

\begin{mdframed}
\textbf{Relevance Selection}
\end{mdframed} & 
\begin{mdframed}
\begin{itemize}
    \item Is the generated text describing the student’s performance?
    \item Is the generated text considering a learning goal?
    \item Is the generated text describing the explanation findings?
    \item Is the generated text suggesting concrete actions that the student can take?
    \item Is the generated text written in the tone of a teacher talking to a student?
    \item Is the generated text concise (readable within 5 minutes)?
    \item Is the generated text describing relevant causes for the student?
    \item Is the generated text describing the information that a student knows already?
    \item Is the generated text describing new insights to the student?
\end{itemize}
\end{mdframed} \\ \hline

\begin{mdframed}
\textbf{Abnormal Conditions}
\end{mdframed} & 
\begin{mdframed}
\begin{itemize}
    \item Is the generated text describing the student’s performance?
    \item Is the generated text considering a learning goal?
    \item Is the generated text describing the explanation findings?
    \item Is the generated text suggesting concrete actions that the student can take?
    \item Is the generated text written in the tone of a teacher talking to a student?
    \item Is the generated text concise (readable within 5 minutes)?
    \item Is the generated text describing abnormal causes?
    \item Is the generated text describing recent events (temporality)?
    \item Is the generated text describing controllable factors?
\end{itemize}
\end{mdframed} \\ \hline

\begin{mdframed}
\textbf{Pearl Explanation}
\end{mdframed} & 
\begin{mdframed}
\begin{itemize}
    \item Is the generated text describing the student’s performance?
    \item Is the generated text considering a learning goal?
    \item Is the generated text describing the explanation findings?
    \item Is the generated text suggesting concrete actions that the student can take?
    \item Is the generated text written in the tone of a teacher talking to a student?
    \item Is the generated text concise (readable within 5 minutes)?
    \item Is the generated text describing factors that are external to the model?
    \item Is the generated text describing internal relationships with other variables?
    \item Is the generated text highlighting features that are actionable and focused on improvement?
\end{itemize}
\end{mdframed} \\ \hline

\begin{mdframed}
\textbf{Necessity Robustness \\Selection}
\end{mdframed} & 
\begin{mdframed}
\begin{itemize}
    \item Is the generated text describing the student’s performance?
    \item Is the generated text considering a learning goal?
    \item Is the generated text describing the explanation findings?
    \item Is the generated text suggesting concrete actions that the student can take?
    \item Is the generated text written in in the tone of a teacher talking to a student?
    \item Is the generated text concise (readable within 5 minutes)?
    \item Is the generated text describing the most necessary causes that explain the result?
    \item Is the generated text describing the robustness of the causes chosen?
\end{itemize}
\end{mdframed} \\ \hline

\begin{mdframed}
\textbf{RaR + Contrastive explanation}
\end{mdframed} & 
\begin{mdframed}
\begin{itemize}
    \item Is the generated text describing the student’s performance?
    \item Is the generated text considering a learning goal?
    \item Is the generated text describing the explanation findings?
    \item Is the generated text suggesting concrete actions that the student can take?
    \item Is the generated text written in in the tone of a teacher talking to a student?
    \item Is the generated text concise (readable within 5 minutes)?
    \item Is the generated text describing a target event that did occur (P)?
    \item Is the generated text describing a contrast event that did not occur (Q)?
    \item Is the generated text describing key differences between the target and the contrast event?
    \item Is the generated text explaining why the key differences are important for understanding the result?
\end{itemize}
\end{mdframed} \\ \hline

\begin{mdframed}
\textbf{Base Contrastive explanation}
\end{mdframed} & 
\begin{mdframed}
\begin{itemize}
    \item Is the generated text describing the student’s performance?
    \item Is the generated text considering a learning goal?
    \item Is the generated text describing the explanation findings?
    \item Is the generated text suggesting concrete actions that the student can take?
    \item Is the generated text written in the tone of a teacher talking to a student?
    \item Is the generated text concise (readable within 5 minutes)?
    \item Is the generated text describing a target event that did occur (P)?
    \item Is the generated text describing a contrast event that did not occur (Q)?
    \item Is the generated text describing key causes that explain the difference between the target and the contrast event?
    \item Is the generated text explaining why the key causes are important for understanding the result?
\end{itemize}
\end{mdframed} \\ \hline

\begin{mdframed}
\textbf{Statistical Relevance}
\end{mdframed} & 
\begin{mdframed}
\begin{itemize}
    \item Is the generated text describing the student’s performance?
    \item Is the generated text considering a learning goal?
    \item Is the generated text describing the explanation findings?
    \item Is the generated text suggesting concrete actions that the student can take?
    \item Is the generated text written in the tone of a teacher talking to a student?
    \item Is the generated text concise (readable within 5 minutes)?
    \item Is the generated text describing the likely causes that most contribute to the result?
    \item Is the generated text describing why the likely causes are important?
    \item Is the generated text concisely explaining why the likely causes lead to the result?
\end{itemize}
\end{mdframed} \\ \hline

\begin{mdframed}
\textbf{Chain of Thought}
\end{mdframed} & 
\begin{mdframed}
\begin{itemize}
    \item Is the generated text describing the student’s performance?
    \item Is the generated text considering a learning goal?
    \item Is the generated text describing the explanation findings?
    \item Is the generated text suggesting concrete actions that the student can take?
    \item Is the generated text written in in the tone of a teacher talking to a student?
    \item Is the generated text concise (readable within 5 minutes)?
    \item Is the generated text describing the explanation in a sequential (step-by-step) manner?
\end{itemize}
\end{mdframed}
\label{appendix:pres-tab-dqs}
\end{longtable}

%%%%%%%%%%%%%%%%%%%%%%%%%%%%%%%%%%%%%%%
\section{Generalization}
%%%%%%%%%%%%%%%%%%%%%%%%%%%%%%%%%%%%%%%
\label{appendix:generalization}

We conducted an additional experiment to test the generalizability of \il to a different educational context. Specifically, we employed the same BiLSTM model used in \cite{swamy2023trusting} and the LIME explainer on a flipped classroom course on Linear Algebra, where students alternated between online and in-person learning components. This new experiment introduced several key variations: a different student population (Bachelor’s students in STEM, taught in French), a novel educational setting (blended learning as opposed to fully online), and a smaller dataset (150 students compared to the original 17,000). As showcased in \cite{swamy2023trusting}, over two years, the flipped classroom course had almost twice as many videos as quizzes, with an average pass rate of 59\%. The exact prompts were used, changing only the details regarding the name of the course and the syllabus.

In these more challenging settings, \il demonstrated robust instruction-following capabilities, with only a minor decline in performance. The model's average faithfulness to the instructions dropped from 97\% to 95\% in Step 1 and from 98\% to 92\% in Step 2. These results highlight the system's capacity to generalize across diverse learning environments and populations, especially to smaller datasets that are harder to predict upon \cite{swamy2023trusting}. Our analysis provides evidence supporting \il's applicability beyond its original MOOC context, underscoring its potential for deployment in varied educational settings, and ease of extension to other domains.

%%%%%%%%%%%%%%%%%%%%%%%%%%%%%%%%%%%%%%%
\section{Student Preference Study}
%%%%%%%%%%%%%%%%%%%%%%%%%%%%%%%%%%%%%%%
\label{appendix:study}
\subsection{Background of Participants}
\label{appendix:study-background}

We recruit 114 participants using \texttt{Prolific},\footnote{\url{www.prolific.com/}}, selecting the ones who identified their current status as a student and are over 18 years old. As our target participants have expertise with being a student after high school, they would be well-suited to understand both the educational context of the study and individually evaluate what explanations are most useful for them. During the study, we ask the participants whether they have ever taken an online course (MOOC), their level of education, and their level of ease with coursework. Detailed demographics distribution can be found in Figure \ref{fig:userstudy_demo}. The sample of participants is gender-balanced, and about half of them have taken or participated in creating a MOOC. The median completion time is 41 minutes, and the average reward per hour is £9.00. 

\begin{figure}[]
    \centering
    \includegraphics[width=0.95\textwidth]{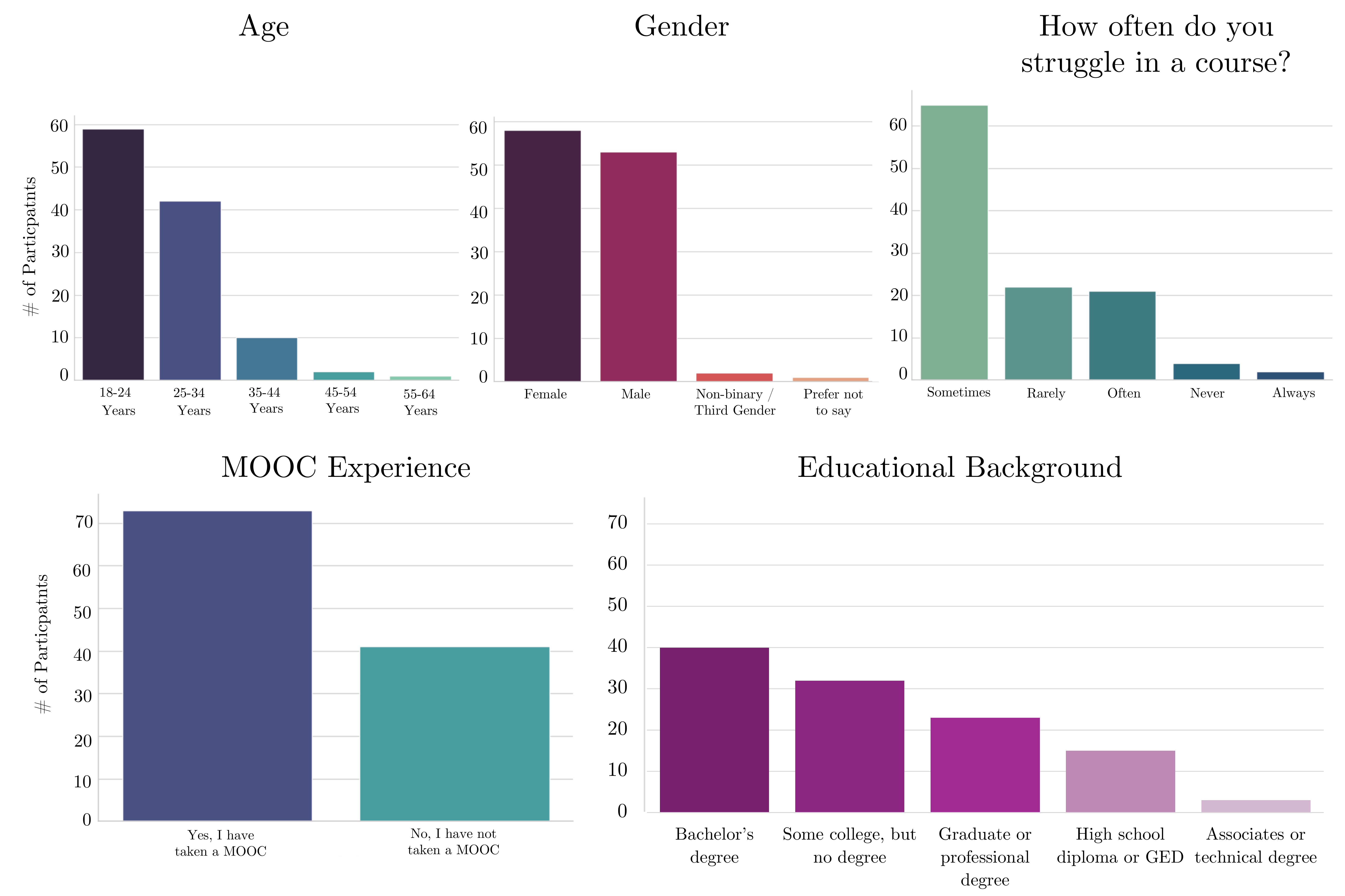}
    % \caption{Distribution of students' ages.}
    \caption{Demographics of students (age, gender, self-identified struggling in courses, MOOC Experience, and educational background) that participated in the user study.}
    \label{fig:userstudy_demo}
\end{figure}

\subsection{Study Materials}
\label{appendix:study-materials}

In this section, we provide a detailed overview of the user study introduced in Section \ref{sec:student-preferences}, covering its design, content, and further analyses of the results. The study was developed over two pilot rounds, involving 6 participants from diverse backgrounds, with continuous and iterative revisions made based on their feedback. The survey received approval from the Human Research Ethics Committee (HREC) under application number HREC 065-2022/27.09.2022.

At the beginning of the study, the participants are shown the following welcome message and disclaimer:\\

\begin{mdframed}[
    backgroundcolor=green!10,
    linecolor=black,
    linewidth=1pt,
    innertopmargin=10pt,
    innerbottommargin=10pt,
    innerrightmargin=10pt,
    innerleftmargin=10pt,
    roundcorner=10pt,
    frametitlebackgroundcolor=gray!30,
    frametitlealignment=\raggedright 
]
Dear participant,\\

Thank you for participating in our study on model explanations. We are very grateful for your participation and your invaluable insight. Please read this Explanatory Statement in full before moving forward. If you would like further information regarding any aspect of this project, please contact us via the email address provided below.\\

We are a group of researchers from the ML4ED Laboratory at EPFL dedicated to improving education through technology. \textbf{The goal of this study is to evaluate different explanations to help a student improve their performance in an online course.}\\

- This survey has been approved by the EPFL Human Research Ethics Committee (HREC) under application number HREC 065-2022/27.09.2022. HREC reviews research proposals involving human participants to ensure that they are ethically acceptable.\\

 - All the personal information will be kept confidential and anonymized. Only demographic information is being recorded and will only be reported as aggregate in a way that prevents identification of any individual participant. You can freely withdraw at any time and any collected data you provided so far will be destroyed.\\

 - All data will be collected and stored safely and reported in an anonymous form, in accordance with the Swiss Federal law on data protection (“Loi fédérale sur la protection des données” – RS 235.1).\\

- Only anonymized or aggregated data may be used in follow-up research (subject to ethics approval), and made available to other researchers for further analysis and for verification of the conclusions reached by the research team.\\

- Only the principal investigator and the aforementioned researchers have access to the original data under strict confidentiality. Results from the project may be published in conference papers and/or journal articles. In any case, no personal data will be published (only aggregated, anonymous and/or anonymized data will be published).\\

- Personal data of participants will be stored for 5 years from the date of collection. During this time, participants have the right to access their data and request information about the processing of their personal data. In order to exercise this right, you need to contact the Principal Investigator.\\

By participating in this survey, you agree that your data can be used for scientific purposes.\\

\textbf{In the following study, you will be asked to compare explanations for approximately 20 minutes.} Please ensure that you have enough time to finish the study correctly. Unfinished or only partially answered studies will not be considered as taken part.\\

We ask you to approach the questions and exercises with seriousness and to complete them to the best of your ability. We will subsequently check questionnaires for seriousness and will have to discard questionnaires that do not meet this requirement.\\

Thank you for your help. If you encounter any problem with the survey, or if you want to give extra feedback, or receive additional information, feel free to contact us (vinitra.swamy@epfl.ch).\\
\end{mdframed}

\subsection{Content of the study}

First, we explain the setting of the study to the participants with the following introductory message:\\

\begin{mdframed}[backgroundcolor=orange!10, linecolor=black, linewidth=1pt, innertopmargin=10pt, innerbottommargin=10pt, innerrightmargin=10pt, innerleftmargin=10pt, roundcorner=10pt]
You are a student taking three online courses (MOOCs): Digital Signal Processing, African Cities, and Elements of Geometry. Since the courses are difficult, often with low passing rates, the teaching team wants to help students who are not doing well to perform better in the course by giving them personalized assistance, and encourage students who are already performing well to continue.\\

To do this, we have a very good model (over 90\% accurate) to predict students' success or failure using various weekly behavior features (such as number of video clicks or how accurately questions are answered on the weekly quizzes). We predict student performance early in the course (before the half-way point) as passing or failing behavior. We use the explanation of the prediction to give students additional, personalized feedback to help pass the course. \\

We want to compare these personalized feedback explanations according to several criteria:\\

\begin{itemize}
    \item Usefulness: This explanation is useful to understand the prediction based on my learning behavior.
    \item Trustworthiness: This explanation lets me judge if I should trust the suggestions.
    \item Actionability: This explanation helps me make a decision on how to improve my learning behavior.
    \item Completeness: This explanation has sufficient detail to understand why the prediction was made based on my learning behavior.
    \item Conciseness: Every detail of this explanation is necessary.
\end{itemize}
\end{mdframed}

We randomly sample 6 students' explanation responses, 2 from each course, and one passing and failing student for each course. For each student, we predict their success or failure with each model and generate an explanation with LIME, MC-LIME, and CEM, then feed it through the \il pipeline for explanation selection, presentation and visualization. We provide them to the participants. The ground truth (student's performance) and the models' exact performance are not provided to the participants so that we do not bias their assessment. 

\begin{figure}[ht!]
    \centering
    \begin{minipage}{0.49\linewidth}
        \centering
        \includegraphics[width=\linewidth]{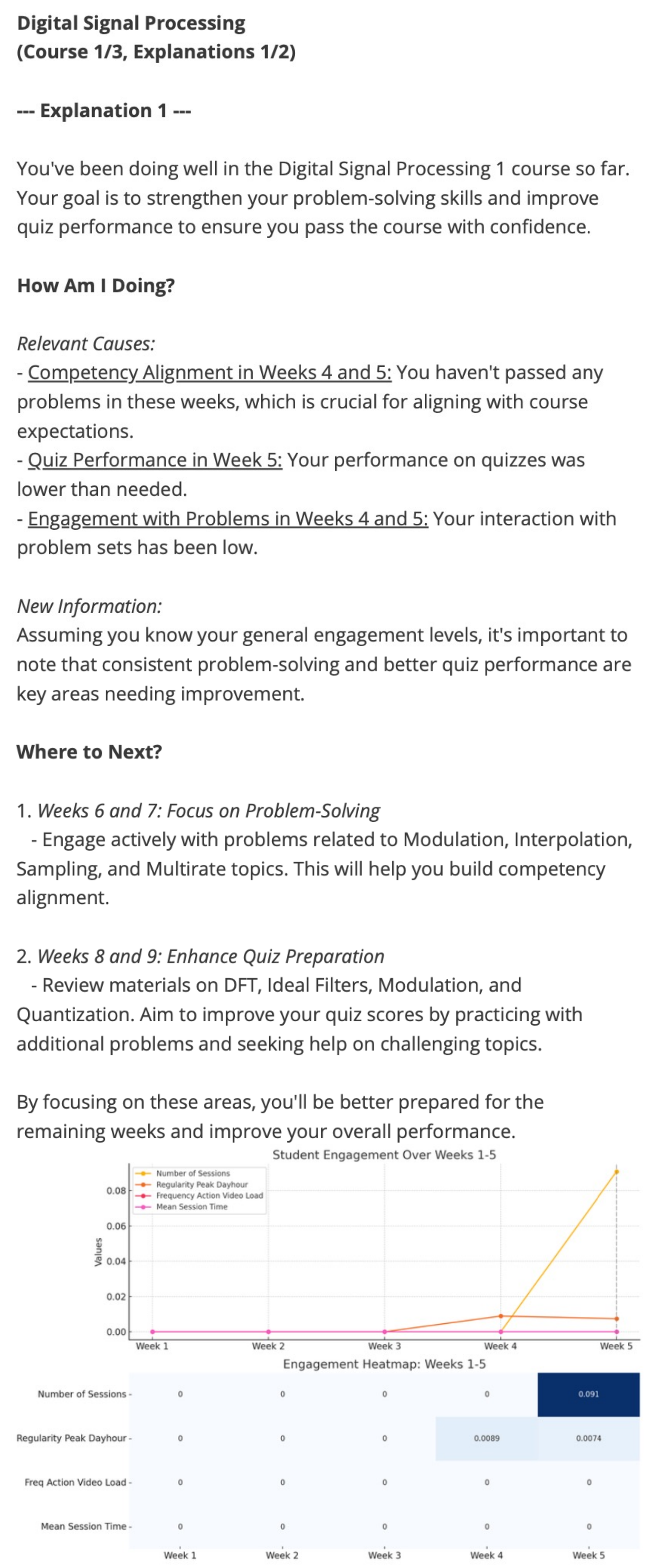}
        \caption{Example CEM \il explanation for Relevance Selection.}
        \label{fig:relevance}
    \end{minipage}
    \hfill
    \begin{minipage}{0.49\linewidth}
        \centering
        \includegraphics[width=\linewidth]{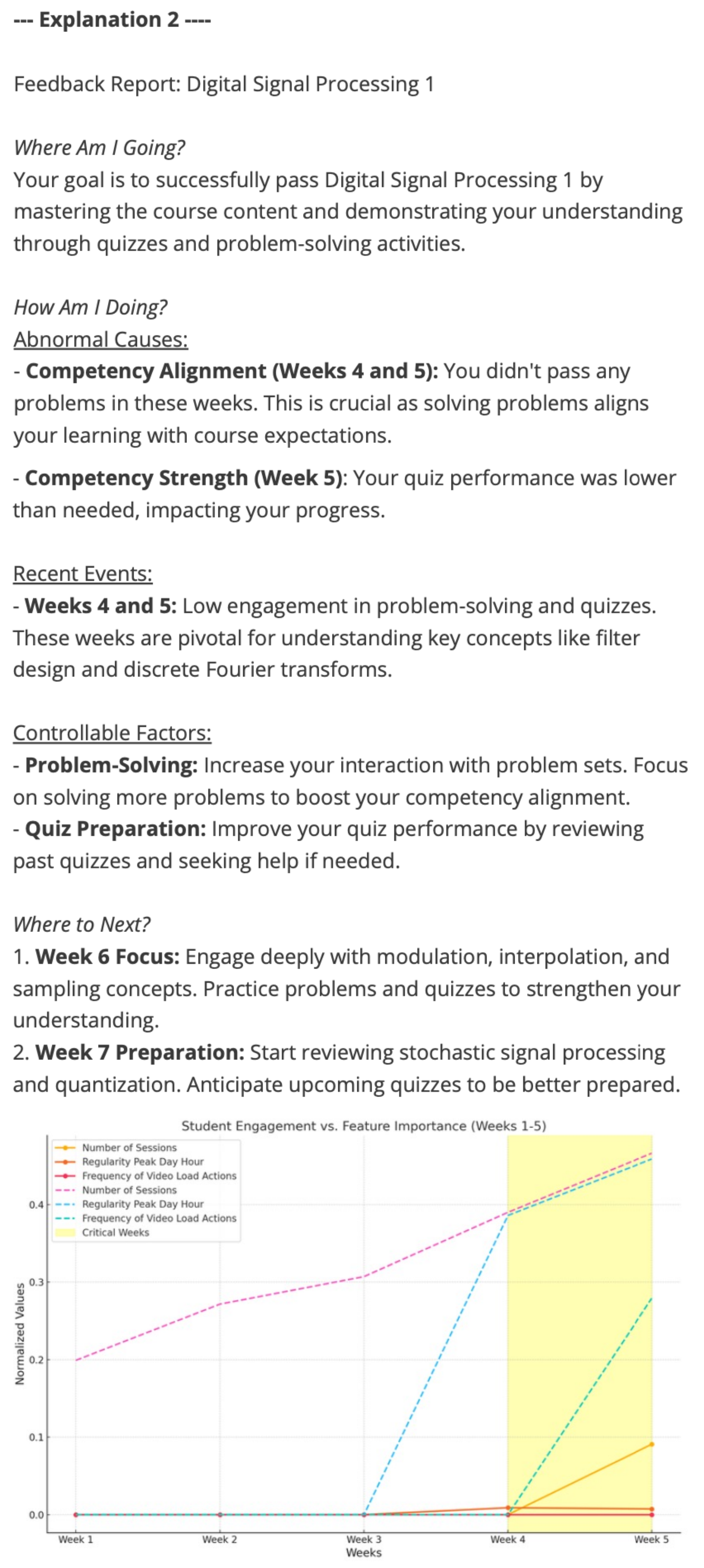}
        \caption{Example CEM \il explanation for Abnormal Conditions.}
        \label{fig:abnormal}
    \end{minipage}
\end{figure}

\begin{figure}[ht!]
    \centering
    \begin{minipage}{0.49\linewidth}
        \centering
        \includegraphics[width=\linewidth]{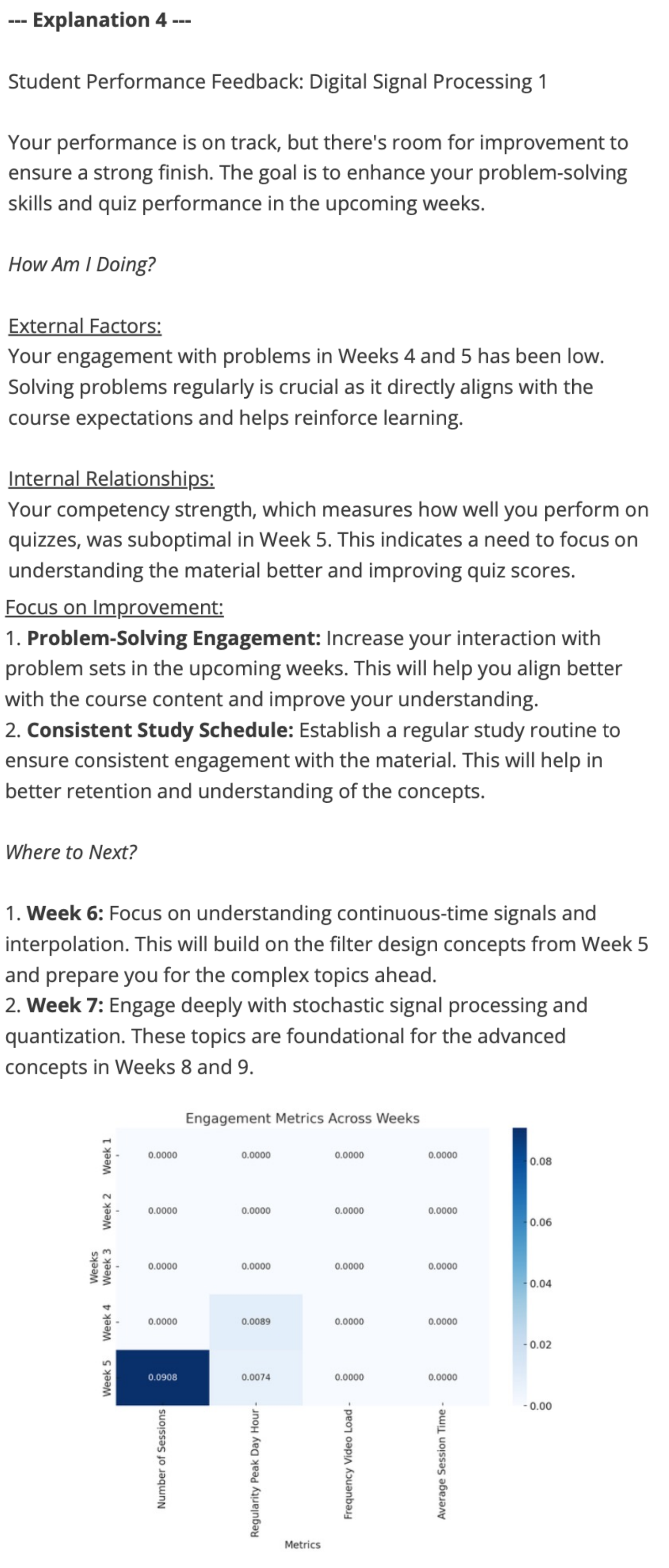}
        \caption{Example CEM \il explanation for Pearl Explanation.}
        \label{fig:pearl}
    \end{minipage}
    \hfill
    \begin{minipage}{0.49\linewidth}
        \centering
        \includegraphics[width=\linewidth]{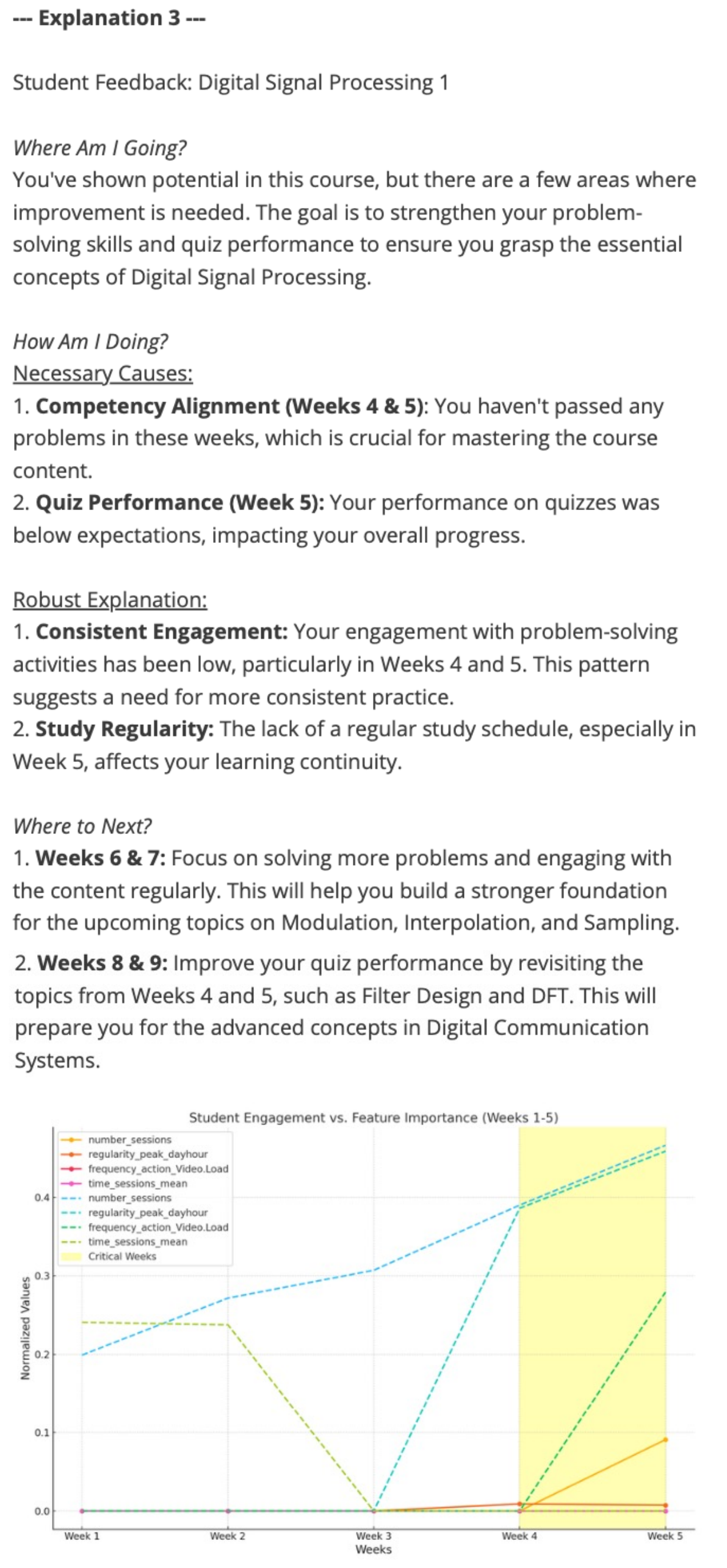}
        \caption{Example CEM \il explanation for Necessity Robustness.}
        \label{fig:nr}
    \end{minipage}
\end{figure}

\begin{figure}[ht!]
    \centering
    \begin{minipage}{0.49\linewidth}
        \centering
        \includegraphics[width=\linewidth]{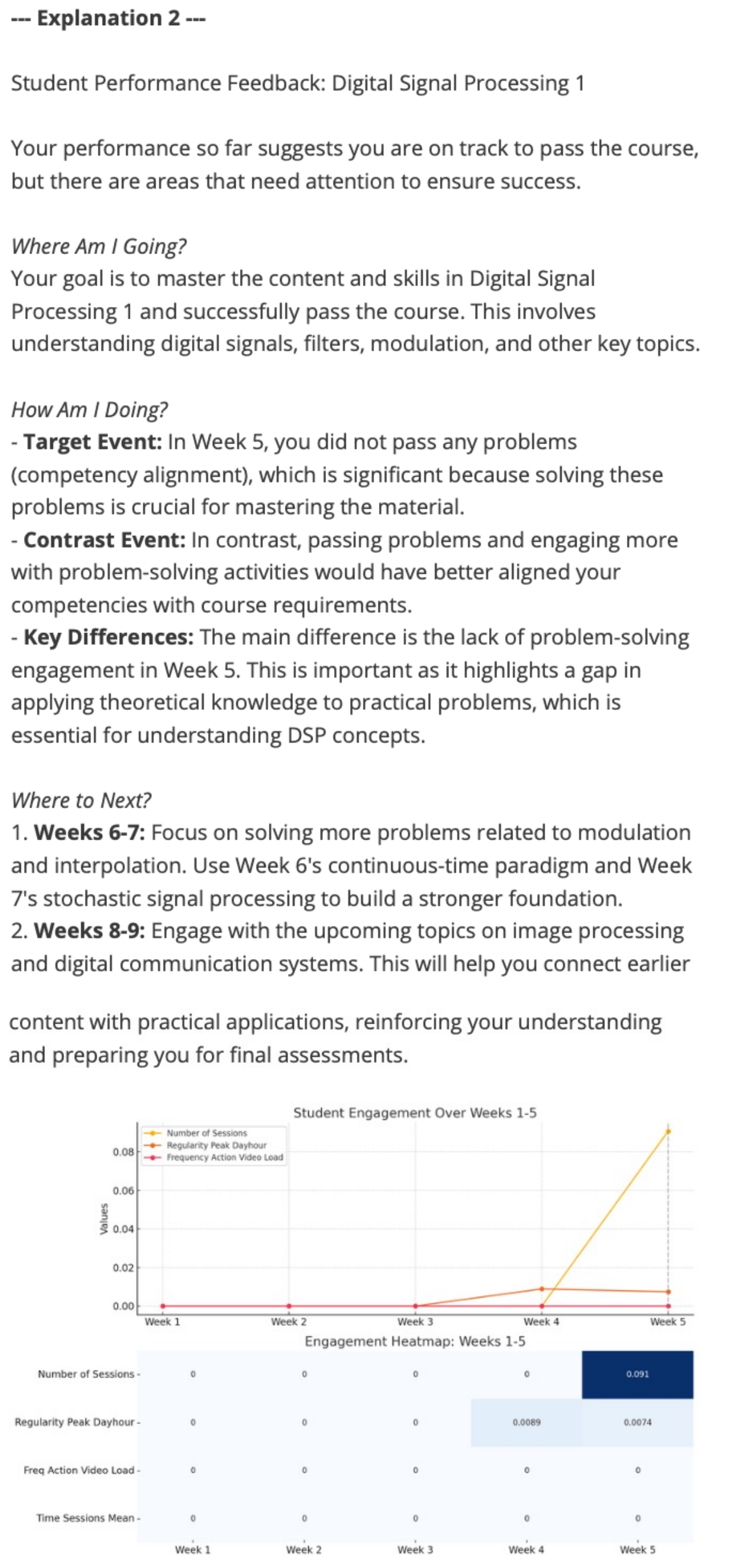}
        \caption{Example CEM \il explanation for Contrastive Explanation.}
        \label{fig:contrastive}
    \end{minipage}
    \hfill
    \begin{minipage}{0.49\linewidth}
        \centering
        \includegraphics[width=\linewidth]{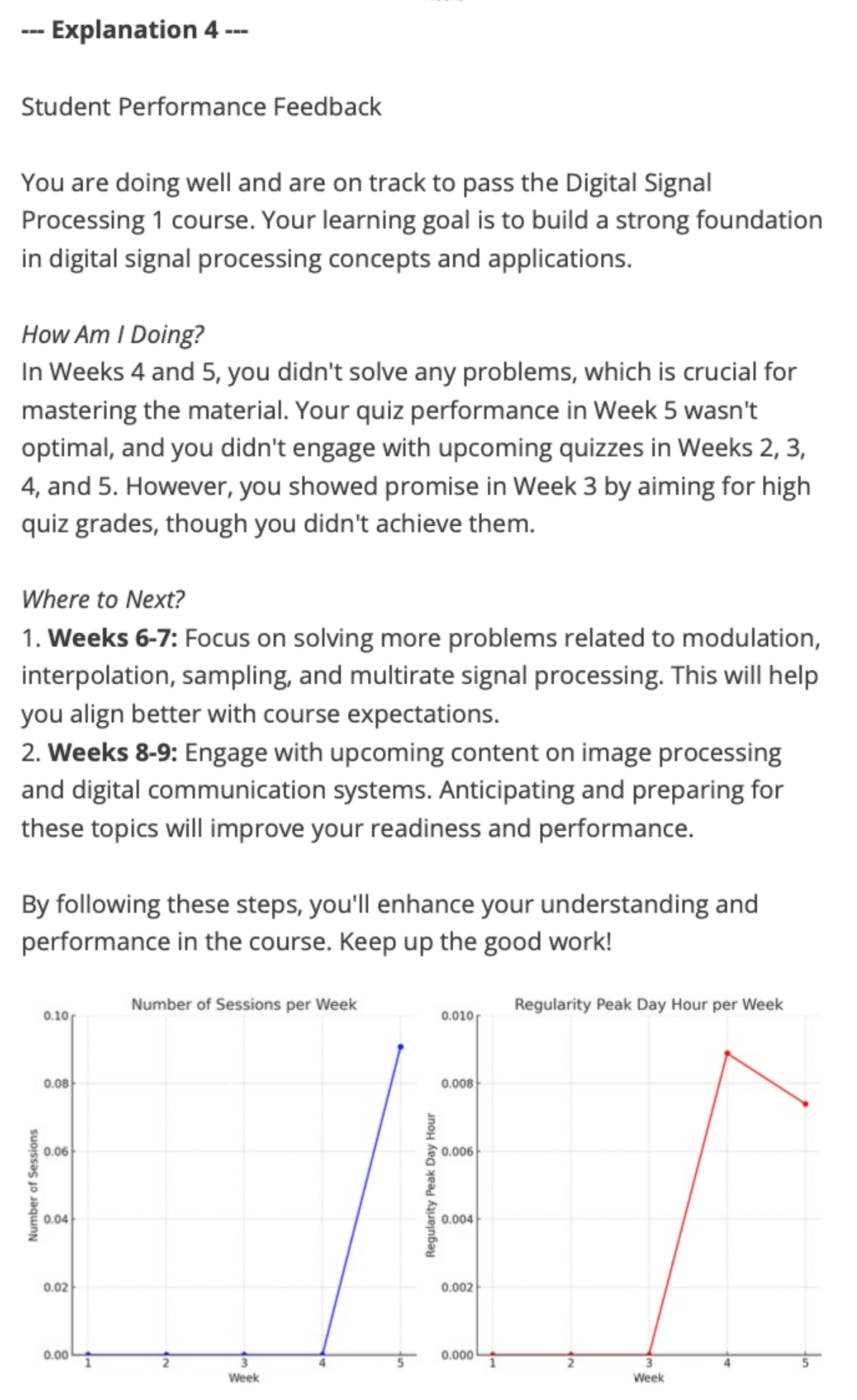}
        \caption{Example chain of thought (CoT) \il explanation.}
        \label{fig:cot}
    \end{minipage}
\end{figure}

\begin{figure}[ht!]
    \centering
    \begin{minipage}{0.3\linewidth}
        \centering
        \includegraphics[width=\linewidth]{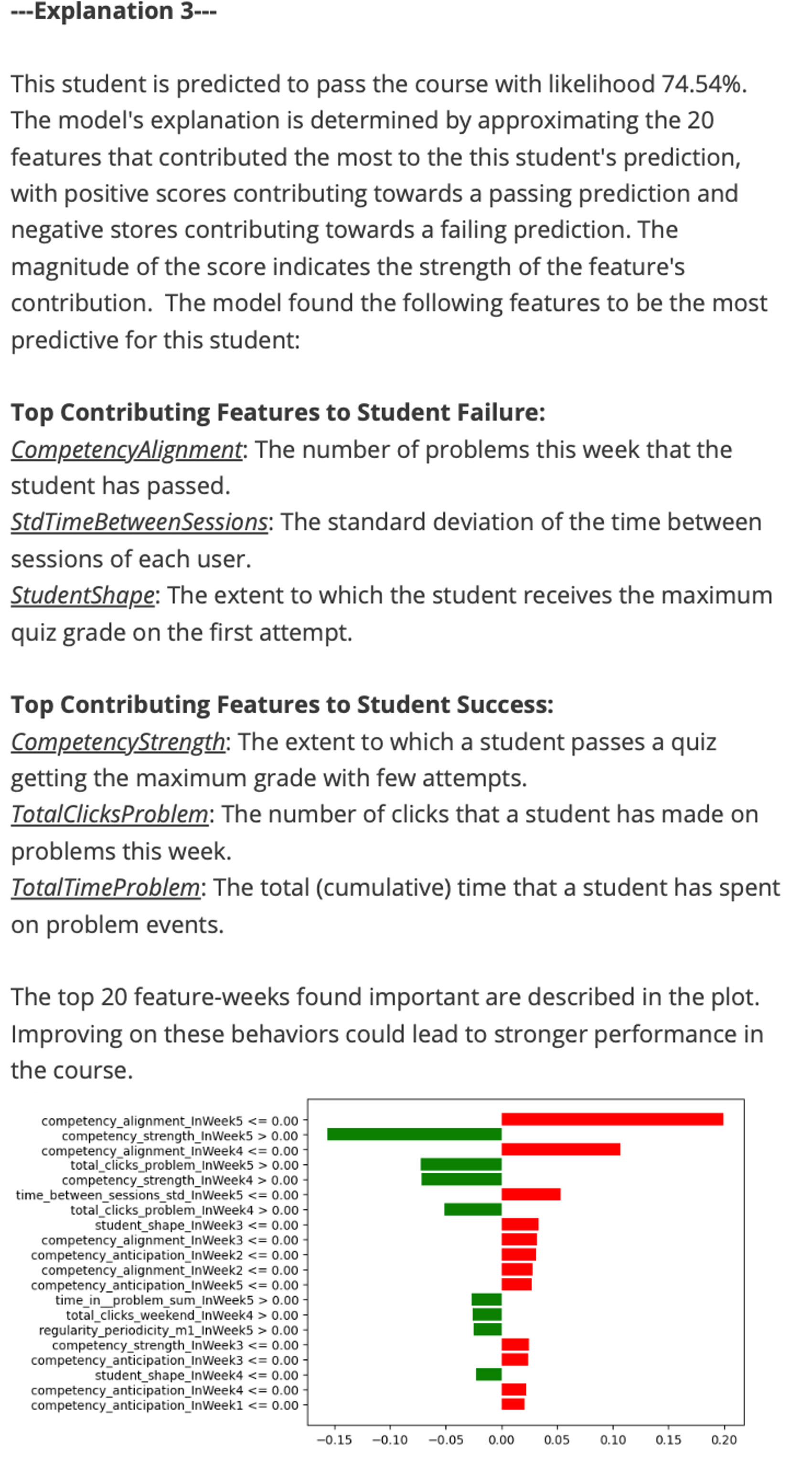}
        \caption{Example LIME (BASE) explanation.}
        \label{fig:LIME}
    \end{minipage}
        \hfill
    \begin{minipage}{0.3\linewidth}
        \centering
        \includegraphics[width=\linewidth]{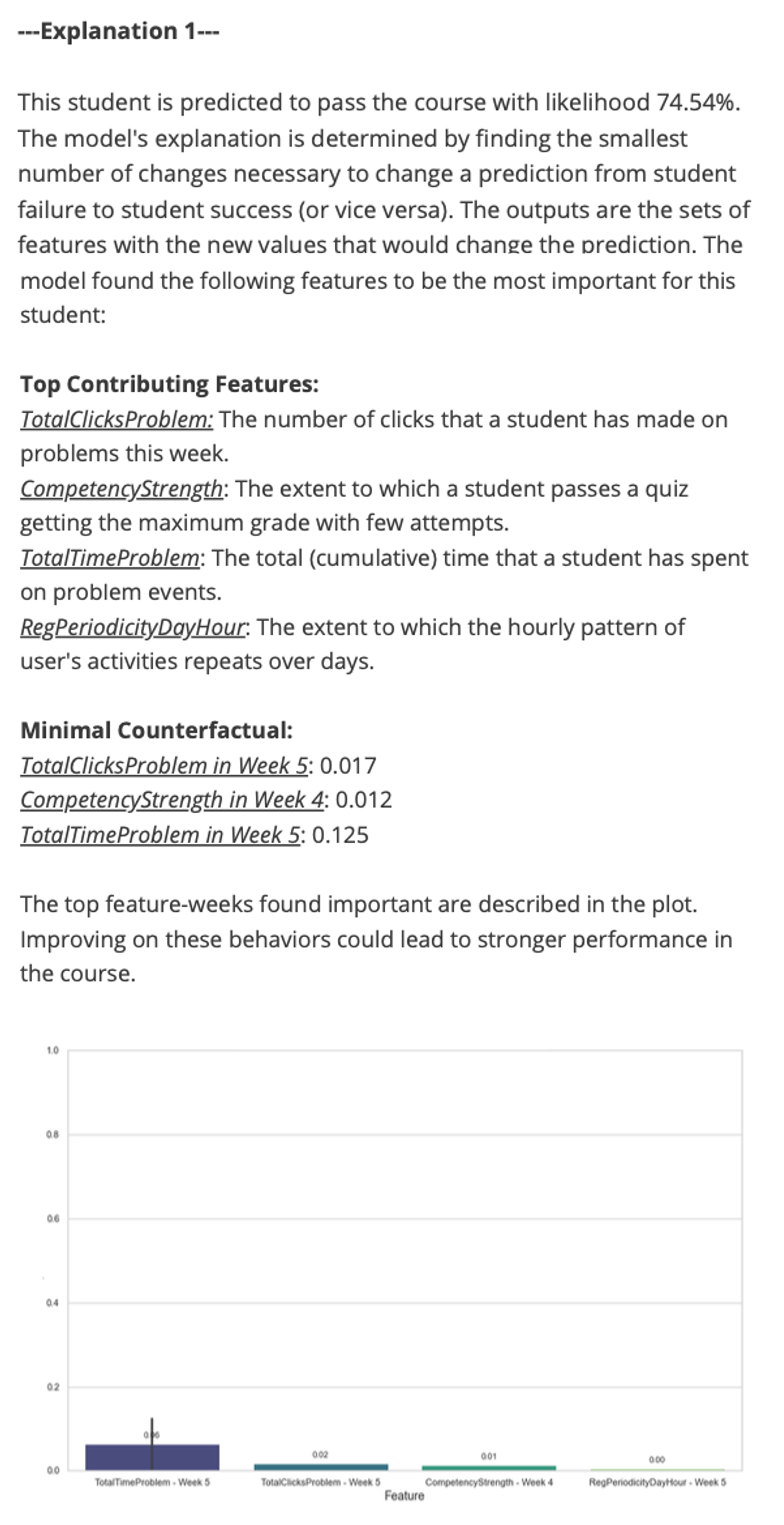}
        \caption{Example MC-LIME (BASE) explanation.}
        \label{fig:MCLIME}
    \end{minipage}
        \hfill
        \begin{minipage}{0.3\linewidth}
        \centering
        \includegraphics[width=\linewidth]{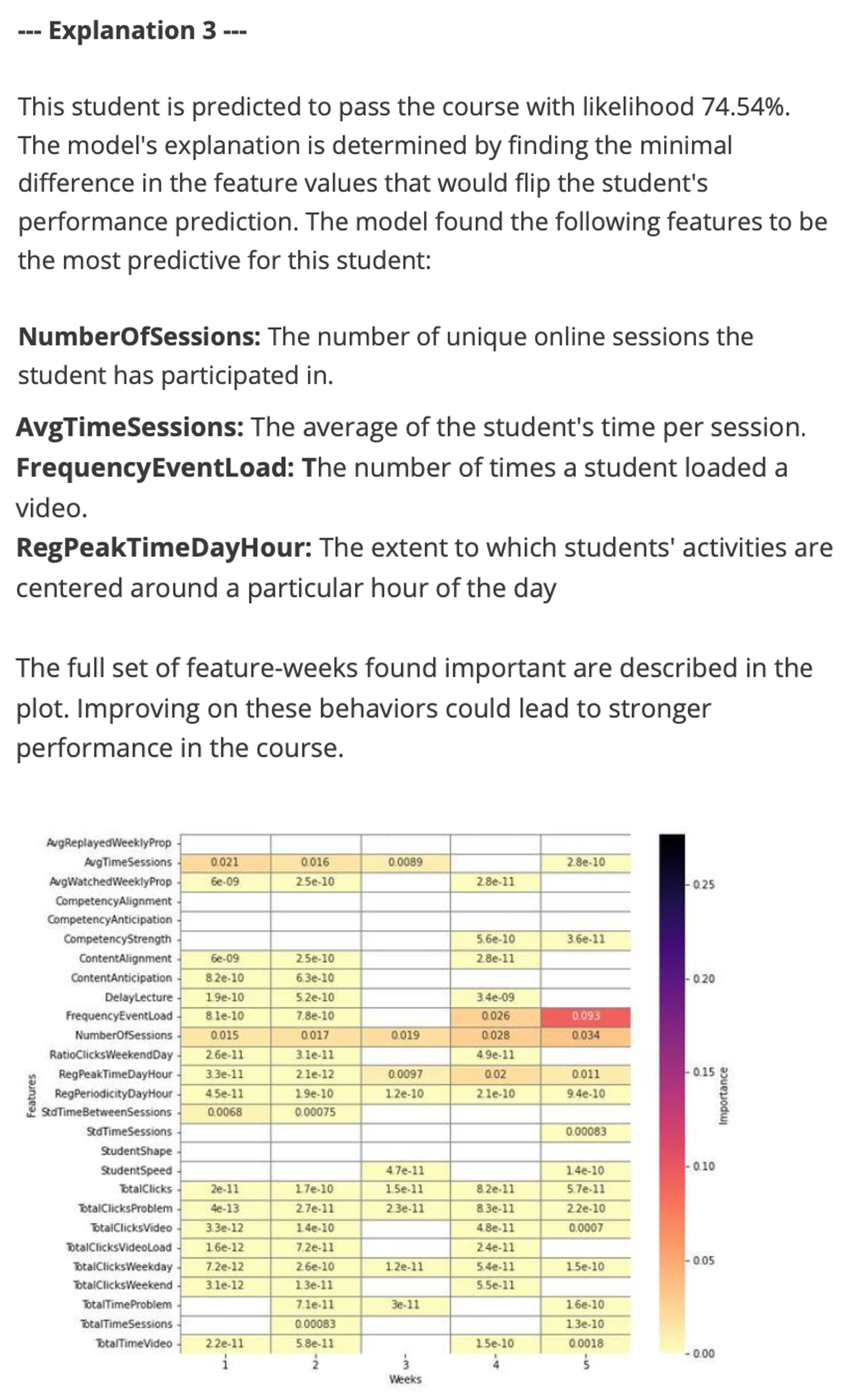}
        \caption{Example CEM (BASE) explanation.}
        \label{fig:CEM}
    \end{minipage}
\end{figure}

The content of the explanation obtained by each method differs greatly. We simplify the explanations and render them in textual and graph format to make them as easy to understand as possible to a human.  For baseline explanations (LIME, CEM, MC-LIME), we craft a text template in line with \cite{swamy2024interpretcc} and iterate on the wording with 2 learning scientists. We provide the description and list of features found important by the model and suggest that improving on these attributes can improve the student performance. For LIME, we show the default graphic that comes with the explanation (Fig. \ref{fig:LIME}). For CEM, we visualize all features found important across the 5 weeks of student progress in a heatmap (Fig. \ref{fig:CEM}), and describe the five most important features in the textual explanation. For MC-LIME, we present only the most concise counterfactual set found important based on the LIME scores and show how much those features need to change to impact the prediction (Fig. \ref{fig:MCLIME}). The remaining figures for all \il explanations are generated directly by GPT-4o, often choosing a line graph, a heatmap, or a bar plot that is directly related to the importance scores or feature values offered in the explanation. Several examples of \il plots are included in Fig. \ref{fig:relevance}, \ref{fig:abnormal}, \ref{fig:pearl}, \ref{fig:nr}, \ref{fig:contrastive}, \ref{fig:cot}.

Consider that the way we chose to present the explanations may influence participants' perceptions. Ideally, we would offer a comprehensive breakdown of how each model utilizes features and how the explanations are generated, allowing participants to fairly evaluate the explanation's quality. However, in a practical setting, the user encountering the explanation may lack the time or background knowledge to fully grasp these details. To address the trade-off between thoroughness and accessibility, we decided on a balanced approach, presenting each explanation with a simple graph and a concise explanatory text.

\begin{figure}[ht!]
    \centering
    \begin{minipage}{0.42\linewidth}
        \centering
        \includegraphics[width=\linewidth]{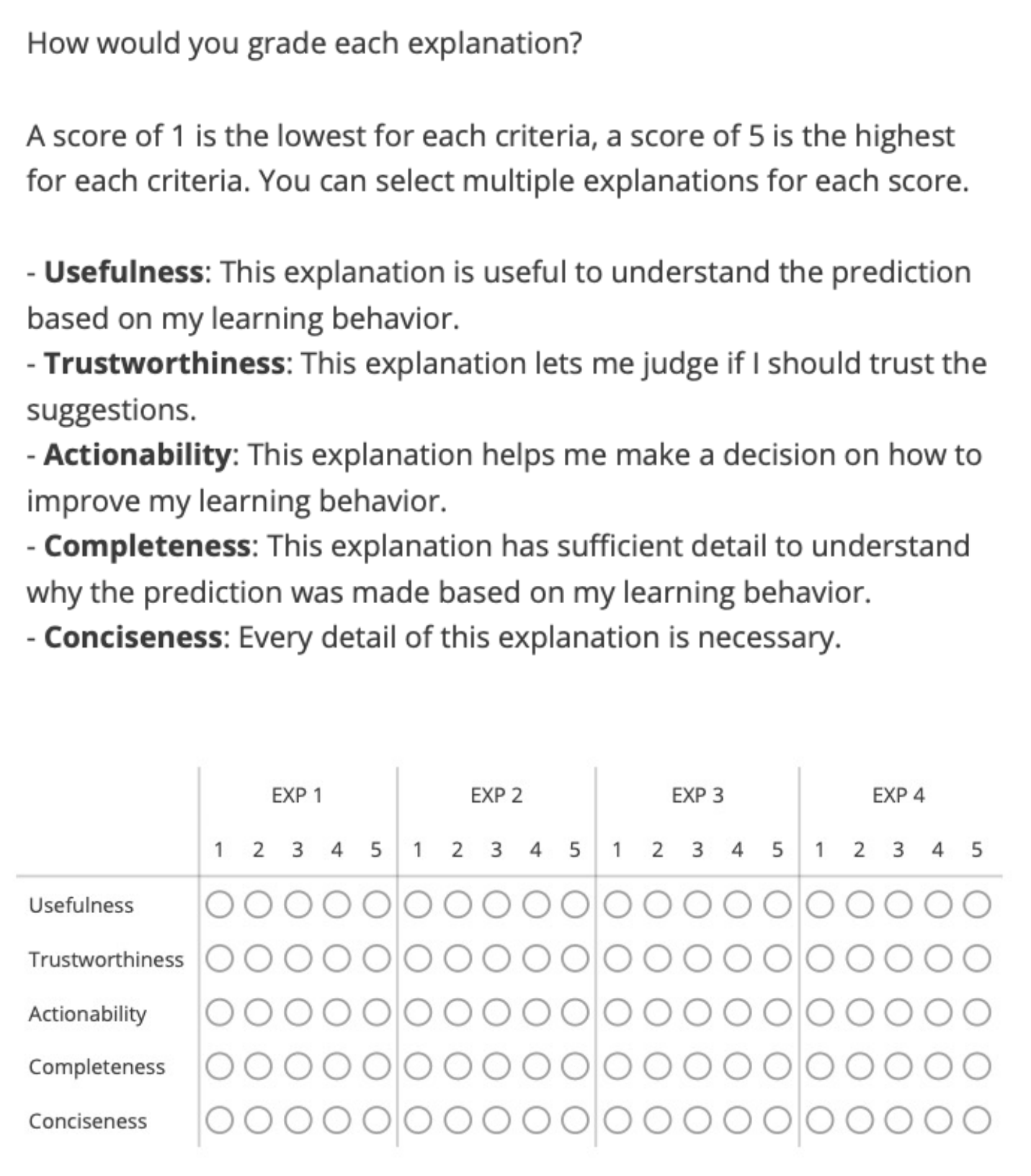}
        \caption{Grading criteria across four explanations presented at a time (one base explanation and three \il explanations).}
        \label{fig:answer-section}
    \end{minipage}
    \hfill
    \begin{minipage}{0.55\linewidth}
        \centering
        \includegraphics[width=\linewidth]{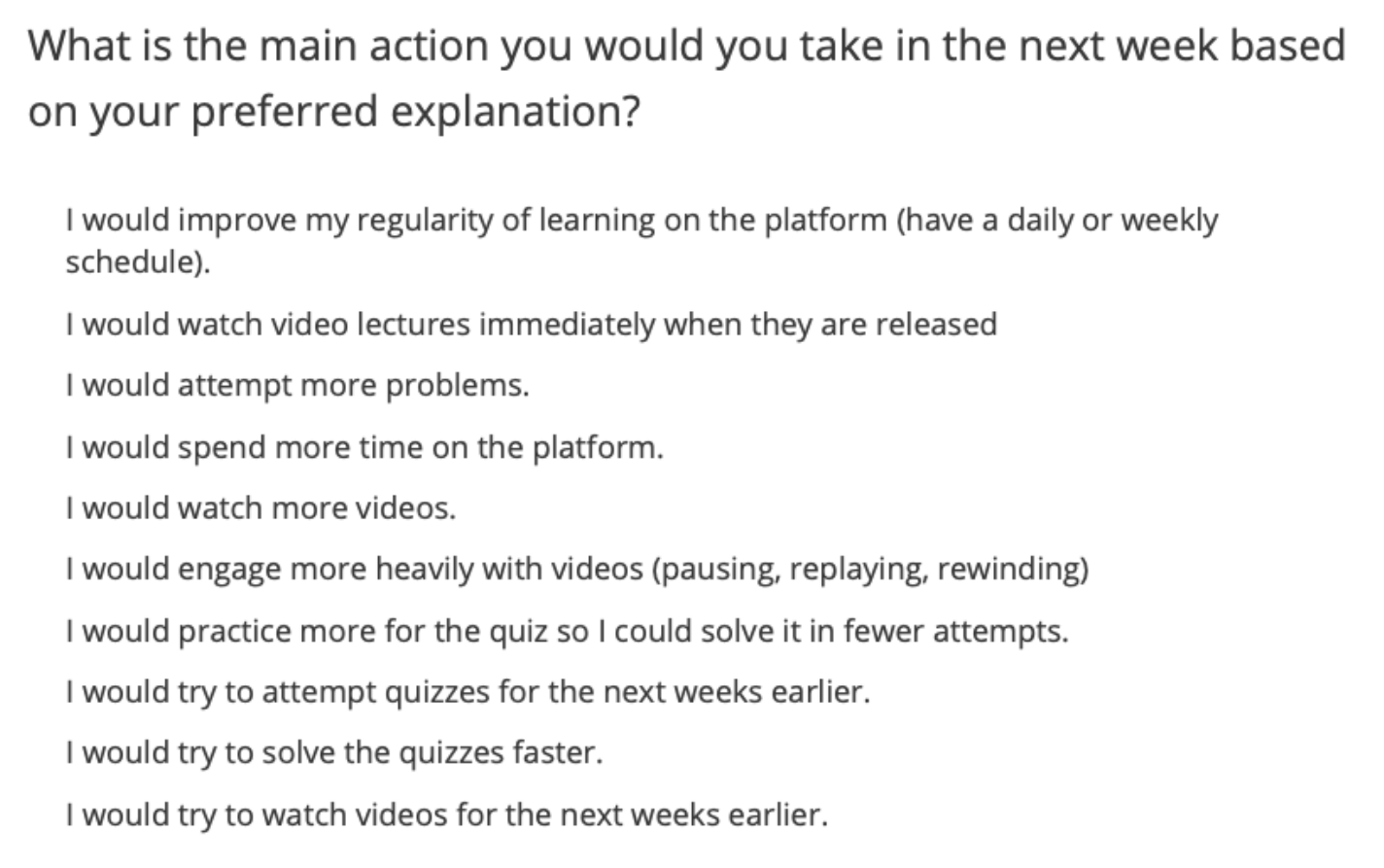}
        \caption{Actionability decision made by the students.}
        \label{fig:actionability}
    \end{minipage}
\end{figure}

For each student sample, participants were presented with four model explanations in a random order. They were then required to evaluate these explanations using Likert scales based on five predefined criteria. To familiarize participants with the evaluation process and to screen for inattentive responses, a practice question was included. Data from three participants who did not correctly answer the practice question were excluded from the final analysis. A screenshot of the answer section can be seen in Figure \ref{fig:answer-section}.

All study materials can be found in the \il repository\footnote{\texttt{http://github.com/epfl-ml4ed/iLLuMinaTE/tree/main/study-materials}}, with the exact versions of the survey shown to students. There are six versions of the survey: two variations with 3 randomly chosen students each (one from each course) and three different source explainer versions (LIME, MC-LIME, CEM) for each student variation.

\pagebreak
\pagebreak
\subsection{Extended Results}
\label{appendix:study-results}

\begin{figure}[ht!]
    \centering
    \includegraphics[width=\linewidth]{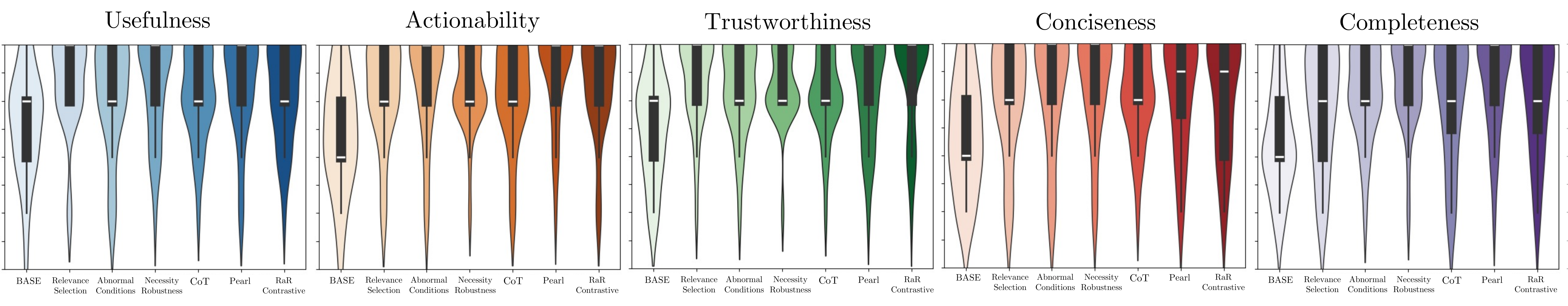}
    \caption{Results of student evaluation of different theories of explanations (Base, Relevance Selection, Abnormal Conditions, Necessity Robustness, CoT, Pearl, RaR + Contrastive) on all five axes of explanation.}
    \label{fig:violins}
\end{figure}

\begin{figure}[ht!]
    \centering
    \includegraphics[width=0.6\linewidth]{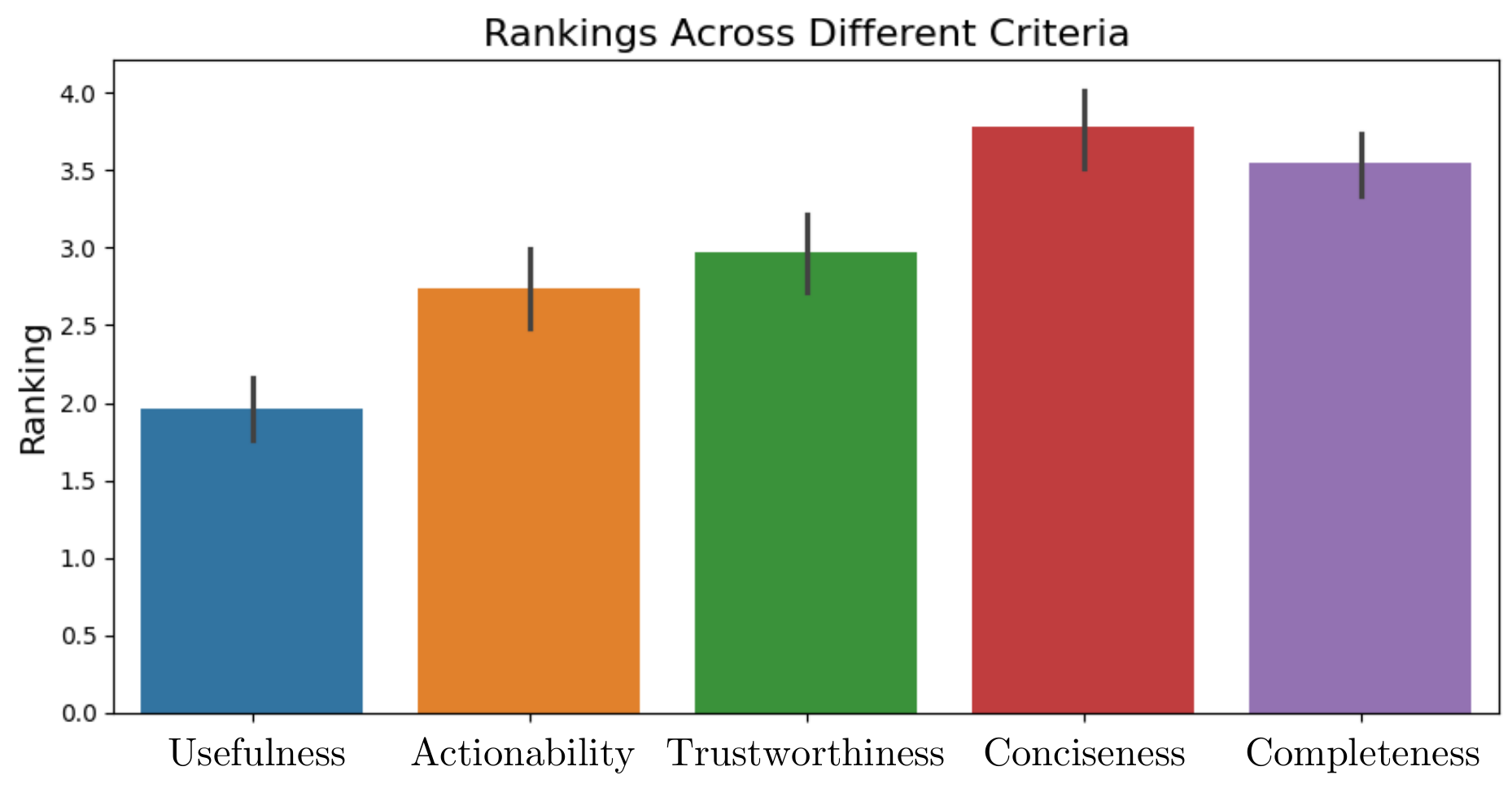}
    \caption{Results of student ranking of different explanation criteria. Lower ranking is better (more important).}
    \label{fig:ranking}
\end{figure}

The extended results of student preferences over five criteria (two of which are included in Fig. \ref{fig:exp3-criteria}) is explored in Fig. \ref{fig:violins}. Violin plots are chosen to reflect the distributions of student rankings of the criteria. Students also ranked the criteria by what they found most important (lower ranks are more important), as reflected in Fig. \ref{fig:ranking}. We note that there is a clear ordering from the student perspective over explainability criteria: usefulness, actionability, trustworthiness and then at a similar level completeness and conciseness.

%%%%%%%%%%%%%%%%%%%%%%%%%%%%%%%%%%%%%%%
\section{Actionability Experiments}
%%%%%%%%%%%%%%%%%%%%%%%%%%%%%%%%%%%%%%%
\label{appendix:actionability}

The following experiments indicate the results of the actionability simulation (with standard deviations) over explainers and theories. We note that MC-LIME is the most performant explainer, and contrastive explanations, closely followed by necessity-robustness leads to the highest quality (simulated) interventions. Note that these experiments were run over 114 study participants with six different base students that were simulated, but a larger scale study would be needed to provide more generalizable conclusions.

\begin{figure}[ht!]
    \centering
    \begin{minipage}[b]{0.41\linewidth}
        \centering
        \includegraphics[width=\linewidth]{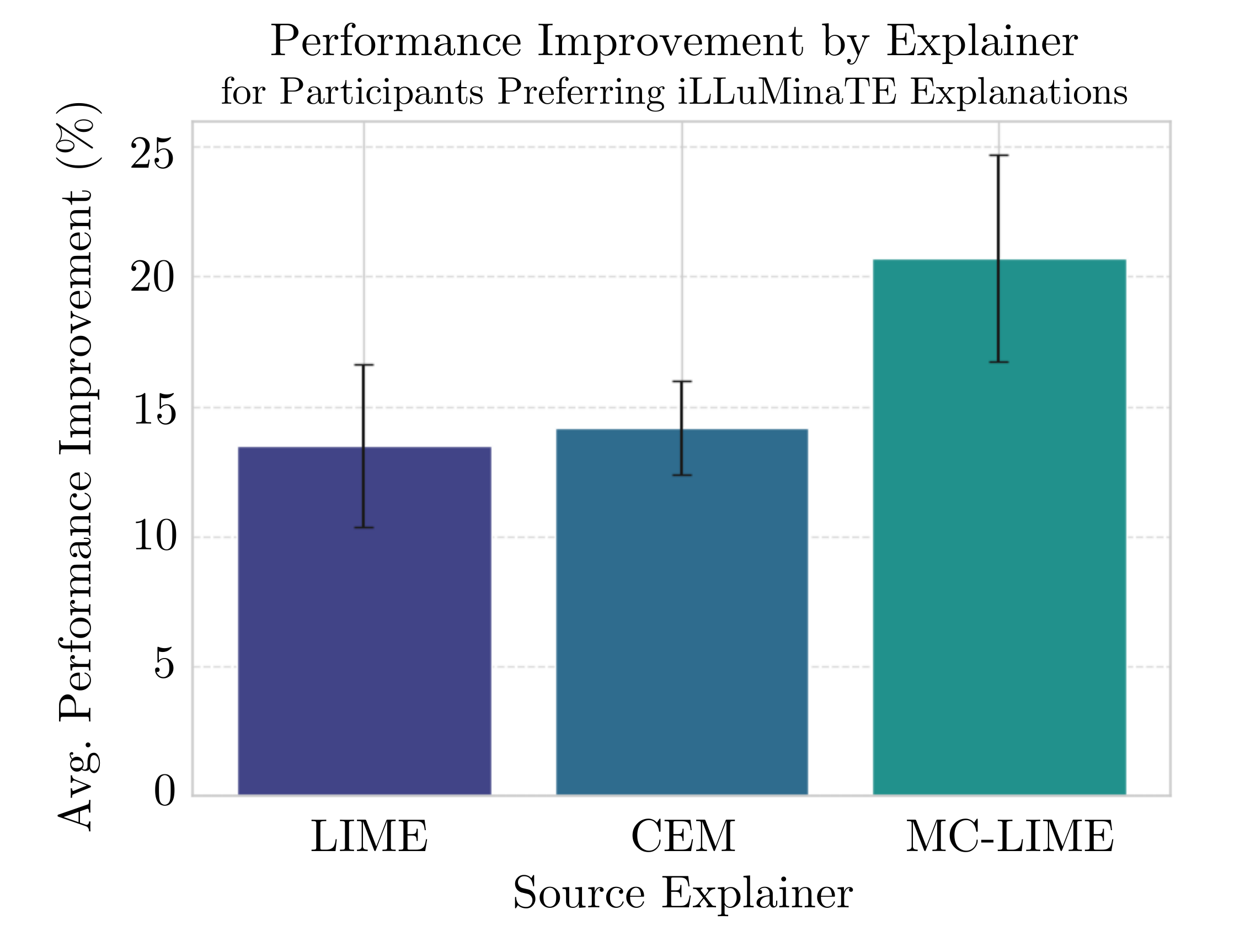}
        \label{fig:action1}
    \end{minipage}
    \hfill
    \begin{minipage}[b]{0.58\linewidth}
        \centering
        \includegraphics[width=\linewidth]{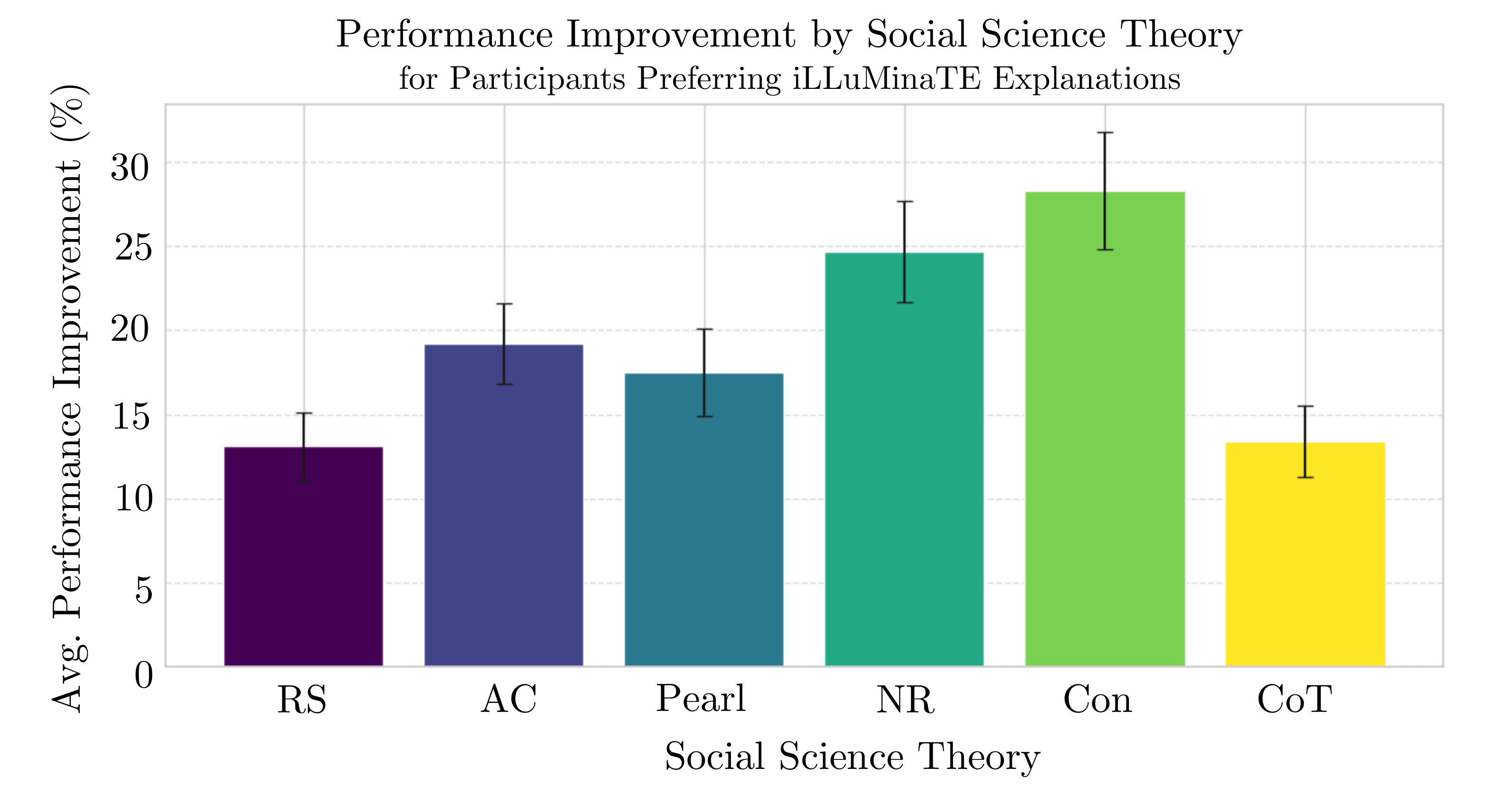}
        \label{fig:action2}
    \end{minipage}
    \caption{(Left) Actionability simulation across source explainers (LIME, CEM, MC-LIME), averaged over courses and theories. (Right) Actionability simulation across six  theories (Relevance Selection, Abnormal Conditions, Pearl's Model, Necessity and Robustness, Contrastive explanations, and Chain of Thought), averaged over courses and source explainer.}
\end{figure}

%%%%%%%%%%%%%%%%%%%%%%%%%%%%%%%%%%%%%%%
\section{Extended Background}
%%%%%%%%%%%%%%%%%%%%%%%%%%%%%%%%%%%%%%%

\subsection{What is a good explanation?}
To address this question, it is important to discuss what an explanation is, generally. Explanation can be seen through different lenses: it can be seen as causal or non-causal, as a process or a product of it. In this section, we will first look at different ideas about what makes a good explanation. Then, we will explore how humans understand and create explanations themselves. Finally, we will discuss how explanations are best communicated.

\subsection{Realist and Pragmatic Theories}

Largely, explanation theories can be divided into realist and pragmatic theories \cite{danks2022governance}. Realist theories emerge from the realms of the philosophy of science and logic, while pragmatic theories of explanation are rooted in studies of psychology and social sciences. 

The difference stands mostly on the consideration of the context as important factor in the definition of an explanation. Realist theories define an explanation a relation between theory and fact, while pragmatic theories a relation between theory, fact, and context.
This difference can be boiled down to \textit{whether radically false (i.e., not even approximately true) statements can be part of an explanation: realists say “no” while pragmatists say “yes"}. In this work, we focus mostly on pragmatic theories of explanation.

\subsection{Explanation as a process or as a product?}

Explanation can be seen as one or the other, or both \cite{lombrozo2006structure}. The theories that define explanation as a process could be categorized in three main groups \cite{miller2019explanation}:
\begin{itemize}
    \item Explanation as a causal process - Explanation is the process of inferring the real causes of an event.
    \item Explanation as a cognitive process - Explanation is a process that takes place in our cognition and it is closely related to abductive reasoning.
    \item Explanation as a social process - Explanation is a process of transferring knowledge between \textit{explainer} (the agent who explains a decision or event) and \textit{explainee} (the agent receiving the explanation).
\end{itemize}

Theories that see explanation as a product define it as an answer to a ``why-question'' \cite{miller2019explanation, molnar2020interpretable}.

\subsection{Explanation as a causal process}

For most of the philosophical theories, explanation is rooted in causality, starting from Hume's regularity theory \cite{hume2016enquiry} of causation that argues \textit{“there is a cause between two types of events if events of the first type are always followed by events of the second”}.

In broad terms, theories defined upon causality have important assumptions in common:
\begin{itemize}
    \item ``Everyday'' explanations don't differ from scientific explanations
    \item Explanation is a causal relationship
    \item Radically false statements can't be part of an explanation
\end{itemize}

Various philosophical theories emerged from causality such as the Deductive-Nomological Model (DN Model)\cite{sep-scientific-explanation} which can be summarized as \textit{``Whenever phenomenon X is observed to occur in the setting of conditions C, Y will be observed''} \cite{cambria2023survey}. An extension of these theories were the probabilistic theories, such as the Statistical Relevance model (SR model) that in general state that\textit{``event C is a cause of event E if and only if the occurrence of C increases the probability of E occurring''} \cite{miller2019explanation}.

Other models have been defined but all the theories in this category have strong limitations when applied to social or human realms. The main problem is that \textit{``the explanations that successfully follow these criteria seem to vary greatly in the extent to which they are explanatory deep or illuminating in the eye of the person receiving the explanation''} \cite{miller2019explanation}.
They miss relevance and bring little impact as in most cases there are no possible explanations that can enter in the realm of certainty and causality. Thus, from this group of theories we will keep only the SR model, as it mainly relies on likelihood and probability.

\subsection{Explanation as a cognitive process}
In general the cognitive process that we follow when we search for an explanation follows a general structure:
\textit{\begin{quote}
When we encounter something confusing, we use abductive reasoning to make sense of it and come up with an explanation. This process includes the mental simulation of hypothetical scenarios where the event didn't happen (counterfactuals) that lead us to the identification of potential causes. From these possibilities, we infer and select an explanation.
\end{quote}}

\subsubsection{Abductive Reasoning}
Abductive reasoning, also known as abduction, is a logical process in which an individual derives a hypothesis to explain an observed phenomenon \cite{peirce1997pragmatism}. It is often used to generate the best possible explanation for surprising or unexpected facts in a context of uncertainty. A common form of abductive inference follows this structure:
\begin{quote}
    \textit{The surprising fact, C, is observed. But if A were true, C would be a matter of course. Hence, there is reason to suspect that A is true.} \cite{peirce1997pragmatism}
\end{quote}

Alternatively, it can be expressed as:
\begin{quote}
    \textit{D is a collection of data (facts, observations, givens). H explains D (would, if true, explain D). No other hypothesis can explain D as well as H does. Therefore, H is probably true.} \cite{josephson1996abductive}
\end{quote}

The process could be summarized in these four steps:

\begin{enumerate}
    \item \textbf{Observation:} Observe some (presumably unexpected or surprising) events.
    \item \textbf{Hypothesis Generation:} Generate one or more hypotheses about these events.
    \item \textbf{Plausibility Judgment:} Judge the plausibility of the hypotheses.
    \item \textbf{Hypothesis Selection:} Select the ‘best’ hypothesis as the explanation.
\end{enumerate}

\subsubsection{Counterfactuals and Contrastive explanations}
Counterfactuals are very relevant to abductive reasoning as studies suggest that people don't simply explain why things happen, but rather why something happened one way (event P) instead of another way (event Q) \cite{molnar2020interpretable}. 
They were firstly described in Hume's work \cite{hume2016enquiry} as the \textit{counterfactual case}:

\begin{quote}
    \textit{“the cause should be understood relative to an imagined, counterfactual case: event C is said to have caused event E if, under some hypothetical counterfactual case the event C did not occur, E would not have occurred.”}
\end{quote}

In other words, explanations are always \textbf{contrastive}. We explain why P occurred by contrasting it with a situation where P didn't happen (Q), even if that situation is purely counterfactual.

\noindent \textbf{The Problem of Implicit Foils.} \cite{lipton1990contrastive} acknowledges a challenge. While explanations often highlight the target event (P), they rarely explicitly mention the foil (Q). For example, the question "Why did Elizabeth open the door?" could have countless foils: leaving it closed, opening the window, or someone else opening it. The unspecified foil, "not(Elizabeth open the door)," creates a vast pool of possibilities.

This theory \cite{van2002remote} further proposes a classification system for explanations based on the type of contrast:

\begin{itemize}
    \item Plain-Fact: Answers basic curiosity about an object's property (e.g., "Why does a magnet attract metal?"). These questions lack a foil.
    \item P-Contrast: Compares properties within an object (e.g., "Why is the sky blue rather than red?").
    \item O-Contrast: Compares properties between objects (e.g., "Why is a car red, while a bus is yellow?").
    \item T-Contrast: Compares properties of the same object across time (e.g., "Why was the water liquid yesterday but solid today?").
\end{itemize}

\cite{hesslow1988problem} argues that the explanandum, the event to be explained, should be viewed not as a simple property, but as a difference between the target object and a reference class (foils) with respect to a particular property. He emphasizes that the selection and weighting of causes depend on their \textit{explanatory relevance}. \cite{lipton1990contrastive}'s difference condition further clarifies that explaining "why P rather than Q" requires identifying a causal difference. This difference consists of a cause for P that is absent in the counterfactual scenario (not-Q).

Contrastive theory suggests that when selecting an explanation from multiple possible causes, people focus on the differences between the target event (P) and the counterfactual foil (Q).

\subsection{Explanation as a social process}
Furthermore, explanation can also be considered finally as a social process that is inherently part of a conversation between an \textit{explainer} and the \textit{explainee}.
Usually the explainer has enough knowledge or understanding of the causes of an event to be able to explain it to others.
The most relevant work in this category is Hilton's conversational model of explanation \cite{hilton1990conversational}.

\subsection{Why people ask for an explanation?}
The primary function of explanation goes beyond simply acquiring new information, it is about \textit{learning} and refining our understanding of the world.

Explanations can ultimately shape our causal beliefs and our prior knowledge. Malle argues that explanation helps us resolve contradictions or inconsistencies in our existing understanding and create a sense of shared meaning within a group of individuals \cite{malle2006mind}. Finally, explanations can serve as a tool for persuasion, influencing how others perceive the world.

Therefore, an explanation can have different goals in the setting of explainable AI. it is up to those designing the AI system to decide what's most important to explain in each situation. For example, an AI could be designed to build trust in the user, thus persuasion and other forms of manipulation could be more relevant than the actual truthfulness of the answer.

\subsection{Structure of explanation}

\subsubsection{Overton's scientific structure of explanation}

Overton’s model \cite{overton2012explanation} of scientific explanation defines five categories of properties or objects that are explained in science. These are:

\begin{itemize}
    \item Theories: These are sets of principles that form the building blocks for models.
    \item Models: An abstraction of a theory that represents the relationships between kinds and their attributes.
    \item Kinds: An abstract universal class that supports counterfactual reasoning. For example, an “arthropod” is a kind.
    \item Entities: An instantiation of a kind. For example, a specific spider is an entity.
    \item Data: Statements about activities, such as measurements and observations. For example, the fact that a particular spider has eight legs is data.
\end{itemize}

Overton argues that explanations of phenomena at one level must be relative to and refer to at least one other level, and that explanations between two such levels must refer to all intermediate levels.
\\
\subsubsection{Malle's theory}

\begin{figure}[t]
    \centering
    \includegraphics[width=0.35\textwidth]{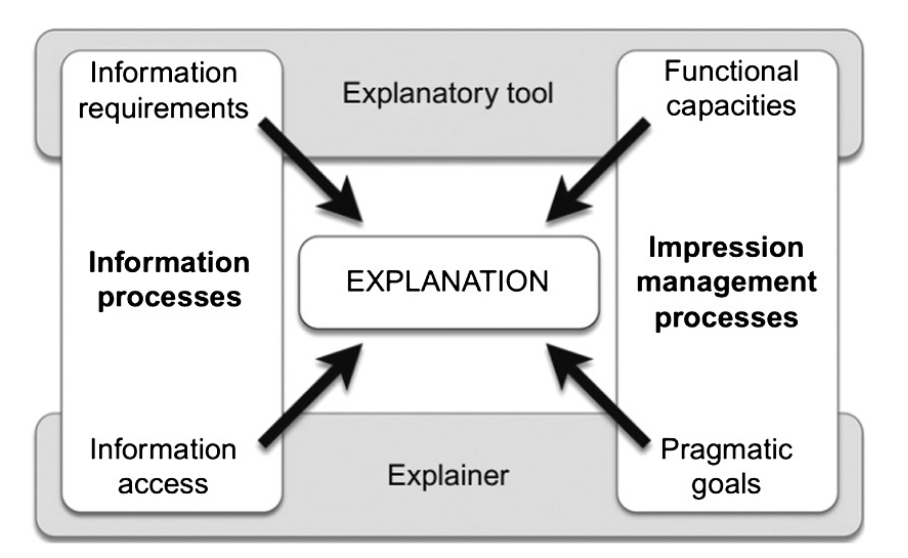}
    \caption{Malle's process model for behavior explanation from \cite{malle2006mind}}
    \label{fig:malle}
\end{figure}

\cite{malle2006mind} proposes a theory that breaks down the cognitive processes of explanation into two main groups:

\begin{enumerate}
    \item \textbf{Information Processes} — These involve the creation and assembly of explanations. 
    \item \textbf{Impression Management Processes} — These concern the social dynamics of providing explanations.
\end{enumerate}

Malle further divides these dimensions into two additional aspects (Figure \ref{fig:malle}), the tools and methods in the hands of the explainer for constructing and delivering explanations and the explainer’s perspective or prior knowledge. By considering these two dimensions, we have four items:

\begin{enumerate}
    \item Information Requirements — These are the necessary elements needed for a proper explanation. 
    \item Information Access — This refers to the information available to the explainer. Sometimes, critical information may be missing.
    \item Pragmatic Goals — This involves the purpose of the explanation, such as imparting knowledge to the listener, portraying someone in a certain light, or building trust.
    \item Functional Capacities — These are the inherent limitations and capabilities of the explanatory tools, which determine what goals can be achieved with them.
\end{enumerate}

In more simple terms these four elements could be described as \textbf{what information is needed}, \textbf{what is available}, \textbf{the intentional goals of the explainer} and the \textbf{constraints of the context and tools}. Malle's framework is fundamental to determine the Context Definition (Sec. \ref{appendix:exp-sel-prompt}) where we define these four elements for our LLMs to give an effective explanation.

\subsubsection{Hilton's conversational model of explanation}

Hilton's conversational model \cite{hilton1990conversational} emphasizes the social nature of explanation, contrasting it with simple causal attribution. He argues that explanations are conversations aimed at resolving a ``puzzle'' in the explainee's mind. This model consist of a two-stage process, which involves:

\begin{enumerate}
    \item Diagnosis: exploration of the possible causes of an event
    \item Explanation presentation: presenting the selected explanation in a socially relevant way
\end{enumerate}

Crucially, good explanations must be relevant to the explainee's specific question, not just provide any identified cause.  Hilton emphasizes that explanations follow the principles of cooperative conversation, including Grice's maxims \cite{grice1975logic} (see Appendix \ref{appendix:exp-pres-prompt}) for quality, quantity, relation, and manner. In essence, explanations should be truthful, provide the right amount of detail, stay focused on the question, and be clear and respectful. It highlights also that explanation should be tailored based on the explainee's existing knowledge, focusing on causes they may not already understand.

\subsection{\il's relationship to post-hoc explainers}

In our study we decided to use post-hoc explainers for the causal connection step. We have an instruction rule in the \il prompt framework to explicitly use the explainer's results. This decision was made due to the popularity of these explainers in the domain of AI for education \cite{hasib2022lime, 2022.EDM-short-papers.33}. However, as we seen from \cite{swamy2023trusting}, explainers in educational settings can disagree with each other. Thus, our work comparing LIME, CEM and MC-LIME begins to answer the question ``is the LLM-based communication of explanations different depending on which post hoc explainer is used?" -- we mostly identify that the choice of explainer does not lead to an overall difference in which theories students prefer (Sec. \ref{sec:student-preferences}). However, we do identify a significant increase of actionability for the MC-LIME source explainer over the other approaches (Sec. \ref{sec:actionability} and Appendix \ref{appendix:actionability}). A longer exploration with more source explainers is required to make generalizable statements (beyond the education use case) regarding how choice of source explainer impacts LLM communication of the explanation. \il is compatible with using interpretable-by-design models (i.e. Concept Bottleneck Models \cite{koh2020concept}, InterpretCC \cite{swamy2024interpretcc}) or in-hoc explainability methods (i.e. concept activation vectors for education \cite{asadi2023ripple}) as a causal connection step instead of post-hoc explainers.

%% file: tail/biblio.tex
\cleardoublepage
\phantomsection
% ************* The working stuff *************
% Print the bibliography just here

\printbibliography

\addcontentsline{toc}{chapter}{Bibliography}

%% file: tail/cv.tex
\cleardoublepage
\thispagestyle{empty}
\phantomsection
\addcontentsline{toc}{chapter}{Curriculum Vitae}

% NOTE: If you want to make your life hard and write your CV in Latex keep the following lines as they are. If you desire to upload a pre-made PDF CV please uncomment the next line and comment the line just after it.


\section*{Supplementary material}

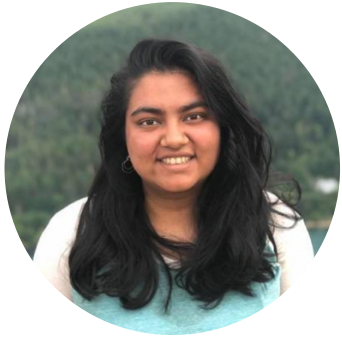

# Vinitra Swamy
Curriculum Vitae

✉ vinitra.swamy@epfl.ch
⬡ vinitra
📍 Lausanne, Switzerland
🔍 eXplainable AI, Human-Centric ML, Multimodality, Natural Language Processing (NLP)
💻 Google Scholar, Twitter, LinkedIn, Website

## EDUCATION

Current  **Ph.D. Computer Science**  Swiss Federal Institute of Technology (EPFL)
Advisor: Tanja Käser (ML4ED), Martin Jaggi (MLO)

5th year PhD on trustworthy, explainable AI for human-centric data. Recipient of the EDIC Fellowship and the G-Research PhD Prize ('24), awarded the IC Distinguished Service Award thrice ('21, '22, '23). Recognized by Stanford, UCSD, and UChicago as a "Rising Star in Data Science" ('24).

2018  **M.S. Electrical Engineering and Computer Science**  University of California, Berkeley
Advisor: David Culler (RISELab)

Research degree with emphasis in deep learning / AI, Graduate Opportunity fellow (full tuition award).

2017  **B.A. Computer Science**  University of California, Berkeley

Graduated 2 years early, awarded EECS Award for Excellence in undergraduate teaching and leadership.

2015  **High School Diploma** Cupertino High School

Relevant Coursework: Fairness in ML[+], Designing and Visualizing Deep Neural Networks[+], Special Topics in Deep Learning[+], Natural Language Processing[+], ML for Education[+], Databases[+], Artificial Intelligence, Algorithms, Advanced Data Science, Operating Systems, Linear Algebra, Linguistics

## PROFESSIONAL EXPERIENCE

2024  **UCL AI Center**  London, UK
Visiting Researcher

Hosted by the NLP lab as a collaborator of Oana-Maria Camburu on natural language explanations.

2018 - 2020  **Microsoft AI**  Bellevue, WA
AI Software Engineer

Worked on the Core Engineering team of the Open Neural Network eXchange (ONNX) Standard (founded by Microsoft, Facebook, and Amazon and now extended to 30+ companies) to enable deep learning framework interoperability. Released ONNX Runtime 1.0 for efficient model inference.

2020  **University of Washington, Paul G. Allen School of CSE**  Seattle, WA
Part-Time Faculty, Machine Learning

Lecturer for CSE/STAT 416: Introduction to Machine Learning to 100+ UW Seattle upper-division undergraduate and graduate students, with a teaching team of 4.

---

[+] Graduate Coursework

2018    **UC Berkeley Division of Data Sciences**  Berkeley, CA
Lecturer, Data Science

Lecturer for DATA 8: Foundations of Data Science to over 250 UC Berkeley summer students,
managed a course staff of 30.

2018    **UC Berkeley Real-time Intelligent Secure Explainable (RISE) Lab**  Berkeley, CA
Graduate Research Assistant

Worked on projects in AI + Systems with an application area of scaling tools for data science education
(JupyterHub architecture, OkPy autograding, deep knowledge tracing).

2017    **IBM Research**  Yorktown Heights, NY
Research Scientist Intern, Machine Learning

Developed ensemble-based ML algorithm for inactive VM identification, filed 2 patents.

2015    **LinkedIn**  Mountain View, CA
Software Engineering Intern, Search Engine Optimization (SEO) Team

Data pipelining and automated testing for LinkedIn public profile pages, UX research experiments.

2011    **Google**  Mountain View, CA
Computing and Programming Experience (CAPE) Intern

Software engineering projects at Google's Mountain View HQ as a high school intern.

## RESEARCH

ICLR     **Intrinsic User-Centric Interpretability through Global Mixture of Experts.**
2025    Vinitra Swamy, Syrielle Montariol, Julian Blackwell, Jibril Frej, Martin Jaggi, Tanja Käser.

AAAI     **iLLuMinaTE: An LLM-XAI Framework Leveraging Social Science Explanation**
2025    **Theories Towards Actionable Student Performance Feedback.** Vinitra Swamy*,
Davide Romano*, Bhargav Srinivasa Desikan, Oana-Maria Camburu, Tanja Käser.

C+E: AI  **The Critical Role of Trust in Adopting AI-Powered Educational Technology for**
2025    **Learning: An Instrument for Measuring Student Perceptions.** Tanya Nazaretsky,
Paola Mejia-Domenzain, Vinitra Swamy, Jibril Frej, Tanja Käser.

ECTEL    **AI or Human? Evaluating Student Feedback Perceptions in Higher Education.**
2024    Tanya Nazaretsky, Paola Mejia-Domenzain, Vinitra Swamy, Jibril Frej, Tanja Käser.
Best    [EPFL News]
Paper

EDM      **Student Answer Forecasting: Transformer-Driven Answer Choice Prediction**
2024    **for Language Learning.** Elena Grazia Gado, Tommaso Martorella, Luca Zunino, Paola
Mejia-Domenzain, Vinitra Swamy, Jibril Frej, Tanja Käser.

AIED     **Interpret3C: Interpretable Student Clustering Through Individualized**
2024    **Feature Selection.** Isadora Salles, Paola Mejia-Domenzain*, Vinitra Swamy*, Julian
Best LBR  Blackwell, Tanja Käser.
Paper

NeurIPS  **MultiModN- Multimodal, Multi-Task, Interpretable Modular Networks.** Vinitra
2023    Swamy*, Malika Satayeva*, Jibril Frej, Thierry Bossy, Thijs Vogels, Martin Jaggi, Tanja
Käser*, Mary-Anne Hartley*.
[EPFL News] [GGBA] [Tech Xplore]

Preprint  **MEDITRON-70B: Scaling Medical Pretraining for Large Language Models.**
(Nature)  Zeming Chen, Alejandro Hernández Cano*, Angelika Romanou*, Antoine Bonnet*, Kyle

2023 Matoba, Francesco Salvi, Matteo Pagliardini, Simin Fan, Andreas Köpf, Amirkeivan Mohtashami, Alexandre Sallinen, Alireza Sakhaeirad, Vinitra Swamy, Igor Krawczuk, Deniz Bayazit, Axel Marmet, Syrielle Montariol, Mary-Anne Hartley*, Martin Jaggi*, Antoine Bosselut*

[AI for Meta] [EPFL News] [BioAlps] [Analytics India] [MarkTechPost]

EMNLP Findings 2023 **Unraveling Downstream Gender Bias from Large Language Models: A Study on AI Educational Writing Assistance.** Thiemo Wambsganss, Xiaotian Su, Vinitra Swamy, Seyed Parsa Neshaei, Roman Rietsche, Tanja Käser.

Preprint (JAIR) **The future of human-centric eXplainable Artificial Intelligence (XAI) is not post-hoc explanations.** Vinitra Swamy, Jibril Frej, Tanja Käser.

LAK 2023 Honorable Mention **Trusting the Explainers: Teacher Validation of Explainable Artificial Intelligence for Course Design.** Vinitra Swamy, Sijia Du, Mirko Marras, Tanja Käser.

[EPFL News]

AAAI 2023 **RIPPLE: Concept-Based Interpretation for Raw Time Series Models in Education.** Mohammad Asadi, Vinitra Swamy, Jibril Frej, Julien Vignoud, Mirko Marras, Tanja Käser. *Educational Advances in AI Symposium.*

COLING 2022 **Bias at a Second Glance: A Deep Dive into Bias for German Educational Peer-Review Data Modeling.** Thiemo Wambsganss*, Vinitra Swamy*, Roman Rietsche, Tanja Käser.

EDM 2022 **Evaluating the Explainers: Black-Box Explainable Machine Learning for Student Success Prediction in MOOCs.** Vinitra Swamy, Bahar Radmehr, Natasa Krco, Mirko Marras, Tanja Käser.

[EPFL News]

L@S 2022 **Meta Transfer Learning for Early Success Prediction in MOOCs.** Vinitra Swamy, Mirko Marras, Tanja Käser.

NeurIPS 2021 **Interpreting Language Models Through Knowledge Graph Extraction.** Vinitra Swamy, Angelika Romanou, Martin Jaggi. *XAI4Debugging Workshop.*

KDD 2019 **Machine Learning for Humanitarian Data: Tag Prediction using the HXL Standard.** Vinitra Swamy, Elisa Chen, Anish Vankayalapati, Abhay Aggarwal, Chloe Liu, Vani Mandava, Simon Johnson. *Social Impact Track @ KDD 2019.*

AIED 2018 **Deep Knowledge Tracing for Free-Form Student Code Progression.** Vinitra Swamy, Allen Guo, Samuel Lau, Wilton Wu, Madeline Wu, Zachary Pardos, David Culler.

MSc Thesis **Pedagogy, Infrastructure, and Analytics for Data Science Education at Scale.** Vinitra Swamy, David Culler. *M.S. Thesis, UC Berkeley 2018.*

[Berkeley Inspire Magazine] [CDSS Berkeley] [Jupyter Infrastructure Team] [JupyterHub Journey] [Berkeley News]

Poster **CSi2: Machine Learning for Cloud Server Idleness Identification.** Vinitra Swamy, Neeraj Asthana, Sai Zeng, Alexei Karve, Aman Chanana, Ivan Dell'Era. *IBM Research, TJ Watson Research Center 2017.*

Poster **Neural Style Transfer for Non-Parallel Text.** Vinitra Swamy, Vasilis Oikonomou, David Bamman. *NLP Research Showcase, UC Berkeley School of Information*, 2017.

## SKILLS

| | |
|---|---|
| Lang. | Python (Expert), Java (Advanced), C (Advanced), C++ (Advanced), SQL, C#, R |
| Data Science | Pandas, NLTK, NumPy, PlotLy, MatPlotLib, Jupyter, SKLearn, Seaborn |
| Deep Learning | PyTorch, TensorFlow, Keras, ONNX, ONNX Runtime |
| Infra. | Docker, Kubernetes, Prometheus, Hadoop, Spark, Azure / AWS / Google Cloud |

## TALKS

2024 **University of Edinburgh NLP Group**
MEDITRON: Open Medical Foundation Models Adapted for Clinical Practice

2024 **Microsoft Research Cambridge**
The Future of Human-Centric XAI (Guest of Advait Sarkar)

2024 **Portugal Data Makers Fest**
The Future of Human-Centric XAI

2024 **University College London NLP Group**
iLLuminaTE - A Zero-Shot, Theory-Driven LLM Framework for Student Feedback

2024 **LauzHack Deep Learning Bootcamp Lecture**
Advanced topics: Explainable AI

2024 **SMART-AI Workshop for the World Health Organization (WHO)**
Using AI for human-readable software-neutral interface, interpretability and transparency

2024 **Applied Machine Learning Days (AMLD)**
MultiModN - Multimodal, Multi-Task, Interpretable Modular Networks

2024 **EDIC Open House Talk**
Explainable, Human-Centric AI (ML4ED)

2024 **BeLEARN center**
Instructor for the JDPLS program (ETH Zurich, EPFL) on Student Modeling

2023 **Red Cross LLM Day at the ICRC Headquarters**
Evaluating and Interpreting LLMs

2023 **AWS Cloud Research Day @ EPFL**
Personalized, Trustworthy Human-Centric Computing

2023 **AAAI 2023 (EAAI Symposium)**
RIPPLE: Concept-Based interpretation for Raw Time Series

2023 **AAAI 2023 (AI4EDU Workshop)**

Encore Track talks (EDM 2022, L@S 2022)

2023 **LAK 2023: Learning Analytics and Knowledge (Honorable Mention)**
Trusting the Explainers

2022 **Oxford Machine Learning Summer Series**
Evaluating Explainable AI

2022 **EDM 2022: Educational Data Mining**
Evaluating the Explainers

2022 **EDM 2022: Opening Remarks at FATED Workshop**
Fairness, Accountability and Transparency in Education

2022 **L@S 2022: ACM Learning at Scale**
Meta Transfer Learning

2022 **ICML 2022: Opening Remarks at WiML "Un-Workshop"**

2022 **WIDS 2022: Women in Data Science, Silicon Valley (SAP)**
Fair and Explainable AI

2021 **NeurIPS 2021: Spotlight at eXplainable AI Workshop**
Interpreting LLMs through Knowledge Graph Extraction

2021 **Tamil Internet Conference (INFITT)**
TamilBERT: Natural Language Modeling for Tamil

2021 **EDIC Orientation for PhDs**
How to navigate a PhD at EPFL

2021 **UC Berkeley Data Science Alumni Panel**

2020 **Tech Gals Podcast (Episode 3)**
Microsoft AI and Creating Berkeley's Division of Data Sciences

2020 **ONNX Workshop**
ONNX Model Zoo, Tutorials, and Cloud Integrations

2020 **WIDS 2020: Women in Data Science: Silicon Valley (SAP)**
Interoperable AI

2020 **Linux Foundation AI Day**
ONNX Converter Updates

2019 **SF Artificial Intelligence Meetup (Microsoft Bay Area)**
Accelerate and Optimize Machine Learning Models with ONNX

2019 **eScience Institute (University of Washington)**
ONNX: An Interoperable Standard for AI Models

2019 **Machine Learning, AI, and Data Science Conference (Microsoft MLADS)**
Model operationalization and acceleration with ONNX

2019 **Channel 9: Microsoft AI Show**
ONNX Runtime speeds up Bing Semantic Precise Image Search

| 2018 | **UC Berkeley Data Science Pedagogy and Practice Workshop** |
|---|---|
| | Autograding at scale in a Data Science classroom (OkPy) |

| 2017 | **Salesforce Dreamforce Conference 2017** |
|---|---|
| | Opening Panelist: Creating your Legacy |

| 2017 | **JupyterCon 2017** |
|---|---|
| | Data Science at UC Berkeley: 2,000 undergraduates, 50 majors, no command line |

| 2015 | **SF Business Times Conference** |
|---|---|
| | Presenter, STEM Education Leadership Summit |

**Organizing Committee**
WiML Program Chair @ ICML 2022
FATED Workshop Organizer @ EDM 2022
HEXED Workshop Organizer @ EDM 2024, EDM 2025

**Reviewer / Program Committee**
NeurIPS 2024
AIED Program Committee 2023
AIED 2021*, 2022* (Subreviewer for Tanja Käser)
EMNLP BlackBoxNLP 2021, 2022, 2023
EACL 2022, ACL 2023
EDM 2023, Journal of Educational Data Mining (JEDM) 2022
LAK 2022*, 2023* (Subreviewer for Tanja Käser)
Editor for Springer Series on Big Data Management (Educational Data Science)

**Working Groups**
Fairness Working Group @ EDM 2022
WiML Workshop Team @ NeurIPS 2021
Lead of the 2020 ONNX SIG for Models and Tutorials

## HONORS AND AWARDS

| 2024 | **Rising Star in Data Science** Stanford University, UCSD, UChicago |
|---|---|

| 2024 | **G-Research PhD Prize** EPFL, Switzerland |
|---|---|

| 2023 | **EPFL IC Distinguished Service Award** EPFL, Switzerland |
|---|---|

| 2023 | **EPFL IC Student Representative Elect** EPFL, Switzerland |
|---|---|

| 2023 | **VP of EPFL PhDs (PolyDoc)** EPFL, Switzerland |
|---|---|

| 2022 | **EPFL IC Distinguished Service Award** EPFL, Switzerland |
|---|---|

| 2021 | **President of EPFL PhDs of IC (EPIC)** EPFL, Switzerland |
|---|---|
| | Unanimously elected for 2 terms |

| 2021 | **EPFL IC Distinguished Service Award** EPFL, Switzerland |
|---|---|

| 2020 | **EDIC Fellowship** EPFL, Switzerland |
|---|---|

| 2018 | **Graduate Opportunity Fellow, EECS** University of California, Berkeley |
|---|---|

2018    **President of Computer Science Honor Society**   University of California, Berkeley
Upsilon Pi Epsilon (UPE), Nu Chapter

2017    **Kairos Entrepreneurship Fellow**   Kairos Society, Berkeley Chapter

2017    **EECS Department Award for Excellence in Teaching and
Leadership**   University of California, Berkeley

2015    **Cal Alumni Leadership Scholar**   University of California, Berkeley

2014    **Google International Trailblazer in Computer Science** Google

## TEACHING

2020 -   **Machine Learning, Discrete Math, Databases**   EPFL, Switzerland
2024    TA for Profs. Tanja Käser, Bob West, Anastasia Alaimaki

     Taught lab and exercise sessions (30-100 students per week) for Advanced Information, Computing
and Complexity (AICC), Machine Learning for Behavioral Data, Applied Data Analysis, Database
Systems courses. Guest lecturer for demonstrations on computational complexity and Halloween Quiz
exercises for 300+ students in AICC 1 in 2022, 2023.

2020    **CSE/STAT 416: Introduction to ML**   University of Washington, Seattle
Part-Time Faculty, Lecturer
Taught 100+ upper-division undergraduate and graduate students a practical introduction to machine
learning. Modules include regression, classification, clustering, retrieval, recommender systems, and
deep learning, with a focus on an intuitive understanding grounded in real-world applications.

2016 -   **DATA 8: Foundations of Data Science**   University of California, Berkeley
2018    Lecturer, Head TA for Profs. John DeNero, David Wagner, Ani Adhikari

     Lecturer to 250+ undergraduate students on fundamentals of statistical inference, computer
programming, and inferential thinking. Head Graduate Student Instructor for over 1000 students per
semester, managed course infrastructure (live deployment of JupyterHub at scale), student teaching
staff of over 100.

2014 -   **Space Cookies Robotics Outreach**   NASA Ames Research Center, Mountain View
2015    Outreach Team Curriculum Developer and Volunteer for FRC Robotics Team

     Conducted community workshops for students from elementary school through high school to
introduce concepts of robotics and Lego Mindstorms robotics.

2012 -   **Bridging the Digital Divide**   Cupertino, CA
2014    Program Founder

     Developed computer literacy fundamentals and basic computer programming course, taught in 3
senior retirement communities and 2 family housing shelters in the Bay Area.

I had the privilege of collaborating with many talented students during my years at EPFL.

**Advised three Summer@EPFL interns:** Bahar Radmehr (Evaluating Explainers), Mohammad Asadi (Concept-
Based Explanations for Raw Time Series), and Isadora Salles (Intrinsically Interpretable Clustering).

**Advised research project students:** Natasa Krco (Generalizing XAI evaluations), Skye Du (Human Validation of
XAI), Xiaotian Su (Autoencoders for MOOCs), Julian Blackwell (Inherently Interpretable MoE), Gabriele D'Angeli
(Explainer Robustness), Lan Tran (Interpretable IRL), Juliette Parchet (Hybrid Explanations), Davide Romano (LLM-
XAI Communication), Amaury George (Teaching via Explanation), Adrien Vauthey (Knowledge Graphs), Jérémy
Chaverot (Federated LLMs), Yannis Laaroussi (Adaptive Question Generation), Othmane Oudghiri (RAG).

 % To upload a PDF CV directly